\documentclass[11pt]{report} 

\usepackage[papersize={8.5 in, 11 in}, nohead, includeheadfoot, left=1.5 in, right = 1 in, vmargin= 1 in]{geometry}

\usepackage[doublespacing]{setspace} 
\usepackage{calc}

\usepackage{array} 
\usepackage{tabularx}
\usepackage{booktabs}

\usepackage{tocloft}
\renewcommand{\cftchappresnum}{CHAPTER }
\renewcommand{\cftchapaftersnum}{ :}
\newlength{\mylen}  
\renewcommand{\cfttabpresnum}{TABLE }

\newlength{\mylent}   
\usepackage{remreset}

\renewcommand{\cftfigpresnum}{FIGURE }

\newlength{\mylenf}   
\usepackage[compact]{titlesec}

\titleformat{\section}[hang]{\large}{\thesection.}{6 pt}{}
\titleformat{\subsection}[hang]{\normalsize\itshape}{\thesubsection.}{6 pt}{}

\titlespacing*{\section} {0pt}{2.5ex plus 1ex minus .2ex}{2.5ex plus .2ex}
\titlespacing*{\subsection} {0pt}{2.5ex plus 1ex minus .2ex}{2.5ex plus .2ex}
\titlespacing*{\subsubsection}{0pt}{2.5ex plus 1ex minus .2ex}{2.5ex plus .2ex}
\titlespacing*{\paragraph} {0pt}{2.5ex plus 1ex minus .2ex}{1em}

\usepackage{parskip} 
\usepackage{xurl} 
\usepackage{natbib} 
\bibpunct{\nolinebreak(}{)}{;}{a}{,}{,} 

\usepackage{amssymb}
\usepackage{amsfonts}
\usepackage{amsmath}
\usepackage{graphicx}
\usepackage{longtable}
\usepackage{dcolumn}
\usepackage{tabularx}
\usepackage{longtable}
\usepackage{rotating}
\usepackage{paralist} 
\usepackage{verbatim}
\usepackage{subfiles} 

\def\biblio{}

\usepackage{bm}
\usepackage{amsthm}
\usepackage{algorithm}
\usepackage{algcompatible}
\usepackage{multirow}
\usepackage{tablefootnote}
\usepackage{caption}
\usepackage{subcaption}
\usepackage{dsfont}

\usepackage{xcolor}
\usepackage{changepage}
\usepackage{nameref}
\usepackage[T1]{fontenc}
\usepackage{hyperref}
\hypersetup{hidelinks}
\usepackage{enumitem}
\usepackage{microtype}
\usepackage{lscape}
\newcommand{\iid}{\textit{i.i.d.}}
\renewcommand\qedsymbol{$\blacksquare$}
\DeclareRobustCommand{\&}{%
  \ifdim\fontdimen1\font>0pt
    \textsl{\symbol{`\&}}%
  \else
    \symbol{`\&}%
  \fi
}

\makeatletter
\renewenvironment{proof}[1][\proofname]{\par
  \vspace{-0.5em}%
  \pushQED{\qed}%
  \normalfont
  \topsep0pt \partopsep0pt 
  \trivlist
  \item[\hskip\labelsep
        \itshape
    #1\@addpunct{.}]\ignorespaces
}{%
  \popQED\endtrivlist\@endpefalse
  \addvspace{6pt plus 6pt} 
}
\makeatother

\newcommand{\ModuleSpace}{\mathcal{M}}
\newcommand{\Modules}{M}
\newcommand{\module}{m}
\newcommand{\modulei}[1][i]{{\module_{#1}}}
\newcommand{\numModules}{k}

\newcommand{\structure}{s}
\newcommand{\structuret}[1][t]{{\structure^{(#1)}}}
\newcommand{\structureti}[2][t]{{\structure^{(#1)}_{#2}}}
\newcommand{\numTasks}{T}
\newcommand{\Loss}{\mathcal{L}}
\newcommand{\Losst}[1][t]{\Loss^{(#1)}}
\newcommand{\inputvec}{\bm{x}}
\newcommand{\Input}{\mathcal{X}}
\newcommand{\Inputt}[1][t]{{\Input^{(#1)}}}
\newcommand{\Output}{\mathcal{Y}}
\newcommand{\Outputt}[1][t]{{\Output^{(#1)}}}
\newcommand{\Task}{\mathcal{Z}}
\newcommand{\Taskt}[1][t]{{\Task^{(#1)}}}
\newcommand{\btheta}{\bm{\theta}}
\newcommand{\thetat}[1][t]{{\btheta^{(#1)}}}
\newcommand{\ModuleParams}{\bm{\phi}}
\newcommand{\ModuleParamsi}[1][i]{{\ModuleParams_{#1}}}
\newcommand{\ModuleParamsMatrix}{\bm{\Phi}}
\newcommand{\StructureParams}{\bm{\psi}}
\newcommand{\StructureParamst}[1][t]{{\StructureParams^{(#1)}}}
\newcommand{\StructureParamsti}[2][t]{{\StructureParams^{(#1)}_{#2}}}
\newcommand{\inputTransform}{\mathcal{E}}
\newcommand{\inputTransformt}[1][t]{{\inputTransform^{(#1)}}}
\newcommand{\outputTransform}{\mathcal{D}}
\newcommand{\outputTransformt}[1][t]{{\outputTransform^{(#1)}}}
\newcommand{\F}{\mathcal{F}}
\newcommand{\f}{f}
\newcommand{\ft}[1][t]{\f^{(#1)}}
\newcommand{\Fisher}{\bm{F}}
\newcommand{\Fishert}[1][t]{{\Fisher^{(#1)}}}

\newcommand{\Reals}{\mathbb{R}}
\newcommand{\nonLinearity}{\sigma}

\newcommand{\kfacA}{{\bm{a}}}
\newcommand{\kfacB}{{\bm{b}}}
\newcommand{\kfacAt}[1][t]{{\kfacA^{(#1)}}}
\newcommand{\kfacBt}[1][t]{{\kfacB^{(#1)}}}

\newcommand{\VAN}{NFT}
\newcommand{\kEWC}{KEWC}
\newcommand{\ER}{ER}
\newcommand{\FM}{FM}

\newcommand{\Landmine}{Landmine}
\newcommand{\FacialRecognition}{FERA}
\newcommand{\LondonSchools}{Schools}
\newcommand{\MNIST}{MNIST}
\newcommand{\Fashion}{Fashion}
\newcommand{\CUB}{CUB}
\newcommand{\CIFAR}{CIFAR}
\newcommand{\Omniglot}{Omniglot}
\newcommand{\Combined}{Combined}
\newcommand{\All}{All data sets}

\newcommand{\bLandmine}{{\bf Landmine}}
\newcommand{\bFacialRecognition}{{\bf FERA}}
\newcommand{\bLondonSchools}{{\bf Schools}}
\newcommand{\bMNIST}{{\bf MNIST}}
\newcommand{\bFashion}{{\bf Fashion}}
\newcommand{\bCUB}{{\bf CUB}}
\newcommand{\bCIFAR}{{\bf CIFAR}}
\newcommand{\bOmniglot}{{\bf Omniglot}}
\newcommand{\bCombined}{{\bf Combined}}

\newcommand{\bepsilon}{\bm{\epsilon}}
\newcommand{\balpha}{\bm{\alpha}}

\newcommand{\bL}{\ModuleParamsMatrix}
\newcommand{\bl}{\bm{l}}
\newcommand{\bs}{\StructureParams}
\newcommand{\bA}{\bm{A}}
\newcommand{\bI}{\bm{I}}
\newcommand{\bb}{\bm{b}}
\newcommand{\bH}{\bm{H}}
\newcommand{\bg}{\bm{g}}
\newcommand{\bF}{\bm{F}}

\newcommand{\that}{\skew{-0.5}\hat{t}}
\newcommand{\thetathat}{\btheta^{(\that)}}

\newcommand{\alphat}{\balpha^{(t)}}
\newcommand{\alphathat}{\balpha^{(\that)}}

\newcommand{\epsilont}{\bepsilon^{(t)}}

\newcommand{\Ht}{\bH^{(t)}}

\newcommand{\Hthat}{\bH^{(\that)}}
\newcommand{\gt}{\bg^{(t)}}

\newcommand{\gthat}{\bg^{(\that)}}
\newcommand{\st}{\StructureParamst}
\newcommand{\sthat}{\StructureParamst[\that]}

\newcommand{\Lt}{\bL_{t}}
\newcommand{\LtMinusOne}{\bL_{t-1}}
\newcommand{\LthatMinusOne}{\bL_{\that-1}}

\newcommand{\Obj}{\mathcal{J}}
\newcommand{\Objt}{\Obj^{(t)}}
\newcommand{\Objthat}{\Obj^{(\that)}}

\newcommand{\trajectories}{\mathbb{T}}

\newcommand{\kron}{\otimes}

\newcommand{\state}{\bm{x}}
\newcommand{\action}{\bm{u}}
\newcommand{\States}{\mathcal{X}}
\newcommand{\Actions}{\mathcal{U}}
\newcommand{\Transitions}{P}
\newcommand{\Rewards}{R}

\newcommand{\lpgftw}{\mbox{LPG-FTW}}

\DeclareMathOperator{\sign}{sign}
\DeclareMathOperator{\tr}{Tr}
\DeclareMathOperator{\vect}{vec}

\DeclareMathOperator*{\argmax}{arg\,max}
\DeclareMathOperator*{\argmin}{arg\,min}

\newcommand{\MDP}{\mathcal{Z}}
\newcommand{\MDPt}[1][t]{\MDP^{(#1)}}
\newcommand{\nextstate}{\bm{s}^\prime}
\newcommand{\reward}{r}

\newcommand{\Inputi}[1][i]{{\Input_{#1}}}
\newcommand{\Outputi}[1][i]{{\Output_{#1}}}
\newcommand{\TaskDescriptors}{\mathcal{T}}

\newcommand{\totalModules}{k}
\newcommand{\subproblem}{F}
\newcommand{\subproblemti}[2][t]{{\subproblem^{(#1)}_{#2}}}
\newcommand{\subproblemi}[1][i]{\subproblem_{#1}}
\newcommand{\taskDescriptor}{t}
\newcommand{\Policies}{\Pi}

\newcommand{\depth}{d}

\newcommand{\compRL}{CompRL}

\newcommand{\NonstatRL}{NonstatRL}

\newcommand{\IIWA}{\texttt{IIWA}}
\newcommand{\Jaco}{\texttt{Jaco}}
\newcommand{\Panda}{\texttt{Panda}}
\newcommand{\Kinova}{\texttt{Gen3}}

\newcommand{\boxObject}{\texttt{box}}
\newcommand{\hollowBox}{\texttt{hollow\_box}}
\newcommand{\dumbbell}{\texttt{dumbbell}}
\newcommand{\plate}{\texttt{plate}}

\newcommand{\objectWall}{\texttt{object\_wall}}
\newcommand{\objectDoor}{\texttt{object\_door}}
\newcommand{\goalWall}{\texttt{goal\_wall}}
\newcommand{\noObstacle}{\texttt{no\_obstacle}}

\newcommand{\pickPlace}{\texttt{pick-{\allowbreak}and-{\allowbreak}place}}
\newcommand{\push}{\texttt{push}}
\newcommand{\shelf}{\texttt{shelf}}
\newcommand{\trashcan}{\texttt{trash\_can}}

\newcommand{\robosuite}{\texttt{robosuite}}

\newcommand{\benchmark}{CompoSuite}

\let\origdoublepage\cleardoublepage
\newcommand{\clearemptydoublepage}{%
  \clearpage
  {\pagestyle{empty}\origdoublepage}%
}
\let\cleardoublepage\clearemptydoublepage

\begin{document}

\makeatletter
\@removefromreset{table}{chapter}
\makeatother
\renewcommand{\thetable}{\arabic{table}}
\makeatletter
\@removefromreset{figure}{chapter}
\makeatother
\renewcommand{\thefigure}{\arabic{figure}}

\def\sym#1{\ifmmode^{#1}\else\(^{#1}\)\fi} 

\def\mytitle{LIFELONG MACHINE LEARNING OF \\FUNCTIONALLY COMPOSITIONAL STRUCTURES} 
\def\mytitlecopyright{LIFELONG MACHINE LEARNING OF FUNCTIONALLY COMPOSITIONAL STRUCTURES} 
\def\myauthor{Jorge Armando M\'endez M\'endez}
\def\myauthorfull{Jorge Armando M\'endez M\'endez}
\def\mysupervisorname{Eric Eaton}
\def\mysupervisortitle{Ph.D., Research Associate Professor of Computer and Information Science}
\newlength{\superlen}   
\settowidth{\superlen}{\mysupervisorname, \mysupervisortitle} 
\def\gradchairname{Mayur Naik, Ph.D.}
\def\gradchairtitle{Professor of Computer and Information Science}
\newlength{\chairlen}   
\settowidth{\chairlen}{\gradchairname, \gradchairtitle} 
\newlength{\maxlen}
\setlength{\maxlen}{\maxof{\superlen}{\chairlen}}
\def\mydepartment{Computer and Information Science}
\def\myyear{2022}
\def\signatures{32 pt}

\pagenumbering{roman}
\pagestyle{plain}

\begin{titlepage}
\thispagestyle{empty} 
\begin{center}

\onehalfspacing

\mytitle

\myauthorfull

A DISSERTATION

in 

\mydepartment

Presented to the Faculties of the University of Pennsylvania

in 

Partial Fulfillment of the Requirements for the

Degree of Doctor of Philosophy

\myyear

\end{center}

\begin{flushleft}

Supervisor of Dissertation\\[\signatures] 

\renewcommand{\tabcolsep}{0 pt}
\begin{table}[h]
\begin{tabularx}{\maxlen}{l}
\toprule
\mysupervisorname, \mysupervisortitle\\ 
\end{tabularx}
\end{table}

Graduate Group Chairperson\\[\signatures] 

\begin{table}[h]
\begin{tabularx}{\maxlen}{l}
\toprule
\gradchairname, \gradchairtitle\\ 
\end{tabularx}
\end{table}
\singlespacing

Dissertation Committee 

Dan Roth (chair), Ph.D., Eduardo D. Glandt Distinguished Professor of Computer and Information Science

Pratik Chaudhari, Ph.D., Assistant Professor of Electrical and Systems Engineering

Kostas Daniilidis, Ph.D., Ruth Yalom Stone Professor of Computer and Information Science

George Konidaris, Ph.D., John E. Savage Assistant Professor of Computer Science, Brown University

\end{flushleft}

\end{titlepage}

\cleardoublepage    
\doublespacing

\thispagestyle{empty} 

\vspace*{\fill}

\begin{flushleft}
\mytitlecopyright

COPYRIGHT
 
\myyear

\myauthorfull\\[24 pt] 

This work is licensed under the \\
Creative Commons Attribution \\
NonCommercial-ShareAlike 4.0 \\
License

To view a copy of this license, visit

\url{http://creativecommons.org/licenses/by-nc-sa/4.0/}

\end{flushleft}
\pagebreak 

\newenvironment{preliminary}{}{}
\titleformat{\chapter}[hang]{\large\center}{\thechapter}{0 pt}{}
\titlespacing*{\chapter}{0pt}{-33 pt}{6 pt} 
\begin{preliminary}

\cleardoublepage    
\addtocounter{page}{1}
\begin{center}
\topskip0pt
\vspace*{\fill}
\textit{To Laura,\\the center of my world,\\my inspiration to do this}
\vspace*{\fill}
\end{center}

\cleardoublepage
\chapter*{ACKNOWLEDGMENTS}
\addcontentsline{toc}{chapter}{ACKNOWLEDGMENTS} 
Eric Eaton, my advisor, deserves my greatest appreciation. Eric took me under his wing as a na\"ive Master's student, and guided me through the road to completing this dissertation. Along the way, Eric taught me about research, writing, presenting, mentoring, ethics, and countless more. He never shied away from awkward conversations and always patiently supported my endeavors, even when those slowed down my research progress. In short, Eric shaped me into an academic, and I am proud to call him my mentor and friend. I will take his lessons with me to every subsequent chapter of my life. To him, my deepest thanks.

I would also like to thank Pratik Chaudhari, Kostas Daniilidis, George Konidaris, and Dan Roth for serving on my dissertation committee. Their collective feedback was instrumental in the formalization of the problem statement of lifelong compositional learning, which became a central element of this research. Special thanks go to Pratik Chaudhari, for our lengthy and insightful discussions on the meaning of lifelong learning, and to George Konidaris, who mentored me from the distance not only about the research in this dissertation, but also more broadly in crafting my career path.

My thanks go to Harm van Seijen, too, for his advice and collaboration on the development of the lifelong compositional reinforcement learning portion of this dissertation. In particular, his insistence on devising a set of tasks that embodied the notion of compositionality shaped much of the direction that this dissertation took. His input led to the definition of the discrete grid-world tasks, which eventually evolved into the CompoSuite benchmark.

Over the years, I had the fortune of working with wonderful people in the Lifelong Machine Learning group: Shashank Shivkumar, Varun Gupta, Seungwon Lee, Boyu Wang, Meghna Gummadi, Kyle Vedder, Srinath Rajagopalan, Wenxuan Zhang, Abdullah Zaini, Marcel Hussing, and David Kent. They all made the lab environment a fun and collaborative space, for which I am grateful. I give my special thanks to Seungwon Lee, with whom I shared almost all of the graduate school journey; Boyu Wang, who helped develop the theoretical proofs of LPG-FTW; and Marcel Hussing, who co-led the development of CompoSuite.

My gratitude also goes to the faculty and staff of the Penn CIS department and the GRASP lab, for their dedication to making Penn a top place to learn and to do research. I especially thank Joe Devietti and Swapneel Sheth for granting me the opportunity to teach at Penn and for their guidance in navigating the often-inescapable difficulties that come with teaching. 

I also thank my close friend, Rafael Rodr\'iguez, for his long-term support. He always pushed me to be better, and from him I learned the value of asking scientific questions and answering them thoroughly and rigorously. 

My sincere appreciation goes to the professors of Universidad Sim\'on Bol\'ivar, who ensured that my undergraduate education was of the highest quality despite the difficult times that the Venezuelan academia was facing during those years.

I count myself lucky to be a member of two wonderful M\'endez families. I thank all my \textit{t\'ios} and \textit{primos}, for having been examples of professional growth, of integrating into foreign cultures as immigrants, and, above all else, of being happy. I give special thanks to my \textit{t\'ia} Mariela, for her encouragement and unconditional support that allowed me to pursue a graduate education at Penn. I also thank my \textit{abuela}, Lila, for raising our family to treasure education, and my \textit{nono}, Eucario, for his memorable lesson, which I paraphrase into: study, to set yourself free. I have taken their lessons to heart and pursued them to the furthest extent of my abilities.

I especially thank my immediate family. My parents, Tirso and Milagros, taught me to give my all to each enterprise I embarked on and to always do so with rectitude. They are, without question, the most loving, encouraging, and dedicated parents one could ask for. I owe much of who I am as a researcher to their upbringing, which gave me the discipline, drive, and self-confidence required to complete this dissertation. I am grateful to my siblings: Ana Elisa, for being the glue that holds us all together and giving her everything to our family; Mar\'ia Ver\'onica, for showing us that there is more to life than work and teaching us to value all people equally; and Jos\'e Alejandro, for his friendship and camaraderie, especially during the time we shared at Penn, and his uncompromising quest for the truth, which I have sought to imitate. 

And my most profound gratitude goes to my dear wife and the love of my life, Laura. Graduate school is a journey of many ups and downs; Laura found in every accomplishment an excuse to celebrate, and held me through every rejected paper, failed project, and series of sleepless nights. She made every sacrifice I asked of her, and many I did not. Through it all, Laura reminded me of the bigger picture and made sure that we continued to live our life, one filled with joy and love. Her companionship made these years go by like a breeze. For all her love, support, and encouragement, I dedicate this dissertation to her. 

The research in this dissertation was partially supported by the DARPA Lifelong Learning Machines program under grant FA8750-18-2-0117, the DARPA SAIL-ON program under contract HR001120C0040, the DARPA ShELL program under agreement HR00112190133, the Army Research Office under MURI grant W911NF-20-1-0080, and Microsoft Research under the research internship program. 

\cleardoublepage
\chapter*{ABSTRACT}
\addcontentsline{toc}{chapter}{ABSTRACT} 
\begin{center}
\mytitle

\myauthor

\mysupervisorname

\end{center}

A hallmark of human intelligence is the ability to construct self-contained chunks of knowledge and reuse them in novel combinations for solving different yet structurally related problems. Learning such compositional structures has been a significant challenge for artificial systems, due to the underlying combinatorial search. To date, research into compositional learning has largely proceeded separately from work on lifelong or continual learning. This dissertation integrated these two lines of work to present a general-purpose framework for lifelong learning of functionally compositional structures. The framework separates the learning into two stages: learning how to best combine existing components to assimilate a novel problem, and learning how to adapt the set of existing components to accommodate the new problem. This separation explicitly handles the trade-off between the stability required to remember how to solve earlier tasks and the flexibility required to solve new tasks. This dissertation instantiated the framework into various supervised and reinforcement learning (RL) algorithms. Empirical evaluations on a range of supervised learning benchmarks compared the proposed algorithms against well-established techniques, and found that 1)~compositional models enable improved lifelong learning when the tasks are highly diverse by balancing the incorporation of new knowledge and the retention of past knowledge, 2)~the separation of the learning into stages permits lifelong learning of compositional knowledge, and 3)~the components learned by the proposed methods represent self-contained and reusable functions. Similar evaluations on existing and new RL benchmarks demonstrated that 1)~algorithms under the framework accelerate the discovery of high-performing policies in a variety of domains, including robotic manipulation, and 2)~these algorithms retain, and often improve, knowledge that enables them to solve tasks learned in the past. The dissertation extended one lifelong compositional RL algorithm to the nonstationary setting, where the distribution over tasks varies over time, and found that modularity permits individually tracking changes to different elements in the environment. The final contribution of this dissertation was a new benchmark for evaluating approaches to compositional RL, which exposed that existing methods struggle to discover the compositional properties of the environment.

\cleardoublepage
\tableofcontents

\singlespacing
\cleardoublepage
\phantomsection
\vspace*{0.6em}
\listoftables
\addcontentsline{toc}{chapter}{LIST OF TABLES}

\cleardoublepage
\phantomsection
\vspace*{0.6em}
\listoffigures
\addcontentsline{toc}{chapter}{LIST OF ILLUSTRATIONS}
\doublespacing

\end{preliminary}

\newenvironment{mainf}{}{}
\titleformat{\chapter}[hang]{\large\center}{CHAPTER \thechapter}{0 pt}{ : }
\titlespacing*{\chapter}{0pt}{-29 pt}{6 pt} 
\begin{mainf}

\newpage
\pagenumbering{arabic}
\pagestyle{plain} 

\setlength{\parskip}{10 pt} 
\setlength{\parindent}{0pt}

\chapter{Introduction}
\label{sec:introduction}

A major goal of artificial intelligence (AI) is to create an agent capable of acquiring a general understanding of the world. Such an agent would require the ability to continually accumulate and build upon its knowledge as it encounters new experiences. Lifelong machine learning (hereafter referred to as simply lifelong learning) addresses this setting, whereby an agent faces a continual stream of problems and must strive to capture the knowledge necessary for solving each new task it encounters. If the agent is capable of accumulating knowledge in some form of compositional representation, it could then selectively reuse and combine relevant pieces of knowledge to construct novel solutions. This dissertation developed algorithms that enable AI agents to accumulate reusable and compositional knowledge over their lifetimes.

\section{Motivation}
Consider the standard supervised machine learning (ML) setting. The learning agent receives a large labeled data set, and processes this entire data set with the goal of making predictions on data that was not seen during training. The central assumption for this paradigm is that the data used for training and the unseen future data are independent and identically distributed (\iid{}). For example, a service robot that has learned a vision model for recognizing plates in a kitchen will continue to predict plates in the same kitchen. 
As AI systems become more ubiquitous, this \iid~assumption becomes impractical. The robot might move to a new kitchen with different plates, or it might need to recognize cutlery at a later time. If the underlying data distribution changes at any point in time, like in the robot example, then the model constructed by the learner becomes invalid, and traditional ML would require collecting a new large data set for the agent to learn to model the updated distribution. In contrast, a lifelong learning robot would leverage accumulated knowledge from having learned to detect plates and adapt it to the novel scenario with little data. 

The lifelong learning problem is therefore that of learning under a nonstationary data distribution, making use of past knowledge when adapting to the updated distribution. One common formalism for modeling the nonstationarity, which this dissertation adopted, is that of learning a sequence of distinct tasks. In the robot example, three tasks could be detecting plates in the first kitchen, detecting plates in the second kitchen, and detecting cutlery. 

As discussed above, one of the requirements that a lifelong learner should satisfy is to accelerate the learning of future tasks by leveraging knowledge of past tasks. This ability, often denoted \textit{forward transfer}, requires the agent to discover knowledge that is reusable in the future without knowing what that future looks like. A second requirement, which is typically in tension with the first, is that the agent should be capable of performing any task seen in the past even long after having learned it. For example, the robot might move back to the first kitchen after adapting to the second, and it should still be able to recognize plates. This necessitates that the agent avoids catastrophic forgetting~\citep{mccloskey1989catastrophic}, but also that, whenever possible, it achieves \textit{backward transfer} to those earlier tasks~\citep{ruvolo2013ella}. This is possible whenever knowledge from future tasks is useful for learning better models for older tasks. While most work sees avoiding forgetting as a requirement solely for being able to perform well on earlier tasks, in many cases it also permits better forward transfer, by enabling the agent to retain more general knowledge that works for a large set of tasks seen over its lifetime. In addition to these two desiderata, the growth of the agent's memory use should be constrained over time. This choice is practical: if the agent requires storing large models or data sets for all past tasks in memory, then it would become impractical to handle very long task sequences. 

All these desiderata can be summarized as discovering knowledge that is reusable: reusable knowledge can be applied to both future and past tasks without uncontrolled growth. Beyond lifelong learning, the autonomous discovery of reusable knowledge has motivated work in transfer learning, multitask learning (MTL), and meta-learning---all of which deal with  learning diverse tasks. These fields have received tremendous attention in the past decade, leading to a large body of literature spanning supervised learning, unsupervised learning, and reinforcement learning (RL). Traditionally, methods for solving this problem have failed to capture the intuition that, in order for knowledge to be maximally reusable, it must capture a self-contained unit that can be composed with other similar pieces of knowledge. For example, a service robot that has learned to both search-and-retrieve objects and navigate across a university building should be able to quickly learn to put these together to deliver a stapler to Jorge's office. Instead, typical methods make assumptions about the way in which different tasks are related, and impose a structure that dictates the knowledge to share across tasks, usually in the form of abstract representations that are not explicitly compositional.

Yet, compositionality is a tremendously promising notion for achieving the three lifelong learning requirements listed above. Knowledge that is compositional can be used in future tasks by combining it in novel ways; this enables forward transfer. Further, not all knowledge must be updated upon learning new tasks to account for them, but only the components of knowledge that are used for solving these new tasks; this prevents catastrophic forgetting of any unused component and could enable backward transfer to tasks that reuse shared components. Finally, compositional knowledge permits solving combinatorially many tasks by combining in different ways; conversely, solving a fixed number of tasks requires logarithmically many knowledge components, thereby inhibiting the agent's memory growth.

Recently, an increasing number of works have focused on the problem of learning compositional pieces of knowledge to share across different tasks~\citep{zaremba2016learning,hu2017learning,kirsch2018modular,meyerson2018beyond}. At a high level, these methods aim to simultaneously discover \textit{what} are the pieces of knowledge to reuse across tasks and \textit{how} to compose them for solving each individual task. To date, studies in this field have made one of two possible assumptions. The first assumption is that the agent has access to a large batch of tasks to learn simultaneously in an MTL setting. This way, the agent can attempt numerous combinations of possible components and explore how useful they are for solving all tasks jointly. While this assumption certainly simplifies the problem by removing the forward and backward transfer requirements, it is unfortunately unrealistic: AI systems in the real world will not have access to batches of simultaneous tasks, but instead will face them in sequence in a lifelong setting. The second assumption is that the agent does face tasks sequentially, but it is capable of learning components on a single task that are reusable for solving many future tasks. This latter assumption, albeit more realistic, relies on the agent being able to find optimal and reusable components from solving a single task, which is not generally possible given the limited data available for the task and the lack of knowledge about which future components the knowledge must be compatible with. 

\section{Thesis Statement}
\label{sec:Thesis}

The thesis of this dissertation is that learning functionally compositional solutions to a lifelong sequence of tasks improves the capabilities of ML agents to achieve forward transfer, avoidance of forgetting, backward transfer, and limited growth. These capabilities extend to both supervised and RL settings, and become increasingly apparent in the presence of long sequences of highly diverse tasks. In particular, one mechanism that enables the discovery of such compositional solutions is separating the learning process into stages for 1)~initializing a set of components that generalize to future tasks, 2)~discovering how to best combine existing components to solve a new task, and 3)~incorporating new knowledge obtained from the current task into the set of components.

\section{Fundamental Questions}

This dissertation posed and addressed the novel question of \emph{how to learn these compositional structures in a lifelong learning setting}. The purpose of this dissertation was to push the boundaries of lifelong learning methods by creating lifelong learning agents that are capable of continually learning to solve new problems, becoming better learners over time. To this end, this dissertation sought to endow agents with the ability to autonomously discover reusable components as tasks arrive sequentially. 

At an intuitive level, compositionality refers to the ability of an agent to tackle parts of each problem individually, and then reuse the solution to each of the subproblems in combination with others to solve multiple bigger problems that contain shared parts. In particular, this dissertation focused on a form of \textit{functional composition}, where each module or component processes an input and produces an output to be consumed by a subsequent module. Each component therefore is analogous to a function in programming, where functions specialize to solve individual subproblems and combine to solve complex problems. 

One key step towards answering the above question was to study how to formalize the problem of lifelong compositional learning in a way that encapsulates both the supervised and RL paradigms. Critically, such a formulation should 1)~capture realistic settings where an agent might benefit from discovering compositional solutions, and 2)~lend itself to the design of learning algorithms that exploit these compositional properties. Given these requisites, this question directly tied to the next inquiry studied in this dissertation: what steps should a learning algorithm take to tackle such compositional problems and achieve the lifelong learning desiderata of forward transfer, avoidance of forgetting, and limited growth.

Another question this dissertation sought to address was how modularity and compositionality could aid a lifelong learner in dealing with a nonstationary environment. In particular, this dissertation considered the case in which different aspects of the environment change at different rates (and potentially some aspects remain unchanged). In such cases, the challenge is to identify these various shifts and then leverage any information from the past that remains relevant about individual elements of the environment.

This dissertation also endeavored to answer how to evaluate approaches specifically in the RL setting to study their compositional capabilities. This involved the investigation of a set of problems that are explicitly compositionally related, and the design of performance metrics that directly measure an RL agent's ability to discover the tasks' compositional structure.

\section{Technical Contributions}

Seeking to answer these fundamental questions, the primary contribution of this dissertation was the development of a general-purpose framework that is agnostic to the specific algorithms used for learning and the form of the structures themselves. The proposed framework is capable of incorporating various forms of compositional structures, techniques for learning to combine these structures, and mechanisms for avoiding catastrophic forgetting. As examples of its flexibility, this dissertation instantiated the framework to incorporate linear parameter combinations and multiple forms of neural net module compositions as the compositional structures; backpropagation, policy gradient (PG) learning, and discrete search to discover the optimal combination of components for each task; and experience replay, elastic weight consolidation~\citep{kirkpatrick2017overcoming}, and off-line RL as knowledge retention mechanisms. Various novel combinations of these examples resulted in new lifelong learning algorithms, and an extensive empirical evaluation validated these methods on new and existing benchmark problems, demonstrating that the proposed framework increases the capabilities of the learning system, reducing catastrophic forgetting and achieving higher overall performance. Moreover, this evaluation verified that the components learned by algorithms within the proposed framework correspond to self-contained, reusable functions. 

Building upon these foundations, this dissertation then developed an extension of the framework to bring lifelong learning closer to real-world deployment. This extension dealt with the problem of nonstationary lifelong learning, in which aspects of the environment change dynamically over time. As one final contribution, this dissertation developed a new large-scale evaluation benchmark specifically for assessing the compositionality of RL methods, as a means for fostering future advancements in this direction.

\subsection{A General-Purpose Framework for Lifelong Learning of Compositional Structures}

The first step toward answering the questions outlined above was to formalize the problem of lifelong learning of compositional structures. In particular, this dissertation defined the problem  around the notion of functional composition, under the assumption that the solution to each task the agent might encounter throughout its lifetime can be solved by executing functions one after the next to process the input and transform it into the desired output. This formulation gave rise to a compositional problem graph, which motivated the design of the neural architectures used in the algorithmic instantiations of the proposed framework. 

As a general-purpose solution to lifelong compositional learning, this dissertation proposed a framework that evokes Piaget's~(\citeyear{piaget1976piaget}) assimilation and accommodation stages of human intellectual development, embodying the benefits of dividing the lifelong learning process into two distinct stages. In the first stage, the learner strives to solve a new task by combining existing components it has already acquired. The second stage uses discoveries from the new task to improve existing components and to construct fresh components if necessary.  

The framework definition is broad by design, yet it still provides significant insight into how to design lifelong learning algorithms. The intuition is that, when learning a new task, the agent has not acquired any knowledge about that task. Therefore, modifying existing components containing solidified knowledge might catastrophically damage them by incorporating likely incorrect information about the current task. Instead, the agent should leverage the existing knowledge as much as possible for discovering information about the new task. Later, once the agent has learned the new task, it should incorporate any knowledge about this current task that might be useful for solving future tasks into the set of existing components. If the agent finds the existing components to be insufficient for solving the new task, then it should create new components to incorporate this new knowledge.

\subsection{Lifelong Composition in Supervised Learning}
The first instantiations of the framework developed in this dissertation were a set of nine lifelong supervised learning algorithms. Each of the methods trains one form of compositional structures from among linear parameter combinations, soft neural layer ordering~\citep{meyerson2018beyond}, and a soft version of neural layer gating~\citep{kirsch2018modular}. On the other hand, in order to avoid catastrophic forgetting of the component parameters when updating them with future knowledge, each method uses a choice from na\"ive fine-tuning, elastic weight consolidation~\citep{kirkpatrick2017overcoming}, and experience replay as knowledge retention mechanisms. Theoretical computational complexity bounds for each of these algorithms demonstrated considerable efficiency gains in training time. 
A comprehensive empirical evaluation tested all the introduced algorithmic instantiations, with two primary goals: 1)~showing that compositional structures enable improved learning, especially in the presence of highly diverse tasks, and 2)~showing that the proposed framework, and in particular the separation of the learning process into two distinct stages, improves the learning of these compositional solutions. Beyond simply ascertaining \textit{if} these results hold, this dissertation sought to discover \textit{why} these results hold. As intermediate questions, the empirical evaluation measured how much knowledge of past tasks the agent forgot after training on future tasks, as well as how general the knowledge stored in the components became as it was updated with more tasks. Additional questions studied in the experiments to test the usefulness of the proposed framework were how rapidly the number of trained components grew, how the results changed in the presence of varying sample sizes, how reusable the learned components were across various tasks, and how the schedule for assimilation and accommodation should be set. As a final test, a brief visualization experiment inspected the meaning of some components learned in an image generation task, with the goal of discovering whether the learned components were indeed self-contained and reusable.

\subsection{Lifelong Composition in Reinforcement Learning}

RL has achieved impressive success at complex tasks, from mastering the game of Go~\citep{silver2016mastering,silver2017mastering} to controlling complex robots~\citep{gu2017deep,openai2020learning}. However, this success has been mostly limited to solving a single problem given enormous amounts of experience. In some settings, this experience is prohibitively expensive, such as when training an actual physical system. If an agent is expected to learn multiple consecutive tasks over its lifetime, then it would be ideal for it to leverage knowledge from previous tasks to accelerate the learning of new tasks. This is the premise of lifelong RL methods. 

Progress in lifelong RL has been substantially slower than in the supervised counterpart, with only a minuscule portion of lifelong learning research being devoted to RL. In the context of this dissertation, this implied that unlike in the supervised setting, it was not possible to develop a multitude of lifelong RL methods by simply combining existing pieces with minor adaptations. Instead, in order to adapt the proposed framework to RL, this dissertation created two novel lifelong RL methods from first principles, in particular developing two new mechanisms for avoiding catastrophic forgetting and achieving backward transfer.

As a first step toward adapting the framework to lifelong RL, this dissertation formulated the novel problem of lifelong RL of functionally compositional tasks, where tasks can be solved by recombining modules of knowledge in different manners. The intuition is that humans' ability to handle diverse problems stems from their capacity to accumulate, reuse, and recombine perceptual and motor abilities in various ways to handle novel circumstances. While RL has studied {\em temporal} compositionality for a long time, such as in the options framework, it has not explored in depth the type of {\em functional} compositionality studied in this dissertation, especially not in the more realistic lifelong learning setting.
Functional compositionality involves a decomposition into subproblems, where the outputs of one subproblem become inputs to others. This moves beyond standard temporal composition to functional compositions of layered perceptual and action modules, akin to programming where functions are used in combination to solve different problems. 
For example, a typical robotic manipulation solution interprets perceptual inputs via a sensor module, devises a path for the robot using a high-level planner, and translates this path into motor controls with a robot driver. Each of these modules can be used in other combinations to handle a variety of tasks. The formalization of this problem adapted the compositional problem graph from the supervised setting to the RL setting.

One of the fundamental differences between lifelong RL and lifelong supervised learning is that, in RL, the agent is in charge of collecting its own experiences. This imposes a trade-off between the ability of the agent to discover new regions of its environment and new behaviors (\textit{exploration}) and its ability to perform well in the environment (\textit{exploitation}). In practice, in order for an agent to learn complex RL tasks with present-day methods, it is necessary for it to learn a strong exploitation policy and apply minor perturbations over this policy for exploration. Consequently, an agent that learns an exploitation policy using fewer exploration steps can learn to solve a task after fewer interactions with the environment. In a lifelong RL setting, this motivates a new metric beyond the standard forward and backward transfer: the speed of learning each new task. Due to its practical value, this dissertation treated speed of learning as the main metric of performance for RL methods. 

The first lifelong compositional RL method developed in this dissertation wraps around base PG methods. This lifelong PG algorithm decomposes the parameters of each task's policy into a set of components that are linearly combined in a task-specific manner to construct the desired policy. The method then leverages the existing components during the training process for each task to accelerate the learning, by searching only over combinations of those components. The approach further exploits the linear combination structure to obtain a closed-form solution for incorporating knowledge about each new task into the existing components without causing forgetting. This simplified structure also permitted deriving theoretical results about the convergence of the knowledge components. An empirical evaluation assessed the performance of this first method on a range of continuous control tasks, including a set of complex and diverse robotic manipulation problems, where it exhibited accelerated learning via lifelong transfer.

The second method more explicitly captures the structure of the compositional problem graph via deep modular architectures. Unlike other methods in this dissertation, this approach separates the initial stage for assimilating the current task into two substages: first, the agent finds the optimal combination of existing modules via discrete search; second, the agent leverages the composed modular architecture to explore the environment via standard deep RL. Then, in order to accommodate new knowledge into the existing components without forgetting and potentially with backward transfer, the agent uses off-line RL on a small replay buffer of data collected from all previously seen tasks. This newfound connection between lifelong RL and off-line RL applies beyond compositional RL to many existing lifelong RL methods, and is a promising mechanism for enabling backward transfer more generally.

Since this latter method learns an architecture that more closely matches the compositional problem graph, the evaluation inspected its compositional properties. For this, the experiments considered two novel suites of explicitly compositional tasks: one in a discrete 2-D environment and another in a continuous robotic manipulation setting. The first part of the experiments studied whether the tasks truly exhibited their intuitive compositional properties. The second part evaluated the proposed algorithm for its ability to quickly learn new tasks and improve performance of earlier tasks with further updates to the knowledge modules.

Due to the inherent complexity of modern RL methods, this dissertation did not develop the final step of incorporating novel components for these two RL algorithms. The question of how to automatically grow the set of policy modules over time in lifelong RL remains open for future investigation.

\subsection{Nonstationary Lifelong Learning}
While lifelong learning is itself a problem of nonstationary learning, in some sense it sidesteps the challenge of nonstationarity by adding a second-layer distribution (over tasks) which is stationary. This enables existing approaches to utilize \iid~learning techniques at the higher level of the task distribution (e.g., by treating each task as analogous to an individual \iid~data point). However, in many realistic deployments, this assumption of task-distribution stationarity does not hold. In such cases, approaches should explicitly consider the nonstationary nature of the environment and incorporate techniques for actively forgetting previous knowledge that has become obsolete in order to enable adaptation to the current state of the world.

This dissertation developed an extension of one lifelong compositional RL method to this nonstationary setting. In particular, the extended algorithm assumes that aspects (or components) of the environment vary independently of each other, and leverages its compositional construction to tackle each component's distributional shift individually. An empirical evaluation explored multiple variants of this approach and evaluated their ability to handle modular nonstationarity on various nonstationary environments.  

\subsection{A Benchmark for Compositional Reinforcement Learning}

Long before this dissertation, AI research has sought to embed compositionality into intelligent systems for decades, from early ideas like hierarchical planning~\citep{sacerdoti1974planning} and logic-based reasoning~\citep{doyle1979truth}, all the way to modern learning-based techniques like neural module networks~\citep{andreas2016neural} and skill discovery~\citep{konidaris2009skilldiscovery}. The ability to decompose a complex problem into easier subproblems would drastically increase the capabilities of learning agents by 1)~making each learning problem easier (because each subproblem is easier than the original problem), and 2)~enabling the agent to very quickly learn to solve new tasks by discovering which components from its earlier solutions are suitable to these new tasks (potentially without ever requiring any data from the new tasks).

Despite the intuitive appeal of these ideas, the RL community has only recently begun to explore the full potential of AI agents to leverage compositional properties of the environment to generalize to \textit{unknown} combinations of \textit{known} components. This dissertation sought to foster progress in this direction by introducing \benchmark{}, a benchmark for compositional RL that exploits the compositionality of robot learning tasks to evaluate the compositional capabilities of learning agents. The benchmark follows the functional compositional RL formulation of previous chapters, which 
is akin to the decomposition of programs for solving robot tasks into software modules for sensing, planning, and acting.

Following this intuition, each \benchmark{} task requires a particular \textit{robot} arm to manipulate one individual \textit{object} to achieve a task \textit{objective} while avoiding an \textit{obstacle}. For example, one \benchmark{} task requires an IIWA arm to circumvent a wall, pick up a dumbbell, and place it in a bin. Another task instructs a Jaco arm to traverse a doorway, pick up a plate, and place it on a shelf. This compositional definition of the tasks endows \benchmark{} with two remarkable properties. First, varying the robot/object/objective/obstacle elements leads to hundreds of RL tasks, each of which requires a meaningfully different behavior. Second, \benchmark{} can evaluate RL approaches specifically for their ability to learn the compositional structure of the tasks. Intuitively, if a learning agent is able to appropriately decompose its solutions to the dumbbell and plate problems into functional components, then it could reuse the IIWA motor module in place of the Jaco motor module in order to solve the plate-on-shelf task without any experience with the IIWA arm on that task. More generally, \benchmark{} can evaluate (noncompositional) multitask and lifelong RL approaches for their ability to handle large numbers of highly varied tasks. This is in stark contrast to most existing multitask RL benchmarks, which are typically limited to at most a few dozen RL tasks: \benchmark{} offers an order of magnitude more tasks, enabling the study of multitask and lifelong RL at scale.

Concretely, \benchmark{} contains $256$ compositional simulated robotic manipulation tasks, all of which require meaningfully different behaviors. In addition, \benchmark{} prescribes various standardized evaluation schemes and metrics in order to foster reproducible evaluation of future approaches. An empirical evaluation of single-task learning (STL), monolithic MTL, and modular MTL agents on various evaluation settings sought to 1)~validate the compositional properties of \benchmark{}, 2)~devise a deeper understanding of the implications of compositionality on RL agents, and 3)~demonstrate that there remains substantial room for improvement in existing RL methods to exploit the compositional properties of \benchmark{}.

\section{Manuscript Structure}

Subsequent chapters of this manuscript are organized as follows:

\paragraph{Chapter~\ref{cha:RelatedWork} -- \nameref{cha:RelatedWork}} contextualizes this dissertation in the broader field of research. In particular, it categorizes existing works along six axes: the learning setting, whether and how the structure of the problem is given, the application domain, the type of composition, the learning mechanism, and the form of the structures. 

\paragraph{Chapter~\ref{cha:Framework} -- \nameref{cha:Framework}} formalizes the problems of lifelong learning and compositional learning, and proposes a framework for addressing the novel problem of lifelong compositional learning. 

\paragraph{Chapter~\ref{cha:Supervised} -- \nameref{cha:Supervised}} instantiates the framework into a suite of nine lifelong supervised learning algorithms and executes an extensive evaluation to understand the properties of the framework.

\paragraph{Chapter~\ref{cha:RL} -- \nameref{cha:RL}} adapts the problem formulation of lifelong composition to the RL setting, proposes two novel mechanisms for lifelong compositional RL, and evaluates each of the algorithms in a set of existing and new benchmark problems.

\paragraph{Chapter~\ref{cha:NonStationary} -- \nameref{cha:NonStationary}} extends one of the lifelong compositional RL algorithms to the setting of nonstationary lifelong RL, and evaluates a variety of choices for how to handle changes in the environment in a range of nonstationary settings.

\paragraph{Chapter~\ref{cha:Benchmark} -- \nameref{cha:Benchmark}} introduces the \benchmark{} benchmark for evaluating compositional properties of RL algorithms, and evaluates existing approaches under a variety of experimental settings.

\paragraph{Chapter~\ref{cha:Conclusion} -- \nameref{cha:Conclusion}} summarizes the technical contributions of this dissertation, as well as the findings obtained from these contributions and how they answer the fundamental questions posed in this introductory chapter. This chapter closes with a discussion of avenues for future investigation that would potentially have significant impact on the field.

\biblio

\chapter{Related Work}
\label{cha:RelatedWork}

\section{Introduction}

This chapter reviews the existing literature on the topics related to this dissertation, with an aim to contextualize this investigation in the broader research landscape. The survey is primarily a separate discussion of two topics that are closely related, yet previously disjointly studied: lifelong or continual learning and compositional knowledge representations. Research into lifelong learning seeks to endow agents with the capability to accumulate knowledge over a nonstationary stream of data, typically presented to the agent in the form of tasks. In principle, if the tasks are related in some way, the agent should be able to detect and extract the commonalities across the tasks in order to leverage shared knowledge and improve its overall performance. On the other hand, the goal of learning compositional knowledge representations is to decompose complex problems into simpler subproblems, such that the solutions to the easier subproblems can be combined to solve the original, harder problem. This formulation makes compositional representations an appealing mechanism for learning the relations across multiple tasks: by discovering subproblems that are common to many tasks, the learner could reuse the solutions to these subproblems as modules that compose in different combinations to solve the many tasks. Despite its intuitive appeal, no prior work had explicitly used compositional representations as a means for transferring knowledge across a lifelong sequence of tasks. This dissertation leveraged techniques from across lifelong learning and compositional representations to define general-purpose algorithms that discover compositional structures in a lifelong learning setting. 

\begin{figure}[p]
\centering
    \begin{subfigure}[b]{0.28\textwidth}
        \raisebox{2.5cm}{No structure given}
    \end{subfigure}%
    \begin{subfigure}[b]{0.2\textwidth}
    \begin{flushright}
        \raisebox{1.2cm}{Supervised}\\
        \raisebox{1.4cm}{Unsupervised}\\
        \raisebox{0.8cm}{RL}
    \end{flushright}
    \end{subfigure}%
    \begin{subfigure}[b]{0.52\textwidth}
        \begin{minipage}[t]{0.19\linewidth}
        \phantom{nl}
        \end{minipage}%
        \begin{minipage}[t]{.27\linewidth}
        \centering
        \subcaption*{Lifelong}
        \end{minipage}%
        \begin{minipage}[t]{.27\linewidth}
        \centering
        \subcaption*{Multitask}
        \end{minipage}%
        \begin{minipage}[t]{.27\linewidth}
        \centering
        \subcaption*{Single-task}
        \end{minipage}  
        \begin{minipage}[t]{0.09\linewidth}
        \end{minipage}%
        \centering
        \vspace{-0.5em}
        \includegraphics[width=0.87\linewidth, trim={0 0 0 0}, clip]{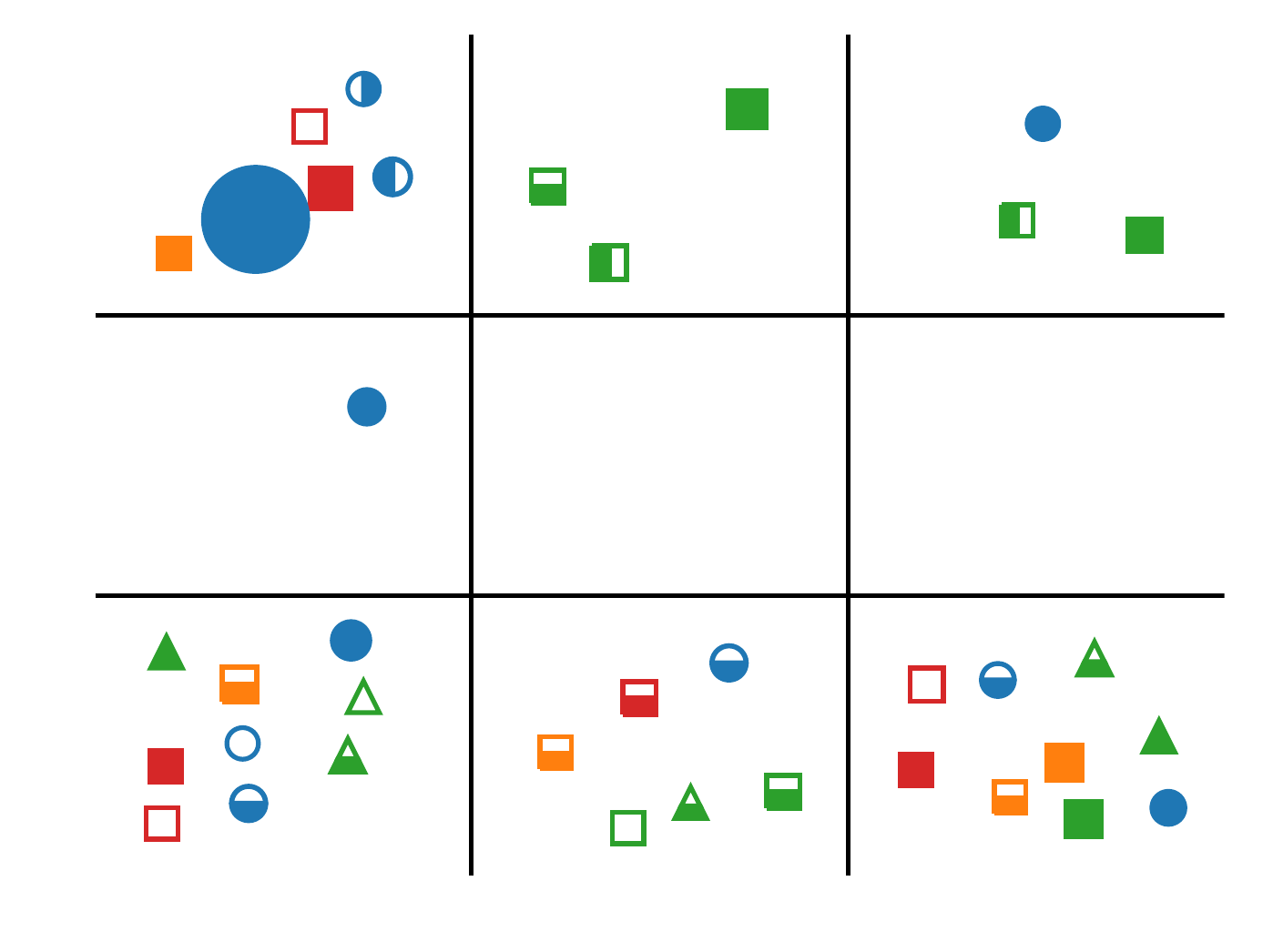}
    \end{subfigure}\\
    \vspace{1em}
    \begin{subfigure}[b]{0.28\textwidth}
        \raisebox{2.5cm}{Structure implicitly given}
    \end{subfigure}%
    \begin{subfigure}[b]{0.2\textwidth}
    \begin{flushright}
        \raisebox{1.2cm}{Supervised}\\
        \raisebox{1.4cm}{Unsupervised}\\
        \raisebox{0.8cm}{RL}
    \end{flushright}
    \end{subfigure}%
    \begin{subfigure}[b]{0.52\textwidth}
        \centering
        \includegraphics[width=0.87\linewidth, trim={0 0 0 0}, clip]{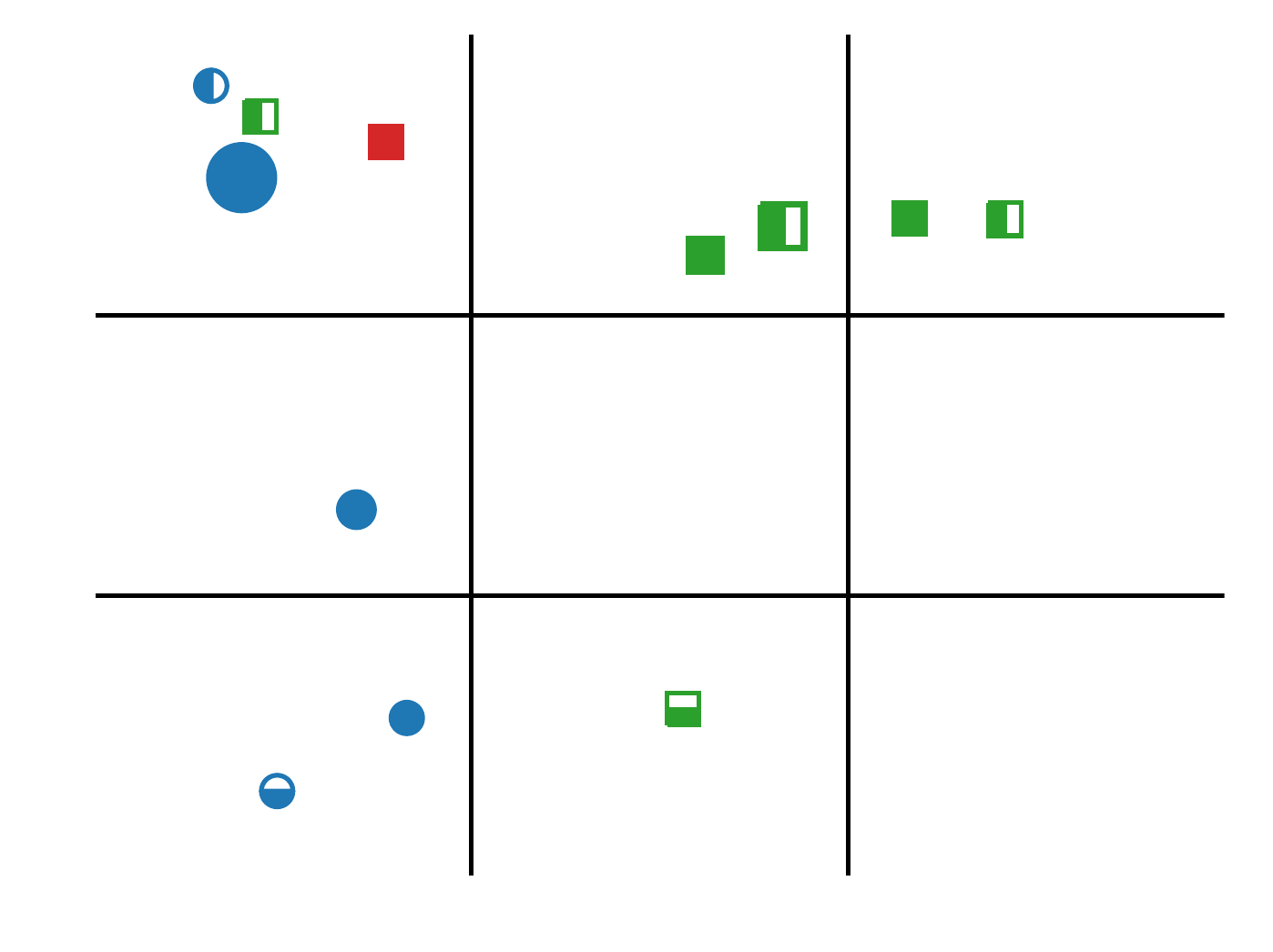}
    \end{subfigure}  
    \\
    \vspace{1em}
    \begin{subfigure}[b]{0.28\textwidth}
        \raisebox{2.5cm}{Structure explicitly given}
    \end{subfigure}%
    \begin{subfigure}[b]{0.2\textwidth}
    \begin{flushright}
        \raisebox{1.2cm}{Supervised}\\
        \raisebox{1.4cm}{Unsupervised}\\
        \raisebox{0.8cm}{RL}
    \end{flushright}
    \end{subfigure}%
    \begin{subfigure}[b]{0.52\textwidth}
        \centering
        \includegraphics[width=0.87\linewidth, trim={0 0 0 0}, clip]{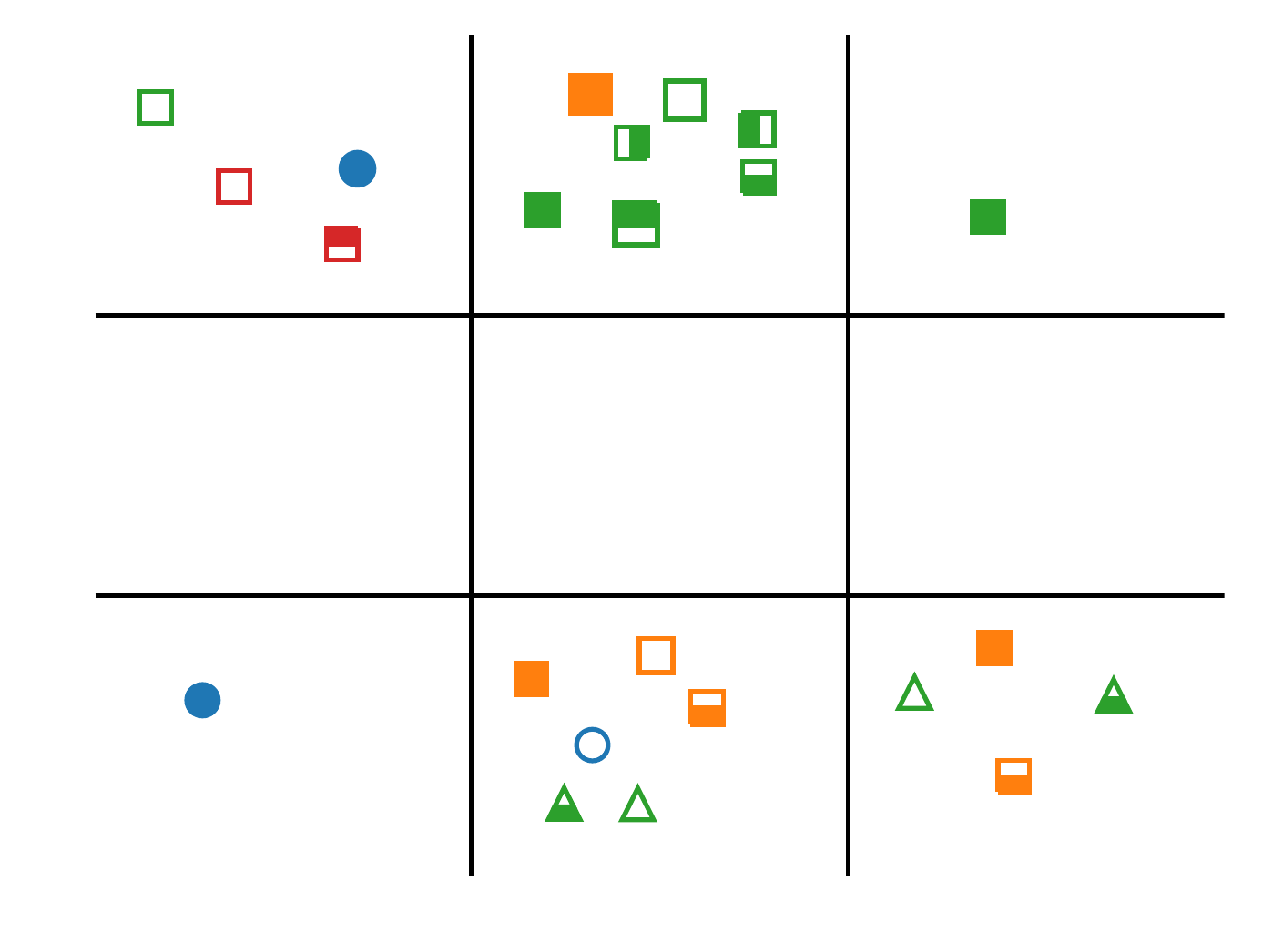}
    \end{subfigure}\\
    \begin{subfigure}[b]{\textwidth}
        Type of composition: 
        \includegraphics[width=0.25cm]{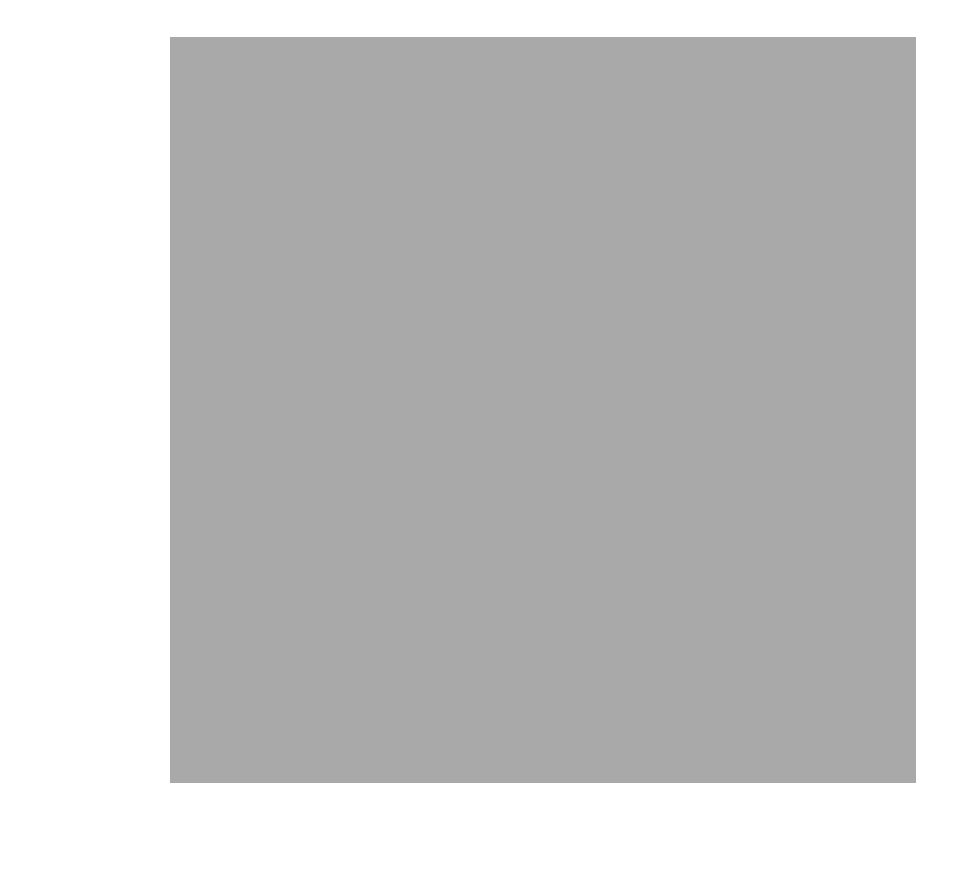} Functional, \includegraphics[width=0.25cm]{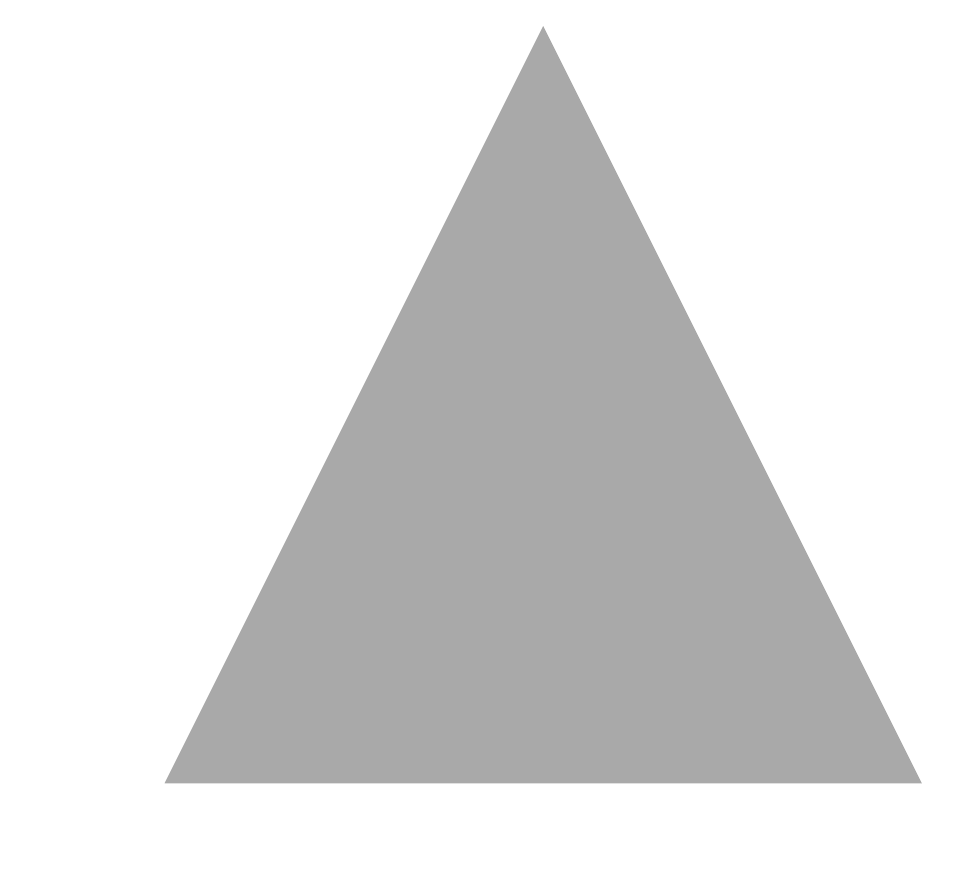} Temporal, \includegraphics[width=0.25cm]{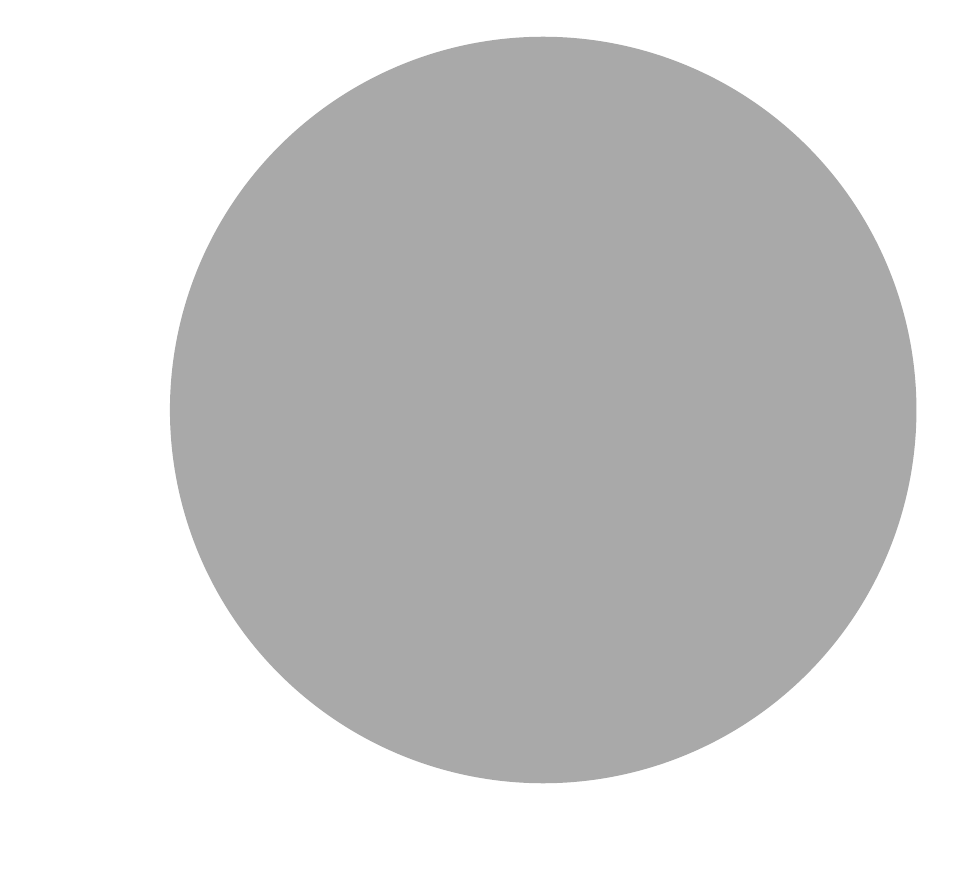} None
    \end{subfigure}
    \begin{subfigure}[b]{\textwidth}
        Type of structural configuration: 
        \includegraphics[width=0.25cm]{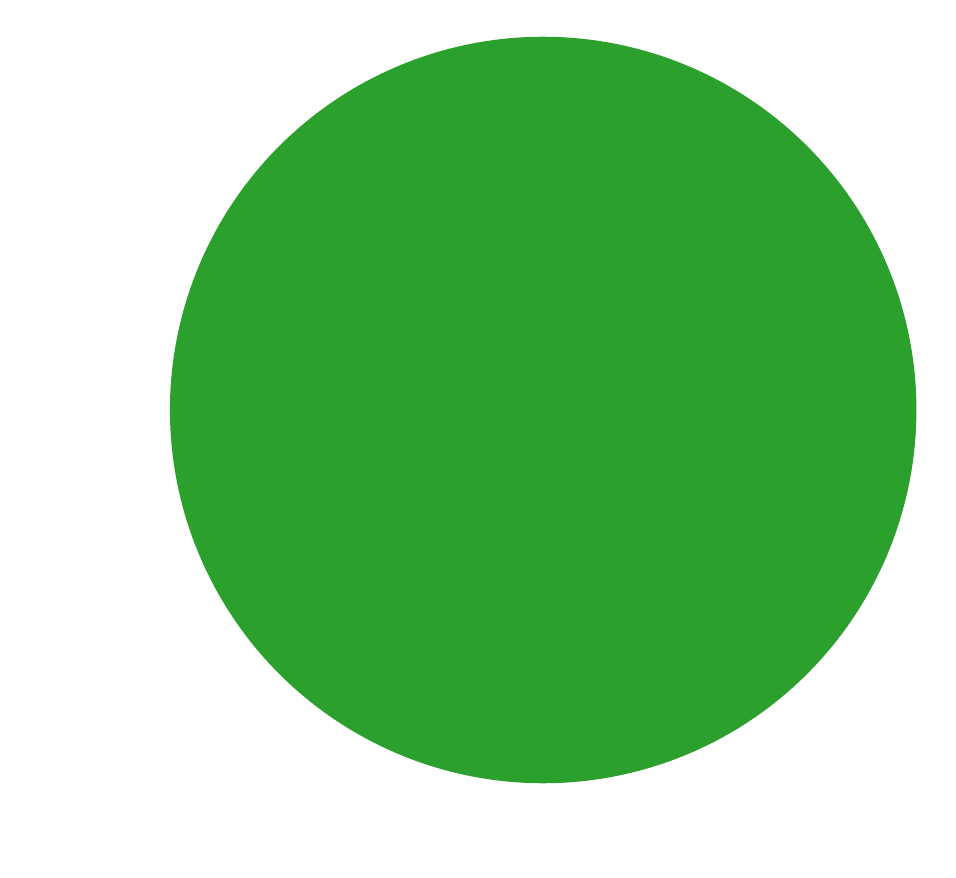} Graph, \includegraphics[width=0.25cm]{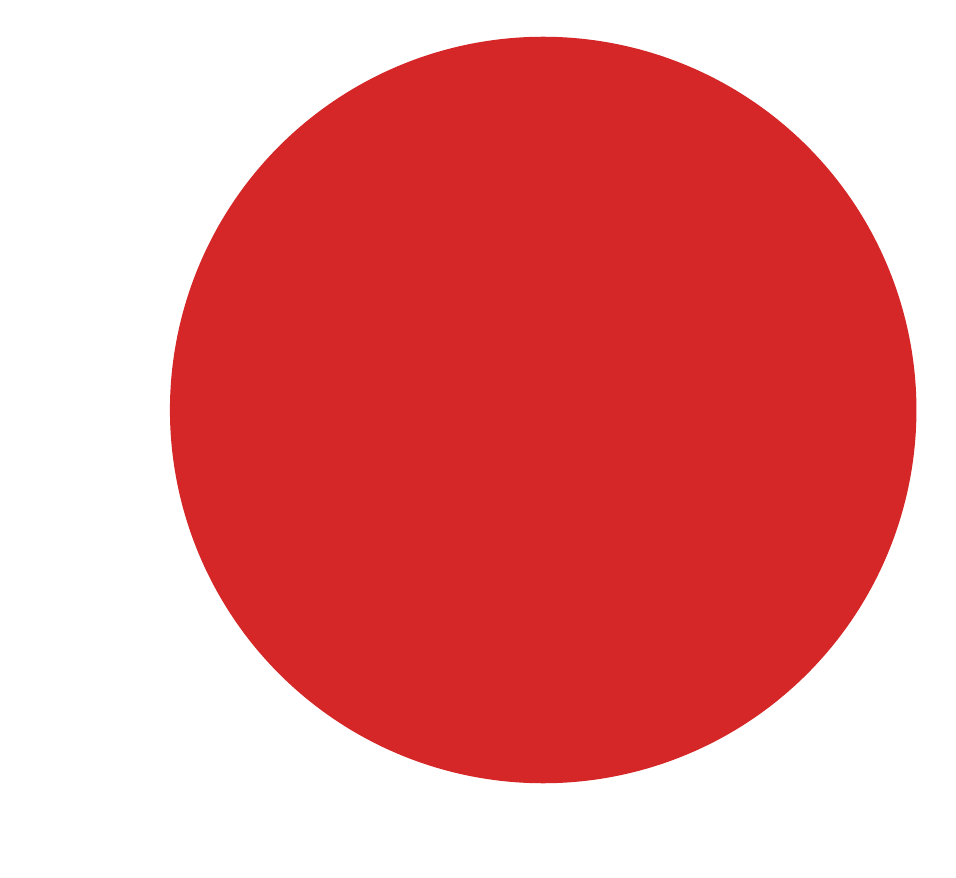} Chaining, \includegraphics[width=0.25cm]{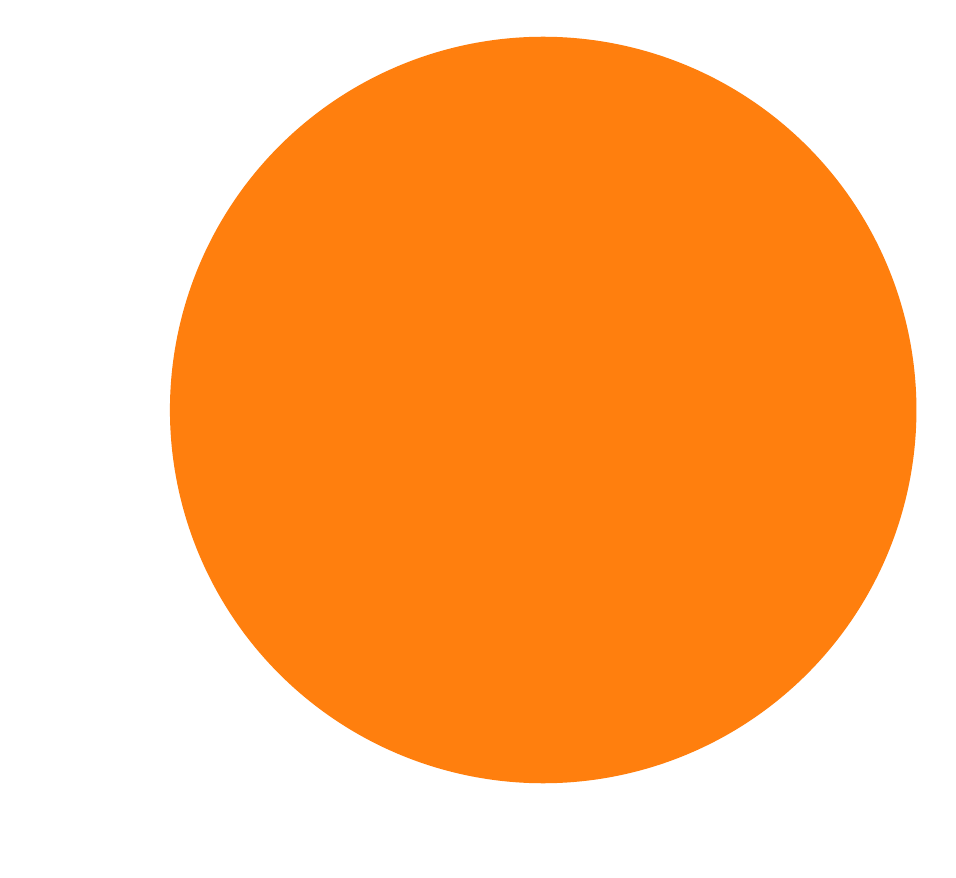} Aggregation, \includegraphics[width=0.25cm]{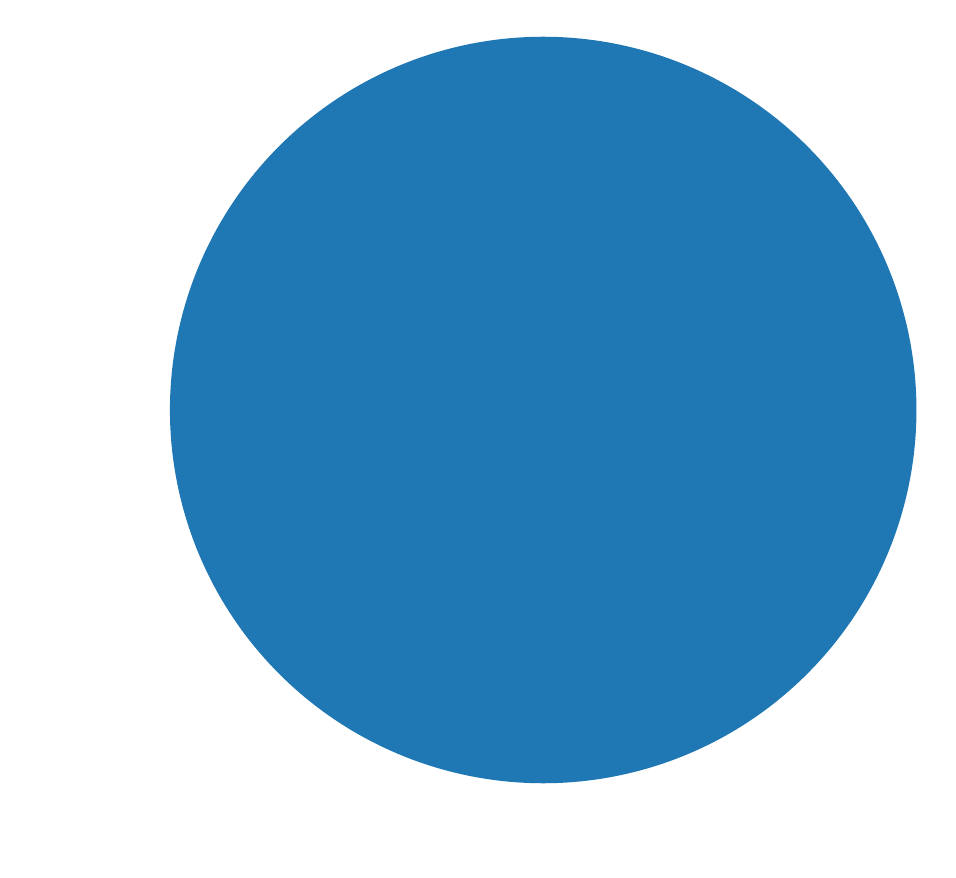} None
    \end{subfigure}
    \begin{subfigure}[b]{\textwidth}
        Application domain: 
        \includegraphics[width=0.25cm]{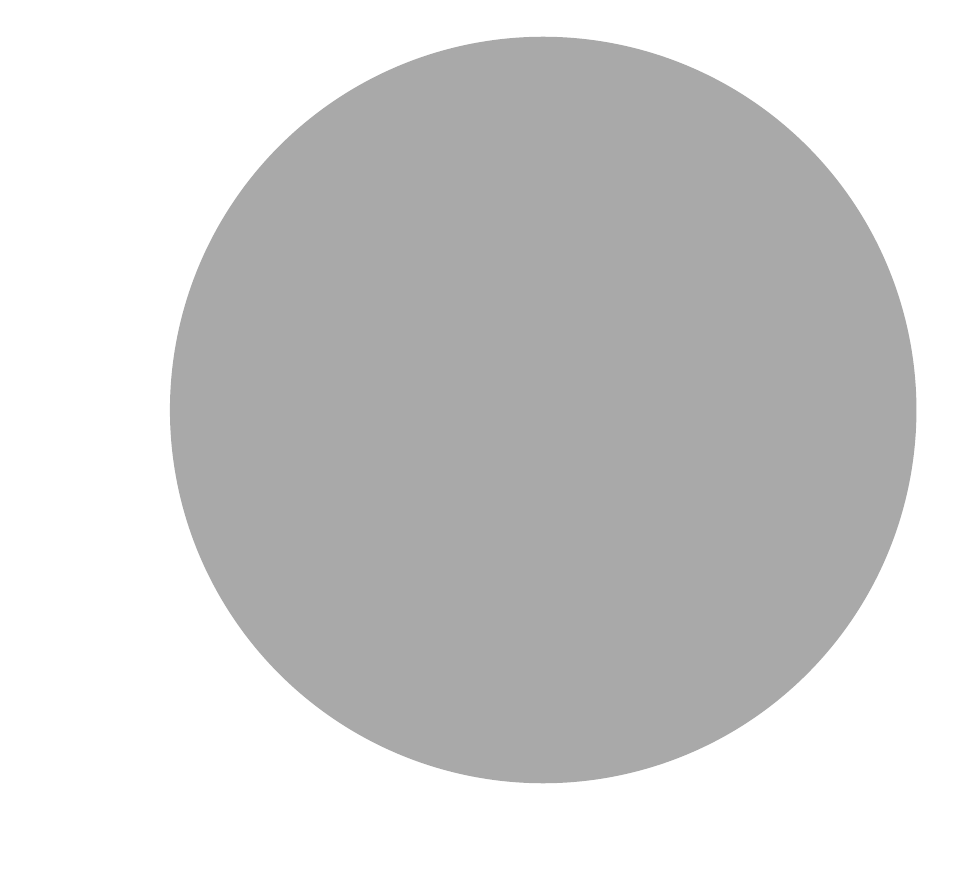} Vision, \includegraphics[width=0.25cm]{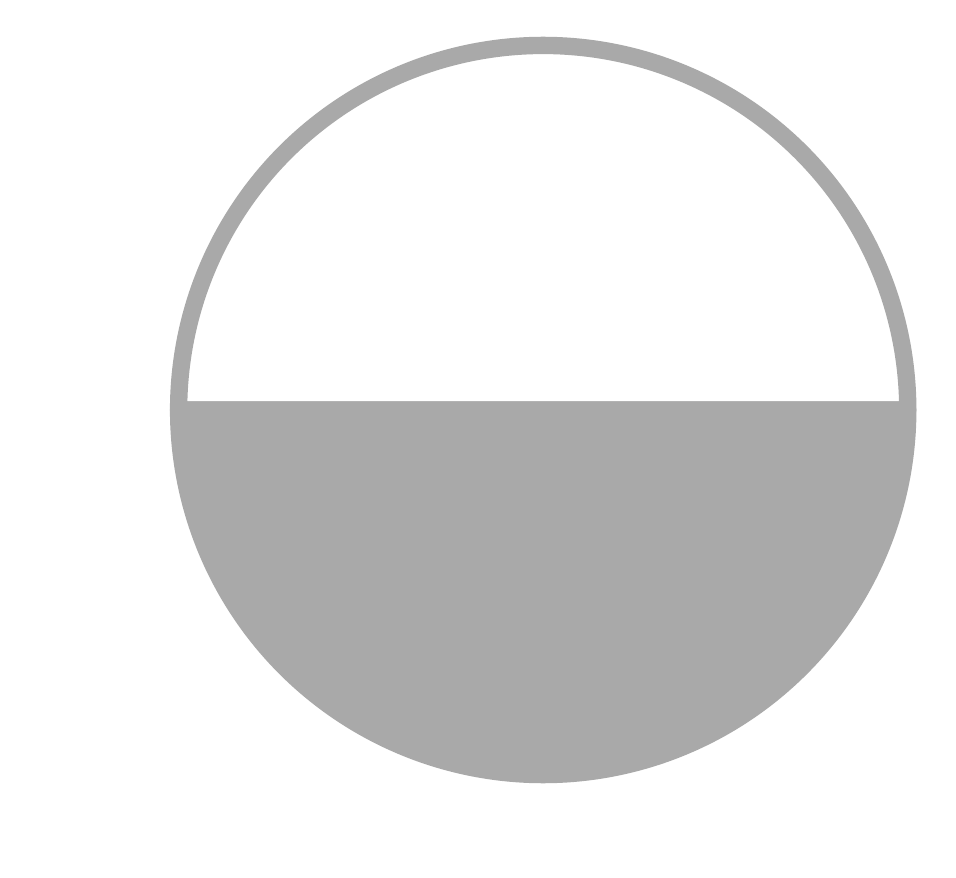} Robotics, \includegraphics[width=0.25cm]{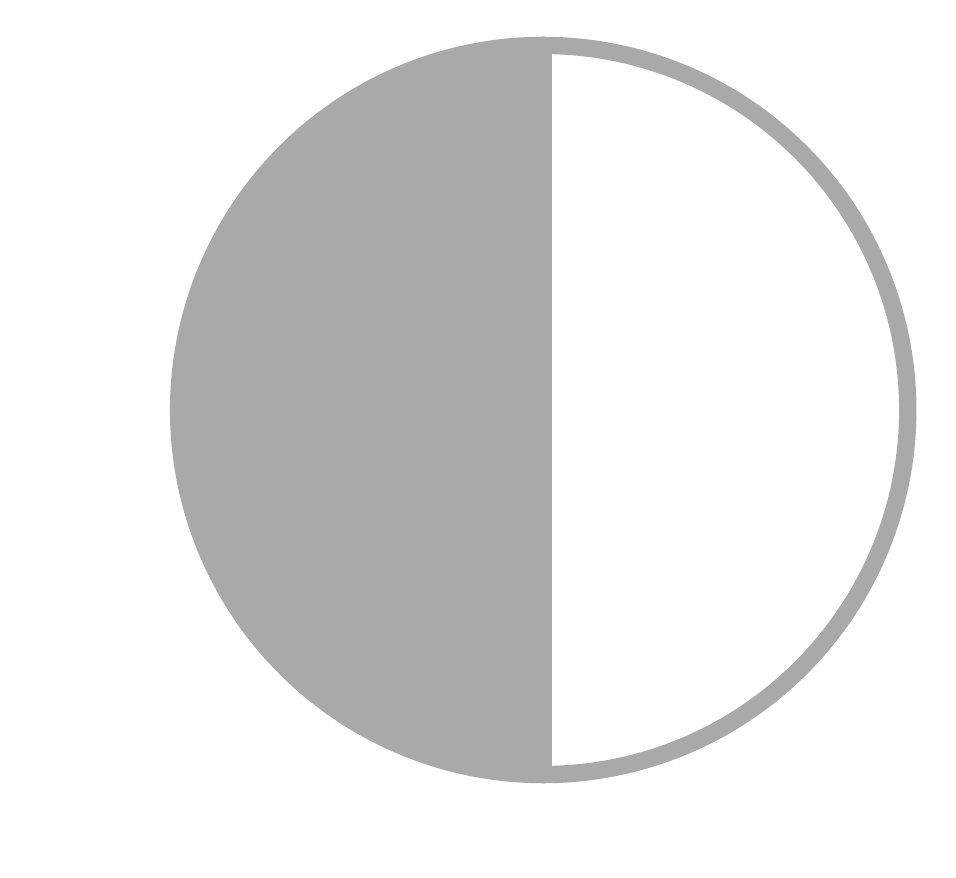} Language, \includegraphics[width=0.25cm]{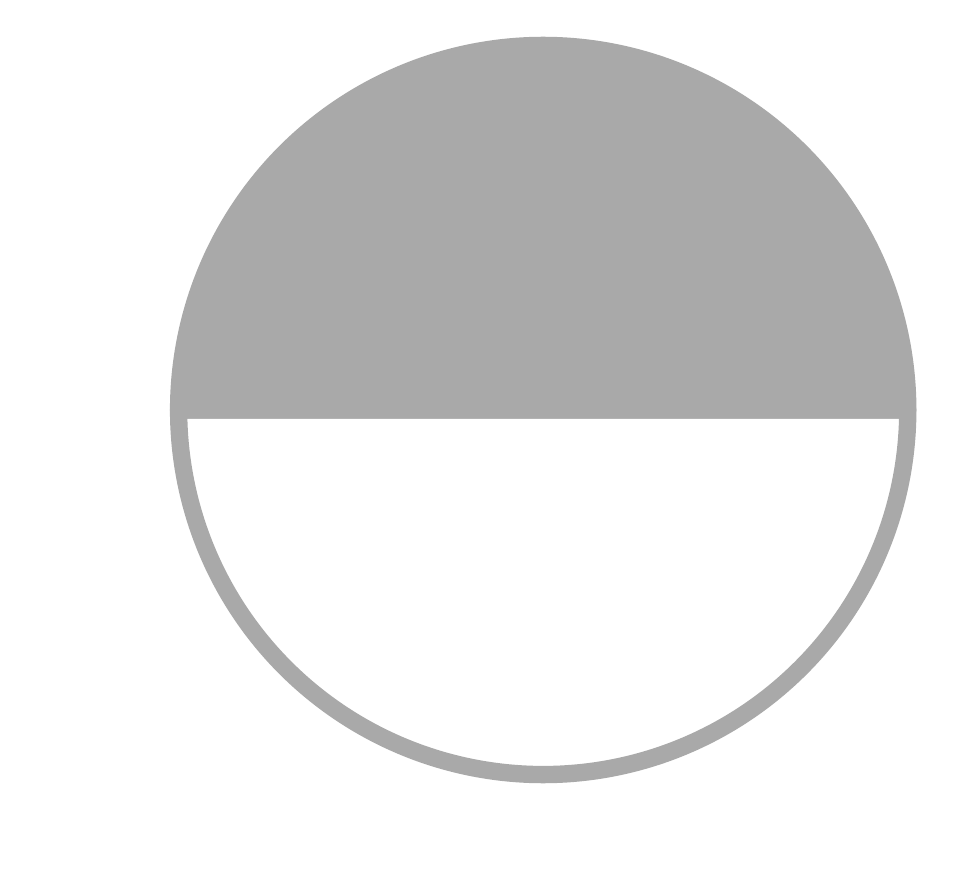} VQA, \includegraphics[width=0.25cm]{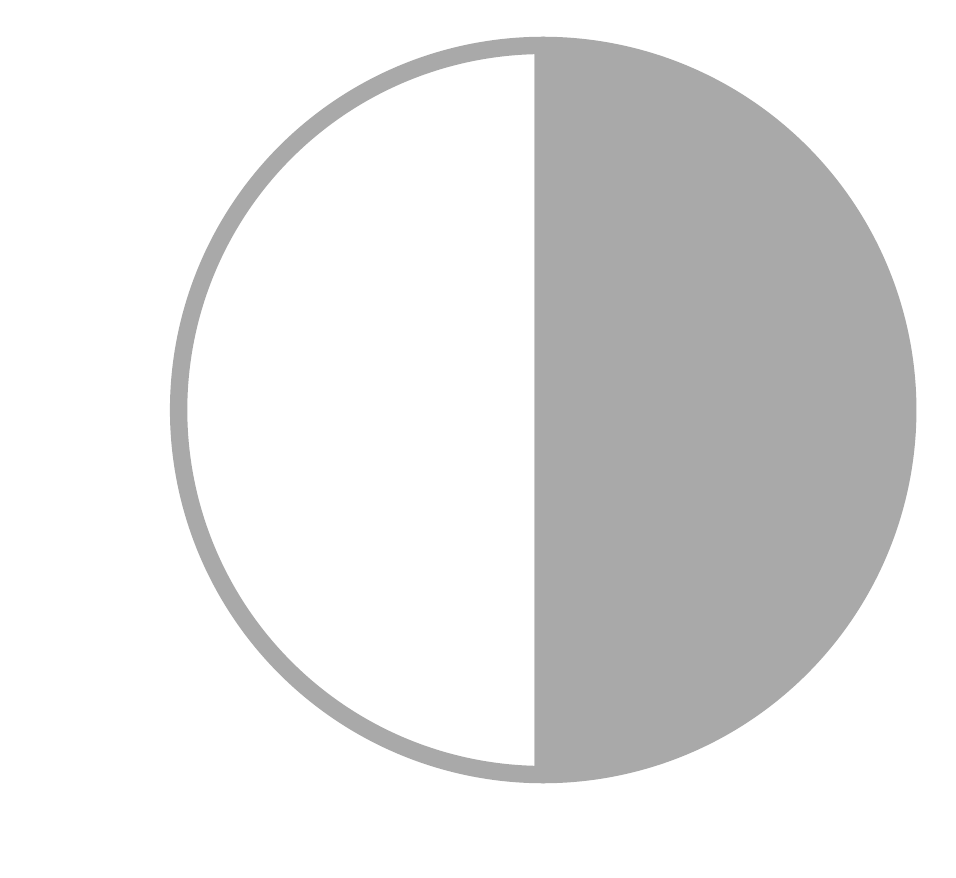} Audio, \includegraphics[width=0.25cm]{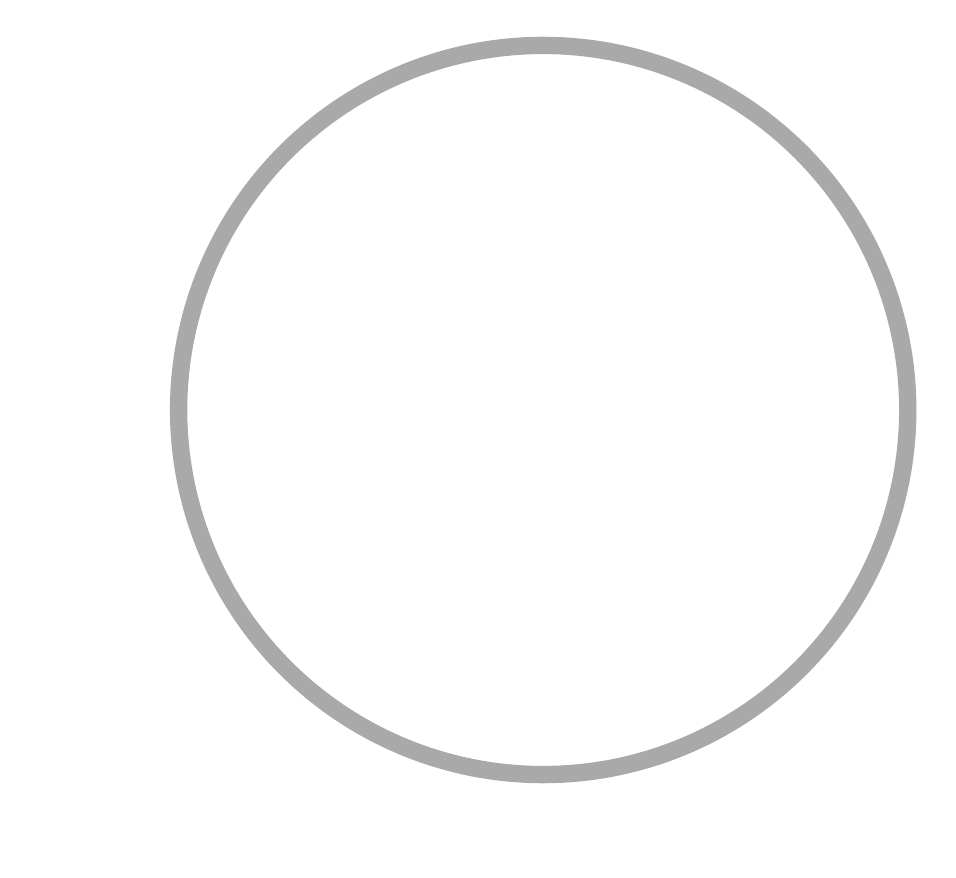} Toy 
    \end{subfigure}
    \caption[Related work categorization.]{A categorization of existing works into six axes, defined in text within the figure; scale represents number of references in this chapter per category. Most work on lifelong or continual learning has not learned explicitly compositional structures, while most efforts on compositional learning have operated in the MTL or STL settings. This dissertation created some of the very first methods developed for lifelong compositional learning. Appendix~\ref{app:RelatedWork} contains a tabular version of this figure, listing all references in their respective categories.}
    \label{fig:RelatedWork}
    \vspace{-1em}
\end{figure}

The discussion further divides works into those carried out in the supervised and the RL settings. While many of the techniques used for one are applicable to the other (albeit with minor-to-major adaptations), research into these two fields has proceeded mostly separately, with the vast majority of work focusing on the supervised setting. In particular, the form of functional composition studied in this dissertation had previously been almost entirely overlooked in the RL literature. The general-purpose framework developed in this investigation adapts to both supervised and RL settings, demonstrating the broad applicability of the notion of lifelong compositionality. 

The discussion ties works together by categorizing them along six axes. Figure~\ref{fig:RelatedWork} includes a visual depiction of the landscape according to this categorization, while Appendix~\ref{app:RelatedWork} lists all cited references in terms of the same categories. The first axis divides works according to the learning setting: lifelong learning, MTL, and STL. The second axis analyzes each approach according to whether the environment provides the structure of the task to the agent and how. The third axis dissects works in terms of the underlying learning mechanism used by the agent for training: supervised, unsupervised, or RL. The fourth axis separates approaches according to the type of compositionality they study: functional composition, temporal composition, or no composition. The fifth axis classifies works with respect to how they structurally combine components: via chaining, aggregation, or a more general graph. The sixth and final axis divides works in terms of the application domain they consider.

There are other avenues of research that are related to aspects of this dissertation. To keep the discussion focused on the concepts of compositionality and lifelong learning, the chapters where they are most relevant summarize those separate lines of work. In particular, Chapter~\ref{cha:RL} discusses the relation between work on off-line RL and catastrophic forgetting, and Chapter~\ref{cha:NonStationary} discusses work on nonstationary lifelong learning to provide context.

\section{Lifelong or Continual Learning}
\label{sec:relatedLifelong}
In lifelong learning, agents must handle a variety of tasks over their lifetimes, and should accumulate knowledge in a way that enables them to more efficiently learn to solve new problems. \citet{thrun1998lifelong} first introduced the concept of lifelong learning, which has received widespread attention in recent years~\citep{chen2018lifelong}. Recent efforts have mainly focused on avoiding catastrophic forgetting~\citep{mccloskey1989catastrophic}. At a high level, existing approaches define parts of parametric models (e.g., deep neural networks) to share across tasks. As the agent encounters tasks sequentially, it strives to retain the knowledge that enabled it to solve earlier tasks. 

The following sections divide lifelong learning methods into task-aware (those that receive a task indicator) and task-agnostic (those that do not receive a task indicator). The discussion then ties this division back to the separation according to how models receive information about the structure of the tasks.

\subsection{Task-Aware}

The majority of works in lifelong learning in recent years fall in the category of task-aware methods. This section summarizes the existing works in this category. Note that, in most cases, the methods that can operate in the task-aware setting can also operate in the task-agnostic setting. In such cases, their categorization into task-aware or task-agnostic follows the (majority of the) experiments used in their original evaluation.

\subsubsection{Regularization}
One common approach to avoid forgetting is to impose data-driven regularization to prevent parameters from deviating in directions that are harmful to performance on the early tasks. The intuition is that similar parameters would lead to similar solutions to the earlier tasks. The canonical example of this idea is elastic weight consolidation (EWC; \citealt{kirkpatrick2017overcoming}), which, inspired by a Bayesian formulation, places a quadratic penalty on the parameters for deviating from the parameters of each previously seen task, weighted by the diagonal of the Fisher information matrix of each task: $\Omega(\btheta)=\sum_\that \Big(\btheta-\thetathat\Big)^\top\Fishert[\that]\Big(\btheta-\thetathat\Big)$. EWC is one of the few exceptional works that was applied to both supervised and RL settings. An extension of EWC uses a Kronecker-factored approximation of the Fisher information matrix instead of a diagonal to improve performance at little additional cost~\citep{ritter2018online}.

Following this same principle, the literature has proposed a variety of methods for computing the regularization terms. Departing from the Bayesian formulation, the approach of \citet{zenke2017continual} computes an estimate of each parameter's importance based on the trajectory of updates to the parameter, and uses this as the weighting for quadratic regularization. A generalized regularizer combines this latter idea with EWC~\citep{chaudhry2018riemannian}. Another well-known mechanism learns task-specific, (nearly-)binary attention masks to select which nodes in a neural net to use for each task, and then constrains updates to parameters based on the previous tasks' masks~\citep{serra2018overcoming}. A similar technique additively decomposes each task's parameters into a shared set of parameters modulated by a task-specific mask and a set of task-adaptive parameters, applying quadratic regularization to prevent previous tasks' parameters from diverging from their original values~\citep{yoon2020scalable}. Recent work proposed a sparsity-based regularizer that reduces the storage space for regularization terms by computing node-wise (as opposed to parameter-wise) importance weights~\citep{jung2020continual}, and applied this approach to RL. \citet{cha2021cpr} proposed a complementary regularizer based on entropy maximization, which encourages wider local minima and can therefore be used in combination with existing regularizers to avoid forgetting. 

While the approaches above were originally proposed and studied solely on vision applications, recent work has extended them to image captioning, demonstrating their applicability to language domains~\citep{delchiaro2020ratt}.

The methods described so far are based on the intuition that parameters that are important for previous tasks should be modified sparingly in order to avoid forgetting. Recent works have considered a different intuition: in order for new tasks to be learned without interfering with past tasks, they should lie on orthogonal subspaces of the parameter space. One mechanism for doing so is to precompute a set of task-specific orthogonal matrices and use them to project the feature space of each task~\citep{chaudhry2020continual}. Alternatively, it is also possible to compute such projection matrices sequentially based on the learned solutions to previous tasks. This can be achieved by exploiting the singular vector space of the activations of the network, which applies to linear and convolutional layers~\citep{saha2021gradient,deng2021flattening}, as well as recurrent layers~\citep{duncker2020organizing}.

Another regularization strategy that has become popular is online variational inference, which approximates the Kullback-Leibler (KL) divergence between the current and previous predictive distributions in a Bayesian setting. \citet{nguyen2018variational} developed the first variational continual learning (VCL) method, which, akin to EWC~\citep{kirkpatrick2017overcoming} in the standard regularization-based setting, requires storing penalty terms for each network parameter. In an effort to reduce storage requirements, \citet{ahn2019uncertainty} modified this method by storing penalty terms for each network node, akin to the work of \citet{jung2020continual}. A generalized variational objective adds a tunable hyperparameter to weight the KL divergence term in the objective function, encompassing EWC and VCL~\citep{loo2021generalized}. These notions can extend to Gaussian mixture distributions. In particular, \citet{zhang2021variational} developed a continual variational inference method using a Chinese restaurant process to automatically determine the number of latent components, while \citet{kumar2021bayesian} did the same using an Indian buffet process for unsupervised and supervised learning.

Other methods instead functionally regularize the outputs of the model directly, penalizing deviations from the predictions of earlier tasks' models on new data~\citep{li2017learning}. \citet{benjamin2018measuring} noted that distance in parameter space (as exploited by regularization-based methods) is not always representative of distance in function space (as advocated for by functional regularization methods), which is what should be preserved in continual learning. The authors subsequently showed that na\"ively storing a small set of samples to estimate function-space distance was sufficient to devise a strong functionally regularized continual learner. 
Other methods convert the learned network into a Gaussian process (GP) and either store a subset of samples from previous tasks to retain in memory~\citep{titsias2020functional,pan2020continual}. Then, the GP posterior on those points penalizes the model for making incorrect predictions on past tasks.

A recent effort used EWC in a different setting, in which the objective is to actively forget knowledge of past tasks that prevents learning new tasks, using only the Fisher information for the current task~\citep{wang2021afec}; this method was used in supervised and RL tasks.

\subsubsection{Replay}

A distinct approach retains a small buffer of data from all tasks, and continually updates the model parameters utilizing data from the current and previous tasks, thereby maintaining the knowledge required to solve the earlier tasks. Most na\"ively, one could simply iterate over past tasks' data when training on the new task~\citep{chaudhry2019tiny}. In the most common case, where the amount of data stored for replay is small, one would expect that the model could overfit to the tiny memory. However, empirical evaluations found that even this na\"ive replay approach performs surprisingly well. 

Other popular approaches use replay data to constrain the directions of gradient updates to regions of the parameter space that do not conflict with the earlier tasks' gradients~\citep{lopez2017gradient,chaudhry2018efficient}. These same techniques have used a meta-learning objective function, whereby the agent trains directly to optimize the network's feature representation to avoid future tasks' gradients from conflicting with previous tasks' gradients~\citep{riemer2018learning,gupta2020lookahead}. 

\citet{mirzadeh2021linear} developed a distinct objective function. The authors found that a linear curve in the objective function connects the solutions of the MTL and continual learning problems, and consequently used replay data to encourage finding a solution that is closer to the (no-forgetting) MTL solution. Similarly, \citet{raghavan2021formalizing} found theoretically that the balance between generalization and forgetting, viewed as a two-player zero-sum game, is stable and corresponds to a saddle point, and developed an algorithm that searches for this saddle point by playing the two-player game. 

Other works have not explicitly modified the objective function, but instead focused on other aspects of the problem. For example, one approach finds a balance between the loss terms corresponding to replay and current data via mixed stochastic gradients~\citep{guo2020improved}. As another example, \citet{pham2021contextual} used a dual memory to train a set of shared neural net layers and a task-specific controller to transform the features of the shared layers.

Despite the appeal of these more advanced replay-based methods, the basic method that simply replays randomly stored data remains a strong and popular baseline~\citep{chaudhry2019tiny}. In practice, replay-based techniques have proven to be much stronger at avoiding forgetting than regularization-based methods. One potential theoretical explanation for this discrepancy is that optimally solving the continual learning problem requires storing and reusing all past data~\citep{knoblauch2020optimal}.

Replay-based approaches have also followed other, less common directions. One nonparametric kernel method leverages the idea of episodic memories, using the memory for detecting the task at inference time instead of for replay~\citep{derakhshani2021kernel}. Other works have learned hypernetworks that take a task descriptor as input and output the parameters for a task-specific network, using replay at the task-descriptor level~\citep{oswald2020continual,henning2021posterior}. In the unsupervised setting, \citet{rostami2021lifelong} used replay to learn a Gaussian mixture model in a latent representation space such that all tasks map to the mixture distribution in the embedding space.

\subsubsection{Generative Replay}

A related technique is to learn a generative model to ``hallucinate'' replay data, potentially reducing the memory footprint by avoiding explicitly storing earlier tasks' data. For example, this can be achieved by training a generative adversarial network (GAN) and using the trained network to generate artificial data for the previous tasks to avoid forgetting~\citep{shin2017continual}. In one of the few rare works that has considered lifelong language learning, the authors leveraged the intuition that a language model itself is a generative model and used it for replaying its own data~\citep{sun2020lamol}. A recent unsupervised method for training GANs learns both global features that are kept fixed after the first task and task-specific transformations to those shared features~\citep{varshney2021camgan}. While the approach does not train the GANs themselves via generative replay, it can use the GANs to generate replay data to train a supervised method.

\subsubsection{Expandable Models}

While approaches described so far are capable of learning sequences of tasks without forgetting, they are limited by one fundamental constraint: the learning of many tasks eventually exhausts the capacity of the model, and it becomes impossible to learn new tasks without forgetting past tasks. Note that, while some replay and regularization approaches described so far add task-specific features that can be considered as a form of capacity expansion, this expansion is na\"ively executed for every task. Some additional methods use per-task growth as their primary mechanism for avoiding forgetting. One example specific to convolutional layers keeps the filter parameters fixed after initial training on a single task and adapts them to each new task via spatial and channel-wise calibration~\citep{singh2020calibrating}. However, these methods still consider the set of \textit{shared} features to be nonexpansive, and these shared features might still run out of capacity. As one potential exception, another technique leverages large pretrained language models that in practice seem to have sufficient capacity for a massive number of tasks (specifically BERT;~\citealp{devlin2019bert}) and adds small modules trained via task-specific masking to learn a sequence of language tasks~\citep{ke2021achieving}.

As a solution to the issue of limited capacity, a similar line of work has studied how to automatically expand the capacity of the model as needed to accommodate new tasks. \citet{yoon2018lifelong} did so via a multistage process that first selects the relevant parameters from past tasks to optimize, then checks the loss on the new task after training, and---if the loss exceeds a threshold---expands the capacity and trains the expanded network with group sparsity regularization to avoid excessive growth. In order to avoid forgetting, the algorithm measures the change in each neuron's input weights, and duplicates the neuron and retrains it if the change is too large. A similar approach sidesteps the need for this duplication step by maintaining all parameters for past tasks fixed~\citep{hung2019compacting}.

Figure~\ref{fig:RelatedWork} categorizes the vast majority of the approaches discussed in this section for task-aware lifelong learning as lifelong supervised learning methods with no composition, no task structure provided, and a focus on vision applications. A handful of exceptions were highlighted that deal with RL or unsupervised learning, as well as with language applications.

While these techniques notably require no information about the way in which tasks are related, they do require access to a task indicator both during learning and during evaluation. This choice enables these task-aware methods to use task-specific parameters to specialize shared knowledge to each individual task, but may be inapplicable in some settings where there are no evident task boundaries or there is no potential for supervision at the level of the task indicator. 

\subsection{Task-Agnostic}

As an alternative, task-agnostic lifelong approaches instead automatically detect tasks. While the techniques for learning these models are typically not fundamentally different from those of the task-aware setting, this survey categorizes them separately to highlight the conceptual difference of learning in the presence of implicit or explicit information about how tasks are related to each other. Note that these methods may or may not assume access to task indicators during \textit{training}, but they all are unaware of the task indicator during \textit{inference}. Since this is inconsequential to the point of the agent receiving information about task relations, the following discussion omits such distinctions.

\subsubsection{Regularization}

Like in the task-aware setting, a number of task-agnostic techniques aim at avoiding forgetting by penalizing deviations from earlier tasks' solutions. One early method uses a diagonal Gaussian approximation in order to obtain a closed-form update rule based on the variational free energy~\citep{zeno2018task}. A later extension of this method handles arbitrary Gaussian distributions by using fixed point equations~\citep{zeno2021task}. Another recent technique combines Kronecker-factored EWC~\citep{ritter2018online} with a novel projection method onto the trust region over the posterior of previous tasks~\citep{kao2021natural}. A distinct approach by \citet{kapoor2021variational} trains a variational GP using sparse sets of inducing points per task. \citet{joseph2020meta} used regularization at a meta level, learning a generative regularized hypernetwork using a variational autoencoder (VAE) to generate parameters for each task based on a task descriptor.  Whereas these methods have operated in the supervised setting, \citet{egorov2021boovae} recently proposed a VAE model with boosting density approximation for the unsupervised setting. 

Functional regularization also applies to the task-agnostic setting for avoiding forgetting. One such method relies on the lottery ticket hypothesis, which states that deep nets with random weight initialization contain much smaller subnetworks that can be trained from the same initialization and reach comparable performance to the original (much larger) network~\citep{frankle2018lottery}. \citet{chen2021long} extended this hypothesis to the lifelong setting by using a pruning and regrowing approach, in combination with functional regularization from unlabeled data from public sources. Another approach combines parameter regularization (specifically, EWC) with functional regularization to avoid forgetting in an approach based on weight and feature calibration~\citep{yin2021mitigating}.

\subsubsection{Replay}

Another popular mechanism to avoid forgetting in the task-agnostic setting is to replay past data stored in memory. One recent approach stores data points along with the network's output probabilities, and uses these to functionally regularize the output of the network to stay close to its past predictions~\citep{buzzega2020dark}. An additional method learns a dual network with different learning rates, training a fast learner to transform a slow learner's output pixel-wise and replaying past data to avoid forgetting~\citep{pham2021dualnet}.

A drastically different technique uses a graph learning approach to discover pairwise similarities between memory and current samples, penalizing forgetting the edges between samples instead of the predictions in order to maintain correlations between samples while still permitting significant changes to the network's representations~\citep{tang2021graphbased}. An extension to the technique of 
\citeauthor{gupta2020lookahead} (\citeyear{gupta2020lookahead}; from the task-aware setting) for lifelong learning via meta-learning  learns an additional binary mask that determines which parameters to learn for each task, leading to sparse gradients~\citep{oswald2021learning}. Note that this discussion categorizes this method as task-agnostic simply because the paper conducted the majority of experiments in that setting.

While these examples have focused on \textit{how} to leverage past examples during training, a related line of work has explored \textit{which} samples to store or replay from the earlier tasks. The method of \citet{aljundi2019gradient} stores samples whose parameter gradients are most diverse. Other work proposed to sample from memory the points whose predictions would be affected most negatively by parameter updates without replay~\citep{aljundi2019online}. \citet{chrysakis2020online} tackled the problem of class imbalance by developing a class-balancing sampling technique to store instances, in combination with a weighted replay strategy. One additional approach determines the optimal set of points to store in memory using bilevel optimization~\citep{borsos2020coresets}. 

Intuitively, it is possible that the samples seen during training are not the best to store in memory for lifelong training (e.g., because they lie far from the decision boundaries). \citet{jin2021gradientbased} leveraged this idea by directly modifying the samples in memory via gradient updates to make them more challenging for the learner. Alternatively, one could imagine that storing high-resolution samples might be wasteful, as many features might be superfluous for retaining performance on past tasks. Prior work has exploited this intuition by automatically compressing data in memory via a multilevel VAE that iteratively compresses samples to meet a fixed storage capacity~\citep{caccia2020onlinelearned}.

As with the rest of the methods, these replay-based approaches have all been used in the vision domain. One exception to this was the work of \citet{demasson2019episodic}, which directly applied memory-based parameter adaptation~\citep{sprechmann2018memorybased} with sparse replay to the language domain.

\subsubsection{Generative Replay}

\citet{achille2018life} developed a VAE-based approach to lifelong unsupervised learning of disentangled representations, which uses generative replay from the VAE to avoid forgetting. A similar technique uses a dynamically expandable mixture of Gaussians to identify when the unsupervised model needs to grow to accommodate new data~\citep{rao2019continual}. \citet{ayub2021eec} developed another unsupervised learning method based on neural style transformers~\citep{gatys2016image}. Unlike prior methods, this latter approach explicitly stores in memory the autogenerated samples in embedding space, and consolidates them into a centroid-covariance representation to maintain a fixed capacity. As unsupervised approaches, these three methods can seamlessly extend to the supervised setting, as demonstrated in the corresponding manuscripts. Alternatively, another approach specifically for supervised learning relies on three model components: a set of shared parameters, a dynamic parameter generator for classification layers on top of the shared parameters, and a data generator~\citep{hu2019overcoming}. The data generator serves a dual purpose: to generate embeddings for the dynamic parameter generator and to be used for replay via functional regularization of the shared parameters. Another similar approach for supervised learning, inspired by the brain, also replays hidden representations to avoid forgetting~\Citep{van2020brain}.

\subsubsection{Expandable Models}

In the vein of dynamically expandable models, \citet{aljundi2017expert} conceived the first task-agnostic method, which trains a separate expert for each task, and automatically routes each data point to the relevant expert at inference time. A distinct method automatically detects distribution shifts during training to meta-learn new components in a mixture of hierarchical Bayesian models~\citep{jerfel2019reconciling}. Similarly, the method of  \citet{lee2020neural} trains a dynamically expandable mixture of experts via variational inference.

Overall, task-agnostic learning might appear at first glance as an unqualified improvement over task-aware learning. However, it comes at the cost of one additional assumption: the task structure must be implicitly embedded in the features of each data point (e.g., one task might be daylight object detection and another nightlight object detection). In some settings, this assumption is not valid, for example if the same data point might correspond to different labels in different tasks (e.g., cat detection and dog detection from images with multiple animals). Figure~\ref{fig:RelatedWork} therefore categorizes these methods as requiring the task structure to be implicitly provided, primarily in the supervised and unsupervised settings, with applications to vision models. To reiterate, note that, in practice, many of the task-aware methods can operate in the task-agnostic setting with minor modifications, and vice versa. 

\subsection{Reusable Knowledge}

Lifelong approaches discussed so far, although effective in avoiding the problem of catastrophic forgetting, make no substantial effort toward the discovery of reusable knowledge. One could argue that these methods learn the model parameters in such a way that they are reusable across all tasks. However, it is unclear what the reusability of these parameters means, and moreover the architecture design hard-codes how to reuse parameters. This latter issue is a major drawback when attempting to learn tasks with a high degree of variability, as the exact form in which tasks connect to one another is often unknown. One would hope that the algorithm could determine these connections autonomously.

The ELLA framework introduced an alternative formulation based on dictionary learning~\citep{ruvolo2013ella}. The elements of the dictionary can be interpreted as a set of models that are reusable across tasks, and task-specific coefficients select how to reuse them. This represents a rudimentary form of functional composition, where each component is a full task model and the new models aggregate the component parameters.

A few other mechanisms identify which knowledge to transfer across tasks. One approach relies on automatically detecting which layers in a neural net should be specific to a task and which should leverage a shared set of parameters. Existing work has achieved this either via variational inference~\citep{adel2020continual} or via expectation maximization~\citep{lee2021sharing}. Another technique is to identify the most similar tasks by training one model via transfer and another via STL and comparing their validation performances~\citep{ke2020continual}. Once the agent has identified similar tasks, it uses an attention mechanism to transfer knowledge from only those tasks. An additional algorithm instead avoids explicitly selecting tasks or layers to transfer, and directly meta-learns a set of features that maximize reuse when task-specific parameters mask the shared weights for transfer~\citep{hurtado2021optimizing}. 

Going back to the illustration of Figure~\ref{fig:RelatedWork}, these methods have included applications to vision and have worked only in the supervised setting. As a sole exception, extensions of the work of \citet{ruvolo2013ella} have applied to RL, as discussed below in Section~\ref{sec:relatedLifelongRL}.  

In a distinct line of work, \citet{yoon2021federated} developed a knowledge-sharing mechanism for lifelong federated learning, selectively transferring knowledge from other clients. In the language domain, \citet{gupta2020neuraltopic} achieved lifelong transfer by sharing latent topics.

\subsection{Additional Approaches}

While the large majority of works on lifelong learning fall into the categories above, some exceptions do not fit this classification. For completeness, this section briefly describes some of the most recent such efforts. 

\citet{javed2019metalearning} developed an online meta-learning algorithm that explicitly trains a representation that avoids catastrophic forgetting. The problem setting studied in their work is distinct from the works described so far: instead of a single lifelong sequence of tasks, the agent faces a pretraining phase, during which it uses \textit{multiple} ``lifelong'' sequences of tasks to meta-learn the representation. A similar method learns a dual network for gating the outputs of a standard net~\citep{beaulieu2020learning}. Another related method extended the work of \citet{javed2019metalearning} to include a generative classifier~\citep{banayeeanzade2021generative}.

A different recent problem formulation is that of continual generalized zero-shot learning, which requires the agent to generalize to unseen tasks as well as perform well on all past tasks~\citep{skorokhodov2021class}. The authors then presented an algorithm for tackling the problem via class normalization.

Other works have instead focused on understanding aspects of existing lifelong learning approaches. \citet{mirzadeh2020understanding} empirically studied the impact of a variety of training hyperparameters (specifically dropout, learning rate decay, and mini-batch size) on the width of the obtained local minima---and therefore, on forgetting. This study led to the development of stable stochastic gradient descent, a now-popular baseline for benchmarking new lifelong approaches. Another study empirically evaluated the effect of task semantics on catastrophic forgetting, finding that intermediate similarity leads to the highest amount of forgetting~\citep{ramasesh2021anatomy}. \citet{lee2021continual} obtained a similar finding for lifelong learning specifically in the teacher-student setting, with additional insight separating task feature similarity and class similarity. A separate work evaluated existing lifelong learning methods on recurrent neural networks, finding that they perform reasonably well and are a solid starting point for the development of lifelong methods specific to recurrent architectures~\citep{ehret2021continual}.

Figure~\ref{fig:RelatedWork} categorizes these last few approaches  as lifelong supervised learning methods without any type of composition. Most of the methods either assume no information about the structure of the tasks (but assume access to a task indicator) or vice versa. The one exception is the work of \citet{skorokhodov2021class}, which assumes explicit task descriptors that enable zero-shot generalization, which equates to explicit information about the task structure. Similarly, all works considered solely vision applications, with the exception of \citet{ehret2021continual}, which additionally considered a simple audio application.

\section{Compositional Knowledge}
\label{sec:RelatedCompositional}
A mostly distinct line of parallel work has explored the learning of compositional knowledge. This section discusses existing methods for functional composition, while Section~\ref{sec:relatedCompositionalRL} discusses other forms of composition specifically used in the RL setting.

\subsection{Multitask Learning}
The majority of compositional learning methods either learn a set of components given a known structure for how to compose them, or learn the structure for piecing together a given set of components. In particular, in the former case, \citet{andreas2016neural} proposed to use neural modules as a means for transferring information across visual question answering (VQA) tasks. Their method parses the questions in natural language and manually transforms them into a neural architecture. Given this fixed architecture, the agent then learns modules for detecting shapes, colors, and spatial relations, and later combines the modules in novel ways to answer unseen questions. In this context, neural modules represent general-purpose, learnable, and composable functions, which permits thinking broadly about composition. Consequently, this dissertation used neural modules as the primary form of learnable components. A related work extended the neural programmer-interpreter (NPI;~\citealp{reed2016neural}) to learn an interpreter for the programming language Forth using neural modules as the primitive functions, given manually specified execution traces~\citep{bovsnjak2017programming}. Another similar study developed the intuition that, in order for neural modules to be composable, they must be invertible, and tested this intuition by manually composing these modules with themselves and other pretrained modules~\citep{wu2021improving}.

In the latter scenario of a given set of components, \citet{cai2017making} improved generalization in the NPI framework by incorporating recursion. Another approach based on programming languages uses RL for rewarding all semantically correct programs and additionally imposes syntactical correctness directly in the training procedure~\citep{bunel2018leveraging}. More advanced RL techniques have tackled the same problem, removing the need for any supervision in the form of annotated execution traces or structures~\citep{pierrot2019learning}. A separate approach  specifically for robot programming tasks uses the application programming interfaces (APIs) of primitive actions to guide the learning~\citep{xu2018neural}. Recently, similar ideas have achieved compositional generalization by directly learning rules over fixed symbols~\citep{nye2020learning} or by providing a curriculum~\citep{chen2020compositional}. In a related line of work, \citet{saqur2020multimodal} trained graph neural networks to couple concepts across different modalities (e.g., image and text), keeping the set of possible symbols fixed.

A more interesting case is when the agent knows neither the structure nor the set of components, and must autonomously discover the compositional structure underlying a set of tasks.  For example, following \citet{andreas2016neural}, several approaches for VQA assume that there exists a mapping from the natural language question to the neural structure and automatically learn this mapping~\citep{pahuja2019structure}. The majority of such methods assume access to a set of ground-truth program traces as a supervisory signal for learning the structures. The first such method simply learns a sequence-to-sequence model from text to network architectures in a supervised fashion~\citep{hu2017learning}. A similar method starts from supervised learning over a small annotated set of programs and subsequently fine-tunes the structure via RL~\citep{johnson2017inferring}. Some recent extensions to these ideas have included using probabilistic modules to parse open-domain text~\citep{gupta2020neuralmodule} and modulating the weights of convolutional layers with language-guided kernels~\citep{akula2021robust}.

Figure~\ref{fig:RelatedWork} categorizes the compositional works described so far as supervised MTL methods with explicitly given task structure, either in the form of fixed modules, fixed structures over the modules, or inputs that directly contain the structure (e.g., in natural language). The compositional structure is any arbitrary graph connecting the components, even allowing for components to be reused multiple times in a single task via recursion. Existing works have used these methods in varied application domains: toy programming tasks~\citep{bovsnjak2017programming,cai2017making,bunel2018leveraging,pierrot2019learning}, VQA~\citep{andreas2016neural,saqur2020multimodal,pahuja2019structure,hu2017learning,johnson2017inferring,gupta2020neuralmodule,akula2021robust}, natural language~\citep{nye2020learning,chen2020compositional}, vision~\citep{wu2021improving}, audio~\citep{wu2021improving}, and even robotics~\citep{xu2018neural}. 

However, some applications require agents (e.g., service robots) to learn more autonomously, without  any kind of supervision on the compositional structures.
Several approaches therefore learn this structure directly from optimization of a cost function. Many such methods assume that the inputs themselves implicitly contain information about their own structure, such as natural language tasks, and therefore use the inputs to determine the structure. One challenge in this setting is that the agent must autonomously discover, in an unsupervised manner, what is the compositional structure that underlies a set of tasks. One approach to this is to train both the structure and the model end-to-end, assuming that the selection over modules is differentiable (i.e., soft module selection;~\citealp{rahaman2021spatially}). Other approaches instead aim at discovering hard modular models, which increases the difficulty of the optimization process. Methods for tackling this variant of the problem have included using RL~\citep{chang2018automatically} or expectation maximization~\citep{kirsch2018modular} as the optimization tool. These ideas have operated on both vision~\citep{rahaman2021spatially,chang2018automatically} and natural language~\citep{kirsch2018modular} tasks.

Other approaches do not assume there is any information about the structure at all given to the agent, and it must therefore blindly search for it for every new task it learns. This often implies that the compositional structure for each task should be fixed across all data points, but often approaches permit reconfiguring the modular structure even \textit{within} a task. On the other hand, much like in the lifelong learning setting without compositional structures, this assumption also implies that the agent requires access to some sort of task indicator. One example of this formulation approximates an arbitrary ordering over a set of neural modules via soft ordering and trains the entire model end-to-end~\citep{meyerson2018beyond}. A related technique decomposes MTL architectures into tensors such that each matrix in the tensor corresponds to a subtask, using hypermodules (akin to hypernetworks) to generate local tensors~\citep{meyerson2019modular}. Another example assumes a hard module selection, and trains the modules via meta-learning so that they are able to quickly find solutions to new, unseen tasks~\citep{alet2018modular}. An extension of this method learns with graph neural networks~\citep{alet2019neural}, and a simplified version discovers whether modules should be task-specific or shared via Bayesian shrinkage~\citep{chen2020modular}. Another technique also learns a hard modular selection, but using RL to select the modules to use for each data point and task~\citep{rosenbaum2018routing}. One of the advantages of keeping the structural configuration fixed for each task (instead of input-dependent) is that the reduced flexibility protects the model from overfitting. This has enabled applying these latter methods to domains with smaller data sets than are typically available in language domains~\citep{meyerson2019modular,chen2020modular}, such as vision~\citep{rosenbaum2018routing,meyerson2018beyond,meyerson2019modular} and robotics~\citep{alet2018modular,alet2019neural}. 

\citet{rosenbaum2019routing} discussed the challenges of optimizing modular architectures with an extensive evaluation with and without task indicators in both vision and language tasks. 

\subsection{Lifelong Learning}
All compositional methods described so far assume that the agent has access to a large batch of tasks for MTL, enabling it to evaluate numerous combinations of components and structures on all tasks simultaneously. In a more realistic setting, the agent faces a sequence of tasks in a lifelong learning fashion. Most work in this line has assumed that the agent can fully learn each component by training on a single task, and then reuse the learned module for other tasks. One example is the NPI, which  assumes that the agent receives supervised module configurations for a set of tasks and can use this signal to learn a mapping from inputs to module configurations~\citep{reed2016neural}. Extensions to the NPI have operated in the MTL setting and were described in the previous section. Other methods do not assume that there is any information in the input about the task structure, and therefore must search for the structure for every new task. \citet{fernando2017pathnet} trained a set of neural modules and chose the paths for each new task using an evolutionary search strategy, applying this technique to both supervised learning and RL. The biggest downside of this technique is that the number of modules is constant, which, added to the fact that the algorithm keeps the weights of the modules fixed after training them on a single task, limits the applicability of the method to a small number of tasks. To alleviate these issues, other methods progressively add new modules, keeping existing modules fixed~\citep{li2019learn}. Some such approaches introduce heuristics for searching over the space of possible module configurations upon encountering a new task to improve efficiency, for example using programming languages techniques~\citep{valkov2018houdini} or data-driven heuristics~\citep{veniat2021efficient}. In the language domain, \citet{kim2019visual} developed an approach that progressively grows a modular architecture for solving a VQA task by providing a curriculum that directly imposes which module solves which subtask, keeping old modules fixed.

Unfortunately, this solution of keeping old modules fixed is infeasible in many real-world scenarios in which the agent has access to little data for each of the tasks, which would render these modules highly suboptimal. Therefore, other methods have permitted further updates to the model parameters. One early example, based on programming languages, simply assumed that future updates would not be harmful to previous tasks~\citep{gaunt2017differentiable}. This limited the applicability of the method to very simplistic settings. \citet{rajasegaran2019random} proposed a more complete approach that uses a combination of regularization and replay strategies to avoid catastrophic forgetting, but requires expensively storing and training multiple models for each task to select the best one before adapting the existing parameters, and is designed for a specific choice of architecture. Another approach routes each data point through a different path in the network, restricting updates to the path via EWC regularization if the new data point is different from past points routed through the same path~\citep{chen2020mitigating}. However, this latter approach heavily biases the obtained solution toward the first task, and does not permit the addition of new modules over time.

Unlike prior methods, the framework developed in this dissertation efficiently learns various forms of compositional structures in a lifelong learning setting and is easily extensible to a wider range of compositional structures. It does not assume access to a large batch of tasks or the ability to learn definitive components after training on a single task. Instead, it initializes components on the first few tasks, and then autonomously accommodates new tasks either by adapting the existing components or by creating new ones. Moreover, the framework applies to both the supervised and the RL settings.

Concurrently to this dissertation, and in particular after the development and publication of the general framework of Chapter~\ref{cha:Framework} and the supervised instantiations of Chapter~\ref{cha:Supervised}, a small number of works have also addressed the shortcomings of prior approaches. \citet{qin2021bns} developed a similar supervised learning approach which automatically grows and updates modules for each new task using an RL-based controller. However, unlike the approaches of Chapter~\ref{cha:Supervised}, which completely avoid forgetting in the structure over modules for each task by making them task-specific, their controller is susceptible to catastrophic forgetting. Another technique uses a local per-module selector that estimates whether each sample is in-distribution for the given module, and chooses the module with the highest value~\citep{ostapenko2021continual}. This mechanism lets this latter method operate in the task-agnostic setting and limits forgetting to local, per-module parameters. While this addresses a large part of the problem of forgetting in the module-selection stage, it enables earlier tasks to select new modules that are likely to malfunction in the presence of old data they did not train on. Notably, this method demonstrated the ability of existing modules to combine in novel ways to solve unseen tasks, exhibiting for the first time compositional generalization in the lifelong learning setting. This dissertation attained a similar result in lifelong RL.
	
The vast majority of the approaches described in this section assume an arbitrary graph structure over the components, and learn to construct paths through this graph. Concretely, in the case of neural modules, this means that each module can be used as input to any other module, or equivalently that modules can be chosen at any depth of the network. Some exceptions, particularly approaches that operate in the lifelong setting, impose a chaining structure by restricting certain modules to be eligible only at certain depths of the network. Note that both of these choices still contemplate an exponential number (in the network's depth) of possible configurations. However, the chaining approach does simplify the problem of learning modules, since it reduces the space of possible inputs and outputs that each module must learn. The framework developed in this dissertation is capable of learning with both these types of compositional structures, as well as simpler aggregated structures with no hierarchy. The experiments evaluated all these choices empirically.

\subsection{Nonmodular Compositional Works}

While modular neural architectures have become popular in recent years for addressing compositional problems, they are not the only solution. In particular, a number of works have dealt with the problem of compositionally generalizing to unseen textual tasks. In this setting, for example, the agent may have learned the concepts of ``walk'', ``twice'', and ``turn left'' in isolation, and later be required to parse an instruction like ``walk twice and turn left''~\citep{lake2018generalization}.

One approach uses meta-learning to explicitly optimize the agent to reason compositionally by generalizing to unseen combinations of language instructions~\citep{lake2019compositional}. Another method, inspired by the emergence of compositionality in human language, uses iterated learning on neural nets to compositionally generalize~\citep{ren2020compositional}. \citet{gordon2020permutation} equated language composition to equivariance on permutations over group actions, and designed an architecture that maintains such equivariances. A similar work imposed invariance to partial permutations on a language understanding system~\citep{guo2020hierarchical}. Another recent technique incorporates a memory of automatically extracted analytical expressions and uses those to compositionally generalize~\citep{liu2020compositional}. A distinct approach by \citet{akyurek2021learning} uses data augmentation to specifically target compositionality, combining prototypes of a generative model into multiprototype samples.

One method in this line of work operates in the lifelong setting, where the vocabulary of the agent grows over time~\citep{li2020compositional}. In this work, the agent separates the semantics and syntax of inputs, keeping the syntax for previously learned semantics parameters fixed and learning additional semantics parameters for each extension of the vocabulary.

The literature on visual object detection has also studied the idea of compositional generalization, under the vein of attribute-based zero-shot classification. At a high level, objects in images contain annotations not only of their class label but also of a set of attributes (e.g., color, shape, texture), and the learning system seeks to detect unseen classes based on their attributes. This requires the agent to learn the semantics of the attributes as well as how to combine them~\citep{huynh2020compositional,atzmon2020causal,ruis2021independent}.

In a similar direction, other approaches compose attributes to generate images. One such method learns one energy-based model per attribute that can later be combined with other attributes in novel combinations---e.g., to generate a smiling man from the attributes ``smiling'' and ``man''~\citep{du2020compositional}. Another approach learns embeddings of manual drawings that can later be composed into complex figures like flowcharts~\citep{aksan2020cose}. Similarly, the mechanism of \citet{arad2021compositional} uses GANs with structural priors to generate scenes by composing multiple objects.

While related, this line of work is farther from the approaches developed in this dissertation, and so a comprehensive overview is outside of the scope of this discussion. 

\subsection{Understanding Composition}

A recent line of work has sought  to understand various aspects of compositionality. An initial study defined a measure of the compositionality of a model as the ability to approximate the output of the model on compositional inputs by combining representational primitives~\citep{andreas2019measuring}. Using this measure, the authors evaluated a set of models and found a correlation (albeit small) between compositionality and generalization on vision and language tasks. A similar study found that the same definition of compositionality is related to zero-shot generalization on vision tasks~\citep{sylvain2020locality}. \citet{damario2021how} showed that explicitly modular (manually defined) neural architectures improve compositional generalization in VQA tasks. Somewhat contradictorily, \citet{agarwala2021one} found theoretically and empirically that a single, monolithic network is capable of learning multiple highly varied tasks. However, this ability requires an appropriate encoding of the tasks that separates them into clusters. One work used a similar intuition to develop a mechanism to compute a description of the execution trace of a modular architecture based on random matrix projections onto separate regions of an embedding space~\citep{ghazi2019recursive}. Given the apparent importance of modularity and compositionality, \citet{csordas2021are} studied two properties of neural nets without explicitly modular architectures: whether they automatically learn specialized modules, and whether they reuse those modules. While they found that neural nets indeed automatically learn highly specialized modules, unfortunately they do not automatically reuse those, thereby inhibiting compositional generalization.

\section{Lifelong Reinforcement Learning}
\label{sec:relatedLifelongRL}

The related works discussed so far primarily deal with supervised learning tasks. The number of approaches that operate in the lifelong RL setting is substantially more reduced. The following paragraphs describe some of the existing methods for lifelong RL, particularly in their relation to the compositional methods developed in this dissertation. 

Much like in the supervised setting, the majority of lifelong RL approaches rely on monolithic or nonmodular architectures, which as discussed in Section~\ref{sec:relatedLifelong} inhibits the discovery of self-contained and reusable knowledge. These methods mainly use regularization techniques for avoiding forgetting. A prominent example is EWC~\citep{kirkpatrick2017overcoming}, a supervised method that has been directly applied to RL, and which imposes a quadratic penalty for deviating from earlier tasks' parameters. One of the challenges of training RL models via EWC is that the vast exploration typically required to learn new RL tasks might be catastrophically damaging to the knowledge stored in the shared parameters. Consequently, an alternative approach first trains an auxiliary model for each new task and subsequently discards it and distills any new knowledge into the single shared model via an approximate version of EWC~\citep{schwarz2018progress}. While these methods in principle can handle task-agnostic settings, assuming that the input contains implicit cues about the task structure, in practice evaluations have tested them most often in the task-aware setting, typically in vision-based tasks (e.g., Atari games;~\citealp{bellemare2013arcade}). Moreover, these works have dealt with limited lifelong settings, with relatively short sequences of tasks and permitting the agent to revisit earlier tasks several times for additional experience. Even in these simplified settings, these methods have failed to achieve substantial transfer over an agent trained independently on each task, without any transfer.  
\citet{kaplanis2019policy} trained a similar monolithic architecture in the task-agnostic setting, including for tasks with continuous distribution shifts. Their approach regularizes the KL divergence of the policy to lie close to itself at different timescales, and was evaluated on simulated continuous control tasks.

Other approaches store experiences for future replay. The use of experience replay to retain performance on earlier tasks requires a number of special considerations in the RL setting. For example, the data collected over the agent's training on each individual task is nonstationary, since the behavior of the agent changes over time.  \citet{isele2018selective} proposed various techniques for selectively storing replay examples and evaluated the impact of these techniques empirically. Another challenge is that, as the agent modifies the policy for earlier tasks, the distribution of the data stored for them no longer matches the distribution imposed by the agent's policy. \citet{rolnick2019experience} proposed using an importance sampling mechanism for limiting the negative effects of this distributional shift. While the former example considered mostly grid-world-style tasks in the task-aware setting, where the input contains no information about the task relations, the latter considered vision-based tasks in the task-agnostic setting, under the assumption that the observation space for each task contains sufficient information for distinguishing it from others. However, the challenges of replay in RL have limited the applicability of these methods to short sequences of two or three tasks, still with the ability to revisit previous tasks. Chapter~\ref{cha:RL} establishes a connection between these issues and off-line RL, which this dissertation leveraged to develop a robust replay mechanism that operates on long sequences of tens of tasks without revisits. 

While most lifelong RL works have considered the use of a single monolithic structure for learning a sequence of tasks, some classical examples have instead followed the ELLA framework of \citet{ruvolo2013ella} to devise similar RL variants. PG-ELLA follows the dictionary-learning mechanics of ELLA, but replaces the supervised models that form ELLA's dictionary by policy factors~\citep{bouammar2014online}. An extension of this approach supports cross-domain transfer by projecting the dictionary onto domain-specific policy spaces~\citep{bouammar2015autonomous}. \citet{zhao2017tensor} followed a similar dictionary-learning formulation for deep nets, replacing all matrix operations with equivalent tensor operations. However, this latter method operates in the easier batch MTL setting. This again represents a rudimentary form of aggregated composition. The primary challenge that ELLA-based approaches face is that the dictionary-learning technique requires first discovering a policy for each task in isolation (i.e., ignoring any information from other tasks) to determine similarity to previous policies, before factoring the parameters to improve performance via transfer. The downside is that the agent does not benefit from prior experience during initial exploration, which is critical for efficient learning in lifelong RL. While these methods target continuous control tasks, their evaluations have considered the interleaved MTL setting, where the agent revisits tasks multiple times before evaluation.

The first approach for lifelong RL developed in this dissertation uses multiple models like the latter category, but it learns these models directly via RL training like the former class. This enables the method to be flexible and handle highly varied tasks while also benefiting from prior information during the learning process, thus accelerating the training. A similar approach in the context of model-based RL models the dynamics of the tasks via an aggregation of supervised models~\citep{nagabandi2018deep}, but the focus of that work was discovering when the agent faced new tasks in the absence of task indicators.

Other approaches instead use a completely separate model for each task. One such method leverages shared knowledge in the form of a metamodel that informs exploration strategies to task-specific models, resulting in linear growth of the model parameters~\citep{garcia2019meta}. Another popular example leverages shared knowledge in the form of lateral connections in a deep net, resulting in quadratic growth of the model parameters~\citep{rusu2016progressive}. Both these approaches are infeasible in the presence of large numbers of tasks.

A separate line of lifelong RL work has departed completely from the notion of tasks and has instead learned information about the environment in a self-guided way. The seminal approach in this area learns a collection of general value functions for a variety of signals and uses those as a knowledge representation of the environment~\citep{sutton2011horde}. A more recent approach learns latent skills that enable the agent to reset itself in the environment in a way that encourages exploration~\citep{xu2020continual}.

\section{Compositional Reinforcement Learning}
\label{sec:relatedCompositionalRL}

The most common form of composition studied in RL has been temporal composition. One influential work in this area is the options framework of \citet{sutton1999between}. At a high level, options represent temporally extended courses of actions, which can be thought of as skills. Once the agent has determined a suitable set of options, it can then learn a higher-level policy directly over the options. In the language used so far, each option represents a module or component, and the high-level policy is the structural configuration over modules. 

Traditional work in temporal composition has assumed that the environment provides the structure a priori as a fixed set of options (or information about how to learn each option, such as subgoal rewards). For example, the approach of \citet{lee2019composing} learns a policy for transitioning from one skill to the next, given a set of pretrained skills. However, other methods automatically discover both the modules and the configuration over them. One such method extends actor-critic methods to handle option discovery via an adaptation to the PG theorem~\citep{bacon2017option}. Recent work has developed mechanisms for skill chaining other than an explicit high-level policy, such as additively combining abstract skill embeddings~\citep{devin2019compositional} or multiplicatively combining policies~\citep{peng2019mcp}. 

The high expressive power of a policy over options enables learning arbitrary graph structures over the modules. However, these approaches have primarily been limited to toy applications, with some exceptions considering simple visual-based or continuous control tasks. 

Crucially, the problem considered in this dissertation differs in that the functional composition occurs at every time step, instead of the temporal chaining considered in the options literature. These two dimensions are orthogonal, and both capture real-world settings in which composition would greatly benefit the learning process of artificial agents.  Chapter~\ref{cha:RL} contains a deeper discussion of these connections. While in principle many of the techniques in Chapter~\ref{cha:RL} could also be applied to option learning, this dissertation left this line of work for future research. With this in mind, note that the discussion of related works on skill discovery, which is a vast literature on its own, is by no means comprehensive.
 
Other forms of hierarchical RL have considered learning state abstractions that enable the agent to more easily solve tasks~\citep{dayan1993feudal,dietterich2000hierarchical,vezhnevets2017feudal}. While these are also related, 
they have mainly focused on a two-layer abstraction. This represents a simple form of composition where the agent executes actions based on a learned abstracted state. Instead, general functional composition considers arbitrarily many layers of abstraction that help the learning of both state and action representations. 

The majority of works on both temporal composition and state abstractions have been in the STL setting, where the agent must simultaneously learn to solve the individual task and learn to decompose its knowledge into suitable components. In practice, this has implied that there is not much benefit of learning such a decomposition, since the learning itself becomes more costly. However, other investigations have considered learning such compositional structures for multiple tasks, in particular in the lifelong setting. \citet{brunskill2014pac} developed a theoretical framework which automatically discovers options and policies over options throughout a sequence of tasks. A more practical approach trains each option separately on a subtask, and later reuses these options for learning subsequent tasks~\citep{tessler2017deep}. A recent model-based approach learns skills in an off-line phase that subsequently enable the agent to learn in a nonstationary lifetime without explicit tasks~\citep{lu2021resetfree}. Other work studied state abstractions from a theoretical perspective in the lifelong setting~\citep{abel2018state}. When learning such compositional structures in a lifelong setting, the agent amortizes the cost of decomposing knowledge over the multiple tasks, yielding substantial benefits when the components capture knowledge that is useful in the future. 

Another form of composition studied in the RL literature has been to learn behaviors that solve different objectives and compose those behaviors to achieve combined objectives. \citet{todorov2009compositionality} showed that the linear composition of value functions is optimal in the case of linearly solvable Markov decision processes (MDPs). A similar result showed that successor features can be combined to solve this type of combined objectives~\citep{barreto2018transfer}. One common terminology for discussing how this process combines objectives is logical composition. Intuitively, if an agent has learned to solve objective $A$ and objective $B$ separately, it can then combine its behaviors to solve $A\ \mathrm{AND}\ B$ or $A\ \mathrm{OR}\ B$. This intuition has driven theoretical and empirical results in the setting of entropy-regularized RL~(\citealp{haarnoja2018composable}; \Citealp{van2019composing}). 
One approach in this setting explicitly modularizes the inputs to a neural net to handle each of the different goals, aided by multi-hot indicators of the active goals~\citep{colas2019curious}. This is similar to the approach of state decomposition used in this dissertation in Chapter~\ref{cha:RL}. \citet{nangue2020boolean} later formalized the intuition of logical composition in the lifelong setting. 
Other recent work developed this idea of composing multiple simultaneous behaviors specifically for robotic control~\citep{cheng2021rmpflow,li2021rmp2,bylard2021composable}.
A related line of work designed a formal language for specifying logically compositional tasks~\citep{jothimurugan2019composable}, and later used a similar language to learn hierarchical policies~\citep{jothimurugan2021compositional}. These compositional approaches require a specification of compositional objectives. Another related vein has sought to decompose the reward into such components, and learn separate policies for each component that can later be combined. These works have decomposed the reward manually ~\Citep{vanseijen2017hybrid} or automatically~\citep{lin2019distributional,lin2020rd2}. In practice, the way in which these logic-based approaches combine behaviors is typically a simple aggregation of value functions (e.g., weighted combination or addition), which limits their applicability to components that represent solutions to entire RL problems. In contrast, the more general functional composition proposed in this dissertation separates each policy itself into components, such that these components combine to form full policies.

A handful of works have considered functional composition in RL with modular neural nets. A first method handles a setting where each task is a combination of one robot and one task objective~\citep{devin2017learning}. Given prior knowledge of this compositional structure, the authors manually crafted chained modular architectures and trained the agent to learn the parameters of the neural modules. Other works have instead assumed no knowledge of the task structure and learned them autonomously, under the assumption that the inputs contain implicit cues of what distinguishes the modular structure of one task from another. In this line, one recent technique learns recurrent independent mechanisms by encouraging modules to become independent via a competition procedure, and combines the modules in general graph structures~\citep{mittal2020learning,goyal2021recurrent}. 
These methods were originally developed for the supervised setting and were directly applied to RL, and have primarily operated in the STL setting. Another closely related method also automatically learns a mapping from inputs to modular structures in the MTL setting, with applications to noncompositional robotic manipulation~\citep{yang2020multi}. 

Compositionality has had a long history in RL, given the promise that learning smaller, self-contained policies might make RL of complex tasks feasible. This has led to a wide diversity of ways to define composition. For completeness, this paragraph discusses other recent approaches to compositionality that have received less attention and bear less connection to the work presented here. As one example, \citet{pathak2019learning} sought to decompose policies via a graph neural network such that each node in the graph corresponds to a link in a modular robot. A later version of this work extended this idea by considering a setting where all links are morphologically equivalent in terms of their size and motor, and ensuring that all modules learn the same policy~\citep{huang2020one}. Others have learned object-centric embeddings in order to generalize to environments with different object configurations~\citep{li2020learning,mu2020refactoring}. \citet{li2021solving} developed an approach related to skill discovery, but instead of combining skills, the agent learns to solve progressively harder tasks by truncating demonstrated trajectories in an imitation learning setting, such that the starting state leads to a task solvable by the current agent. 

The understanding of the modularity of RL agents at a fundamental level has received very little attention. One exception has been the recent work of \citet{chang2021modularity}, which studied the modularity of credit assignment as the ability of an algorithm to learn mechanisms for choosing actions that can be modified independently of the mechanisms for choosing other actions. The conclusion of this study was that some single-step temporal difference methods are modular, but PG methods are not.

This dissertation formalized the problem of lifelong compositional RL in terms of a compositional problem graph. This formulation led to the design of a second lifelong RL approach, which differs from these existing compositional RL methods in that 1)~it operates in a lifelong learning setting, where tasks arrive sequentially and the agent may not gain further experience in a previous task after learning it, and 2)~it applies to tasks that are explicitly compositional at multiple hierarchical levels, enabling in-depth study of the functionality of each component.

\subsection{Benchmarking Compositional Reinforcement Learning}

The development of large-scale, standardized benchmarks was key to the acceleration of deep learning research (e.g., ImageNet; \citealp{deng2009imagenet}). Inspired by this, multiple attempts have  sought to construct equivalent benchmarks for deep RL research, leading to popularly used evaluation domains in both discrete-~\citep{bellemare2013arcade,vinyals2017starcraft} and continuous-action~\citep{brockman2016openai,tunyasuvunakool2020dmcontrol} settings.

While these benchmarks have promoted deep RL advancements, they are restricted to STL---i.e., they design each task to be learned in isolation. Consequently, work in multitask and lifelong RL has resorted to ad hoc evaluation settings, slowing down progress. Recent efforts have sought to bridge this gap by creating evaluation domains with multiple tasks that share a common structure that is (hopefully) transferable across the tasks. One example varied dynamical system parameters of continuous control tasks (e.g., gravity) to create multiple related tasks~\citep{henderson2017multitask}. Other work created a grid-world evaluation domain with tasks of progressive difficulty~\citep{chevalier-boisvert2018babyai}. In the continual learning setting, a recent benchmark evaluates approaches in a multiagent coordination setting~\citep{nekoei2021continuous}. Specifically in the context of robotics, recent works have created large sets of tasks  for evaluating MTL, lifelong learning, and meta-learning algorithms~\citep{yu2019meta,james2020rlbench, wolczyk2021continual}.

Despite this recent progress, it remains unclear exactly what an agent can transfer between tasks in these benchmarks, and so existing algorithms are typically limited to transferring neural net parameters in the hopes that they discover reusable information. Unfortunately, typical evaluations of compositional learning use such standard benchmarks in both the supervised and RL settings. While this enables fair performance comparisons, it fails to give insight into the agent's ability to find meaningful compositional structures.  Some notable exceptions exist for evaluating compositional generalization in supervised learning~\citep{bahdanau2018systematic,lake2018generalization,sinha2020evaluating,keysers2020measuring}. 

This dissertation extended the ideas of compositional generalization to RL, and introduced the separation of zero-shot compositional generalization and fast adaptation, which is particularly relevant in RL. To evaluate these notions, this dissertation further created various evaluation benchmarks of explicitly compositional tasks for evaluating compositional RL methods. In particular, Chapter~\ref{cha:Benchmark} introduces one benchmark comprising hundreds of highly diverse RL tasks with explicit functionally compositional structure. 

Concurrently to this dissertation, \citet{gur2021environment} developed a benchmark for \textit{temporal} (instead of functional) compositional generalization in RL, which is complementary to the benchmarks presented here. Another related work procedurally created robotics tasks by varying dynamical parameters to study causality in RL~\citep{ahmed2021causalworld}, but considered a single robot arm and continuous variations in the physical properties of objects.

\section{Summary}
This chapter reviewed the state of prior research on the topics most closely related to those studied in this dissertation, and categorized them along six dimensions. In summary, lifelong or continual learning has primarily focused on the problem of catastrophic forgetting in the supervised setting, but has mostly overlooked how to obtain knowledge that can be reusable for future tasks. On the other hand, compositional learning has developed methods for obtaining reusable knowledge, but has done so in the simpler case of MTL, where the agent trains on all tasks simultaneously. This dissertation combined these two lines of work by developing algorithms that discover reusable compositional knowledge in a lifelong setting. Few works have attempted to port lifelong learning techniques to the RL setting, and their shortcomings have prevented their application to complex and diverse sequences of tasks, which this dissertation overcame by leveraging lifelong composition. In particular, the form of functional composition studied here had been severely understudied in the RL literature. 

\biblio

\chapter{A General-Purpose Framework for Lifelong Learning of Compositional Structures}
\label{cha:Framework}

\section{Introduction}
Despite their intuitive connections, lifelong learning and compositional learning have largely proceeded as disjoint lines of work. This chapter describes the concrete problem formulations for both lifelong learning and compositional learning as studied in this dissertation. At a high level, lifelong learning is the problem of accumulating knowledge over time and reusing it to solve related tasks, while compositional learning is the problem of decomposing knowledge into maximally reusable components. To address the joint problem of lifelong compositional learning, this chapter further presents the first general-purpose framework for lifelong learning of compositional structures. 

Unlike the handful of lifelong compositional learning methods in previous work, this framework is agnostic to the form of structures learned by the agent, the methods used to discover the structures, and the mechanisms used for avoiding catastrophic forgetting. As examples of the structures that the framework supports, this chapter discusses linear combinations of models (a form of aggregated composition, according to the categorization in Chapter~\ref{cha:RelatedWork}), soft neural layer ordering and soft neural gating (two approximate forms of arbitrary graph compositions), and hard modular neural nets (a form of chained composition).  

In order to learn these structures, the framework separates the learning process into two distinct stages. First, the learner leverages components that it has already acquired to discover how to maximally reuse them for solving the current task. Then, once the agent has assimilated the current task, it accommodates any new knowledge required to solve the current task either by adapting those existing components or by adding novel components if needed. These stages evoke Piaget's assimilation and accommodation stages of cognitive development~\citep{piaget1976piaget}, and so the stages of the framework adopt those terms.

\section{The Lifelong Learning Problem} 
\label{sec:lifelongLearningProblemAddendum}
At the highest level, lifelong learning involves learning over a nonstationary and potentially never-ending stream of data. From this high-level definition, different works have proposed multiple concrete instantiations of the problem. This section dissects common problem formulations in the literature, defines the problem as was addressed throughout this dissertation, and provides example problems that can be captured under this definition.

\Citet{van2019three} categorized lifelong learning problem definitions in terms of how nonstationarity is presented to the agent, proposing the following three variations:

\begin{itemize}
    \item The most common problem definition, denoted \textbf{task-incremental learning}, introduces nonstationarity into the learning problem in the form of \textit{tasks}. Each task $\Task^{(t)}$ is itself a standard \iid~learning problem, with its own input space $\Inputt$ and output space $\Outputt$. There exists a ground-truth mapping $\ft:\Inputt \mapsto \Outputt$ that defines the individual task, as well as a cost function $\Losst\Big(\hat{\f}^{(t)}\Big)$ that measures how well a learned $\hat{\f}^{(t)}$ matches the true $\ft$ under the task's data distribution $\mathcal{D}^{(t)}\Big(\Inputt, \Outputt\Big)$. During the learning process, the agent faces a sequence of tasks $\Taskt[1], \ldots, \Taskt[t], \ldots$. The learner receives a data set $\bm{X}^{(t)}, \bm{Y}^{(t)} \sim \mathcal{D}^{(t)}\Big(\Inputt, \Outputt\Big)$ along with a task indicator $t$ that reveals which is the current task, but not how it relates to other tasks. Upon facing the $t$-th task, the goal of the learner is to solve (an online approximation of) the MTL objective: $z=\frac{1}{t}\sum_{\that=1}^{t}\Losst[\that]\Big(\hat{\f}^{(\that)}_t\Big)$, where $\hat{\f}^{(\that)}_t$ is the predictor for task $\Taskt[\that]$ at time $t$.
    
    \item Another common definition is \textbf{domain-incremental learning}. The key distinguishing factor of this setting is that the learning problem does not inform the agent of the task indicator $t$. Instead, the tasks vary only in their input distribution $\mathcal{D}^{(t)}\Big(\Inputt\Big)$, but there exists a single common solution that solves all tasks. For example, the problem could be a binary classification problem between ``cat'' and ``dog'', and each different task could be a variation in the input domain (e.g., changing light conditions or camera resolutions). The goal of the learner is still to optimize the approximate MTL objective. 
    
    \item Yet another problem setting is \textbf{class-incremental learning}. In this setting, there is a single multiclass classification task with a large number of classes. The agent observes classes sequentially, and must be able to predict the correct class among all previously seen classes. For example, the task could be ImageNet~\citep{deng2009imagenet} classification, and classes could be presented to the agent ten at a time, with later stages not containing previous classes in the \textit{training} data, but indeed requiring accurate prediction across all seen classes in the \textit{test} data. One alternative way to define this same problem is that each learning stage (i.e., each group of classes) is a distinct task, and the goal of the agent is to simultaneously detect which is the current task indicator $t$ and which is the current class within that task $\Taskt$. The latter equivalent formulation, though less intuitive, enables framing the learning objective in exactly the same way as the previous two problem settings.
\end{itemize}

This dissertation considered the task-incremental learning setting. Concretely, a vector $\thetat$ parameterizes each task's solution, such that $\ft=\f_{\thetat}$. After training on $\numTasks$ tasks, the goal of the lifelong learner is to find parameters $\thetat[1],\ldots,\thetat[\numTasks]$ that minimize the cost across all tasks: $\frac{1}{\numTasks}\sum_{t=1}^{\numTasks} \Losst\Big(\ft\Big)$. The agent does not know the total number of tasks, the order in which tasks will arrive, or (unless otherwise stated) how tasks are related to each other. 

Given limited data for each new task, typically insufficient for obtaining optimal performance without leveraging information from prior tasks, the agent must strive to discover any relevant information to 1)~relate it to previously stored knowledge in order to permit transfer and 2)~store any new knowledge for future reuse. The environment may require the agent to perform any previous task, implying that it must perform well on \textit{all} tasks. In consequence, the agent must strive to retain knowledge from even the earliest tasks.

While prior work in supervised learning has described this task-incremental setting as artificial, it contains some desirable properties that are missing from other common definitions. First, unlike in domain-incremental learning, it is not necessary that the inputs are the only aspect that changes over time. This is useful, for example, when extending these definitions to the RL setting, where different tasks naturally correspond to different reward functions or transition dynamics. Second, unlike in class-incremental learning, extension to RL is straightforward, since the learning objective can still easily be averaged across tasks. 

In the task-incremental supervised setting, the standard way to craft the different tasks for benchmarking purposes originates from the class-incremental setting: split each data set into multiple smaller tasks, each containing a subset of the classes. This is the setting that was used for  most supervised learning experiments in Chapter~\ref{cha:Supervised}. As an example, CUB-200, a data set of $200$ classes corresponding to bird species, was split randomly into $20$ individual $10$-way classification tasks to evaluate lifelong agents. However, to show that this is not the only possible setting that the proposed methods can handle, the experiments also evaluated the proposed methods in a more complex setting with tasks from various distinct data sets (concretely, MNIST, Fashion MNIST, and CUB-200).

Note that all the above definitions assume that the agent is required to perform well on all previously seen tasks. However, this is often not realistic. For example, consider a service robot that has provided assistance for a long time in a small one-floor apartment, and is later moved to a much larger two-floor home. While knowledge from the small apartment may be useful for quickly adapting to the larger home, over time retaining full knowledge about how to traverse the small apartment might become counterproductive as that information becomes obsolete. Therefore, this dissertation also proposed adaptations of the developed techniques to the nonstationary lifelong learning problem, where not only the \emph{data} distribution changes from task to task, but also the distribution over \emph{tasks} itself changes from time to time (like in the small-apartment-to-large-home example). In this setting, the objective must change, since it is no longer desirable to retain performance on tasks from out-dated distributions. Chapter~\ref{cha:NonStationary} discusses more appropriate objectives for these settings.

\section{The Compositional Learning Problem} 
\label{sec:compositionalLearningProblemAddendum}

Section~\ref{sec:lifelongLearningProblemAddendum} describes the lifelong learning problem in terms of how nonstationarity is presented to the agent. However, it does not provide insight about how different tasks might be related to each other. In particular, this dissertation assumed that tasks are \textit{compositionally} related, and developed methods that explicitly exploit these compositional assumptions. 

Following the problem formulation from \citet{chang2018automatically}, this dissertation assumed that each task can be decomposed into subtasks. Equivalently, the predictive function $\ft$ characterizing each task can be decomposed into multiple subfunctions $\subproblemti{1}, \subproblemti{2}, \ldots$, such that $\ft = \subproblemti{1}\circ\subproblemti{2}\circ\cdots(\inputvec)$. This assumption trivially holds for any function $\ft$. Critically, the formulation further assumes that there exists a set of $k$ subfunctions that are common to all tasks the agent might encounter: $\subproblemti{i}\in\{\subproblemi[1], \ldots, \subproblemi[k]\} \,\,\forall t,i$.

This way, the full learning problem can be characterized by a directed graph $\mathcal{G}=(\mathcal{V},\mathcal{E})$. There are two types of nodes in the graph. The first type represents the inputs and outputs of each task as random variables. Concretely, each task has an input node $u^{(t)}$ with in-degree zero and an output node $v^{(t)}$ with out-degree zero such that $u^{(t)},v^{(t)}\sim\mathcal{D}^{(t)}\Big(\Inputt, \Outputt\Big)$. The second type of nodes $\subproblem$ represents functional transformations such that:
\begin{enumerate}
    \item for every edge $\langle u, \subproblem \rangle$ the function $\subproblem$ takes as input the random variable $u$,
    \item for every edge $\langle \subproblem, \subproblem^\prime\rangle$ the output of $\subproblem$ feeds into $\subproblem^\prime$, and
    \item for every edge $\langle \subproblem, v \rangle$ $v$ is the output of $\subproblem$.
\end{enumerate}
With this definition, the paths in the graph from $u^{(t)}$ to $v^{(t)}$ represent all possible solutions to task $\Taskt$ given a set of functional nodes.

This formalism also has an equivalent generative formulation. In particular, a compositional function graph $\mathcal{G}$ generates a task $\Taskt$ by choosing one input node $u^{(t)}$ and a path $p^{(t)}$ through the graph to some node $v^{(t)}$. Then, the following two steps define the generative distribution for task $\Taskt$. First, instantiate the random variable $u^{(t)}$ by sampling from the input distribution  $u^{(t)}=x^{(t)}\sim \mathcal{D}^{(t)}$. Next, generate the corresponding labels $y^{(t)}$ by compositionally applying all functions in the chosen path $p^{(t)}$ to the sampled $u^{(t)}$. As noted by \citet{chang2018automatically}, there are generally multiple possible compositional solutions to each task. This dissertation assumed that the generative problem graph is that with the minimum number of possible nodes, such that nodes (i.e., subtasks) are maximally shared across different tasks. This choice intuitively implies the maximum amount of possible knowledge transfer across tasks.

\begin{figure}[b!]
\centering
    \begin{subfigure}[b]{0.31\textwidth}
    \centering
        \includegraphics[scale=0.3, trim={8.3cm 6.2cm 5.5cm 5.5cm}, clip]{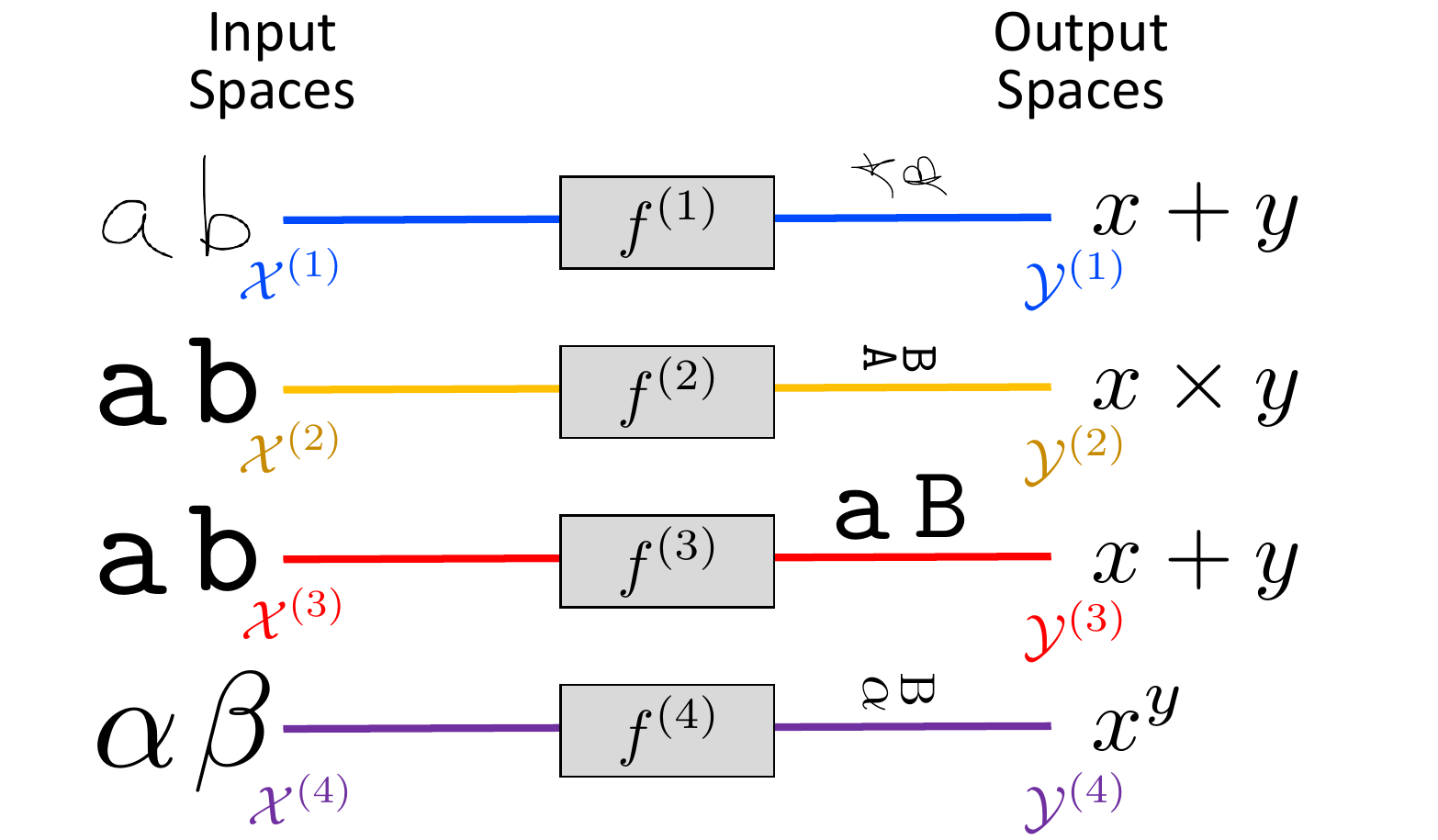}
        \caption{Single-task}
        \label{fig:compositionalFunctionGraphSTL}
    \end{subfigure}%
    \begin{subfigure}[b]{0.31\textwidth}
    \centering
        \includegraphics[scale=0.3, trim={8cm 6.5cm 5.5cm 5.5cm}, clip]{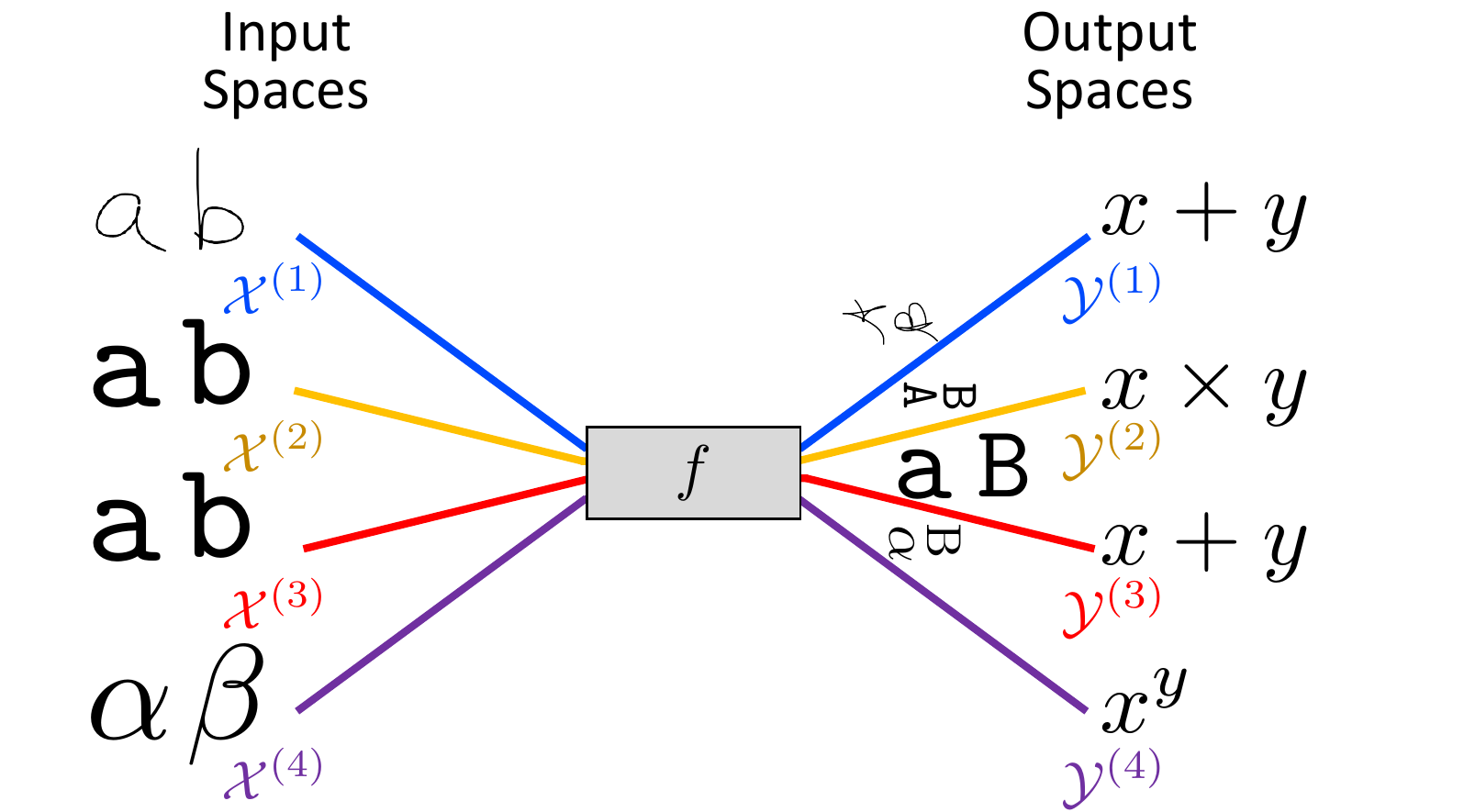}
        \caption{Multitask}
        \label{fig:compositionalFunctionGraphMTL}
    \end{subfigure}%
    \begin{subfigure}[b]{0.38\textwidth}
    \centering
        \includegraphics[scale=0.3, trim={5cm 6cm 5cm 5.5cm}, clip]{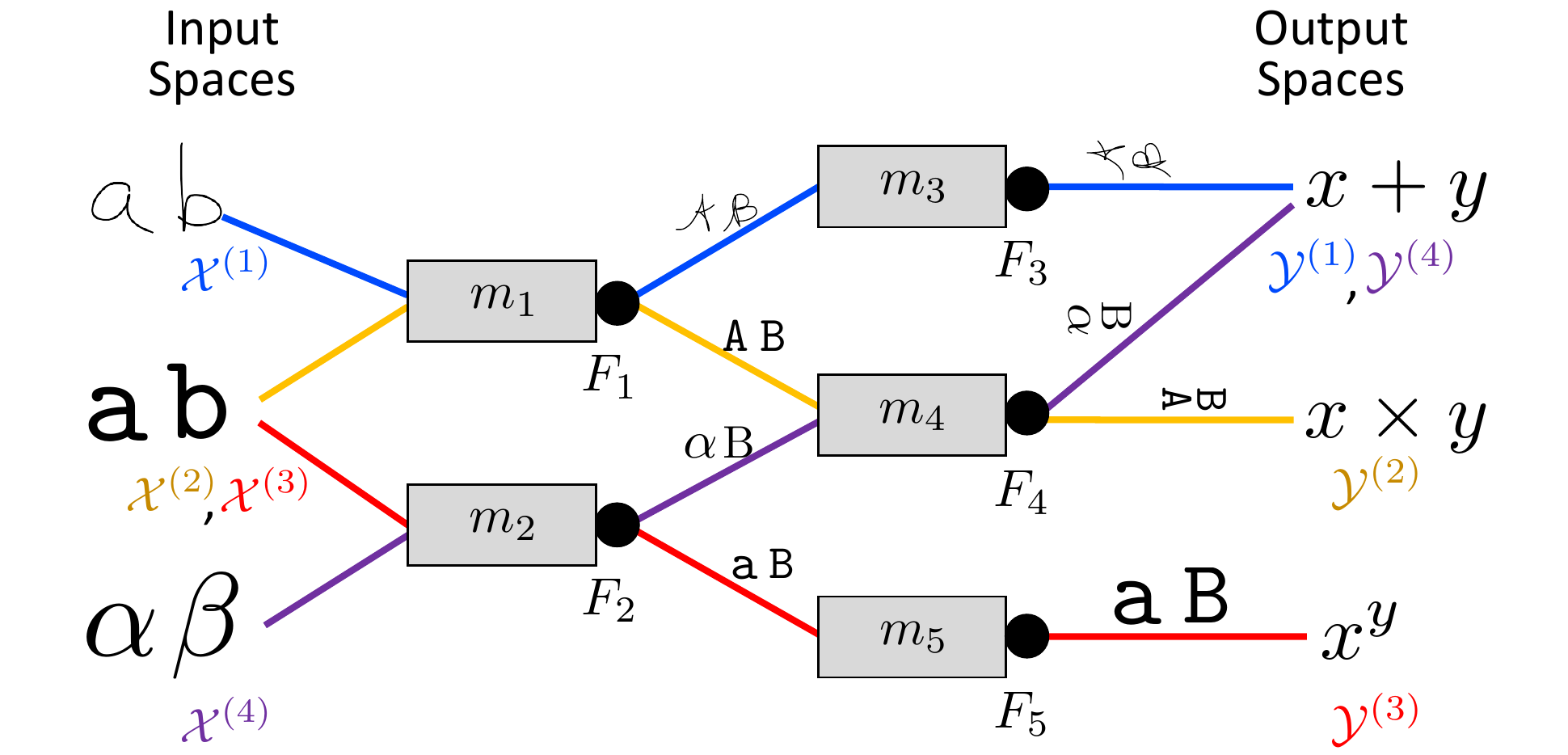}
        \caption{Compositional multitask}
        \label{fig:compositionalFunctionGraphComp}
    \end{subfigure}%
    \caption[Compositional problem graph.]{Compositional problem graph. Each node in the graph represents a random variable for a representational space, produced by the output of a module or function. STL agents assume that tasks are unrelated and learn modules in isolation, while monolithic MTL agents assume that all tasks share a single module. In contrast, more general compositional MTL agents assume that tasks selectively share a set of modules, yielding different solutions to each task constructed from common solutions to subtasks.}
    \label{fig:compositionalFunctionGraph}
    \vspace{-1em}
\end{figure}

Figure~\ref{fig:compositionalFunctionGraph} shows three different assumptions that learning algorithms make over the space of tasks. The left-most graph (Figure~\ref{fig:compositionalFunctionGraphSTL}) shows the standard STL assumption: each task $\Taskt$ is completely independent from the others, and therefore the agent learns the predictive functions $\ft$ in isolation. Note that this doesn't explicitly prohibit learning compositional solutions: each $\ft$ could itself be decomposed into multiple subtasks, but the subtasks would still be individual to each task. The center graph (Figure~\ref{fig:compositionalFunctionGraphMTL}) shows the typical monolithic MTL assumption: all different tasks can be decomposed in such a way that all subtasks are common to all tasks. The right-most graph (Figure~\ref{fig:compositionalFunctionGraphComp}) shows the assumption made throughout this dissertation: each task can be decomposed into a task-specific sequence of subtasks, but the set of possible subtasks is common to all tasks.

As a first example that matches the latter formulation, consider the following set of tasks:
\begin{itemize}
    \item $\Taskt[1]$: count the number of cats in an image
    \item $\Taskt[2]$: locate the largest cat in an image
    \item $\Taskt[3]$: locate the largest dog in an image
    \item $\Taskt[4]$: count the number of dogs in an image
\end{itemize}
These tasks can be decomposed into: detect cats, detect dogs, locate largest, and count. If an agent learns tasks $\Taskt[1]$, $\Taskt[2]$, and $\Taskt[3]$, and along the way discovers generalizable solutions to each of the four subtasks, then solving $\Taskt[4]$ would simply involve reusing the solutions to the dog detector and the general counter.

Consider another example from the language domain. In particular, the set of tasks may be:
\begin{itemize}
    \item $\Taskt[1]$: translate text from English to Spanish
    \item $\Taskt[2]$: translate text from Spanish to Italian
    \item $\Taskt[3]$: translate text from English to Italian
\end{itemize}
As above, if the learner has learned to solve tasks $\Taskt[1]$ and $\Taskt[2]$, it could solve task $\Taskt[3]$ by first translating the text from English to Spanish and subsequently translating the resulting text from Spanish to Italian. Here, Spanish would act as a \textit{pivot language}.

This definition can be applied to RL problems as well. Consider the following task components in a robotic manipulation setting:
\begin{itemize}
    \item Robot manipulator: diverse robotic arms with different dynamics and kinematic configurations can be used to solve each task.
    \item Objective: each task might have a different objective, like placing an object on a shelf or throwing it in the trash.
    \item Obstacle: various obstacles may impede the robot's actions, such as a door frame the robot needs to go through or a wall the robot needs to circumvent.
    \item Object: different objects require different grasping strategies.
\end{itemize}
One way to solve each robot-objective-obstacle-object task is to decompose it into subtasks: grasp the object, avoid the obstacle, reach the objective, and drive the robot's joints. Chapter~\ref{cha:Benchmark} instantiates these ideas into a compositional evaluation benchmark for RL. Note that this is not equivalent to the temporal composition of skills or options. Instead, each time step requires solving all subtasks simultaneously (e.g., the actions must be tailored to the current robot arm at all times). A detailed description of the problem formulation for the RL case, along with an in-depth comparison to temporal composition, is given in Chapter~\ref{cha:RL}.

The purpose of the empirical evaluations in this dissertation was two-fold: to demonstrate that compositional solutions arise in a range of interesting, realistic problems, and to show that the proposed algorithms can find such compositional solutions in complex problems. With these objectives in mind, the experiments evaluated supervised learning methods on standard benchmark tasks that are not explicitly compositional, while they evaluated RL methods on explicitly compositional tasks. The latter enabled a more in-depth analysis of the modularity of the learned solutions.

\section{The Lifelong Compositional Learning Framework}
\label{sec:Framework}
This section describes the main contribution developed in this dissertation: a general-purpose framework for lifelong learning of compositional structures. The framework admits a variety of forms of modules (e.g., linear factors and deep neural net modules), compositional structures (e.g., soft weighting and hard selection), base learning algorithms (e.g., stochastic gradient descent and PG learning), and knowledge retention mechanisms (e.g., elastic weight consolidation and experience replay). To demonstrate the flexibility of the framework, the remainder of this dissertation presents multiple algorithmic instantiations of it for the supervised and RL settings, along with extensive empirical evaluations of each method.

The framework  stores knowledge in a set of $\numModules$ shared components \mbox{$\Modules = \{\modulei[1], \ldots, \modulei[\numModules]\}$} that are acquired and refined over the agent's lifetime. Each component or module $\modulei = \module_{\ModuleParamsi} \in \ModuleSpace$ is a self-contained, reusable function parameterized by $\ModuleParamsi$ that can be combined with other components. The agent reconstructs each task's predictive function $\ft$ via a task-specific structure $\structuret: \Inputt \times \ModuleSpace^{\numModules} \mapsto \F$, with $\ModuleSpace^{\numModules}$ being the set of possible collections of $\numModules$ components, such that $\ft(\inputvec) = \structuret(\inputvec, \Modules)(\inputvec)$, where $\structuret$ is parameterized by a vector $\StructureParamst$. Note that $\structuret$ yields a function from $\F$. The structure functions select the modules from $\Modules$ and the order in which to compose them to construct the model for each task (the $\ft$'s).

\subsection{Example Compositional Structures}
\label{sec:CompositionalStructures}

The following paragraphs describe specific examples of components and structures that can be learned within the proposed framework. Each example structure is accompanied by a description of how it relates to the graph in Figure~\ref{fig:compositionalFunctionGraphComp}.

\begin{figure}[b!]
\centering
    \begin{subfigure}[b]{0.33\textwidth}
        \centering
        \raisebox{4em}{\includegraphics[scale=0.7, trim={0 0 0 0}, clip]{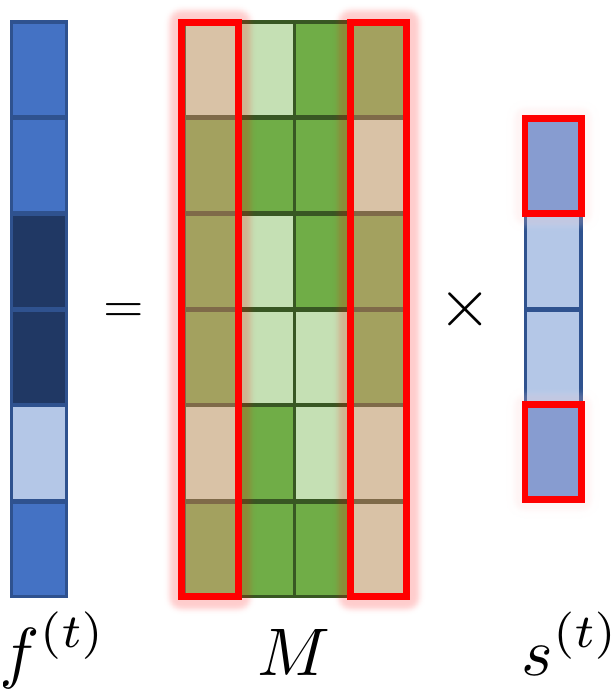}}
        \caption{Linear model combinations}
        \label{fig:linearModelCombination}
    \end{subfigure}%
    \hfill
    \begin{subfigure}[b]{0.33\textwidth}
        \centering
        \includegraphics[scale=0.7, trim={0 0 0 0}, clip]{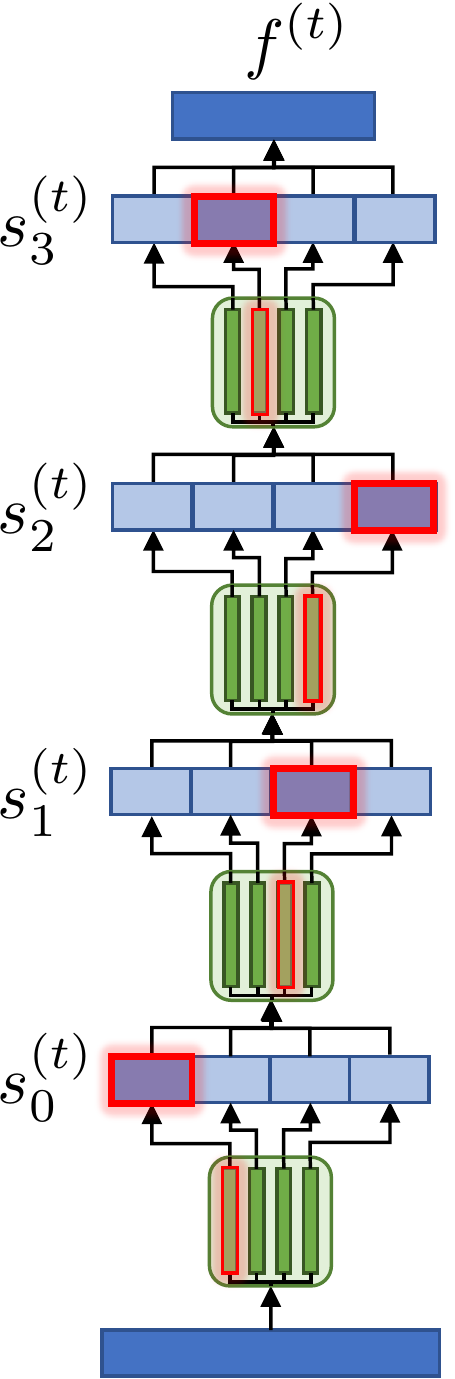}
        \caption{Soft layer ordering}
        \label{fig:softOrdering}
    \end{subfigure}%
    \hfill
    \begin{subfigure}[b]{0.33\textwidth}
        \centering
        \includegraphics[scale=0.7, trim={0 0 0 0}, clip]{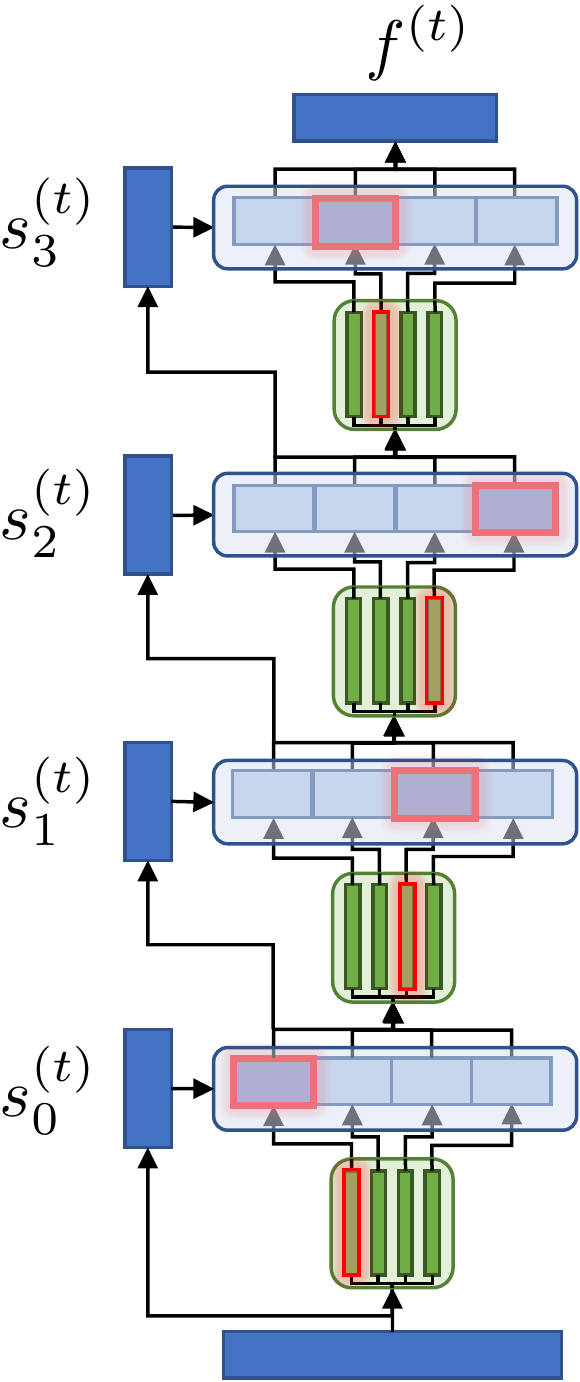}
        \caption{Soft gating}
        \label{fig:softGating}
    \end{subfigure}%
    \caption[Compositional structure examples.]{Compositional structure examples.}
    \label{fig:compositionalStructures}
\end{figure}

\paragraph{Linear combinations of models} In the simplest setting, depicted in Figure~\ref{fig:linearModelCombination}, each component is a linear model, and they are composed via linear combinations. Specifically, assume that $\Inputt\subseteq \Reals^{d}$, and that each task-specific function is given by $\f_{\thetat}(\inputvec)={\thetat}^{\top}\inputvec$, with $\thetat\in\Reals^{d}$. The model constructs each predictive function from a set of linear modules $\module_{\ModuleParamsi}(\inputvec)=\ModuleParamsi^\top\inputvec$, with $\ModuleParamsi\in\Reals^{d}$, by linearly combining them via a task-specific weight vector $\StructureParamst \in \Reals^{k}$, yielding:
\begin{equation}
    \ft(\inputvec)=\structure_{\StructureParamst}(\inputvec, \Modules)(\inputvec)={\StructureParamst}^{\top}\big(\ModuleParamsMatrix^\top\inputvec\big)\enspace,
\end{equation}
where the matrix  $\ModuleParamsMatrix=[\ModuleParamsi[1],\ldots,\ModuleParamsi[\numModules]]$ collects all $\numModules$ components. This corresponds to a crude approximation to the compositional function graph, where each task function $\ft$ contains a single subtask solution $f_i$, approximated by softly combining all the $f_i$'s. This dissertation used this formulation in both supervised and RL evaluations; prior lifelong learning works had also used this formulation~\citep{ruvolo2013ella,bouammar2014online,bouammar2015autonomous,isele2016using,rostami2020using}. Note that there is no strict requirement that the models must be linear in order to linearly combine them. However, the linear combination of nonlinear models is no longer equivalent to the linear combination of their model parameters. Yet, as shown in Chapter~\ref{cha:RL}, linear parameter combinations accept efficient optimization in the lifelong setting. Critically, these techniques can also be used to train nonlinear models (e.g., deep neural networks), at the cost of making the approximation to the compositional problem graph even cruder.

\paragraph{Soft layer ordering}
In order to handle more complex models, compositional deep nets  compute each layer's output as a linear combination of the outputs of multiple modules. As proposed by \citet{meyerson2018beyond}, this model assumes that each module is one layer, the number of components matches the network's depth, and all components share the input and output dimensions. Concretely, each component is a deep net layer $\module_{\ModuleParamsi}(\inputvec)=\nonLinearity\big(\ModuleParamsi^\top\inputvec\big)$, where $\nonLinearity$ is any nonlinear activation and $\ModuleParamsi\in\Reals^{\tilde{d}\times\tilde{d}}$. Each task network contains input $\inputTransformt$ and output $\outputTransformt$ transformations such that \mbox{$\inputTransformt:\Inputt\mapsto\Reals^{\tilde{d}}$} and $\outputTransformt: \Reals^{\tilde{d}} \mapsto \Outputt$. A set of parameters $\StructureParamst\in\Reals^{k\times k}$  weights the output of the components at each depth: 
\begin{equation}
    \structuret = \outputTransformt \circ \sum_{i=1}^{\numModules} \StructureParamsti{i,1}\modulei \circ \cdots \circ\sum_{i=1}^{\numModules}\StructureParamsti{i,k}\modulei \circ \inputTransformt\enspace,
\end{equation}
where the weights are restricted to sum to one at each depth $j$ to approximate a hard selection: $\sum_{i=1}^{k}\StructureParamsti{i,j} = 1$. This architecture, illustrated in Figure~\ref{fig:softOrdering}, is a soft approximation of the compositional problem graph, where the number of possible subtasks is $k$. This dissertation used this structure in the supervised learning evaluations. 

\paragraph{Soft gating}
In the presence of large data, it is often beneficial to modify the architecture for each input $\inputvec$~\citep{rosenbaum2018routing,kirsch2018modular}, unlike both approaches above which use a constant structure for each task. Consequently, the next architecture modifies the soft layer ordering architecture by weighting each component's output at depth $j$ by an input-dependent soft gating net ${\structureti{j}:\Inputt \mapsto \Reals^{k}}$, giving a predictive function:
\begin{equation}
    \structuret =\outputTransformt \circ \sum_{i=1}^{\numModules} \Big[\structureti{1}(\inputvec)\Big]_{i}\modulei \circ \cdots \circ \sum_{i=1}^{\numModules}\Big[\structureti{k}(\inputvec)\Big]_{i}\modulei \circ \inputTransformt\enspace.
\end{equation}
As above, the weights are restricted to sum to one at each depth: ${\sum_{i=1}^{k}\Big[\structureti{j}(\inputvec)\Big]_{i} = 1}$. This structure, visualized in Figure~\ref{fig:softGating}, still represents a soft approximation to the compositional problem graph, with the difference that the agent chooses the path through the graph dynamically based on the current input. Consequently, the evaluations assessed this architecture in the supervised setting, too.

\paragraph{Hard modular nets} The three example structures described so far resort to soft combinations of modules to approximate compositional solutions. The primary reason for doing this is that it permits optimizing the models directly via gradient-based training. However, this intuitively comes at the cost of yielding less differentiated and self-contained modules. As an alternative, one can in principle replace the soft weighting scheme by a hard selection mechanism and use discrete optimization techniques to find a solution. Examples for how to do this have included expectation-maximization methods~\citep{kirsch2018modular}, reinforcement learning~\citep{rosenbaum2018routing,chang2018automatically}, and explicit discrete search~\citep{alet2018modular}. Chapter~\ref{cha:RL} describes an instantiation of a hard modular net for RL, where the choice of modules is specific to each layer (unlike the soft ordering and soft gating structures) and domain knowledge is used to decompose the input into module-specific components.

\subsection{Stages of Lifelong Compositional Learning}
\label{sec:StagesOfFramework}

The intuition behind the proposed framework is that, at any point in time $t$, the agent has acquired a set of components suitable for solving tasks it encountered previously---$\Taskt[1], \ldots, \Taskt[t-1]$. If these components, with minor adaptations, can be combined to solve the current task $\Taskt[t]$, then the agent should first learn how to reuse these components before making any modifications to them. The rationale for this idea of keeping components fixed during the early stages of training on the current task $\Taskt[t]$, before the agent has acquired sufficient knowledge to perform well on $\Taskt[t]$, is that premature modification could be catastrophically damaging to the set of existing components. Once the agent has learned the structure $\structuret$, the framework considers that it has captured sufficient knowledge about the current task, and it would be sensible to update the components to better accommodate that knowledge. If, instead, it is not possible to capture the current task with the existing components, then new components should be added.  These notions loosely mirror the stages of assimilation and accommodation in Piaget's~(\citeyear{piaget1976piaget}) theories of intellectual development, and so the stages of the framework adopt those terms. Algorithms under this framework take the form of Algorithm~\ref{alg:LifelongCompositionalLearning}, split into the following steps, illustrated in Figure~\ref{fig:lifelongLearningAgent}.

\begin{algorithm}[t] 
  \caption{Lifelong Compositional Learning}
  \label{alg:LifelongCompositionalLearning}
\begin{algorithmic} 
    \STATE Initialize components $\Modules$ for reusability across tasks
    \WHILE {$\,\,\Taskt\gets\mathtt{getTask}()$}
        \FOR{$\,\,i = 1,\ldots,\,\mathtt{structureUpdates}$, keeping $\Modules$ fixed}
            \STATE Take assimilation step on structure parameters $\StructureParamst$ to find optimal structure $\structuret$
            \IF{$\,\,i \!\mod \mathtt{adaptationFrequency} = 0$}
                \FOR{$\,\,j = 1,\ldots,\,\mathtt{componentUpdates}$, keeping $\structuret$ fixed and unfixing $\Modules$}
                    \STATE Take adaptation step on module parameters $\ModuleParams$ to add new knowledge into $\Modules$
                \ENDFOR
            \ENDIF
        \ENDFOR
        \STATE Conditionally add new components via expansion
        \STATE Store any necessary information for future adaptation
    \ENDWHILE
\end{algorithmic}
\end{algorithm}

\begin{figure}[t!]
    \centering
    \includegraphics[width=0.68\textwidth, trim={4.7cm 9.9cm 4.65cm 10.0cm}, clip]{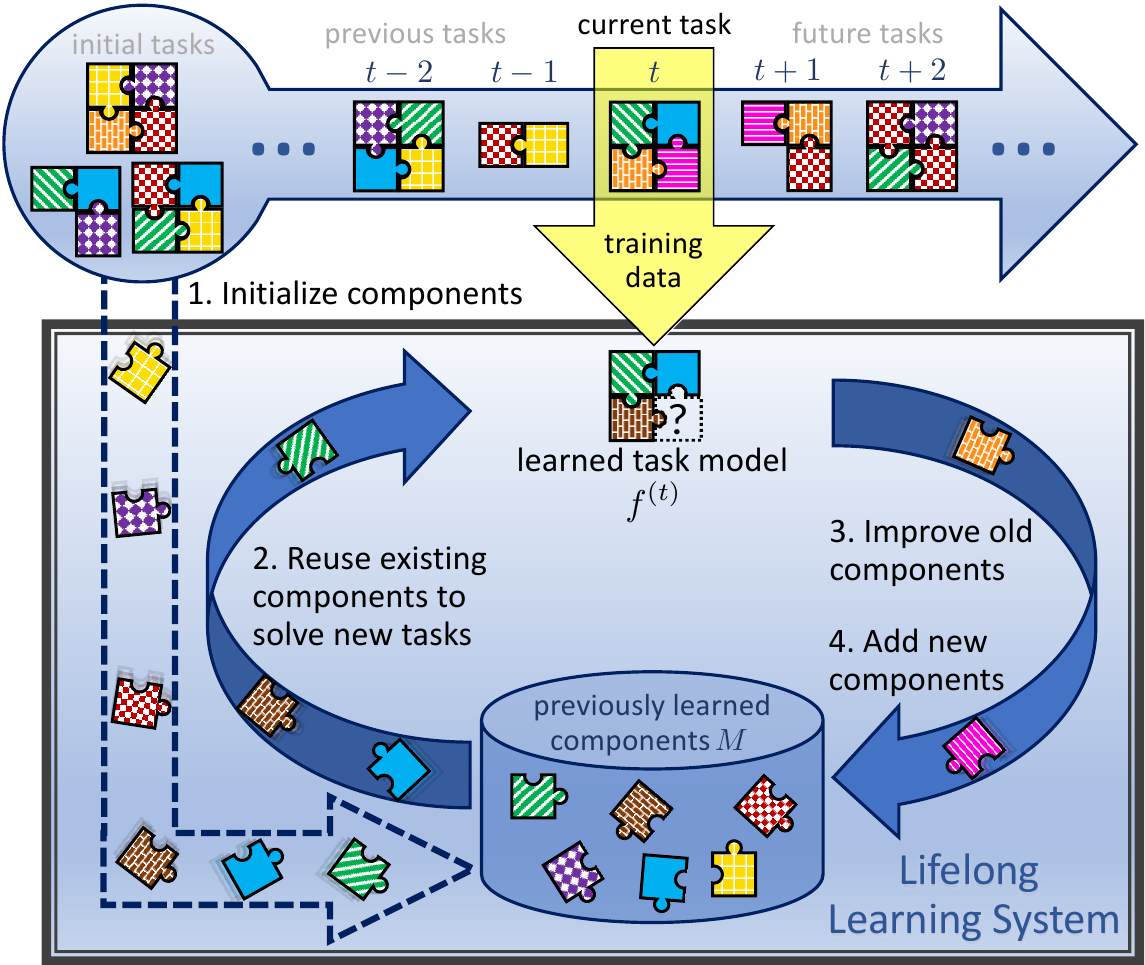}
    \caption[Lifelong compositional learner.]{Lifelong compositional learner. (1)~The framework for lifelong compositional learning initializes a set of components by training on a small set of tasks. (2)~The agent learns each new task by composing the relevant components. Subsequently, the it (3)~improves imperfect components with newly discovered knowledge, and (4)~adds any new components that were discovered during training on the current task.}
    \label{fig:lifelongLearningAgent}
\end{figure}

\paragraph{Initialization} The components $M$ should be initialized encouraging reusability, both across tasks and within different structural configurations of task models. The former signifies that the components should solve a particular
subproblem regardless of the objective of the task. The latter means that the structure for a single task's model may reuse components multiple times, or that different tasks may use them at different structural orders. For example, in deep nets, this means that the components could be used at different depths---of course, this is only relevant for architectures that permit selecting a module at various depths, unlike the hard modular nets described in the previous section. In the simplest case of linear module combinations, one simple example for how to do this is to incrementally train new modules on each new task until the capacity of $\Modules$ is met, reusing earlier components \textit{and} the new module upon training on the new task. In the case of deep modular nets, one could train the first few tasks the agent encounters jointly (in a small batch MTL setting) to initialize $M$, keeping a fixed (but random) structure that reuses components to encourage reusability. Alternatively, one could instead impose a curriculum over tasks with known compositional structures, such that the modules trained on the first few tasks are disjoint and therefore become specialized. This dissertation used these three examples as building blocks to develop the various algorithms under the framework.

\paragraph{Assimilation} The framework assimilates the current task by keeping the components $\Modules$ fixed and learning only the structure $\structuret$.  Existing algorithms for finding compositional knowledge vary in how they optimize each task's structure. In modular nets,  component selection can be learned via RL~\citep{johnson2017inferring,rosenbaum2018routing,chang2018automatically,pahuja2019structure}, stochastic search~\citep{fernando2017pathnet,alet2018modular}, or backpropagation~\citep{shazeer2017outrageously,kirsch2018modular,meyerson2018beyond}. The framework uses any of these mechanisms for assimilation. Approaches supported by the framework must accept decoupling the learning of the structure from the learning of the components themselves; this requirement holds for all the above examples.

\paragraph{Accommodation} An effective approach should maintain performance on earlier tasks, yet be flexible to incorporate new knowledge. To accommodate new knowledge from the current task, the learner may {\em adapt} existing components or {\em expand} to include new components: 
\begin{itemize}
\item {\em Adaptation step}\hspace{2em}Once the agent has obtained a suitable structure over modules, the framework incorporates new knowledge into the selected modules, keeping the structure fixed. Approaches for incorporating new knowledge into noncompositional structures have been to na\"ively  fine-tune models with data from the current task, to impose regularization to selectively freeze weights~\citep{kirkpatrick2017overcoming, ritter2018online}, or to store a portion of data from previous tasks and use experience replay~\citep{lopez2017gradient,isele2018selective}. Note that the last two approaches avoid catastrophically forgetting knowledge of how to solve earlier tasks. The framework can be instantiated by using any of these methods to accommodate new knowledge into existing components after assimilating the current task. For this to be possible, the method must be able to be selectively applied to only the component parameters $\ModuleParams$. 
\item {\em Expansion step}\hspace{2em}Often, existing components, even with some adaptation, are insufficient to solve the current task. In this case, the learner would incorporate novel components, which should encode knowledge distinct from existing components and combine with those components to solve the new task. The ability to discover new components endows the learner with the flexibility required to learn over a lifetime. Unlike for prior stages, there are no widespread mechanisms for automatically expanding the set of modules in a compositional architecture over time. Consequently, this dissertation created \textit{component dropout}, described in Chapter~\ref{cha:Supervised}, as a technique for assessing the effect of adding new modules without the need for storing and training multiple separate models. 
\end{itemize}

Chapter~\ref{cha:Supervised} presents concrete instantiations of Algorithm~\ref{alg:LifelongCompositionalLearning} for the supervised setting and a corresponding evaluation, while Chapter~\ref{cha:RL} includes the corresponding description of algorithms and evaluation for the RL setting. 

\section{Summary}
This chapter formalized the problem of lifelong compositional learning studied throughout this dissertation. At a high level, a lifelong learner is faced with a sequence of learning tasks, each of which is characterized by a compositional function. The goal of the agent is to find the set of shared modules or components that make up the functional solutions to each task, in such a way that there is maximum sharing across the different tasks. 

Next, this chapter described a general-purpose framework for lifelong compositional learning. This constitutes the central contribution of this dissertation. The key insight of the framework is that in order to learn meaningful knowledge decompositions sequentially, it is useful to split the learning process into phases, such that in the early stages of learning a new task, the agent does not need to worry about retaining knowledge of past tasks and can instead focus solely on acquiring knowledge of the current task. Then, once the agent has acquired sufficient knowledge about the current task, it may seek to combine this newly obtained knowledge with any knowledge from the earlier tasks to consolidate it all into the shared set of components. These stages have connections to Piagetian theory of development, and therefore they are adequately denoted assimilation and accommodation per Piagetian terms. These connections open the door for future investigations that bridge between lifelong learning and developmental psychology. 

The proposed framework is simple conceptually, and as shown in Chapter~\ref{cha:Supervised}, it is easy to combine with existing continual or compositional learning techniques, and effective in trading off the flexibility and stability required for lifelong learning. The framework further permits developing novel algorithms, even in the RL setting, as shown in Chapter~\ref{cha:RL}.

\biblio

\chapter{Application of Lifelong Composition to Supervised Learning}
\label{cha:Supervised}

\section{Introduction}
Chapter~\ref{cha:Framework} introduced the problems of lifelong and compositional learning, and additionally proposed a general-purpose framework for the joint problem of lifelong compositional learning, which this dissertation adopted to develop a multitude of algorithms. This chapter presents the first set of such algorithms, specifically for the supervised learning setting. The framework splits the learning process into broad stages: initialization, assimilation, and accommodation with its own adaptation and expansion substages. In particular, these stages are agnostic to the specific form of the compositional structures used for learning and to the base lifelong learning methods used for adaptation of the components with new knowledge. 

This chapter first presents a simple mechanism for determining when to expand the set of components when those components correspond to neural modules. This technique, based on dropout regularization, permits training and comparing networks with and without new components to assess the relative benefits of expanding after training on each new task. 

The exposition then turns to introduce nine concrete algorithms developed under the framework. These methods combine variants of existing modular structures---linear combinations of models, soft layer ordering, and soft gating---and lifelong learning mechanisms for avoiding catastrophic forgetting---na\"ive fine-tuning, elastic weight consolidation, and experience replay. The chapter then describes the derivations of computational complexity bounds for each of the nine resulting algorithms, and the obtained costs demonstrate that methods under the framework are more efficient at training than existing approaches.

The remainder of the chapter is devoted to an extensive empirical evaluation to validate the design choices and demonstrate the power of compositional knowledge representations for lifelong learning and of the two-stage mechanism dictated by the proposed framework. To this end, the primary baselines used for comparison consider training noncompositional models or compositional models without two-stage training. The results across a range of data sets of varying complexity demonstrate that methods under the framework consistently outperform competing methods. It becomes apparent that the main driver of this improved performance is the reduced amount of forgetting suffered by these approaches, which itself stems from the design choice to maintain the set of shared components fixed throughout most of the training (while the agent is still assimilating the current task).

\section{Expansion of the Set of Components \texorpdfstring{$\Modules$}{M} via Component Dropout}
\label{sec:ComponentDropout}
To enable deep compositional learners to discover new components, this dissertation created an expansion step where the agent considers adding a single new component per task. In order to assess the benefit of the new component, the agent learns two different networks: with and without the novel component. Dropout enables training multiple neural networks without additional storage~\citep{hinton2012improving}, and prior work has used it to prune neural net nodes in noncompositional settings~\citep{gomez2019learning}. The proposed dropout strategy deterministically alternates backpropagation steps with and without the new component, which is denoted in this chapter {\em component dropout}. Intermittently bypassing the new component ensures that existing components can compensate for it if it is discarded. After training, the agent applies a post hoc criterion (in the experiments herein, a validation error check) to potentially prune the new component.

\section{Framework Instantiations for the Supervised Setting}
\label{sec:FrameworkInstantiationsSupervised}

The experiments evaluated the framework with the three examples of soft compositional structures of Chapter~\ref{cha:Framework}: linear combinations of models, soft layer ordering, and soft gating. All methods assimilate task $\Taskt$ via backpropagation on the structure's parameters $\StructureParamst$; the training considers the input $\inputTransformt$ and output $\outputTransformt$ transformations as part of each task's structure, and so they are only updated during assimilation. The experiments trained each model structure with three instantiations of Algorithm~\ref{alg:LifelongCompositionalLearning}, varying the adaptation method: 
\begin{itemize}
\item \textit{Na\"ive fine-tuning}~(\VAN) updates components via backpropagation, ignoring past tasks. 

\item \textit{Kronecker-factored elastic weight consolidation}~(\kEWC; \citealp{ritter2018online}) penalizes modifying model parameters via $\frac{\lambda}{2}\!\sum_{t=1}^{T-1}\!\|\btheta - \thetat\|_{\Fishert}^2$, where $\Fishert$ is the Fisher information around $\thetat$, approximated with Kronecker factors. Adaptation steps carry out backpropagation on the regularized loss.  

\item \textit{Experience replay}~(\ER) stores $n_m$ samples per task in a replay buffer, and during adaptation takes backpropagation steps with both replay and current-task data. 
\end{itemize}
The evaluation explored two variations for each adaptation method, with and without the expansion step: \textbf{dynamic + compositional} methods use component dropout to add new modules, while {\bf compositional} methods keep a fixed-size set. 

For simplicity, the evaluation fixed the values of $\mathtt{structureUpdates}$, $\mathtt{adaptationFrequency}$, and $\mathtt{componentUpdates}$  such that the learning process would be split into multiple epochs of assimilation followed by a single final epoch of adaptation. This is the most extreme separation of the learning into assimilation and accommodation: the agent accommodates no knowledge into existing components until after assimilation has finished. Section~\ref{sec:ResultsStandardBenchmarks} studies the effects of this choice.  

Algorithms~\ref{alg:CompositionalVAN}--\ref{alg:DynamicER} summarize the implementations of all framework instantiations used in the experiments. Algorithms~\ref{alg:Initialization}--\ref{alg:AssimilationDynamic} contain all shared subroutines, and blank lines in compositional methods highlight missing steps from their dynamic + compositional counterparts. The learner first initializes the components by jointly training on the first $\numTasks_{\mathtt{init}}$ tasks it encounters. At every subsequent time $t$, during the first $(\mathtt{numEpochs}-1)$ epochs, the agent assimilates the new task by training the task-specific structure parameters $\StructureParamst$ via backpropagation, learning how to combine existing components for task $\Taskt$. For dynamic + compositional methods, this assimilation step incorporates component dropout and simultaneously optimizes the parameters of the newly added component $\ModuleParamsi[k+1]$. The adaptation step varies according to the base lifelong learning method, applying techniques for avoiding forgetting to the whole set of component parameters $\ModuleParamsMatrix$ for one epoch. This step also incorporates component dropout for dynamic + compositional methods. Finally, dynamic + compositional methods discard the fresh component if it does not improve performance by more than a threshold $\tau$ on the current task $\Taskt$, and otherwise keep it for future training. 

\subsection{Shared Subroutines for Lifelong Compositional Algorithms}
\vspace{-1.75em}
\begin{minipage}[t]{0.49\linewidth}
\begin{algorithm}[H] 
\renewcommand\algorithmicthen{}
\renewcommand\algorithmicdo{}
  \caption{Initialization}
  \label{alg:Initialization}
\begin{algorithmic}[1] 
    \STATE $\structuret\gets\mathtt{randomInitialization}()$
    \STATE $\mathtt{init\_buff}\gets \mathtt{init\_buff} \,\,\cup\,\, \Taskt.\mathtt{train}$%
    \IF {$\,\,t=\numTasks_\mathtt{init}-1$}
        \FOR{$\,\,i = 1,\ldots,\,\mathtt{numEpochs
        }$}
            \FOR {$\,\,\that, \inputvec \gets \mathtt{init\_buff}$}
                \STATE $\ModuleParamsMatrix \gets \ModuleParamsMatrix - \eta\nabla_{\ModuleParamsMatrix}\Losst[\that]\Big(\ft[\that](\inputvec)\Big)$
            \ENDFOR
        \ENDFOR \COMMENT{backprop on components}
    \ENDIF
\end{algorithmic}
\end{algorithm}
\end{minipage}
\hfill
\begin{minipage}[t]{0.49\linewidth}
\begin{algorithm}[H] 
\renewcommand\algorithmicthen{}
\renewcommand\algorithmicdo{}
  \caption{Expansion}
  \label{alg:Expansion}
\begin{algorithmic}[1] 
    \STATE $a_1\gets$ $\mathtt{accuracy}\Big(\Taskt.\mathtt{validation}\Big)$
    \STATE $\mathtt{hideComponent}(k+1)$
    \STATE $a_2\gets \mathtt{accuracy}\Big(\Taskt.\mathtt{validation}\Big)$
    \STATE $\mathtt{recoverComponent}(k+1)$
    \IF {$\,\,\frac{a_1-a_2}{a_2} < \tau$ \COMMENT{validation error check}}
        \STATE $\mathtt{discardComponent}(k+1)$
        \STATE $k \gets k+1$
    \ENDIF
\end{algorithmic}
\end{algorithm}
\end{minipage}

\begin{minipage}[t]{0.49\linewidth}
\begin{algorithm}[H] 
\renewcommand\algorithmicthen{}
\renewcommand\algorithmicdo{}
  \caption{Assimilation 
    (Comp.)}
  \label{alg:AssimilationCompositional}
\begin{algorithmic}[1] 
    \STATEx {\color{white}  $\ModuleParamsMatrix \gets [\ModuleParamsMatrix; \mathtt{randomVector}()]$ new comp.}
    \STATEx {\color{white} $\StructureParamsti{k+1,1:k} \gets 1$}
    \FOR{$\,\,i = 1,\ldots,\,\mathtt{numEpochs}-1$}
        \FOR {$\,\,\inputvec \gets \Taskt.\mathtt{train}$}
            \STATE $\StructureParamst \gets \StructureParamst - \eta\nabla_{\StructureParamst}\Losst\Big(\ft(\inputvec)\Big)$%
                    \STATEx {\color{white} ${\ModuleParamsi[k+1] \!\gets\! \ModuleParamsi[k+1] - \eta\nabla_{\!\!\ModuleParamsi[k+1]}\Losst\Big(\ft(\inputvec)\Big)}$}%
                    \STATEx {\color{white} $\mathtt{hideComponent}(k+1)$
                    \STATE $\StructureParamst \gets \StructureParamst - \eta\nabla_{\StructureParamst}\Losst\Big(\ft(\inputvec)\Big)$}%
                    \STATEx {\color{white}  $\mathtt{recoverComponent}(k+1)$}
        \ENDFOR  
    \ENDFOR \COMMENT{backprop on structure}
\end{algorithmic}
\end{algorithm}
\end{minipage}
\hfill
\begin{minipage}[t]{0.49\linewidth}
\begin{algorithm}[H] 
\renewcommand\algorithmicthen{}
\renewcommand\algorithmicdo{}
  \caption{Assimilation (Dyn.~+ Comp.)}
  \label{alg:AssimilationDynamic}
\begin{algorithmic}[1] 
        \STATE $\ModuleParamsMatrix \gets [\ModuleParamsMatrix; \mathtt{randomVector}()]$ \COMMENT{new comp.}%
        \STATE $\StructureParamsti{k+1,1:k} \gets 1$
            \FOR{$\,\,i = 1,\ldots,\,\mathtt{numEpochs}-1$}
                \FOR {$\,\,\inputvec \gets \Taskt.\mathtt{train}$}
                    \STATE $\StructureParamst \gets \StructureParamst - \eta\nabla_{\StructureParamst}\Losst\Big(\ft(\inputvec)\Big)$%
                    \STATE ${\ModuleParamsi[k+1] \!\gets\! \ModuleParamsi[k+1] - \eta\nabla_{\!\!\ModuleParamsi[k+1]}\Losst\Big(\ft(\inputvec)\Big)}$%
                    \STATE $\mathtt{hideComponent}(k+1)$
                    \STATE $\StructureParamst \gets \StructureParamst - \eta\nabla_{\StructureParamst}\Losst\Big(\ft(\inputvec)\Big)$%
                    \STATE $\mathtt{recoverComponent}(k+1)$
                \ENDFOR
            \ENDFOR \COMMENT{component dropout}
\end{algorithmic}
\end{algorithm}
\end{minipage}

\subsection{Lifelong Compositional Algorithms Using Na\"ive Fine-Tuning}
\vspace{-1.75em}
\begin{minipage}[t]{0.49\linewidth}
\begin{algorithm}[H] 
\renewcommand\algorithmicthen{}
\renewcommand\algorithmicdo{}
  \caption{Compositional \VAN{}}
  \label{alg:CompositionalVAN}
\begin{algorithmic}[1] 
    \WHILE {$\,\,\Taskt\gets \mathtt{getTask}()$}
            \IF{$\,\,t < \numTasks_{\mathtt{init}}$}
                \STATE Call Algorithm~\ref{alg:Initialization} \COMMENT{initialization}
            \ELSE
                \STATE Call Algorithm~\ref{alg:AssimilationCompositional} \COMMENT{assimilation}
            \FOR {$\,\,\inputvec \gets \Taskt.\mathtt{train}$}
                \STATE $\ModuleParamsMatrix \gets \ModuleParamsMatrix - \eta\nabla_{\ModuleParamsMatrix}\Losst\Big(\ft(\inputvec)\Big)$
                \STATEx {\color{white}$\mathtt{hideComponent}(k+1)$}
                \STATEx {\color{white}$\ModuleParamsMatrix \gets \ModuleParamsMatrix - \eta\nabla_{\ModuleParamsMatrix}\Losst\Big(\ft(\inputvec)\Big)$}
                \STATEx {\color{white}$\mathtt{recoverComponent}(k+1)$}
            \ENDFOR \COMMENT{adaptation}%
            \STATEx {\color{white}Call Algorithm~\ref{alg:Expansion} expansion}
        \ENDIF
    \ENDWHILE
\end{algorithmic}
\end{algorithm}
\end{minipage}
\hfill
\begin{minipage}[t]{0.49\linewidth}
\begin{algorithm}[H] 
\renewcommand\algorithmicthen{}
\renewcommand\algorithmicdo{}
  \caption{Dyn.~+ Compositional \VAN{}}
  \label{alg:DynamicVAN}
\begin{algorithmic}[1] 
    \WHILE {$\,\,\Taskt\gets \mathtt{getTask}()$}
            \IF{$\,\,t < \numTasks_{\mathtt{init}}$}
                \STATE Call Algorithm~\ref{alg:Initialization} \COMMENT{initialization}
            \ELSE
                \STATE Call Algorithm~\ref{alg:AssimilationDynamic} \COMMENT{assimilation}
            \FOR {$\,\,\inputvec \gets \Taskt.\mathtt{train}$}
                \STATE $\ModuleParamsMatrix \gets \ModuleParamsMatrix - \eta\nabla_{\ModuleParamsMatrix}\Losst\Big(\ft(\inputvec)\Big)$
                \STATE $\mathtt{hideComponent}(k+1)$
                \STATE $\ModuleParamsMatrix \gets \ModuleParamsMatrix - \eta\nabla_{\ModuleParamsMatrix}\Losst\Big(\ft(\inputvec)\Big)$
                \STATE $\mathtt{recoverComponent}(k+1)$
            \ENDFOR \COMMENT{adaptation}
            \STATE Call Algorithm~\ref{alg:Expansion} \COMMENT{expansion}
        \ENDIF
    \ENDWHILE
\end{algorithmic}
\end{algorithm}
\end{minipage}

\subsection{Lifelong Compositional Algorithms Using Elastic Weight Consolidation}
\vspace{-1.75em}

\begin{minipage}[t]{0.49\linewidth}
\begin{algorithm}[H] 
\renewcommand\algorithmicthen{}
\renewcommand\algorithmicdo{}
  \caption{Compositional \kEWC{}}
  \label{alg:CompositionalEWC}
\begin{algorithmic}[1] 
    \WHILE {$\,\,\Taskt\gets \mathtt{getTask}()$}
        \IF{$\,\,t < \numTasks_{\mathtt{init}}$}
            \STATE Call Algorithm~\ref{alg:Initialization} \COMMENT{initialization}
        \ELSE
            \STATE Call Algorithm~\ref{alg:AssimilationCompositional} \COMMENT{assimilation}
            \FOR {$\,\,\inputvec \gets \Taskt.\mathtt{train}$}
                \STATE $\bm{A} \gets \sum_{\that}^{t-1}\kfacAt[\that] \ModuleParamsMatrix \kfacBt[\that]$
                \STATE $\bm{g} \gets \nabla_{\ModuleParamsMatrix} \Losst\Big(\ft(\inputvec)\Big)  + \lambda (\bm{A}-\bm{B})$
                \STATE $\ModuleParamsMatrix \gets \ModuleParamsMatrix - \eta\bm{g}$
                \STATEx {\color{white}$\mathtt{hideComponent}(k+1)$}
                \STATEx {\color{white}$\bm{A} \gets \sum_{\that}^{t-1}\kfacAt[\that] \ModuleParamsMatrix \kfacBt[\that]$}
                \STATEx {\color{white}$\bm{g} \gets \nabla_{\ModuleParamsMatrix} \Losst\Big(\ft(\inputvec)\Big)  + \lambda (\bm{A}-\bm{B})$}
                \STATEx {\color{white}$\ModuleParamsMatrix \gets \ModuleParamsMatrix - \eta\bm{g}$}
                \STATEx {\color{white}$\mathtt{recoverComponent}(k+1)$}
            \ENDFOR \COMMENT{adaptation}%
            \STATEx {\color{white} Call Algorithm~\ref{alg:Expansion} expansion}
        \ENDIF
        \STATE $\kfacAt, \kfacBt \gets \mathtt{KFAC}\Big(\Taskt.\mathtt{train}, \ModuleParamsMatrix\Big)$
        \STATE $\bm{B} \gets \bm{B} - \kfacAt\ModuleParamsMatrix\kfacBt$
    \ENDWHILE
\end{algorithmic}
\end{algorithm}
\end{minipage}
\hfill
\begin{minipage}[t]{0.49\linewidth}
\begin{algorithm}[H] 
\renewcommand\algorithmicthen{}
\renewcommand\algorithmicdo{}
  \caption{Dyn.~+ Compositional \kEWC{}}
  \label{alg:DynamicEWC}
\begin{algorithmic}[1] 
    \WHILE {$\,\,\Taskt\gets \mathtt{getTask}()$}
        \IF{$\,\,t < \numTasks_{\mathtt{init}}$}
            \STATE Call Algorithm~\ref{alg:Initialization} \COMMENT{initialization}
        \ELSE
            \STATE Call Algorithm~\ref{alg:AssimilationDynamic} \COMMENT{assimilation}
            \FOR {$\,\,\inputvec \gets \Taskt.\mathtt{train}$}
                \STATE $\bm{A} \gets \sum_{\that}^{t-1}\kfacAt[\that] \ModuleParamsMatrix \kfacBt[\that]$
                \STATE $\bm{g} \gets \nabla_{\ModuleParamsMatrix} \Losst\Big(\ft(\inputvec)\Big)  + \lambda (\bm{A}-\bm{B})$
                \STATE $\ModuleParamsMatrix \gets \ModuleParamsMatrix - \eta\bm{g}$
                \STATE $\mathtt{hideComponent}(k+1)$
                \STATE $\bm{A} \gets \sum_{\that}^{t-1}\kfacAt[\that] \ModuleParamsMatrix \kfacBt[\that]$
                \STATE $\bm{g} \gets \nabla_{\ModuleParamsMatrix} \Losst\Big(\ft(\inputvec)\Big)  + \lambda (\bm{A}-\bm{B})$
                \STATE $\ModuleParamsMatrix \gets \ModuleParamsMatrix - \eta\bm{g}$
                \STATE $\mathtt{recoverComponent}(k+1)$
            \ENDFOR \COMMENT{adaptation}
            \STATE Call Algorithm~\ref{alg:Expansion} \COMMENT{expansion}
        \ENDIF
        \STATE $\kfacAt, \kfacBt \gets \mathtt{KFAC}\Big(\Taskt.\mathtt{train}, \ModuleParamsMatrix\Big)$
        \STATE $\bm{B} \gets \bm{B} - \kfacAt\ModuleParamsMatrix\kfacBt$
    \ENDWHILE
\end{algorithmic}
\end{algorithm}
\end{minipage}

\subsection{Lifelong Compositional Algorithms Using Experience Replay}
\vspace{-1.75em}

\begin{minipage}[t]{0.49\linewidth}
\begin{algorithm}[H] 
\renewcommand\algorithmicthen{}
\renewcommand\algorithmicdo{}
  \caption{Compositional \ER{}}
  \label{alg:CompositionalER}
\begin{algorithmic}[1] 
    \WHILE {$\,\,\Taskt\gets \mathtt{getTask}()$}
        \IF{$\,\,t < \numTasks_{\mathtt{init}}$}
            \STATE Call Algorithm~\ref{alg:Initialization} \COMMENT{initialization}
        \ELSE
            \STATE Call Algorithm~\ref{alg:AssimilationCompositional} \COMMENT{assimilation}
            \FOR {$\,\,\that, \inputvec \gets \!\Big(t, \Taskt.\mathtt{train}\Big) \cup\, \mathtt{buffer}$}
                \STATE $\ModuleParamsMatrix \gets \ModuleParamsMatrix - \eta\nabla_{\ModuleParamsMatrix}\Losst[\that]\Big(\ft[\that](\inputvec)\Big)$%
                \STATEx {\color{white}$\mathtt{hideComponent}(k+1)$}
                \STATEx {\color{white}$\ModuleParamsMatrix \gets \ModuleParamsMatrix - \eta\nabla_{\ModuleParamsMatrix}\Losst[\that]\Big(\ft[\that](\inputvec)\Big)$}%
                \STATEx {\color{white}$\mathtt{recoverComponent}(k+1)$}
            \ENDFOR \COMMENT{adaptation}%
            \STATEx {\color{white}Call Algorithm~\ref{alg:Expansion} expansion}
        \ENDIF
        \STATE $\mathtt{buffer}[t] \gets \mathtt{sample}\Big(\Taskt.\mathtt{train}, n_m\Big)$
    \ENDWHILE
\end{algorithmic}
\end{algorithm}
\end{minipage}
\hfill
\begin{minipage}[t]{0.49\linewidth}
\begin{algorithm}[H] 
\renewcommand\algorithmicthen{}
\renewcommand\algorithmicdo{}
  \caption{Dyn.~+ Compositional \ER{}}
  \label{alg:DynamicER}
\begin{algorithmic}[1] 
    \WHILE {$\,\,\Taskt\gets \mathtt{getTask}()$}
        \IF{$\,\,t < \numTasks_{\mathtt{init}}$}
            \STATE Call Algorithm~\ref{alg:Initialization} \COMMENT{initialization}
        \ELSE
            \STATE Call Algorithm~\ref{alg:AssimilationDynamic} \COMMENT{assimilation}
            \FOR {$\,\,\that, \inputvec \gets \!\Big(t, \Taskt.\mathtt{train}\Big) \cup\, \mathtt{buffer}$}
                \STATE $\ModuleParamsMatrix \gets \ModuleParamsMatrix - \eta\nabla_{\ModuleParamsMatrix}\Losst[\that]\Big(\ft[\that](\inputvec)\Big)$%
                \STATE $\mathtt{hideComponent}(k+1)$
                \STATE $\ModuleParamsMatrix \gets \ModuleParamsMatrix - \eta\nabla_{\ModuleParamsMatrix}\Losst[\that]\Big(\ft[\that](\inputvec)\Big)$%
                \STATE $\mathtt{recoverComponent}(k+1)$
            \ENDFOR \COMMENT{adaptation}
            \STATE Call Algorithm~\ref{alg:Expansion} \COMMENT{expansion}
        \ENDIF
        \STATE $\mathtt{buffer}[t] \gets \mathtt{sample}\Big(\Taskt.\mathtt{train}, n_m\Big)$
    \ENDWHILE
\end{algorithmic}
\end{algorithm}
\end{minipage}

\subsection{Computational Complexity}
\label{sec:ComputationalCost}

Approaches to lifelong learning tend to be computationally intensive, revisiting data or parameters from previous tasks at each training step. The proposed framework only carries out these expensive operations during (infrequent) adaptation steps. This section contains the derivations of asymptotic bounds for the computational complexity of all algorithms within the framework  described in Section~\ref{sec:FrameworkInstantiationsSupervised}, as well as the baselines used for the empirical evaluation. Briefly, joint baselines train compositional structures in a single stage, while no-components baselines optimize a monolithic architecture across all tasks. Section~\ref{sec:Baselines} provides more detailed descriptions of the baselines. These derivations assume the network architecture uses fully connected layers, and soft layer ordering for compositional structures. Extending these results to convolutional layers and soft gating is straightforward.

A single forward and backward pass through a standard fully connected layer of $d_i$ inputs and $d_o$ outputs requires $O(d_id_o)$ computations, and is additive across layers. Assuming a binary classification net, the no-components architecture contains one input layer $\inputTransformt$ with $d$ inputs and $\tilde{d}$ outputs, $\numModules$ layers with $\tilde{d}$ inputs and $\tilde{d}$ outputs, and one output layer $\outputTransformt$ with $\tilde{d}$ inputs and one output. Consequently, training such a net in the standard STL setting requires $O\big(d\tilde{d} + \tilde{d}^2\numModules + \tilde{d}\big)$ computations per input point. For a full epoch of training on a data set with $n$ data points, the training cost would then be $O\big(n\tilde{d}\big(\tilde{d}\numModules+d\big)\big)$. This is exactly the computational cost of no-components \VAN{}, since it ignores any information from past tasks during training, and leverages only the initialization of parameters.

On the other hand, a soft layer ordering net evaluates all $\numModules$ layers of size $\tilde{d}\times\tilde{d}$ at every one of the $\numModules$ depths in the network, resulting in a cost of $O\big(\tilde{d}^2\numModules^2\big)$ for those layers. This results in an overall cost per epoch of $O\big(n\tilde{d}\big(\tilde{d}\numModules^2 + d\big)\big)$ for single-task training, and therefore also for joint \VAN{} training. Since compositional methods do not use information from earlier tasks during assimilation, because they only train the task-specific structure $\structuret$ during this stage, then the cost per epoch of assimilation is also $O\big(n\tilde{d}\big(\tilde{d}\numModules^2+d\big)\big)$. Dynamic + compositional methods can at most contain $\numTasks$ components if they add one new component for every seen task. This leads to a cost of $O\big(\tilde{d}^2\numModules\numTasks\big)$ for the shared layers, and an overall cost per epoch of assimilation of $O\big(n\tilde{d}\big(\tilde{d}\numModules\numTasks+d\big)\big)$.

\kEWC{} requires computing two $O\big(\tilde{d}\times\tilde{d}\big)$ matrices, $\kfacAt$ and $\kfacBt$, for every observed task. It then modifies the gradient of component $\modulei$ by adding \mbox{$\lambda\sum_{t=1}^{\numTasks} \kfacBt[t]\ModuleParamsi\kfacBt[t] - \kfacBt[t]\ModuleParamsi^{(t)}\kfacBt[t]$} at each iteration, where $\ModuleParamsi^{(t)}$ are the parameters of component $\modulei$ obtained after training on task $\Taskt[t]$. While the second term of this sum can be precomputed and stored in memory, it is not possible to precompute the first term. Theoretically, one can apply Kronecker product properties to store a (prohibitively large) $O\big(\tilde{d}^2\times\tilde{d}^2\big)$ matrix and avoid computing the per-task sum, but practical implementations avoid this and instead compute the sum for every task, at a cost of $O\big(\numTasks\tilde{d}^3\numModules\big)$ per mini-batch. With $O(n)$ mini-batches per epoch, this yields an additional cost with respect to joint and no-components \VAN{} of $O\big(n\numTasks\tilde{d}^3\numModules\big)$. Note that the learning carries out this step after obtaining the gradients for each layer, and thus there is no additional $\numModules^2$ term for joint \kEWC{}.

Deriving the complexity bound of \ER{} simply requires extending the size of the batch of data from $n$ to $(\numTasks n_m + n)$ for a replay buffer size of $n_m$ per task.

Table~\ref{tab:computationalComplexity} summarizes these results, highlighting that the assimilation step of the proposed methods with expansion (dynamic + compositional) is comparable to joint baselines in the \textit{worst case} (one new component per task), and the method without expansion (compositional) is always at least as fast.

\begin{table}
    \centering
    \addtolength{\extrarowheight}{0.5em}
    \caption[Time complexity of supervised lifelong compositional methods.]{Time complexity per epoch (of assimilation, where applicable) for $n$ samples of $d$ features, $k$ components of $\tilde{d}$ nodes, $T$ tasks, and $n_m$ replay samples per task.}
    \label{tab:computationalComplexity}
    \begin{tabular}{l|c|c|c}
        & \ER & \kEWC\tablefootnote{While it is theoretically possible for \kEWC{} to operate in constant time with respect to $\numTasks$, practical implementations use per-task Kronecker factors due to the enormous computational requirements of the constant-time solution.} & \VAN\\
        \hline
        \hline
         Dyn.~+ Comp. & \multicolumn{3}{c}{$O\big(n  \tilde{d}  \big(\tilde{d} \numModules\numTasks + d\big)\big)$} \\
         \hline
         Compositional & \multicolumn{3}{c}{$O\big(n  \tilde{d}  \big(\tilde{d}\numModules^2 + d\big)\big)$}\\
         \hline
         Joint & $O\big((\numTasks n_m + n)  \tilde{d}  \big(\tilde{d}\numModules^2 + d\big)\big)$ & $O\big(n  \tilde{d}  \big(\numTasks \tilde{d}^2\numModules + \tilde{d}\numModules^2 + d\big)\big)$ & $O\big(n  \tilde{d}  \big(\tilde{d}\numModules^2 + d\big)\big)$\\
         \hline
         No Comp. & $O\big((\numTasks n_m + n)  \tilde{d}  \big(\tilde{d}\numModules + d\big)\big)$ & $O\big(n \tilde{d}  \big(\numTasks \tilde{d}^2\numModules + \tilde{d}\numModules + d\big) \big)$ & $O\big(n \tilde{d}  \big(\tilde{d}\numModules + d\big)\big)$\\
    \end{tabular}
\end{table}

To put the computational complexity of dynamic + compositional methods into perspective, consider the number of components required to solve $\numTasks$ tasks with networks with \textit{hard} layer ordering, assuming that all $\numTasks$ tasks can be represented by different orders over the same set of components. Given a network with $\numModules$ depths and $\tilde{\numModules}$ components, it is possible to create ${\tilde{\numModules}}^{\numModules}$ different layer orderings. If all $\numTasks$ tasks require different orderings, then the architecture requires at least $\tilde{\numModules}=\sqrt[\numModules]{\numTasks}$ components. Designing a lifelong learning algorithm that can provably attain this bound in the number of components, or any sublinear growth in $\numTasks$, remains an open problem. 

For completeness, note that the (very infrequent) adaptation steps for compositional methods incur the same computational cost as any epoch of joint methods. On the other hand, to obtain the cost of adaptation steps for dynamic + compositional methods, replace $\numModules^2$ terms in the expressions for joint methods by $\numModules\numTasks$, again noting that this corresponds to the \textit{worst case}, where the agent adds a new component for every task it encounters.

\section{Experimental Evaluation}
\label{sec:ExperimentalEvaluationSupervised}

The primary contribution of this portion of the dissertation was a large-scale empirical evaluation, conducted to assess the capabilities of compositional lifelong learning. 
In particular, the evaluation considered multiple combinations of the algorithms described in the previous section with the compositional structures of Chapter~\ref{cha:Framework}: linear model combinations, soft layer ordering, and soft gating. The primary focus of this study was to verify that compositionality improves the overall performance of lifelong learning systems, and that the two-stage process prescribed by the proposed framework enables learning such compositional solutions. In summary, the obtained results demonstrate that methods under the proposed framework achieve higher overall performance in standard lifelong learning benchmarks, more complex benchmarks combining highly varied tasks, and toy compositional tasks.

The evaluation repeated each experiment for ten trials, varying the random seed which controlled the tasks (whenever tasks were not fixed by definition), the splits for training/validation/test, and the order in which the agent encountered the tasks. Only dynamic + compositional learners used the validation set, for deciding whether to keep a new component.

Code and data sets are at \url{https://github.com/Lifelong-ML/Mendez2020Compositional}. 

\subsection{Baselines}
\label{sec:Baselines}
The evaluation considered two baselines for every adaptation method listed above:
\begin{itemize}
    \item {\it Joint} baselines use compositional structures, but do not separate assimilation and accommodation, and instead update components and structures jointly.
    \item {\it No-components} baselines optimize a single architecture to be used for all tasks, with additional task-specific input and output mappings, $\inputTransformt$ and $\outputTransformt$.
\end{itemize} 
The latter baselines correspond to the most common lifelong learning approach, which learns a monolithic structure shared across tasks, while the former are the na\"ive extensions of those methods to a compositional setting. Additionally, the experiments trained an ablated version of the framework that keeps all components fixed after initialization~(\FM), only taking assimilation steps for each new task.

\subsection{Data Sets} 
The evaluation tested linear combinations of models on three data sets used previously for evaluating linear lifelong learning~\citep{ruvolo2013ella}. The Facial Recognition~(\bFacialRecognition{}) data set tasks involve recognizing one of three facial expression action units for one of seven people, for a total of $\numTasks=21$ tasks. 
The \bLandmine{} data set consists of $\numTasks=29$ tasks, which require detecting land mines in radar images from different regions. Finally, the London Schools~(\bLondonSchools{}) data set contains  $\numTasks=139$ regression tasks, each corresponding to exam score prediction in a different school. These three data sets underwent the same processing and train/test split of \citet{ruvolo2013ella}. 

The experiments for deep compositional methods, with soft ordering and soft gating, used five benchmark data sets, all of which split multiclass computer vision tasks into multiple tasks. Binary MNIST~(\bMNIST{}; \citealp{gradient1998lecun}) is a common lifelong learning benchmark, where each task is a binary classification problem between a pair of digits. \MNIST{} evaluations constructed $\numTasks=10$ tasks by randomly sampling the digits, allowing digits to be reused across tasks. The Binary Fashion MNIST~(\bFashion{}; \citealp{xiao2017fashion}) data set is similar to \MNIST{}, but images correspond to items of clothing. A more complex lifelong learning problem commonly used in the literature is Split CUB-200~(\bCUB{}; \citealp{welinder2010caltech}), where the agent must classify bird species. \CUB{} evaluations constructed $\numTasks=20$ tasks by randomly sampling ten species for each, without reusing classes across tasks. A preprocessing step cropped \CUB{} images by the provided bounding boxes and resized them to $224\times224$.  For these first three data sets, all architectures were fully connected networks. To show that the proposed framework supports more complex convolutional architectures, the evaluation used two additional data sets. The first such data set was a lifelong learning version of CIFAR-100~(\bCIFAR{}; \citealp{krizhevsky2009learning}) with $\numTasks=20$ tasks, each of them constructed by randomly sampling five classes, without reusing classes across tasks. For these four data sets, the experiments used the standard train/test split, and further divided the training set into $80\%$ for training and $20\%$ for validation. Finally, the evaluation used the \bOmniglot{}~\citep{lake2015human} data set, which consists of $\numTasks=50$ multiclass classification problems, each corresponding to detecting handwritten symbols in a given alphabet. \Omniglot{} evaluations split the data set into $80\%$ for training, $10\%$ for validation, and $10\%$ for test, for each task.

\begin{table}[t!]
    \centering
    \caption[Data set summary for supervised lifelong composition evaluations.]{Data set summary.}
    \label{tab:datasetDetails}
    \begin{tabular}{@{}l|c|c|c|c|c|c|c@{}}
     & tasks & classes & features & feature extractor & train & val & test\\
\hline
\hline
\FacialRecognition{} & $21$ & $2$ & $100$ & PCA & $225$--$499$ & --- & $225$--$500$ \\
\Landmine{} & $29$ & $2$ & $9$ & --- & $222$--$345$ & --- & $223$--$345$ \\
\LondonSchools{} & $139$ & --- & $27$ & --- & $11$--$125$ & --- & $11$--$126$ \\
\MNIST{} & $10$ & $2$& $784$  & --- & $\sim9500$ & $\sim2500$ & $\sim2000$ \\
\Fashion{} & $10$ & $2$ & $784$ & --- & $\sim9500$ & $\sim2500$ & $2000$ \\
\CUB{} & $20$ & $10$ & $512$ & ResNet-18 & $\sim120$ & $\sim30$ & $\sim150$ \\
\CIFAR{} & $20$ & $5$ & $32\times32\times3$ & --- & $\sim2000$ & $\sim500$ & $500$ \\
\Omniglot{} & $50$ & $14$--$55$ & $105\times105$ & --- & $224$--$880$ & $28$--$110$ & $28$--$110$\\
    \end{tabular}
\end{table}

Table~\ref{tab:datasetDetails} summarizes the details of these data sets and the splits.

\subsection{Network Architectures}
\label{sec:ModelArchitecturesSupervised}
All compositional algorithms with fixed $\numModules$ used $\numModules\!=\!4$ components, and methods with dynamic expansion used $\numModules=4$ components for initialization. This is the only architectural choice for linear models. The following paragraphs describe the architectures used for other experiments.

\paragraph{Soft layer ordering} The soft layer ordering architectures followed those used by \citet{meyerson2018beyond}, whenever possible. For \MNIST{} and \Fashion{}, agents used a task-specific linear input transformation layer $\inputTransformt$ initialized at random and kept fixed throughout training, to ensure that the input spaces were sufficiently different, and each component was a fully connected layer of $64$ units. For \CUB{}, all tasks shared a fixed (i.e., not trained) ResNet-18 pretrained on ImageNet\footnote{The evaluation used the pretrained ResNet-18 provided by PyTorch, and followed the preprocessing recommended at \url{https://pytorch.org/docs/stable/torchvision/models.html}.}~\citep{deng2009imagenet} as a shared input transformation $\inputTransform$, followed by a task-specific input transformation $\inputTransformt$ given by a linear trained layer, and each component was a fully connected layer of $256$ units. For \CIFAR{}, there was no input transformation $\inputTransform$, and each component was a convolutional layer of $50$ channels with $3\times3$ kernels and padding of $1$ pixel, followed by a max-pooling layer of size $2\times2$. Finally, for \Omniglot{}, there was also no input transformation $\inputTransform$, and each component was a convolutional layer of $53$ channels with $3\times3$ kernels and no padding, followed by max-pooling of $2\times2$ patches. The input images to the convolutional nets on \CIFAR{} and \Omniglot{} were padded with all-zero channels in order to match the number of channels required by all component layers ($50$ and $53$, respectively). All component layers were followed by ReLU activation and a dropout layer with dropout probability $p=0.5$. The output of each network was a linear task-specific output transformation $\outputTransformt$ trained individually on each task. The architectures for jointly trained baselines were identical to these, and those for no-components baselines had the same layers but no mechanism to select the order of the layers.
\paragraph{Soft gating} The soft gating architectures mimicked those of the soft layer ordering architectures closely, all having the same input and output transformations, as well as the same components. The only difference was in the structure architectures. For fully connected nets, at each depth, the structure function $\structuret$ was a linear layer that took as input the previous depth's output and whose output was a soft selection over the component layers for the current depth. For convolutional nets, there was one gating net per task with the same architecture as the main network. The model computed the structure $\structuret$ by passing the previous depth's output in the main network through the remaining depths in the gating network (e.g., the model passed the output of depth $2$ in the original network through depths $3$ and $4$ in the gating network to compute the structure over modules at depth $3$).

\subsection{Algorithm Details} 
All agents trained for $100$ epochs on each task, with a mini-batch of $32$ samples. Compositional agents used the first $99$ epochs solely for assimilation and the final epoch for adaptation. Dynamic + compositional agents followed this same process, but executed every assimilation step via component dropout; after the adaptation step, the agent kept the new component if its  validation performance with the added component represented at least a $5\%$ relative improvement over the performance without the additional component. Joint agents trained all components and the structure for the current task jointly during all $100$ epochs, keeping the structure for the previous tasks fixed, while no-components agents trained the whole model at every epoch. 

\ER{}-based algorithms used a replay buffer of a single mini-batch per task. Similarly, \kEWC{}-based algorithms used a single mini-batch to compute the approximate Fisher information matrix required for regularization, and used a fixed regularization parameter $\lambda=10^{-3}$.

To ensure a fair comparison, all algorithms, including the baselines, used the same initialization procedure by training the first $\numTasks_{\mathtt{init}}\!=\!4$ tasks jointly, in order to encourage the network to generalize across tasks. For soft ordering nets, the model initialized the order of modules for the initial tasks as a random one-hot vector for each task at each depth, selecting each component at least once, and for soft gating nets, the model initialized the gating nets randomly. The model kept the structures over initial tasks fixed during training, modifying only the parameters of the components.

\subsection{Results on Standard Benchmarks}
\label{sec:ResultsStandardBenchmarks}

The first evaluation considered tasks with no evident compositional structure, in order to demonstrate that there is no strict requirement for a certain type of compositionality. Section~\ref{sec:ResultsCompositionalDataSet} introduces a simple compositional data set, and shows that the results naturally extend to that setting.  

\subsubsection{Linear Combinations of Models}

\begin{table}[t!]
    \centering
    \caption[Performance of supervised lifelong composition: linear models.]{Average final performance across tasks using factored linear models---accuracy for \FacialRecognition{} and \Landmine{} (higher is better) and RMSE for \LondonSchools{} (lower is better). Compositional methods were best on \FacialRecognition{} and \Landmine{}, and no-components methods performed best on \LondonSchools{}, demonstrating that the latter data set contains very similar tasks. Standard errors across ten seeds reported after the $\pm$.}
    \label{tab:linearResults}
    \begin{tabular}{@{}l|l|c|c|c@{}}
Base & Algorithm & \FacialRecognition & \Landmine & \LondonSchools\\
\hline
\hline
\multirow{3}{*}{\ER} & Compositional & $\bf{79.0}${\tiny$\bf{\pm0.4}$}$\%$ & $\bf{93.6}${\tiny$\bf{\pm0.1}$}$\%$ & $10.65${\tiny$\pm0.04$}\\
& Joint & $78.2${\tiny$\pm0.4$}$\%$ & $90.5${\tiny$\pm0.3$}$\%$ & $11.55${\tiny$\pm0.09$}\\
& No Comp. & $66.4${\tiny$\pm0.3$}$\%$ & $93.5${\tiny$\pm0.1$}$\%$ & $\bf{10.34}${\tiny$\bf{\pm0.02}$}\\
\hline
\multirow{3}{*}{\kEWC} & Compositional & $\bf{79.0}${\tiny$\bf{\pm0.4}$}$\%$ & $\bf{93.7}${\tiny$\bf{\pm0.1}$}$\%$ & $10.55${\tiny$\pm0.03$}\\
& Joint & $72.1${\tiny$\pm0.7$}$\%$ & $92.2${\tiny$\pm0.2$}$\%$ & $10.73${\tiny$\pm0.17$}\\
& No Comp. & $60.1${\tiny$\pm0.5$}$\%$ & $93.5${\tiny$\pm0.1$}$\%$ & $\bf{10.35}${\tiny$\bf{\pm0.02}$}\\
\hline
\multirow{3}{*}{\VAN} & Compositional & $\bf{79.0}${\tiny$\bf{\pm0.4}$}$\%$ & $\bf{93.7}${\tiny$\bf{\pm0.1}$}$\%$ & $\bf{10.87}${\tiny$\bf{\pm0.07}$}\\
& Joint & $67.9${\tiny$\pm0.6$}$\%$ & $72.8${\tiny$\pm2.5$}$\%$ & $25.80${\tiny$\pm2.35$}\\
& No Comp. & $57.0${\tiny$\pm0.9$}$\%$ & $92.7${\tiny$\pm0.4$}$\%$ & $18.01${\tiny$\pm1.04$}\\
    \end{tabular}
\end{table}

Table~\ref{tab:linearResults} summarizes the results obtained with linear models on the \FacialRecognition{}, \Landmine{}, and \LondonSchools{} data sets. The compositional versions of \ER{}, \kEWC{}, and \VAN{} clearly outperformed all the joint versions, which learn the same form of models but by jointly optimizing structures and components. This suggests that the separation of the learning process into assimilation and accommodation stages enables the agent to better capture the structure of the problem. Interestingly, the no-components variants, which learn a single linear model for all tasks, performed better than the jointly trained versions on two of the data sets, and even outperformed the compositional algorithms on one. This indicates that the tasks in those two data sets (\Landmine{} and \LondonSchools{}) are so similar that a single model can capture them.

\subsubsection{Deep Compositional Learning With Soft Layer Ordering}
\label{sec:ResultsSoftOrdering}
The next evaluation studied the performance of the algorithms when learning deep nets with soft layer ordering, using five data sets: \MNIST{}, \Fashion{}, \CUB{}, \CIFAR{}, and \Omniglot{}. 

\begin{table}[t!]
    \centering
    \caption[Performance of supervised lifelong composition: soft layer ordering.]{Average final accuracy across tasks using soft layer ordering. Dynamic + compositional methods performed best, followed closely by compositional methods without dynamic expansion, except on \CIFAR{}. Standard errors across ten seeds reported after the $\pm$.}
    \label{tab:softOrderingResults}
    \begin{tabular}{@{}l|l|c|c|c|c|c@{}}
Base & Algorithm & \MNIST & \Fashion & \CUB & \CIFAR & \Omniglot\\
\hline
\hline
\multirow{4}{*}{\ER} & Dyn.~+ Comp. & $\bf{97.6}${\tiny$\bf{\pm0.2}$}$\%$ & $\bf{96.6}${\tiny$\bf{\pm0.4}$}$\%$ & $79.0${\tiny$\pm0.5$}$\%$ & $\bf{77.6}${\tiny$\bf{\pm0.3}$}$\%$ & $\bf{71.7}${\tiny$\bf{\pm0.5}$}$\%$\\
& Compositional & $96.5${\tiny$\pm0.2$}$\%$ & $95.9${\tiny$\pm0.6$}$\%$ & $\bf{80.6}${\tiny$\bf{\pm0.3}$}$\%$ & $58.7${\tiny$\pm0.5$}$\%$ & $71.2${\tiny$\pm1.0$}$\%$\\
& Joint & $94.2${\tiny$\pm0.3$}$\%$ & $95.1${\tiny$\pm0.7$}$\%$ & $77.7${\tiny$\pm0.5$}$\%$ & $65.8${\tiny$\pm0.4$}$\%$ & $70.7${\tiny$\pm0.3$}$\%$\\
& No Comp. & $91.2${\tiny$\pm0.3$}$\%$ & $93.6${\tiny$\pm0.6$}$\%$ & $44.0${\tiny$\pm0.9$}$\%$ & $51.6${\tiny$\pm0.6$}$\%$ & $43.2${\tiny$\pm4.2$}$\%$\\
\hline
\multirow{4}{*}{\kEWC} & Dyn.~+ Comp. & $\bf{97.2}${\tiny$\bf{\pm0.2}$}$\%$ & $\bf{96.5}${\tiny$\bf{\pm0.4}$}$\%$ & $\bf{73.9}${\tiny$\bf{\pm1.0}$}$\%$ & $\bf{77.6}${\tiny$\bf{\pm0.3}$}$\%$ & $\bf{71.5}${\tiny$\bf{\pm0.5}$}$\%$\\
& Compositional & $96.7${\tiny$\pm0.2$}$\%$ & $95.9${\tiny$\pm0.6$}$\%$ & $73.6${\tiny$\pm0.9$}$\%$ & $48.0${\tiny$\pm1.7$}$\%$ & $53.4${\tiny$\pm5.2$}$\%$\\
& Joint & $66.4${\tiny$\pm1.4$}$\%$ & $69.6${\tiny$\pm1.6$}$\%$ & $65.4${\tiny$\pm0.9$}$\%$ & $42.9${\tiny$\pm0.4$}$\%$ & $58.6${\tiny$\pm1.1$}$\%$\\
& No Comp. & $66.0${\tiny$\pm1.1$}$\%$ & $68.8${\tiny$\pm1.1$}$\%$ & $50.6${\tiny$\pm1.2$}$\%$ & $36.0${\tiny$\pm0.7$}$\%$ & $68.8${\tiny$\pm0.4$}$\%$\\
\hline
\multirow{4}{*}{\VAN} & Dyn.~+ Comp. & $\bf{97.3}${\tiny$\bf{\pm0.2}$}$\%$ & $\bf{96.4}${\tiny$\bf{\pm0.4}$}$\%$ & $73.0${\tiny$\pm0.7$}$\%$ & $\bf{73.0}${\tiny$\bf{\pm0.4}$}$\%$ & $\bf{69.4\pm0.4}\%$\\
& Compositional & $96.5${\tiny$\pm0.2$}$\%$ & $95.9${\tiny$\pm0.6$}$\%$ & $\bf{74.5}${\tiny$\bf{\pm0.7}$}$\%$ & $54.8${\tiny$\pm1.2$}$\%$ & $68.9${\tiny$\pm0.9$}$\%$\\
& Joint & $67.4${\tiny$\pm1.4$}$\%$ & $69.2${\tiny$\pm1.9$}$\%$ & $65.1${\tiny$\pm0.7$}$\%$ & $43.9${\tiny$\pm0.6$}$\%$ & $63.1${\tiny$\pm0.9$}$\%$\\
& No Comp. & $64.4${\tiny$\pm1.1$}$\%$ & $67.0${\tiny$\pm1.3$}$\%$ & $49.1${\tiny$\pm1.6$}$\%$ & $36.6${\tiny$\pm0.6$}$\%$ & $68.9${\tiny$\pm1.0$}$\%$\\
\hline
\multirow{2}{*}{\FM} & Dyn.~+ Comp. & $\bf{99.1}${\tiny$\bf{\pm0.0}$}$\%$ & $\bf{97.3}${\tiny$\bf{\pm0.3}$}$\%$ & $78.3${\tiny$\pm0.4$}$\%$ & $\bf{78.4}${\tiny$\bf{\pm0.3}$}$\%$ & $\bf{71.0}${\tiny$\bf{\pm0.4}$}$\%$\\
& Compositional & $84.1${\tiny$\pm0.8$}$\%$ & $86.3${\tiny$\pm1.3$}$\%$ & $\bf{80.1}${\tiny$\bf{\pm0.3}$}$\%$ & $48.8${\tiny$\pm1.6$}$\%$ & $63.0${\tiny$\pm3.3$}$\%$\\
    \end{tabular}
\end{table}

Results in Table~\ref{tab:softOrderingResults} show that all the algorithms conforming to the framework outperformed the joint and no-components learners. 
On four out of the five data sets, the dynamic addition of new components yielded either no or marginal improvements. However, on \CIFAR{} it was crucial for the agent to be capable of detecting when it needed new components. This added flexibility enables compositional learners to handle more varied tasks, where new problems may not be solved without substantially new knowledge. Algorithms with adaptation outperformed the ablated compositional \FM{} agent, showing that it is necessary to accommodate new knowledge into the set of components in order to handle a diversity of tasks. When \FM{} was allowed to dynamically add new components (keeping old ones fixed), it yielded the best performance on \MNIST{} and \Fashion{} by adding far more components than methods with adaptation, as shown in Table~\ref{tab:numberOfLearnedComponentsSoftOrdering}, and \CIFAR{} exhibited a similar trend.

\begin{figure}[b!]
\centering
\captionsetup[subfigure]{aboveskip=1pt}
    \begin{subfigure}[b]{0.315\textwidth}
        \includegraphics[height=1.4cm, trim={0.35cm 0cm 0cm 1.6cm}, clip]{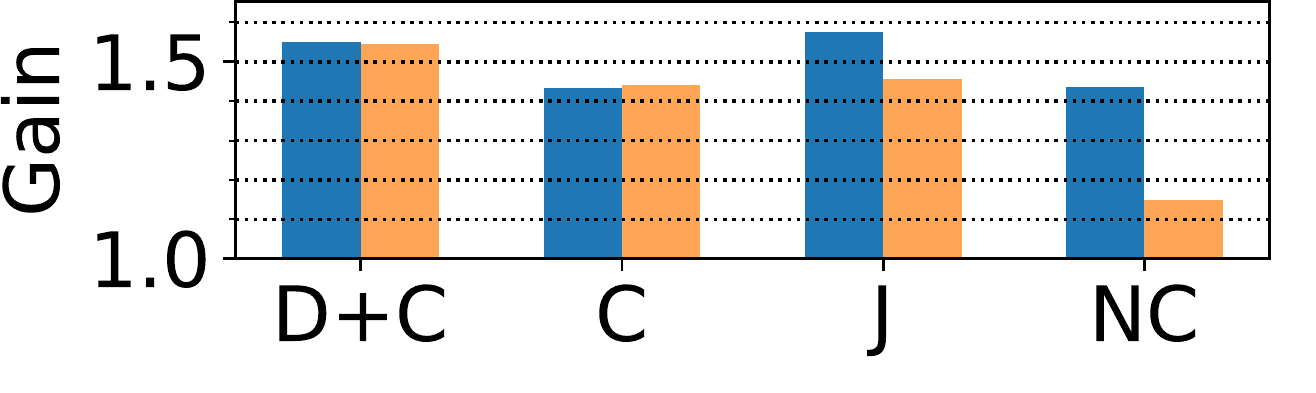}
        \caption{\ER}
    \end{subfigure}%
    \begin{subfigure}[b]{0.285\textwidth}
        \includegraphics[height=1.4cm, trim={1.3cm 0cm 0cm 1.6cm}, clip]{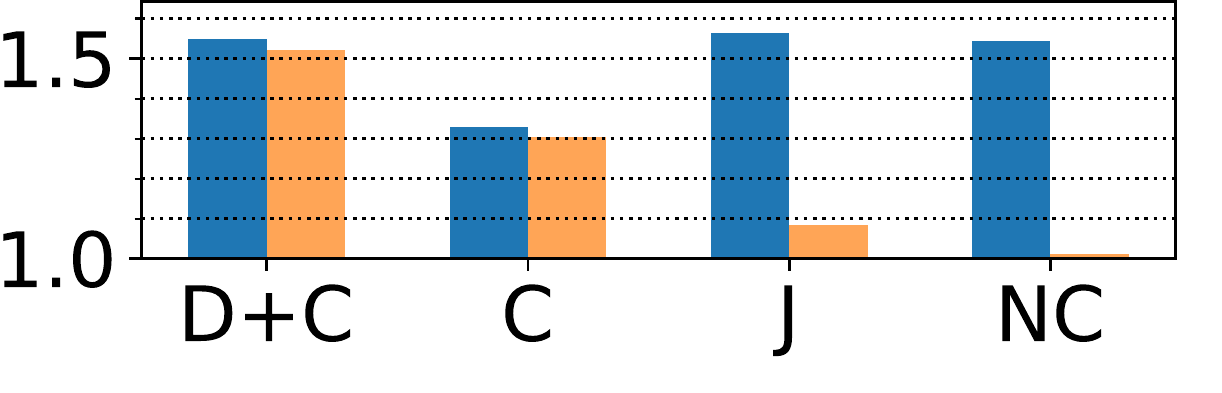}
         \caption{\kEWC}
    \vspace{-1pt}
    \end{subfigure}%
    \begin{subfigure}[b]{0.285\textwidth}
        \includegraphics[height=1.4cm, trim={1.3cm 0cm 0cm 1.6cm}, clip]{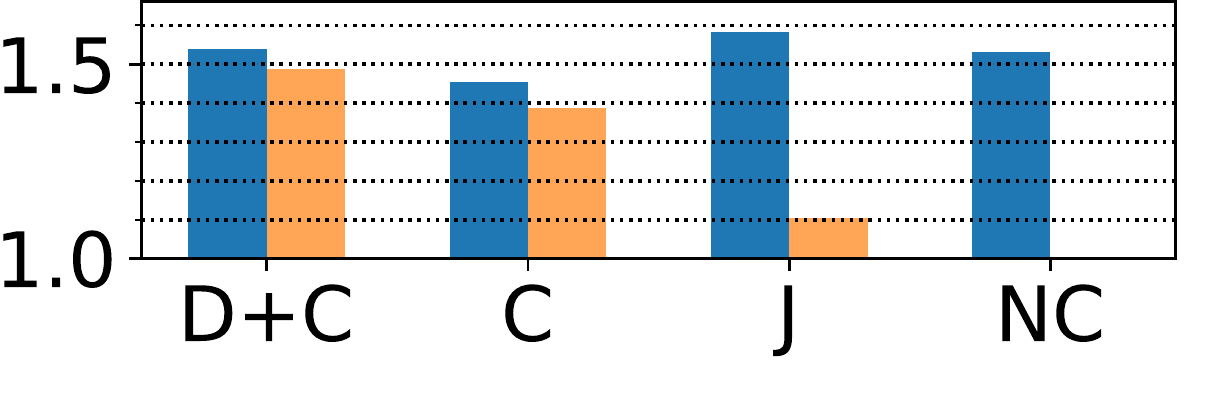}
        \caption{\VAN}
    \vspace{-1pt}
    \end{subfigure}%
    \begin{subfigure}[b]{0.115\textwidth}
        \centering
        \raisebox{2.7em}{\includegraphics[width=0.9\linewidth, trim={0.1cm, 0.1cm, 0.1cm, 0.1cm}, clip]{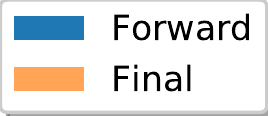}}
    \vspace{-1pt}
    \end{subfigure}
    \caption[Performance of supervised lifelong composition after each task and after all tasks: soft ordering.]{Average gain with respect to no-components \VAN{} across tasks and data sets using soft ordering, immediately after training on each task (forward) and after training on all tasks (final), using soft ordering (top) and soft gating (bottom). Algorithms within the proposed framework (C and D+C) outperformed baselines. Gaps between forward and final performance indicate that the framework exhibits less forgetting.}
    \label{fig:softOrderingBars}
\end{figure}

To study how flexibly compositional agents learn new tasks and how stably they retain knowledge about earlier tasks, Figure~\ref{fig:softOrderingBars} shows accuracy gains immediately after learning each task (forward) and after learning all tasks (final), with respect to no-components \VAN{} (final). Compositional learners without expansion struggled to match the forward performance of joint baselines, indicating that learning the ordering over existing layers during much of the training is less flexible than modifying the layers themselves, as expected. However, the added stability dramatically decreased forgetting with respect to joint methods.

\begin{figure}[b!]
\centering
\captionsetup[subfigure]{aboveskip=2pt}
    \begin{subfigure}[b]{0.54\textwidth}
        \includegraphics[height=4.5cm, trim={0.5cm 0cm 0.1cm 2.6cm}, clip]{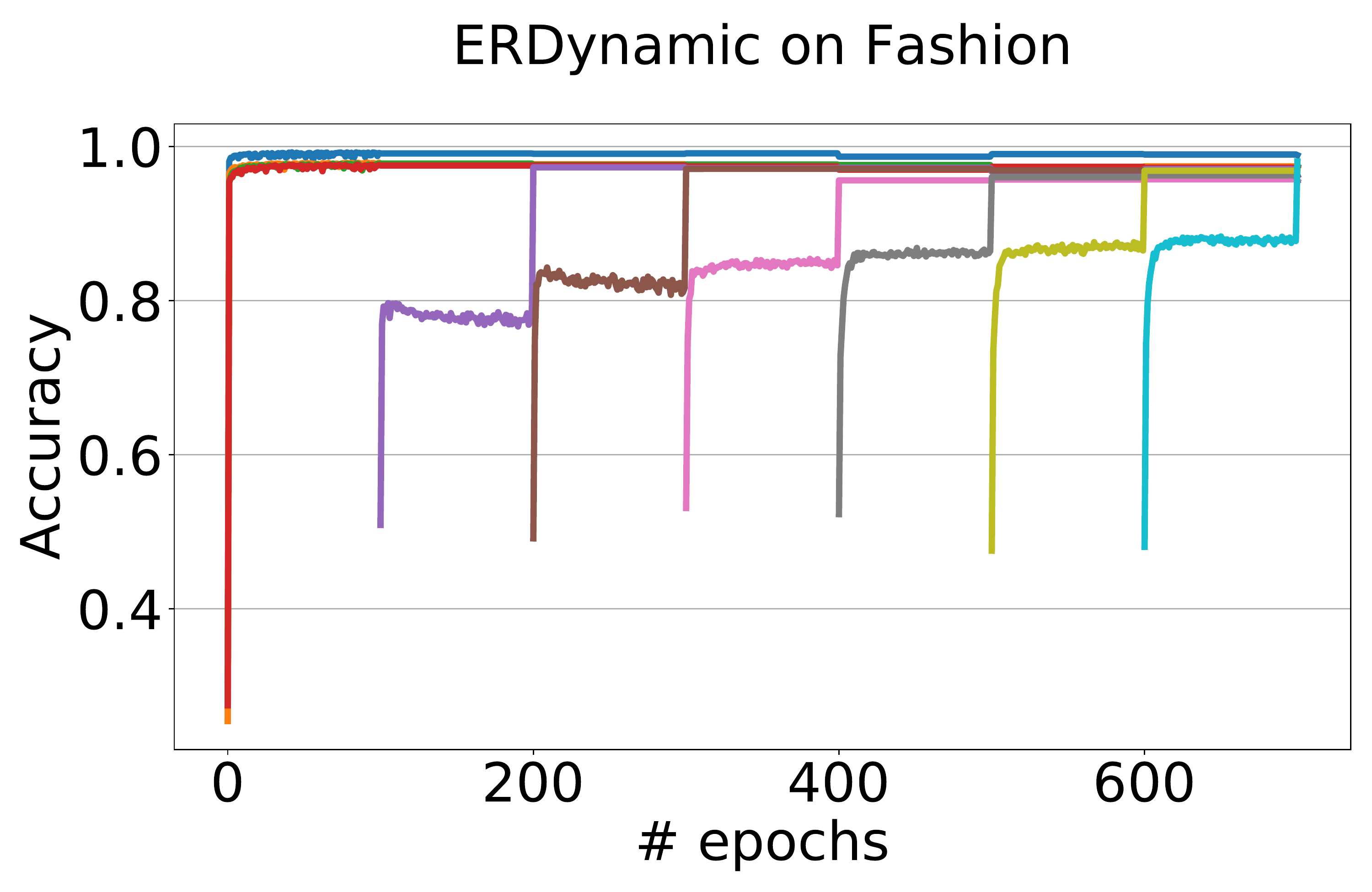}
        \caption{ER Dyn.~+ Comp.}
    \end{subfigure}%
    \begin{subfigure}[b]{0.46\textwidth}
        \includegraphics[height=4.5cm, trim={3.8cm 0cm 0.1cm 2.6cm}, clip]{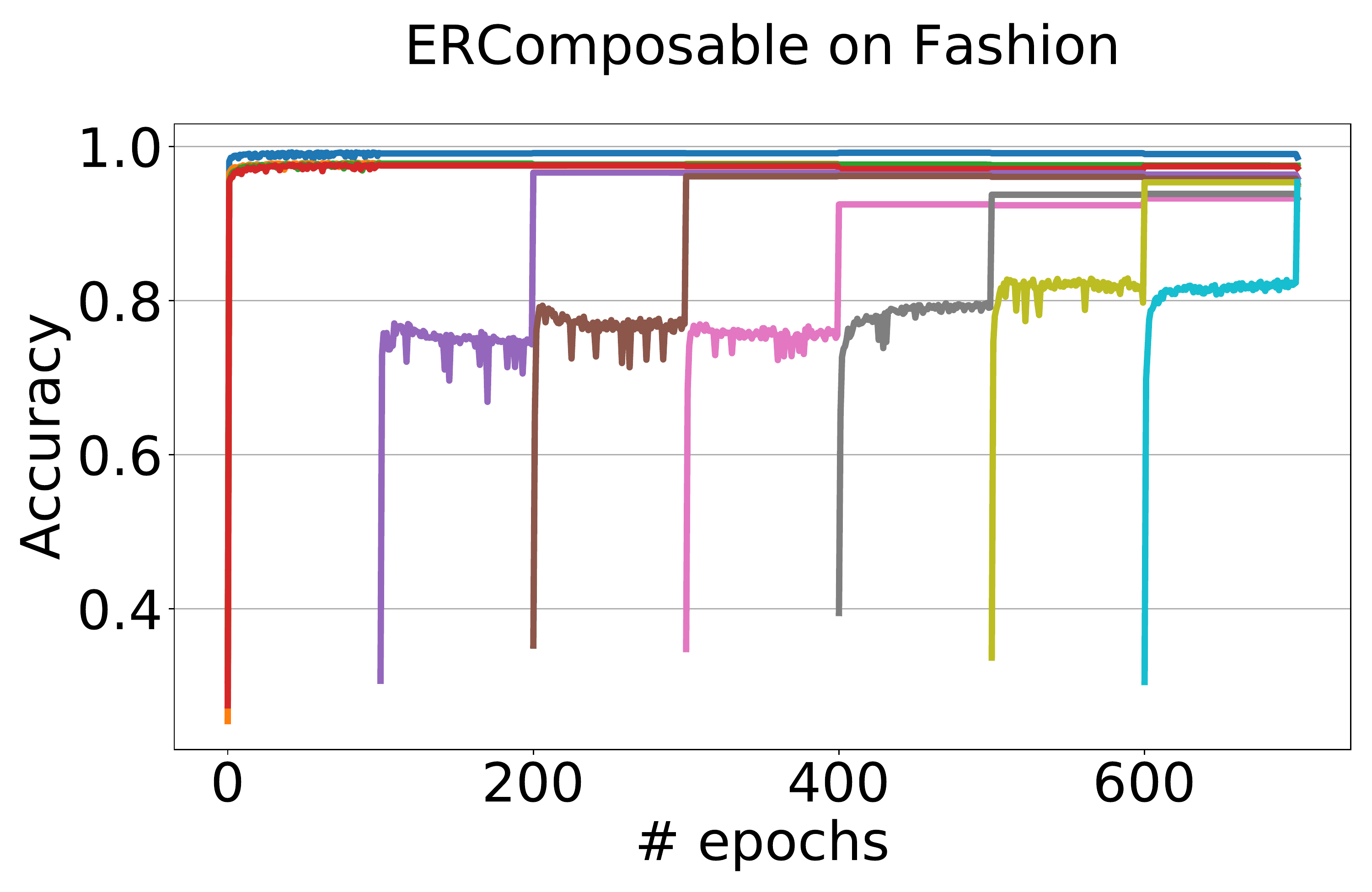}
        \caption{ER Compositional}
    \end{subfigure}\\
    \vspace{1em}
    \begin{subfigure}[b]{0.54\textwidth}
        \includegraphics[height=4.5cm, trim={0.5cm 0cm 0.1cm 2.6cm}, clip]{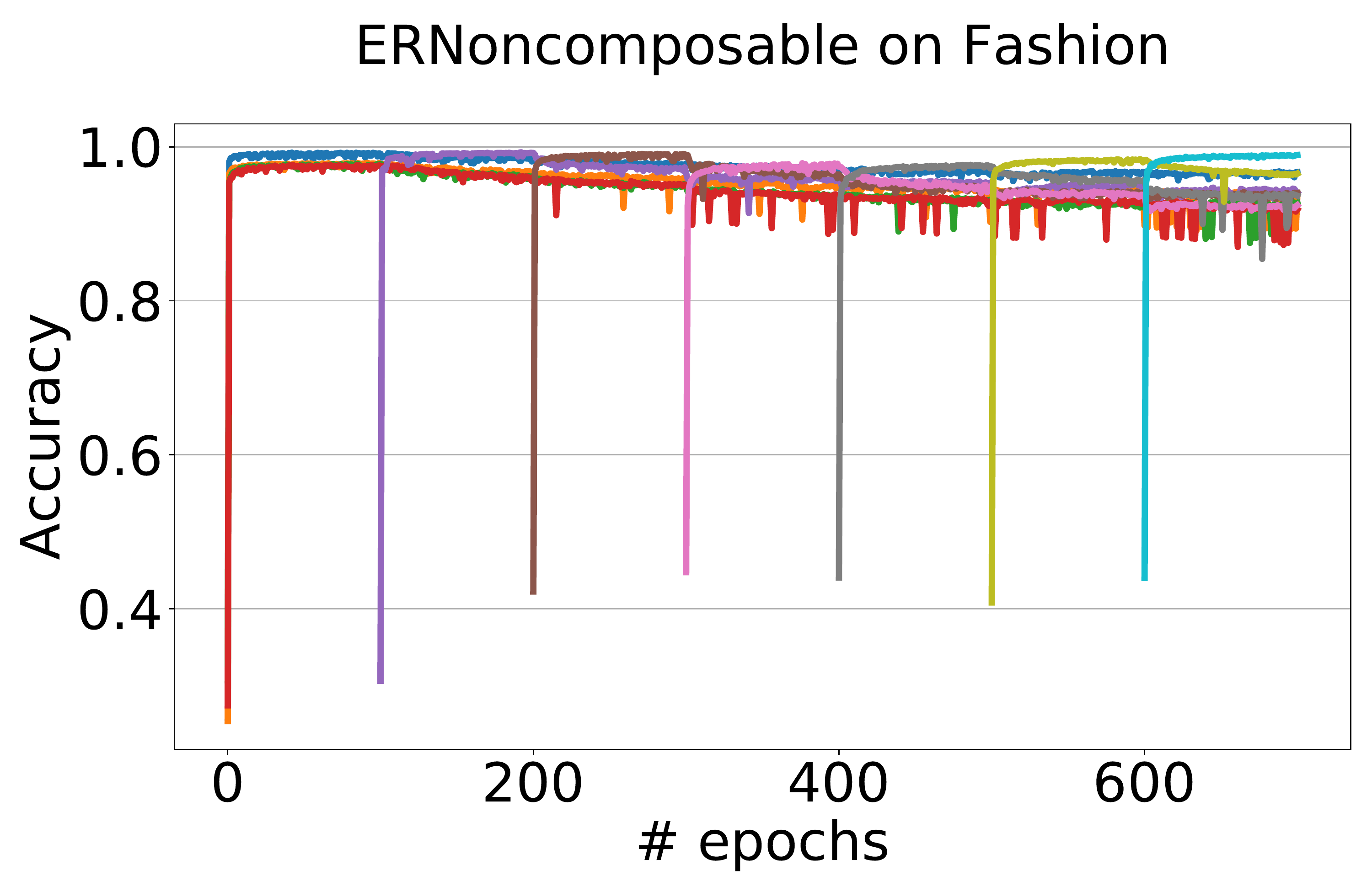}
        \caption{ER Joint}
    \end{subfigure}%
    \begin{subfigure}[b]{0.46\textwidth}
        \includegraphics[height=4.5cm, trim={3.8cm 0cm 0.1cm 2.6cm}, clip]{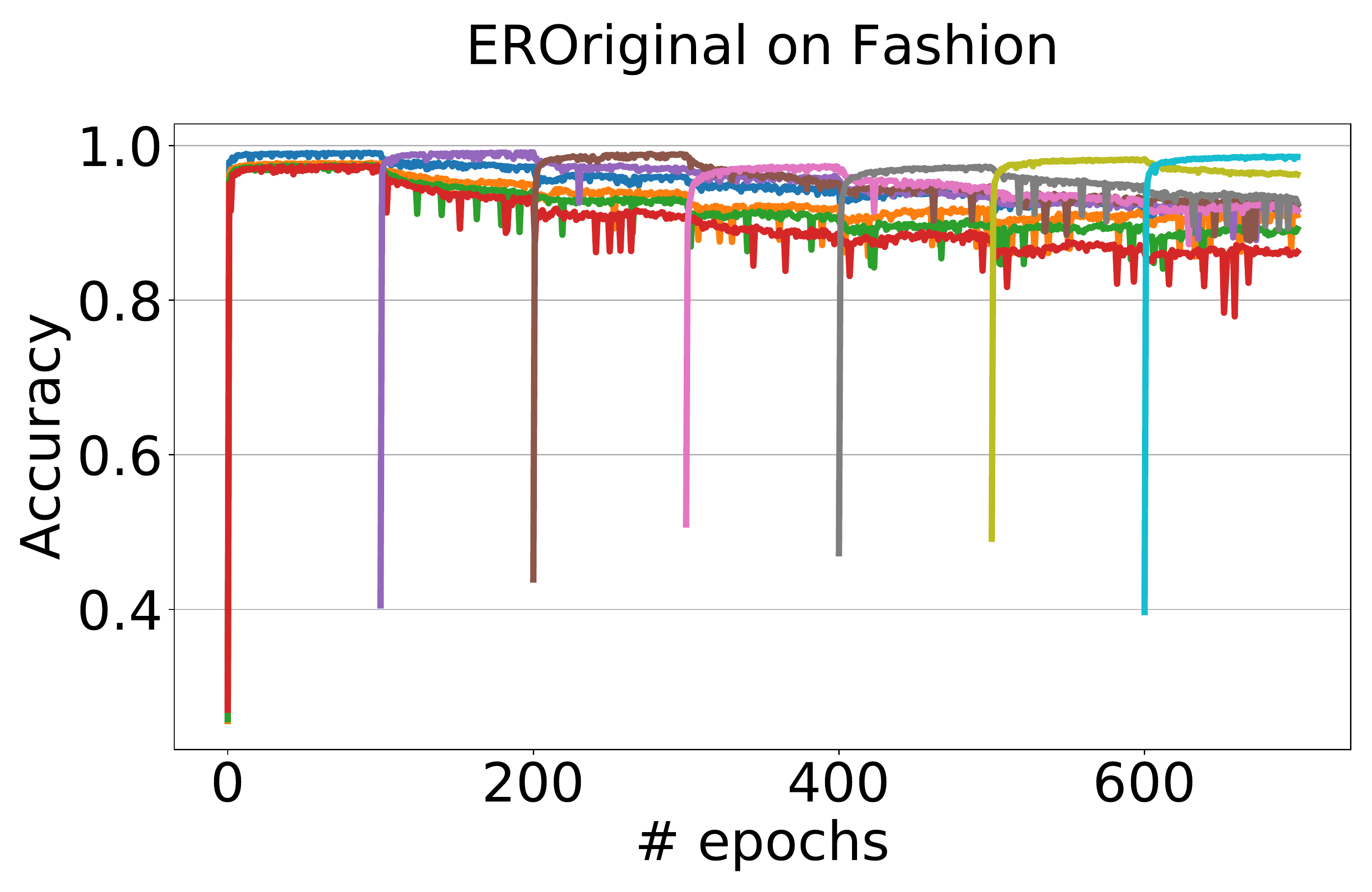}
        \caption{ER No Components}
    \end{subfigure}%
    \caption[Averaged learning curves for supervised lifelong composition.]{Learning curves averaged across \MNIST{} and \Fashion{} using \ER{} and soft ordering. Each curve shows a single task trained for $100$ epochs and continually evaluated during and after training. Algorithms under the proposed framework displayed no forgetting. For \ER{} dynamic + compositional, as the agent saw and accommodated more tasks, assimilation performance of later tasks improved. Joint and no-components versions dropped performance of early tasks during the learning of later tasks.}
    \label{fig:softOrderingCurves}
\end{figure}
The dynamic addition of new layers yielded substantial improvements in the forward stage, while still reducing catastrophic forgetting with respect to the baselines. Figure~\ref{fig:softOrderingCurves} shows the learning curves on \MNIST{} and \Fashion{} tasks using \ER{}, the best adaptation method. Performance jumps in $100$-epoch intervals show adaptation steps incorporating knowledge about the current task into the existing components without noticeably impacting earlier tasks' performance. Compositional and dynamic + compositional \ER{} exhibited almost no performance drop after training on a task, whereas accuracy for the joint and no-components versions diminished as the agent learned subsequent tasks. Most notably, as dynamic \ER{} saw more tasks, the existing components became better able to assimilate new tasks, shown by the trend of increasing performance as the number of tasks increases. This suggests that the later tasks' accommodation stage can successfully determine which new knowledge should be incorporated into existing components (enabling those components to better generalize across tasks), and which must be incorporated into a new component. See Appendix~\ref{app:FullResultsSupervised} for versions of  Figure~\ref{fig:softOrderingBars} and~\ref{fig:softOrderingCurves} separated by each data set.

\begin{figure}[b!]
\centering
    \begin{subfigure}[b]{\textwidth}
        \centering
        \includegraphics[height=0.45cm, trim={0.1cm, 0.1cm, 0.1cm, 0.15cm}, clip]{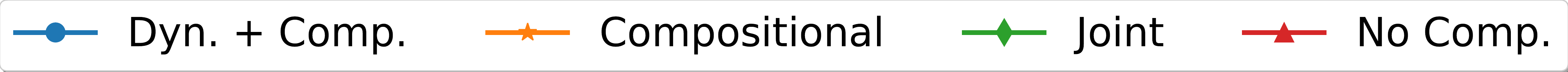}
    \end{subfigure}\\
    \begin{subfigure}[b]{0.34\textwidth}
        \includegraphics[height=4.5cm, trim={0.44cm 0.4cm 0cm 3.cm}, clip]{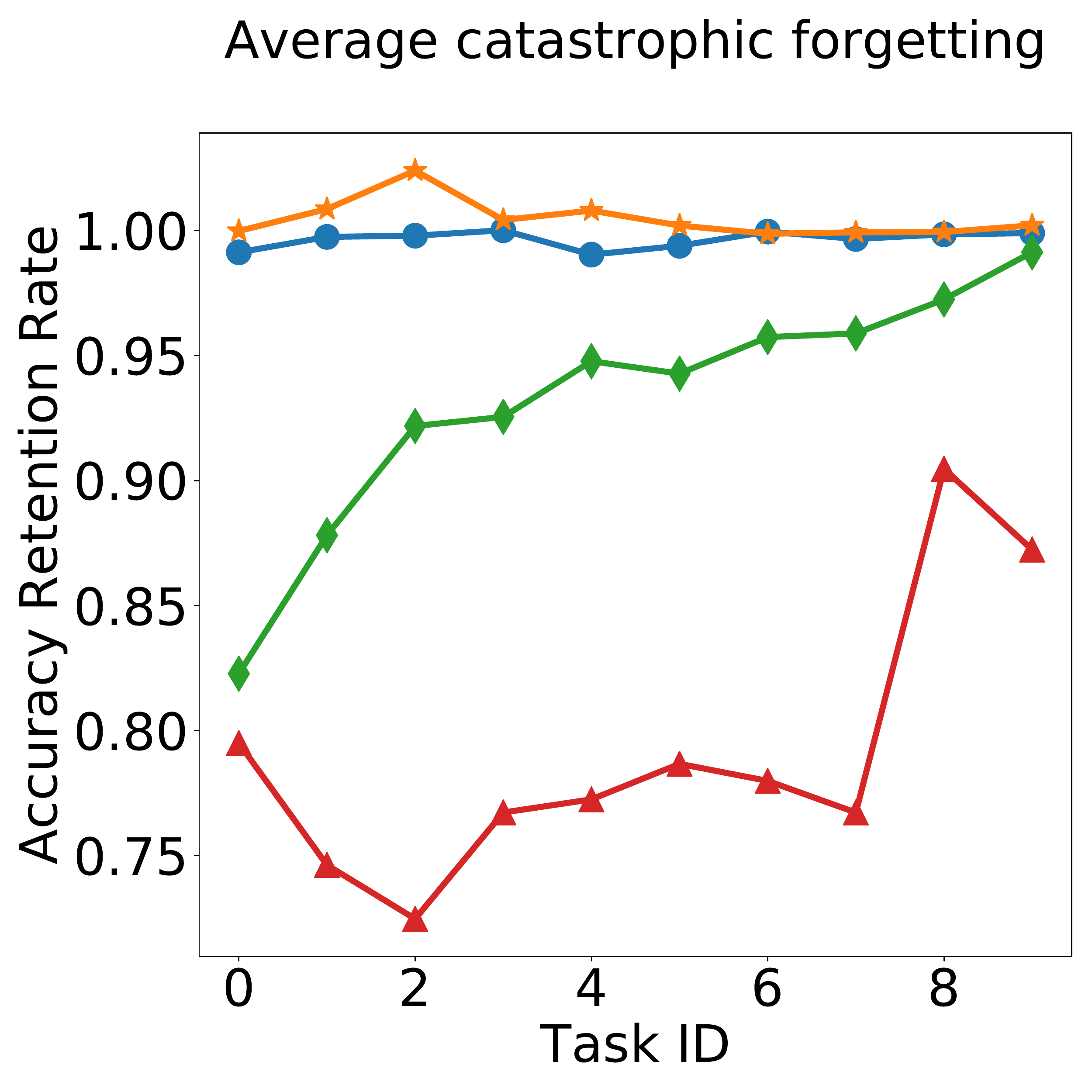}
        \caption{\ER}
    \end{subfigure}%
    \begin{subfigure}[b]{0.33\textwidth}
        \includegraphics[height=4.5cm, trim={1.6cm 0.4cm 0cm 3.cm}, clip]{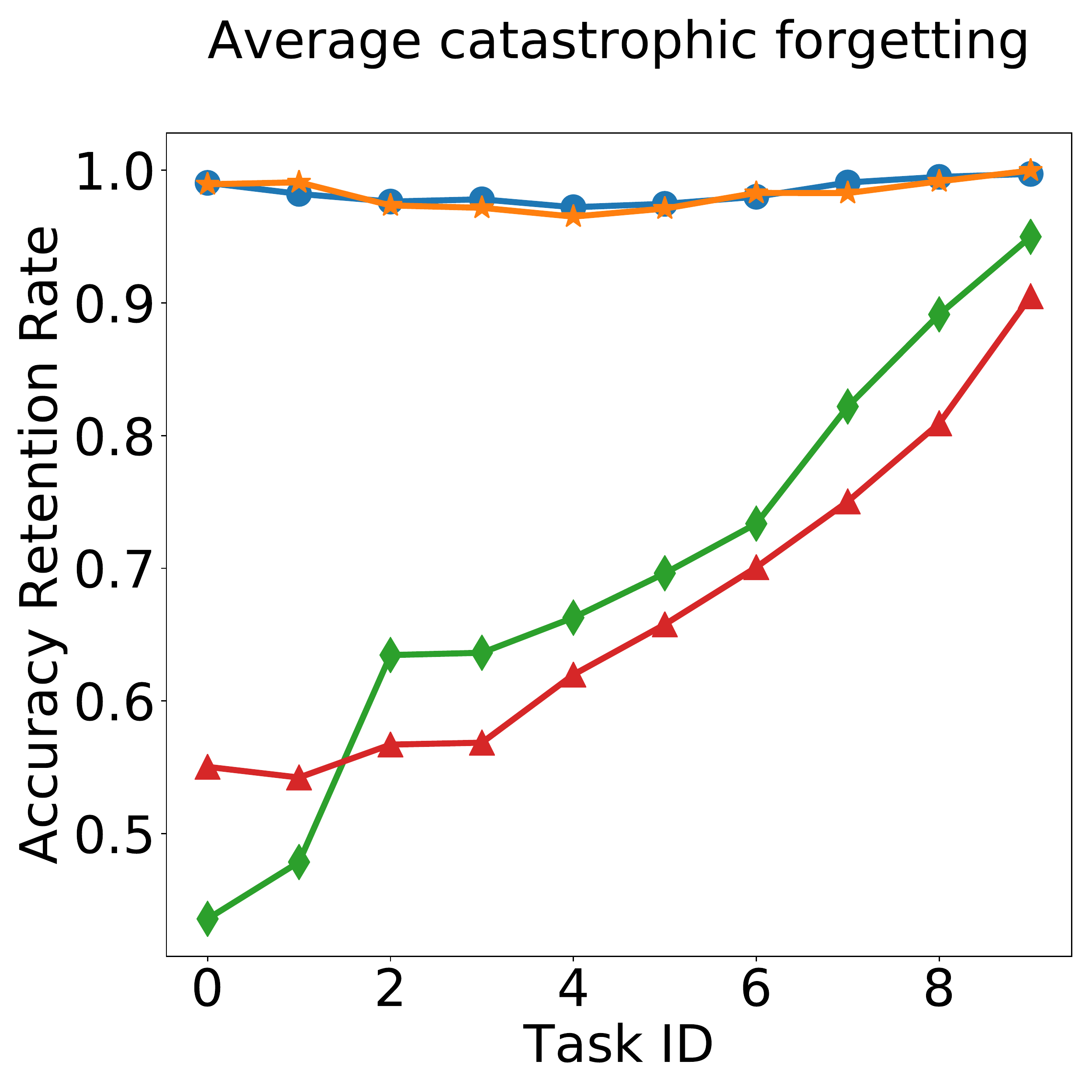}
        \caption{\kEWC}
    \end{subfigure}%
    \begin{subfigure}[b]{0.33\textwidth}
        \includegraphics[height=4.5cm, trim={1.6cm 0.4cm 0cm 3.cm}, clip]{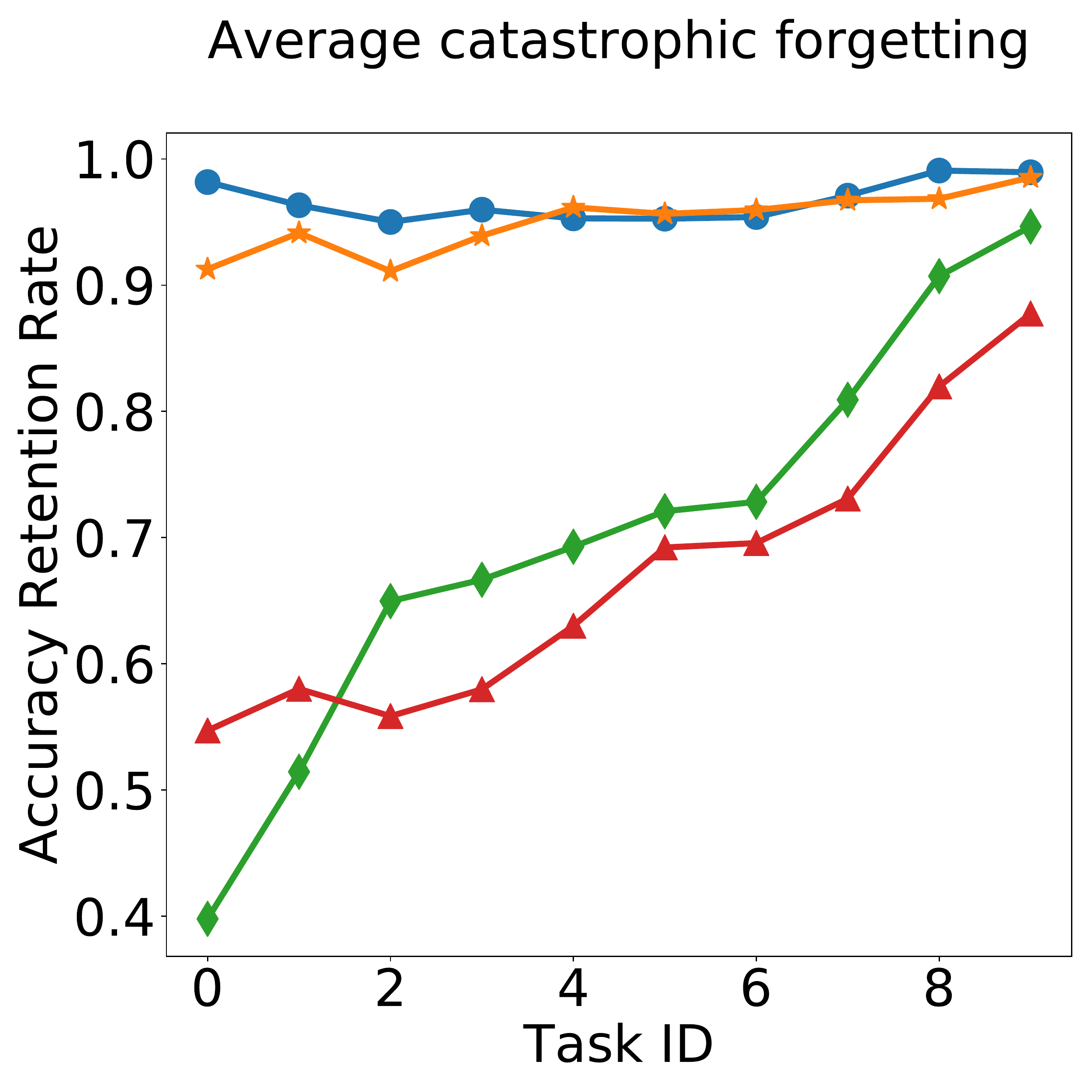}
        \caption{\VAN}
    \end{subfigure}
    \caption[Catastrophic forgetting on supervised lifelong composition: soft ordering.]{Catastrophic forgetting across data sets using soft ordering. Ratio of accuracy immediately after learning a task to after learning all tasks. For data sets with more than ten tasks, the evaluation sampled ten interleaved tasks to match all the x-axes. Compositional algorithms had practically no forgetting, whereas jointly trained and no-components baselines forgot knowledge required to solve earlier tasks.}
    \label{fig:catastrophicForgettingSoftOrdering}
\end{figure}

The gap between the first and second bar for each algorithm in Figure~\ref{fig:softOrderingBars} is an indicator of the amount of catastrophic forgetting. However, it hides details of how forgetting affects each individual task. On the other hand, the decay rate of each task in Figure~\ref{fig:softOrderingCurves} shows how the learners forget each task over time, but does not measure quantitatively how much forgetting occurs. Based on prior work~\citep{lee2019learning}, Figure~\ref{fig:catastrophicForgettingSoftOrdering} shows knowledge retention quantitatively as the ratio of performance after training on each task to after training on all tasks. Results show that compositional methods exhibit substantially less catastrophic forgetting, particularly for the earlier tasks seen during training.

\begin{table}[t!]
    \centering
    \caption[Number of learned components in supervised lifelong composition: soft ordering.]{Number of learned components using soft ordering. \MNIST{}, \Fashion{}, \CUB{}, and \Omniglot{} required few additional components, except for \FM{}. \CIFAR{} required many more modules, explaining the gap in performance between compositional and dynamic + compositional methods. Standard errors across ten seeds reported after the $\pm$.}
    \label{tab:numberOfLearnedComponentsSoftOrdering}
    \begin{tabular}{l|c|c|c|c|c}
Base & \MNIST & \Fashion & \CUB & \CIFAR & \Omniglot\\
\hline\hline
\ER & $5.2${\tiny$\pm0.3$} & $4.9${\tiny$\pm0.3$} & $5.9${\tiny$\pm0.3$} & $19.1${\tiny$\pm0.3$} & $9.3${\tiny$\pm0.3$}\\
\kEWC & $5.0${\tiny$\pm0.3$} & $4.7${\tiny$\pm0.2$} & $5.8${\tiny$\pm0.2$} & $19.6${\tiny$\pm0.2$} & $10.1${\tiny$\pm0.3$}\\
\VAN & $5.0${\tiny$\pm0.2$} & $4.8${\tiny$\pm0.3$} & $6.1${\tiny$\pm0.3$} & $17.7${\tiny$\pm0.3$} & $10.0${\tiny$\pm0.7$}\\
\FM & $10.0${\tiny$\pm0.0$} & $8.8${\tiny$\pm0.2$} & $6.5${\tiny$\pm0.4$} & $19.1${\tiny$\pm0.4$} & $10.2${\tiny$\pm0.6$}\\
    \end{tabular}
\end{table}

In the experiments discussed so far, it was in many cases necessary to incorporate an expansion step in order for compositional algorithms to be sufficiently flexible to handle the stream of incoming tasks. This expansion step enables compositional methods to dynamically add new components if the existing ones are insufficient to achieve good performance on the new task. Table~\ref{tab:numberOfLearnedComponentsSoftOrdering} shows the number of components learned by each dynamic algorithm, averaged across all ten trials. Notably, in order for dynamic methods to work on the \CIFAR{} data set, they required learning almost one component per task. This explains why compositional algorithms without dynamic component additions performed poorly on \CIFAR{}.

\begin{figure}[t!]
\centering
    \begin{subfigure}[b]{\textwidth}
        \centering
        \includegraphics[height=0.45cm, trim={0.1cm, 0.1cm, 0.1cm, 0.15cm}, clip]{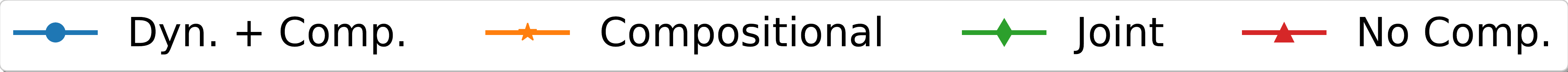}
    \end{subfigure}\\
    \begin{subfigure}[b]{0.34\textwidth}
        \includegraphics[height=4.2cm, trim={0.5cm 0.4cm 0cm 3.1cm}, clip]{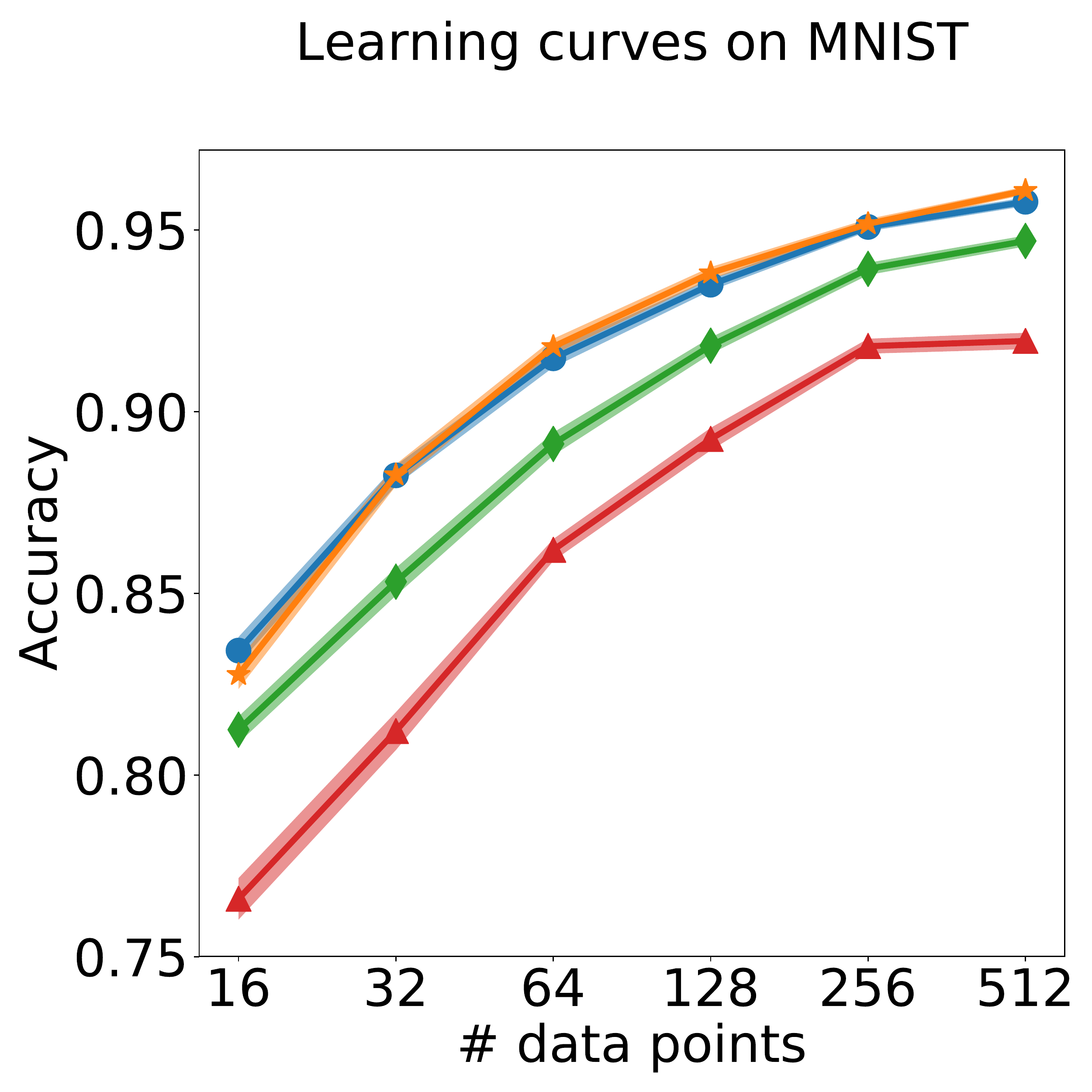}
        \caption{\MNIST{}}
    \end{subfigure}%
    \begin{subfigure}[b]{0.33\textwidth}
        \includegraphics[height=4.2cm, trim={1.cm 0.4cm 0cm 3.1cm}, clip]{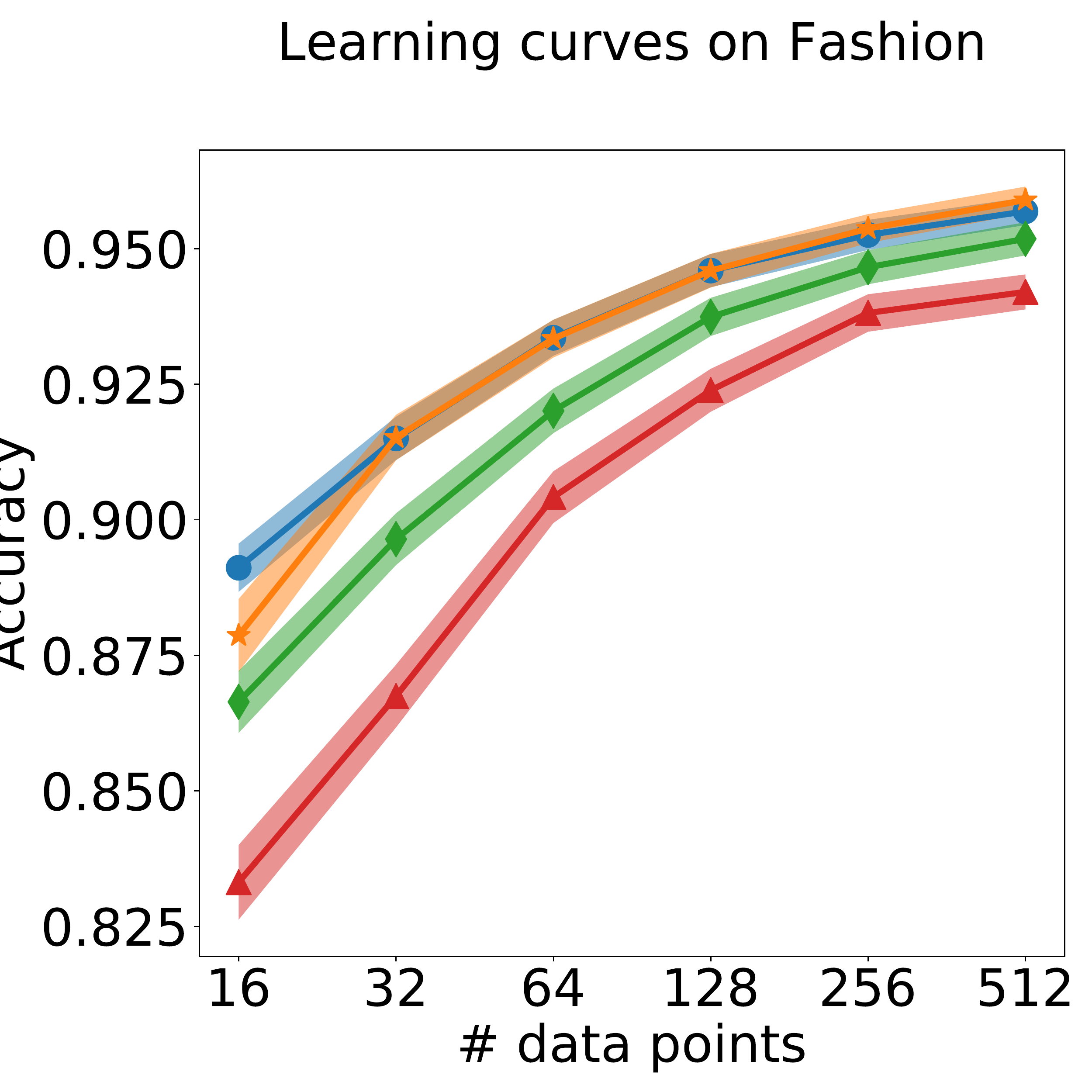}
        \caption{\Fashion{}}
    \end{subfigure}%
    \begin{subfigure}[b]{0.33\textwidth}
        \includegraphics[height=4.2cm, trim={1.6cm 0.4cm 0cm 3.1cm}, clip]{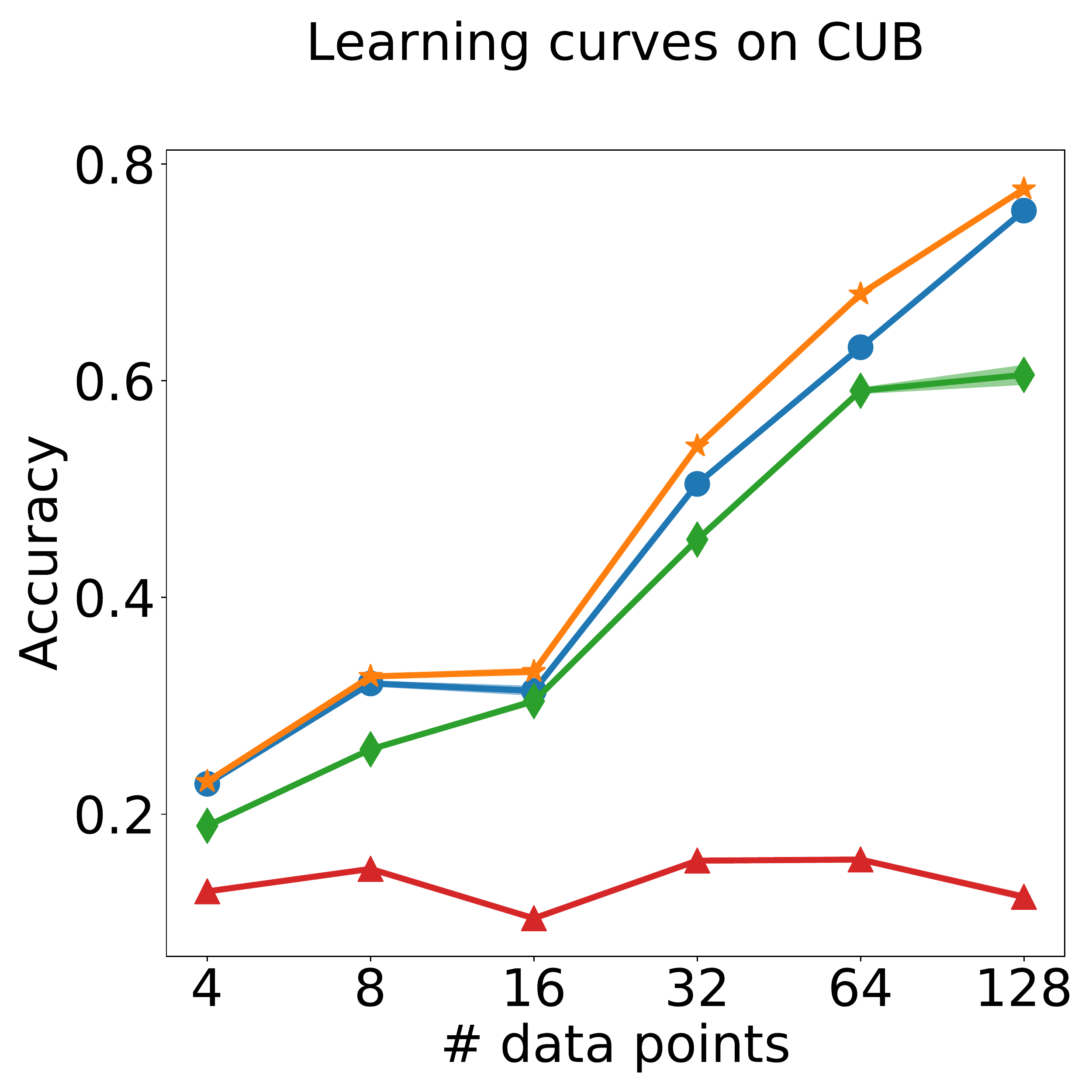}
        \caption{\CUB{}}
    \end{subfigure}
    \caption[Performance of supervised lifelong composition with respect to data set size.]{Accuracy of \ER{}-based methods with varying data sizes. Compositional methods performed better even with extremely little data per task. The shaded area represents standard errors across $50$ seeds.}
    \label{fig:limitedData}
\end{figure}

One of the key aspects of lifelong learning is the ability to learn in the presence of little data for each task, using knowledge acquired from previous tasks to acquire better generalization for new tasks. To evaluate the sample efficiency of compositional algorithms, the following experiment varied the number of data points used for training for \MNIST{}, \Fashion{}, and \CUB{}, using \ER{} as the adaptation method. 
The evaluation was repeated for $50$ trials, each with a different random seed controlling the selection of classes and samples for each task, and the order over tasks. Note that this experiment required more trials to obtain statistically significant results because the smaller sample sizes led to higher variance in the learning performance of the various algorithms. Learners trained for $1{,}000$ epochs, with compositional methods alternating nine epochs of assimilation and one epoch of adaptation. For each sample size $n$, all agents used a mini-batch of size $b=32$, and the replay buffer size was limited to $10\%$ of the sample or a single mini-batch, whichever was smaller: $\min(\max(\lfloor0.1n\rfloor, 1), b)$. Figure~\ref{fig:limitedData} shows the learning accuracy for \ER{}-based algorithms as a function of the number of training points, revealing that compositional algorithms work better than baselines even in the presence of very little data.

\begin{table}[t!]
    \centering
    \caption[Component reusability in supervised lifelong composition.]{Number of tasks that reuse a component. A task reuses a component if its accuracy drops by more than $5\%$ relative when not using the component. Several tasks reuse most modules, except on \CIFAR{}. Standard errors across ten seeds reported after the $\pm$.}
    \label{tab:ModuleReuse}
\begin{tabular}{@{}l|l|c|c|c|c|c@{}}
            Algorithm & Comp.    &            MNIST &          Fashion &               CUB &             CIFAR &          Omniglot \\
\hline\hline
\multirow{4}{*}{Compositional} & 0  &  $6.40 ${\tiny$\pm 0.43$} &  $6.00 ${\tiny$\pm 0.24$} &  $13.40 ${\tiny$\pm 0.78$} &  $18.90 ${\tiny$\pm 0.30$} &  $46.40 ${\tiny$\pm 0.98$} \\
             & 1  &  $4.90 ${\tiny$\pm 0.26$} &  $4.70 ${\tiny$\pm 0.28$} &   $7.90 ${\tiny$\pm 0.41$} &  $16.20 ${\tiny$\pm 0.80$} &  $30.60 ${\tiny$\pm 2.86$} \\
             & 2  &  $4.10 ${\tiny$\pm 0.26$} &  $4.10 ${\tiny$\pm 0.17$} &   $5.90 ${\tiny$\pm 0.57$} &  $11.90 ${\tiny$\pm 1.14$} &  $18.80 ${\tiny$\pm 3.76$} \\
             & 3  &  $3.00 ${\tiny$\pm 0.32$} &  $2.60 ${\tiny$\pm 0.32$} &   $3.20 ${\tiny$\pm 0.49$} &   $5.70 ${\tiny$\pm 0.97$} &  $10.90 ${\tiny$\pm 2.17$} \\
\hline
\multirow{15}{*}{Dyn.~+ Comp.} & 0  &  $4.70 ${\tiny$\pm 0.38$} &  $5.10 ${\tiny$\pm 0.26$} &   $9.80 ${\tiny$\pm 0.98$} &  $13.30 ${\tiny$\pm 1.27$} &  $21.90 ${\tiny$\pm 1.82$} \\
             & 1  &  $3.60 ${\tiny$\pm 0.25$} &  $3.90 ${\tiny$\pm 0.22$} &   $6.20 ${\tiny$\pm 0.56$} &   $6.20 ${\tiny$\pm 0.61$} &  $12.30 ${\tiny$\pm 0.68$} \\
             & 2  &  $2.80 ${\tiny$\pm 0.28$} &  $3.30 ${\tiny$\pm 0.28$} &   $4.40 ${\tiny$\pm 0.62$} &   $4.00 ${\tiny$\pm 0.35$} &   $9.20 ${\tiny$\pm 0.61$} \\
             & 3  &  $1.90 ${\tiny$\pm 0.26$} &  $2.10 ${\tiny$\pm 0.22$} &   $2.70 ${\tiny$\pm 0.20$} &   $3.10 ${\tiny$\pm 0.26$} &   $7.40 ${\tiny$\pm 0.47$} \\
             & 4  &              --- &              --- &               --- &   $3.00 ${\tiny$\pm 0.32$} &   $6.50 ${\tiny$\pm 0.49$} \\
             & 5  &              --- &              --- &               --- &   $1.80 ${\tiny$\pm 0.13$} &   $5.00 ${\tiny$\pm 0.51$} \\
             & 6  &              --- &              --- &               --- &   $1.50 ${\tiny$\pm 0.16$} &   $4.20 ${\tiny$\pm 0.46$} \\
             & 7  &              --- &              --- &               --- &   $1.10 ${\tiny$\pm 0.09$} &   $3.40 ${\tiny$\pm 0.47$} \\
             & 8  &              --- &              --- &               --- &   $1.00 ${\tiny$\pm 0.00$} &   $1.90 ${\tiny$\pm 0.30$} \\
             & 9  &              --- &              --- &               --- &   $1.00 ${\tiny$\pm 0.00$} &               --- \\
             & 10 &              --- &              --- &               --- &   $1.00 ${\tiny$\pm 0.00$} &               --- \\
             & 11 &              --- &              --- &               --- &   $1.00 ${\tiny$\pm 0.00$} &               --- \\
             & 12 &              --- &              --- &               --- &   $1.00 ${\tiny$\pm 0.00$} &               --- \\
             & 13 &              --- &              --- &               --- &   $0.90 ${\tiny$\pm 0.09$} &               --- \\
             & 14 &              --- &              --- &               --- &   $0.90 ${\tiny$\pm 0.09$} &               --- \\
\end{tabular}
\end{table}

The proposed compositional learning framework seeks to discover a set of components that are reusable across multiple tasks. To verify that this occurs, the next evaluation studied how many tasks reused each component. Taking the models pretrained via compositional and dynamic + compositional \ER{}, this evaluation measured the accuracy of the models on each task if the model discarded any individual component. A task counted as reusing a given component if removing it caused a relative drop in accuracy of more than $5\%$. Table~\ref{tab:ModuleReuse} shows the number of tasks that reused each component. Since there is no fixed ordering over components across trials, the evaluation sorted each trial's components in descending order of the number of tasks that reused each component. Moreover, for dynamic + compositional \ER{}, the results only considered components that were created across all trials for a given data set, to ensure that all averages were statistically significant. Results show that across all data sets and algorithms, multiple tasks reused all $\numModules=4$ components available from initialization. For the \Omniglot{} data set, this behavior persisted even for components that were dynamically added in the expansion step. However, this was not the case for the \CIFAR{} data set, for which  multiple tasks indeed reused the first few dynamically added components, but only a single task used subsequent ones. This indicates that the agent added those components merely for increasing performance on that individual task, but found no reusable knowledge useful for future tasks.

\begin{figure}[b!]
\centering
    \begin{subfigure}[b]{\textwidth}
        \centering
        \includegraphics[height=0.45cm, trim={0.1cm, 0.1cm, 0.1cm, 0.15cm}, clip]{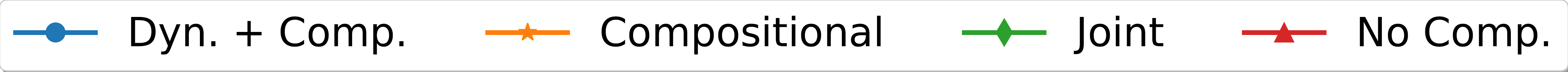}
    \end{subfigure}\\
    \begin{subfigure}[b]{0.34\textwidth}
        \includegraphics[height=4.5cm, trim={0.5cm 0.4cm 0cm 3.cm}, clip]{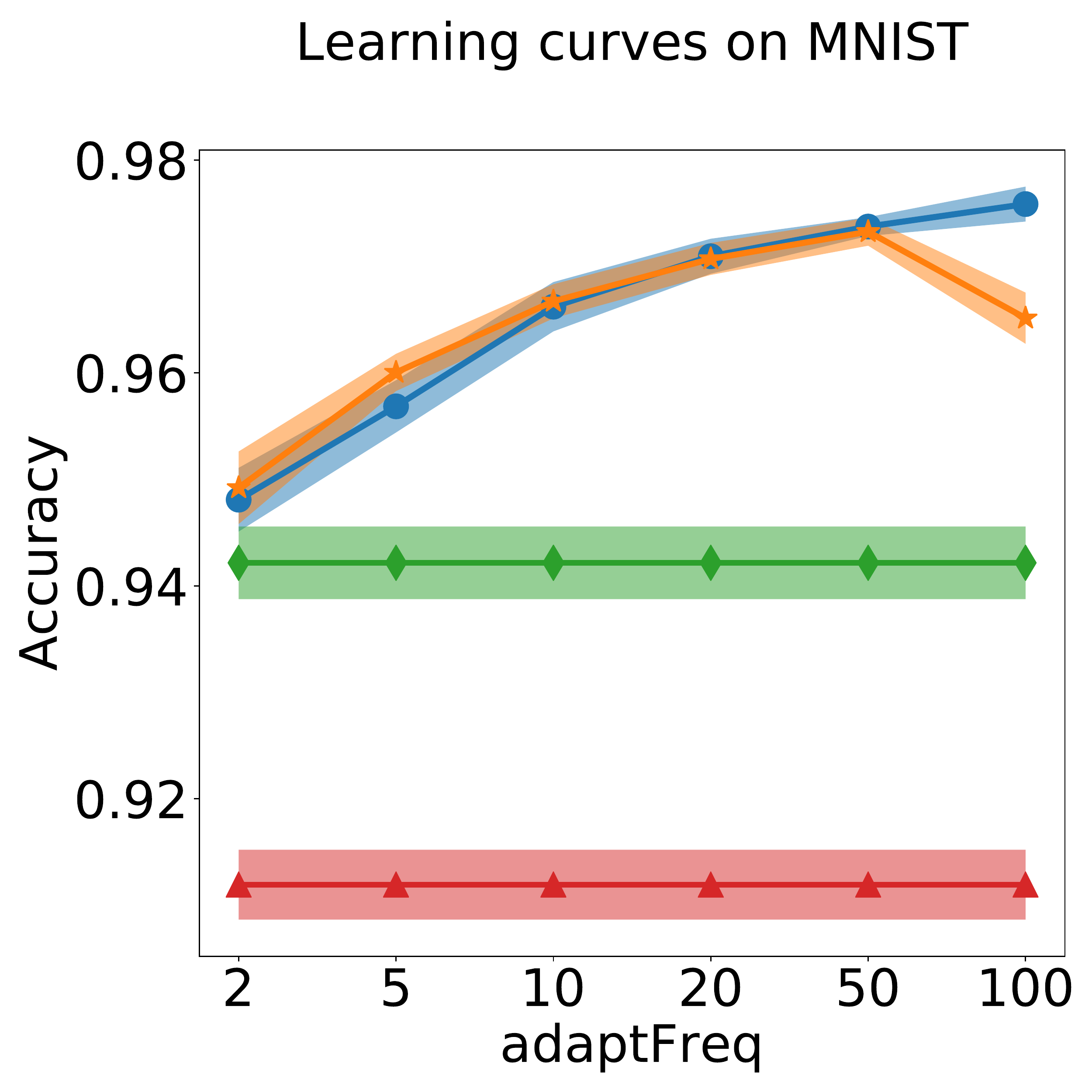}
        \caption{\ER{} on \MNIST{}}
    \end{subfigure}%
    \begin{subfigure}[b]{0.33\textwidth}
        \includegraphics[height=4.5cm, trim={2.35cm 0.4cm 0cm 3.cm}, clip]{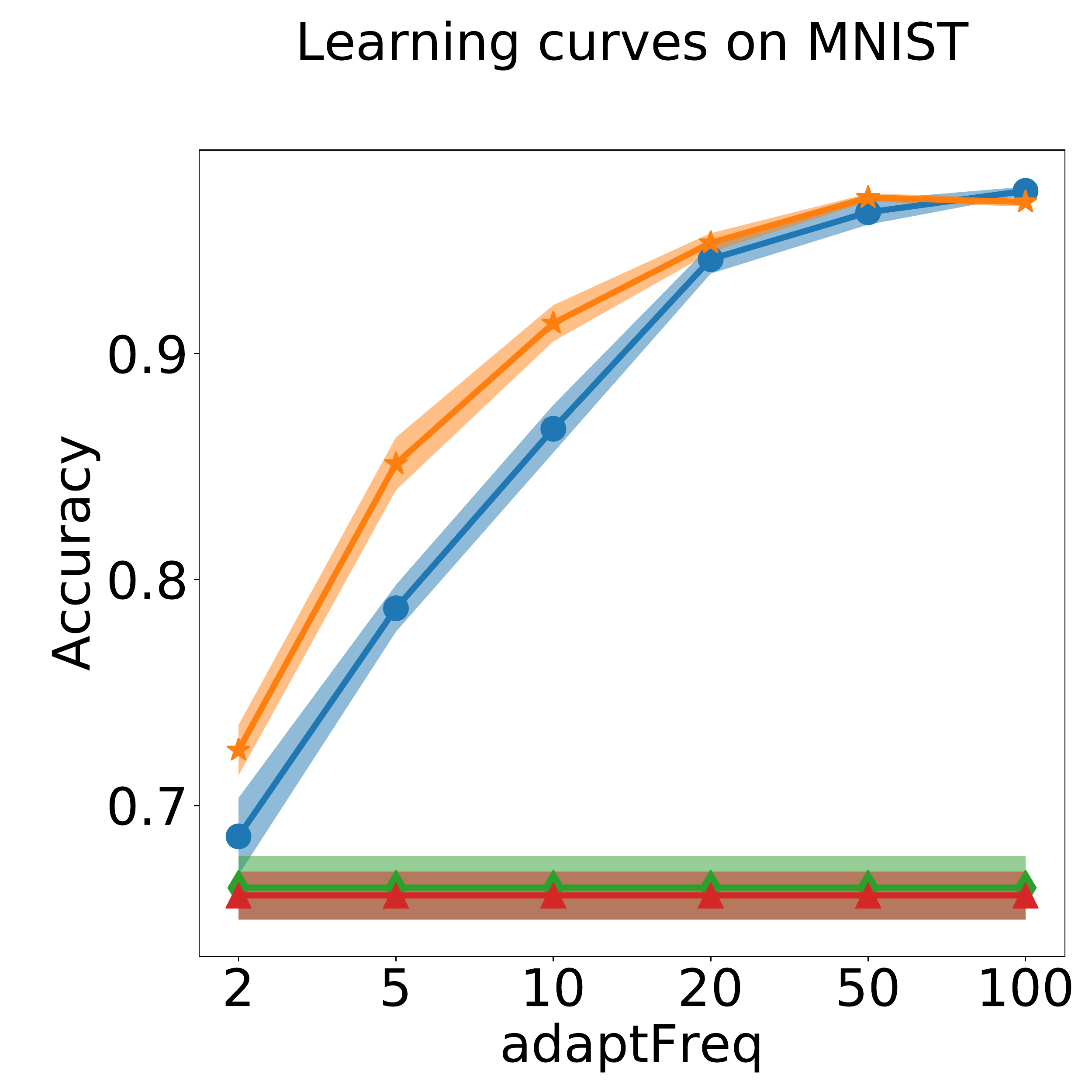}
        \caption{\kEWC{} on \MNIST{}}
    \end{subfigure}%
    \begin{subfigure}[b]{0.33\textwidth}
        \includegraphics[height=4.5cm, trim={2.35cm 0.4cm 0cm 3.cm}, clip]{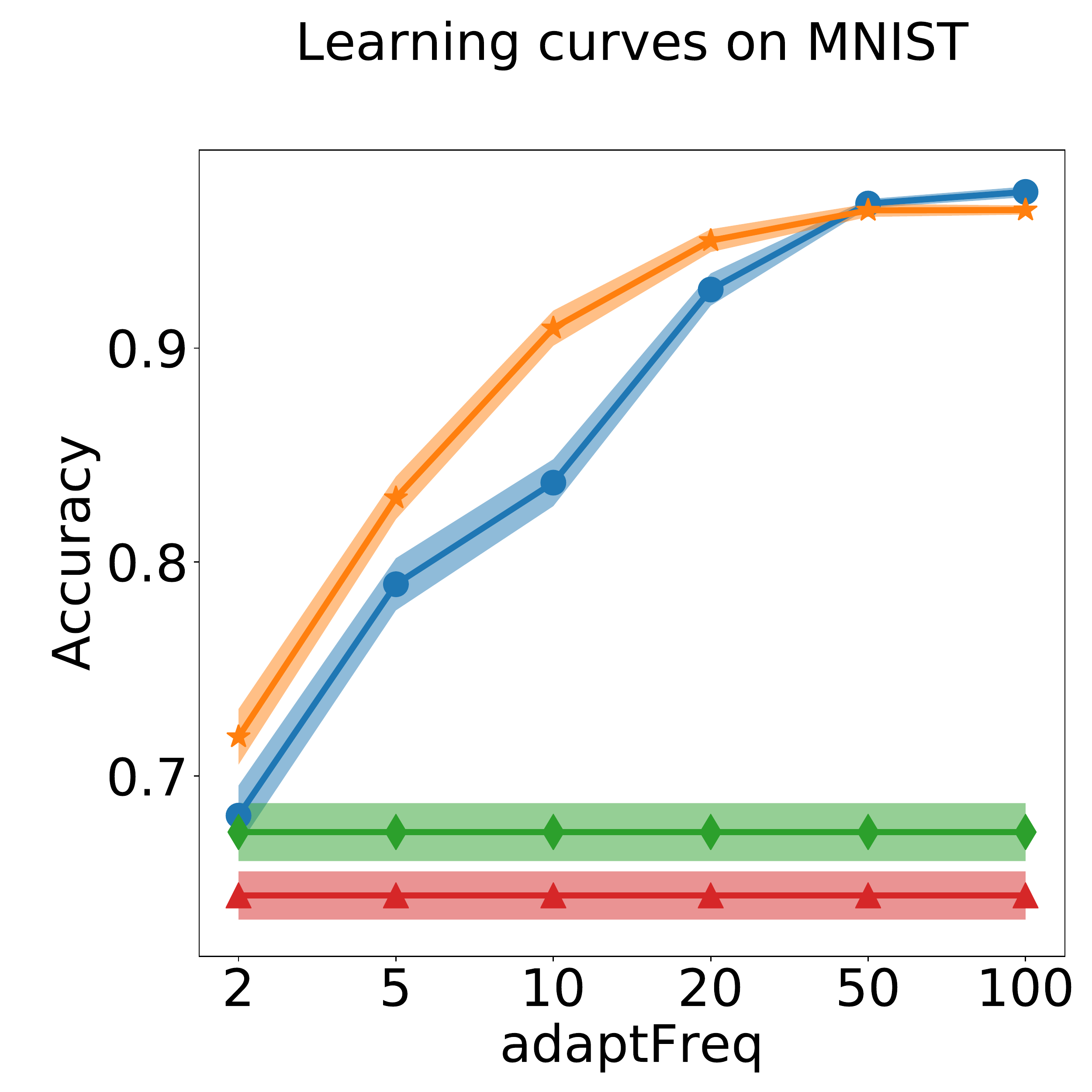}
        \caption{\VAN{} on \MNIST{}}
    \end{subfigure}\\
    \vspace{0.5em}
    \begin{subfigure}[b]{0.34\textwidth}
        \includegraphics[height=4.5cm, trim={0.5cm 0.4cm 0cm 3.cm}, clip]{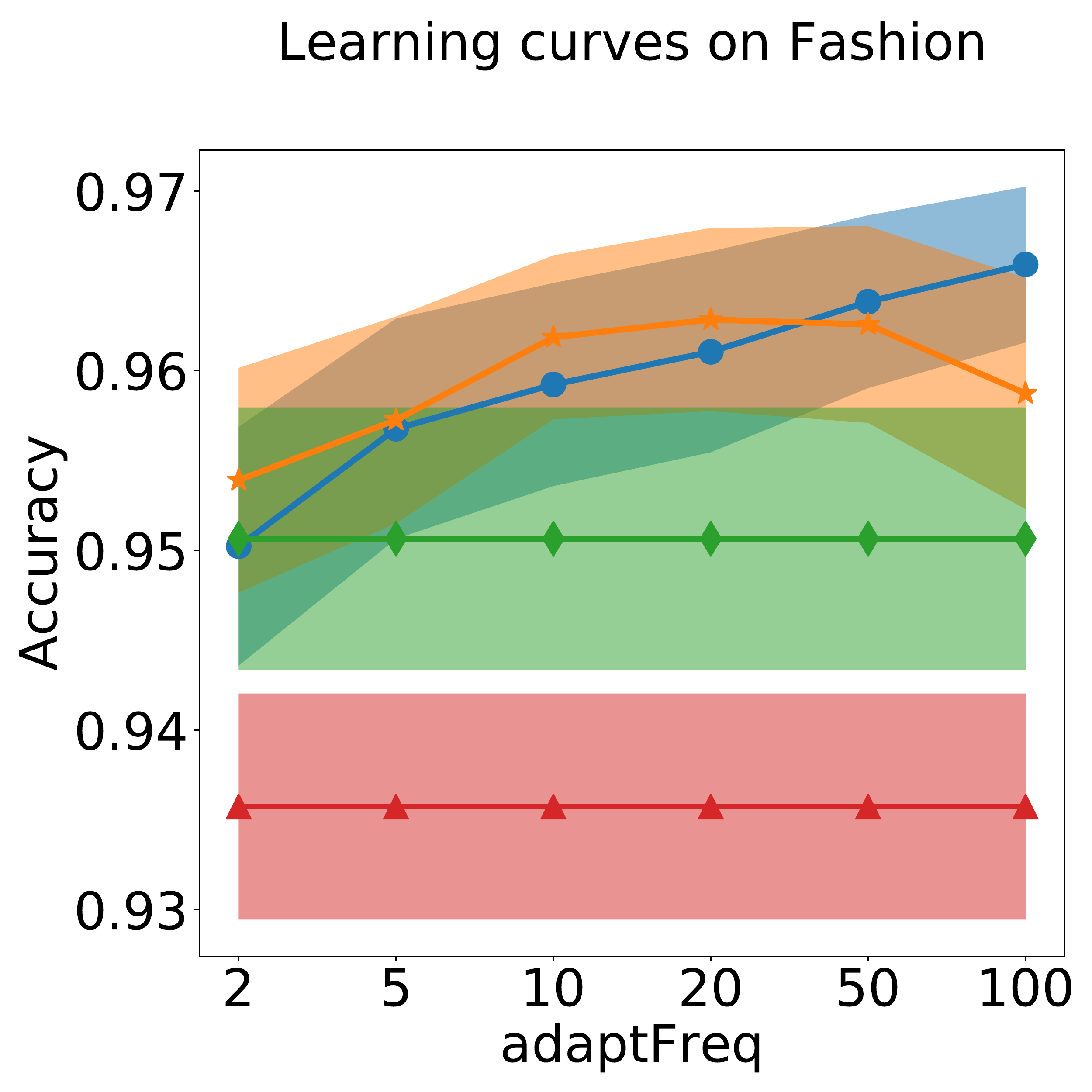}
        \caption{\ER{} on \Fashion{}}
    \end{subfigure}%
    \begin{subfigure}[b]{0.33\textwidth}
        \includegraphics[height=4.5cm, trim={1.6cm 0.4cm 0cm 3.cm}, clip]{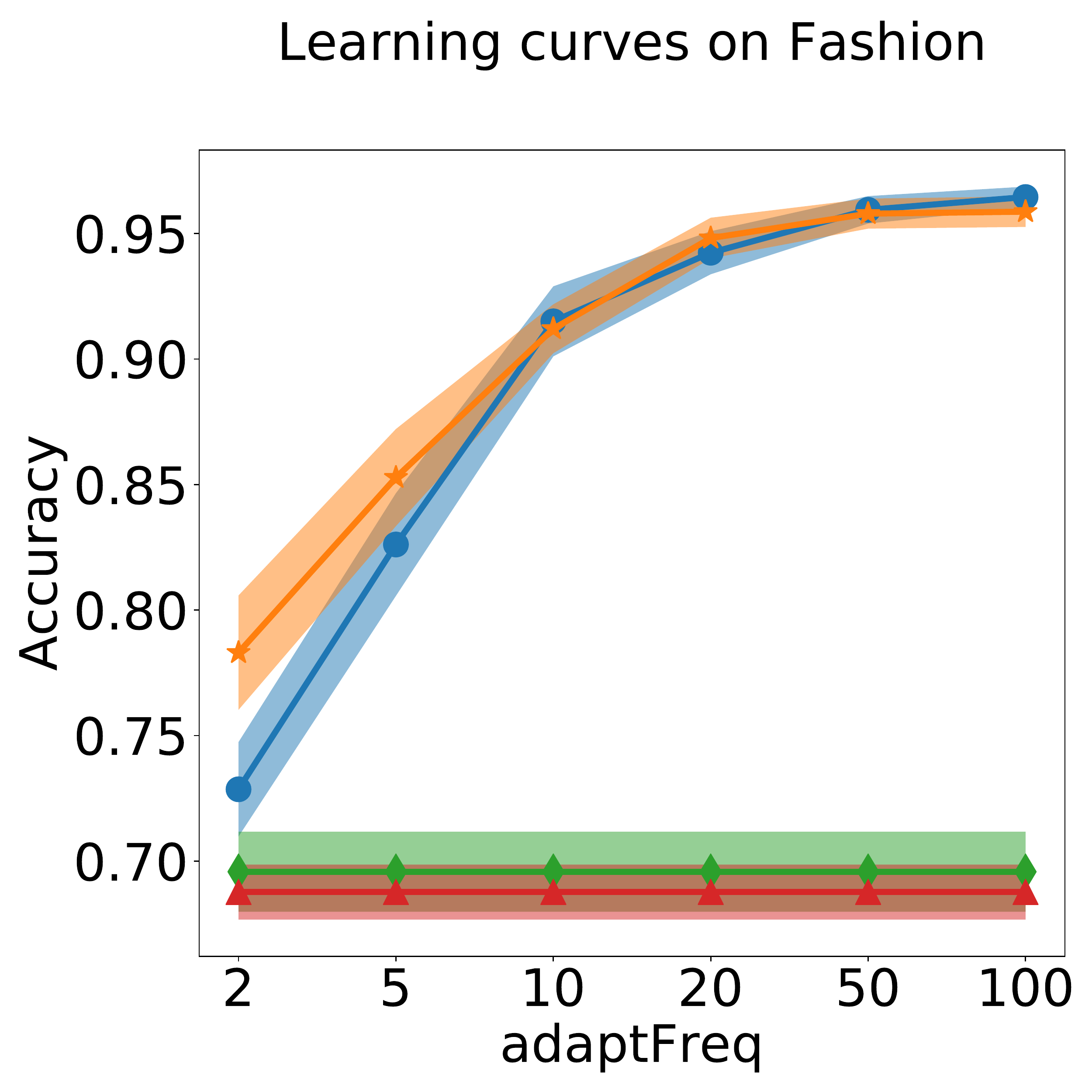}
        \caption{\kEWC{} on \Fashion{}}
    \end{subfigure}%
    \begin{subfigure}[b]{0.33\textwidth}
        \includegraphics[height=4.5cm, trim={1.6cm 0.4cm 0cm 3.cm}, clip]{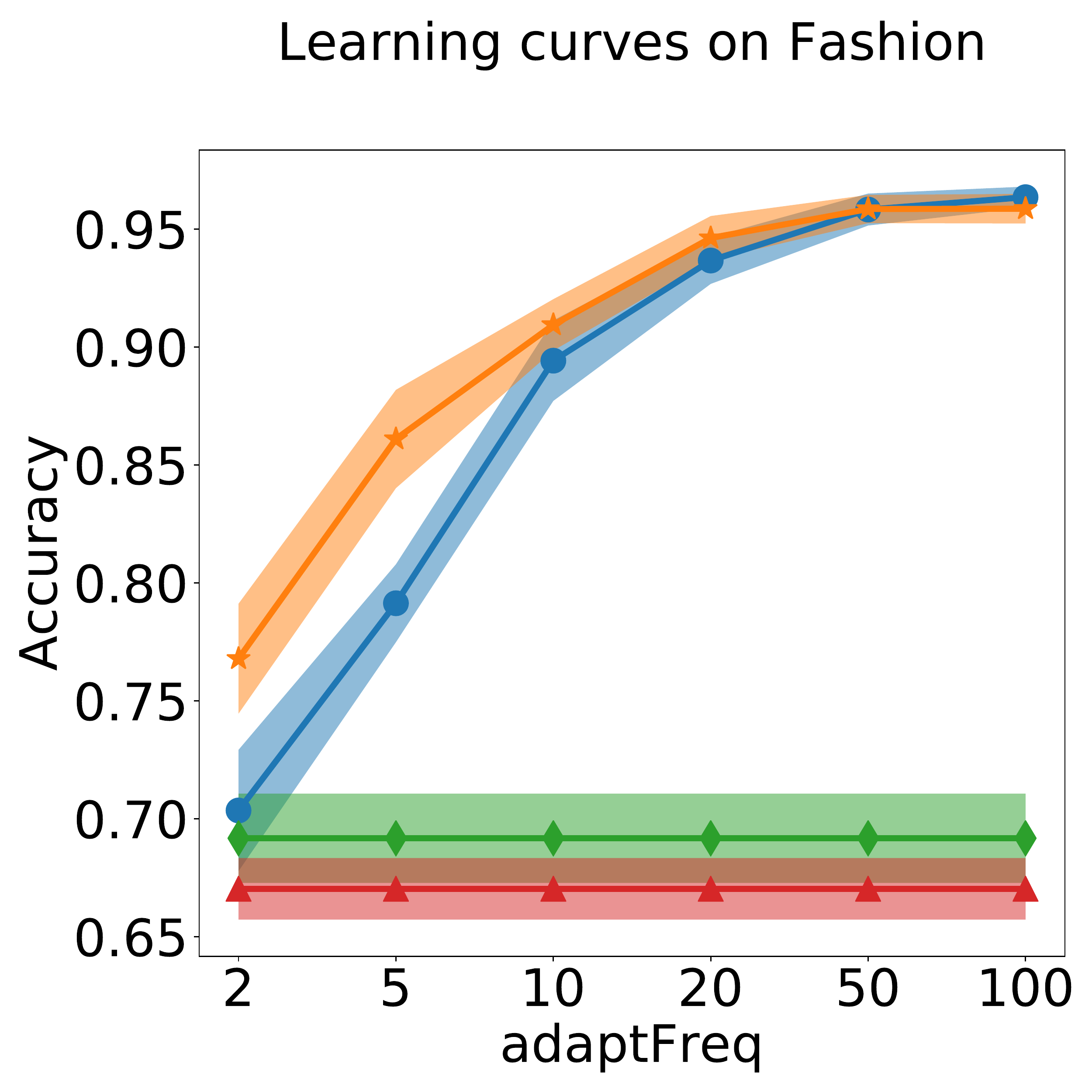}
        \caption{\VAN{} on \Fashion{}}
    \end{subfigure}\\
    \vspace{0.5em}
    \begin{subfigure}[b]{0.34\textwidth}
        \includegraphics[height=4.5cm, trim={0.5cm 0.4cm 0cm 3.cm}, clip]{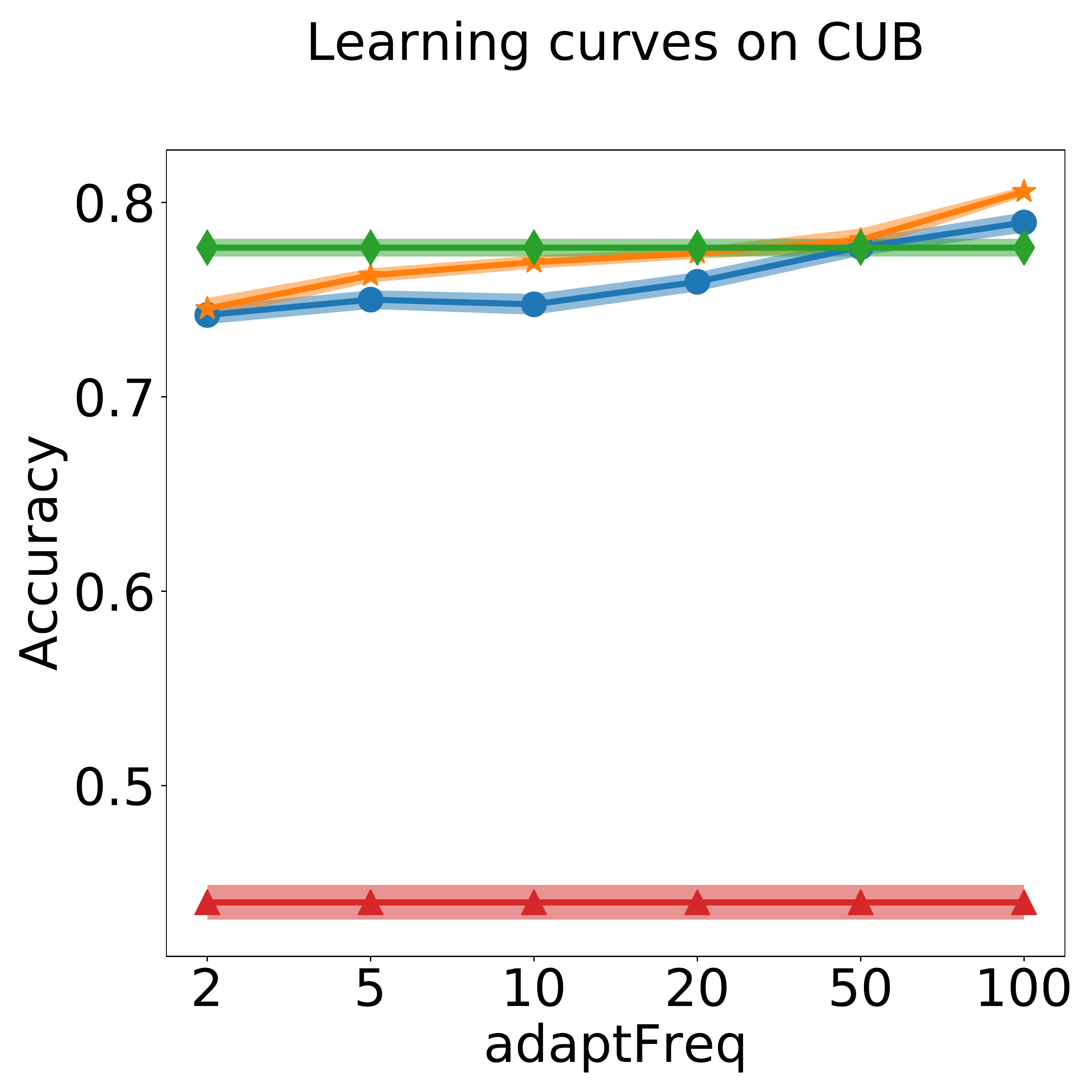}
        \caption{\ER{} on \CUB{}}
    \end{subfigure}%
    \begin{subfigure}[b]{0.33\textwidth}
        \includegraphics[height=4.5cm, trim={1.6cm 0.4cm 0cm 3.cm}, clip]{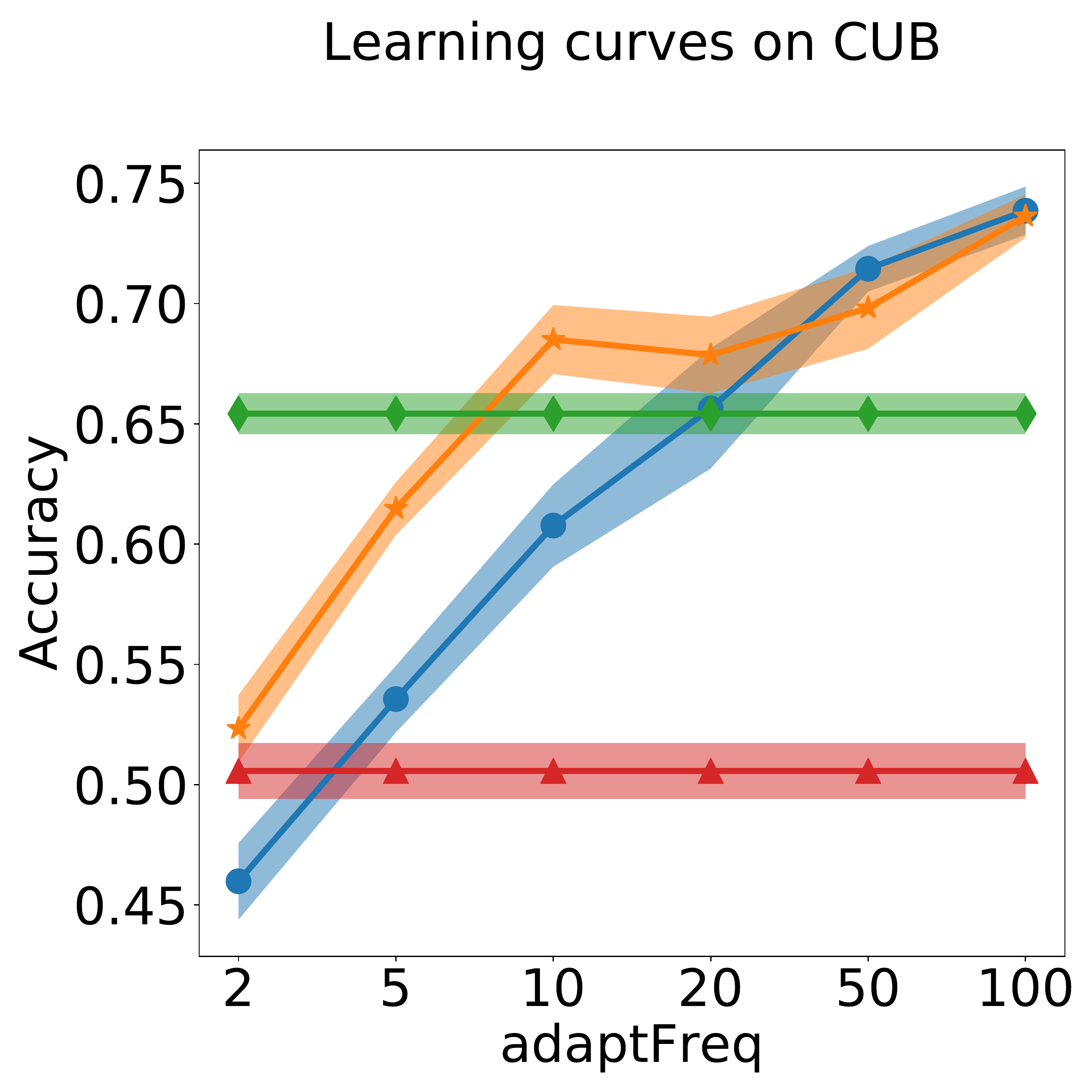}
        \caption{\kEWC{} on \CUB{}}
    \end{subfigure}%
    \begin{subfigure}[b]{0.33\textwidth}
        \includegraphics[height=4.5cm, trim={2.35cm 0.4cm 0cm 3.cm}, clip]{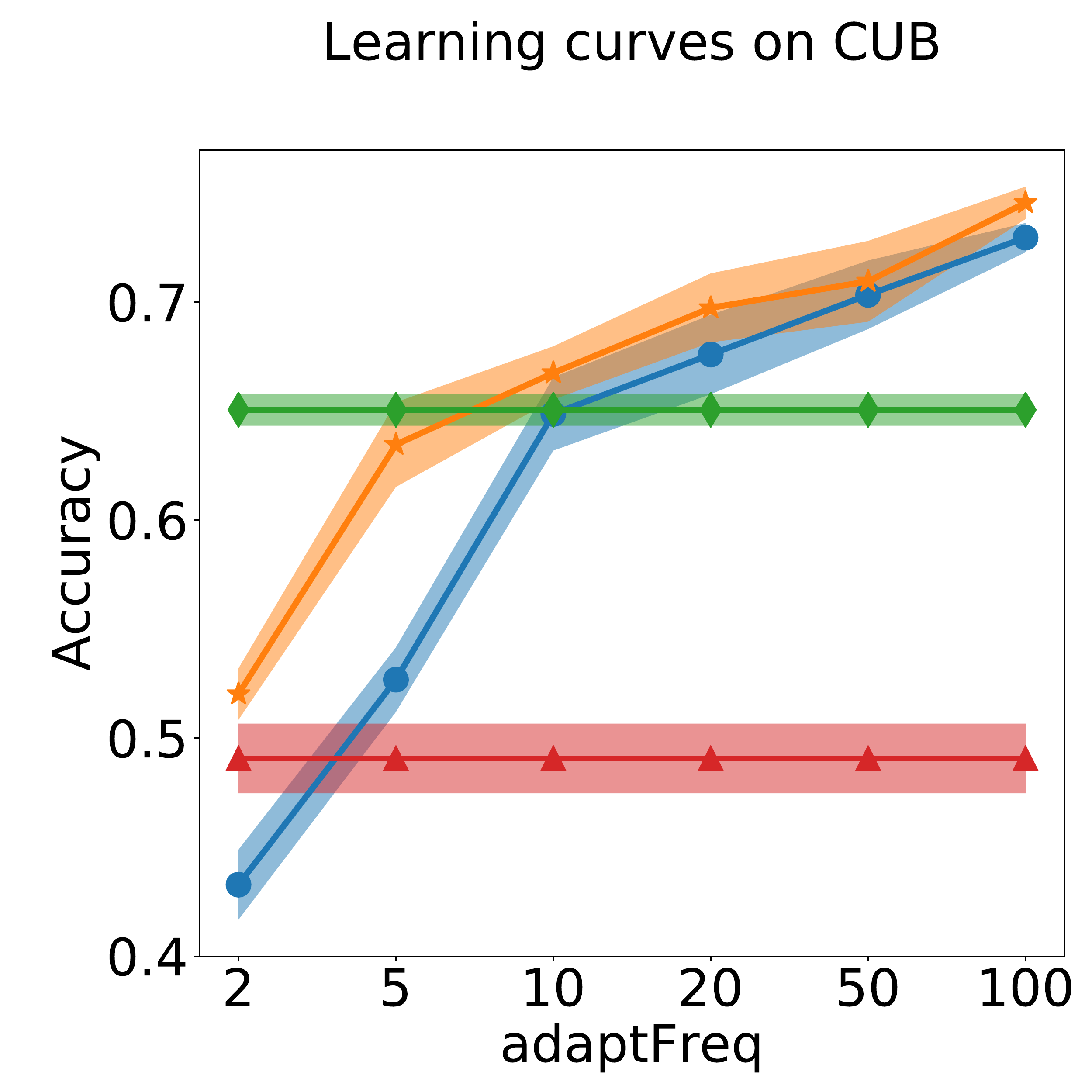}
        \caption{\VAN{} on \CUB{}}
    \end{subfigure}
    \caption[Sensitivity to the rate of assimilation vs.~accommodation in supervised lifelong composition.]{Effect of the assimilation and accommodation schedule. Average accuracy across tasks with respect to the number of assimilation epochs between accommodation epochs. Broadly, methods under the proposed framework performed better with a scheduled that favored stability, taking more assimilation steps before accommodating any new knowledge into the set of existing components.}
    \label{fig:ScheduleCurves}
\end{figure}

When designing algorithms under the framework, one needs to choose how to alternate the processes of assimilation and accommodation. Most experiments so far considered the simplest case, where the learning carries out adaptation entirely after finishing assimilation. However, it is possible that other choices yield better results, enabling the learner to incorporate knowledge about the current task that further enables it to assimilate it better. To study this question, the evaluation executed additional experiments using \ER{} variants on the \MNIST{}, \Fashion{}, and \CUB{} data sets with soft layer ordering. Instead of executing the adaptation step only after completing assimilation, the agent alternated epochs of assimilation with epochs of adaptation with various frequencies. Figure~\ref{fig:ScheduleCurves} displays the obtained results. Generally, it was beneficial to carry out adaptation steps infrequently, with a clear increasing trend in performance as the learner took more assimilation steps before each adaptation step. For \MNIST{} and \Fashion{}, all choices of schedule led to improved performance over baselines, highlighting the benefits of splitting the learning process into assimilation and accommodation. For \CUB{}, the results were more nuanced, with very fast accommodation rates achieving lower accuracy than the baselines. This is consistent with the results in Table~\ref{tab:softOrderingResults}, where compositional \FM{}---equivalent to compositional \ER{} with  a schedule of infinite assimilation steps per accommodation step---performed nearly as well as compositional \ER{} with a single adaptation epoch. 

\subsubsection{Deep Compositional Learning With Soft Gating}
\label{sec:ResultsSoftGating}

\begin{table}[b!]
    \centering
    \captionof{table}[Performance of supervised lifelong composition: soft gating.]{Average final accuracy across all tasks using soft gating. The gap between compositional and dynamic + compositional methods was smallar than with soft ordering, and in particular \FM{} methods were competitive, due to the increased flexibility of the gating approach. Standard errors across ten seeds reported after the $\pm$.}
    \label{tab:softGatingResults}
    \begin{tabular}{@{}l|l|c|c|c|c@{}}
Base & Algorithm & \MNIST & \Fashion & \CIFAR & \Omniglot\\
\hline
\hline
\multirow{4}{*}{\ER} & Dyn.~+ Comp. & $\bf{98.2}${\tiny$\bf{\pm0.1}$}$\%$ & $\bf{97.1}${\tiny$\bf{\pm0.4}$}$\%$ & $74.9${\tiny$\pm0.3$}$\%$ & $73.7${\tiny$\pm0.3$}$\%$\\
& Compositional & $98.0${\tiny$\pm0.2$}$\%$ & $97.0${\tiny$\pm0.4$}$\%$ & $\bf{75.9}${\tiny$\bf{\pm0.4}$}$\%$ & $\bf{73.9}${\tiny$\bf{\pm0.3}$}$\%$\\
& Joint & $93.8${\tiny$\pm0.3$}$\%$ & $94.6${\tiny$\pm0.7$}$\%$ & $72.0${\tiny$\pm0.4$}$\%$ & $72.6${\tiny$\pm0.2$}$\%$\\
& No Comp. & $91.2${\tiny$\pm0.3$}$\%$ & $93.6${\tiny$\pm0.6$}$\%$ & $51.6${\tiny$\pm0.6$}$\%$ & $43.2${\tiny$\pm4.2$}$\%$\\
\hline
\multirow{4}{*}{\kEWC} & Dyn.~+ Comp. & $\bf{98.2}${\tiny$\bf{\pm0.1}$}$\%$ & $\bf{97.0}${\tiny$\bf{\pm0.4}$}$\%$ & $76.6${\tiny$\pm0.5$}$\%$ & $73.6${\tiny$\pm0.4$}$\%$\\
& Compositional & $98.0${\tiny$\pm0.2$}$\%$ & $\bf{97.0}${\tiny$\bf{\pm0.4}$}$\%$ & $\bf{76.9}${\tiny$\bf{\pm0.3}$}$\%$ & $\bf{74.6}${\tiny$\bf{\pm0.2}$}$\%$\\
& Joint & $68.6${\tiny$\pm0.9$}$\%$ & $69.5${\tiny$\pm1.8$}$\%$ & $49.9${\tiny$\pm1.1$}$\%$ & $63.5${\tiny$\pm1.2$}$\%$\\
& No Comp. & $66.0${\tiny$\pm1.1$}$\%$ & $68.8${\tiny$\pm1.1$}$\%$ & $36.0${\tiny$\pm0.7$}$\%$ & $68.8${\tiny$\pm0.4$}$\%$\\
\hline
\multirow{4}{*}{\VAN} & Dyn.~+ Comp. & $\bf{98.2}${\tiny$\bf{\pm0.1}$}$\%$ & $\bf{97.1}${\tiny$\bf{\pm0.4}$}$\%$ & $66.6${\tiny$\pm0.7$}$\%$ & $69.1${\tiny$\pm0.9$}$\%$\\
& Compositional & $98.0${\tiny$\pm0.2$}$\%$ & $96.9${\tiny$\pm0.5$}$\%$ & $\bf{68.2}${\tiny$\bf{\pm0.5}$}$\%$ & $\bf{72.1}${\tiny$\bf{\pm0.3}$}$\%$\\
& Joint & $67.3${\tiny$\pm1.7$}$\%$ & $66.4${\tiny$\pm1.9$}$\%$ & $51.0${\tiny$\pm0.8$}$\%$ & $65.8${\tiny$\pm1.3$}$\%$\\
& No Comp. & $64.4${\tiny$\pm1.1$}$\%$ & $67.0${\tiny$\pm1.3$}$\%$ & $36.6${\tiny$\pm0.6$}$\%$ & $68.9${\tiny$\pm1.0$}$\%$\\
\hline
\multirow{2}{*}{\FM} & Dyn.~+ Comp. & $\bf{98.4}${\tiny$\bf{\pm0.1}$}$\%$ & $\bf{97.0}${\tiny$\bf{\pm0.4}$}$\%$ & $\bf{77.2}${\tiny$\bf{\pm0.3}$}$\%$ & $74.0${\tiny$\pm0.4$}$\%$\\
& Compositional & $94.8${\tiny$\pm0.4$}$\%$ & $96.3${\tiny$\pm0.4$}$\%$ & $\bf{77.2}${\tiny$\bf{\pm0.3}$}$\%$ & $\bf{74.1}${\tiny$\bf{\pm0.3}$}$\%$\\
    \end{tabular}
\end{table} 

\begin{figure}[t!]
\centering
\captionsetup[subfigure]{aboveskip=1pt}
    \begin{subfigure}[b]{0.315\textwidth}
        \includegraphics[height=1.4cm, trim={0.35cm 0cm 0cm 1.6cm}, clip]{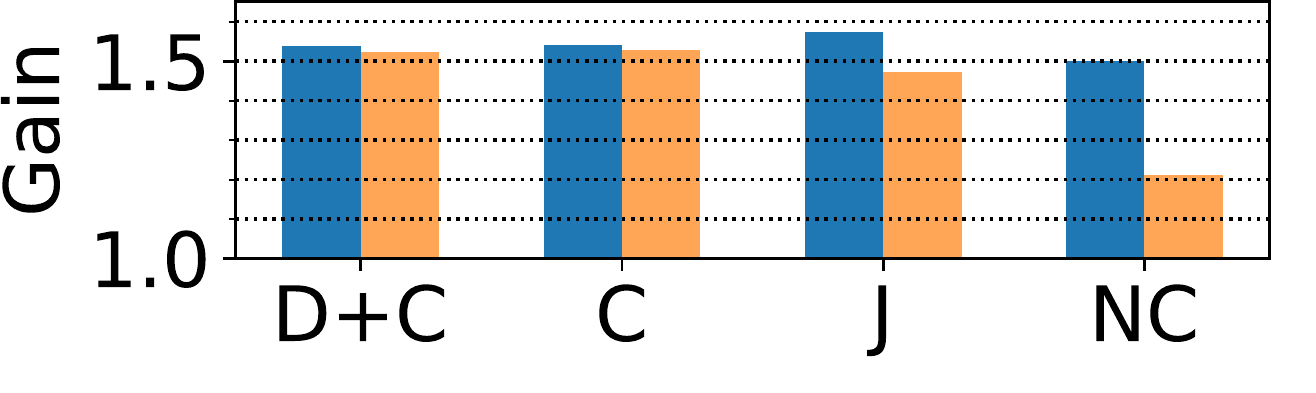}
        \caption{\ER}
    \end{subfigure}%
    \begin{subfigure}[b]{0.285\textwidth}
        \includegraphics[height=1.4cm, trim={1.3cm 0cm 0cm 1.6cm}, clip]{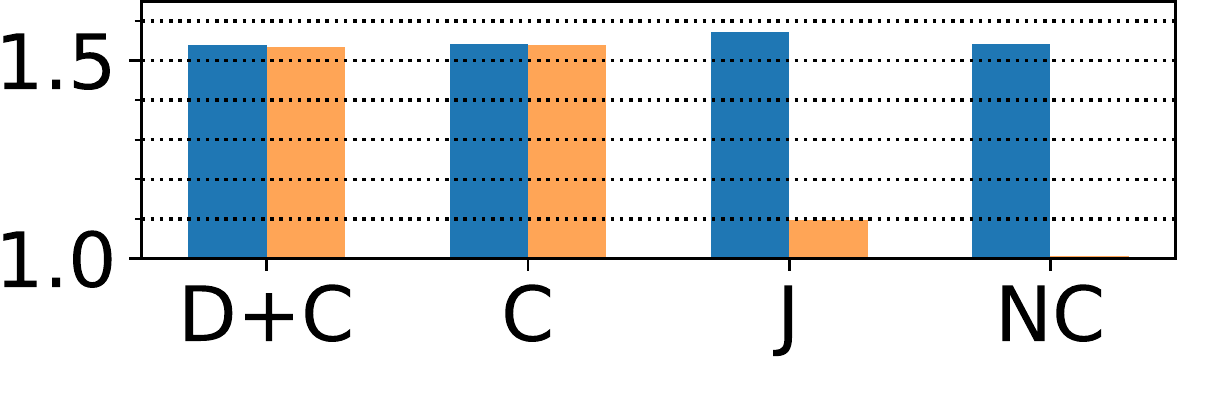}
         \caption{\kEWC}
    \end{subfigure}%
    \begin{subfigure}[b]{0.285\textwidth}
        \includegraphics[height=1.4cm, trim={1.3cm 0cm 0cm 1.6cm}, clip]{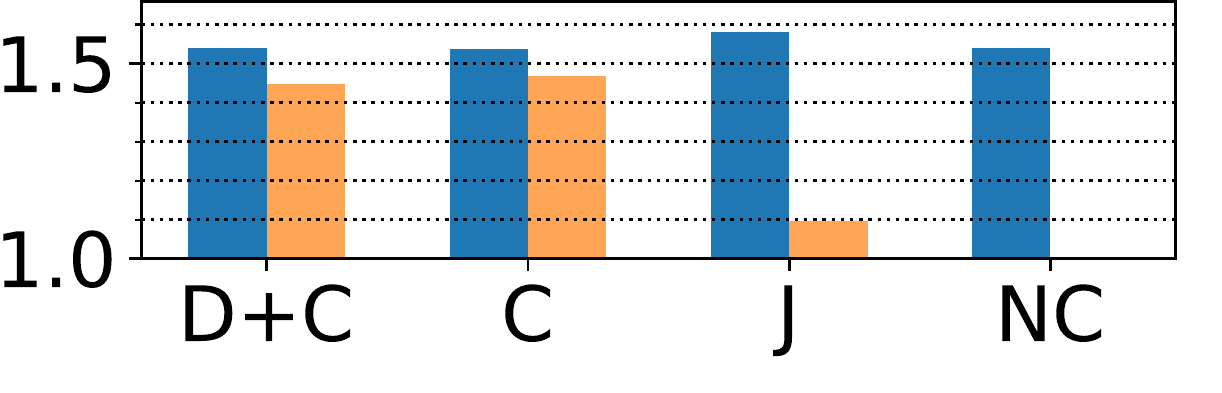}
        \caption{\VAN}
    \end{subfigure}%
    \begin{subfigure}[b]{0.115\textwidth}
        \centering
        \raisebox{2.7em}{\includegraphics[width=0.9\linewidth, trim={0.1cm, 0.1cm, 0.1cm, 0.1cm}, clip]{chapter3/Figures/barcharts/bar_legend.pdf}}
    \end{subfigure}
    \caption[Performance of supervised lifelong composition after each task and after all tasks: soft gating.]{Average gain with respect to no-components \VAN{} across tasks and data sets using soft ordering, immediately after training on each task (forward) and after training on all tasks (final), using soft ordering (top) and soft gating (bottom). Algorithms within the proposed framework (C and D+C) outperformed baselines. Gaps between forward and final performance indicate that the framework exhibits less forgetting.}
    \label{fig:softGatingBars}
\end{figure}

\begin{figure}[b!]
\centering
\captionsetup[subfigure]{aboveskip=5pt}
    \begin{subfigure}[b]{\textwidth}
        \centering
        \includegraphics[height=0.45cm, trim={0.1cm, 0.1cm, 0.1cm, 0.15cm}, clip]{chapter3/Figures/catastrophic_forgetting/catastrophic_forgetting_legend.pdf}
    \end{subfigure}\\
    \begin{subfigure}[b]{0.34\textwidth}
        \includegraphics[height=4.5cm, trim={0.5cm 0.4cm 0cm 3.cm}, clip]{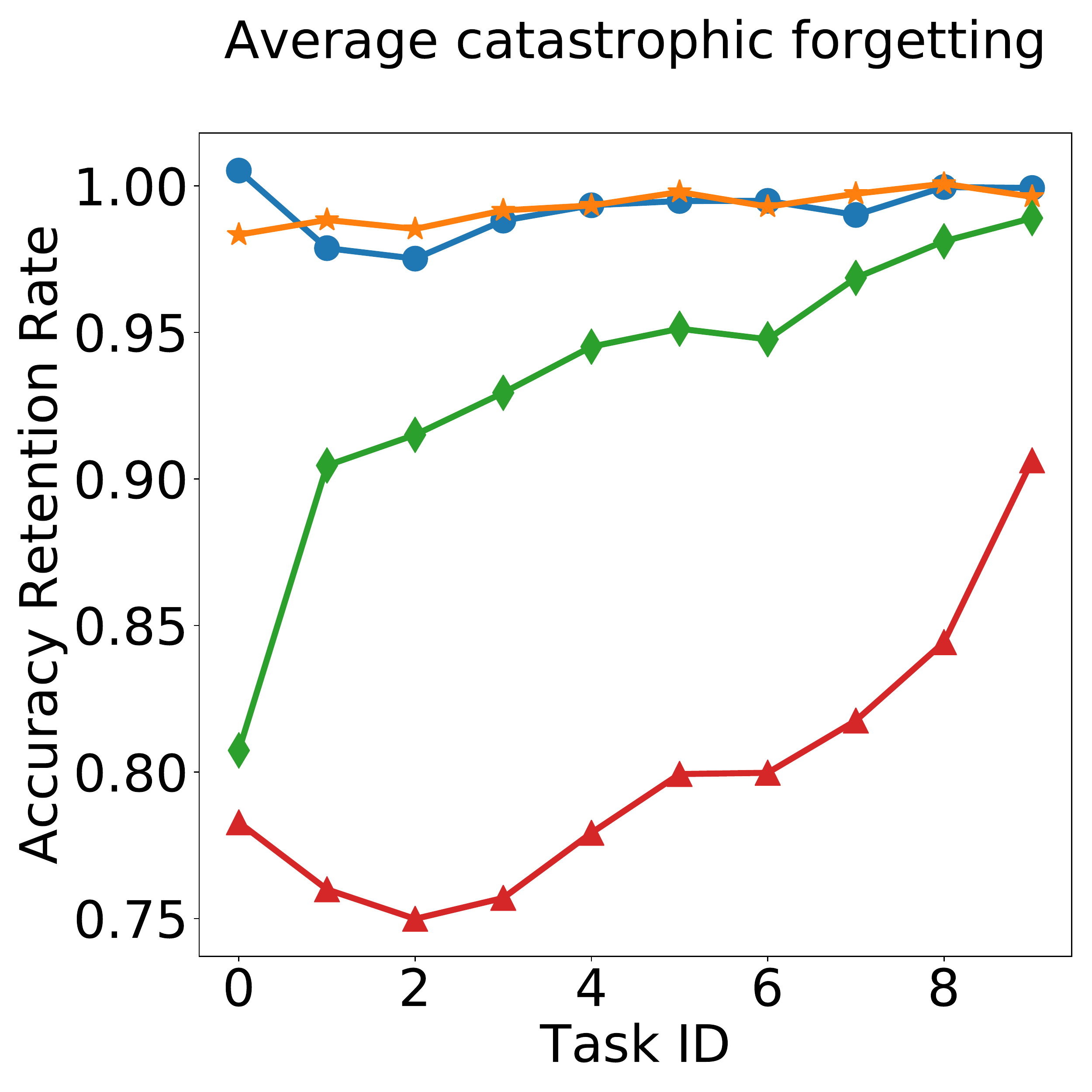}
        \caption{\ER}
    \end{subfigure}%
    \begin{subfigure}[b]{0.33\textwidth}
        \includegraphics[height=4.5cm, trim={1.6cm 0.4cm 0cm 3.cm}, clip]{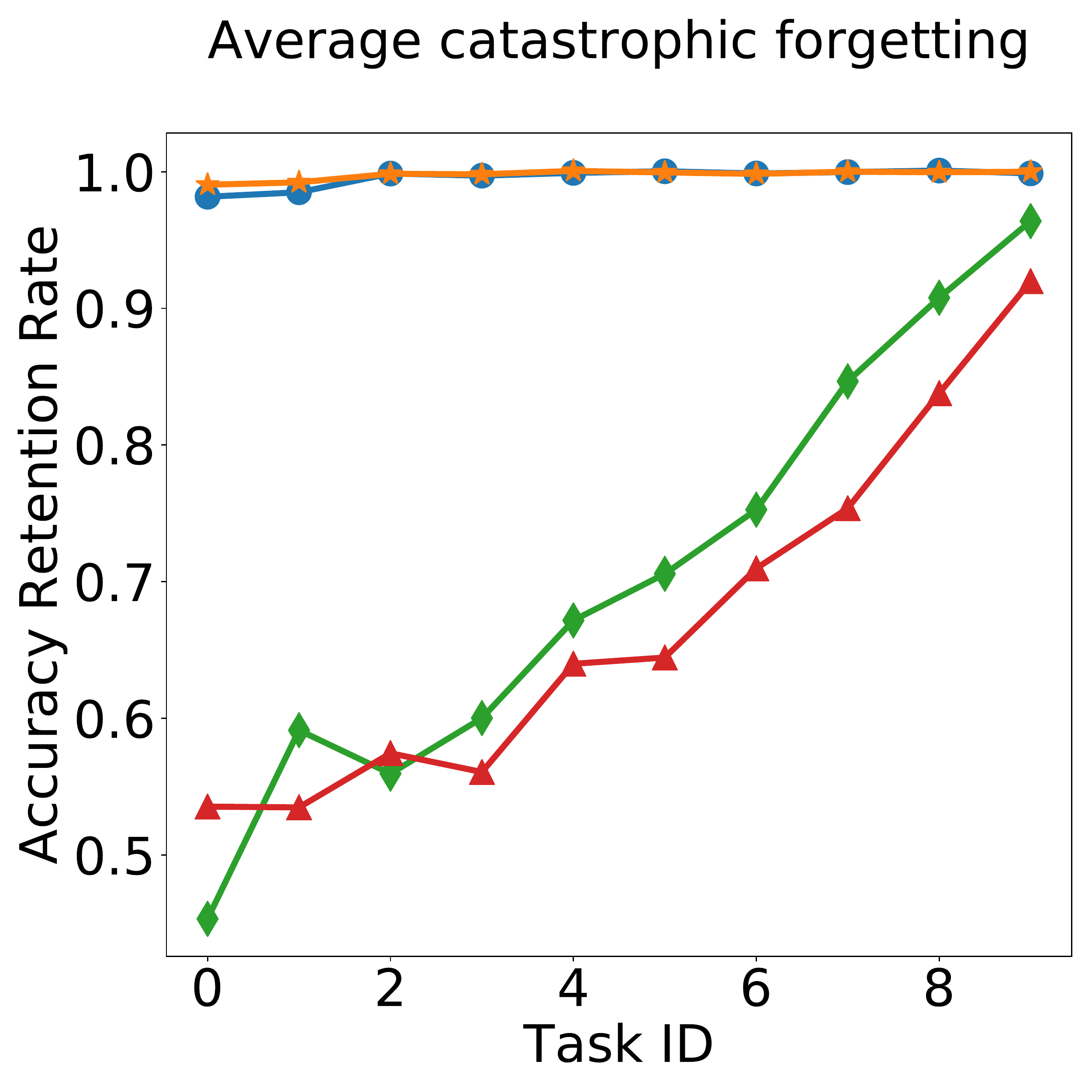}
        \caption{\kEWC}
    \end{subfigure}%
    \begin{subfigure}[b]{0.33\textwidth}
        \includegraphics[height=4.5cm, trim={1.6cm 0.4cm 0cm 3.cm}, clip]{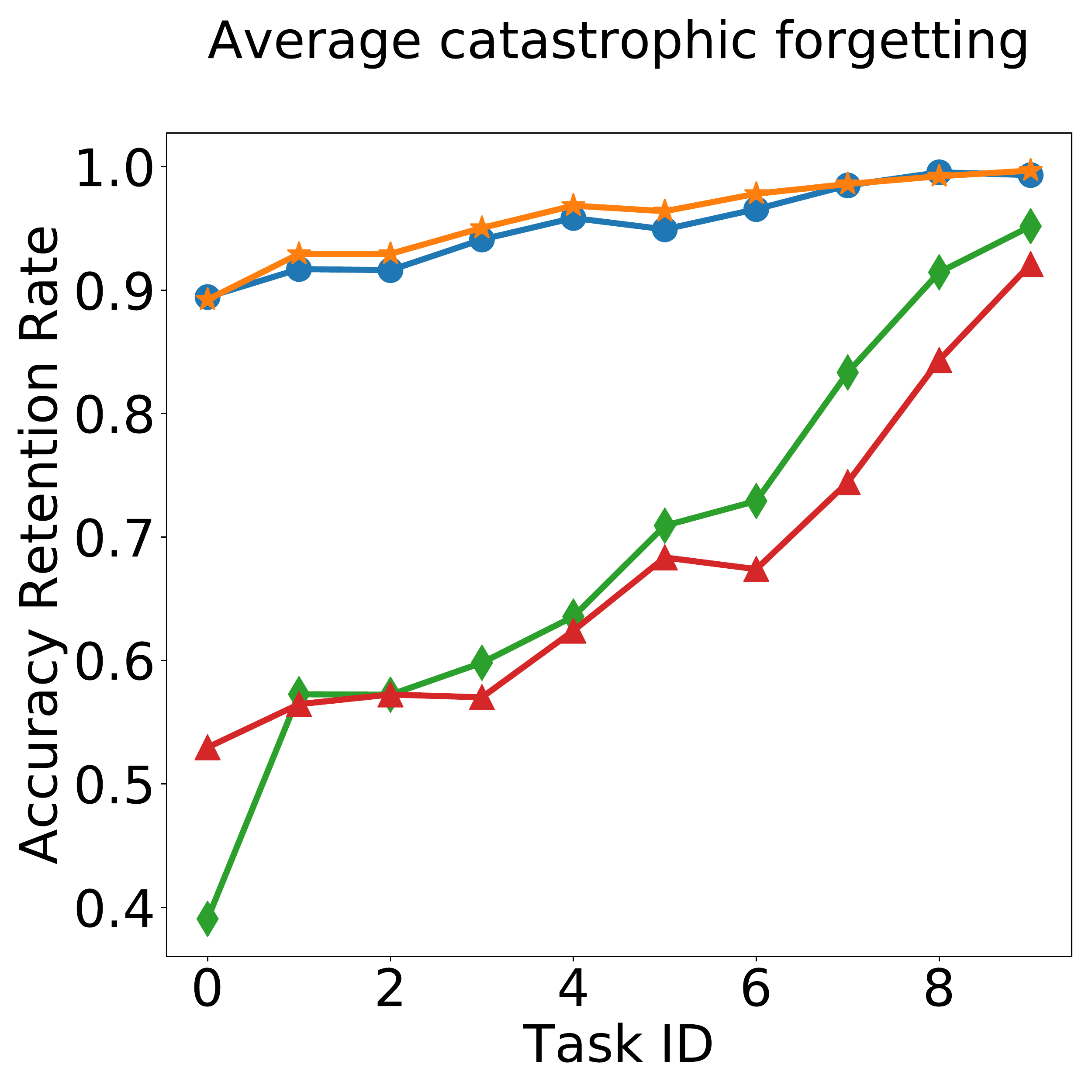}
        \caption{\VAN}
    \end{subfigure}
    \caption[Catastrophic forgetting on supervised lifelong composition: soft gating.]{Catastrophic forgetting across data sets using soft gating. Ratio of accuracy immediately after learning a task to after learning all tasks. For data sets with more than ten tasks, the evaluation sampled ten interleaved tasks to match all the x-axes. Compositional algorithms had practically no forgetting, whereas jointly trained and no-components baselines forgot knowledge required to solve earlier tasks.}
    \label{fig:catastrophicForgettingSoftGating}
\end{figure}

Having completed an extensive evaluation on the performance of compositional algorithms using soft ordering nets as the compositional structures, a similar study considered using the proposed algorithms for training soft gating nets. Table~\ref{tab:softGatingResults} shows a substantial improvement of compositional algorithms with respect to baselines in overall performance. One reasonable hypothesis is that the gating net should grant the assimilation step more flexibility, which is confirmed in Figure~\ref{fig:softGatingBars}---the forward accuracy of compositional methods was nearly identical to that of jointly trained and no-components versions. This added flexibility enabled the simplest version of a compositional algorithm, \FM{}, to perform better than the full versions of the algorithm on the \CIFAR{} data set with convolutional gating nets, showing that even the components initialized with only a few tasks are sufficient for top lifelong learning performance. Early experiments with this method on the \CUB{} data set revealed that all algorithms (including the baselines) were incapable of generalizing to test data. This is consistent with findings in prior work, which showed that gating nets require vast amounts of data, unavailable on \CUB{}~\citep{rosenbaum2018routing, kirsch2018modular}.

Results in Figure~\ref{fig:catastrophicForgettingSoftGating} show the knowledge retention ratio for all tested methods using soft gating nets, once more revealing that compositional methods exhibit substantially less catastrophic forgetting, particularly for the earlier tasks seen during training. On the other hand, Table~\ref{tab:numberOfLearnedComponentsSoftGating} shows that soft gating nets typically required adding fewer new components than soft ordering nets, which is to be expected, since the gating structure gives each component substantially more flexibility to adapt to a larger number of tasks. Notably, this result holds even for \FM{}, which does not get to adapt existing modules to new tasks. 

\begin{table}[t!]
    \centering
    \caption[Number of learned components in supervised lifelong composition: soft gating.]{Number of learned components using soft gating. The model growth was substantially more controlled with this more flexible architecture. Standard errors across ten seeds reported after the $\pm$.}
    \label{tab:numberOfLearnedComponentsSoftGating}
    \begin{tabular}{l|c|c|c|c}
Base & \MNIST & \Fashion  & \CIFAR & \Omniglot\\
\hline\hline
\ER & $4.0${\tiny$\pm0.0$} & $4.2${\tiny$\pm0.1$} & $4.1${\tiny$\pm0.1$} & $7.1${\tiny$\pm0.4$}\\
\kEWC & $4.1${\tiny$\pm0.1$} & $4.0${\tiny$\pm0.0$} & $4.8${\tiny$\pm0.2$} & $7.4${\tiny$\pm0.4$}\\
\VAN & $4.1${\tiny$\pm0.1$} & $4.2${\tiny$\pm0.1$} & $4.1${\tiny$\pm0.1$} & $7.2${\tiny$\pm0.3$}\\
\FM & $5.4${\tiny$\pm0.2$} & $4.7${\tiny$\pm0.2$} & $4.4${\tiny$\pm0.2$} & $7.3${\tiny$\pm0.4$}\\
    \end{tabular}
\end{table}

 \begin{table}[t!]
    \centering
    \caption[Performance of supervised lifelong composition on \Combined{} data set: soft ordering, all methods.]{Average final accuracy across tasks on the \Combined{} data set, which consists of highly varied tasks. Each column shows accuracy on the subset of tasks from each given data set, as labeled. No-components methods were incapable of learning \CUB{} tasks, and the full dynamic + compositional version of the framework with \ER{} achieved the highest performance. Standard errors across ten seeds reported after the $\pm$.}
    \label{tab:CombinedAll}
    \begin{tabular}{@{}l|l|c|c|c|c@{}}
Base & Algorithm & \All & \MNIST & \Fashion & \CUB\\
\hline
\hline
\multirow{4}{*}{\ER} & Dyn.~+ Comp. & $\bf{86.5}${\tiny$\bf{\pm1.8}$}$\%$ & $\bf{99.5}${\tiny$\bf{\pm0.0}$}$\%$ & $\bf{98.0}${\tiny$\bf{\pm0.3}$}$\%$ & $\bf{74.2}${\tiny$\bf{\pm2.0}$}$\%$\\
 & Compositional & $82.1${\tiny$\pm2.5$}$\%$ & $\bf{99.5}${\tiny$\bf{\pm0.0}$}$\%$ & $97.8${\tiny$\pm0.3$}$\%$ & $65.5${\tiny$\pm2.4$}$\%$\\
 & Joint & $72.8${\tiny$\pm4.1$}$\%$ & $98.9${\tiny$\pm0.3$}$\%$ & $97.0${\tiny$\pm0.7$}$\%$ & $47.6${\tiny$\pm6.2$}$\%$\\
 & No Comp. & $47.4${\tiny$\pm4.5$}$\%$ & $91.8${\tiny$\pm1.3$}$\%$ & $83.5${\tiny$\pm2.5$}$\%$ & $7.1${\tiny$\pm0.4$}$\%$\\
\hline
\multirow{4}{*}{\kEWC} & Dyn.~+ Comp. & $\bf{75.1}${\tiny$\bf{\pm3.2}$}$\%$ & $98.7${\tiny$\pm0.5$}$\%$ & $\bf{97.1}${\tiny$\bf{\pm0.7}$}$\%$ & $\bf{52.4}${\tiny$\bf{\pm2.9}$}$\%$\\
 & Compositional & $71.3${\tiny$\pm4.0$}$\%$ & $\bf{99.4}${\tiny$\bf{\pm0.0}$}$\%$ & $96.1${\tiny$\pm0.9$}$\%$ & $44.8${\tiny$\pm3.5$}$\%$\\
 & Joint & $52.2${\tiny$\pm5.0$}$\%$ & $85.1${\tiny$\pm5.5$}$\%$ & $88.6${\tiny$\pm3.8$}$\%$ & $17.5${\tiny$\pm1.5$}$\%$\\
 & No Comp. & $28.9${\tiny$\pm2.8$}$\%$ & $52.9${\tiny$\pm1.6$}$\%$ & $52.5${\tiny$\pm1.4$}$\%$ & $5.0${\tiny$\pm0.4$}$\%$\\
\hline
\multirow{4}{*}{\VAN} & Dyn.~+ Comp. & $\bf{75.5}${\tiny$\bf{\pm3.2}$}$\%$ & $\bf{99.1}${\tiny$\bf{\pm0.3}$}$\%$ & $\bf{96.2}${\tiny$\bf{\pm0.9}$}$\%$ & $\bf{53.3}${\tiny$\bf{\pm2.8}$}$\%$\\
 & Compositional & $70.6${\tiny$\pm3.8$}$\%$ & $98.5${\tiny$\pm0.5$}$\%$ & $95.6${\tiny$\pm0.8$}$\%$ & $44.2${\tiny$\pm3.5$}$\%$\\
 & Joint & $52.7${\tiny$\pm4.9$}$\%$ & $85.5${\tiny$\pm4.9$}$\%$ & $88.5${\tiny$\pm3.7$}$\%$ & $18.4${\tiny$\pm1.7$}$\%$\\
 & No Comp. & $34.6${\tiny$\pm3.7$}$\%$ & $61.3${\tiny$\pm3.8$}$\%$ & $59.8${\tiny$\pm3.6$}$\%$ & $8.7${\tiny$\pm0.5$}$\%$\\
\hline
\multirow{2}{*}{\FM} & Dyn.~+ Comp. & $\bf{83.8}${\tiny$\bf{\pm2.0}$}$\%$ & $\bf{99.6}${\tiny$\bf{\pm0.0}$}$\%$ & $\bf{98.3}${\tiny$\bf{\pm0.3}$}$\%$ & $\bf{68.7}${\tiny$\bf{\pm1.5}$}$\%$\\
 & Compositional & $74.6${\tiny$\pm3.1$}$\%$ & $99.5${\tiny$\pm0.0$}$\%$ & $98.1${\tiny$\pm0.3$}$\%$ & $50.3${\tiny$\pm2.0$}$\%$\\
    \end{tabular}
\end{table}

\subsection{Results on Combined Data Set of Diverse Tasks}
\label{sec:ResultsCombined}
One of the key advantages of learning compositional structures is that they enable learning a more diverse set of tasks, by recombining components in novel ways to solve each problem. In this setting, noncompositional structures struggle to capture the diversity of the tasks in a single monolithic architecture. To verify that this is indeed the case, this evaluation created a novel data set that combines  the $10$ \MNIST{} tasks, $10$ \Fashion{} tasks, and $20$ \CUB{} tasks into a single \bCombined{} lifelong learning data set of $\numTasks=40$ tasks, and compared all methods and baselines on this new data set. The experiment gave no indication to the agents that each task came from a different data set, and they all trained following the exact same setup of Section~\ref{sec:ResultsSoftOrdering}. The framework instantiations and baselines trained with the soft layer ordering method, using the same architecture as used for \CUB{} in Section~\ref{sec:ResultsSoftOrdering}. The pretrained ResNet-18 processed only the \CUB{} images, whereas the task-specific input transformation $\inputTransformt$ consumed directly the \MNIST{} and \Fashion{} images.

Table~\ref{tab:CombinedAll} summarizes the results. As expected, compositional methods clearly outperformed all baselines, by a much wider margin than in the single-data-set settings of Section~\ref{sec:ResultsSoftOrdering}. In particular, no-components baselines (those with monolithic architectures) were completely incapable of learning to solve the \CUB{} tasks, showing that handling this more complex setting requires compositional architectures. Even the jointly trained variants, which do have compositional structures but learn them na\"ively with existing lifelong methods, failed drastically. Compositional methods performed remarkably well, especially when using \ER{} as the base adaptation method and dynamically adding components.

Note that, in order to match the requirements of the \CUB{} data set, the architecture used for these experiments gave \MNIST{} and \Fashion{} a higher capacity (layers of size $256$ vs.~$64$) and the ability to train the input transformation for each task individually (instead of keeping it fixed) compared to the architecture described in Section~\ref{sec:ModelArchitecturesSupervised}. This explains the higher performance of most methods in those two data sets compared to the results in Table~\ref{tab:softOrderingResults}.

\subsection{Evaluation on a Toy Compositional Data Set}
\label{sec:ResultsCompositionalDataSet}
The results of Sections~\ref{sec:ResultsStandardBenchmarks} and~\ref{sec:ResultsCombined} show the compositional learning performance on a suite of data sets that does not explicitly require any compositional structure. This deliberate choice permitted studying the generality of the proposed framework, and revealed that algorithms that instantiate it work well across data sets with a range of feature representations, relations across tasks, numbers of tasks, and sample sizes. This section introduces a data set that explicitly embodies a compositional structure that intuitively matches the assumptions of the soft layer ordering architecture, and shows that the results obtained for noncompositional data sets still hold for this class of problems.

The new {\bf Objects} data set consists of $48$ classes, each corresponding to an object composed of a shape (circle, triangle, or square), color (orange, blue, pink, or green), and location (each of the four quadrants in the image). The data set contains $n=100$ images of size $28\times28$ per class. The data generation process sampled the center location, color, and size of each data point uniformly at random from each attribute's corresponding range. The range for the centers was ${[c_x-3,c_x+3]}, {[c_y-3,c_y+3]}$, where $c_x$ and $c_y$ are the centers of the quadrant for each class, respectively. The range for the red-green-blue (RGB) values was ${[r-16,r+16]}, {[g-16,g+16]},$ ${[b-16,g+16]}$, where $r$, $g$, and $b$ are the nominal RGB values for the color of each class. Finally, the range of the object sizes was $[3, 7]$ pixels. 

To test the proposed framework in this setting, the experiment created a lifelong version of the Objects data set by randomly splitting the data into $\numTasks=16$ three-way classification tasks. The evaluation used $50\%$ of the instances for each class as training data, $20\%$ as validation data, and $30\%$ as test data. The learners used soft ordering nets with $\numModules=4$ components of $64$ fully connected hidden units shared across tasks, and a linear input transformation $\inputTransformt$ trained for each task. All agents trained for $100$ epochs per task using a mini-batch of size $32$, with compositional agents using $99$ epochs for assimilation and a single epoch for adaptation. The regularization hyperparameter for \kEWC{} was set to $\lambda=1e-3$, and \ER{} was given a replay buffer of size $n_m=5$. Each experiment was repeated for $50$ trials with different random seeds controlling class splits for each task, training/validation/test splits for each class, and the ordering of tasks.

\begin{table}[b!]
    \centering
    \caption[Performance of supervised lifelong composition on compositional data set: soft ordering.]{Average final accuracy across tasks on the compositional Objects data set using soft layer ordering. Column labels indicate which component was held out for final tasks. Compositional approaches outperformed the baselines in this setting that explicitly matches the compositional assumptions. Standard errors across ten seeds reported after the $\pm$.}
    \label{tab:ObjectsResults}
    \begin{tabular}{@{}l|l|c|c|c|c@{}}
Base & Algorithm & Circle & Top-left & Orange & Random\\
\hline
\hline
\multirow{4}{*}{\ER} & Dyn.~+ Comp. & $\bf{93.4}${\tiny$\bf{\pm0.7}$}$\%$ & $\bf{85.9}${\tiny$\bf{\pm1.0}$}$\%$ & $\bf{89.4}${\tiny$\bf{\pm1.0}$}$\%$ & $\bf{91.8}${\tiny$\bf{\pm0.6}$}$\%$\\
 & Compositional & $92.2${\tiny$\pm0.9$}$\%$ & $84.9${\tiny$\pm1.2$}$\%$ & $88.7${\tiny$\pm1.2$}$\%$ & $90.9${\tiny$\pm0.9$}$\%$\\
 & Joint & $92.0${\tiny$\pm0.8$}$\%$ & $83.5${\tiny$\pm1.0$}$\%$ & $87.8${\tiny$\pm1.2$}$\%$ & $89.1${\tiny$\pm0.6$}$\%$\\
 & No Comp. & $91.2${\tiny$\pm1.0$}$\%$ & $83.5${\tiny$\pm1.1$}$\%$ & $88.4${\tiny$\pm0.8$}$\%$ & $89.8${\tiny$\pm1.0$}$\%$\\
\hline
\multirow{4}{*}{\kEWC} & Dyn.~+ Comp. & $\bf{93.4}${\tiny$\bf{\pm0.7}$}$\%$ & $\bf{85.9}${\tiny$\bf{\pm1.0}$}$\%$ & $\bf{89.5}${\tiny$\bf{\pm1.1}$}$\%$ & $\bf{91.6}${\tiny$\bf{\pm0.7}$}$\%$\\
 & Compositional & $92.0${\tiny$\pm1.1$}$\%$ & $85.3${\tiny$\pm1.2$}$\%$ & $88.7${\tiny$\pm1.2$}$\%$ & $90.9${\tiny$\pm0.9$}$\%$\\
 & Joint & $91.1${\tiny$\pm0.8$}$\%$ & $82.4${\tiny$\pm1.3$}$\%$ & $87.0${\tiny$\pm1.4$}$\%$ & $90.1${\tiny$\pm0.7$}$\%$\\
 & No Comp. & $88.1${\tiny$\pm1.8$}$\%$ & $81.0${\tiny$\pm1.5$}$\%$ & $83.3${\tiny$\pm2.5$}$\%$ & $86.3${\tiny$\pm2.1$}$\%$\\
\hline
\multirow{4}{*}{\VAN} & Dyn.~+ Comp. & $\bf{93.3}${\tiny$\bf{\pm0.7}$}$\%$ & $\bf{86.3}${\tiny$\bf{\pm1.0}$}$\%$ & $\bf{89.6}${\tiny$\bf{\pm1.1}$}$\%$ & $\bf{91.5}${\tiny$\bf{\pm0.6}$}$\%$\\
 & Compositional & $92.3${\tiny$\pm0.8$}$\%$ & $85.7${\tiny$\pm1.1$}$\%$ & $88.7${\tiny$\pm1.2$}$\%$ & $90.6${\tiny$\pm0.9$}$\%$\\
 & Joint & $90.6${\tiny$\pm0.9$}$\%$ & $81.8${\tiny$\pm1.2$}$\%$ & $86.6${\tiny$\pm1.3$}$\%$ & $88.4${\tiny$\pm1.2$}$\%$\\
 & No Comp. & $89.2${\tiny$\pm2.0$}$\%$ & $77.5${\tiny$\pm1.9$}$\%$ & $86.8${\tiny$\pm1.2$}$\%$ & $85.7${\tiny$\pm1.5$}$\%$\\
\hline
\multirow{2}{*}{\FM} & Dyn.~+ Comp. & $\bf{93.0}${\tiny$\bf{\pm0.7}$}$\%$ & $\bf{86.0}${\tiny$\bf{\pm1.0}$}$\%$ & $\bf{89.5}${\tiny$\bf{\pm1.1}$}$\%$ & $\bf{91.4}${\tiny$\bf{\pm0.5}$}$\%$\\
 & Compositional & $90.8${\tiny$\pm1.4$}$\%$ & $83.8${\tiny$\pm1.5$}$\%$ & $88.1${\tiny$\pm1.2$}$\%$ & $89.4${\tiny$\pm0.9$}$\%$\\
    \end{tabular}
\end{table}

The experiments considered four different evaluation settings. The Random setting split classes and ordered them into task sequences randomly, matching the experimental setting of Section~\ref{sec:ResultsStandardBenchmarks}. The other, more challenging settings held out one shape, location, or color only for the final four (for color and location) or five (for shape) tasks, requiring the agents to adapt to never-seen components dynamically. Results in Table~\ref{tab:ObjectsResults} show that each of the compositional methods outperformed all baselines in all settings, showcasing the ability of the proposed framework to discover the underlying compositional structures.

\subsection{Visualization of the Learned Components}
\label{sec:VisualizationOfComponents}

The primary motivation for the proposed framework was the creation of lifelong learning algorithms capable of discovering self-contained, reusable components, useful for solving a variety of tasks. This section visually inspects the components learned by the framework to verify that they are indeed self-contained and reusable.  
 
\begin{figure}[p]
\centering
\captionsetup[subfigure]{aboveskip=0pt,belowskip=0pt}
    \begin{subfigure}[b]{0.04\textwidth}
        \raisebox{0.1cm}{\rotatebox{90}{$\hspace{1em}4\longleftarrow$\!depth$\longrightarrow1$}}
    \end{subfigure}%
    \begin{subfigure}[b]{0.46\textwidth}
        $\,0\hfill\longleftarrow$ \hfill intensity\hfill  $\longrightarrow\hfill1\,$
        \includegraphics[width=\linewidth, trim={0cm 0cm 0cm 0cm}, clip]{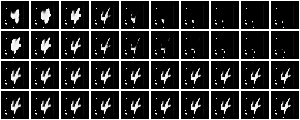}
        \caption*{Task 9}
    \end{subfigure}\hspace{0.25cm}
    \begin{subfigure}[b]{0.46\textwidth}
        $\,0\hfill\longleftarrow$ \hfill intensity\hfill  $\longrightarrow\hfill1\,$
        \includegraphics[width=\linewidth, trim={0cm 0cm 0cm 0cm}, clip]{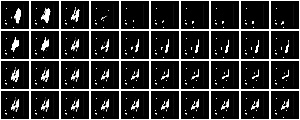}
        \caption*{Task 9}
    \end{subfigure}\\
    \vspace{0.1cm}
    \begin{subfigure}[b]{0.04\textwidth}
        \raisebox{0.1cm}{\rotatebox{90}{$\hspace{2.2em}4\longleftarrow$\!depth$\longrightarrow1$}}
    \end{subfigure}%
    \begin{subfigure}[b]{0.46\textwidth}
        \includegraphics[width=\linewidth, trim={0cm 0cm 0cm 0cm}, clip]{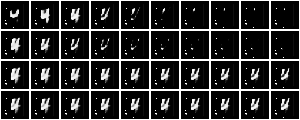}
        \caption*{Task 10}
        \vspace{0.1cm}
        \caption{ER Compositional}
    \end{subfigure}\hspace{0.25cm}
    \begin{subfigure}[b]{0.46\textwidth}
        \includegraphics[width=\linewidth, trim={0cm 0cm 0cm 0cm}, clip]{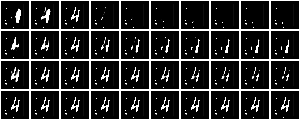}
        \caption*{Task 10}
        \vspace{0.1cm}
        \caption{ER Joint}
    \end{subfigure}%
    \caption[Visualization of first component learned via supervised learning with soft layer ordering.]{Visualization of reconstructed MNIST ``4'' digits on the last two tasks seen by compositional and joint \ER{} with soft layer ordering, varying the intensity of component $i=0$. Compositional \ER{} learned a functional primitive: the more intensely the component is selected (left to right), the thinner the lines of the digit become. The magnitude of this effect decreases with depth (top to bottom), with the digit disappearing as the component is more intensely selected at the earliest layers, but only becoming slightly sharper with intensity at the deepest layers. This effect is consistent across both tasks. Joint \ER{} did not exhibit this consistent behavior, with different effects observed across depths and tasks.}
    \label{fig:MNISTVisualization}
\end{figure}

\begin{figure}[p]
\centering
\captionsetup[subfigure]{aboveskip=0pt,belowskip=0pt}
    \begin{subfigure}[b]{0.04\textwidth}
        \raisebox{0.1cm}{\rotatebox{90}{$\hspace{1em}4\longleftarrow$\!depth$\longrightarrow1$}}
    \end{subfigure}%
    \begin{subfigure}[b]{0.46\textwidth}
        $\,0\hfill\longleftarrow$ \hfill intensity\hfill  $\longrightarrow\hfill1\,$
        \includegraphics[width=\linewidth, trim={0cm 0cm 0cm 0cm}, clip]{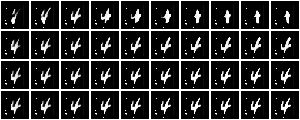}
        \caption*{Task 9}
    \end{subfigure}\hspace{0.25cm}
    \begin{subfigure}[b]{0.46\textwidth}
        $\,0\hfill\longleftarrow$ \hfill intensity\hfill  $\longrightarrow\hfill1\,$
        \includegraphics[width=\linewidth, trim={0cm 0cm 0cm 0cm}, clip]{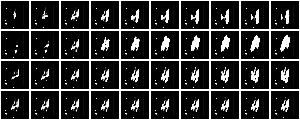}
        \caption*{Task 9}
    \end{subfigure}\\
    \vspace{0.1cm}
    \begin{subfigure}[b]{0.04\textwidth}
        \raisebox{0.1cm}{\rotatebox{90}{$\hspace{2.2em}4\longleftarrow$\!depth$\longrightarrow1$}}
    \end{subfigure}%
    \begin{subfigure}[b]{0.46\textwidth}
        \includegraphics[width=\linewidth, trim={0cm 0cm 0cm 0cm}, clip]{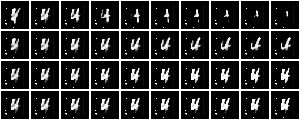}
        \caption*{Task 10}
        \vspace{0.1cm}
        \caption{ER Compositional}
    \end{subfigure}\hspace{0.25cm}
    \begin{subfigure}[b]{0.46\textwidth}
        \includegraphics[width=\linewidth, trim={0cm 0cm 0cm 0cm}, clip]{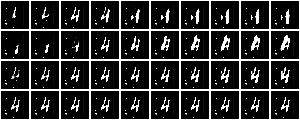}
        \caption*{Task 10}
        \vspace{0.1cm}
        \caption{ER Joint}
    \end{subfigure}%
    \caption[Visualization of second component learned via supervised learning with soft layer ordering.]{Visualization of reconstructed MNIST ``4'' digits, varying the intensity of component $i=1$. The component learned via compositional \ER{} consistently decreases the length of the left side of the digit and increases that of the right side. Again, it was not possible to detect any consistency in the effect of the component learned via joint \ER{}.}
    \label{fig:MNISTVisualization1}
\end{figure}

\begin{figure}[t!]
\centering
\captionsetup[subfigure]{aboveskip=0pt,belowskip=0pt}
    \begin{subfigure}[b]{0.04\textwidth}
        \raisebox{0.1cm}{\rotatebox{90}{$\hspace{1em}4\longleftarrow$\!depth$\longrightarrow1$}}
    \end{subfigure}%
    \begin{subfigure}[b]{0.46\textwidth}
        $\,0\hfill\longleftarrow$ \hfill intensity\hfill  $\longrightarrow\hfill1\,$
        \includegraphics[width=\linewidth, trim={0cm 0cm 0cm 0cm}, clip]{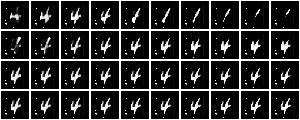}
        \caption*{Task 9}
    \end{subfigure}\hspace{0.25cm}
    \begin{subfigure}[b]{0.46\textwidth}
        $\,0\hfill\longleftarrow$ \hfill intensity\hfill  $\longrightarrow\hfill1\,$
        \includegraphics[width=\linewidth, trim={0cm 0cm 0cm 0cm}, clip]{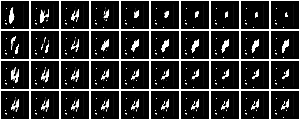}
        \caption*{Task 9}
    \end{subfigure}\\
    \vspace{0.1cm}
    \begin{subfigure}[b]{0.04\textwidth}
        \raisebox{0.1cm}{\rotatebox{90}{$\hspace{2.2em}4\longleftarrow$\!depth$\longrightarrow1$}}
    \end{subfigure}%
    \begin{subfigure}[b]{0.46\textwidth}
        \includegraphics[width=\linewidth, trim={0cm 0cm 0cm 0cm}, clip]{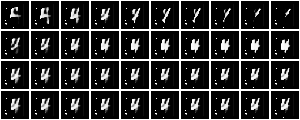}
        \caption*{Task 10}
        \vspace{0.1cm}
        \caption{ER Compositional}
    \end{subfigure}\hspace{0.25cm}
    \begin{subfigure}[b]{0.46\textwidth}
        \includegraphics[width=\linewidth, trim={0cm 0cm 0cm 0cm}, clip]{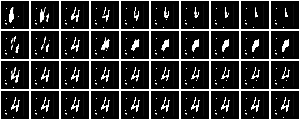}
        \caption*{Task 10}
        \vspace{0.1cm}
        \caption{ER Joint}
    \end{subfigure}%
    \caption[Visualization of third component learned via supervised learning with soft layer ordering.]{Visualization of reconstructed MNIST ``4'' digits, varying the intensity of component $i=2$. As the intensity of the component learned via compositional \ER{} increased, the digit changed from very sharp to very smooth. Joint \ER{} did not exhibit any consistent behavior.}
    \label{fig:MNISTVisualization2}
\end{figure}

\begin{figure}[t!]
\centering
\captionsetup[subfigure]{aboveskip=0pt,belowskip=0pt}
    \begin{subfigure}[b]{0.04\textwidth}
        \raisebox{0.1cm}{\rotatebox{90}{$\hspace{1em}4\longleftarrow$\!depth$\longrightarrow1$}}
    \end{subfigure}%
    \begin{subfigure}[b]{0.46\textwidth}
        $\,0\hfill\longleftarrow$ \hfill intensity\hfill  $\longrightarrow\hfill1\,$
        \includegraphics[width=\linewidth, trim={0cm 0cm 0cm 0cm}, clip]{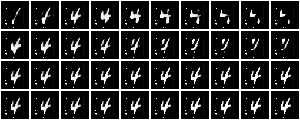}
        \caption*{Task 9}
    \end{subfigure}\hspace{0.25cm}
    \begin{subfigure}[b]{0.46\textwidth}
        $\,0\hfill\longleftarrow$ \hfill intensity\hfill  $\longrightarrow\hfill1\,$
        \includegraphics[width=\linewidth, trim={0cm 0cm 0cm 0cm}, clip]{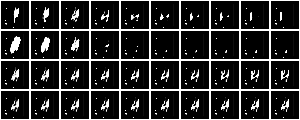}
        \caption*{Task 9}
    \end{subfigure}\\
    \vspace{0.1cm}
    \begin{subfigure}[b]{0.04\textwidth}
        \raisebox{0.1cm}{\rotatebox{90}{$\hspace{2.2em}4\longleftarrow$\!depth$\longrightarrow1$}}
    \end{subfigure}%
    \begin{subfigure}[b]{0.46\textwidth}
        \includegraphics[width=\linewidth, trim={0cm 0cm 0cm 0cm}, clip]{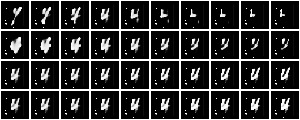}
        \caption*{Task 10}
        \vspace{0.1cm}
        \caption{ER Compositional}
    \end{subfigure}\hspace{0.25cm}
    \begin{subfigure}[b]{0.46\textwidth}
        \includegraphics[width=\linewidth, trim={0cm 0cm 0cm 0cm}, clip]{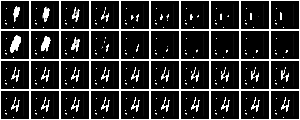}
        \caption*{Task 10}
        \vspace{0.1cm}
        \caption{ER Joint}
    \end{subfigure}%
    \caption[Visualization of fourth component learned via supervised learning with soft layer ordering.]{Visualization of reconstructed MNIST ``4'' digits, varying the intensity of component $i=3$. This component also interpolates between sharper and smoother digits, while also rotating the digit. There was  no consistency in the component learned by Joint \ER{}.}
    \label{fig:MNISTVisualization3}
\end{figure}

The experimental setting followed that of \citet{meyerson2018beyond}, where each task corresponded to a single image of the digit ``4'', and each pixel in the image constituted one data-point. The $x, y$ coordinates of the pixel constituted the features, and the pixel's intensity was the associated label. The preprocessing normalized the pixel coordinates and intensities to $[0,1]$. The agent trained using all pixels in the image, since the goal was understanding the learned representations, as opposed to generalizing to unseen data. The agent used a network with $\numModules=4$ components shared across all tasks, and used soft layer ordering to learn the structure $\structuret$ for each task. The network first processed the input with a linear transformation layer $\inputTransform$ shared across all tasks, and after the compositional structure passed its output through a shared sigmoid output transformation layer $\outputTransform$. Sharing the input and output transformations across tasks ensured that the only differences across the models of the different tasks were due to the structure of each task over the components. The network minimized the binary cross-entropy loss on $\numTasks=10$ tasks for $1{,}000$ epochs via the compositional and jointly trained versions of \ER{} with a replay buffer and mini-batch of $32$ pixel instances, updating the components of the compositional version every $100$ epochs.

To assess the ability of compositional \ER{} to capture reusable functional primitives, the evaluation varied the intensity $\StructureParamsti{i,j}$ with which the network chose one specific component $\modulei$ at different depths $j$, and observed the reconstructed images output by the network. The evaluation focused on the last two tasks seen by the learner, in order to disregard the effects of catastrophic forgetting, which rendered the visualizations of the outputs of the joint \ER{} baseline incomprehensible for earlier tasks. Figures~\ref{fig:MNISTVisualization}--\ref{fig:MNISTVisualization3} show these reconstructions as the intensity of each component individually varies at different depths. The components trained with compositional \ER{} learned to produce effects on the digits consistent across tasks, with more extreme effects at the initial layers. For example, Figure~\ref{fig:MNISTVisualization} reveals that the discovered component learned to vary the thickness of the digit regardless of the task at hand, with the effect being more pronounced at the initial layers. In contrast, joint \ER{} learned components whose effects are different for different tasks and at different depths.

\section{Summary}
This chapter developed concrete instantiations for the supervised setting of the general-purpose framework for lifelong learning of compositional structures presented in Chapter~\ref{cha:Framework}. 
These instantiations demonstrated the flexibility of the proposed framework by capturing nine different concrete algorithms, varying the choice of compositional structures and the back-bone mechanism for accommodating knowledge of multiple tasks sequentially. An extensive evaluation empirically tested each instantiation, showing that these algorithms are stronger lifelong learners than existing approaches. In particular, the evaluation demonstrated that both learning traditional monolithic architectures and na\"ively training compositional structures via existing methods lead to substantially degraded performance. In contrast, methods that follow the proposed framework suffered from minimal amounts of forgetting, improved the quality of the model over time as they trained on more tasks, performed well even in the face of extremely little data, and achieved higher overall performance. Therefore, not only is the framework simple conceptually, but it is also easy to combine with existing continual or compositional learning techniques, and effective in trading off the flexibility and stability required for lifelong learning.

This chapter showed the potential of compositional structures to enable strong lifelong learning. One major line of open work remains properly understanding how to measure the quality of the obtained compositional solutions, especially in settings without obvious decompositions, like those considered in Section~\ref{sec:ResultsStandardBenchmarks}. While the visualizations in Section~\ref{sec:VisualizationOfComponents} and results in Table~\ref{tab:ModuleReuse} suggest that instantiations of the framework obtain reusable components, so far this lacks a proper metric to assess the degree to which the learned structures are compositional. Chapter~\ref{cha:RL} develops benchmarks for the RL setting that are explicitly compositional, and which consequently permit assessing whether the obtained solutions are indeed compositional.

\biblio

\chapter{Application of Lifelong Composition to Reinforcement Learning}
\label{cha:RL}

\newtheorem{claim}{Claim}
\newtheorem{lemma}{Lemma}
\newtheorem{proposition}{Proposition}
\newtheorem{remark}{Remark}
\newtheorem{corollary}{Corollary}

\section{Introduction}
\label{sec:IntroductionRL}
So far, the discussion of the contributions of this dissertation has centered around supervised learning, demonstrating the promise of the proposed framework for lifelong discovery of compositional knowledge representations. This chapter expands on this discussion by demonstrating that functionally compositional knowledge is a powerful tool for RL, too. 

To this end, this chapter first formalizes the problem of discovering functionally compositional knowledge specifically for RL, adapting the supervised learning definition described in Chapter~\ref{cha:Framework} into the RL setting. Intuitively, each functional module can be thought of as akin to a function in a data processing pipeline for a programmed robot controller. In particular, this is fundamentally different from the typical \textit{temporal} composition considered by the majority of hierarchical RL works. 

Subsequently, the chapter presents two novel algorithms for lifelong RL, following the lifelong compositional learning framework of Chapter~\ref{cha:Framework}. Unlike in the supervised setting, there is no wealth of existing modular architectures and lifelong mechanisms for RL. Consequently, this dissertation developed the two methods described in this chapter from basic principles. 

The first mechanism uses the compositional structure with linear model combinations described in Chapter~\ref{cha:Framework}. In the assimilation stage, this method directly trains the selection over existing components via PG learning; in the accommodation stage, it approximates the overall MTL cost via a second-order Taylor expansion and solves the resulting optimization problem in closed form. These simplifying assumptions permitted deriving theoretical proofs that the proposed algorithm finds an optimum of the approximate MTL objective. An empirical evaluation of this method on a range of continuous control problems, including a challenging set of robotic manipulation tasks from Meta-World~\citep{yu2019meta}, demonstrated the capability of the method to accelerate the learning of existing single-task PG methods. 

The second mechanism uses a form of hard modular net, introduced in Chapter~\ref{cha:Framework} and described in detail here in Section~\ref{sec:ModularArchitectureRL}. This method separates the assimilation stage into two substages: a discrete search stage, where the agent chooses the optimal module combination, and an exploration stage, where the agent trains via standard RL using the chosen modules as an initialization. The accommodation stage uses data collected during the exploration stage from the current and previous tasks to perform experience replay. Learning a policy off-line from fixed data introduces a distributional shift between the data and the learned policy, which strains the capabilities of standard methods~\citep{levine2020offline}. One recent technique is to constrain the departure from the observed data distribution~\citep{fujimoto2019off,laroche2019safe,kumar2019stabilizing}. This issue bears close connections to lifelong RL: obtaining backward transfer after training on future tasks requires modifying the behavior on the earlier tasks without additional experience. The accommodation stage exploits this connection by storing a portion of the data collected on each task and replaying it for off-line RL after training on future tasks.
Given the fact that the hard modular architecture used by this method closely matches the functional composition assumption, the evaluation constructed two sets of RL tasks that specifically match these compositional assumptions and tested the proposed method on those tasks. The second of these benchmarks, which consists of a set of tens of robotic manipulation tasks, was subsequently extended in Chapter~\ref{cha:Benchmark} into \textit{hundreds} of far more complex and diverse robotics tasks.

\section{Background on Reinforcement Learning for Continuous State-Action Spaces}
\label{sec:RLviaPolicySearch}

Formally, an RL problem is given by an MDP $\MDP = \langle \States, \Actions, \Rewards, \Transitions, \gamma\rangle$, where $\States\subseteq\Reals^{d}$ is the set of states, $\Actions\subseteq\Reals^{m}$ is the set of actions, $\Transitions:\States\times\Actions\times\States\mapsto[0,1]$ is the probability distribution $p\big(\state^\prime \mid \state, \action\big)$ of transitioning to state $\state^\prime$ upon executing action $\action$ in state $\state$, $\Rewards: \States\times\Actions\mapsto\Reals$ is the reward function measuring the goodness of a given state-action pair, with $\Rewards_i=\Rewards(\state_i, \action_i)$ being the reward obtained at step $i$, and $\gamma\in[0,1)$ is the discount factor that reduces the importance of rewards obtained far in the future. A policy $\pi:\States\times\Actions\mapsto[0,1]$ dictates the agent's behavior, giving the probability $p(\action\mid\state)$ of selecting action $\action$ in state $\state$. The goal of the agent is to find the policy $\pi^*\in\Policies$ that maximizes the expected discounted long-term returns $\mathbb{E}\big[\sum_{i=0}^\infty 
\gamma^i \Rewards_i\big]$~\citep{sutton2018reinforcement}. 

Existing approaches to solve the RL problem rely on two common quantities to evaluate the quality of a given policy $\pi$. The state-value function, given by:
\begin{align}
    V^{\pi}(\state) = \mathbb{E} \left[ \Rewards(\state, \action\sim\pi(\state)) +  \sum_{i=1}^\infty \gamma^i \Rewards(\state_i\sim\Transitions(\state_{i-1}, \action_{i-1}, \state_i), \action_i\sim\pi(\state_i)) \right]\enspace,
\end{align}
represents how good a given state $\state$ is under a given policy $\pi$. If evaluated on the (potentially stochastic) initial state $\state_0$, $V$ gives the expected overall performance of the policy $\pi$. Similarly, the action-value function, given by:
\begin{align}
    Q^{\pi}(\state, \action) =  \mathbb{E} \left[ \Rewards(\state, \action) +  \sum_{i=1}^\infty \gamma^i \Rewards(\state_i\sim\Transitions(\state_{i-1}, \action_{i-1}, \state_i), \action_i\sim\pi(\state_i)) \right]\enspace,
\end{align}
characterizes how good a given state-action pair $\state,\action$ is under the policy $\pi$. From these two quantities, one can also obtain the advantage function $A^\pi(\state, \action) = Q^\pi(\state,\action) - V^\pi(\state)$ as the improvement of the action $\action$ over the average action in the current state $\state$. In principle, one can find the optimal policy $\pi^*$ by optimizing the $Q$-function by repeatedly applying the Bellman equation:
\begin{equation}
\label{equ:Bellman}
    Q^*(\state, \action) = \Rewards(\state,\action) + \max_{\action^\prime}\gamma Q^*\big(\state^\prime\sim\Transitions(\state, \action, \state^\prime), \action^\prime\big)\enspace.
\end{equation}
This is at a high level the principle followed by $Q$-learning~\citep{watkins1989learning} and its variants, including the deep $Q$-network algorithm~(DQN; \citealp{mnih2015human}). However, in cases where the action space is continuous, as considered in most of this dissertation, the maximization in Equation~\ref{equ:Bellman} is infeasible to compute.

PG algorithms have shown success in solving continuous-action RL problems by assuming that a vector $\btheta\in\Reals^{d}$ parameterizes the policy $\pi_{\btheta}$ and searching for the set of parameters $\btheta^*$ that optimizes the long-term rewards: $\Obj(\btheta)=\mathbb{E}\left[\sum_{i=0}^{\infty}\gamma^{i}\Rewards_i\right]$~\citep{sutton2000policy,schulman2015trust,lillicrap2015continuous}. Different approaches use varied strategies for estimating the gradient $\nabla_{\btheta}\Obj(\btheta)$. 
However, the common high-level idea is to use the current policy $\pi_{\btheta}$ to sample trajectories of interaction with the environment for exploration, and then estimate the gradient as the average of some function of the state features and rewards encountered through the trajectories.

Alternatively, actor-critic methods simultaneously learn a policy and a $Q$-function to estimate the quality of the policy. Broadly, the learning trains a parametric policy $\pi_{\btheta_\pi}$ to solve the inner maximization problem of Equation~\ref{equ:Bellman}, and a parametric $Q$-function $Q_{\btheta_{Q}}$ to solve the outer optimization~\citep{konda1999actor,mnih2016asynchronous,schulman2017proximal}. 

\section{The Lifelong Reinforcement Learning Problem}
\label{LifelongLearningProblem}

Chapter~\ref{cha:Framework} described the problem of lifelong supervised learning used to develop the algorithms in Chapter~\ref{cha:Supervised}, where the agent must learn predictive functions for a sequence of tasks by leveraging information that is common to multiple tasks. This section adapts the problem definition to the RL setting, where the agent instead must learn policies and value functions based on shared information. Concretely, the agent faces a sequence of tasks $\Task^{(1)},\dots,\Task^{(\numTasks)}$, each of which is an MDP $\Task^{(t)}=\langle\States^{(t)},\Actions^{(t)},\Transitions^{(t)},\Rewards^{(t)},\gamma \rangle$. The environment draws tasks \iid~from a fixed, stationary environment. Section~\ref{sec:TheoreticalGuarantees} formalizes this stationarity assumption for a simplified setting, and the remaining sections of this chapter use it in an informal sense; Chapter~\ref{cha:NonStationary} extends the problem definition to a nonstationary setting. The goal of the agent is to find the policy parameters $\left\{\btheta^{(1)}_{\pi},\ldots,\btheta^{(\numTasks)}_{\pi}\right\}$ and value-function parameters $\left\{\btheta^{(1)}_{Q},\ldots,\btheta^{(\numTasks)}_{Q}\right\}$ that maximize the performance across all tasks: $\frac{1}{\numTasks}\sum_{t=1}^{\numTasks} \mathbb{E}\Big[\sum_{i=0}^{\infty}\gamma^{i}\Rewards_i^{(t)}\Big]$. As in the supervised case, the agent is unaware of the total number of tasks it will face or the order in which tasks will arrive. However, the agent does receive a task indicator $t$ that lets it differentiate tasks from one another. 

Upon observing each task, the agent is allowed to interact with the environment for a limited time, typically insufficient for obtaining optimal performance without exploiting information from prior tasks. During this time, the learner  explores the effects of different behaviors on the environment, striving to discover any relevant information from the current task to 1)~relate it to previously stored knowledge in order to permit transfer and 2)~store any newly discovered knowledge for future reuse. At any time, the environment may evaluate the agent on any previously seen task, so it must retain knowledge from all early tasks. 

As discussed in Chapter~\ref{cha:RelatedWork}, unlike in the supervised setting, there is no wealth of lifelong or continual learning methods for RL. In consequence, this dissertation developed two novel methods for incorporating knowledge from multiple consecutive tasks into a repository of shared components. Sections~\ref{sec:LPG-FTWAlgo} and~\ref{sec:LifelongCompositionalRLAlgo} describe how these two new accommodation techniques yield two corresponding new lifelong RL approaches.

\section{The Problem of Lifelong Functional Composition in Reinforcement Learning}
\label{sec:lifelongCompositionInRL}

The problem formulation of compositional learning described in Chapter~\ref{cha:Framework} for the supervised setting is natural, and existing works have extensively used it for designing modular supervised learning architectures: each predictive function is a composition of multiple simpler functions, and the goal of the agent is to find the optimal function decomposition to maximize transfer across multiple tasks.

In the case of RL, complex problems can also often be divided into easier subproblems. However, most works in RL consider temporal composition, where the agent executes skills one after another. While this dissertation shares the premise of compositional solutions with such hierarchical RL efforts, those works do not consider the formulation presented here, where \textit{functional} composition occurs at multiple hierarchical levels of abstraction.

Recent years have taught us that, given unlimited experience,  artificial agents can tackle such complex problems without any sense of their compositional structures~\citep{silver2016mastering,gu2017deep}. However, discovering the underlying subproblems and learning to solve those would enable learning with substantially less experience, especially when faced with numerous tasks that share a common structure.

An RL problem $\MDP$ is a composition of subproblems $\subproblem_1,\subproblem_2,\ldots$ if its optimal policy $\pi^*$ can be constructed by combining solutions to those subproblems: $\pi^*(\state)=\modulei[1]\circ \modulei[2] \circ \cdots$, where each $\modulei\in\ModuleSpace: \Inputi\mapsto\Outputi$ is the solution to the corresponding $\subproblemi$. 
From an intuitive perspective, in RL each subproblem could involve pure sensing, pure acting, or a combination of both. For instance, recall the robotic manipulation example of Chapter~\ref{cha:Framework}, where a variety of robot arms learn to achieve various objectives with different objects while avoiding a choice of obstacles. In that case, each task can be decomposed into recognizing objects (sensing), detecting the obstacle and devising a plan to avoid it (combined), detecting the target location and devising a plan to reach it (combined), and actuating the robot to follow the plans (acting).

\begin{figure}[b!]
    \centering
    \includegraphics[width=\linewidth,clip,trim=2.2in 2.2in 2.2in 2.2in]{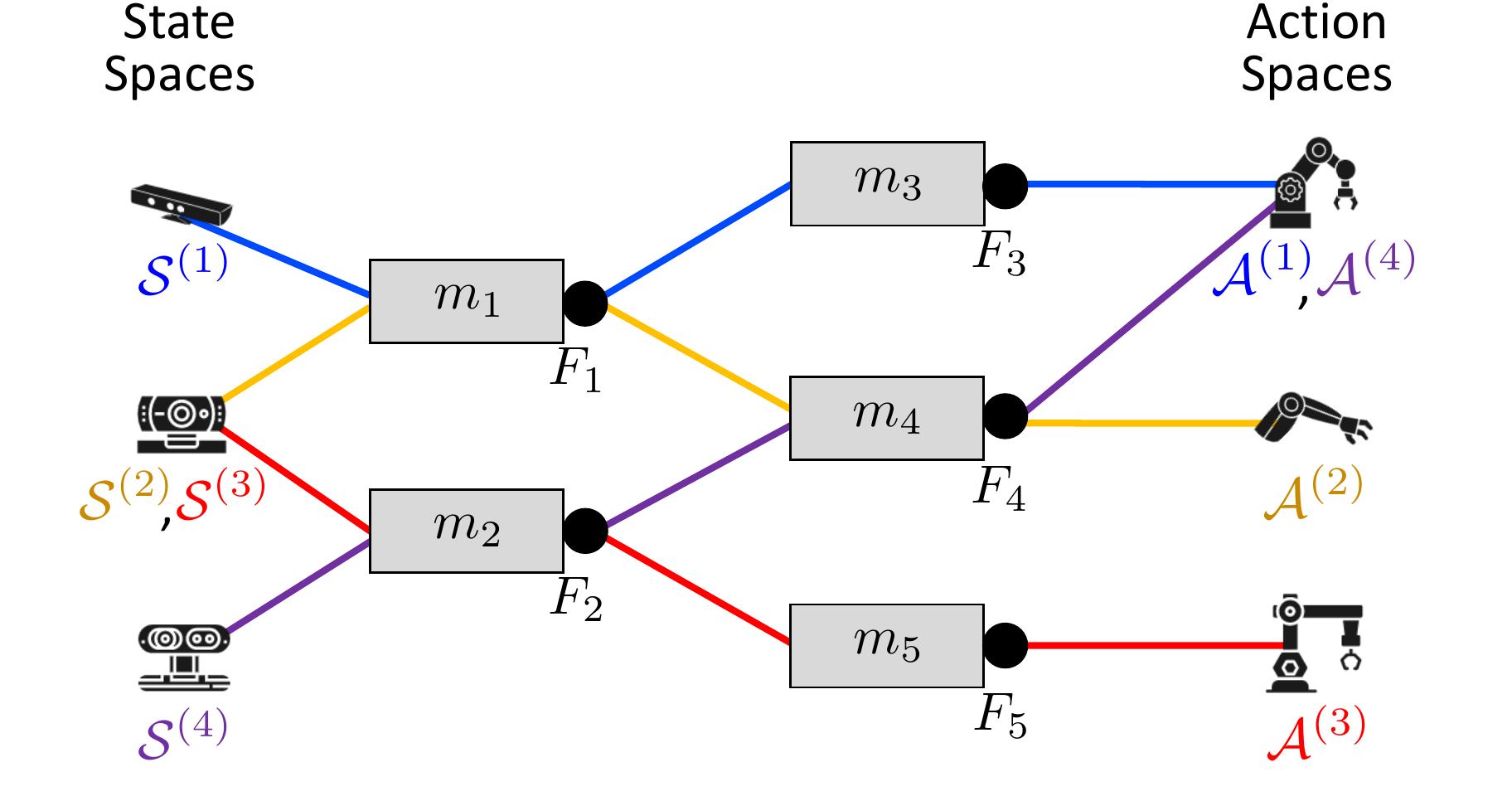}
    \caption[Compositional RL problem graph.]{Compositional RL problem graph. Each node in the graph represents a state space, an action space, or the representational space corresponding to one subproblem that must be solved as part of a set of tasks. Different paths through the graph apply a sequence of transformations to the input and yield the solutions to different tasks.}
    \label{fig:compositionalRLGraph}
\end{figure}

In lifelong compositional RL, the agent faces a sequence of MDPs $\MDPt[1],\ldots,\MDPt[\numTasks]$
over its lifetime. All MDPs are compositions of different subsets from $\totalModules$ shared subproblems $\F = \{\subproblemi[1],\ldots,\subproblemi[\totalModules]\}$. The goal of the lifelong learner is to find the set of solutions to these subproblems as a set of modules $\Modules =\{\modulei[1], \ldots, \modulei[\totalModules]\}$, such that learning to solve a new 
problem reduces to finding how to combine these modules optimally. Each module can be viewed as a processing stage in a hierarchical processing pipeline, or equivalently as functions in a program, and the goal of the agent is to find the correct module to execute at each stage and the instantiation of that module (i.e., its parameters).

Similar to the supervised setting, the compositional RL problem can be formalized as a graph $\mathcal{G} = (\mathcal{V}, \mathcal{E})$ (e.g., Figure~\ref{fig:compositionalRLGraph}) whose
nodes are the subproblem solutions augmented with the state and 
action spaces of the MDPs: $\mathcal{V} = \F \bigcup \breve{\States} \bigcup \breve{\Actions}$, where $\breve{\States} = \mathrm{unique}\Big(\Big\{\States^{(1)}, \ldots, \States^{(\numTasks)}\Big\}\Big)$ are 
the unique\footnote{This process combines comparable spaces into one node, such as one state space used by multiple tasks.} state spaces and $\breve{\Actions} = \mathrm{unique}\Big(\Big\{\Actions^{(1)}, \ldots, \Actions^{(\numTasks)}\Big\}\Big)$ are the unique action spaces. Each subproblem  $\subproblemi$ corresponds 
to a latent representational space $\Outputi$, generated by the corresponding module 
$\modulei \in \ModuleSpace: \Inputi \mapsto \Outputi$. Similarly, the state and action spaces $\States^{(t)}$'s and $\Actions^{(t)}$'s can serve as representation spaces ($\Input_t, \Output_t$).

A pair of state and action nodes $\Big(\States^{(t)}, \Actions^{(t)}\Big)$ in the graph then specify a corresponding problem $\MDPt[t]$, and the goal of the compositional learner 
is to find a path between those nodes corresponding to a policy ${\pi^{(t)*}}$ that maximizes $\Rewards^{(t)}$.
More generally, the graph formalism allows for recurrent computations via walks with cycles, and parallel computations via concurrent multipaths; an extended definition of \textit{multiwalks} trivially captures both settings. The methods  in this chapter consider only the path formulation, and restrict the number of edges in the graph by organizing the modules into layers, as explained in Section~\ref{sec:ModularArchitectureRL}. 

\subsection{Connection Between Functionally Compositional Reinforcement Learning and Hierarchical Reinforcement Learning}
\label{sec:compositionalRLvsHierarchicalRL}

This section discusses the close connections between the proposed functionally compositional RL problem and the popular hierarchical RL setting. 

The primary conceptual difference between hierarchical RL and the proposed functionally compositional RL problem is that hierarchical RL considers composition of sequences of actions \emph{in time}, whereas the proposed problem considers composition \emph{of functions} that, when combined, form a full policy. In  particular, for a given compositional task, the agent uses all functions that make up its modular policy at every time step to determine the action to take (given the current state).

Going back to the example of robot programming from Section~\ref{sec:IntroductionRL}, modules in the compositional RL formulation might correspond to sensor processing drivers, path planners, or robot motor drivers. In programming, at every time step, the sensory input passes through modules in some preprogrammed sequential order, which finally outputs the motor torques to actuate the robot. Similarly, in compositional RL, the state observation passes through the different modules, used in combination to execute the agent's actions. 

Hierarchical RL takes a complementary approach. Instead, each ``module'' (e.g., an option) is a self-contained policy that receives as input the state observation and outputs an action. Each of these options operates in the environment, for example to reach a particular subgoal state. Upon termination of an option, the agent selects a different option to execute, starting from the state reached by the previous option. In contrast, the compositional RL framework assumes that the agent uses a single policy to solve a complete task.

An integrated approach is possible that decomposes the problem along both a functional axis and a temporal axis. This would enable selecting a different functionally modular policy at different stages of solving a task, simplifying the amount of information that each module should encode. Consequently, the proposed framework could be used to learn individual options, which the agent would then compose sequentially. This would enable options to be made up of functionally modular components, simplifying the form of the options themselves and enabling reuse {\em across} options. Research in this direction could drastically improve the data efficiency of RL approaches.

\subsection{Deployment Scenarios for Lifelong Compositional Learning}

This section describes two possible deployment settings for compositional RL agents, depending on whether the agent receives information that enables predicting the optimal compositional structure for a task without collecting data, or data is necessary for discovering the compositional structure. 

\paragraph{Zero-shot generalization with full information} In some scenarios, the agent may have access to a task descriptor that encodes how the current task relates to others in terms of their compositional structures. This descriptor might be sufficient to combine modules into a solution (i.e., zero-shot generalization), provided that the agent has learned to map the descriptors into a solution structure. 
The descriptor could take different forms, such as a multi-hot encoding of the various components, natural language, or highlighting the target objects in the input image. The experiments in this chapter studied multi-hot descriptors as a means to provide the compositional structure. Formally, zero-shot generalization assumes that the agent receives a descriptor as an external input $\taskDescriptor\in\TaskDescriptors$, and that there exists some function $\structure: \TaskDescriptors \times \ModuleSpace \mapsto \Policies$ that can map this input and an optimal set of modules $\Modules$ into an optimal policy for the current task  $\structure(\taskDescriptor, \Modules) = {\pi^{(t)*}}$. This would enable the agent to achieve \emph{compositional generalization}: the ability to solve a new task entirely by reusing components learned from previous tasks. 

\paragraph{Fast adaptation with restricted information} In other scenarios, the agent does not have the luxury of information about the compositional structure. This is common in RL, where the only supervisory signal is typically the reward. In this case, the agent would be incapable of zero-shot transfer. Instead, this setting measures generalization as the agent's ability to \emph{learn from limited experience} a task-specific function $\structuret: \ModuleSpace \mapsto \Policies$ that combines  existing modules $M$ into the optimal policy for the current task  $\structuret(\Modules)={\pi^{(t)*}}$. Intuitively, if the agent has accumulated a useful set of modules $\Modules$, then one would expect it to be capable of quickly discovering how to combine and adapt them to solve new tasks.

\section{\lpgftw{}: Approximate Modular Lifelong Reinforcement Learning via Parameter Factorization}
\label{sec:LPG-FTWAlgo}

The first lifelong RL algorithm developed in this dissertation, \textit{lifelong PG: faster training without forgetting} (\lpgftw{}), uses the compositional architecture that linearly combines model parameters, described in Chapter~\ref{cha:Framework}. Concretely, 
\lpgftw{} assumes that the policy parameters for task $\Taskt$ can factor into $\btheta^{(t)}\approx \bL \st$, where $\bL\in\Reals^{d\times k}$ is a shared dictionary of policy factors and $\st\in\Reals^{k}$ are task-specific coefficients that select components for the current task. \lpgftw{} further assumes access to some base PG algorithm that can find a parametric policy that performs well on a single task, although not necessarily optimally.

In the case of linear parameterization of the policies, this linear factorization induces a soft approximation of the graph in Figure~\ref{fig:compositionalRLGraph}. In particular, it assumes that all paths from states to actions have only one intermediate node, corresponding to the functional policy (i.e., the length of the path is two edges). Instead of selecting a single one of the $k$ possible intermediate nodes or components, \lpgftw{} softly combines all $k$ components via the $\st$'s. If the policy is instead parameterized by a deep net, this approximation is much cruder, since linear combinations of parameters do not correspond to linear combinations of the outputs of the network. Despite the coarseness of this approximation, results in Chapter~\ref{sec:ExperimentalResultsChallenging} show that parameter factorization of deep nets can still achieve accelerated learning in complex robotic manipulation tasks.

\begin{figure}[t!]
  \centering
  \begin{minipage}{0.95\linewidth}
    \begin{algorithm}[H] 
      \caption{\lpgftw{}($d, k, \lambda, \mu, M$)}
      \label{alg:LPGFTW}
\begin{algorithmic} 
        \STATE $T\gets0$,\ \ \ \ $\bL\gets\mathtt{ initializeModules}(d,k)$
        \WHILE {$\,\,\Taskt\gets\mathtt{getTask}()$}
            \IF{{\,\,\tt isNewTask$\Big(\Taskt\Big)$}}
                \STATE $\st\gets\mathtt{initializeStructure}(k)$
                \STATE $T\gets T + 1$
            \ELSE[subtract from cost]
                \STATE $\bA \gets \bA - 2\left(\st{\st}^\top\right) \kron \Ht$
                \STATE $ \bb \gets \bb -  \st \kron \Big(-\gt + 2\Ht\alphat\Big)$
            \ENDIF
            \FOR[assimilation via PGs]{$\,\,i = 1,\ldots,\,\mathtt{structureUpdates}$}
                \STATE $\trajectories \gets \mathtt{getTrajectories}\Big(\bL\st\Big)$
                \STATE $\st \gets\mathtt{PGStep}\Big(\trajectories,\bL,\st,\mu\Big)$
                \IF[accommodation via 2nd-order]{$\,\,i \!\mod \mathtt{adaptationFrequency} = 0$}
                    \STATE $\alphat \gets \bL\st$
                    \STATE $\gt, \Ht \gets\mathtt{gradientAndHessian}\Big(\alphat\Big)$
                    \STATE $\bA_{\mathrm{tmp}} \gets \bA + 2\left(\st{\st}^\top\right) \kron \Ht$
                    \STATE $\bb_{\mathrm{tmp}} \gets \bb +  \st \kron \Big(-\gt + 2\Ht\alphat \Big)$%
                    \STATE $\vect(\bL) \gets \left(\frac{1}{T}\bA_{\mathrm{tmp}} -2 \lambda \bI\right)^{-1} \left(\frac{1}{T}\bb_{\mathrm{tmp}}\right)$
                \ENDIF
            \ENDFOR
            \STATE $\bA \gets \bA_{\mathrm{tmp}}$,\ \ \ \ $\bb \gets \bb_{\mathrm{tmp}}$
        \ENDWHILE
    \end{algorithmic}
    \end{algorithm}
  \end{minipage}
\end{figure}

Following the framework prescribed in Chapter~\ref{cha:Framework}, \lpgftw{} splits its learning process into the following stages, summarized in Algorithm~\ref{alg:LPGFTW}.

\begin{figure}[t!]
  \centering
  \begin{minipage}{.85\linewidth}
    \begin{algorithm}[H] 
      \caption{InitializeModules($d, k, \lambda, \mu$)}
      \label{alg:LifelongPolicyInitialization}
    \begin{algorithmic} 
        \STATE $T\gets0$,\ \ \ \ $\bL\gets\mathtt{empty}(d,0)$ \COMMENT{create empty set of modules}
        \WHILE{$\,\,T < k$} 
            \STATE $\Taskt\gets\mathtt{getTask}()$
            \STATE $\st\gets\mathtt{initializeStructure}(k)$
            \STATE $T\gets T + 1$
            \FOR[PG training of new module]{$\,\,i = 1,\ldots,\,\mathtt{structureUpdates}$}
                \STATE $\trajectories \gets\mathtt{getTrajectories}\Big(\bL\st\Big)$
                \STATE $\st, \epsilont \gets\mathtt{PGStep}\Big(\trajectories,\bL,\st,\epsilont,\mu\Big)$
            \ENDFOR
            \STATE $\bL \gets\mathtt{addColumn}\Big(\bL, \epsilont\Big)$ \COMMENT{incorporate new module}
            \STATE $\alphat \gets \bL\st + \epsilont$
            \STATE $\gt, \Ht \gets\mathtt{gradientAndHessian}\Big(\alphat\Big)$
            \STATE $\bA \gets \bA + 2\left(\st{\st}^\top\right) \kron \Ht$
            \STATE $ \bb \gets \bb +  \st \kron \Big(-\gt + 2\Ht\alphat\Big)$
        \ENDWHILE
    \end{algorithmic}
    \end{algorithm}
  \end{minipage}
\end{figure}

\paragraph{Initialization} Like other approaches in this dissertation, \lpgftw{} requires proper initialization of the shared components $\bL$. If $\bL$ is na\"ively initialized at random, then the $\st$'s are unlikely to find a well-performing policy, and so updates to $\bL$ would not leverage any useful information. One common alternative is to initialize the $k$
columns of $\bL$ with the STL solutions to the first
$k$ tasks. However, this method prevents tasks $\Task^{(2)}, \ldots, \Task^{(k)}$  from leveraging information from  earlier tasks, impeding them from achieving potentially higher performance. Moreover, several tasks might rediscover redundant information, leading to wasted training time and capacity of $\bL$. The initialization method for \lpgftw{}, presented as Algorithm~\ref{alg:LifelongPolicyInitialization}, enables early tasks to leverage knowledge from previous tasks and prevents the discovery of redundant information. The algorithm starts from an empty set of components and adds a new error vector $\epsilont \in \Reals^d$ as a new component for each of the initial $k$ tasks. For each task $\Task^{(t)}$, the agent uses the base learner to simultaneously learn $\st$ to combine initialized components and an additional set of learnable parameters, $\epsilont$:
\begin{equation}
    \label{equ:LPGFTWInit}
    \st,\epsilont= \argmax_{\bs,\bepsilon} \Objt(\LtMinusOne\bs + \bepsilon) - \mu\|\bs\|_1 - \lambda\|\bepsilon\|_2^2\enspace.
\end{equation}
Intuitively, each $\epsilont$ finds knowledge of task $\Task^{(t)}$ that is not currently contained in $\bL$. Then, $\epsilont$ is incorporated as a new component in $\bL$. Once $\epsilont$ is included as a component, the agent makes no further modifications to it during initialization training via Equation~\ref{equ:LPGFTWInit}. 

\paragraph{Assimilation} Upon encountering a new task $\Task^{(t)}$, \lpgftw{} uses the base learner to optimize the task-specific coefficients $\st$, without modifying the shared components in $\bL$. This corresponds to searching for the optimal policy that can be obtained by combining the factors of $\bL$. 
Concretely, the agent strives to solve the following optimization:
\begin{equation}
    \st = \argmax_{\bs} \ell(\LtMinusOne, \bs)
   = \argmax_{\bs}\, \Objt(\LtMinusOne\bs) - \mu\|\bs\|_1\enspace,
 \label{equ:UpdateStGeneral}
\end{equation}
where $\LtMinusOne$ denotes the $\bL$ trained up to task $\Task^{(t-1)}$, $\Objt(\cdot)$ is any PG objective, and the $\ell_1$ norm encourages sparsity.  

\paragraph{Accommodation} Every $\texttt{adaptationFrequency}\gg1$ steps, the agent updates the components in $\bL$ with any relevant information collected from $\Task^{(t)}$ up to that point. 
Similar to \citet{bouammar2014online}, the agent approximates the MTL objective via a second-order Taylor expansion, yielding the following optimization objective:
\begin{align}
    \label{equ:UpdateL}
    \Lt = \argmax_{\bL} \hat{g}_{t}(\bL) =& \argmax_{\bL} -\lambda\|\bL\|_{\mathsf{F}}^2 + \frac{1}{t} \sum_{\that=1}^{t} \hat{\ell}\Big(\bL, \sthat, \alphathat, \Hthat, \gthat\Big)\\
    \hat{\ell}\Big(\bL, \sthat, \alphathat, \Hthat, \gthat\!\Big)=& -\!\mu\Big\|\sthat\Big\|_{1} \!+ \Big\|\alphathat\!-\bL\sthat\Big\|_{\Hthat}^2 \!+ \gthat{}^\top\Big(\bL\sthat\!-\alphathat\!\Big)\enspace,
\end{align}%
where $\hat{\ell}$ is the second-order approximation to the objective of a previously seen task $\Task^{(\that)}$. The learner evaluates the gradient \mbox{${\gthat=\nabla_{\btheta} \Objthat(\btheta)}$} and Hessian \mbox{${\Hthat=\frac{1}{2} \nabla_{\btheta,\btheta^\top} \Objthat(\btheta)}$} at the policy for task $\Task^{(\that)}$ immediately after training, \mbox{$\alphathat=\LthatMinusOne\sthat$}. 
The solution to this optimization can be obtained in closed form as $\vect(\Lt) =\bA^{-1} \bb$, where:
\begin{align}
    \bA =& -2\lambda\bI + \frac{2}{t}\sum_{\that=1}^{t} \left(\sthat{\sthat}^\top\right) \kron \Hthat\\ \qquad 
    \bb =& \frac{1}{t}\sum_{\that=1}^{t} \sthat \kron \left(-\gthat + 2\Hthat\alphathat\right) \enspace,
\end{align}
using $\otimes$ to denote the Kronecker tensor product.
Notably, these can be computed incrementally as each new task arrives, so that $\bL$ can be updated without preserving data or parameters from earlier tasks. Moreover, the Hessians $\Hthat$ needed to compute $\bA$ and $\bb$ can be discarded after each task if the agent does not expect to revisit tasks for further training. If instead the environment allows the agent to revisit tasks multiple times (e.g., for interleaved MTL), then each $\Hthat$ must be stored at a cost of $O(d^2\numTasks)$.

Intuitively, in Equation~\ref{equ:UpdateStGeneral} the agent leverages knowledge from all past tasks while training on task~$\Task^{(t)}$, by searching for $\thetat$ in the span of $\bL_{t-1}$. This makes \lpgftw{} fundamentally different from prior multimodel methods that learn each task's parameter vector in isolation and subsequently combine prior knowledge to improve performance. 
One potential drawback is that, by restricting the search to the span of $\LtMinusOne$, the agent might miss other, potentially better policies. However, any set of parameters far from the space spanned by $\LtMinusOne$ would be uninformative for the MTL objective, 
since the approximations to the previous tasks would be poor near the current task's parameters and vice versa. 
Then, in Equation~\ref{equ:UpdateL}, \lpgftw{} approximates the loss around the current set of parameters $\alphat$ via a second-order expansion and finds the $\Lt$ that optimizes the average approximate cost over all tasks seen so far, accommodating new knowledge while ensuring that the agent does not forget the knowledge required to solve the previously learned tasks.

\subsection{Computational Complexity}

The computational complexity of \lpgftw{} can be trivially obtained by following the derivations of \citet{ruvolo2013ella}. In particular, the assimilation stage introduces an (additive) overhead of $O(k\times d)$ per PG step, due to the multiplication of the gradient by $\bL^\top$. Additionally, every $\texttt{adaptationFrequency}\gg1$ steps, the accommodation step of $\bL$ takes an additional $O(d^3k^2)$. Notably, this latter value is constant with respect to the number of tasks seen so far, since Equation~\ref{equ:UpdateL} is solved incrementally, unlike other approaches in the supervised setting from Chapter~\ref{cha:Supervised}. This is possible thanks to the strong assumption that the inverse of $\bA$ in Equation~\ref{equ:UpdateL} can be computed at a cost of $O(d^3k^2)$, which is only feasible for reasonably small models.

If the number of parameters $d$ is too high, the learner could use faster techniques for solving the inverse of $\bA$, like the conjugate gradient method, or approximate the Hessian with a Kronecker-factored or diagonal matrix. While this dissertation did not test these approximations, they work well in related methods~\citep{bouammar2014online,ritter2018online}, so one would expect LPG-FTW to behave similarly. However, while the time complexity of LPG-FTW would remain constant with respect to the number of tasks for diagonal approximations, it would scale linearly for Kronecker-factored approximations, which require storing all Hessians and recomputing the cost for every new task.

\subsection{Base Policy Gradient Algorithms}
\label{sec:BaseLearners}

This section describes how two single-task PG learning algorithms can be used as the base learner of \lpgftw{}. 

\subsubsection{Episodic REINFORCE} 

The vanilla PG learning method, REINFORCE~\citep{williams1992simple}, updates parameters as:
\begin{align}
    \btheta_{j} \gets& \btheta_{j-1} + \eta_{j} \bg_{\btheta_{j-1}}\\
    \bg_{\btheta}=\nabla_{\btheta}\Obj(\btheta) =& \mathbb{E}\left[\sum_{i=0}^{\infty}\nabla_{\btheta}\log\pi_{\btheta}(\state_i,\action_i)A(\state_i, \action_i)\right]\enspace,
\end{align}
where $\bg_{\btheta}$ is the gradient of the policy and $A(\state, \action)$ is the advantage function. 
\lpgftw{} would then update the $\st$'s as:
\begin{align}
    \st_j \gets& \st_{j-1} + \eta_{j}\nabla_{\bs}\left[\Objt(\LtMinusOne\bs) - \mu\|\bs\|_1\right]\Big\lvert_{\bs=\st_j}\\
    \nabla_{\bs}\left[\Objt(\LtMinusOne\bs) - \mu\|\bs\|_1\right] =& \LtMinusOne^\top\bg_{\LtMinusOne\bs} - \mu\sign(\bs)\enspace.
\end{align}
The Hessian for Equation~\ref{equ:UpdateL} is given by $\bH = \frac{1}{2}\mathbb{E}\left[\sum_{i=0}^{\infty} \nabla_{\btheta,\btheta^\top}\log\pi_{\btheta}(\state_i,\action_i)A(\state_i,\action_i)\right]$, which evaluates to ${\bH = -\frac{1}{2\sigma^{2}}\mathbb{E}\left[\sum_{i=0}^{\infty}\state\state^\top A(\state_i,\action_i)\right]}$
in the case where the policy is a linear Gaussian (i.e., ${\pi_{\btheta}=\mathcal{N}\big(\btheta^\top\state,\sigma\big)}$). 
One major drawback of this is that the Hessian is not guaranteed to be negative definite, so Equation~\ref{equ:UpdateL} might move the policy arbitrarily far from the original policy used for sampling trajectories. 

\subsubsection{Natural Policy Gradient}
\label{sec:NaturalPolicyGradient}

The natural PG (NPG) algorithm gets around this issue. In particular, the formulation followed by \citet{rajeswaran2017towards} at each iteration optimizes:
\begin{align}
    \max_{\btheta} \qquad&\bg_{\btheta_{j-1}}^\top(\btheta - \btheta_{j-1}) \\
    \text{s.t.} \qquad&\|\btheta - \btheta_{j-1}\|_{\bF_{\btheta_{j-1}}}^2\leq \delta\enspace,
\end{align}
where ${\bF_{\btheta} = \mathbb{E}\left[\nabla_{\btheta}\log\pi_{\btheta}(\state,\action) \nabla_{\btheta}\log\pi_{\btheta}(\state,\action)^\top\right]}$ is the approximate Fisher information matrix of $\pi_{\btheta}$~\citep{kakade2002natural}. The base learner would then update the policy parameters at each iteration as:
\begin{equation}
    \btheta_{j} \gets \btheta_{j-1} + \eta_{\btheta} \bF^{-1}_{\btheta_{j-1}}\bg_{\btheta_{j-1}\enspace,
}
\end{equation}
with $\eta_{\btheta}=\sqrt{\frac{\delta}{\bg_{\btheta_{j-1}}^\top\bF^{-1}_{\btheta_{j-1}}\bg_{\btheta_{j-1}}}}$. To use NPG as the base learner, at each step \lpgftw{} solves:
\begin{align}
    \max_{\bs} \qquad& \bg_{\st_{j-1}}^\top(\bs - \st_{j-1}) \\
    \text{s.t.} \qquad& \Big\|\bs-\st_{j-1}\Big\|_{\bF_{\st_{j-1}}}^2\leq \delta\enspace,
\end{align}
which yields the update: 
\begin{align}
    \st_{j} \gets \st_{j-1} + \eta_{\st}\bF_{\st_{j-1}}^{-1}\bg_{\st_{j-1}}\enspace.
\end{align}
The Hessian for Equation~\ref{equ:UpdateL} is computed by  using the equivalent soft-constrained problem: 
\begin{equation}
 \widehat{\Obj}(\btheta) = \bg_{\btheta_{j-1}}^\top(\btheta - \btheta_{j-1}) + \frac{\|\btheta - \btheta_{j-1}\|_{\bF_{\btheta_{j-1}}}^2 - \delta}{2\eta_{\btheta}}\enspace,
\end{equation}
which gives $
    \bH = -\frac{1}{\eta_{\btheta}}\bF_{\btheta_{j-1}}
$. 
This Hessian \textit{is} negative definite, and thus encourages the parameters to stay close to the original ones, where the approximation is valid. 
\subsection{Connections to PG-ELLA}
\label{sec:ConnectionsToPGELLA}

\lpgftw{} and PG-ELLA~\citep{bouammar2014online} both learn a factorization of policies into $\bL$ and $\st$. To optimize the factors, PG-ELLA first trains individual task policies via STL, potentially leading to policy parameters that are incompatible with a shared $\bL$. In contrast, \lpgftw{} learns the $\st$'s directly via PG learning, leveraging shared knowledge in $\bL$ to accelerate the learning and restricting the $\alphat$'s to the span of $\bL$. This choice implies that, even if the agent finds the (typically infeasible) optimal $\st$, this may not result in an optimal policy, so Equation~\ref{equ:UpdateL} explicitly includes a linear term, which PG-ELLA omits. On the other hand, PG-ELLA typically initializes the shared components with the policies of the first $k$ tasks. Instead, \lpgftw{} exploits the policies from the few previously observed tasks to 1)~accelerate the learning of the earliest tasks and 2)~discover $k$ distinct knowledge components. These improvements enable the proposed method to operate in a true lifelong setting, where the agent encounters tasks sequentially. In contrast, PG-ELLA was evaluated in the easier interleaved MTL setting, in which the agent experiences each task multiple times, alternating between tasks frequently. These modifications also enable applying \lpgftw{} to far more complex dynamical systems than PG-ELLA, including domains requiring deep policies, previously out of reach for factored policy learning methods.

\subsection{Theoretical Guarantees}
\label{sec:TheoreticalGuarantees}

Part of the appeal of using linear combinations of model parameters is that it permits deriving strong theoretical guarantees. This dissertation proved that \lpgftw{} converges to an optimum of the (approximate) MTL objective for any ordering over tasks, despite the online approximation of keeping the $\st$'s fixed after initial training. In other words, \lpgftw{} finds a set of components that are in some approximate sense optimal under the observed task distribution. The proofs follow those of \citet{ruvolo2013ella} from the supervised setting, but are substantially adapted to handle the nonoptimality of the $\alphat$'s and the fact that the $\st$'s and $\bL$ optimize  different objectives. This section contains sketches of the main proofs, while complete proofs are available in Appendix~\ref{app:proofsOfTheoreticalGuarantees}.

The objective defined in Equation~\ref{equ:UpdateL}, $\hat{g}$, considers the optimization of each $\st$ separately with the $\Lt$ known up to that point, and is a surrogate for the true objective:
\begin{align}
    g_{t} (\bL) =& \frac{1}{t}\sum_{\that=1}^{t}\max_{\sthat} \bigg[\! \Big\|\alphathat \!-\bL\sthat\Big\|_{\Hthat}^2 \!\!+ {\gthat}^\top\!\!\Big(\bL\sthat\!-\alphathat\Big) -  \mu\Big\|\sthat\Big\|_1 \!\bigg] - \lambda\|\bL\|_{\mathsf{F}}^2\enspace,
\end{align}
which considers the simultaneous optimization of all $\st$'s. The expected objective is:
\begin{equation}
    g(\bL) = \mathbb{E}_{\Ht, \gt, \alphat} \Big[\max_{\bs}\hat{\ell}\Big(\bL, \bs, \alphat, \Ht, \gt\Big) \Big]\enspace,
\end{equation}
which measures how well $\bL$ can represent a random future task without accommodation. 

The main theoretical results presented in this section are that: 1)~$\Lt$ becomes increasingly stable, 2)~$\hat{g}_{t}$, $g_{t}$, and $g$ converge to the same value, and 3)~$\Lt$ converges to a stationary point of $g$. These results are based on the following assumptions:
\renewcommand{\theenumi}{\Alph{enumi}}
\begin{enumerate}
\itemsep0em
    \item \label{asmp:iid}The tuples $\Big(\Ht,\gt\Big)$ are drawn \iid~from a distribution with compact support.
    \item \label{asmp:mixing}The sequence $\Big\{\alphat\Big\}_{t=1}^{\infty}$ is stationary and $\phi$-mixing.
    \item \label{asmp:magnitude} The magnitude of $\Objt(\bm{0})$ is bounded by $B$.
    \item \label{asmp:uniqueness}For all $\bL$, $\Ht$, $\gt$, and $\alphat$, the largest eigenvalue (smallest in magnitude) of \mbox{$\bL_{\gamma}^\top\Ht\bL_{\gamma}$} is at most $-\kappa$, with $\kappa>0$, where $\gamma$ is the set of nonzero indices of \mbox{$\st=\argmax_{\bs}\hat{\ell}\Big(\bL, \bs,\Ht, \gt, \alphat\Big)$}. The nonzero elements of the unique maximizing $\st$ are given by: \mbox{$\st_{\gamma} = \big(\bL_{\gamma}^\top\Ht\bL_{\gamma}\big)^{-1}\Big(\bL^\top\Big(\Ht\alphat - \gt\Big) - \mu\sign\Big(\st_{\gamma}\Big)\Big)$}.
\end{enumerate}
\renewcommand{\theenumi}{\arabic{enumi}}

\begin{proposition}
\label{prop:Stability}
    $\bL_t - \bL_{t-1} = O(\frac{1}{t})\enspace$.
\end{proposition}
\begin{proof}[Proof sketch]
    The first step in the proof is to show that the entries of $\bL$, $\st$, and $\alphat$ are bounded by Assumptions~\ref{asmp:iid} and \ref{asmp:magnitude} and the regularization terms. The next step is showing that $\hat{g}_{t}\!-\!\hat{g}_{t-1}$ is $O\left(\frac{1}{t}\right)$--Lipschitz. The facts that $\LtMinusOne$ maximizes $\hat{g}_{t-1}$ and the eigenvalues of the Hessian of $\hat{g}_{t-1}$ are bounded complete the proof.
\end{proof}

The critical step for adapting the proof from \citet{ruvolo2013ella} to \lpgftw{} is to introduce the following lemma, which shows the equality of the maximizers of $\ell$ and $\hat{\ell}$.

\begin{lemma}
\label{lma:EqualityOfMaximizers} ${\hat{\ell}\Big(\bL_{t}, \bs^{(t+1)}, \balpha^{(t+1)}, \bH^{(t+1)}, \bg^{(t+1)}\Big) }= {\max_{\bs} \hat{\ell}\Big(\bL_{t}, \bs, \balpha^{(t+1)}, \bH^{(t+1)}, \bg^{(t+1)}\Big)}\enspace$.
\end{lemma}
\begin{proof}[Proof sketch]
    The facts that the STL objective $\hat{\ell}$ is a second-order approximation at $\bs^{(t+1)}$ of $\ell$ and that  $\bs^{(t+1)}$ is a maximizer of $\ell$ imply that $\bs^{(t+1)}$ is also a maximizer of $\hat{\ell}$.
\end{proof}

\begin{proposition}
\label{prop:Convergence}
~
\begin{enumerate}
        \item \label{part:firstPropositionConvergence} $\hat{g}_{t}(\bL_{t})$ converges a.s.
        \item \label{part:secondPropositionConvergence}$g_{t}(\bL_{t}) -\hat{g}_{t}(\bL_{t})$ converges a.s.~to $0$
        \item \label{part:thirdPropositionConvergence} {$g_{t}(\bL_{t}) - \hat{g}(\bL_{t})$ converges a.s.~to $0$}
        \item \label{part:fourthPropositionConvergence} $g(\bL_{t})$ converges a.s.\\
\end{enumerate}
\end{proposition}
\begin{proof}[Proof sketch]
    The first step is to use Lemma~\ref{lma:EqualityOfMaximizers} to show that the sum of negative variations of the stochastic process $u_{t}=\hat{g}_{t}(\Lt)$ is bounded. Given this result, the next step shows that $u_{t}$ is a quasi-martingale that converges almost surely (Part~\ref{part:firstPropositionConvergence}). This fact, along with a simple lemma of positive sequences, permits proving Part~\ref{part:secondPropositionConvergence}. The final two parts can be shown due to the equivalence of $g$ and $g_{t}$ as $t\to\infty$.
\end{proof}

\begin{proposition}
\label{prop:StationaryPoint}
    The distance between $\Lt$ and the set of all stationary points of $g$ converges a.s.~to $0$ as $t\to\infty$.
\end{proposition}
\begin{proof}[Proof sketch]
    The following two facts lead to completing the proof: 1)~$\hat{g}_{t}$ and $g$ have Lipschitz gradients with constants independent of $t$, and 2)~$\hat{g}_{t}$ and $g$ converge almost surely.
\end{proof}

\section{\compRL{}: Modular Lifelong Reinforcement Learning via Neural Composition}
\label{sec:LifelongCompositionalRLAlgo}

This section describes \compRL{}, the second of the lifelong RL approaches developed in this dissertation. This second method utilizes a version of the hard modular architecture introduced in Chapter~\ref{cha:Framework}, which is the representation that most accurately represents the problem of compositional learning as described in Section~\ref{sec:lifelongCompositionInRL}. 

At a high level, the agent constructs a different neural net policy for every task by selecting from a set of available modules. The modules themselves are used to accelerate the learning of each new task and are then automatically improved by the agent with new knowledge from this latest task. 

\subsection{Neural Modular Policy Architecture}
\label{sec:ModularArchitectureRL}

Few recent works have attempted to solve modular RL problems via neural composition, but without a substantial effort to study their applicability to truly compositional problems~\citep{yang2020multi,goyal2021recurrent}. One notable exception proposed a specialized modular architecture to handle multitask, multirobot problems~\citep{devin2017learning}. This latter architecture inspired the design of the new modular policy described in this section, which tackles compositional RL more generally. 

Following the assumptions of Section~\ref{sec:lifelongCompositionInRL}, each neural module $\modulei$ is in charge of solving one specific subproblem $\subproblemi$ (e.g., finding an object's grasping point in the robot tasks), such that there is a one-to-one and onto mapping from subproblems to modules. All tasks that require solving $\subproblemi$ share $\modulei$. 
To construct the network for a task, the model chains modules in sequence, thereby replicating   
the graph structure depicted in Figure~\ref{fig:compositionalRLGraph} with neural modules. 

Most modular architectures consider a pure chaining structure, in which the complete input passes through a sequence of modules. In such architectures, each module must not only process the information needed to solve its subproblem (e.g., the obstacle in the robot examples), but also pass through information required by subsequent modules. Additionally, the chaining structure induces brittle dependencies among the modules, and so a tiny change to the first module can have cascading effects.  While in MTL it is viable to learn such complex modules, in the lifelong setting the modules must generalize to unseen combinations with other modules after training on just a few tasks in sequence. A better solution is to let each module $m_i$ only receive  information needed to solve its subproblem $F_i$, ensuring that it only needs to output the solution to $F_i$. \citet{devin2017learning} used this insight to design a similar modular architecture. Therefore, the architecture used by \compRL{} assumes that the state can factor into module-specific components, such that each subproblem $F_i$ requires only access to a subset of the state components and passes only the relevant subset to each module. For example, in the robotics domain, robot modules only receive as input the state components related to the robot state. Equivalently, the model treats each element of the state vector as a variable and feeds only the variables necessary for solving each subproblem $F_i$ into $m_i$. This process requires only high-level information about the semantics of the state representation, similar to the architecture of \citet{devin2017learning}.

At each depth $\depth$ in the modular net, the agent has access to $\totalModules_\depth$ modules. Each module is a small neural net that takes as inputs the module-specific state component and the output of the module at the previous depth $d\!-\!1$. Note that this differs from gating networks in that the modular structure is fixed for each individual task, instead of modulated by the input.  The number of modular layers $d_{\mathrm{max}}$ is the number of subproblems that must be solved (e.g., $d_{\mathrm{max}}=3$ for (1)~grasping an object and (2)~avoiding an obstacle with (3)~a robot arm).  Section~\ref{sec:ModelArchitecturesRL} describes the exact architectures used for the experiments.

\subsection{Sequential Learning of Neural Modules}
\label{sec:sequentialLearning}

\compRL{} follows the framework prescribed in Chapter~\ref{cha:Framework}, yet is in some ways fundamentally different from the supervised learning methods from Chapter~\ref{cha:Supervised} and even \lpgftw{} from Section~\ref{sec:LPG-FTWAlgo}. The key difference is that the assimilation stage does not restrict the agent to only searching over combinations of fixed, existing modules. Instead, it allows the agent to explore beyond those rigid modules, which was necessary to achieve reasonable performance given the strict modularity assumptions of the architecture. The accommodation stage then uses off-line RL techniques to incorporate new knowledge obtained during the exploration stage into the existing modules. The following paragraphs describe the stages of the approach in detail, and Algorithm~\ref{alg:LifelongCompositionalRL} summarizes the overall process.

\begin{figure}[t!]
  \centering
  \begin{minipage}{.85\linewidth}
    \begin{algorithm}[H]   \caption{Lifelong Compositional RL}
  \label{alg:LifelongCompositionalRL}
\begin{algorithmic}[1]
    \STATE $T\gets0$
    \WHILE {$\,\,\Taskt\gets\mathtt{getTask}()$}
        \IF [initialization]{$\,\,T \leq k$}%
            \STATE ${\tt steps} \gets 0; \hspace{1em} s_d \gets T \,\, \forall d$
        \ELSE[find module combination]
            \STATE $s, {\tt steps} \gets {\tt discreteSearch()}$
            \STATE ${\tt bckModules} \gets {\tt clone}(M)$
        \ENDIF
        \STATE $\pi^{(t)}, Q^{(t)} \gets {\tt modularNets}(M, s)$
        \WHILE[online exploration]{$\,\,{\tt steps} < {\tt onlineSteps}$}
            \STATE $\state, \action, \reward, \nextstate \gets {\tt rollouts}\Big(\pi^{(t)}, {\tt iterSteps}\Big)$%
            \STATE ${\tt buffer}[t].{\tt push}(\state, \action, \reward, \nextstate)$
            \STATE $\pi^{(t)}\!, Q^{(t)} \gets {\tt PPOStep}\Big(\state, \action, \reward, \nextstate, \pi^{(t)}\!, Q^{(t)}\Big)$%
            \STATE ${\tt steps} \gets {\tt steps} + {\tt iterSteps}$
        \ENDWHILE
        \IF {$\,\,T < {\tt numModules}$}
            \STATE $M \gets {\tt bckModules}$
        \ENDIF
        \FOR[accommodation via off-line RL] {$\,\,i = 1,\ldots,{\tt offlineSteps}$}
            \FOR {$\,\, t = 1,\ldots,{\tt seenTasks}$}
                \STATE $\state, \action, \reward, \nextstate \gets {\tt buffer}[t].{\tt sample()}$
                \STATE $\pi^{(t)} \gets \argmin_{\tilde{\pi}} {\tt NLL}(\tilde{\pi}(\state), \action)$
                \STATE $\action^\prime \gets \argmax_{\tilde{\action}: \pi^{(t)}(\nextstate, \tilde{\action}) > \tau} Q^{(t)}(\nextstate, \tilde{\action})$
                \STATE ${\tt targetQ} \gets r + {Q^{(t)}}^\prime(\nextstate,\action^\prime)$
                \STATE $Q^{(t)} \gets \argmin_{\tilde{Q}} (\tilde{Q}(\state, \action) - {\tt targetQ})$
            \ENDFOR
        \ENDFOR
    \ENDWHILE
\end{algorithmic}
\end{algorithm}
\end{minipage}
\end{figure}

\paragraph{Initialization} Finding a good set of initial modules is a major challenge. In order to achieve lifelong transfer to future tasks, the agent must start from a sufficiently strong and diverse set of modules. To achieve this, \compRL{} learns each new task on disjoint sets of neural modules until it has initialized all modules. Intuitively, this corresponds to the assumption that the first tasks contain disjoint sets of components. The empirical evaluation of Section~\ref{sec:resultsMiniGrid} shows that this assumption is not necessary in practice.

\paragraph{Assimilation} Like other methods in this dissertation, the assimilation stage keeps shared modules fixed in order to avoid incorporating new (likely incorrect) information about the current task during the initial stages of training on that task. However, the strict modular architecture used by \compRL{} makes it difficult to combine existing modules into a solution that achieves high performance without improving the modules. While in principle subsequent accommodation stages could achieve such improvements, in practice it becomes infeasible for an RL agent to discover high-performing behaviors without ever having experienced them online. Therefore, \compRL{} subdivides the assimilation stage into two stages:
\begin{itemize}
\item {\em Module selection}\hspace{2em}Upon encountering a new task, the agent selects the best modules to solve the task without any modifications to the module parameters. This ensures that the choice of modules requires minimal modifications to those parameters, as is needed to retain knowledge of the earlier tasks. The evaluation considered two versions of the method. In one case, the environment gives the agent the correct choice of modules, and the agent does not need to search over module combinations. In the other (more general) case, the agent must discover which modules are relevant to the current task. Since the previous stage has already initialized a diverse set of modules, this is achieved via exhaustive search of the possible module combinations in terms of the reward they yield when combined.
This type of combinatorial search requires the least amount of additional assumptions. In practice, this process rolls out the policy resulting from each of the possible combinations in the environment for some number of episodes, and chooses the combination that yields the highest average return.

\item {\em Exploration}\hspace{2em}Once the agent has chosen the set of modules to use for the current task, it might be able to perform reasonably well on the task. However, especially on the earliest tasks, when the agent has not yet learned fully general modules, it is unlikely that modules that have never combined into a trained policy work together perfectly. In the supervised setting, the agent has access to a data set that is representative of the data distribution of the current task, which enables the agent to incorporate knowledge into the modules directly using the provided data after module selection. However, in RL there is no given data set, and the agent must instead explore the environment to collect data that represents near-optimal behavior. In order to avoid catastrophic damage to existing neural components and at the same time enable full flexibility for exploring the current task, \compRL{} executes this exploration via standard RL training on a \textit{copy} of the shared modules, leveraging the selected modules to initialize the policy, but without modifying the actual shared module parameters. 
\end{itemize}
The rationale for executing selection and exploration separately is that the selected modules should be updated as little as possible in the next and final stage. If the learner instead jointly explores via module adaptation and searches over module configurations, it would likely find a solution that makes drastic modifications to the selected modules. If instead it restricts module selection to the fixed modules, then this stage is more likely to find a solution that requires little module adaptation.

\paragraph{Accommodation} While the exploration stage enables learning about the current task, this is typically insufficient for training a lifelong learner, since 1)~the learner would need to store all copies of the modules in order to perform well on each task, and 2)~the modules obtained from initialization are often suboptimal, limiting their potential for future transfer. For this reason, once the agent has gained enough experience on the current task, it incorporates newly discovered knowledge into the existing modules. It is crucial that this accommodation step does not discard knowledge from earlier tasks, which is not only necessary for solving those earlier tasks (thereby avoiding catastrophic forgetting) but possibly also for future tasks (enabling forward transfer). One popular strategy in the supervised setting to incorporate new knowledge into shared parameters is to train parameters with a mix of new and replay data. However, while experience replay has been tremendously successful in the supervised setting, its success in RL has been limited to very short sequences of tasks~\citep{isele2018selective,rolnick2019experience}. One of the challenges of using experience replay in RL is that training RL policies off-line tends to degrade performance, due to a mismatch between the off-line data distribution and the data distribution imposed by the updated policy. Consequently, this dissertation proposed a novel method for experience replay based on recent off-line RL techniques, designed precisely to avoid this issue. Concretely, at this stage \compRL{} uses off-line RL over the replay data from all previous tasks as well as the current task, keeping the selection over components fixed and modifying only the shared modules. Note that this is a general solution that applies beyond compositional algorithms to any lifelong method that includes a stage of incorporating knowledge into a shared repository of knowledge after online exploration. Section~\ref{sec:Ablations}  validates that this drastically improves the performance of one noncompositional method.

\subsection{Base Online and Off-Line Reinforcement Learning Techniques}

The implementation of \compRL{} used for experiments in Section~\ref{sec:ExperimentalEvaluationRL} uses \textit{proximal policy optimization} (PPO; \citealp{schulman2017proximal}) for assimilation via RL exploration and \textit{batch-constrained $Q$-learning} (BCQ; \citealp{fujimoto2019off,fujimoto2019benchmarking}) for accommodation via off-line RL. BCQ simultaneously trains an actor $\pi$ to mimic the behavior in the data and a critic $Q$ to maximize the values of actions that are likely under the actor $\pi$. In the lifelong setting, this latter constraint ensures that the $Q$-values are not over-estimated for states and actions that were not observed during the assimilation stage. 

One caveat of using PPO and BCQ together in the proposed algorithm is that BCQ requires training an action-value function $Q$, whereas PPO trains a state-value function $V$ as the critic. For compatibility, \compRL{} uses a modified version of PPO that instead trains a $Q$-function and computes $V$ from the learned $Q$. In the discrete setting, this is done by assuming a deterministic actor and computing the maximum value from the computed $Q$-values: $V=\max_a Q(s, a)$. In the continuous setting, the learner instead samples $n$ actions and computes the average $Q$-value across all of them to obtain an approximation of $V$.

One of the primary reasons to choose BCQ as the off-line RL mechanism of \compRL{}, beside its conceptual simplicity, was its flexibility to be applied in both discrete- and continuous-action settings. The following paragraphs provide additional details about the BCQ implementations used in \compRL{}. Note that other off-line RL mechanisms could be used in place of BCQ.

\paragraph{Discrete BCQ} In the discrete-action setting, BCQ trains an actor and a critic. The actor trains to imitate the behavior distribution in the data, and the critic trains to compute the $Q$-values, constrained to regions of the data distribution with high probability mass. Concretely, instead of computing $Q(s,a)=r + \gamma\max_{a^\prime}Q(s^\prime, a^\prime)$ as dictated by the standard Bellman equation, BCQ computes $Q=r + \gamma\max_{a^\prime: \pi(s, a^\prime) > \tau} Q(s^\prime, a^\prime)$, where $\tau$ is a threshold below which actions are considered too unlikely under the data distribution. In the experiments, the accommodation stage updated the PPO actor $\pi$ as the behavior cloning actor in BCQ, and the PPO critic $Q$ as the BCQ critic. After this stage, the evaluations of the policy rolled out actions via Boltzmann sampling of $Q$ over actions above the likelihood threshold $\tau$.
\paragraph{Continuous BCQ} The original continuous-action BCQ is substantially more complex, since the actor should mimic a general distribution over the continuous-valued actions, for which BCQ uses a variational autoencoder to represent the arbitrary distribution. However, in the case considered in \compRL{}, the data is itself generated by a Gaussian actor, and therefore one can assume that a Gaussian policy can represent the data distribution. Note that, since the training updates the exploration policy over time, this is merely an approximation. Therefore, like in the discrete case, the PPO actor $\pi$ trains to imitate the data distribution. This choice also permits dropping the perturbation model trained in BCQ to model a separate actor. BCQ trains two separate $Q$-functions in a manner similar to clipped double $Q$-learning~\citep{fujimoto2018functionapprox}. The target value is computed by:
\begin{equation}
r+\gamma \max_{a^\prime\sim\pi(s)} \lambda \min_{j=\{1,2\}}Q_j(s^\prime, a^\prime) + (1-\lambda)\max_{j=\{1,2\}} Q_j(s^\prime, a^\prime) \enspace, 
\end{equation}
which is estimated by sampling $n$ actions from the actor $\pi$. Subsequent evaluations in the experiments also sampled actions directly from the actor $\pi$.

\section{Experimental Evaluation}
\label{sec:ExperimentalEvaluationRL}

The experiments evaluated the capabilities of the proposed lifelong RL methods. Similar to the supervised learning experiments of Chapter~\ref{cha:Supervised}, this chapter divides tests according to the compositional structure used by the learner. First, the model with linear combinations of policy parameters trained via \lpgftw{} was evaluated on a set of tasks that are not explicitly compositional. This evaluation demonstrated that \lpgftw{} substantially increases the learning speed and a dramatically reduces catastrophic forgetting. Next, the experiments tested the model with hard neural module compositions trained via \compRL{} on a set of explicitly compositional tasks especially created for this evaluation. This study showed that these novel tasks are indeed compositional: once the agent has trained a suitable set of components, novel combinations of those components can solve unseen tasks. Moreover, the experiments revealed that the proposed algorithm can learn these components sequentially in a lifelong setting, dramatically increasing the learning speed and decreasing forgetting.

Code and training videos for experiments involving \lpgftw{} are available at~\url{https://github.com/Lifelong-ML/LPG-FTW}, and code for experiments involving \compRL{} is available at~\url{https://github.com/Lifelong-ML/Mendez2022ModularLifelongRL}.

\subsection{Evaluation Domains}
\label{sec:EvaluationDomainsRL}

The evaluation used four different domains to assess the two lifelong RL methods developed in this dissertation. The first two domains evaluated \lpgftw{} on noncompositional tasks and were taken directly or with minor modifications from existing literature, whereas the latter two evaluated \compRL{} on explicitly compositional tasks and were created specifically for this evaluation to assess the compositional properties of \compRL{}.

\subsubsection{OpenAI Gym MuJoCo Domains} The first experiments evaluated \lpgftw{} on simple MuJoCo environments from OpenAI Gym~\citep{todorov2012mujoco, brockman2016openai}. For this, the agent trained on two different evaluation domains for each of the HalfCheetah, Hopper, and Walker-2D environments: a \textit{gravity} domain, where each task corresponded to a random gravity value between $0.5g$ and $1.5g$, and a \textit{body-parts} domain, where the size and mass of each of four parts of the body (head, torso, thigh, and leg) were randomly set to values between $0.5\times$ and $1.5\times$ their nominal values. These choices lead to highly diverse tasks, as demonstrated in Section~\ref{sec:ExperimentalResultsSimple}. A modified version of the {\tt gym-extensions}~\citep{henderson2017multitask} package generated the tasks, scaling each body part independently to achieve a higher diversity.

The evaluation considered $\numTasks=20$ tasks for the HalfCheetah and Hopper domains, and $\numTasks=50$ for the Walker-2D domains. The agents trained on each task for a fixed number of iterations before moving on to the next.

\subsubsection{Meta-World Domains} To study the flexibility of the framework, the next evaluation tested \lpgftw{} on Meta-World~\citep{yu2019meta}, a substantially more challenging benchmark. The tasks involve using a simulated Sawyer robotic arm to manipulate various objects in diverse ways, and have been notoriously difficult for state-of-the-art MTL and meta-learning algorithms. Concretely, \lpgftw{} trained on sequential versions of the MT10 benchmark, with $\numTasks=10$ tasks, and the MT50 benchmark, using a subset of $\numTasks=48$ tasks. All Meta-World tasks were simulated on version 1.5 of the MuJoCo physics simulator~\citep{todorov2012mujoco}. The observation space for each task was six-dimensional, comprising the robot hand and the object locations. Note that the goal, which was kept fixed for each task, was not given to the agent. For this reason, the evaluation removed two tasks from MT50 which are incompatible with the six-dimensional observations---{\tt stick pull} and {\tt stick push}---for a total of $\numTasks=48$ tasks.

\subsubsection{Discrete 2-D World}

The discrete 2-D world was the first truly compositional set of tasks used for RL experiments in this dissertation. Each task in the domain consists of an $8\times8$ grid of cells populated with a variety of objects, and is built upon \texttt{gym-minigrid}~\citep{gym_minigrid}. The agent's goal is to reach a specific target in an environment populated with static objects that have different effects. 
The tasks are compositional in three hierarchical levels: 
\begin{itemize}
\item{\em Agent dynamics}\hspace{2em}The domain contains four different artificially created dynamics, each corresponding to a permutation of the actions (e.g., the \texttt{turn\_left} action moves the agent forward, \texttt{turn\_right} rotates left). For each task, the chosen dynamics determine the effect of the agent's actions. 
\item{\em Static objects}\hspace{2em}Each task contains a chain of static objects, with a single gap cell. \texttt{Wall}s block the agent's movement; the agent must open a \texttt{door} in the gap cell to move to the other side. \texttt{Floor} cells have an object indicator, but have no effect on the agent. \texttt{Food} gives a small positive reward if picked up. Finally, \texttt{lava} gives a small negative reward and terminates the episode.
\item{\em Target objects}\hspace{2em}There are four colors of targets; each task involves reaching one color.
\end{itemize}

There are $\numTasks=64$ tasks of the form ``reach the \texttt{COLOR} target with action permutation \texttt{N} while interacting with \texttt{OBJECT}.''  If the agent has learned to ``reach the red target with permutation $0$'' and ``reach the green target with permutation $1$'', then it should be able to quickly discover how to ``reach the red target with permutation $1$'' by recombining the relevant knowledge. 

\begin{figure}[b!]
\centering
    \begin{subfigure}[b]{0.4\textwidth}
        \centering
        \includegraphics[trim={1.7cm 0.4cm 1.7cm 0cm}, clip=True, height=5cm]{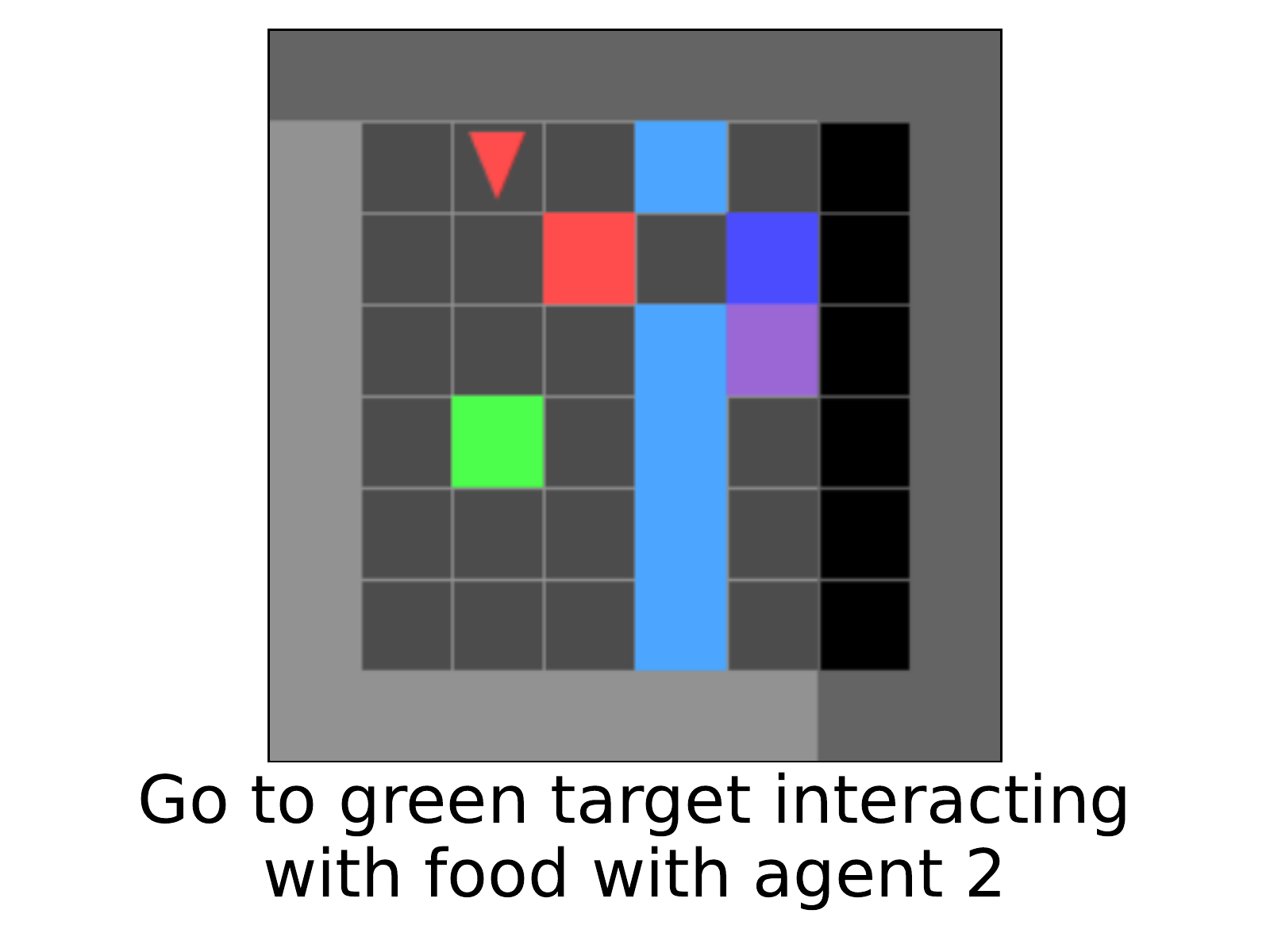}
    \end{subfigure}%
    \begin{subfigure}[b]{0.4\textwidth} 
        \centering
        \includegraphics[trim={1.7cm 0.4cm 1.7cm 0cm}, clip=True, height=5cm]{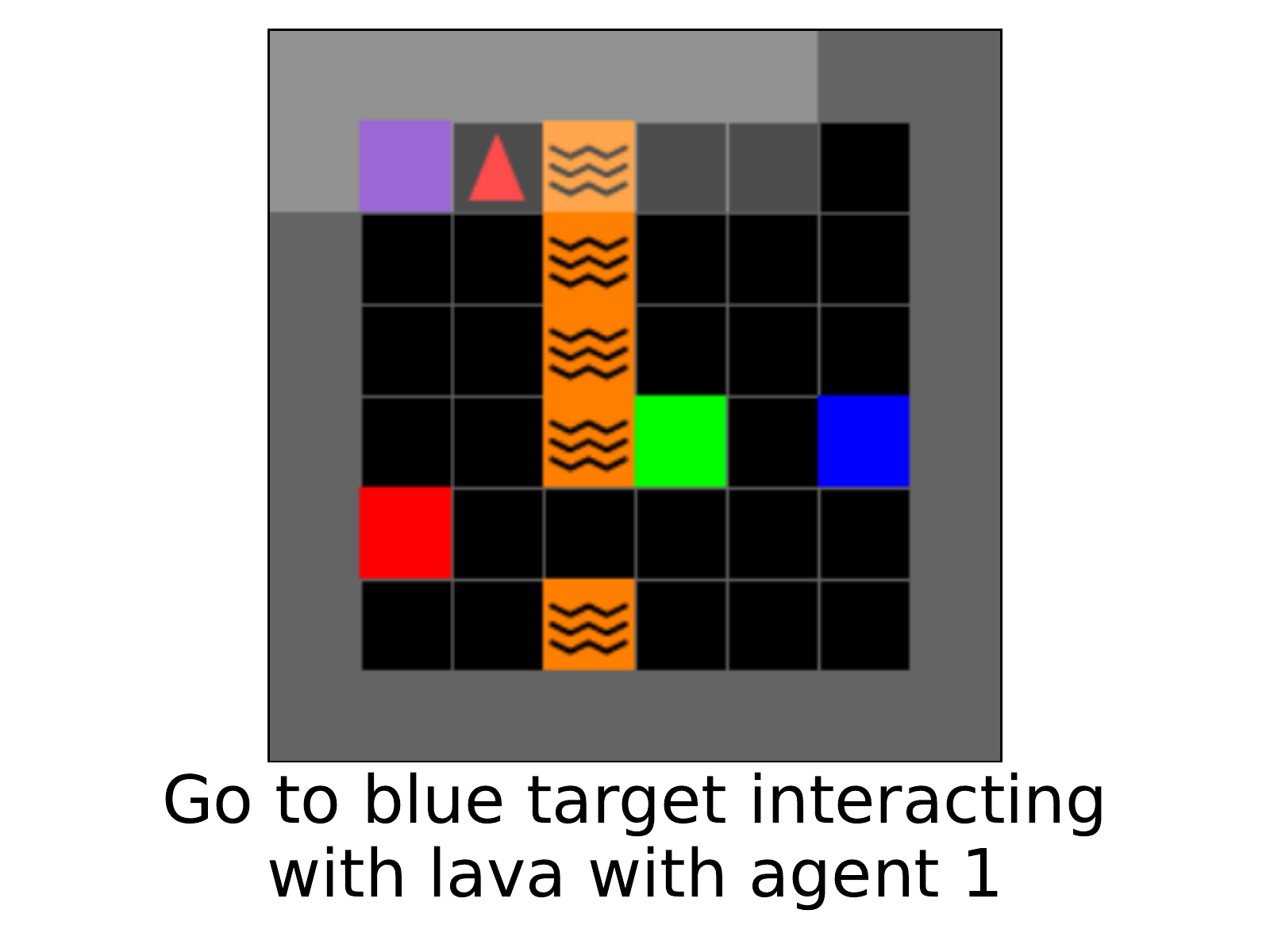}
    \end{subfigure}\\
    \vspace{1em}
    \begin{subfigure}[b]{0.4\textwidth}  
        \centering
        \includegraphics[trim={1.7cm 0.4cm 1.7cm 0cm}, clip=True, height=5cm]{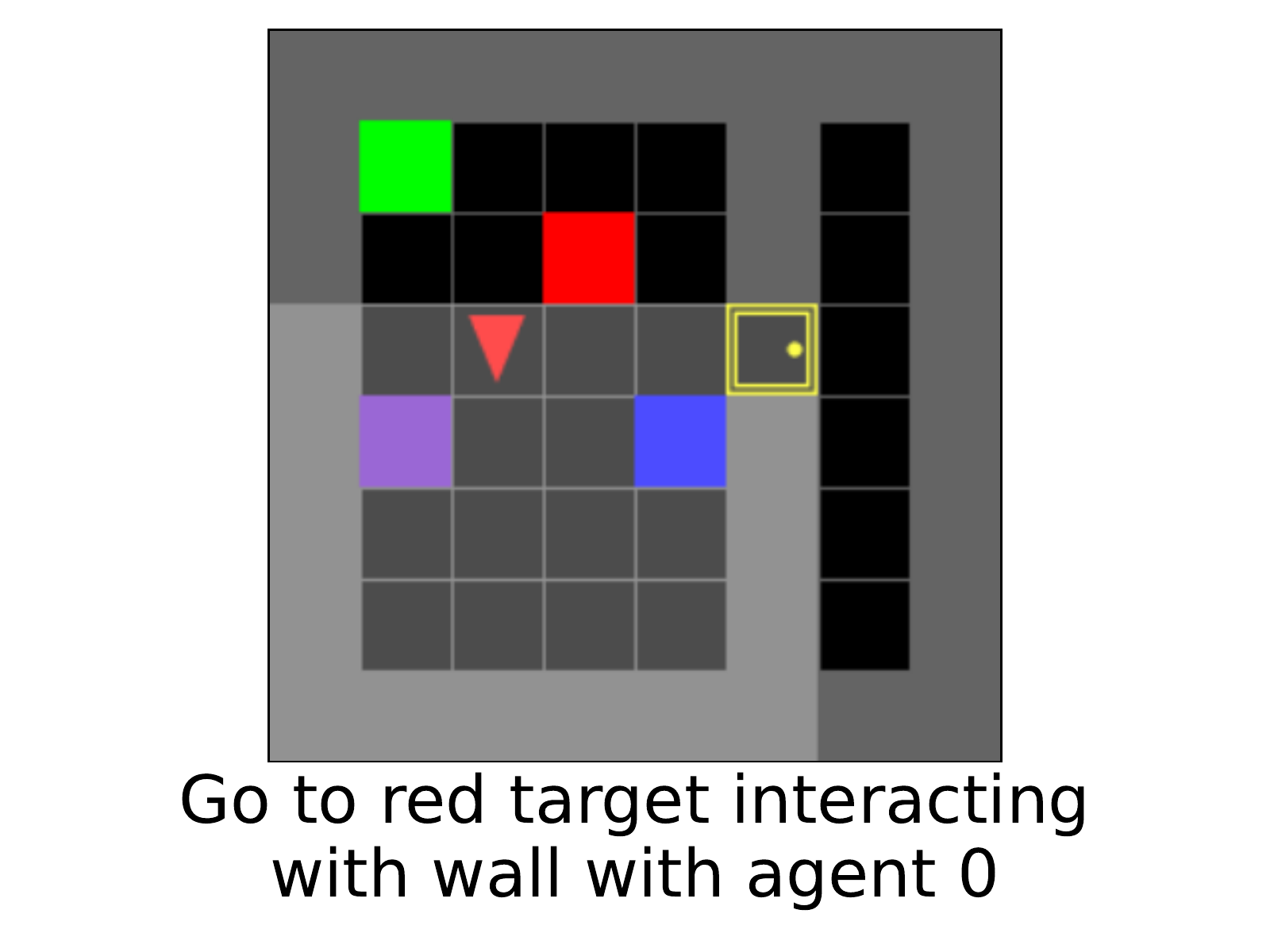}
    \end{subfigure}%
    \begin{subfigure}[b]{0.4\textwidth}  
        \centering
        \includegraphics[trim={1.7cm 0.4cm 1.7cm 0cm}, clip=True, height=5cm]{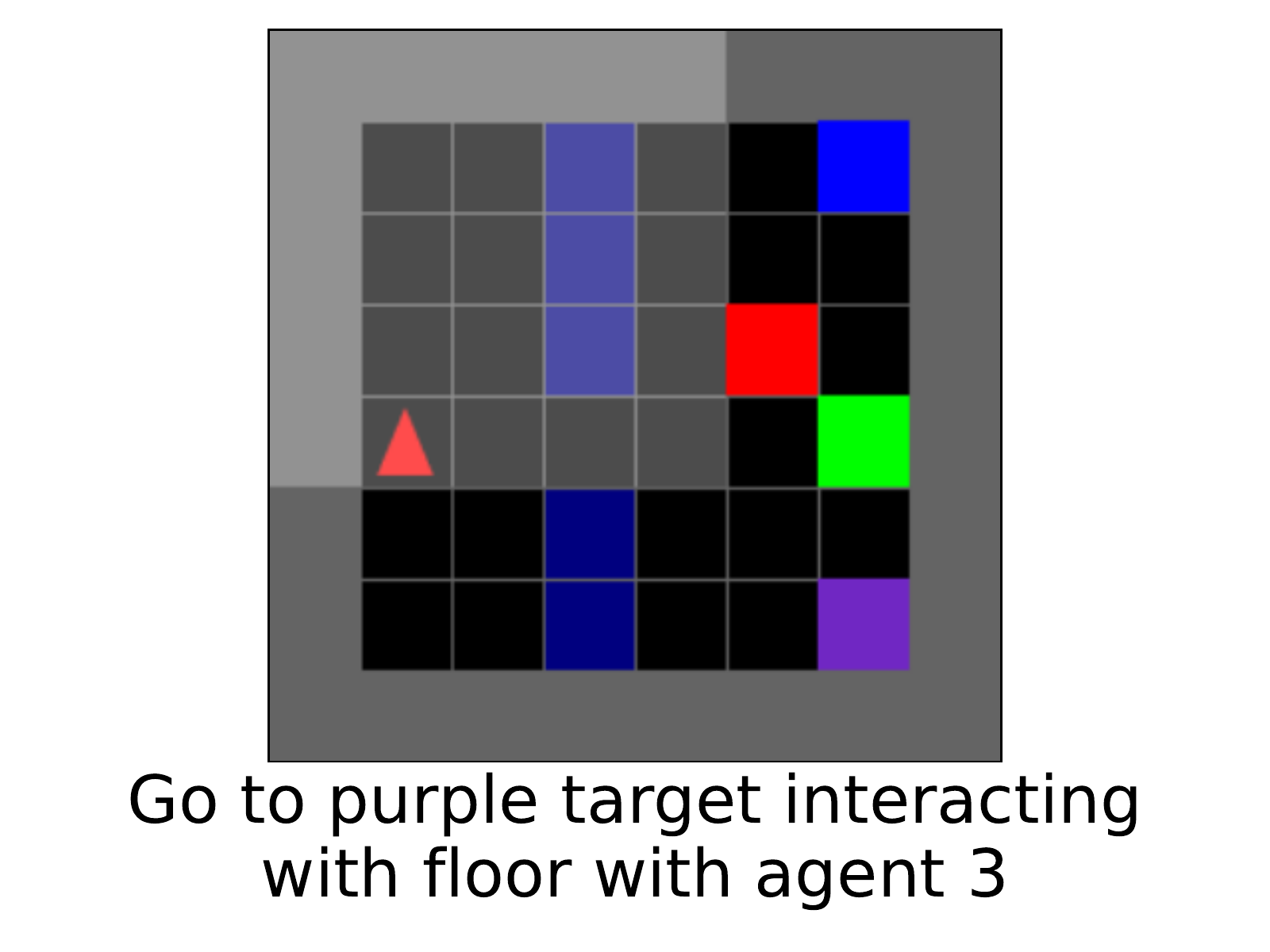}
    \end{subfigure}%
    \caption[Visualization of explicitly compositional discrete 2-D RL tasks.]{Visualization of various instantiations of the compositional discrete 2-D tasks. The highlighted area represents the agent's field of view. }
    \label{fig:MiniGridTasks}
 \end{figure}

Figure~\ref{fig:MiniGridTasks} shows example tasks created by sampling one component of each type. 
The paragraphs below describe the core elements of the proposed environment and how they vary according to the task components. 

\paragraph{Observation space} The learner receives a partially observable view of the $7\times7$ window in front of it, organized as an $ h \times w \times c$ image-like tensor, where $h=w=7$ are the height and width of the \texttt{agent}'s field of view, and $c=7$ is the number of channels. Each channel corresponds to one of \texttt{wall}, \texttt{floor}, \texttt{food}, \texttt{lava}, \texttt{door}, \texttt{target}, and \texttt{agent}. The first four channels are binary images, with ones at locations populated with the relevant objects. The \texttt{door} channel contains a zero for any cell without a \texttt{door} or with an open \texttt{door}, and a one for any cell with a closed \texttt{door}. The \texttt{target} channel has all-zeros except for the locations of the \texttt{target}, which are populated with an integer indicator of the color between one and four. Finally, the \texttt{agent} channel has all-zeros except for the location of the \texttt{agent}, which is populated with an integer indicator of the \texttt{agent}'s orientation (\texttt{right}, \texttt{down}, \texttt{left}, or \texttt{up}) between one and four. This separation into channels matches the assumptions of the architecture defined in Section~\ref{sec:ModularArchitectureRL}. The \texttt{agent} can observe past all objects except \texttt{walls} and closed \texttt{doors}, which occlude any objects beyond them.

\paragraph{Action space} Every time step, the agent can execute one of six actions: \texttt{turn\_left}, \texttt{turn\_right}, \texttt{move\_forward}, \texttt{pick\_object}, \texttt{drop\_object}, and \texttt{open\_door}. These discrete actions are deterministic, such that they always have the intended outcome. However, tasks have distinct dynamics that permute the ordering of the actions in the following four orders; for readability, actions whose effect stays constant across permutations are grayed out:%
\begin{enumerate}
\setcounter{enumi}{-1}
    \item {\color{gray}\texttt{turn\_left}}, \texttt{turn\_right}, \texttt{move\_forward}, {\color{gray}\texttt{pick\_object}}, {\color{gray}\texttt{drop\_object}}, \texttt{open\_door}
    \item {\color{gray}\texttt{turn\_left}}, \texttt{turn\_right}, \texttt{open\_door}, {\color{gray}\texttt{pick\_object}}, {\color{gray}\texttt{drop\_object}}, \texttt{move\_forward}
    \item {\color{gray}\texttt{turn\_left}}, \texttt{move\_forward}, \texttt{turn\_right}, {\color{gray}\texttt{pick\_object}}, {\color{gray}\texttt{drop\_object}}, \texttt{open\_door}
    \item {\color{gray}\texttt{turn\_left}}, \texttt{move\_forward}, \texttt{open\_door}, {\color{gray}\texttt{pick\_object}}, {\color{gray}\texttt{drop\_object}}, \texttt{turn\_right}
\end{enumerate}

\paragraph{Reward function} The main component of the reward is the original sparse reward function provided by \texttt{gym-minigrid}, which gives a zero at every time step except at the end of a successful episode. The environment computes the terminal reward value as $\Rewards_i=1 - 0.9 (i / H)$, where $i$ is the time step at which the agent reaches the target, and $H$ is the horizon of the environment. In tasks where \texttt{food} is present, the agent gets an additional reward of $0.05$ for every piece of food it picks up with the \texttt{pick\_object} action. In contrast, the agent receives a penalty of $-0.05$ if it steps on a \texttt{lava} object.

\paragraph{Initial conditions} The $8\times 8$ grid is surrounded by a wall. Each episode sets the initial state by randomly sampling the locations of all objects in the scene. First, the initialization places the static object at a horizontal location $x$ in the range $[2, w-2]$. All cells $i,j$ such that $i=x$ are populated with the task's static object, except for one individual cell at a random vertical location $y: i,j=x,y$. Cell $x,y$ is empty in all tasks except those whose static object is \texttt{wall}, in which case the cell contains a closed \texttt{door}. The agent starts at some random location not occupied by the static object, and facing randomly in any of the four possible directions. Finally, the environment places one target object of each of the four possible colors randomly in any remaining free spaces in the environment. 

\paragraph{Episode termination} For all tasks, the episode terminates upon reaching the correct target, or after $H=64$ time steps, whichever happens first. The episode immediately terminates if the agent steps on a \texttt{lava} object. 

These 2-D tasks capture the notion of functional composition studied in this dissertation, and proved to be notoriously difficult for existing lifelong RL methods. This demonstrates both the difficulty of the problem of knowledge composition and the plausibility of neural composition as a solution. However, artifices like action permutations and lava obstacles fail to show the real-world applicability of the proposed problem. 
Therefore, as a more realistic example, this dissertation introduced a second domain of different robot arms performing a variety of manipulation tasks that vary in three hierarchical levels. 

\clearpage
\subsubsection{Robot Manipulation}

The primary component of this domain is a set of four popular commercial robotic arms with seven degrees of freedom (7-DoF): Rethink Robotics' Sawyer, KUKA's IIWA, Kinova's Gen3, and Franka's Panda. Each task consists of one robot arm in a continuous state-action space, with a single object and (optionally) an obstacle. All robots use a general-purpose gripper by Rethink Robotics with two parallel fingers. The task components are:
\begin{itemize}
\item{\em Robots}\hspace{2em}These manipulators have varying kinematic configurations, joint limits, and torque limits, requiring specialized policies to actuate each of the arms. All dynamics were simulated in \robosuite{}~\citep{robosuite2020}.
\item {\em Objects}\hspace{2em}The robot must grasp and lift a can, a milk carton, a cereal box, or a loaf of bread. Their varying geometry implies that no common strategy can manipulate all these objects to solve the tasks.  
\item {\em Obstacles}\hspace{2em}The robot's workspace may be free (i.e., no obstacle), blocked by a wall the robot needs to circumvent, or limited by a door frame that the robot must traverse. 
\end{itemize}
\begin{figure}[t!]
\centering
    \begin{subfigure}[b]{0.4\textwidth}
        \centering
        \includegraphics[height=5.5cm]{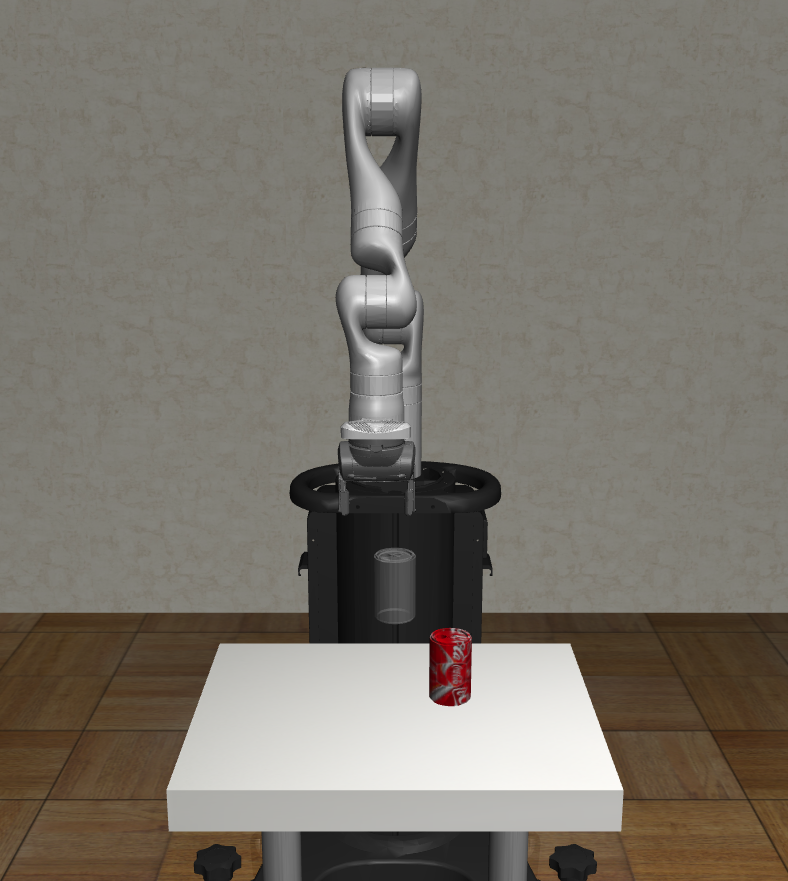}
        \vspace{1em}
    \end{subfigure}%
    \begin{subfigure}[b]{0.4\textwidth} 
        \centering
        \includegraphics[height=5.5cm]{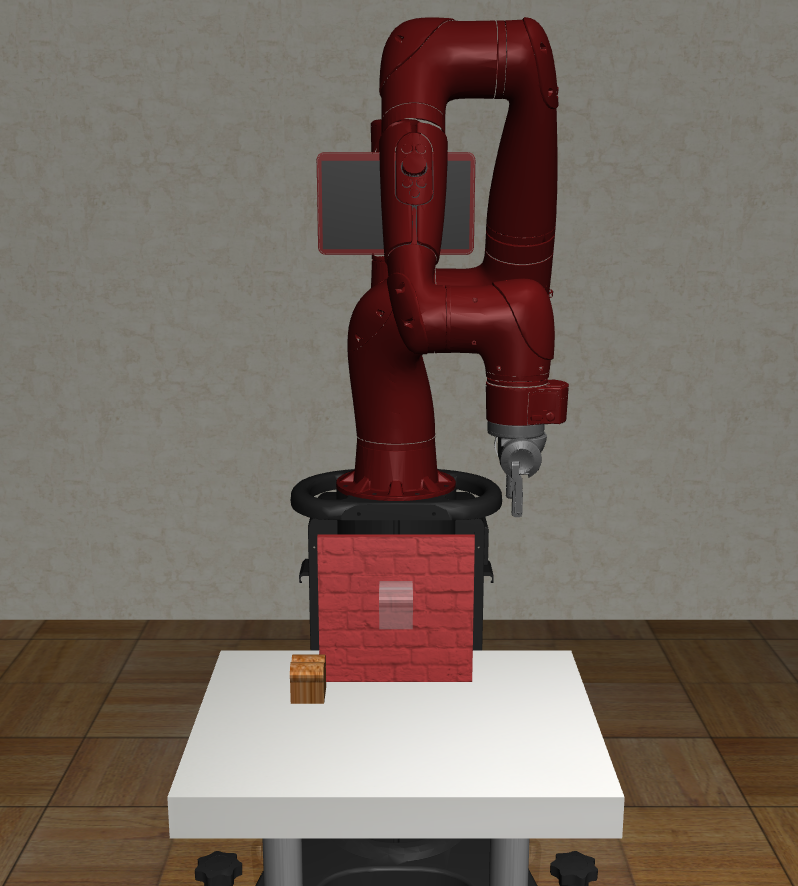}
        \vspace{1em}
    \end{subfigure}\\
    \begin{subfigure}[b]{0.4\textwidth}  
        \centering
        \includegraphics[height=5.5cm]{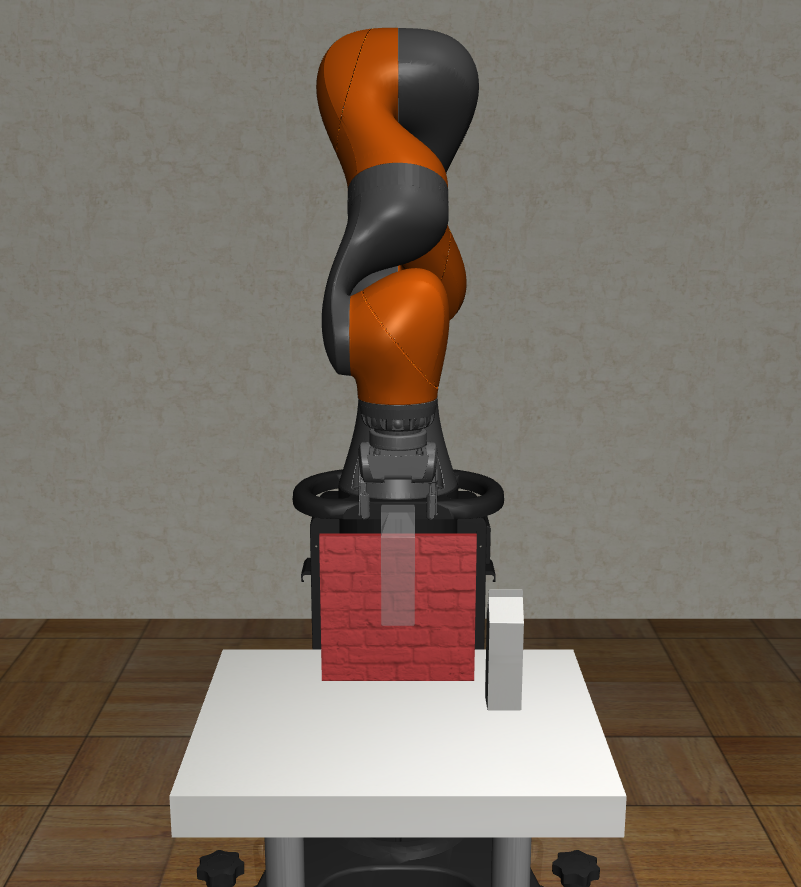}
    \end{subfigure}%
    \begin{subfigure}[b]{0.4\textwidth}  
        \centering
        \includegraphics[height=5.5cm]{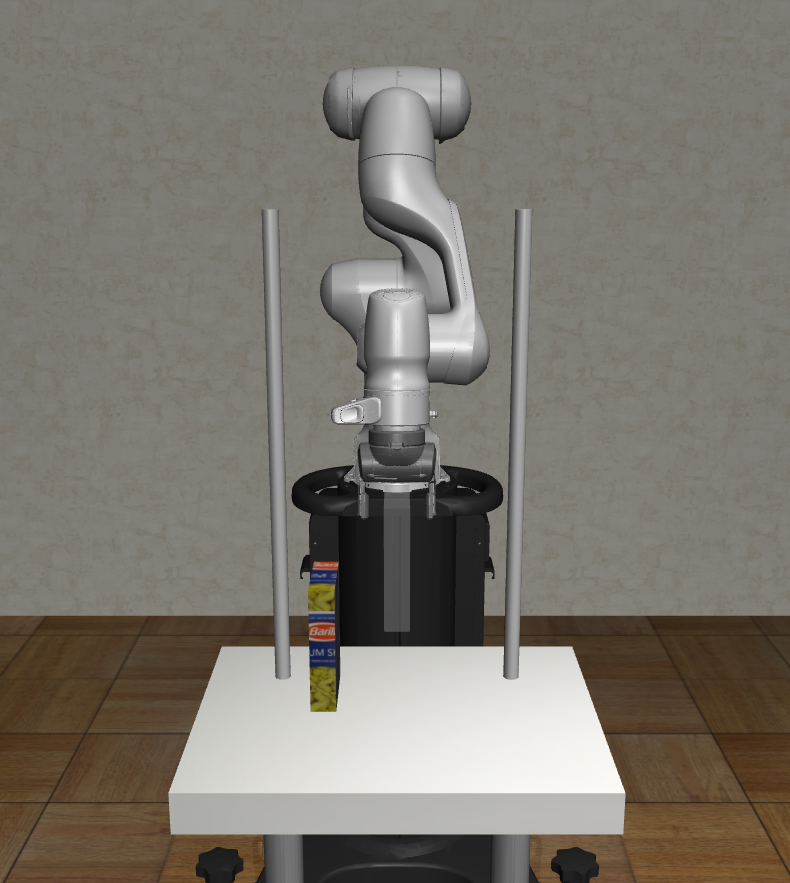}
    \end{subfigure}%
    \caption[Visualization of explicitly compositional robotic RL tasks.]{Visualization of various instantiations of the compositional robotic tasks.}
    \label{fig:RobosuiteTasks}
 \end{figure}
Each task is one of the $\numTasks=48$ combinations of the above elements, just like in the 2-D case. Intuitively, if the agent has learned to manipulate the milk carton with the IIWA arm and the cereal box with the Panda arm, then it could recombine knowledge to manipulate the milk carton with the Panda arm.

Figure~\ref{fig:RobosuiteTasks} shows example tasks created by sampling one component of each type. Chapter~\ref{cha:Benchmark} extends this evaluation domain into a large-scale benchmark of hundreds of compositionally related tasks.
The following paragraphs provide details of the underlying MDP of tasks within this domain, and how it varies according to the task components.

\paragraph{Observation space} Each time step, the agent receives a rich observation that describes all elements in the task. The robot arm state comprises a $32$-dimensional vector, concatenating the sine and cosine of the joint positions, the joint velocities, the end-effector position, the end-effector orientation in unit quaternions, and the gripper fingers' positions and velocities. The target object's state is a $14$-dimensional vector that concatenates the position and orientation of the object in global coordinates, and the position and orientation of the object relative to the end-effector. The observation similarly describes the obstacle with its position and orientation in global and end-effector coordinates. The observation also describes the goal (i.e., the target height) by its position and orientation in both coordinate frames, as well as the relative position of the object with respect to the goal. Note that many of these elements are redundant and in principle unnecessary for solving the task at hand. However, the initial evaluations found that this combination of observations leads to tasks that are much more easily learned by the STL agent. 

\paragraph{Action space} The agent's actions are continuous-valued, eight-dimensional vectors, indicating the change in each of the seven joint positions and the gripper. For reference, this corresponds to the \texttt{JOINT\_POSITION} controller in \texttt{robosuite}.

\paragraph{Reward function} The environment provides the agent with dense rewards. At a high level, the agent receives an increasingly large reward for approaching, grasping, and lifting the object. In tasks with the \texttt{wall} obstacle, the reward additionally encourages the agent to lift the object past the \texttt{wall}. Concretely, in tasks with no \texttt{wall}, the reward is given by:
\begin{equation}
    \Rewards = \begin{cases}
         0.1 (1 - \tanh(5 d)) \hspace{2em}& \text{if not grasping}\\
         0.25 + 0.5 (1 - \tanh(25 h)) \hspace{2em}& \text{if grasping and not success}\\
         1 \hspace{2em}& \text{if success} \enspace,
    \end{cases}
\end{equation}
and in tasks with \texttt{wall}, the reward is given by:
\begin{equation}
    \Rewards = \begin{cases}
         0.05 (1 - \tanh(5 d_w)) \hspace{2em}& \text{if not past wall}\\
         0.05 + 0.05 (1 - \tanh(5 d)) \hspace{2em}& \text{if past wall and not grasping}\\
         0.25 + 0.5 (1 - \tanh(25 h)) \hspace{2em}& \text{if grasping and not success}\\
         1 \hspace{2em}& \text{if success} \enspace,
    \end{cases}
\end{equation}
where $d$ is the distance from the gripper to the center of the object, $h$ is the vertical distance between the object and the target height truncated at $0$, and $d_w$ is the $x,z$ distance from the gripper to the wall.

\paragraph{Initial conditions} The robot arm starts at the center of the left edge of a flat table of width $w=0.39$ and depth $d=0.49$. The environment places the obstacle one quarter of the way from left to right in the table, and at the center. The environment further samples the object location uniformly at random in the right half of the table, so that the robot must always surpass the obstacle before reaching the object. 

\paragraph{Episode termination} The episode may only terminate after reaching the task's horizon of $H=500$ time steps.

\subsection{Model Architectures}
\label{sec:ModelArchitecturesRL}

This section describes the model architectures used to represent each learnable function. In the MuJoCo and Meta-World domains, the architectures were simple linear or neural net models constructed by linearly combining components from $\bL$ and trained via \lpgftw{}. On the other hand, the learners constructed the architectures for the 2-D world and robotics domain following the modular structure described in Section~\ref{sec:ModularArchitectureRL}.

\subsubsection{MuJoCo} 
For these simple experiments, the policy was a Gaussian with mean linear in the observations and a trainable standard deviation vector per task. \lpgftw{} used $k=5$ components in $\bL$ for all domains except  Walker-2D \textit{body-parts} domain, which in practice required a higher capacity of $k=10$. Additionally, the baseline model leveraged by NPG to compute the advantage function used a task-specific multilayer perceptron (MLP) with two hidden layers of $128$ units and ReLU activation.

\subsubsection{Meta-World} 
The architecture for these more complex tasks was a Gaussian policy parameterized by a neural net with two hidden layers of $32$ units and $\tanh$ activation. \lpgftw{} constructed these two layers by linearly combining components stored in $k=3$ components in $\bL$. Given the high diversity of the tasks considered in this evaluation, the agent used task-specific output layers, in order to specialize policies to each individual task. The standard deviation and baseline for NPG were identical to those used in MuJoCo domains.

\subsubsection{2-D Domains} 
The architecture for the discrete 2-D tasks assumes that the underlying graph structure first processes the static object, then the target object, and finally the agent dynamics. Intuitively, the agent's actions require all information about target and static objects, and the agent module should be closest to the output to actuate the agent. 
The plan for reaching a target object requires information about how to interact with the static object. Interacting with the static object instead could be done without information about the target object. 

Static object modules consume as input the five channels corresponding to static objects, and pass those through one convolutional block of $c=8$ channels, kernel size $k=2$, ReLU activation, and max pooling of kernel size $k=2$, and another block of $c=16$ channels, kernel size $k=2$, and ReLU activation. Subsequent modules, which require incorporating the outputs of the previous modules, include a preprocessing network that transforms the module-specific state into a representation compatible with the previous module's output, and then concatenate the two representations. The concatenated representation passes through a post-processing network to compute a representation of the overall state up to that point. Target object modules preprocess the target object channel with a block with the same architecture as static object modules, concatenate the preprocessed state with the static object module's output, and pass this through a single convolutional block of $c=32$ channels, kernel size $k=2$, and ReLU activation. Similarly, agent modules pass the agent channel through a preprocessing net with the same architecture as the target object module (minus the concatenation with the static object module output), concatenate the preprocessed state with the output of the target object module, and pass this through separate MLPs for the actor and the critic with a single hidden layer of $n=64$ hidden units and $\tanh$ activation. Since the domain contains four objects of each type, the model also contains four modules of each type. The learner then constructs a separate architecture for each task by combining one module of each type. Figure~\ref{fig:MiniGridArchitecture} shows the resulting architecture.
\begin{figure}[t!]
    \centering
    \begin{subfigure}[b]{\textwidth}
        \centering
        \includegraphics[height=9cm]{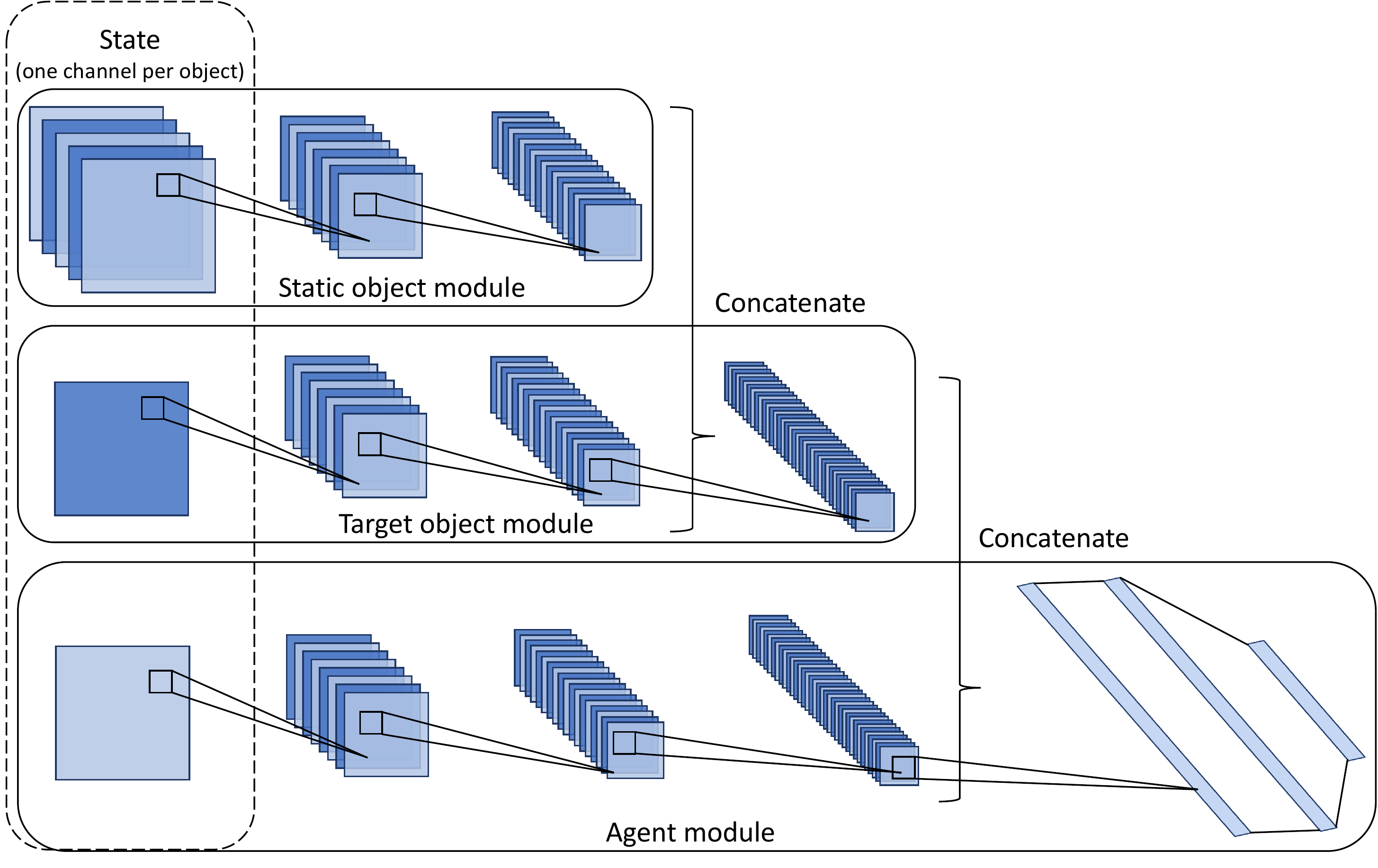}
    \end{subfigure}%
    \caption[Modular architecture for explicitly compositional discrete 2-D RL tasks.]{Modular architecture for discrete 2-D tasks. The input factors into elements corresponding to each task component and passed only to the corresponding type of module. The output of each module feeds into the next as additional input.}
    \label{fig:MiniGridArchitecture}
\end{figure}

\subsubsection{Robotics} 
After validating empirically that the architecture from the discrete 2-D domain correctly captured the underlying structure of the tasks, the graph structure for the robotics domain followed that of the discrete 2-D setting. In particular, the model processes the  obstacle first, the object next, and the robot last. The obstacle module passes the obstacle state through a single hidden layer of $n=32$ hidden units with $\tanh$ activation. The object module preprocesses the object state with another $\tanh$ layer of $n=32$ nodes, concatenates the preprocessed state with the output of the obstacle module, and passes this through another $\tanh$ layer with $n=32$ units. Finally, the robot module takes the robot and goal states as input, processes those with two hidden $\tanh$ layers of $n=64$ hidden units each, concatenates this with the output of the object module, and passes the output through a linear output layer. Following standard practice for continuous control with PPO, the agent uses completely separate networks for the actor and the critic (instead of sharing the early layers). However, the graph structure is the same for both networks. Moreover, for the critic, the robot module takes the action as an input along with the robot and goal states. Figure~\ref{fig:RobosuiteArchitecture} shows the resulting architecture.

\begin{figure}[t!]
    \centering
    \begin{subfigure}[b]{\textwidth}
        \centering
        \includegraphics[height=5cm]{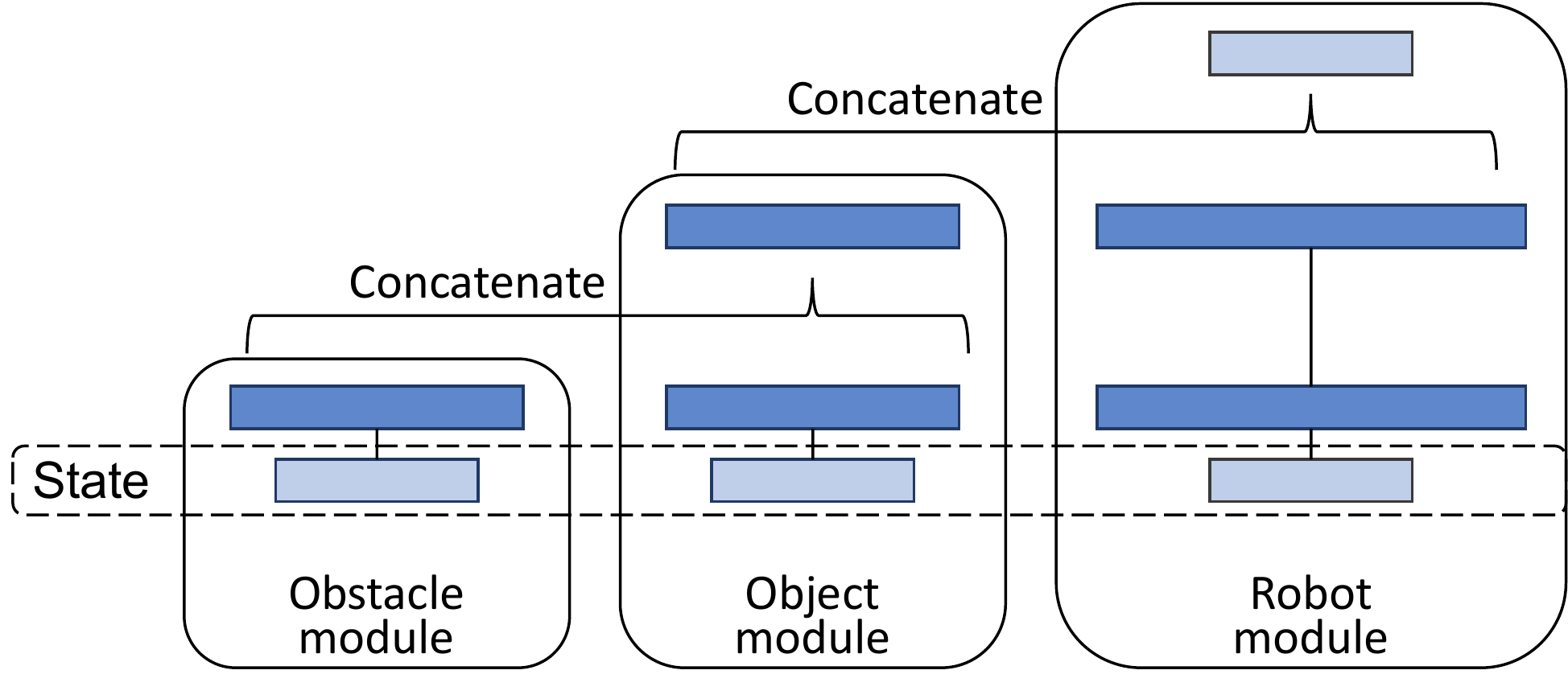}
    \end{subfigure}%
    \caption[Modular architecture for explicitly compositional robot tasks.]{Modular architecture for robot tasks. Similar to the 2-D case, the input is decomposed into module-specific components, and the output of each module is used as input to the next.}
    \label{fig:RobosuiteArchitecture}
\end{figure}

\subsection{Baselines} 
\label{sec:BaselinesRL}

The evaluation used the following baselines to validate performance. Each experiment considered only a subset of them, to ensure compatibility and usefulness of the comparisons. 

\begin{itemize}
    \item {\em STL} is the default baseline that does not transfer knowledge across tasks and serves to study whether agents are achieving forward transfer: the ability to improve the learning on one task by leveraging knowledge from earlier tasks. All experiments compared against STL, which trains a separate model for each task via NPG (in MuJoCo and Meta-World domains) or PPO (in compositional discrete 2-D and robotics domains). Each such model follows the architecture described above, but uses a single module of each type. In the first two domains, this corresponds to the standard STL architecture. However, in compositional domains this architecture is nonstandard, as it does not process the input observation entirely in the first layer of the network. This choice ensured a fair comparison, as it yielded substantially better results in initial evaluations. 
    \item {\em EWC}, described in Chapter~\ref{cha:Supervised}, failed to work in any of the initial evaluations, and so the experiment on the compositional robotics domain omitted it. For fairness, in noncompositional domains EWC used the full Hessian instead of diagonal or Kronecker approximations, since \lpgftw{} also used the full Hessian. On the other hand, for simplicity, online EWC was used in compositional domains~\citep{schwarz2018progress}. The architecture of EWC was the same as of STL, with additional multi-hot indicators of the ground-truth components in compositional domains. To validate that performance differences were not due to the higher capacity of compositional methods, a high-capacity variant of EWC (denoted EWC\_h) was used in Meta-World experiments. 
    \item {\em PG-ELLA}, described in Section~\ref{sec:ConnectionsToPGELLA}, bears close connections to \lpgftw{}, and therefore experiments in noncompositional settings compared these two methods. PG-ELLA used exactly the same architecture as \lpgftw{}.
    \item {\em CLEAR} is a replay-based method that leverages previous tasks' trajectories~\citep{rolnick2019experience}. However, it only applies to the discrete-action setting, and therefore only the discrete 2-D evaluations used it. In particular, in order to prevent forgetting, for every sample collected online and used to compute the assimilation loss, the agent replays $\eta$ samples from previous tasks to compute the custom CLEAR loss, which balances an importance sampling PG objective ($V$-trace) and an imitation objective. In the experiments, the agent evenly split replay samples across all previously seen tasks, and the architecture was the same as used by EWC.
    \item {\em ER} is a simple replay baseline, used in place of CLEAR for continuous-action, noncompositional domains. ER uses importance sampling over previous tasks' data to encourage knowledge retention. It was tested in Meta-World domains and failed to yield good performance, so it was not considered for remaining evaluations.
    \item \textit{P\&C} is similar to the proposed methods, separating the training into a forward transfer stage, where the agent keeps shared parameters fixed, and a consolidation stage, where the agent pushes new knowledge into the shared parameters~\citep{schwarz2018progress}. Due to the difficulty of the implementation (and the lack of open-source code),  only explicitly compositional evaluations considered P\&C. Unlike \compRL{}, P\&C uses a monolithic network, making it much harder to find a solution that works across all tasks. In the experiments, it used a similar architecture to that of EWC for its knowledge base (KB) and active columns, with additional lateral connections from the KB to the active column. In the original P\&C algorithm, the {\em progress} phase trains the active column online and the {\em compress} phase trains the KB column online as well, but imposing an EWC penalty on the shared parameters. The experiments in this chapter used a slightly modified version closer to \compRL{}: the compress phase instead used replay data from the current task to distill knowledge from the active column into the KB. By doing this only over the latest portion of the data, this closely matches the P\&C formulation that generates data online with the distribution imposed by the (fixed) active policy. This also enables P\&C to leverage the entirety of the online interactions for the progress phase, instead of trading off which portion to use for progressing and which portion to use for compressing. 
\end{itemize}

\subsection{Hyperparameters}
\label{sec:HyperParametersRL}

This section describes how the evaluation chose hyperparameters for each algorithm. At a high level, some search method selected the STL hyperparameters, and then all lifelong learners, which wrap around STL, used these same hyperparameters. This reduced the search space over hyperparameters considerably while ensuring that the evaluation was fair. 

\paragraph{MuJoCo} A manual hyperparameter tuning selected the hyperparameters for NPG by running an evaluation on the nominal task for each domain (without gravity or body part modifications). The agent trained with various combinations of the number of iterations, number of trajectories per iteration, and step size, until the search reached a learning curve that was fast and reached proficiency. All lifelong learning algorithms used the obtained hyperparameters. \lpgftw{} used typical hyperparameters (regularization parameters \mbox{$\mu=\lambda=10^{-5}$}, number of columns of $\bL$ $k=5$) and held them fixed through all experiments, forgoing potential additional benefits from a hyperparameter search. The only exception was the number of latent components used for the Walker-2D \textit{body-parts} domain, as it was found empirically that $k=5$ led to saturation of the learning process early on. \lpgftw{}  used the simplest setting for the update schedule of $\bL$, $\mathtt{adaptationFrequency}=\mathtt{structureUpdates}$, which corresponds to only updating the components in $\bL$ once assimilation had completed. PG-ELLA used the same hyperparameters as used for \lpgftw{}, since they are used in exactly the same way for both methods. Finally, for EWC, the evaluation ran a grid search over the value of the regularization term, $\lambda$, among $\{1e\!-\!7,1e\!-\!6, 1e\!-\!5, 1e\!-\!4, 1e\!-\!3\}$. For this, the agent trained on five consecutive tasks for $50$ iterations over five trials with different random seeds. Each domain chose the value of $\lambda$ independently to maximize the average performance after training on all tasks. Upon discovering that EWC was struggling in some of the domains, the evaluation considered various versions of EWC, as described in Appendix~\ref{app:EWCAdditionalResultsMuJoCo}, modifying the regularization term and selecting whether to share the policy's variance across tasks. The only version that worked in all domains was the original EWC penalty with a shared variance across tasks, so the remainder of the evaluation used only that version. To make comparisons fair, EWC used the full Hessian instead of the diagonal Hessian proposed by the authors. Table~\ref{tab:hyperparametersNonCompositional} summarizes the obtained hyperparameters.

\begin{table}[b!]
    \centering
    \caption[Summary of hyperparameters of \lpgftw{} and baselines for lifelong compositional RL.]{Summary of optimized hyperparameters used by \lpgftw{} and the baselines.}
    \label{tab:hyperparametersNonCompositional}
    \addtolength{\tabcolsep}{-0.05em}
    \begin{tabular}{@{}l|c|c|c|c|c|c|c|c@{}}
    & Hyperparam. & HC-G & HC-BP & Ho-G & Ho-BP & W-G & W-B & MT10/50\\
    \hline\hline
    \multirow{5}{*}{NPG} & \# iterations & $50$ & $50$ & $100$ & $100$ & $200$ & $200$ & $200$\\
    & \# traj.~/ iter. & $10$ & $10$ & $50$ & $50$ & $50$ & $50$ & $50$\\
    & step size & $0.5$ & $0.5$ & $0.005$ & $0.005$ & $0.05$ & $0.05$ & $0.005$\\
    & $\lambda$ (GAE) & $0.97$ & $0.97$ & $0.97$ & $0.97$ & $0.97$ & $0.97$ &$0.97$\\
    & $\gamma$ (MDP) & $0.995$ & $0.995$ & $0.995$ & $0.995$ & $0.995$ & $0.995$ & $0.995$\\
    \hline
    \multirow{3}{*}{\lpgftw{}} & $\lambda$ & $1e\!-\!5$ & $1e\!-\!5$ & $1e\!-\!5$ & $1e\!-\!5$ & $1e\!-\!5$ & $1e\!-\!5$ & $1e\!-\!5$\\
    & $\mu$ & $1e\!-\!5$ & $1e\!-\!5$ & $1e\!-\!5$ & $1e\!-\!5$ & $1e\!-\!5$ & $1e\!-\!5$ & $1e\!-\!5$\\
    & $k$ & $5$ & $5$ & $5$ & $5$ & $5$& $10$ & $3$\\
    \hline
    \multirow{3}{*}{PG-ELLA} & $\lambda$ & $1e\!-\!5 $& $1e\!-\!5$ & $1e\!-\!5$ & $1e\!-\!5$ & $1e\!-\!5$ & $1e\!-\!5$ & $1e\!-\!5$\\
    & $\mu$ & $1e\!-\!5$ & $1e\!-\!5$ & $1e\!-\!5$ & $1e\!-\!5$ & $1e\!-\!5$ & $1e\!-\!5$ & $1e\!-\!5$\\
    & $k$ & $5$ & $5$ & $5$ & $5$ & $5$& $10$ & $3$\\
    \hline
    EWC & $\lambda$ & $1e\!-\!6$ & $1e\!-\!6$ & $1e\!-\!7$ & $1e\!-\!4$ & $1e\!-\!7$ & $1e\!-\!7$ & $1e\!-\!7$
    \end{tabular}
\end{table}

\paragraph{Meta-World} In this case, hyperparameter tuning manually chose the parameters for NPG on the {\tt reach} task, which is the easiest task to solve in the benchmark. Once again, all lifelong learners used these fixed hyperparameters. \lpgftw{} and PG-ELLA used typical values for $k$, $\lambda$, and $\mu$. In particular, they used fewer latent components than in the previous evaluation ($k=3$), since MT10 contains only $\numTasks=10$ tasks and using more than three policy factors would give \lpgftw{} an unfair advantage over single-model methods. The evaluation also tested a version of EWC with a higher capacity (EWC\_h), with $50$ hidden units for MT10 and $40$ for MT50, to ensure that it had access to approximately the same number of parameters as \lpgftw{}. For EWC and EWC\_h, the evaluation ran a grid search for $\lambda$ in the same way as for MuJoCo experiments. ER used a fixed experience replay ratio of $50$-$50$, as suggested by \citet{rolnick2019experience}, and each mini-batch sampled from the replay buffer contained the same number of trajectories from each previous task. \lpgftw{}, PG-ELLA, and EWC all had access to the full Hessian, and EWC did not to share the variance across tasks since the outputs of the policies were task-specific. Table~\ref{tab:hyperparametersNonCompositional} summarizes the obtained hyperparameters.

\begin{table}[t!]
    \centering
    \caption[Summary of hyperparameters used for evaluations on explicitly compositional RL tasks.]{Summary of optimized hyperparameters used by \compRL{} and the baselines.}
    \label{tab:hyperparametersCompositional}
    \begin{tabular}{@{}l|l|c|c@{}}
    Algorithm & Hyperparameter & Discrete 2-D world & Robotic manipulation\\
    \hline\hline
    \multirow{10}{*}{PPO} & \# env. steps & $1M$ & $3.6M$ \\
    & \# env. steps / update & 4,096 & $8{,}000$ \\
    & learning rate & $1e\!\!-\!\!3$ & $1e\!\!-\!\!3$\\
    & mini-batch size & $256$ & $8{,}000$\\
    & epochs per update & $4$ & $80$\\
    & $\lambda$ (GAE) & $0.95$ & $0.97$\\
    & $\gamma$ (MDP) & $0.99$ & $0.995$\\
    & entropy coefficient & $0.5$ & ---\\
    & Gaussian policy variance & --- & fixed\\
    & \# parameters & $1{,}080{,}320$ & $1{,}040{,}304$\\    
    \hline
    \multirow{5}{*}{Compositional} & \# modules per depth & $4$ & $4$\\
    & \# rollouts / module comb. & $10$ & $10$\\
    & \# replay samples / task & $100{,}000$ & $100{,}000$\\
    & \# BCQ epochs & $10$ & $100$\\
    & \# parameters & $86{,}350$ & $165{,}970$\\
    \hline
    \multirow{5}{*}{P\&C} & $\lambda$ (EWC) & $10$ & $10$\\
    & $\gamma$ (EWC) & $1$ &  $1$\\
    & \# replay samples / task & $100{,}000$ & $100{,}000$\\
    & \# distillation epochs & $10$ & $100$\\
    & \# parameters & $70{,}282$ & $104{,}384$\\
    \hline
    \multirow{3}{*}{EWC} & $\lambda$ (EWC) & $10{,}000$ & ---\\
    & $\gamma$ (EWC) & $1$ &  ---\\
    & \# parameters & 18,928 & ---\\
    \hline
    \multirow{2}{*}{CLEAR} & $\eta$ & $1.5$ & ---\\
    & \# parameters & $18{,}928$ & ---
    \end{tabular}
\end{table}

\paragraph{Discrete 2-D} Hyperparameter tuning on STL executed a grid search over the learning rate (from \{$1e\!\!-\!\!6, 3e\!\!-\!\!6, 1e\!\!-\!\!5, 3e\!\!-\!\!5, 1e\!\!-\!\!4, 3e\!\!-\!\!4, 1e\!\!-\!\!3, 3e\!\!-\!\!3, 1e\!\!-\!\!2, 3e\!\!-\!\!2$\}) and the number of environment interactions per training step (from \{$256$, $512$, $1024$, $2048$, $4096$, $8192$\}). Since the static object affects the difficulty of the task, the search considered one task with each static object (\texttt{wall}, \texttt{floor}, \texttt{food}, and \texttt{lava}) for each hyperparameter combination, and optimized for the average performance across the four objects. All lifelong agents reused the obtained PPO hyperparameters. For lifelong agents, the evaluation tuned their main hyperparameter by training on five tasks over five random seeds. For P\&C and EWC, the search was over the regularization $\lambda$ in \{$1e\!-\!3, 1e\!-\!2, 1e\!-\!1, 1e0,1e1, 1e2, 1e3, 1e4$\}. For CLEAR, the search was over the replay ratio $\eta$ in \{$0.25, 0.5, 0.75, 1, 1.25, 1.5, 1.75, 2$\}. Table~\ref{tab:hyperparametersCompositional} summarizes the obtained hyperparameters.

\paragraph{Robotics} Instead of hyperparameter tuning, the tasks were created ensuring that STL performed well. Given the fixed tasks, the evaluation perturbed key PPO hyperparameters (e.g., the learning rate and the entropy coefficient) and verified that the changes led to decreased performance. In particular, this setting used a modified version of  \texttt{Spinning Up}'s~\citep{SpinningUp2018} PPO implementation to encourage improved exploration, because initial experiments with the original implementation suffered from premature convergence.

Each agent used an MLP to represent the mean of a Gaussian policy. Against popular wisdom, which encourages using linear activations in the final layer, initial analyses yielded that adding a $\tanh$ activation led to substantially improved exploration. The rationale is that simulators typically cap the magnitude of robot actions to emulate physical limits in real robots. Therefore, if the MLP outputs high-magnitude means for the Gaussian distribution, the sampled actions are all likely to reach the range limits, regardless of the variance of the Gaussian. In consequence, the agent could ``cheat'' existing techniques to avoid premature convergence (e.g., entropy regularization) by learning a high variance but being deterministic in practice by saturating the actions. The $\tanh$ activation ensures that the actions are never too large in magnitude, which permits the sampling to induce stochasticity (and, consequently, exploration). 

The second modification was to use a constant variance for the Gaussian policy, instead of propagating gradients through it. The reason this was necessary is that, with learnable variance, the agent was finding pathological regions of the optimization landscape that (once more) cheated existing entropy regularization approaches. Concretely, the agent was inflating variance along dimensions where actions were inconsequential (e.g., joints that rotate in directions orthogonal to the motion of the gripper), and reducing variance to a minimum along critical dimensions. The resulting policy was therefore deterministic along all interesting dimensions, and so the exploration the agent was engaging in was ineffective. Setting a fixed variance of $\sigma^2=1$ ($\log(\sigma) = 0$) for the seven joint actions and $\sigma^2= 1/e$ ($\log(\sigma)=-0.5$) for the gripper action ensured that the robot consistently explored throughout the learning process and was critical toward enabling learning the compositional robotics tasks. 

The PPO hyperparameters were fixed for lifelong training. The one modification that was required for lifelong compositional training was to downscale the output layers of the policy and critic networks by a factor of $0.01$ whenever the initial policy achieved no success; this ensured that, if the policy was not close to solving the task, the agent would not be following a highly (and incorrectly) specialized policy, while still leveraging the compositional representations at lower layers. The regularization coefficient $\lambda$ for P\&C was tuned by training on five tasks over three random seeds, varying $\lambda$ in \{$1e\!-\!3, 1e\!-\!2, 1e\!-\!1, 1e0,1e1, 1e2, 1e3, 1e4$\}. Table~\ref{tab:hyperparametersCompositional} summarizes the hyperparameters used for these robotic manipulation experiments.

\subsection{Empirical Evaluation on OpenAI Gym MuJoCo Domains}
\label{sec:ExperimentalResultsSimple}

\begin{figure}[b!]
\captionsetup[subfigure]{belowskip=1em}
    \begin{subfigure}[b]{\textwidth}
        \centering
        \includegraphics[height=0.45cm, trim={0.1cm 0.1cm 17cm 0.1cm}, clip]{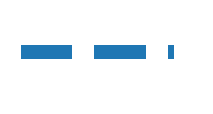}
        \hspace{-0.1em}\raisebox{0.13cm}{\fontfamily{DejaVuSans-TLF}\selectfont\fontsize{8.7pt}{9pt}\selectfont LPG-FTW}\hspace{1.em}
        \includegraphics[height=0.45cm, trim={5.5cm 0.1cm 0.1cm 0.1cm}, clip]{chapter4/Figures/LPG-FTW/learning_curves/learning_curve_legend_cameraready.pdf}
    \end{subfigure}\\
    \begin{subfigure}[b]{0.53\textwidth}
        \includegraphics[height=3.3cm, trim={0.2cm 5.8cm 0.9cm 2.1cm}, clip]{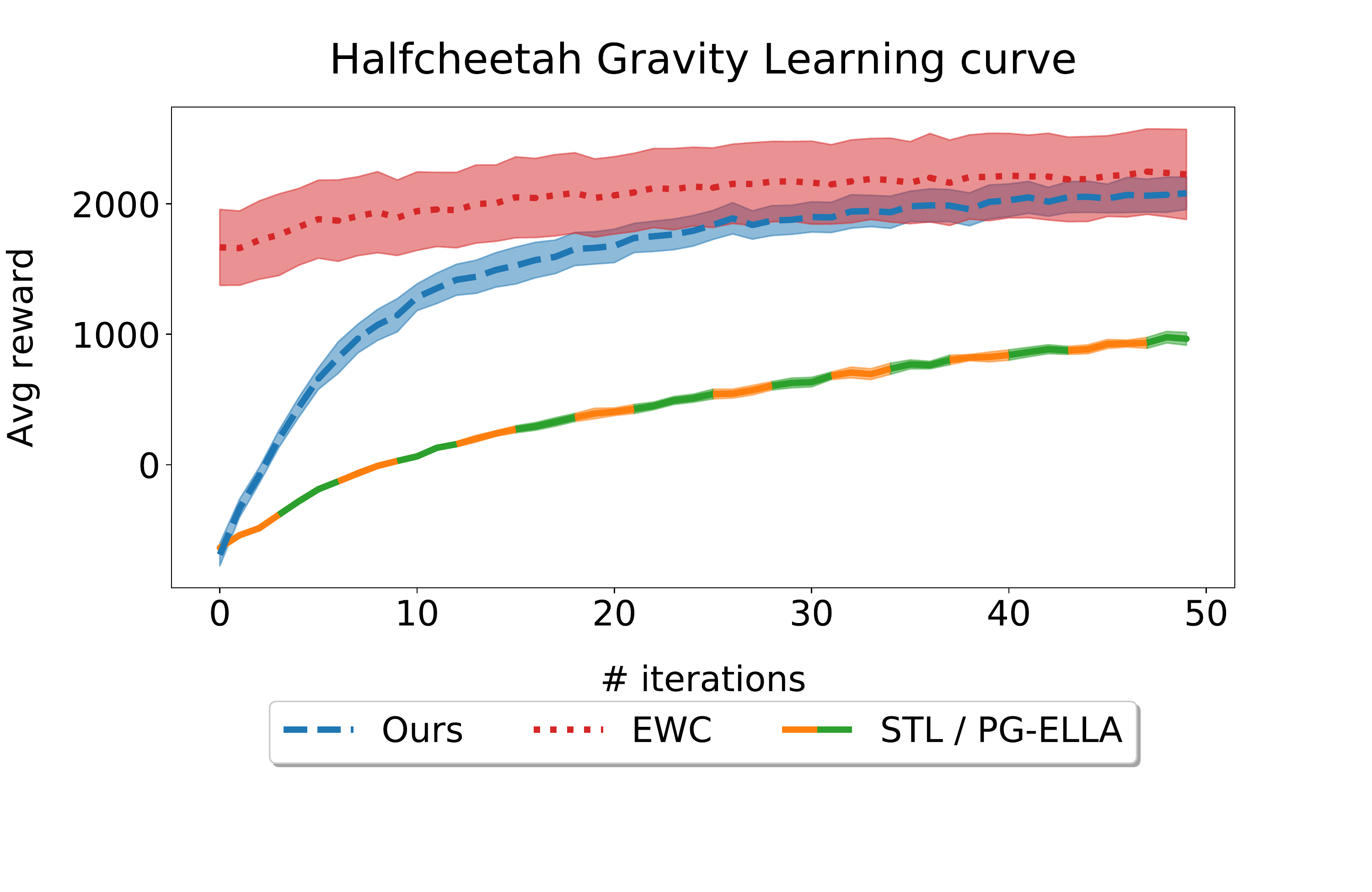}
        \caption{HalfCheetah \textit{gravity}}
    \end{subfigure}%
    \begin{subfigure}[b]{0.47\textwidth}
        \includegraphics[height=3.3cm, trim={1.6cm 5.8cm 0.9cm 2.1cm}, clip]{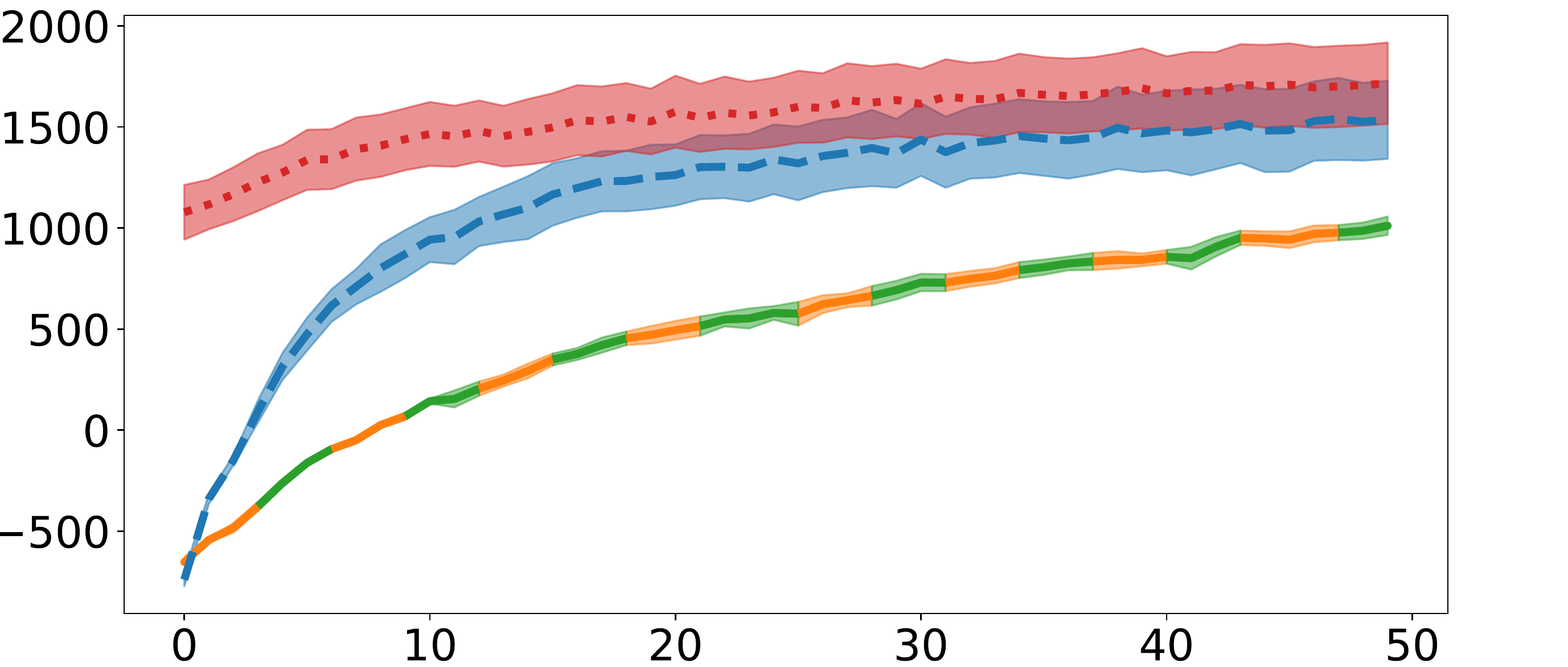}
        \caption{HalfCheetah \textit{body-parts}}
    \end{subfigure}\\
    \begin{subfigure}[b]{0.53\textwidth}
        \includegraphics[height=3.3cm, trim={0.2cm 5.8cm 0.9cm 2.1cm}, clip]{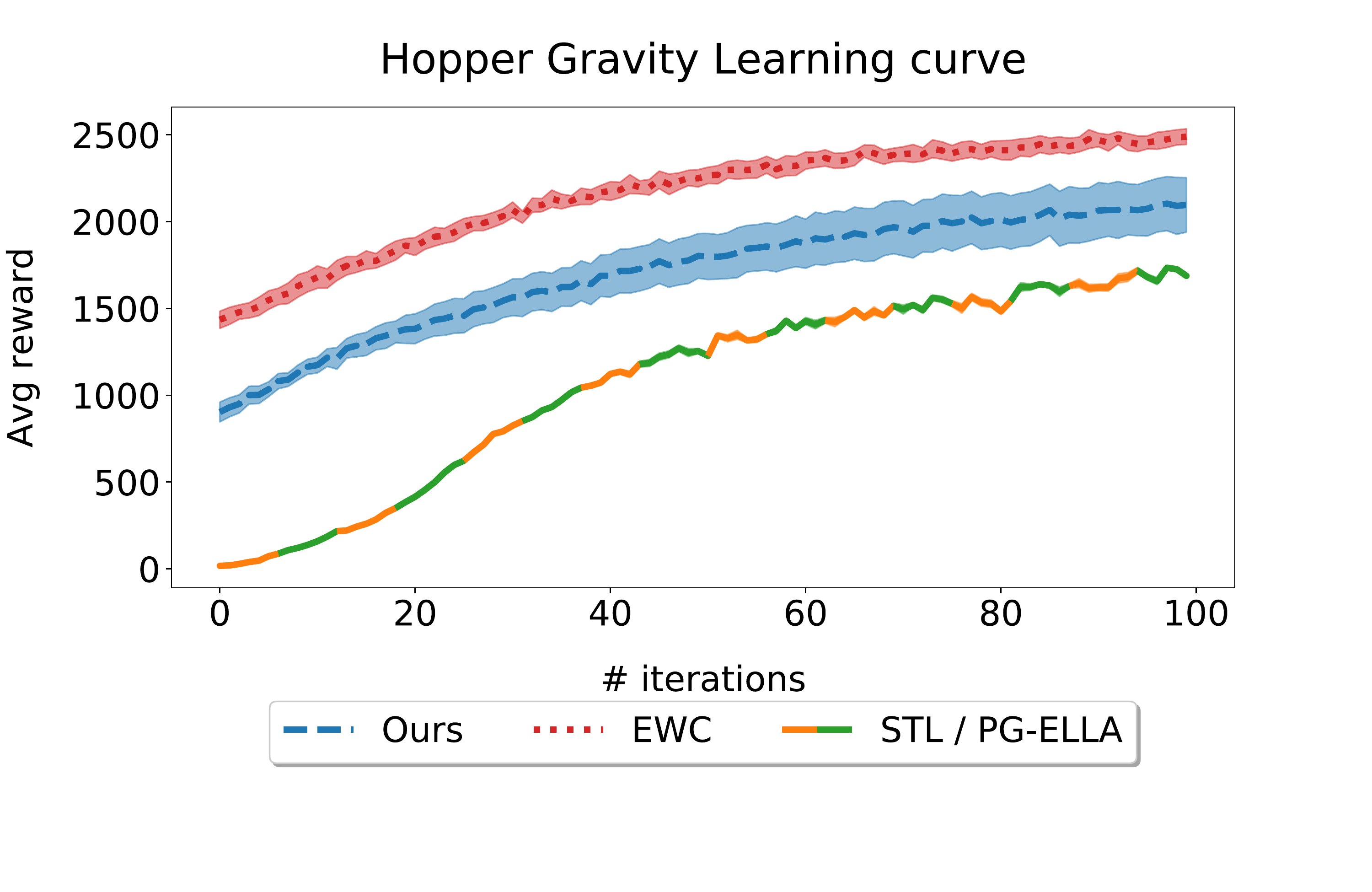}
         \caption{Hopper \textit{gravity}}
    \end{subfigure}%
    \begin{subfigure}[b]{0.47\textwidth}
        \includegraphics[height=3.3cm, trim={1.6cm 5.8cm 0.9cm 2.1cm}, clip]{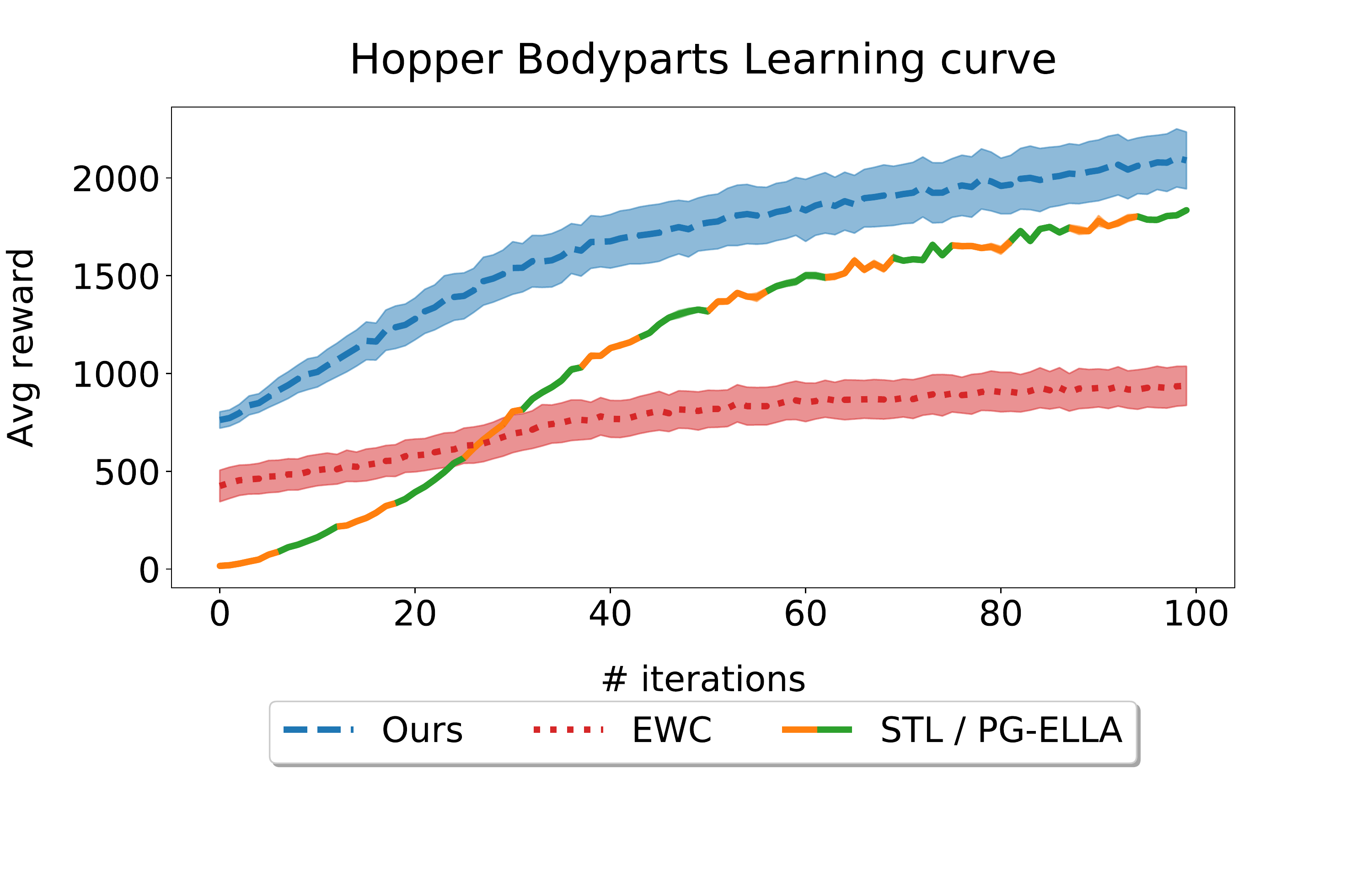}
         \caption{Hopper \textit{body-parts}}
    \end{subfigure}\\
    \begin{subfigure}[b]{0.53\textwidth}
        \includegraphics[height=3.72cm, trim={0.2cm 4.3cm 0.9cm 2.1cm}, clip]{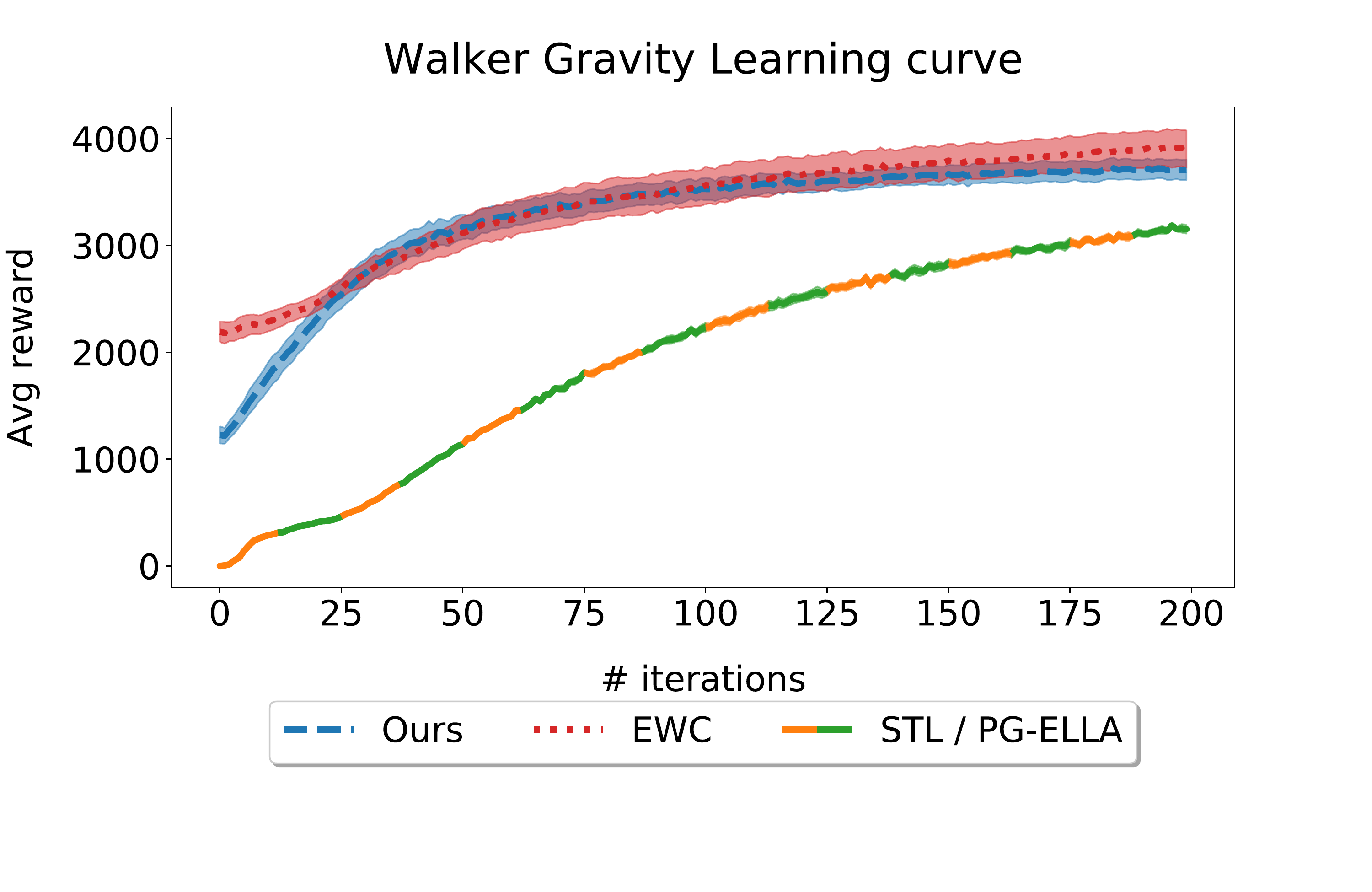}
         \caption{Walker-2D \textit{gravity}}
         \vspace{-1em}
    \end{subfigure}%
    \begin{subfigure}[b]{0.47\textwidth}
        \includegraphics[height=3.72cm, trim={1.6cm 4.3cm 0.9cm 2.1cm}, clip]{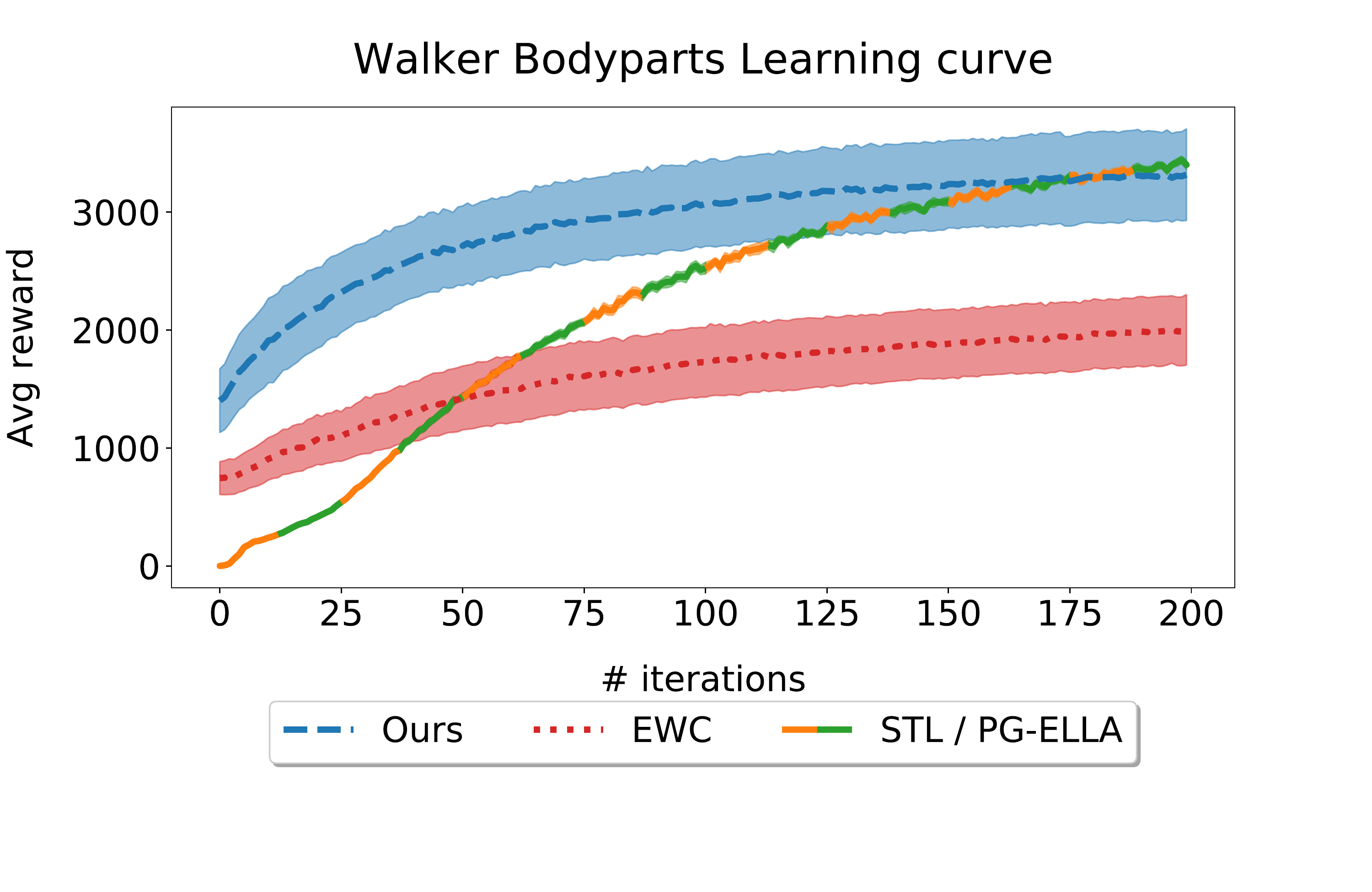}
         \caption{Walker-2D \textit{body-parts}}
         \vspace{-1em}
    \end{subfigure}
    \caption[Learning curves of \lpgftw{} in lifelong compositional RL on MuJoCo domains with linear policies.]{Average performance during training across all tasks for six MuJoCo domains. \lpgftw{} is consistently faster than STL and PG-ELLA (which by definition learn at the same pace) in achieving proficiency, and achieves better final performance in five domains and equivalent performance in the remaining one. EWC is faster and converges to higher performance than \lpgftw{} in some domains, but completely fails to learn in others. Shaded error bars denote standard errors across five random seeds.}
    \label{fig:learningCurves}
\end{figure}

The first evaluation presented here tested \lpgftw{} on the MuJoCo environments from OpenAI Gym, varying the gravity or body-part sizes as described in Section~\ref{sec:EvaluationDomainsRL}, using linear policies. The evaluation repeated all experiments in this section over five trials, varying the random seed controlling the parameter initialization and the ordering over tasks.

Figure~\ref{fig:learningCurves} shows the average performance over all tasks as a function of the NPG training iterations.  Even though the agents trained on tasks sequentially, the averaged curves serve to study the ability to accelerate learning: any curve above STL indicates that the corresponding method achieved forward transfer. \lpgftw{} consistently learned faster than STL, and obtained higher final performance on five out of the six domains. Learning the task-specific coefficients $\st$ directly via policy search increased the learning speed of \lpgftw{}, whereas PG-ELLA was limited to the learning speed of STL, as indicated by the shared learning curves. EWC was faster than \lpgftw{} in reaching high-performing policies in four domains, primarily due to the fact that EWC uses a single shared policy across all tasks, which enables it to have starting policies with high performance. However, EWC failed to even match the STL performance in two of the domains. One hypothesis that might explain this failure is that the tasks in those two domains are highly varied (particularly in the \textit{body-parts} domains, since it has four different axes of variation), and the single shared policy was unable to perform well on all tasks in those domains. Appendix~\ref{app:EWCAdditionalResultsMuJoCo} shows an evaluation with various versions of EWC attempting (unsuccessfully) to alleviate these issues.

\begin{figure}[t!]
\captionsetup[subfigure]{belowskip=1em}
    \begin{subfigure}[b]{\textwidth}
        \centering
        \includegraphics[height=0.45cm, trim={0.1cm 0.1cm 0.1cm 0.1cm}, clip]{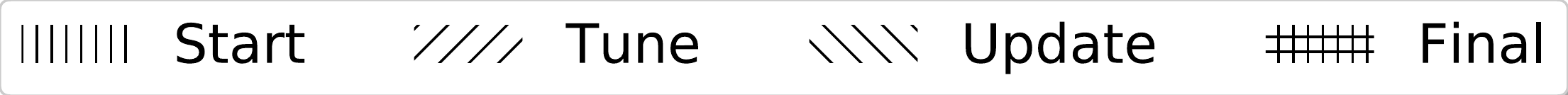}
    \end{subfigure}\\
    \begin{subfigure}[b]{0.53\textwidth}
    \vspace{1.5em}
    \begin{picture}(100, 100)
        \put(0,4.5){\includegraphics[height=3.9cm, trim={0.8cm 0.8cm 0.9cm 2.2cm}, clip]{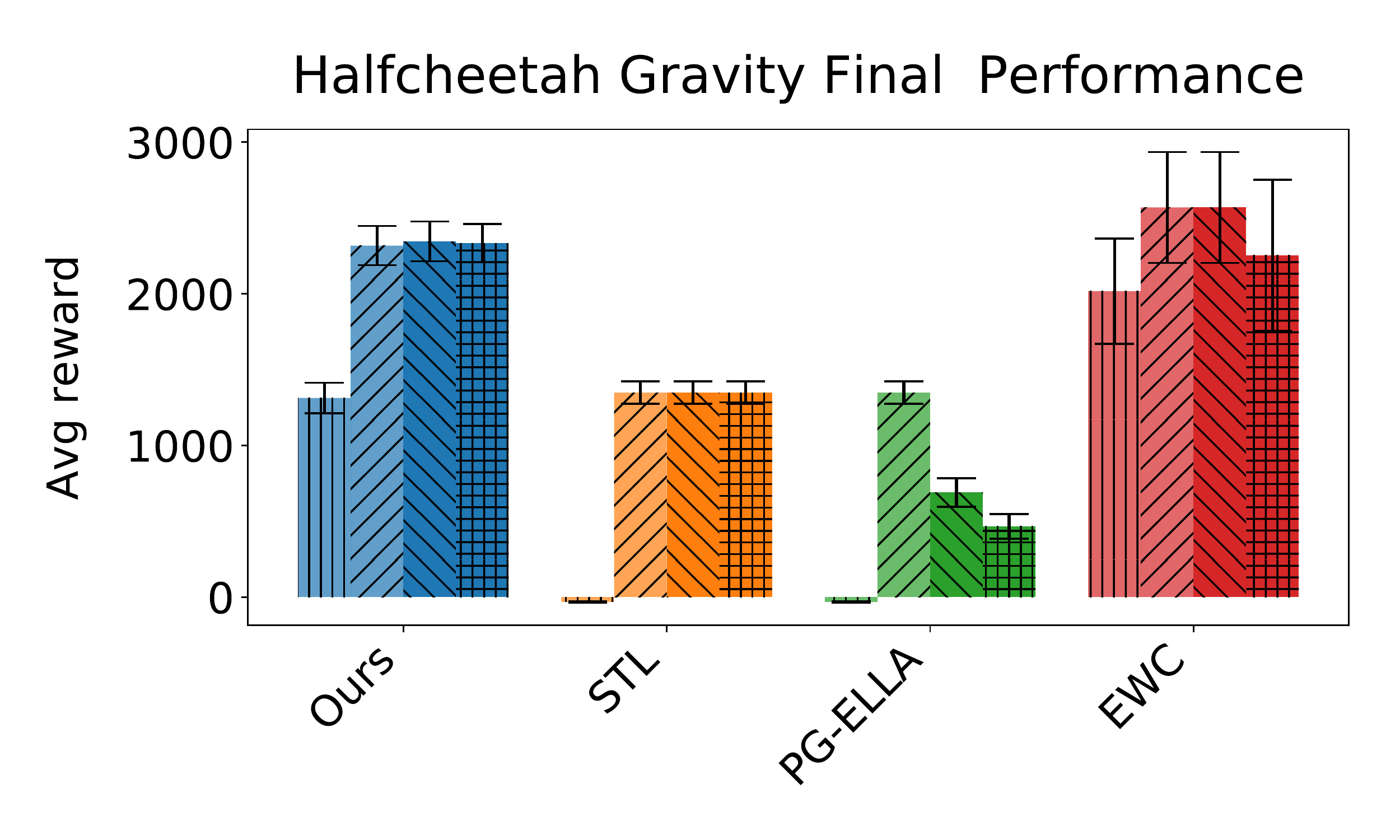}}
        \put(28,0){\rotatebox{45}{\fontfamily{DejaVuSans-TLF}\selectfont\fontsize{7pt}{7.5pt}\selectfont \colorbox{white}{LPG-FTW}}}
    \end{picture}
        \vspace{-0.5em}
        \caption{HalfCheetah \textit{gravity}}
    \end{subfigure}%
    \begin{subfigure}[b]{0.47\textwidth}
    \vspace{1.5em}
    \begin{picture}(100, 100)
        \put(0,4.5){\includegraphics[height=3.9cm, trim={2.3cm 0.8cm 0.9cm 2.2cm}, clip]{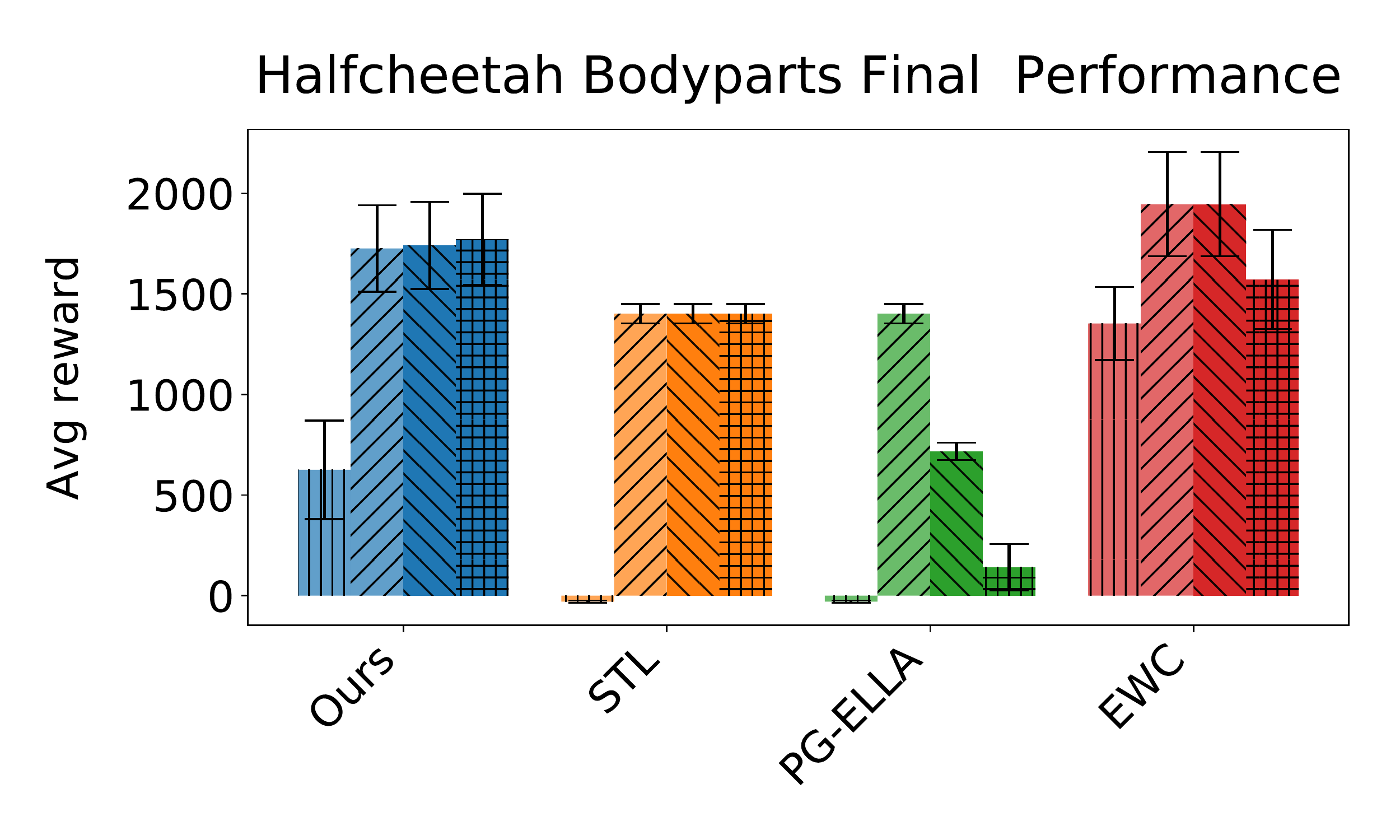}}
        \put(13,0){\rotatebox{45}{\fontfamily{DejaVuSans-TLF}\selectfont\fontsize{7pt}{7.5pt}\selectfont \colorbox{white}{LPG-FTW}}}
    \end{picture}
        \vspace{-0.5em}
        \caption{HalfCheetah \textit{body-parts}}
    \end{subfigure}\\
    \begin{subfigure}[b]{0.53\textwidth}
    \vspace{1em}
    \begin{picture}(100,100)
        \put(0,4.5){\includegraphics[height=3.9cm, trim={0.8cm 0.8cm 0.9cm 2.2cm}, clip]{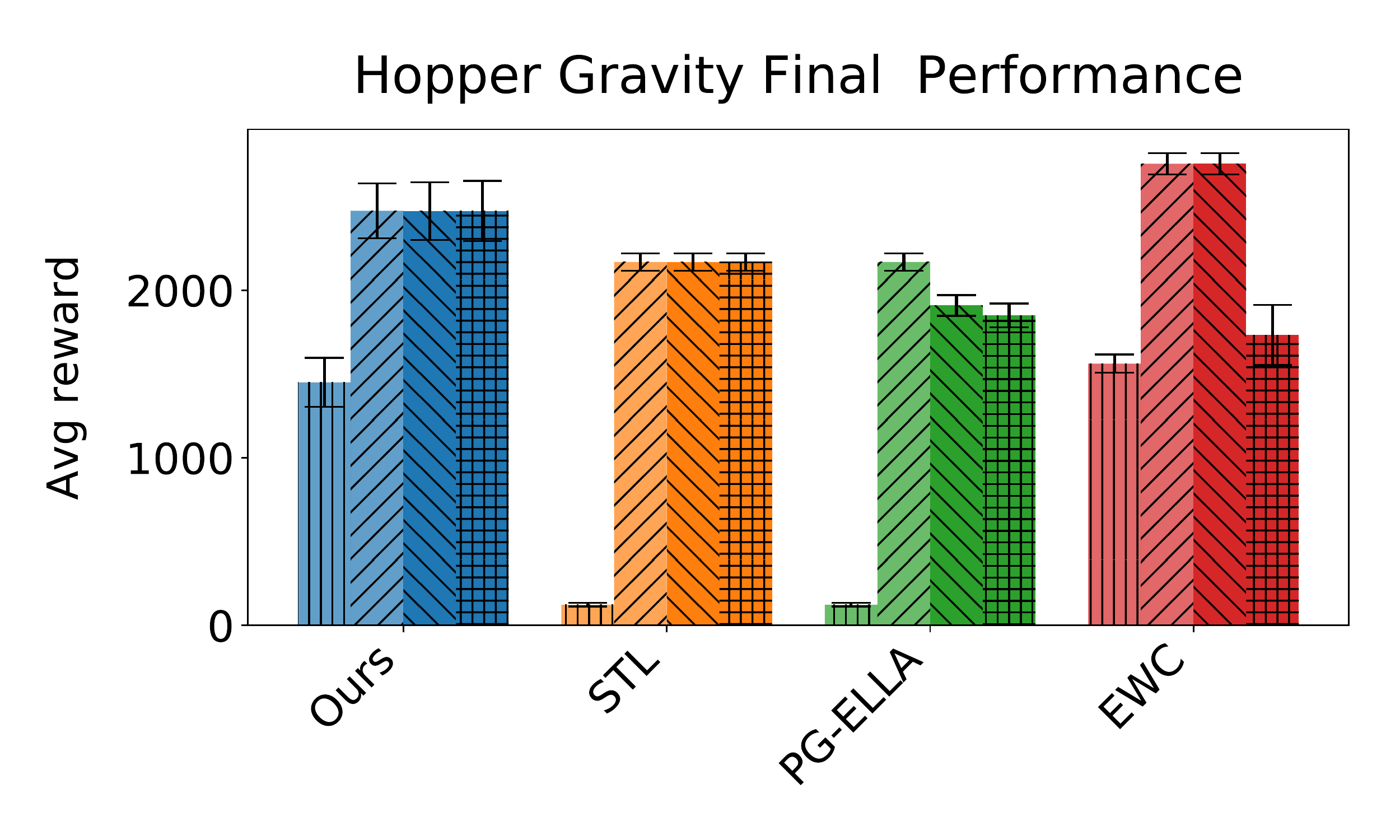}}
        \put(28,0){\rotatebox{45}{\fontfamily{DejaVuSans-TLF}\selectfont\fontsize{7pt}{7.5pt}\selectfont \colorbox{white}{LPG-FTW}}}
    \end{picture}
        \vspace{-0.5em}
         \caption{Hopper \textit{gravity}}
    \end{subfigure}%
    \begin{subfigure}[b]{0.47\textwidth}
    \vspace{1em}
    \begin{picture}(100,100)
        \put(0,4.5){\includegraphics[height=3.9cm, trim={2.3cm 0.8cm 0.9cm 2.2cm}, clip]{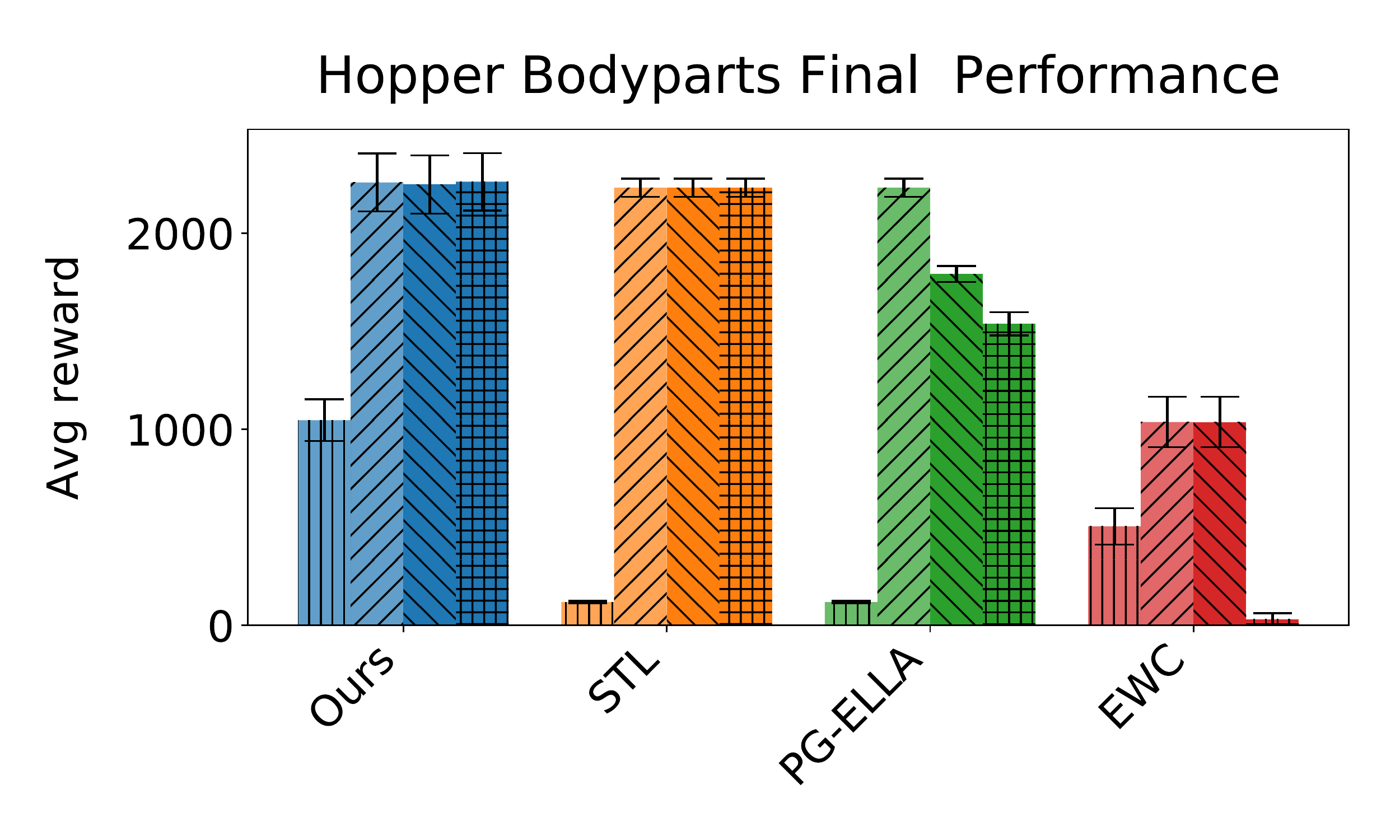}}
        \put(13,0){\rotatebox{45}{\fontfamily{DejaVuSans-TLF}\selectfont\fontsize{7pt}{7.5pt}\selectfont \colorbox{white}{LPG-FTW}}}
    \end{picture}
        \vspace{-0.5em}
         \caption{Hopper \textit{body-parts}}
    \end{subfigure}\\
    \begin{subfigure}[b]{0.53\textwidth}
    \vspace{1.5em}
    \begin{picture}(100,100)
        \put(0,4.5){\includegraphics[height=3.9cm, trim={0.8cm 0.8cm 0.9cm 2.2cm}, clip]{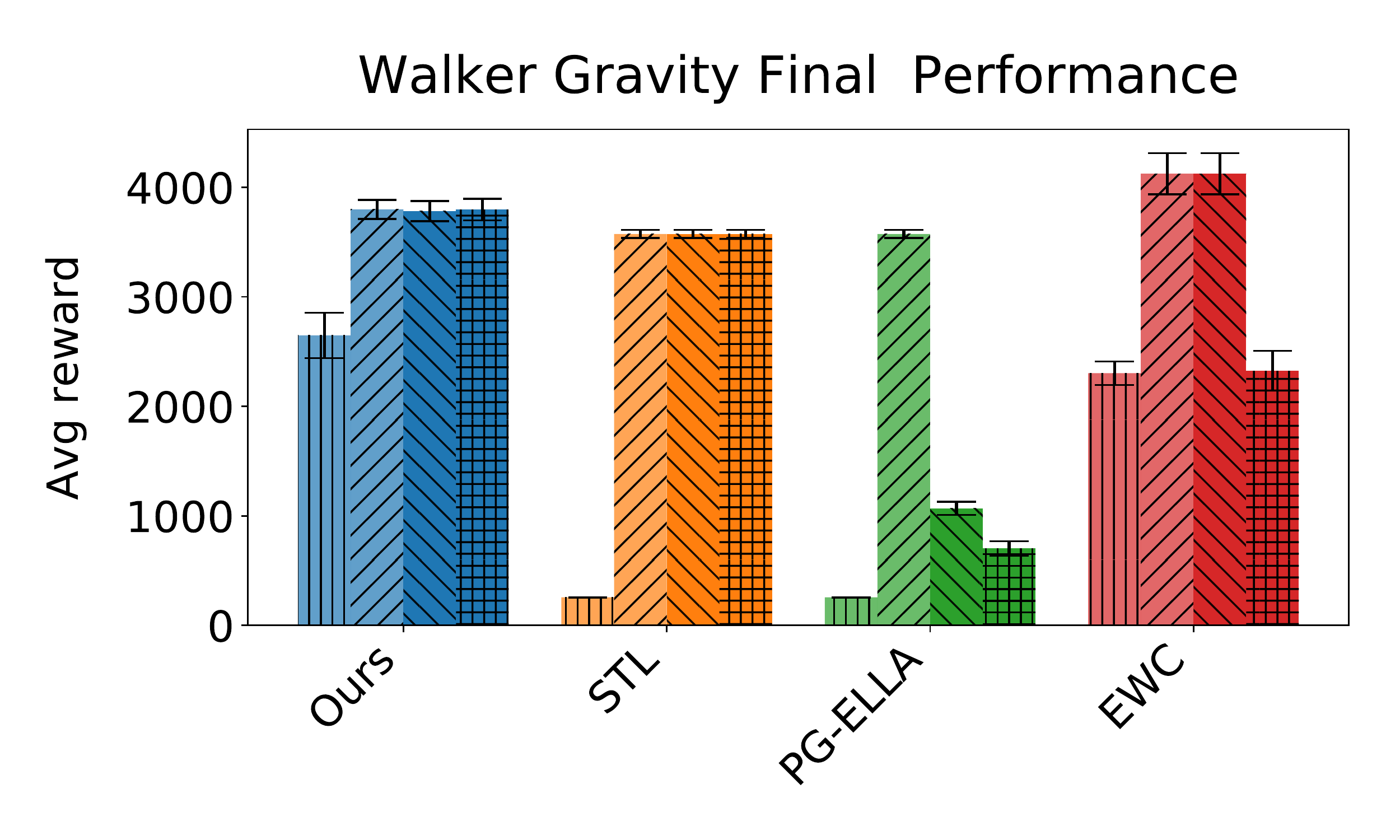}}
        \put(28,0){\rotatebox{45}{\fontfamily{DejaVuSans-TLF}\selectfont\fontsize{7pt}{7.5pt}\selectfont \colorbox{white}{LPG-FTW}}}
    \end{picture}
        \vspace{-0.5em}
         \caption{Walker-2D \textit{gravity}}
         \vspace{-0.5em}
    \end{subfigure}%
    \begin{subfigure}[b]{0.47\textwidth}
    \vspace{1.5em}
    \begin{picture}(100,100)
        \put(0,4.5){\includegraphics[height=3.9cm, trim={2.3cm 0.8cm 0.9cm 2.2cm}, clip]{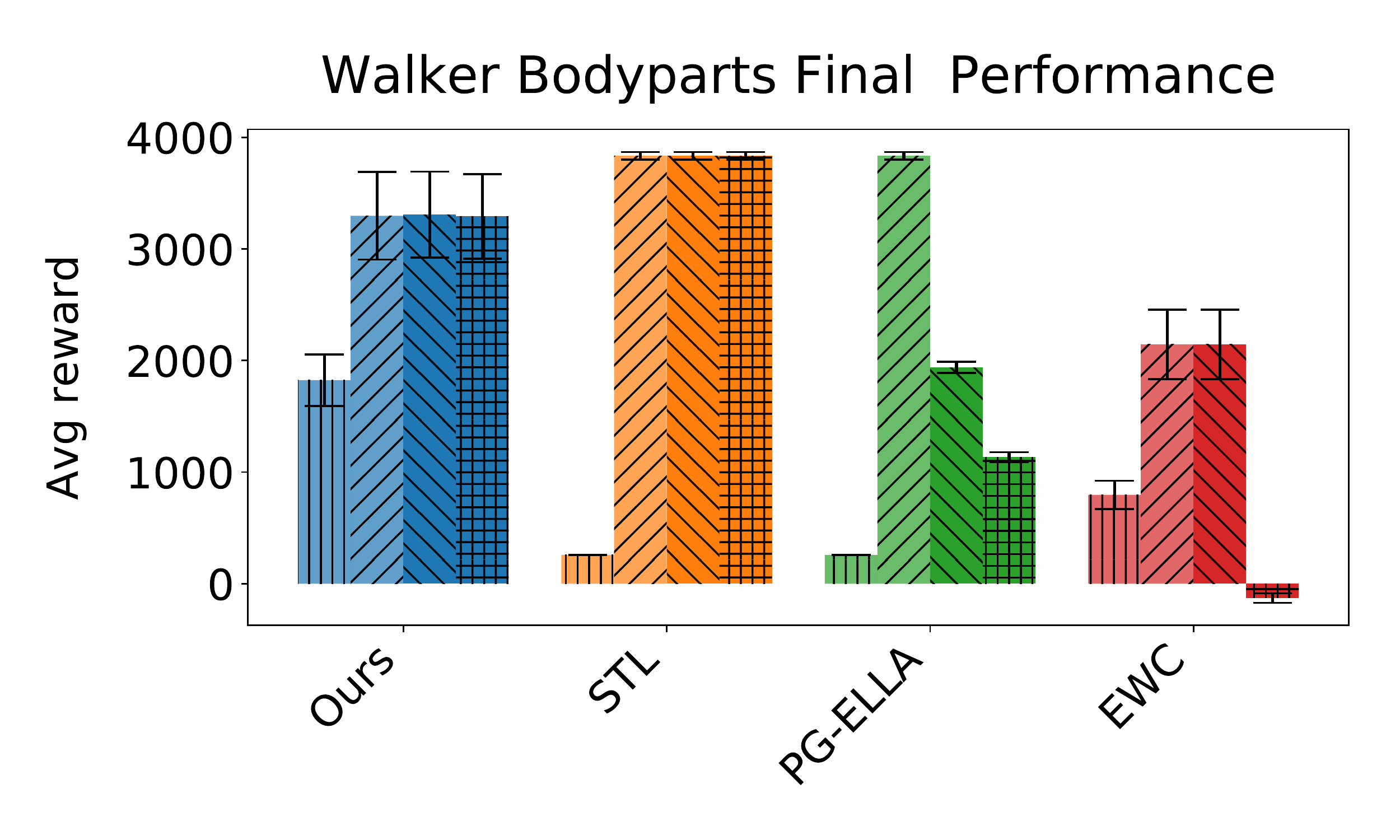}}
        \put(13,0){\rotatebox{45}{\fontfamily{DejaVuSans-TLF}\selectfont\fontsize{7pt}{7.5pt}\selectfont \colorbox{white}{LPG-FTW}}}
    \end{picture}
        \vspace{-0.5em}
         \caption{Walker-2D \textit{body-parts}}
         \vspace{-0.5em}
    \end{subfigure}
    \caption[Performance of \lpgftw{} in lifelong compositional RL before and after each task, and after all tasks on MuJoCo domains with linear policies.]{Average performance at the beginning of training (start), after all training iterations (tune, equivalent to the final point in Figure~\ref{fig:learningCurves}), after the update step for PG-ELLA and \lpgftw{} (update), and after training on all tasks (final). The update step in \lpgftw{} never hinders performance, and even after training on all tasks the agent maintains performance. PG-ELLA always performed worse than STL. EWC suffered from catastrophic forgetting in five  domains, in two resulting in degradation below initial performance. Error bars denote standard errors across five seeds.}
    \label{fig:testPerformance}
\end{figure}

Results in Figure~\ref{fig:learningCurves} consider only how fast the agent learns a new task using information from earlier tasks. 
PG-ELLA and \lpgftw{} then perform an update step (Equation~\ref{equ:UpdateL} for \lpgftw{}) where they incorporate knowledge from the current task into $\bL$. The third bar from the left per each algorithm in Figure~\ref{fig:testPerformance} shows the average performance after this step, revealing that \lpgftw{} maintained performance, whereas PG-ELLA's performance decreased. This is because \lpgftw{} ensures that the learner computes the approximate objective near points in the parameter space that the current basis $\bL$ can generate, by finding $\alphat$ via a search over the span of $\bL$. 
A critical component of lifelong learning algorithms is their ability to avoid catastrophic forgetting. To assess the capacity of \lpgftw{} to retain knowledge from earlier tasks, the evaluation measured the performance of the policies obtained from the knowledge base $\bL$ after training on all tasks, without modifying the $\st$'s. The rightmost bar in each algorithm in Figure~\ref{fig:testPerformance} shows the average final performance across all tasks. \lpgftw{} successfully retained knowledge of all tasks, showing no signs of catastrophic forgetting on any of the domains. The PG-ELLA baseline suffered from forgetting on all domains, and EWC on all but one of the domains. Moreover, the final performance of \lpgftw{} was the best among all baselines on all but one domain.

\begin{figure}[b!]
\centering
    \includegraphics[width=0.7\linewidth, trim={0.5cm 0.6cm 0.5cm 1.7cm}, clip]{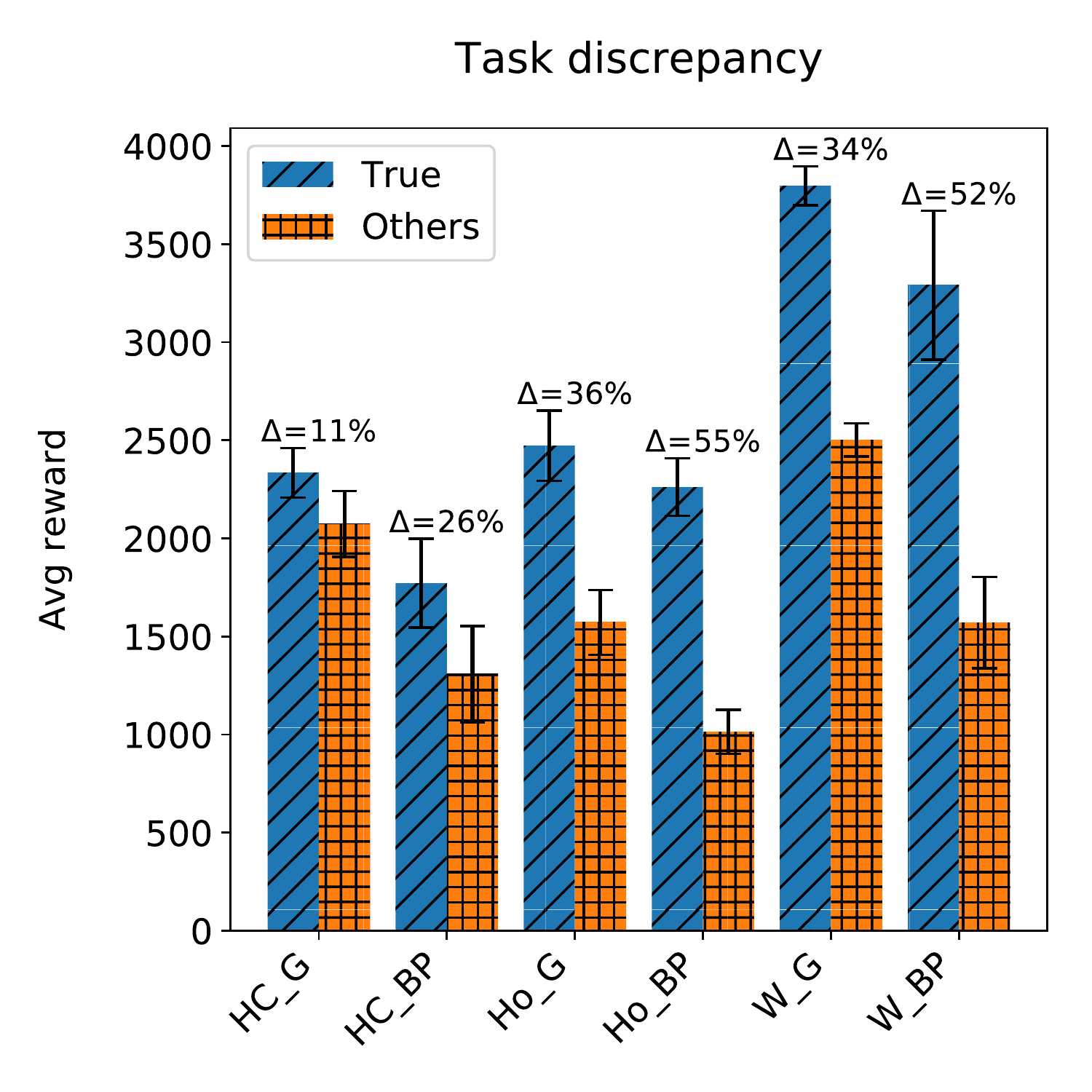}
    \caption[Diversity of tasks in MuJoCo domains used to evaluate lifelong compositional RL with \lpgftw{}.]{Performance with the true policy vs.~other policies. Percent gap ($\Delta$) indicates task diversity. Body parts (BP) domains are more diverse than gravity (G) domains, and Walker-2D (W) and Hopper (Ho) domains are more varied than HalfCheetah (HC) domains. Error bars denote standard errors across five seeds.}
    \label{fig:taskDiscrepancy}
\end{figure}

One important question in the study of lifelong RL is how diverse the tasks used for evaluation are. To measure this in the OpenAI Gym MuJoCo domains, the next experiment compared the performance on each task using the final policy trained by \lpgftw{} on the correct task and the average performance using the policies trained on all other tasks. Figure~\ref{fig:taskDiscrepancy} shows that the policies do not work well across different tasks, demonstrating that the tasks are diverse. Moreover, the most highly-varying domains, Hopper and Walker-2D \textit{body-parts}, are precisely those for which EWC struggled the most, suffering from catastrophic forgetting, as shown in Figure~\ref{fig:testPerformance}. This is consistent with the fact that a single policy does not work across various tasks. In those domains, \lpgftw{} reached the performance of STL with a high speedup while retaining knowledge from early tasks.

\subsection{Empirical Evaluation on More Challenging Meta-World Domains}
\label{sec:ExperimentalResultsChallenging}

Results so far show that \lpgftw{} improves performance and completely avoids forgetting in simple settings. To study the flexibility of the framework to handle more complex RL tasks, the evaluation tested it on sequential versions of the Meta-World MT10 and MT50 benchmarks~\citep{yu2019meta}, repeating each experiment with five different random seeds controlling the parameter initialization and the ordering over tasks.

\begin{figure}[b!]
\centering
\captionsetup[subfigure]{belowskip=4pt}
    \begin{subfigure}[b]{\textwidth}
        \centering
        \includegraphics[height=0.45cm, trim={0.1cm 0.1cm 28cm 0.1cm}, clip]{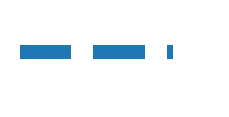} 
        \hspace{-0.1em}\raisebox{0.13cm}{\fontfamily{DejaVuSans-TLF}\selectfont\scriptsize LPG-FTW}\hspace{1.em}
        \includegraphics[height=0.45cm, trim={5.5cm 0.1cm 0.1cm 0.1cm}, clip]{chapter4/Figures/LPG-FTW/learning_curves/learning_curve_legend_metaworld_horizontal.pdf} 
    \end{subfigure}\\
    \begin{subfigure}[b]{0.53\textwidth}
    \centering
        \includegraphics[height=3.72cm, trim={2.1cm 4.3cm 3cm 2.2cm}, clip]{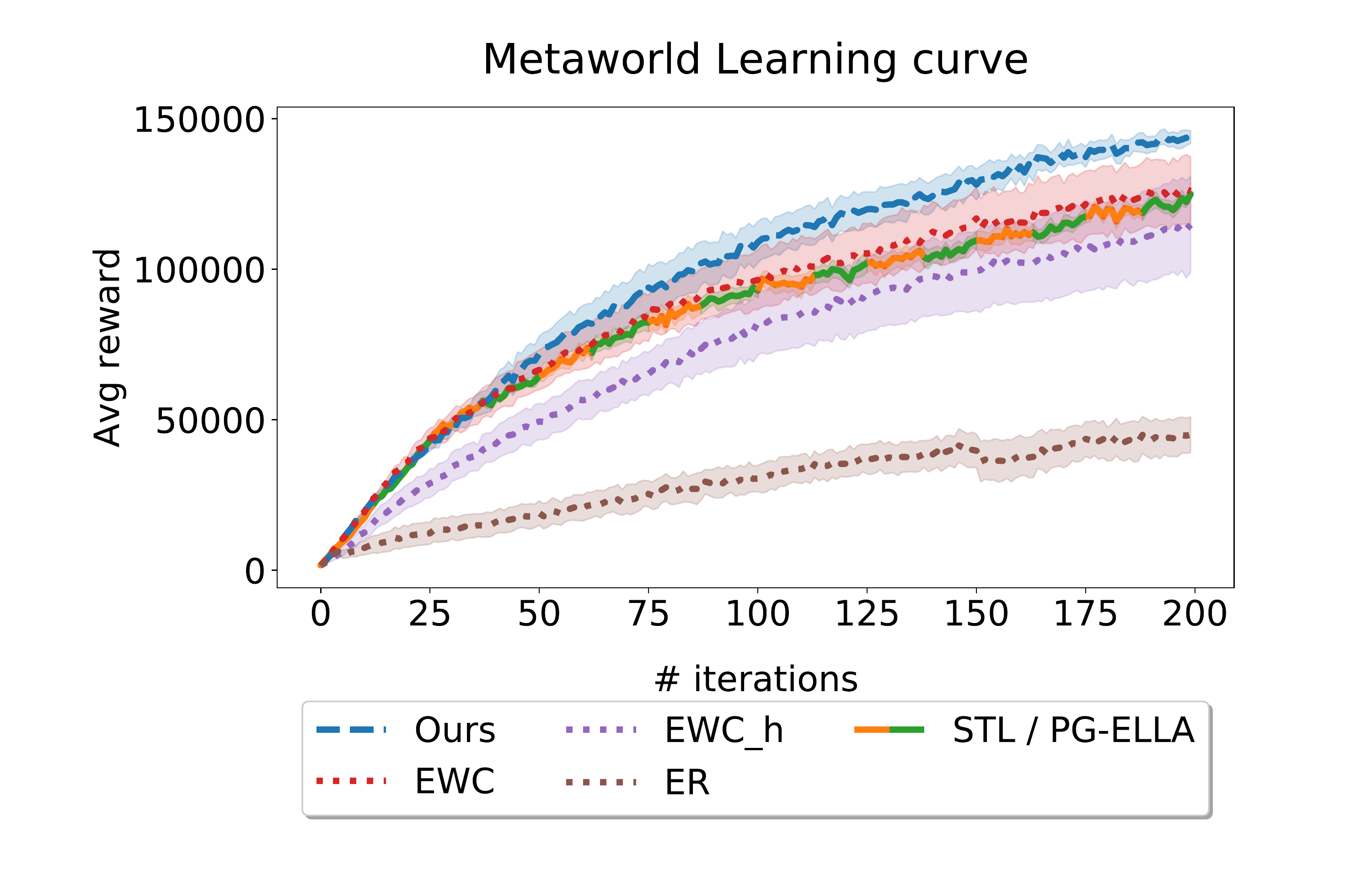}
        \vspace{1em}
    \end{subfigure}%
    \begin{subfigure}[b]{0.47\textwidth}
    \centering
        \includegraphics[height=3.72cm, trim={3cm 4.3cm 3cm 2.3cm}, clip]{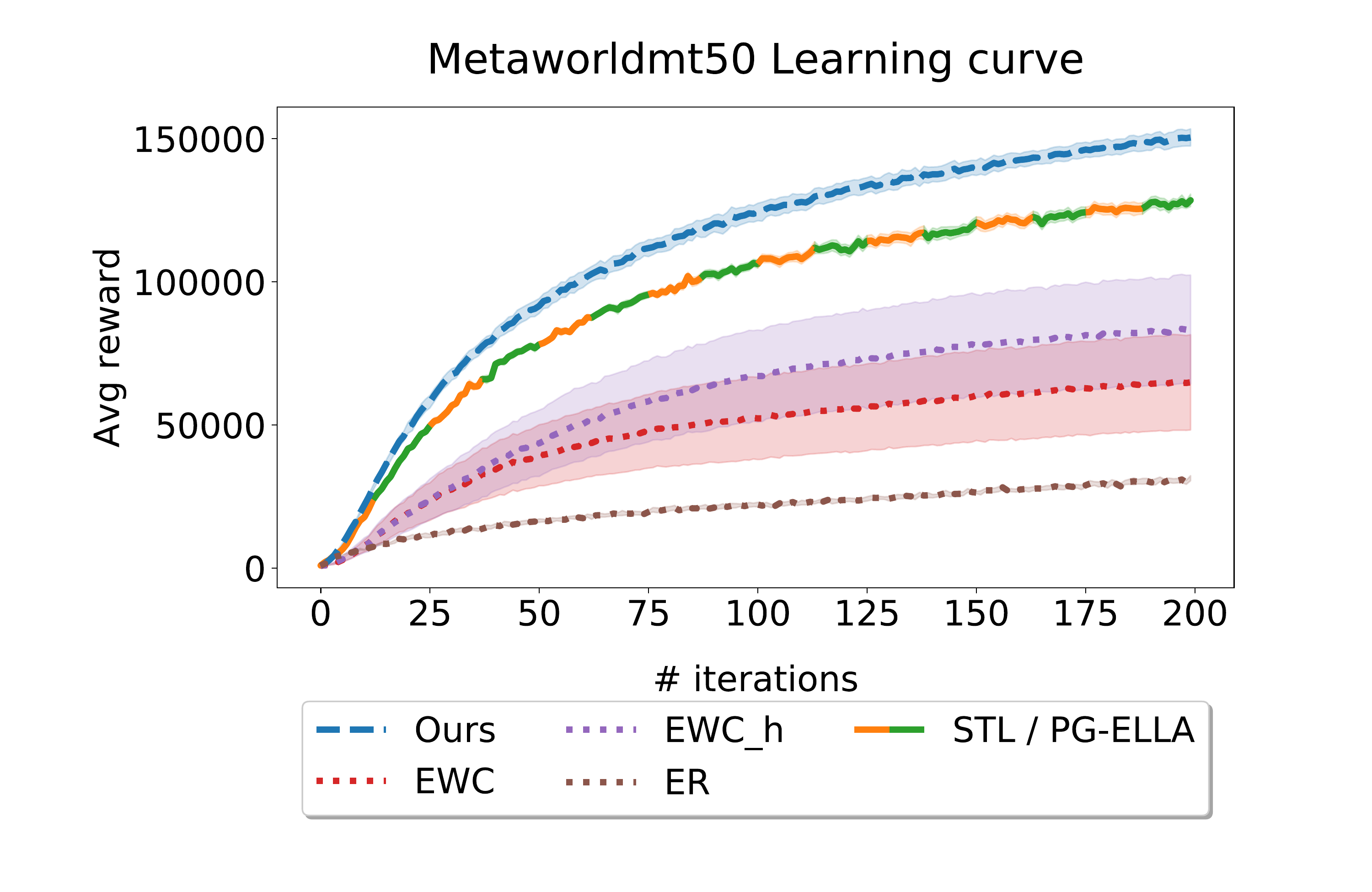}
        \vspace{1em}
    \end{subfigure}\\
    \begin{subfigure}[b]{\textwidth}
        \centering
        \includegraphics[height=0.45cm, trim={0.1cm 0.1cm 0.1cm 0.1cm}, clip]{chapter4/Figures/LPG-FTW/final_performance/bar_chart_legend_cameraready.pdf}
    \end{subfigure}\\
    \begin{subfigure}[b]{0.53\textwidth}
    \centering
    \vspace{1.5em}
    \hspace{-3.7cm}
    \begin{picture}(100,100)
        \put(0,1.5){\includegraphics[height=3.95cm, trim={1.5cm 3.3cm 3cm 2.3cm}, clip]{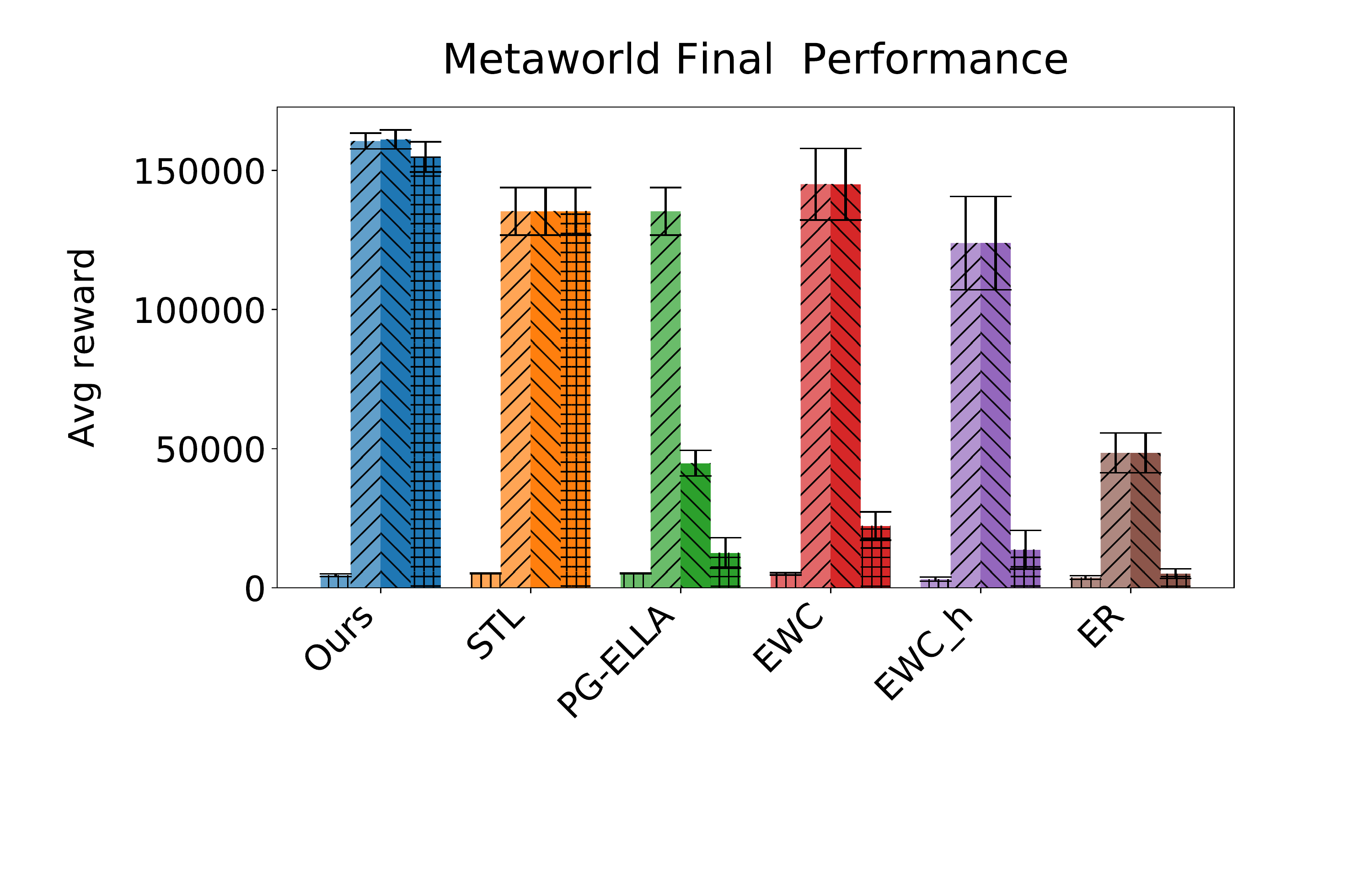}}
        \put(28,0){\rotatebox{45}{\fontfamily{DejaVuSans-TLF}\selectfont\tiny \colorbox{white}{LPG-FTW}}}
    \end{picture}
        \vspace{-0.5em}
        \caption{Meta-World MT10}
    \end{subfigure}%
    \begin{subfigure}[b]{0.47\textwidth}
    \centering
    \vspace{1.5em}
    \hspace{-3.7cm}
    \begin{picture}(100,100)
        \put(0,1.5){\includegraphics[height=3.95cm, trim={3cm 3.3cm 3cm 2.3cm}, clip]{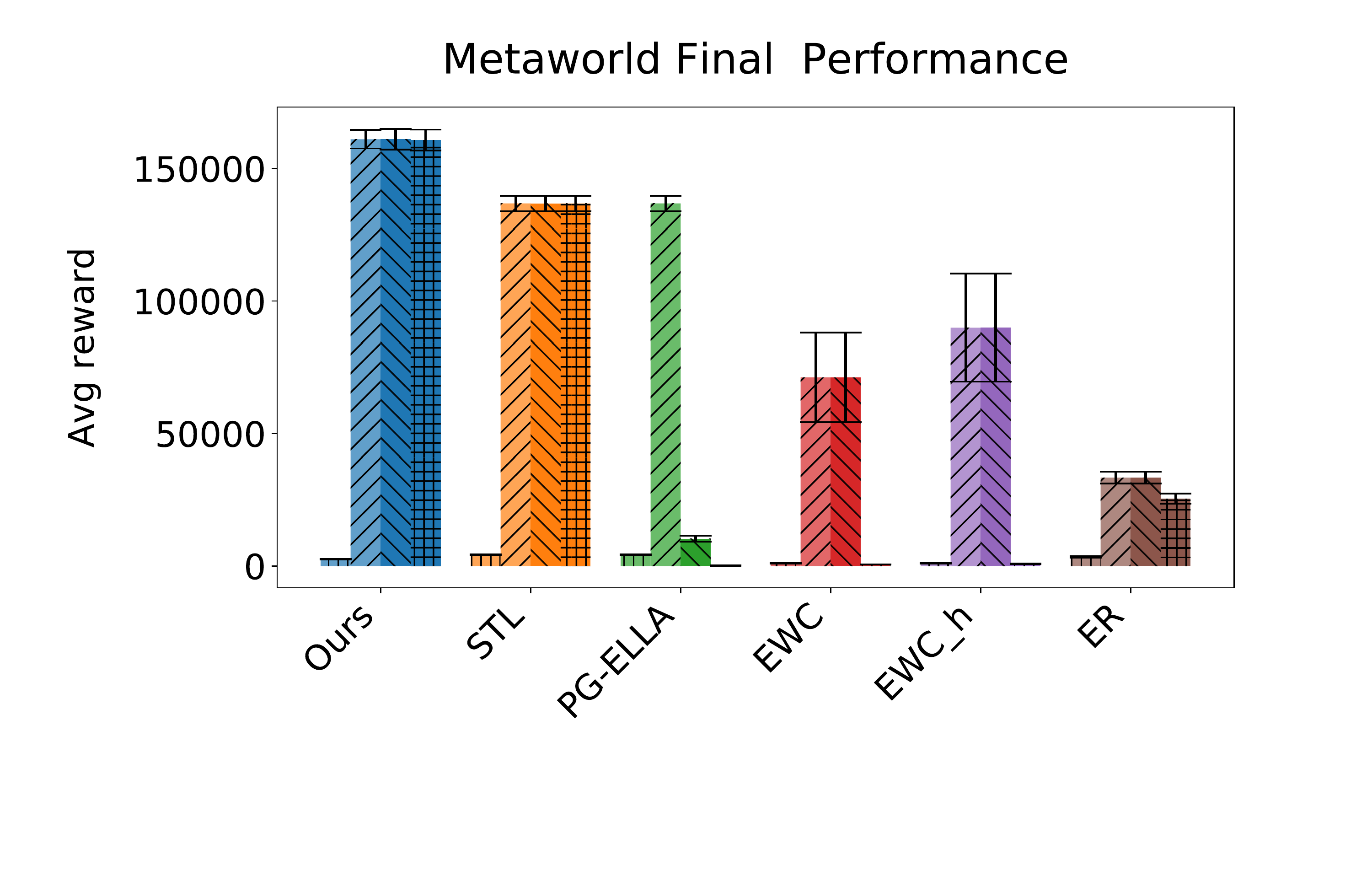}}
        \put(15,0){\rotatebox{45}{\fontfamily{DejaVuSans-TLF}\selectfont\tiny \colorbox{white}{LPG-FTW}}}
    \end{picture}
        \vspace{-0.5em}
         \caption{Meta-World MT50}
    \end{subfigure}\\
    \caption[Performance of \lpgftw{} in lifelong compositional RL on Meta-World domain with deep policies.]{Performance on the Meta-World benchmark. Top: average performance during training across all tasks. Bottom: average performance at the beginning of training (start), after all training iterations (tune), after the update step for PG-ELLA and \lpgftw{} (update), and after all tasks have been trained (final). In this notoriously challenging benchmark, \lpgftw{} still improves the performance of STL and all baselines, and suffers from no catastrophic forgetting. All lifelong baselines suffer from catastrophic forgetting. Error bars and shaded regions denote standard errors across five seeds.}
    \label{fig:PerformanceMetaWorld}
\end{figure}

The top row of Figure~\ref{fig:PerformanceMetaWorld} shows average learning curves across tasks. \lpgftw{} again was faster in training, showing that the restriction that the agent only train the $\st$'s for each new task does not harm its ability to solve complex, highly diverse problems. The difference in learning speed was particularly noticeable on MT50, where single-model methods became saturated. This was the first published display of lifelong transfer on the challenging Meta-World benchmark. The bottom row of Figure~\ref{fig:PerformanceMetaWorld} shows that \lpgftw{} suffered from a small amount of forgetting on MT10. However, on MT50, where $\bL$ trained on sufficient tasks for convergence, \lpgftw{} suffered from no forgetting. In contrast, none of the baselines was capable of accelerating the learning, and they all suffered from dramatic forgetting, particularly on MT50, when needing to learn more tasks. Adding capacity to EWC did not substantially alter these results, showing that the ability of \lpgftw{} to handle highly varied tasks does not stem from its higher capacity from using $k$ factors, but instead to the use of \textit{different} models for each task by intelligently combining the shared components.

\subsection{Zero-Shot Transfer to Unseen Discrete 2-D Task Combinations via Multitask Learning}
\label{sec:zeroShotEvaluation}

To assess whether the compositional tasks described in Section~\ref{sec:EvaluationDomainsRL} exhibit the expected composition, the agent received a hard-coded graph structure following the formalism of Section~\ref{sec:lifelongCompositionInRL}. This way, the agent knows a priori which modules to use for each task and only needs to learn the module parameters. The architecture uses four modules of each of three types, one for each task component (static object, target object, and agent dynamics). 
Each task policy contains one module of each type, chained as {\textsf{ static object}} $\rightarrow$ \textsf{ target object} $\rightarrow$ {\textsf{ agent}}. Modules are convolutional nets whose inputs are the module-specific state components and the outputs of the previous modules. Agent modules output both the action and the $Q$-values. The agent trained in a batch MTL fashion using PPO, collecting data from all tasks at each training step and 
computing the average gradient across tasks to update the parameters. Since each task uses a single module of each type, gradient updates only affect the relevant modules to each task.

\begin{figure}[t]
    \centering
    \begin{subfigure}[b]{0.56\textwidth}
        \centering
        \includegraphics[trim={0.1cm 0.1cm 0.1cm 0.15cm}, clip, height=0.45cm]{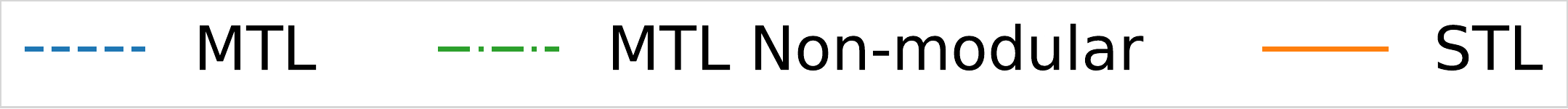}
    \end{subfigure}%
    \hfill
    \begin{subfigure}[b]{0.42\textwidth}
        \centering
        \includegraphics[trim={0.1cm 0.1cm 0.1cm 0.15cm}, clip,height=0.45cm]{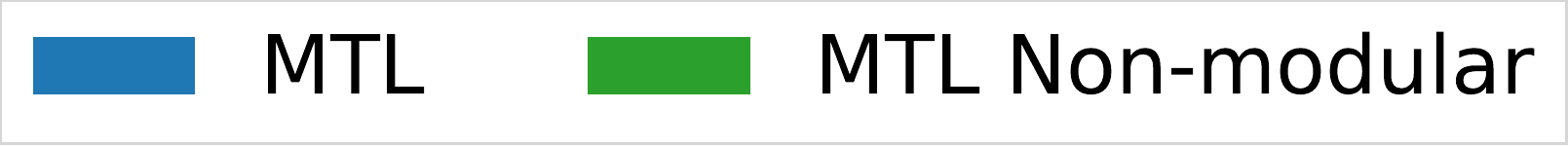}
    \end{subfigure}\\
    \begin{subfigure}[b]{0.56\textwidth}    
        \centering
       \includegraphics[trim={1.1cm 0.5cm 0cm 1.2cm}, clip,height=4.5cm]{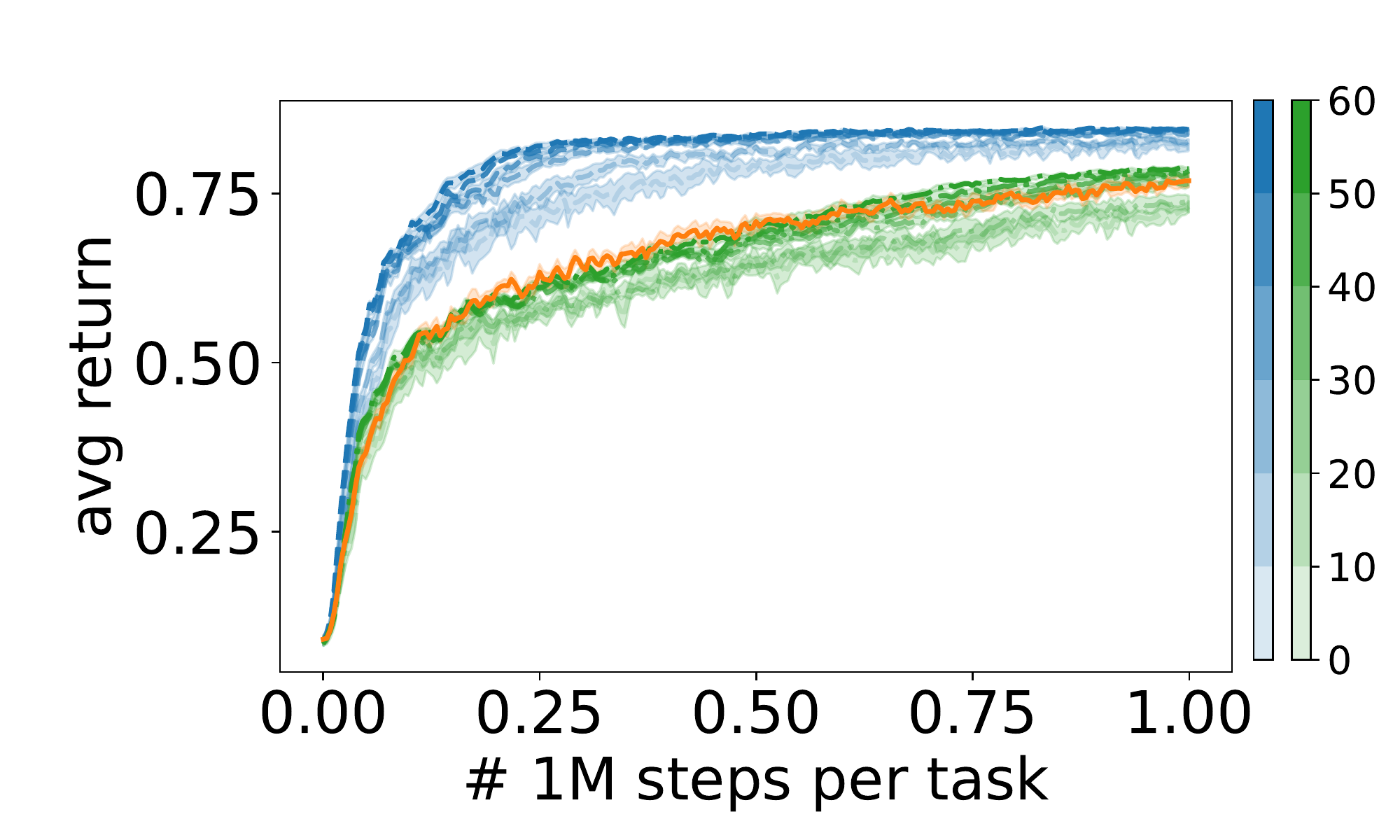}
       \caption{Learning curves}
       \label{fig:compositionalMinigridMTLcurves}
    \end{subfigure}%
    \hfill
    \begin{subfigure}[b]{0.42\textwidth}  
        \centering
       \includegraphics[trim={0.5cm 0.5cm 0.3cm 0.3cm}, clip,height=4.5cm]{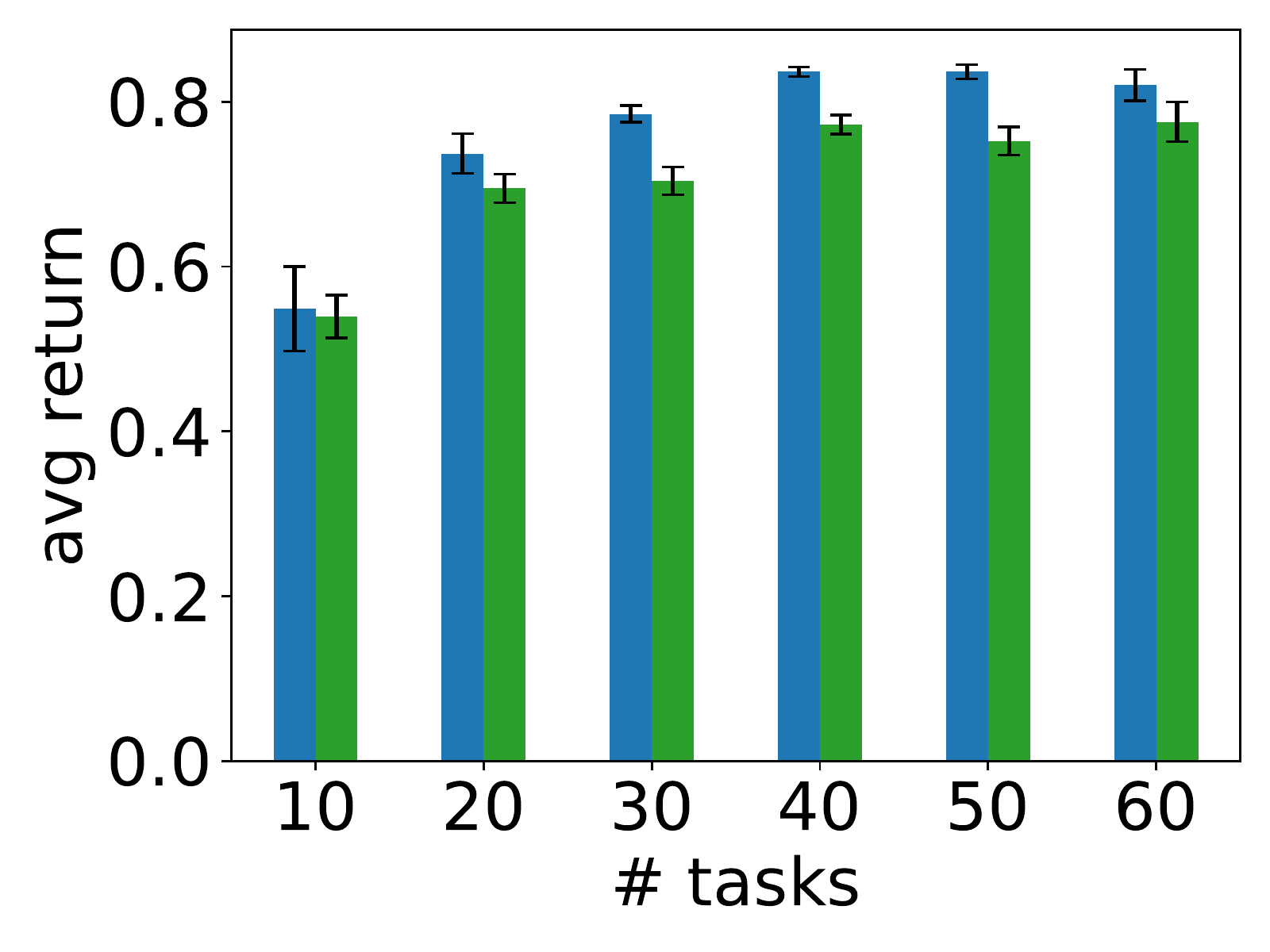}
       \caption{Zero-shot generalization}
       \label{fig:compositionalMinigridMTLzeroshot}
    \end{subfigure}%
    \caption[Performance of modular MTL agent on explicitly compositional discrete 2-D RL tasks.]{Average returns of STL (trained on $\numTasks=64$ tasks) and MTL (trained on various tasks, as indicated by the color bar) on 2-D discrete tasks. (a) The modular architecture correctly captures the relations across tasks, accelerating learning. Training on more tasks further improves results. (b) Generalization of pretrained modules to unseen combinations as a function of the number of training tasks. Modules can be combined in novel ways to achieve high performance without additional training. Shaded regions and error bars show standard errors across six seeds.}
    \label{fig:compositionalMinigridMTL}
\end{figure}

The agent trained on various possible combinations of discrete 2-D tasks, and the evaluation compared its performance against two baselines: training a separate STL agent on each task, and training a single monolithic network across all tasks in the same MTL fashion. To ensure a fair comparison, the monolithic MTL network received as input a multi-hot encoding of the components that constituted each task.  The evaluation repeated this experiment  with six different random parameter initialization configurations and samples of training tasks. Figure~\ref{fig:compositionalMinigridMTLcurves} shows that the compositional method was substantially faster than the two alternatives by sharing relevant information across tasks. 

These results suggest that the proposed modular architecture accurately captures the relations across tasks in the form of modules. To verify this, the next experiment evaluated the agent on tasks it had not encountered during training in a zero-shot manner. To construct the network for each task, the agent again received the hard-coded graph structure, but it kept the parameters of the modules fixed after multitask training. Figure~\ref{fig:compositionalMinigridMTLzeroshot} shows the high zero-shot performance of the modular method, revealing that the modules can be combined in novel ways to solve unseen tasks without any additional training.

\subsection{Lifelong Discovery of Modules on Discrete 2-D Tasks}
\label{sec:resultsMiniGrid}

The results of Figure~\ref{fig:compositionalMinigridMTL} encourage envisioning a lifelong learner improving modules over a sequence of tasks and reusing those modules to learn new tasks faster. This section studies the ability of \compRL{} to achieve this kind of lifelong composition. This evaluation considered two instances of \compRL{}: one in which the agent receives the hard-coded graph structure (Comp.+Struct.) and one in which the agent receives no information about which tasks share which components and must therefore discover these relations autonomously via discrete search (Comp.+Search). Once the agent selects the structure for one particular task, it collects data for that task via PPO training, starting from the parameters of the selected modules. Finally, the agent uses this collected data for incorporating knowledge about the current task into the selected modules via off-line RL with data from the current task and all tasks that reuse any of those modules to avoid forgetting. To match the assumptions of \compRL{}, unless otherwise stated, the initial tasks presented to the agent contain disjoint sets of components. The evaluation carried out the experiments in this section with six different random seeds controlling the parameter initialization and the ordering over tasks.

\begin{figure}[t!]
    \centering
    \begin{subfigure}[b]{\textwidth}
        \centering
        \includegraphics[trim={0.1cm 0.1cm 0.1cm 0.15cm}, clip,height=0.9cm]{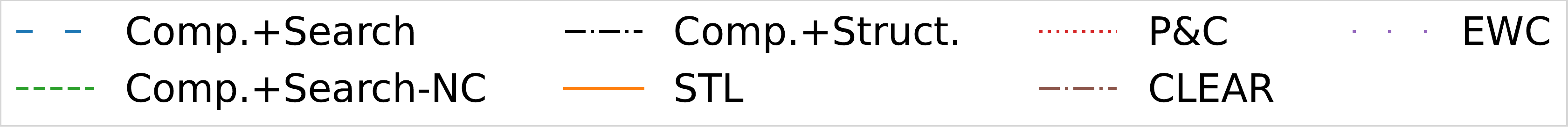}
    \end{subfigure}\\
    \begin{subfigure}[b]{0.5\textwidth}
        \centering
        \includegraphics[trim={0.2cm 0.3cm 1.3cm 1.3cm}, clip=True, height=5.5cm]{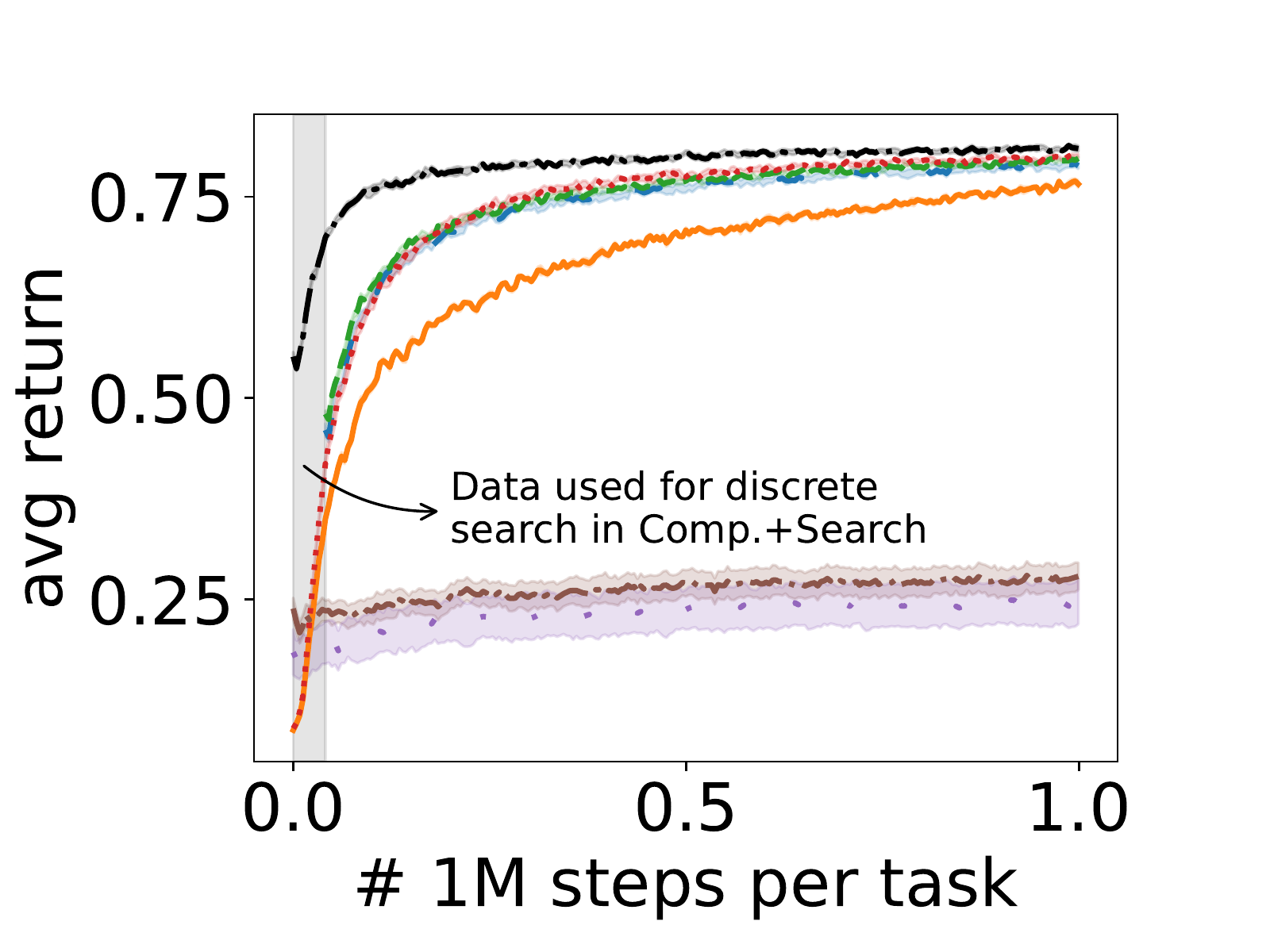}
        \caption{Learning curves}
        \label{fig:compositionalMinigridLLcurves}
    \end{subfigure}%
    \begin{subfigure}[b]{0.5\textwidth} 
        \centering
        \includegraphics[trim={0.2cm 0.3cm 0.3cm 0.3cm}, clip=True, height=5.5cm]{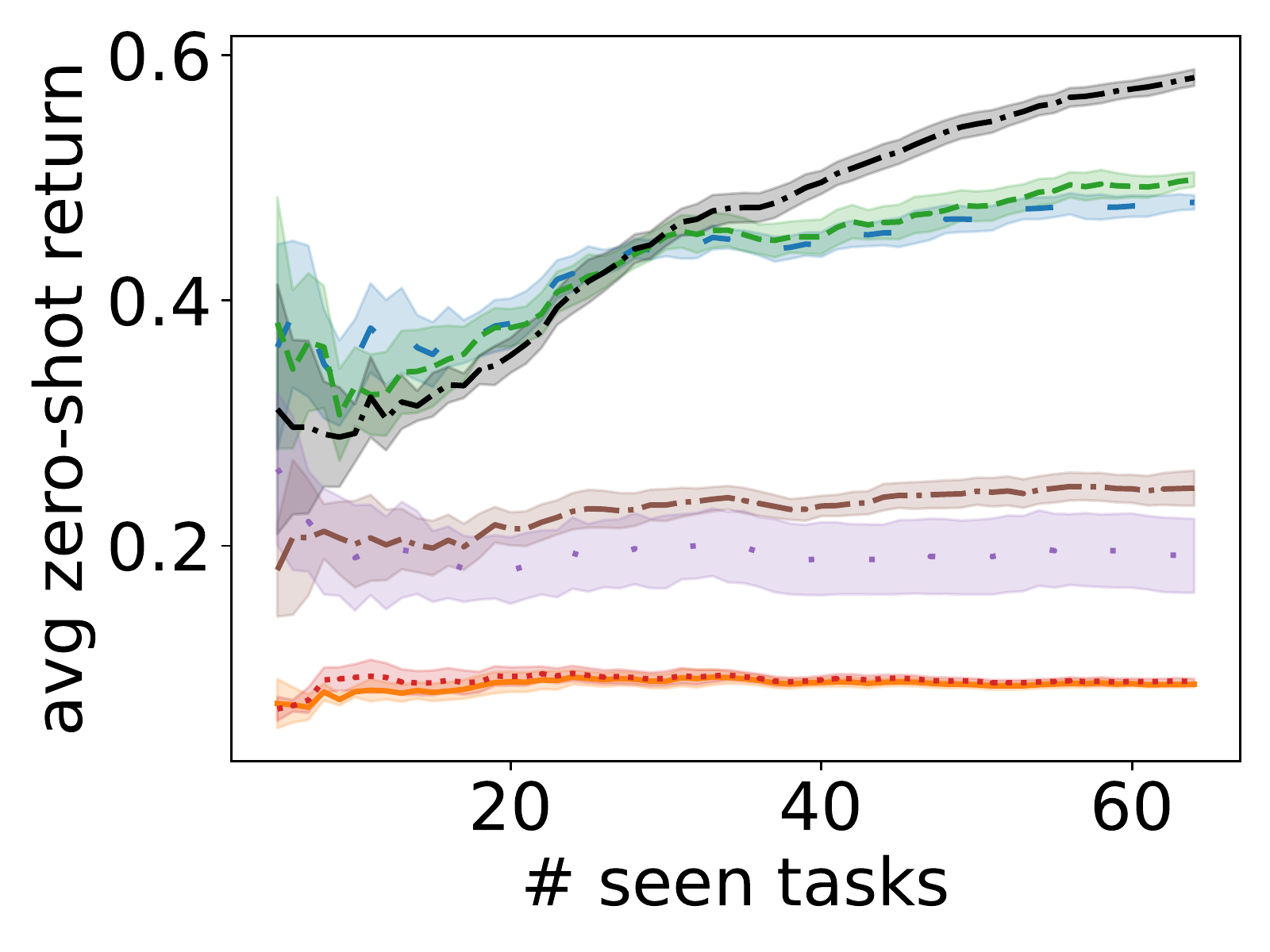}
        \caption{Zero-shot performance over time}
        \label{fig:compositionalMinigridLLzeroshot}
    \end{subfigure}\\
    \vspace{1em}
    \begin{subfigure}[b]{0.66\textwidth}
        \centering
        \includegraphics[trim={0.1cm 0.1cm 0.1cm 0.15cm}, clip,height=0.9cm]{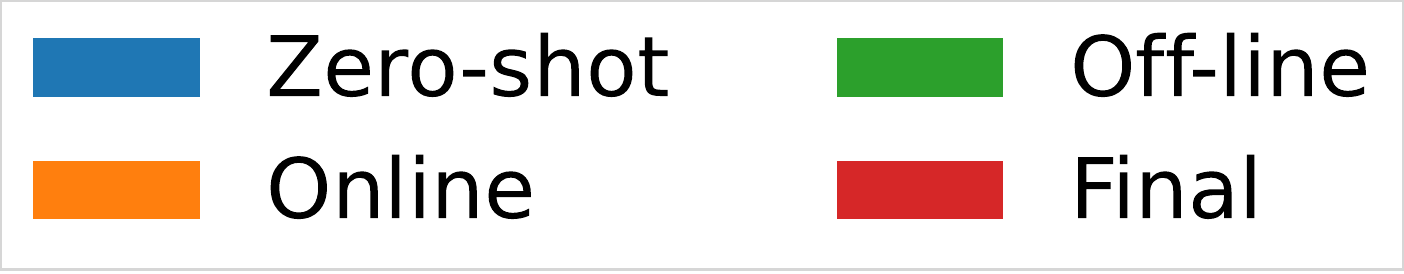}
    \end{subfigure}\\
    \begin{subfigure}[b]{0.66\textwidth}  
        \centering
        \includegraphics[trim={0.2cm 0.5cm 0.3cm 0.3cm}, clip=True, height=5.2cm]{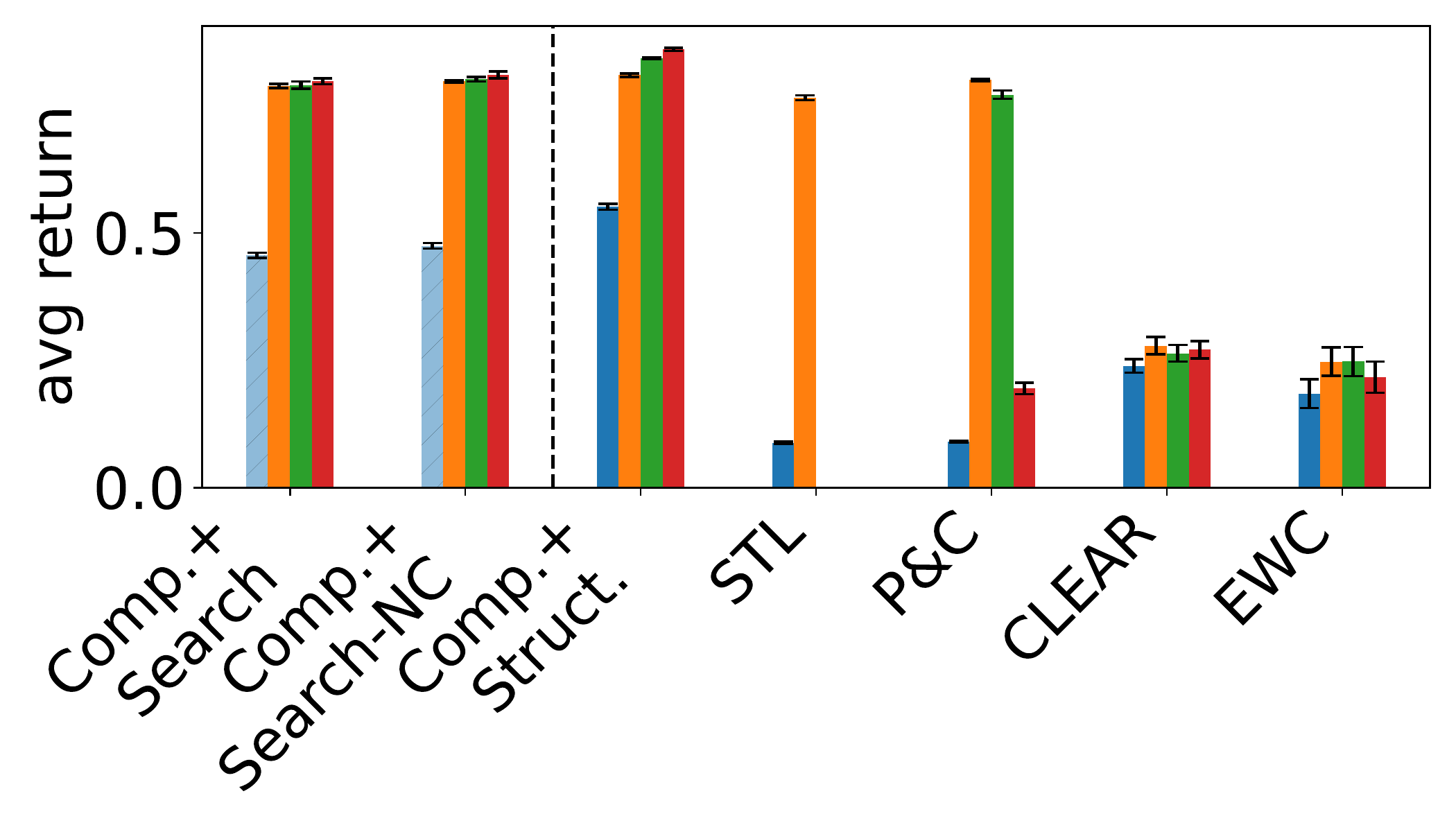}
        \caption{Performance at the various stages}
        \label{fig:compositionalMinigridLLstages}
    \end{subfigure}%
    \caption[Performance of lifelong compositional RL in explicitly compositional discrete 2-D tasks.]{Average returns of STL and lifelong agents on $\numTasks=64$ compositional 2-D discrete tasks. (a) Compositional methods accelerate the training with respect to STL, demonstrating forward transfer. (b) As compositional methods train on more tasks, they improve modules, achieving higher zero-shot performance when combined in novel ways. (c) P\&C also achieves forward transfer, but it forgets how to solve earlier tasks, while compositional methods retain performance---Comp.+Struct. even achieves backward transfer. Comp.+Search performs better than baselines that receive multi-hot task descriptors. ``Zero-shot'' for Comp.+Search (shaded) is after discrete search, which does require data. Shaded regions and error bars represent standard errors across six seeds.}
    \label{fig:compositionalMinigridLL}
 \end{figure}

Figure~\ref{fig:compositionalMinigridLLcurves} shows the average learning curves of the lifelong agents trained on all $\numTasks=64$ possible 2-D tasks. Once again, the averaged curves demonstrate that the variants of \compRL{} achieve forward transfer, accelerating the learning with respect to STL. Note that this acceleration occurred despite \compRL{} using an order of magnitude fewer trainable parameters than STL ($86{,}350$ vs.~$1{,}080{,}320$). Additionally, both \compRL{} methods improved the modules over time, as demonstrated by the trend of increasing zero-shot performance as the agent sees more tasks, shown in Figure~\ref{fig:compositionalMinigridLLzeroshot}. P\&C also learned faster than STL, but as discussed in the following paragraph, P\&C catastrophically forgot how to solve earlier tasks. Other lifelong baselines performed substantially worse than STL, since they are designed to keep the solutions to later tasks close to those of earlier tasks, which fails in compositional settings where optimal task policies vary drastically. 

In these 2-D tasks, lifelong learners should completely avoid forgetting, since there exist models (compositional and monolithic) that can learn to solve all possible tasks (see Figure~\ref{fig:compositionalMinigridMTL}). Figure~\ref{fig:compositionalMinigridLLstages} shows the average performance as each task progressed through various stages: the beginning (zero-shot) and end (online) of assimilation via online training, the accommodation of knowledge into the shared parameters (off-line; for Comp.~and P\&C only), and the evaluation after training on all tasks (final). \compRL{} variants were the only ones that achieved forward transfer without suffering from any forgetting. Moreover, Comp.+Struct. achieved \textit{backward transfer}: improving the earlier tasks' performance after training on future tasks, as indicated by the increase in performance from the off-line to final bars. 

To study the flexibility of \compRL{}, the evaluation also assessed its performance on a random sequence of tasks, without forcing the initial tasks to be composed of distinct components. Figure~\ref{fig:compositionalMinigridLL} shows that this lack of curriculum (Comp.+Search-NC) did not hinder the forward or backward transfer of \compRL{}, since its performance was similar to Comp.+Search.

\begin{figure}[b!]
    \centering
    \begin{subfigure}[b]{0.49\textwidth}
        \includegraphics[trim={0.1cm 0.1cm 0.1cm 0.15cm}, clip,height=0.9cm]{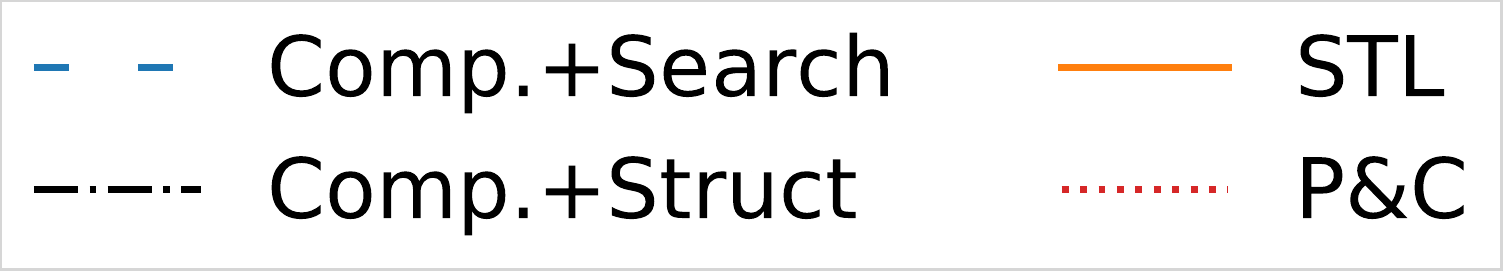}
    \end{subfigure}%
    \hfill
    \begin{subfigure}[b]{0.49\textwidth}
        \centering
        \includegraphics[trim={0.1cm 0.1cm 0.1cm 0.15cm}, clip,height=0.9cm]{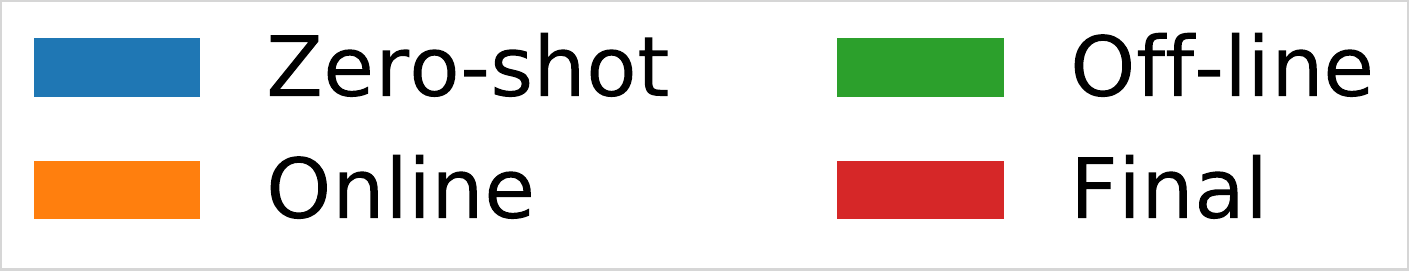}
    \end{subfigure}
    \begin{subfigure}[b]{0.49\textwidth}
        \includegraphics[trim={0.3cm 0.3cm 0.3cm 0.3cm}, clip=True, height=5cm]{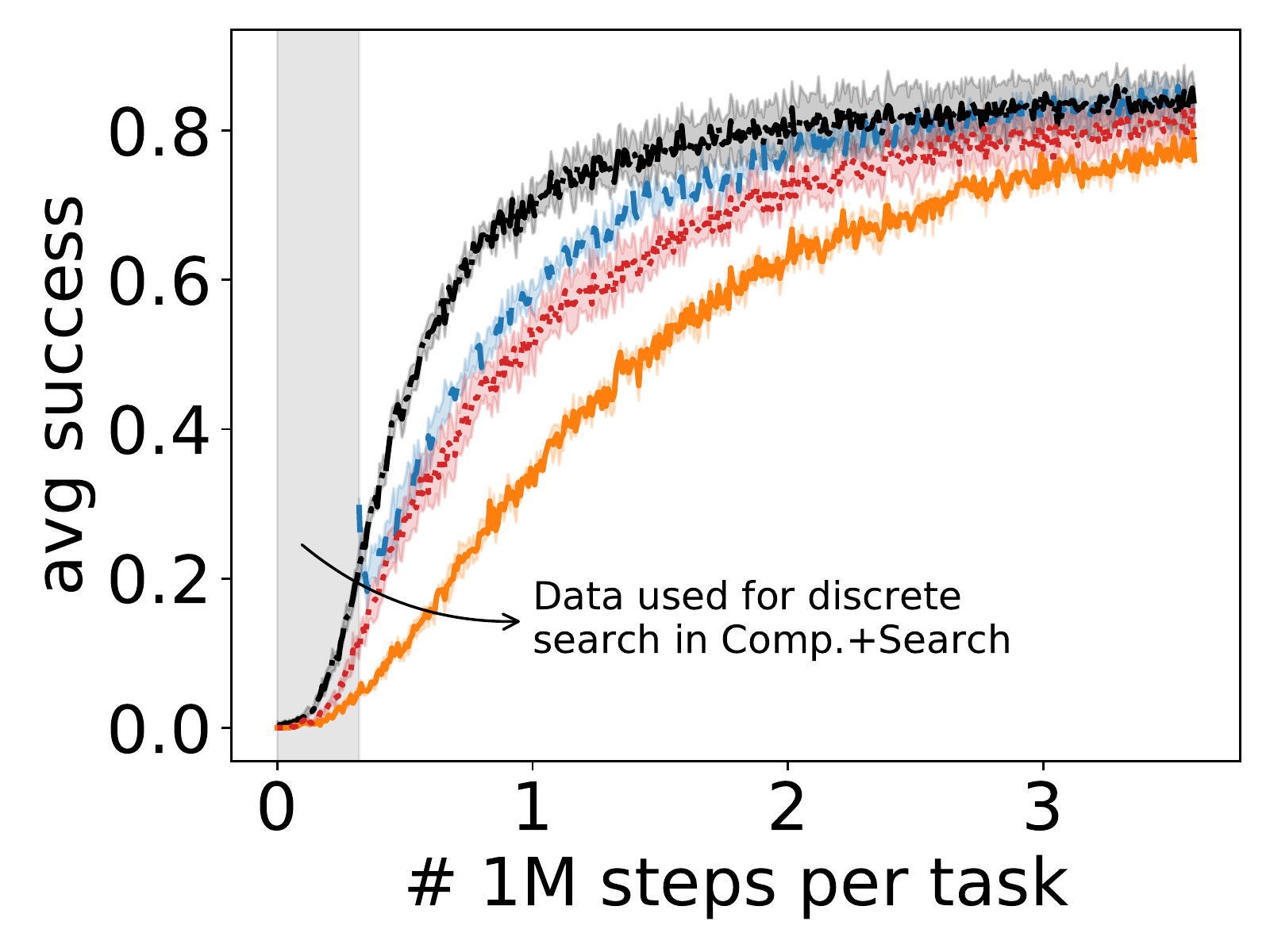}
        \caption{Learning curves}
        \label{fig:compositionalRobosuiteLLcurves}
    \end{subfigure}%
    \hfill
    \begin{subfigure}[b]{0.49\textwidth} 
        \centering
        \includegraphics[trim={0.cm 0.3cm 0.3cm 0.3cm}, clip=True, height=5cm]{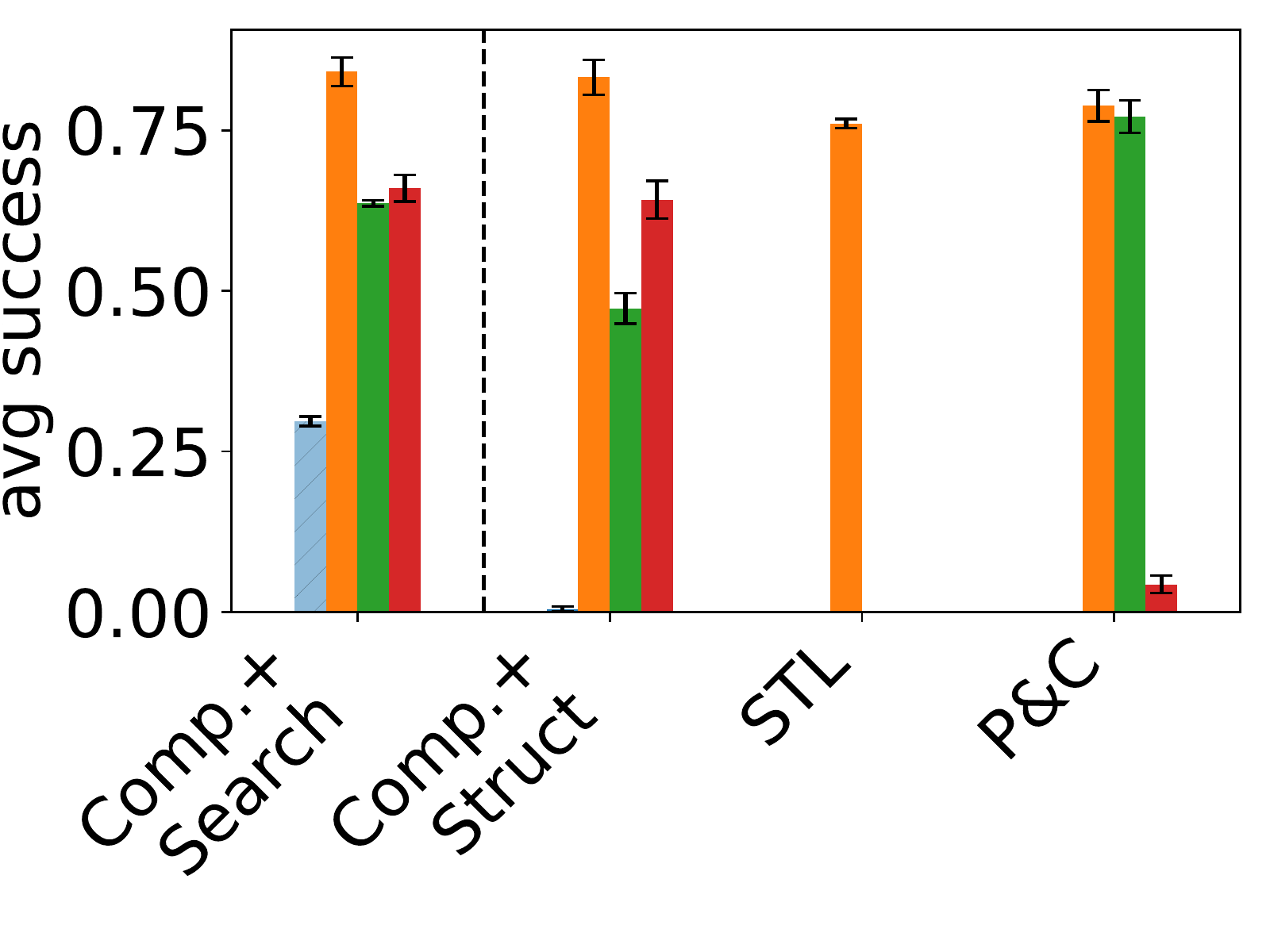}
        \caption{Performance at the various stages}
        \label{fig:compositionalRobosuiteLLstages}
    \end{subfigure}%
    \caption[Performance of lifelong compositional RL on explicitly compositional robotics tasks.]{Average success of STL and lifelong agents on $\numTasks=48$ compositional robot manipulation tasks. (a) Compositional methods again achieve forward transfer. (b) The off-line stage causes a drop in performance, but further training the modules on future tasks achieves backward transfer and partially recovers the lost performance. Shaded regions and error bars represent standard errors across three seeds.}
    \label{fig:compositionalRobosuiteLL}
\end{figure}

\subsection{Lifelong Discovery of Modules on Realistic Robotic Manipulation Tasks}

Having validated that \compRL{} achieves lifelong transfer on the 2-D tasks, an equivalent evaluation tested the agents on the more complex and realistic robotic manipulation suite. The architecture similarly chained one module of each type for the obstacle, object, and robot arm, and each such module was an MLP. The evaluation repeated this experiment over three trials, varying the random seed controlling the parameter initialization and the ordering over tasks. In this evaluation, PPO would output overconfident actions when initialized from the existing modules directly, which limited the agent's ability to learn proficient policies. Therefore, the experiments used a modified version of PPO which permitted successful training of all tasks at the cost of often inhibiting zero-shot transfer (see Section~\ref{sec:HyperParametersRL}). The learning curves in Figure~\ref{fig:compositionalRobosuiteLLcurves} show that all lifelong agents learned noticeably faster than the base STL agent, and compositional methods were fastest, despite using an order of magnitude fewer trainable parameters than STL ($165{,}970$ vs.~$1{,}040{,}304$).  Figure~\ref{fig:compositionalRobosuiteLLstages} also shows that the off-line stage led to a decrease in performance. However, like in the 2-D domain, training on subsequent tasks led to backward transfer, partially recovering the performance of the earlier tasks as the agent learned future tasks. As expected, P\&C was incapable of retaining knowledge of past tasks, leading to massive catastrophic forgetting.

\subsection{Ablative Tests on Compositional Domains}
\label{sec:Ablations}

One natural question that arises when evaluating methods with various algorithmic and architectural building blocks is: which of these constituent parts are crucial for the obtained performance? This section empirically validates the design choices behind \compRL{}.

The first study verified that the modules required to solve the discrete 2-D tasks are diverse. Results from  Section~\ref{sec:zeroShotEvaluation} showed that the discrete 2-D tasks are truly compositional: if the agent discovers a good set of modules, it can recombine them and reuse them to solve unseen tasks. However, it is possible that some of these components are essentially the same. For example, perhaps the module for learning to reach the green target could be replaced with the module to reach the red target. This would severely limit the usefulness of the evaluations. As a sanity check, the evaluation tested the effect of using the incorrect module for evaluation on a task, separated by type of module. Figure~\ref{fig:moduleDiversity} reveals that using the incorrect static object modules leads to a small (but noticeable) drop in performance, while using the incorrect target object or agent module leads to a drastic drop to nearly random performance. This validates that the modules differ substantially.

\begin{figure}[t!]
    \centering
    \begin{subfigure}[b]{0.52\textwidth}
        \centering
        \includegraphics[trim={0.1cm 0.1cm 0.1cm 0.15cm}, clip,height=0.9cm]{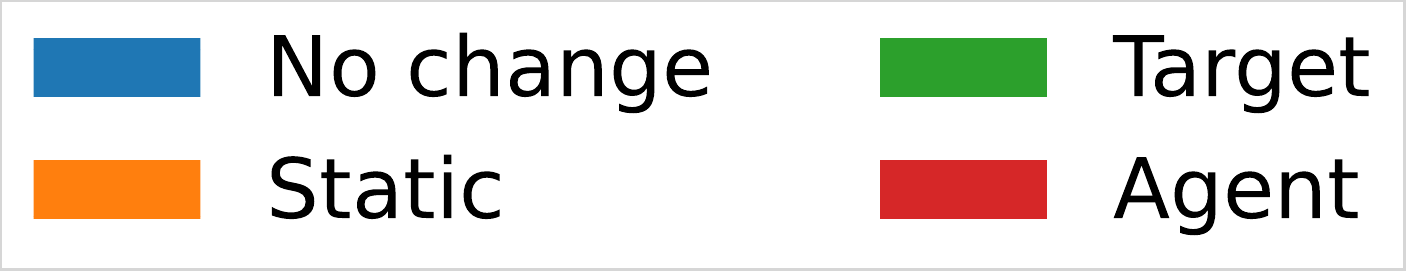}
    \end{subfigure}\\    
    \begin{subfigure}[b]{0.52\textwidth}         
        \includegraphics[trim={0.3cm 0.5cm 0.3cm 0.3cm}, clip, height=5.1cm]{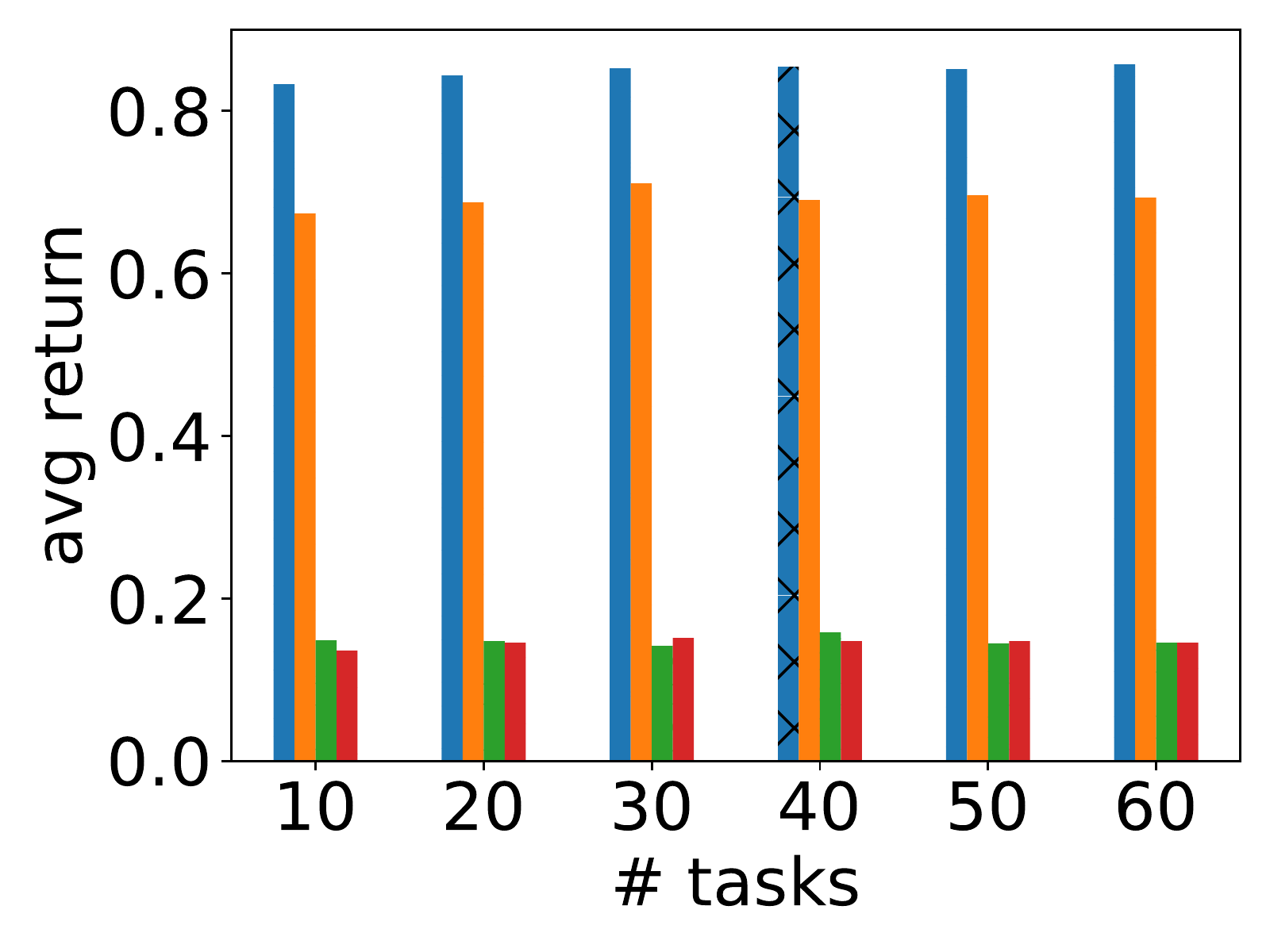}
    \end{subfigure}%
    \caption[Ablative analysis on explicitly compositional discrete 2-D RL tasks: module diversity.]{Ablative analysis on the diversity of modules in discrete 2-D tasks. Performance of the modular MTL agent when constructing the policy with incorrect modules, demonstrating that modules specialize to solve their assigned subproblem.}
    \label{fig:moduleDiversity}
\end{figure}

\begin{figure}[b!]
    \centering
    \begin{subfigure}[b]{0.49\textwidth}
        \centering
        \raisebox{0.2cm}{\includegraphics[trim={0.1cm 0.1cm 0.1cm 0.15cm}, clip,height=0.45cm]{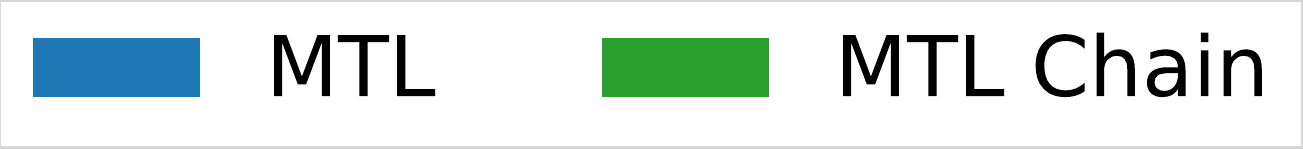}}
    \end{subfigure}\\    
    \begin{subfigure}[b]{0.49\textwidth}         
        \includegraphics[trim={0.5cm 0.5cm 0.3cm 0.3cm}, clip, height=5.1cm]{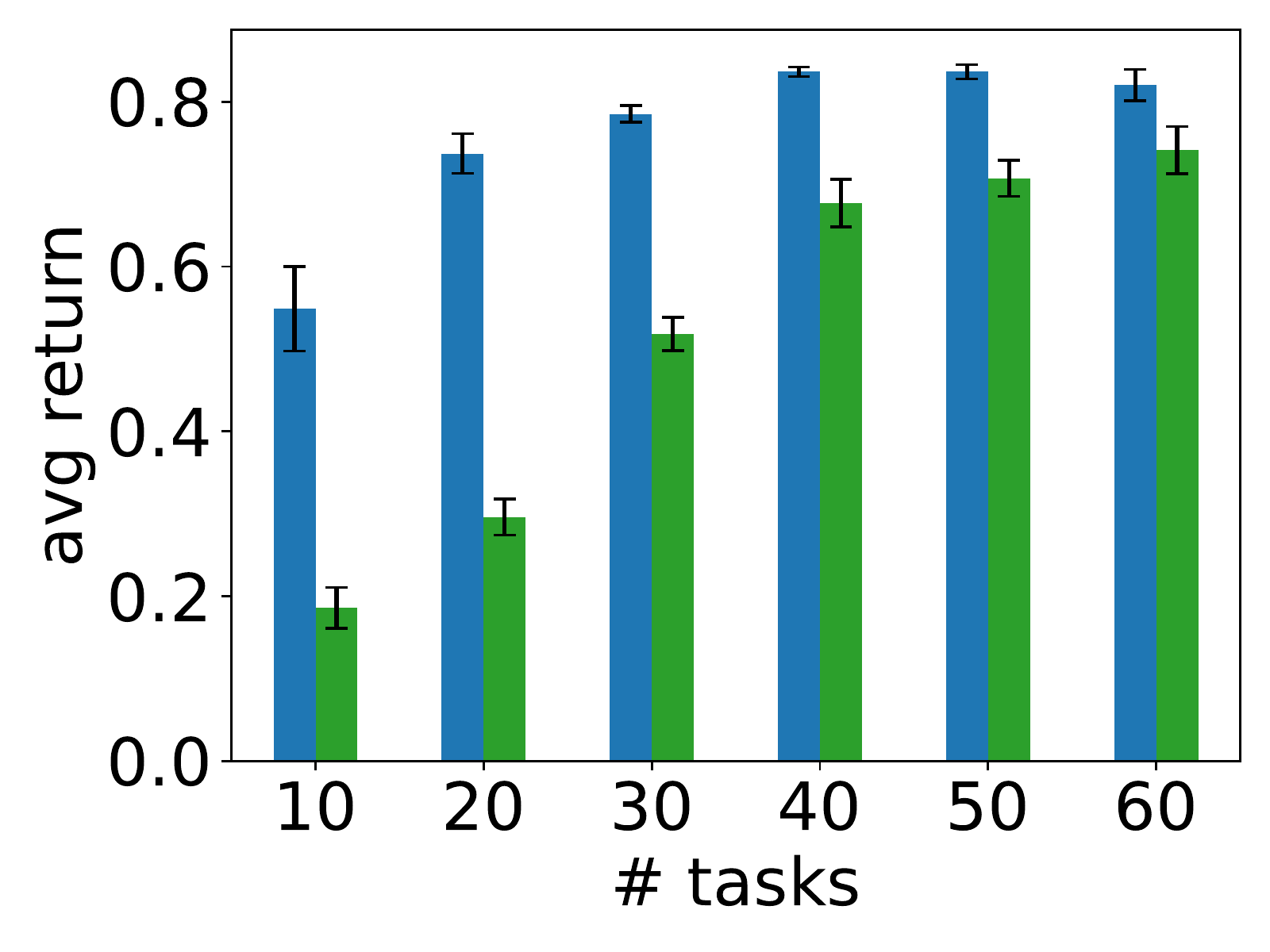}
    \end{subfigure}\\
    \caption[Ablative analysis on explicitly compositional discrete 2-D RL tasks: modular architecture.]{Ablative analysis on the modular architecture design for discrete 2-D tasks. Generalization to unseen combinations with the proposed modular architecture (MTL) and with a standard chained modular architecture (MTL Chain), showing that modules require far less data to generalize if they are trained on decomposed state representations. Error bars denote standard errors across six seeds.}
    \label{fig:zeroShotAblation}
\end{figure}
The next study analyzed the architecture choice described in Section~\ref{sec:ModularArchitectureRL}. A more natural choice of architecture, which was considered early in the development cycle of \compRL{}, is a simple module chaining, where the input passes entirely through a first module, whose output passes to the next module, and so on. This is in contrast to the proposed architecture, where the input factors into task components and feeds separately into distinct modules. The next experiment repeated the evaluation of Section~\ref{sec:zeroShotEvaluation} with a purely chained architecture, and Figure~\ref{fig:zeroShotAblation} shows the obtained results. This chained architecture cannot generalize nearly as quickly as the proposed modified architecture. This is intuitively reasonable; consider for example the first module, which is in charge of static object detection. In the chained architecture, this module is further in charge of passing information to subsequent modules about all remaining task components, whereas the proposed architecture only needs to focus each module on the relevant component, without distractor features from other task components. One alternative view of the same problem is that, to achieve zero-shot generalization, the output of all modules at one depth needs to be compatible with all modules at the next depth. This requires that the output spaces of all modules are compatible. One way to encourage this compatibility is to restrict the inputs to the modules to only the relevant task information, as achieved by the proposed architecture.

\begin{figure}[b!]
    \centering
    \begin{subfigure}[b]{0.53\textwidth}
        \includegraphics[trim={0.1cm 0.1cm 0.1cm 0.15cm}, clip,height=0.9cm]{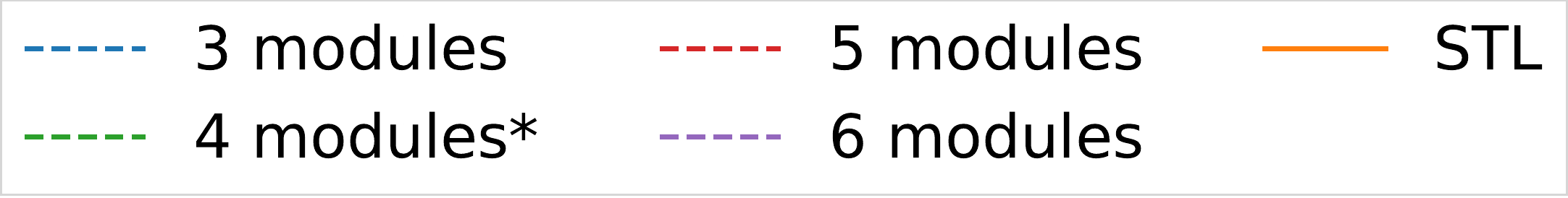}
    \end{subfigure}%
    \begin{subfigure}[b]{0.47\textwidth}
        \centering
        \includegraphics[trim={0.1cm 0.1cm 0.1cm 0.15cm}, clip,height=0.9cm]{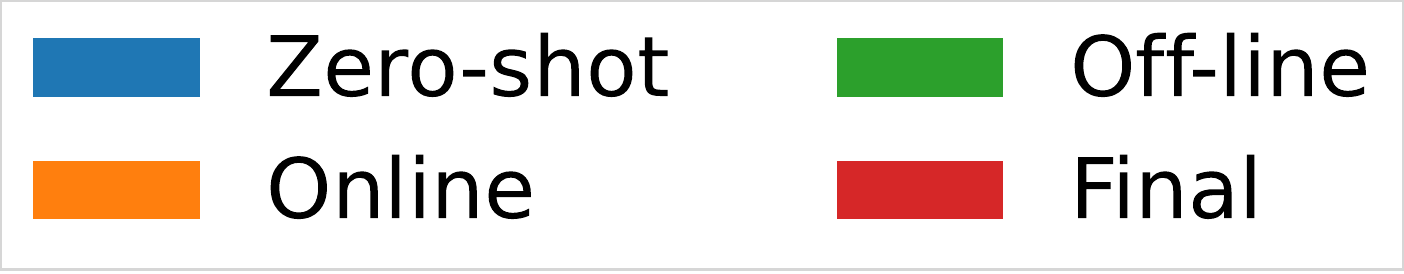}
    \end{subfigure}\\
    \begin{subfigure}[b]{0.4\textwidth}
        \includegraphics[trim={0.3cm 0.5cm 0.3cm 1.3cm}, clip=True, height=4.5cm]{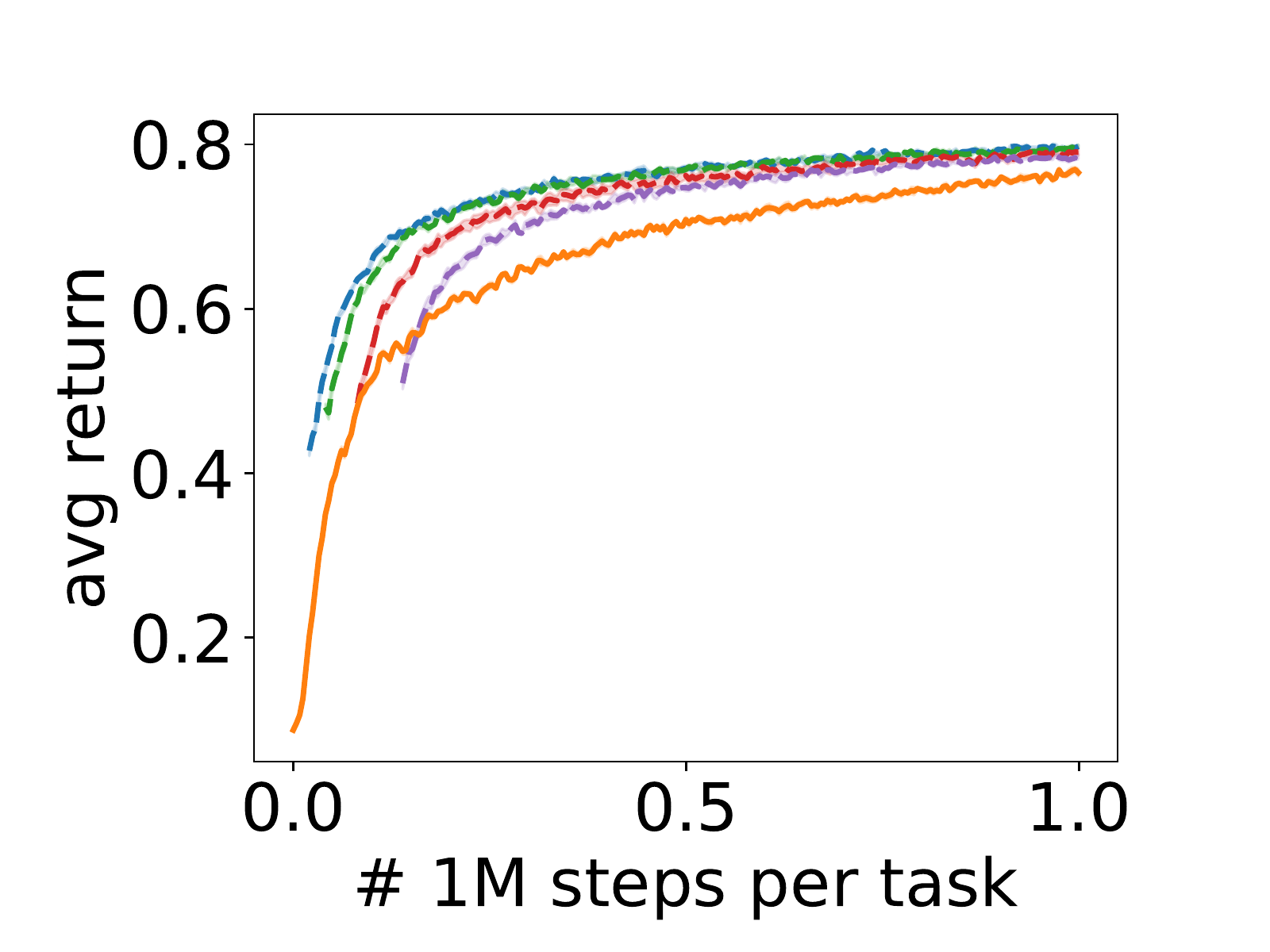}
        \caption{Learning curves}
    \end{subfigure}%
    \begin{subfigure}[b]{0.6\textwidth} 
        \centering
        \includegraphics[trim={0.7cm 0.5cm 0.3cm 0.3cm}, clip=True, height=4.5cm]{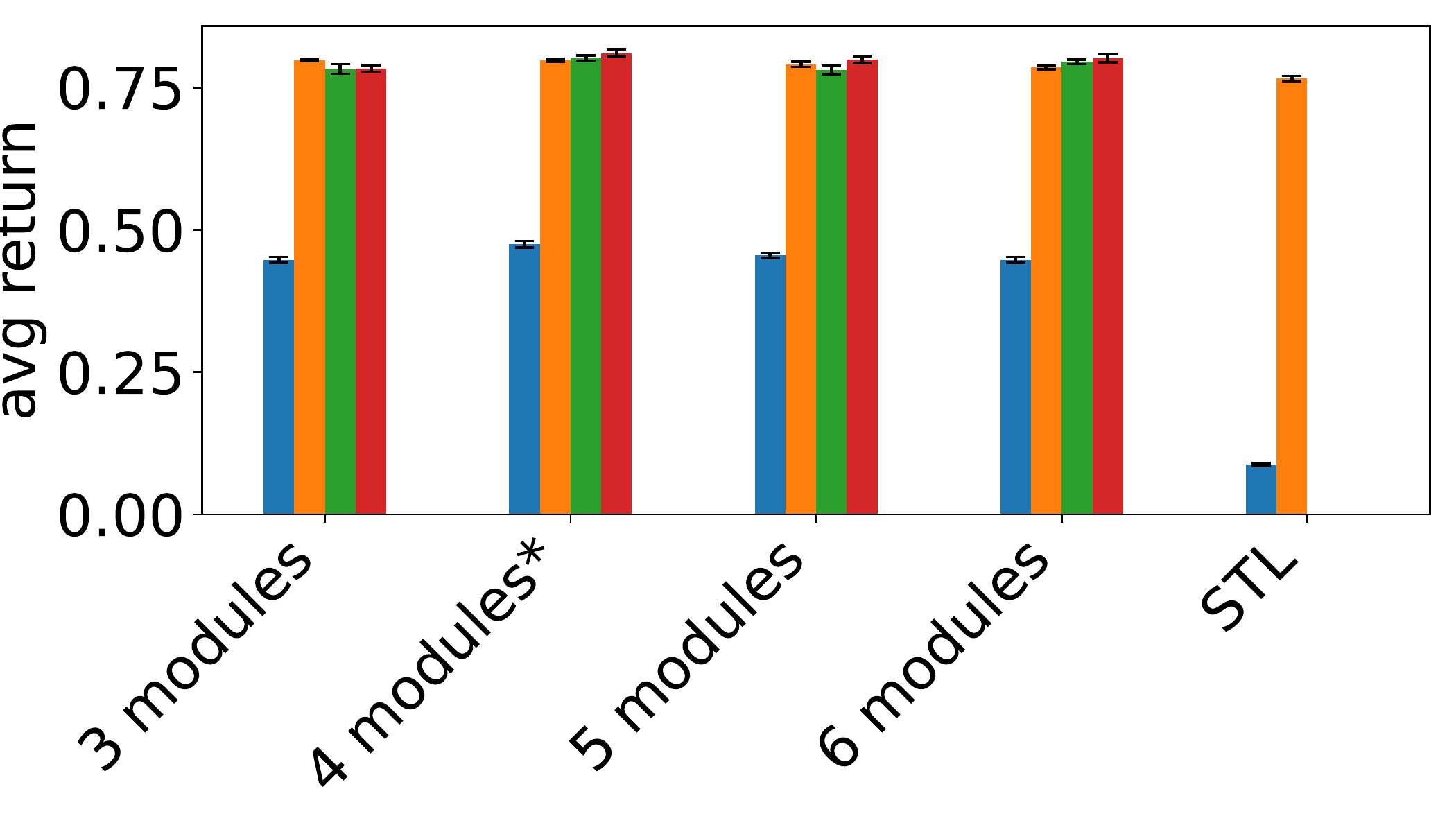}
        \caption{Performance at the various stages}
    \end{subfigure}%
    \caption[Ablative analysis on explicitly compositional discrete 2-D RL tasks: number of modules.]{Ablative analysis on the number of modules for discrete 2-D tasks. Average returns of Comp.+Search-NC with varying number of modules. The number of modules has only minor effects on the overall performance of \compRL{}. Shaded regions and error bars represent standard errors across six seeds. \\\**$4$ modules is the original (correct) value from Section~\ref{sec:resultsMiniGrid}.}
    \label{fig:numModAblationMiniGrid}
\end{figure}

To test how \compRL{} performs if the architecture uses more or fewer modules of each type than task components, the evaluation repeated the experiments on  the 2-D domain with varying numbers of neural components, using completely random task sequences (i.e., no curriculum). Figure~\ref{fig:numModAblationMiniGrid} shows that \compRL{} is remarkably insensitive to the number of modules, performing well with a range of choices.

So far, the experiments have shown that modular architectures enable improved performance in the compositional tasks. Since all the lifelong baselines in the evaluation use monolithic architectures, one could think that the improved performance of \compRL{} might come solely from the use of a better architecture. To verify that this is not the case, a learner trained the proposed modular architecture using CLEAR. As shown in Figure~\ref{fig:learningCurveAblation}, while the performance of CLEAR indeed improved, it fell substantially short of matching the performance of \compRL{}. Since CLEAR uses an objective that closely mimics off-line RL, but does so online during training of new tasks, this highlights that the advantage of \compRL{} comes primarily from the separation of the learning process into assimilation and accommodation stages. 

\begin{figure}[t!]
    \centering
    \begin{subfigure}[b]{0.49\textwidth}
        \centering
        \includegraphics[trim={0.1cm 0.1cm 0.1cm 0.15cm}, clip,height=0.9cm]{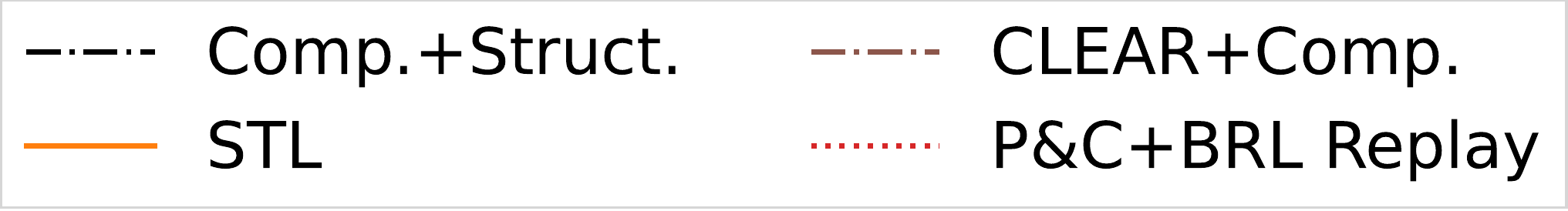}
    \end{subfigure}
    \begin{subfigure}[b]{0.49\textwidth}
        \centering
        \includegraphics[trim={0.1cm 0.1cm 0.1cm 0.15cm}, clip,height=0.9cm]{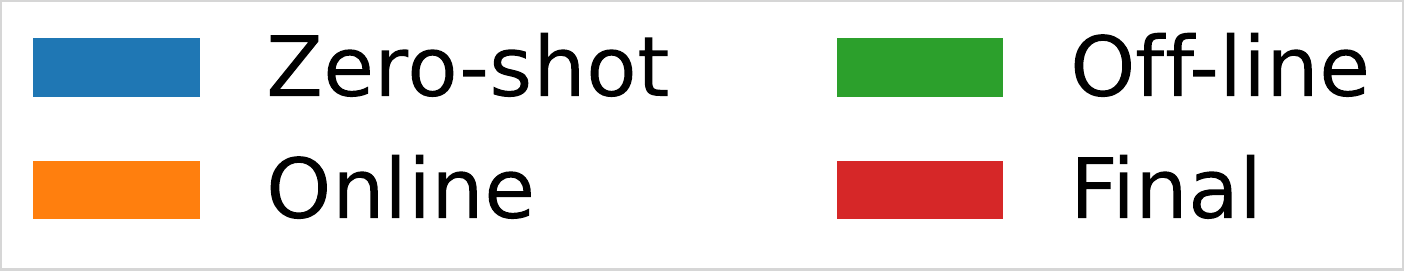}
    \end{subfigure}\\
    \begin{subfigure}[b]{0.49\textwidth}         
        \includegraphics[trim={0.cm 0.5cm 1.3cm 1.3cm}, clip,height=5.1cm]{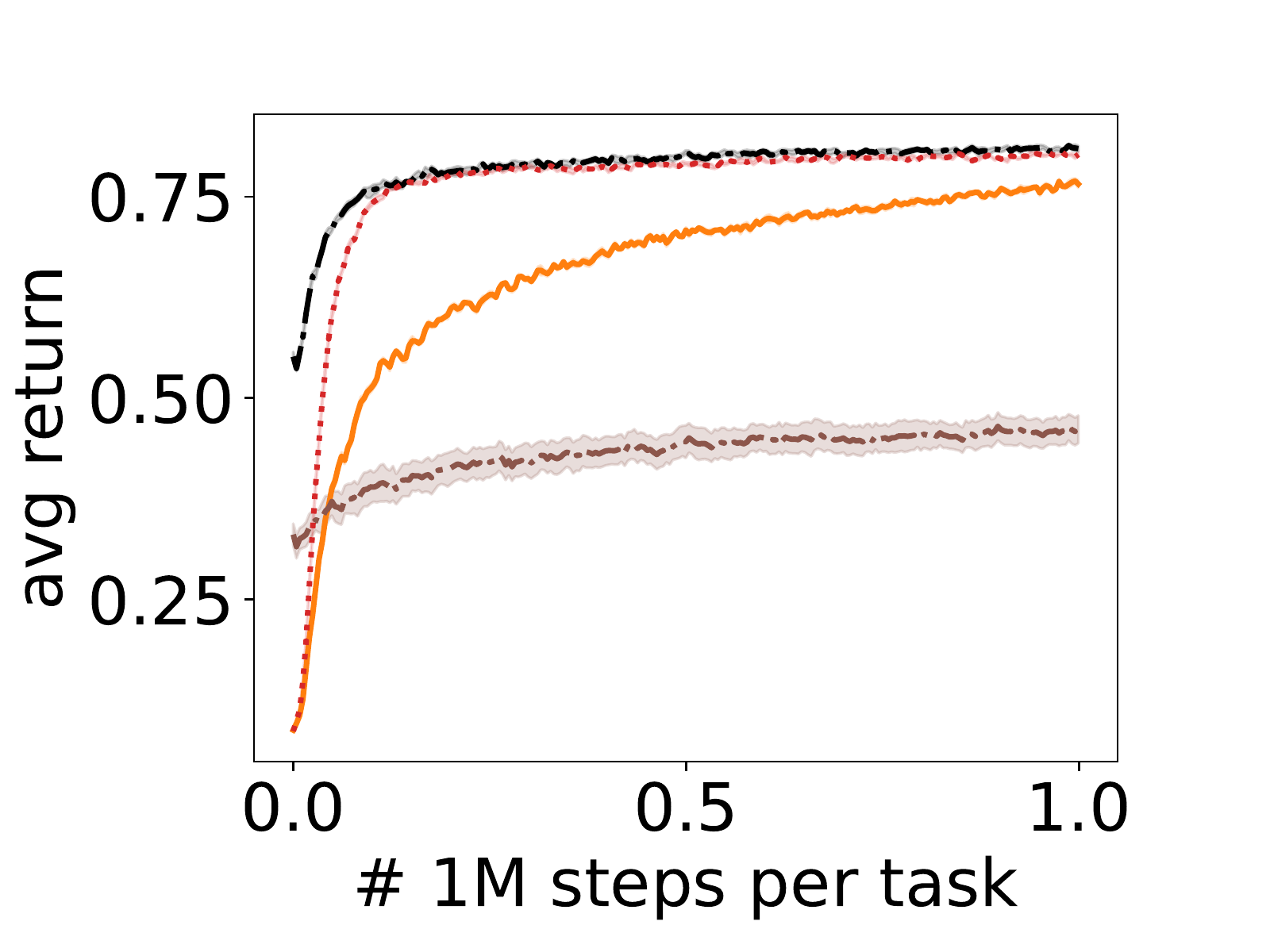}
        \caption{Learning curves}
        \label{fig:learningCurveAblation}
    \end{subfigure}%
    \begin{subfigure}[b]{0.49\textwidth}         
        \includegraphics[trim={0.7cm 1.3cm 0.3cm 0.3cm}, clip,height=5.1cm]{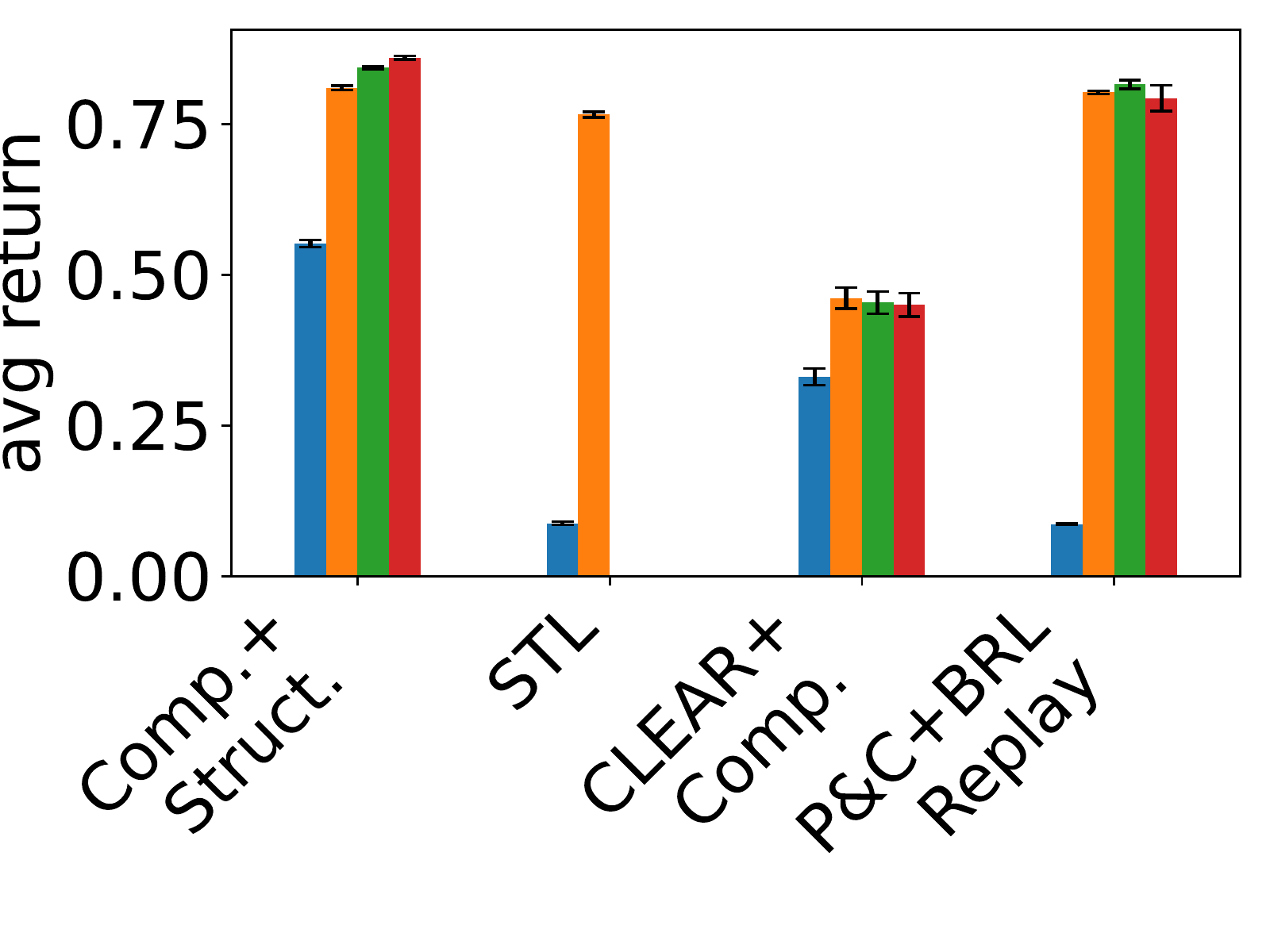}
        \caption{Performance at the various stages}
        \label{fig:barchartAblation}
    \end{subfigure}\\
    \caption[Ablative analyses on explicitly compositional discrete 2-D RL tasks: modularity and off-line RL.]{Ablative analyses on the algorithmic design of \compRL{} on discrete 2-D tasks. Performance of the proposed modular architecture trained via CLEAR is much lower than with \compRL{}, and the proposed off-line RL mechanism to avoid forgetting drastically improves the performance of P\&C, almost entirely preventing forgetting. Shaded regions and error bars denote standard errors across six seeds.}
    \label{fig:compositionalMinigridAblations}
\end{figure}

Another major contribution of this dissertation was the use of off-line RL to avoid forgetting. To analyze the effect of this choice, P\&C was trained replacing EWC in its compress stage with off-line RL. Notably, as shown in Figure~\ref{fig:barchartAblation}, this almost entirely suppressed the effect of forgetting. Moreover, this led to even-better forward transfer of P\&C. This result stresses the fact that avoiding forgetting is necessary not only for retaining performance of earlier tasks, but also for accumulating knowledge that better transfers to future tasks. 

\begin{figure}[t!]
    \centering
    \begin{subfigure}[b]{0.49\textwidth}
        \includegraphics[trim={0.1cm 0.1cm 0.1cm 0.15cm}, clip,height=0.9cm]{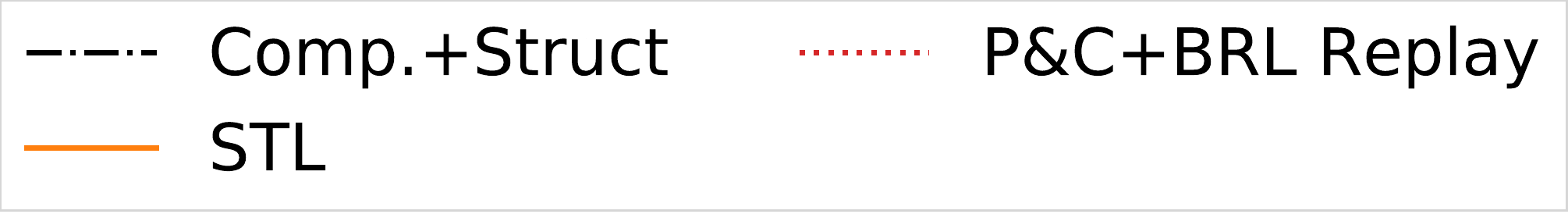}
    \end{subfigure}%
    \hfill
    \begin{subfigure}[b]{0.49\textwidth}
        \centering
        \includegraphics[trim={0.1cm 0.1cm 0.1cm 0.15cm}, clip,height=0.9cm]{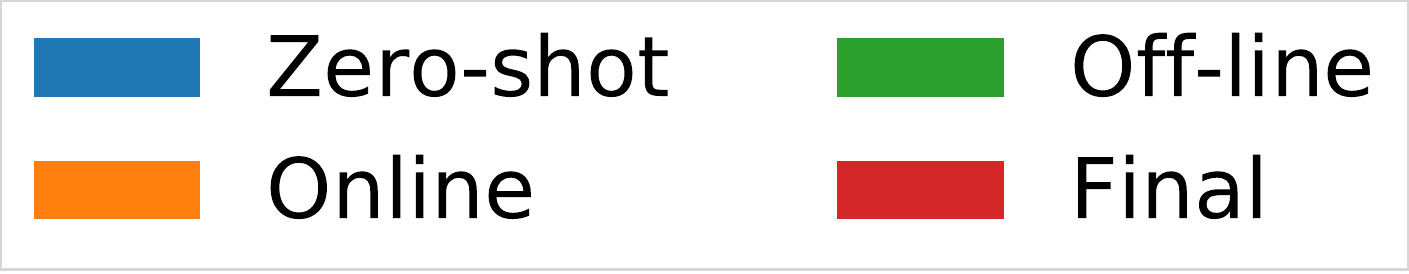}
    \end{subfigure}
    \begin{subfigure}[b]{0.49\textwidth}
        \includegraphics[trim={0.3cm 0.5cm 0.35cm 0.3cm}, clip=True, height=5.1cm]{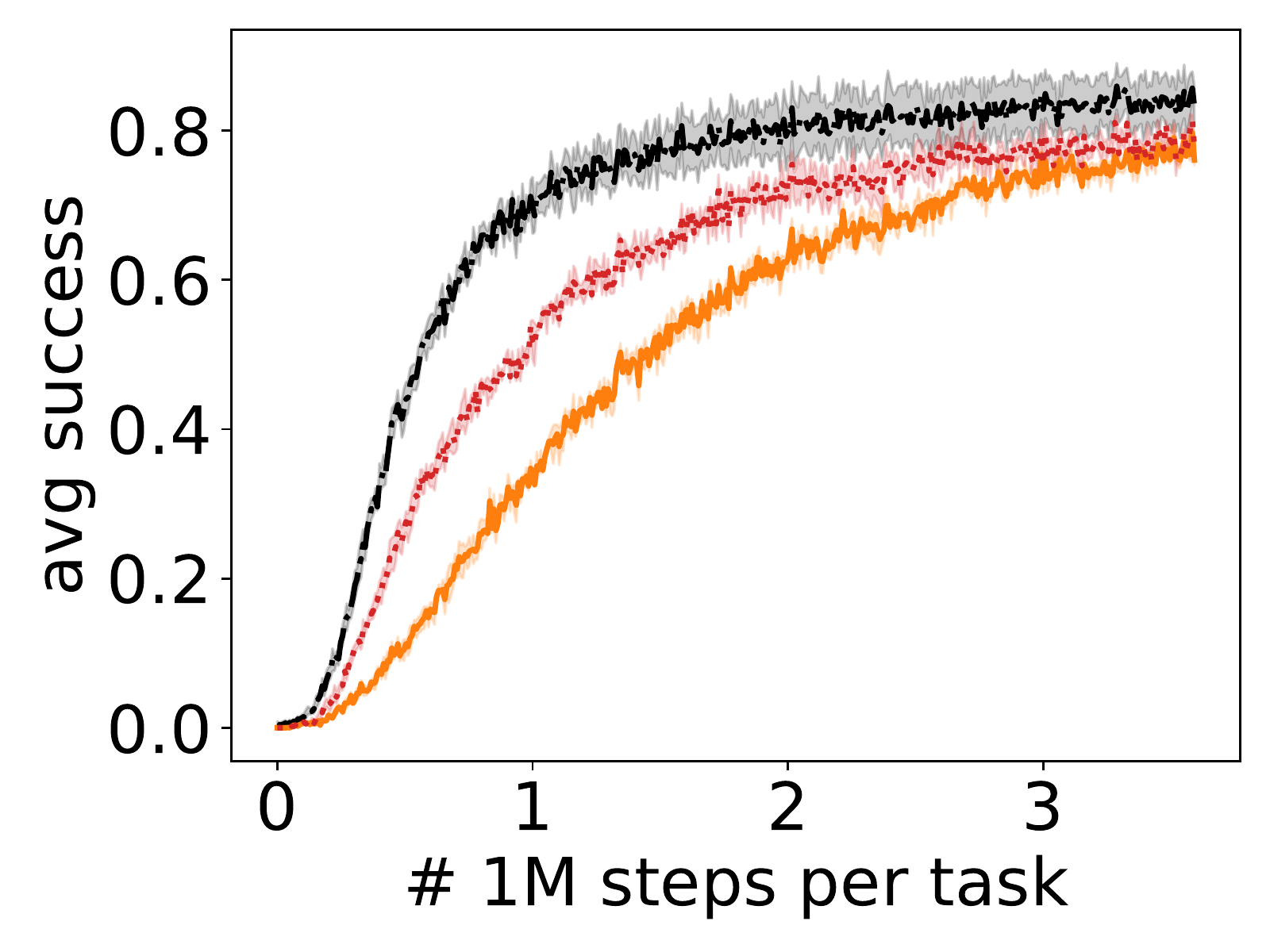}
        \caption{Learning curves}
    \end{subfigure}%
    \hfill
    \begin{subfigure}[b]{0.49\textwidth} 
        \centering
        \includegraphics[trim={0.05cm 0.5cm 0.3cm 0.3cm}, clip=True, height=5.1cm]{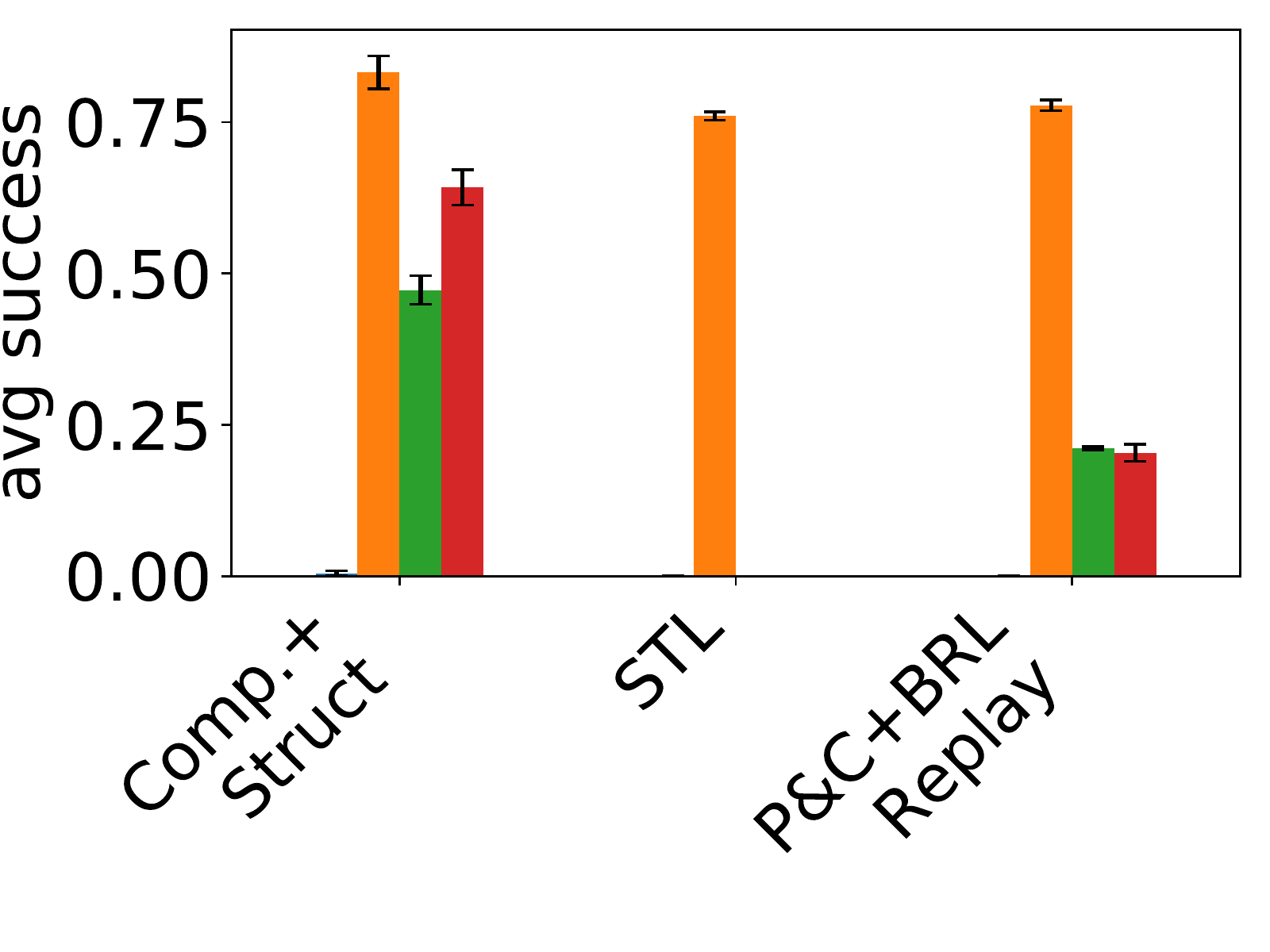}
        \caption{Performance at the various stages}
    \end{subfigure}%
    \caption[Ablative analysis on explicitly compositional robotics RL tasks: off-line RL.]{Ablative analysis on the algorithmic design of \compRL{} on robotic manipulation tasks. P\&C using off-line RL replay avoids forgetting, but cannot fully incorporate new knowledge due to the monolithic structure. Shaded regions and error bars denote standard errors across three seeds.}
    \label{fig:compositionalRobosuiteAblations}
\end{figure}

The evaluation repeated the off-line RL ablative test on the compositional robotic manipulation tasks, yielding that the training of P\&C was accelerated beyond that of STL, and off-line RL avoided forgetting. However, in this harder setting, the monolithic structure was insufficient to express policies for all tasks and therefore off-line and final performance were substantially degraded, as shown in Figure~\ref{fig:compositionalRobosuiteAblations}.

\section{Summary}
This chapter formulated the problem of lifelong compositional RL, presenting a graph formalism along with intuitive compositional domains, which are extended in Chapter~\ref{cha:Benchmark}. It then evaluated the flexibility of the framework for lifelong learning of compositional structures presented in Chapter~\ref{cha:Framework} by instantiating it with two powerful lifelong RL algorithms. 

The first proposed method, \lpgftw{}, linearly combines model parameters as the components, and leverages this simplified structure to apply a closed-form update to components in the accommodation stage. This choice also enables deriving theoretical proofs that \lpgftw{} converges to the approximate MTL objective, despite operating completely online. Empirically, \lpgftw{} enables RL agents to quickly learn to solve new tasks by leveraging knowledge accumulated from earlier tasks and does not suffer from catastrophic forgetting, and therefore permits learning a large number of tasks in sequence. This first method can be viewed as an improvement over the popular PG-ELLA algorithm~\citep{bouammar2014online}, which had not previously been applicable to highly complex and diverse RL problems like studied in this dissertation, especially in the evaluation on the Meta-World benchmark. 

The second algorithm developed in this chapter, \compRL{}, learns neural compositional models that more closely match the problem formulation of compositional learning. The evaluation demonstrated that \compRL{} is capable of leveraging accumulated components to more quickly learn new tasks without forgetting earlier tasks and while enabling backward transfer. As a core component, \compRL{} uses off-line RL as a mechanism to avoid forgetting. This proved to be a strong choice for avoiding forgetting more broadly in lifelong RL methods with multistage training processes.

Two of the primary limitations of these methods are the reliance on task indicators $(t)$ to reconstruct individual task policies via the $\st$'s, and the assumption that the world is stationary and tasks are drawn \iid{} One possible way to address the former challenge is to assume that each new batch of experiences comes from a new task, as prior work has done~\citep{nagabandi2018deep}. However, note that this would require a small amount of retraining at evaluation time, since the agent would need to discover which task it is currently facing. To address the latter challenge of nonstationarity, Chapter~\ref{cha:NonStationary} introduces an extension of \compRL{} that specifically handles environments with shifting components.  

In the specific case of \compRL{}, another limitation is its scalability with respect to the number of modules, requiring to attempt all possible combinations for the discrete search step. While the experiments showed that this is feasible even on relatively long sequences of $\numTasks=64$ tasks, specialized heuristics to reduce the search space would be needed if searching over many more possible combinations. 
Additionally, while modular lifelong RL improved performance in the robotic experiments, this improvement was relatively modest. In particular, it was not possible to combine existing modules for zero-shot transfer. Achieving lifelong zero-shot compositional generalization in such a complex RL setting remains an open problem. 

One of the key contributions of this line of work is reducing the amount of experience required by RL agents to achieve proficiency at a multitude of tasks. The methods presented here are some of the first plausible solutions to solving highly diverse sets of RL tasks in a lifelong setting. Research in this direction that further reduces the amount of experience required to learn proficient policies would enable RL training on systems where experience is expensive, such as training real robotic systems or learning policies for medical treatments. In these settings, training RL policies has been impractical to date, but could potentially have a large positive impact by discovering policies superior to those conceivable by human experts with domain knowledge.

\biblio

\chapter{Extension of Lifelong Composition to Nonstationary Environments}
\label{cha:NonStationary}

\renewcommand\qedsymbol{$\blacksquare$}

\section{Introduction}
One key limitation of the approaches presented so far is that they assume that the agent operates in a stationary environment, where it must perform a variety of tasks, but the tasks are drawn from a fixed distribution. This implies that the agent must remember how to solve any of the tasks seen so far, because they are all equally likely under its observed distribution. However, in practical situations it is often the case that the environment changes over time, and tasks that were likely at an earlier time are no longer relevant in the present. 

This chapter presents a compositional view of the problem of nonstationary lifelong learning, where individual components of the environment vary over time. Following these compositional assumptions, the chapter presents a variant of the general-purpose framework that tackles the nonstationary lifelong learning problem. Intuitively, given the separation of the learning into stages, only the accommodation stage requires accounting for the nonstationarity in the environment, since this stage combines knowledge about the current task with that of earlier tasks into the components. 

As a proof of concept, this chapter proposes a lifelong RL method that extends \compRL{} from Chapter~\ref{cha:RL} to handle the nonstationary lifelong RL problem. To account for the changes in the environment, during the adaptation stage the agent performs off-line RL replay only over previous tasks that are likely under the current task distribution. The chapter discusses several methods for detecting such tasks, and evaluates these methods on nonstationary variants of the discrete 2-D tasks of Chapter~\ref{cha:RL}.

\section{Related Work on Nonstationary Lifelong Learning}
\label{sec:RelatedWorkNonstat}

Few investigations have considered the nonstationary lifelong learning setting, where not only the \textit{data} distribution changes across tasks, but also the \textit{task} distribution changes over time. 
Some works treat the lifelong learning problem itself as nonstationary learning. As opposed to considering a sequence of tasks, these works consider a stream of data, and the objective of the learner is to learn a model that works well across all seen data. In practice, experiments to evaluate these approaches still often instantiate nonstationarity via different tasks, but the task boundaries are hidden from the agent. Typical approaches in the supervised~\citep{riemer2018learning} and reinforcement~\citep{nagabandi2018deep} settings include change-point detection and clustering techniques. However, note that these methods deal with a single level of distributional change---the change in the data distribution--- and do not consider higher-level changes in the task distribution.
  
Other formulations assume that the temporal changes are smooth and predictable. Consequently, if the agent learns to predict future changes, then it can adapt its model preemptively to handle future tasks. One early work provided theoretical performance bounds under two settings: the tasks are dependent but identically distributed, or the distribution changes over time in a consistent fashion~\citep{pentina2015lifelong}. Another common assumption is that the distribution is Markovian: the next environment depends only on the current (and not the previous) environment. Meta-RL works have proposed approaches to learn a mechanism that predicts the future policy to deal with the modified environment~\citep{al-shedivat2018continuous,clavera2018learning,xie2020deep}.

When dealing with nonstationary tasks, it is often desirable to forget irrelevant information that might be harmful for future learning. Domain adaptation has studied this setting, where the data distribution changes over time and the environment does not require the agent to perform well on past data~\citep{kurle2020Continual,lao2020continuous,kumar2020understanding}. In a lifelong setting, the agent must still retain relevant information for future transfer and to perform well on previous tasks that remain relevant. The approaches in this chapter deal with this issue via compositional learning mechanisms that target the learning to specific components of the environment that have changed over time.

Recent efforts have attempted to unify approaches by mathematically formulating the nonstationary learning problem in terms of appropriate performance metrics, which make specific assumptions about the nature of the distribution shifts~\citep{caccia2020onlinefast,ren2021wandering}. 
This chapter considers a complementary problem, whereby the distribution shift is modular, but the variations in each component are arbitrary. Combining these two lines, by making assumptions about how each component's distribution changes over time, remains open for future work.

\section{The Compositional Nonstationary Lifelong Learning Problem}
\label{sec:LifelongNonstationaryProblem}

Approaches for nonstationary lifelong learning typically assume that the environment changes over time as a whole. If an earlier task is unlikely under the current distribution, the agent down-weights the entirety of that task in the objective---or, more extremely, completely discards the task for any future learning. However, if certain aspects of the environment remain unchanged over time, it would be desirable to leverage all the knowledge about those aspects from every previous task to better generalize to future tasks. Similarly, it is possible that different aspects of the environment vary at different rates or in different patterns (e.g., cyclical), making it difficult for monolithic nonstationary learners to track all the changes.

As an illustrative example, consider the compositional suite of robotic manipulation tasks from Chapter~\ref{cha:RL}. If the agent executes these tasks on physical systems, the robot motors would likely degrade over time, each at its own distinct rate. Meanwhile, the visual sensors in charge of detecting the object locations would be affected by cyclical changes in lighting conditions throughout the day. In this example, object modules for tasks executed in morning light could leverage object knowledge from all tasks in the past executed in morning light, regardless of the status of the robot motors. On the other hand, robot modules would only be able to leverage information of recent tasks where the motors had similar levels of degradation, but regardless of whether the agent learned the tasks in the morning or at night.

\begin{figure}
    \centering
    \includegraphics[width=0.95\linewidth,clip,trim=2.2in 2.2in 2.2in 2.2in]{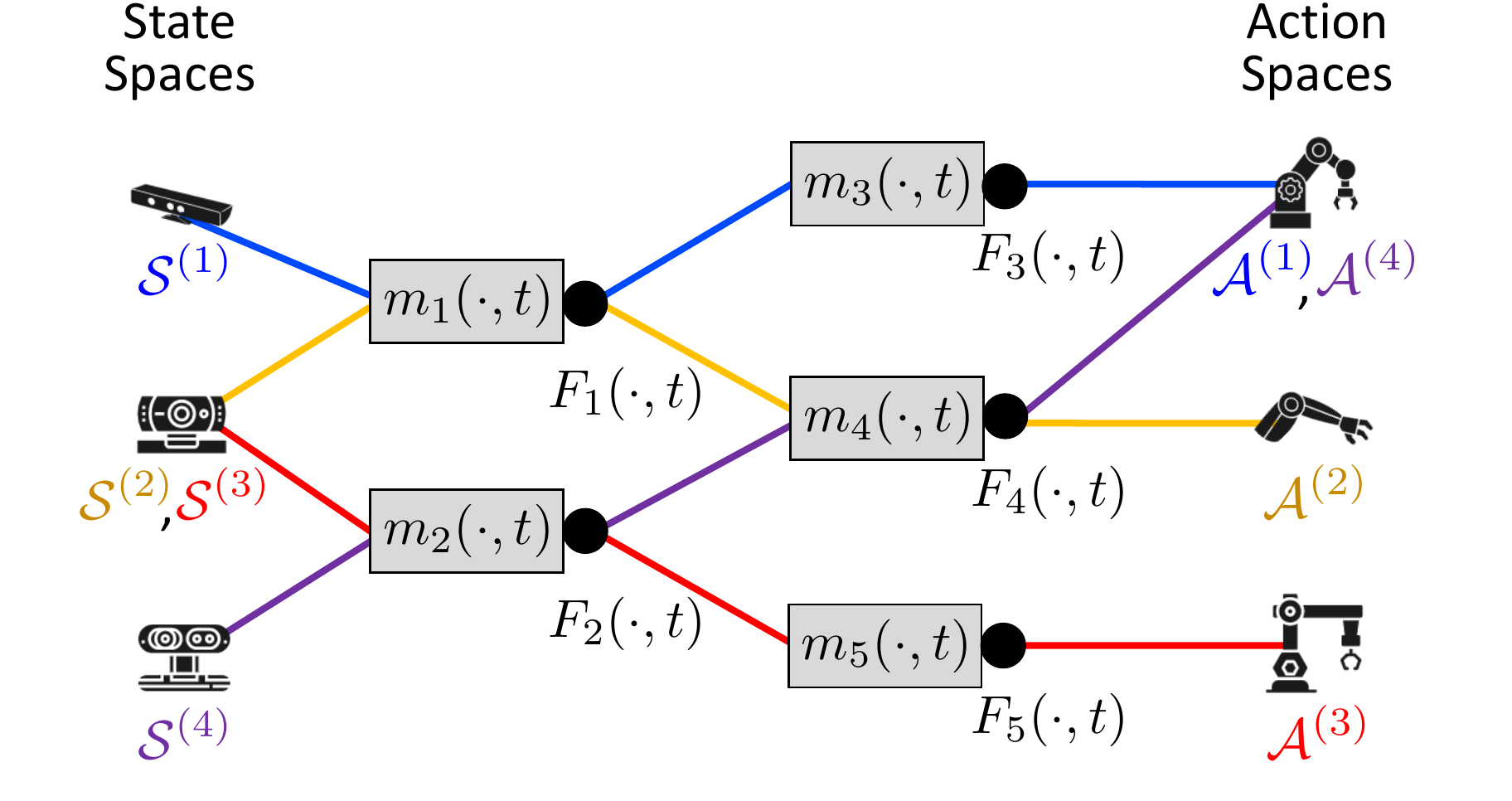}
    \caption[Compositional nonstationary RL problem graph.]{Compositional nonstationary RL problem graph. Like in the stationary counterpart, nodes represent state spaces, action spaces, and subproblems. In particular, the subproblems are time-dependent, inducing nonstationarity in the environment and requiring the agent to track their changes over time.}
    \label{fig:compositionalRLGraphNonstat}
\end{figure}

The problem of compositional nonstationary lifelong RL\footnote{The supervised variant follows trivially from the RL definition. Since the algorithm and evaluation dealt with the RL setting, the problem definition for the supervised setting was not explicitly included here.} closely follows that of compositional lifelong RL of Chapter~\ref{cha:RL}. The one difference, as illustrated in Figure~\ref{fig:compositionalRLGraphNonstat}, is that the subproblems---and consequently, the modules---that constitute the tasks are \emph{time dependent}. In particular, the problem assumes that the time dependence is individual to each module, which permits separately considering their evolution over time.

The goal of the agent is then to not only learn the compositional structures over a lifelong sequence of tasks, as in the problem considered in Chapter~\ref{cha:RL}, but also to track the changes in each of the components over time. Ideally, the learner should be able to leverage any knowledge from previous tasks about the components that are still relevant under the current distribution of the environment.

\section{\NonstatRL{}: Nonstationary Lifelong Reinforcement Learning via Composition}
\label{sec:LifelongNonstationaryRLAlgo}

Lifelong compositional learning, as described in previous chapters, permits handling this more challenging nonstationary lifelong learning problem. To demonstrate the feasibility of discovering functional modules under this setting, this section presents \NonstatRL{}, an adapted version of \compRL{} that explicitly deals with nonstationarity. 

The compositional architecture is a hard modular network, exactly like that described in Chapter~\ref{cha:RL}. Concretely, the state factors into module-specific state components, which are passed only to the relevant modules. The modules themselves are chained such that the output of one acts as (part of) the input to the next.

While much of \NonstatRL{} follows exactly the same process of \compRL{} for discovering these modules, with the only differences arising during the accommodation stage, the following paragraphs describe all stages of the algorithm for consistency and clarity. 

\paragraph{Initialization} In the same way as \compRL{}, \NonstatRL{} trains a disjoint set of neural modules on each of the initial tasks until all modules are initialized. 

\paragraph{Assimilation} Recall that the purpose of the assimilation stage in the RL setting is two-fold: 1)~to find the correct set of components to solve the current task and 2)~to explore the space of solutions to the current task. \NonstatRL{} follows \compRL{} by splitting the assimilation stage into two substages. The first substage selects the correct set of modules, either by manually setting them from domain knowledge or by conducting an exhaustive search over the possible choices. The second substage performs RL training  on a copy of the selected modules to maximize performance solely on the current task.

\paragraph{Accommodation} The accommodation stage updates the shared modules with any relevant information from the current task. The mechanism developed in Chapter~\ref{cha:RL} for \compRL{} retrains the modules via off-line RL with data from past tasks. In typical (stationary) lifelong learning scenarios, this enables the agent to perform well on the current task, continue to perform well on past tasks, and consolidate knowledge to transfer to future tasks. However, in the nonstationary setting, it is possible that some of the knowledge acquired in the past is outdated. If the agent trains the modules to simultaneously perform well on the current data as well as on data from an outdated distribution, current and future performance would be compromised. Consequently, the agent should select data from previous tasks that remains relevant under the current distribution. In particular, if the agent is capable of detecting \emph{which} components of the environment have changed, then all the data from tasks that share \emph{other} components with the current task remains valid. Only data from tasks that share the modified components would need to be appropriately filtered. The key question therefore becomes how to identify which of the previous tasks' modules match the present state of the current task's modules. This dissertation considered the following three possibilities:
\begin{itemize}
    \item {\em Oracle}\hspace{2em} In some situations, the agent might know exactly which past modules match the current state of the world. For example, if the changes in the environment are due to varying daylight, the agent could use time-stamps to infer this information. Moreover, evaluating an oracle-based approach permits validating whether the nonstationarity can be captured compositionally, regardless of the agent's ability to detect it.
    \item {\em Loss weighting}\hspace{2em} More generally, the agent must automatically detect changes and determine which past tasks' data to leverage. Intuitively, this can be achieved by measuring how much the current task's solution departs from previous tasks' solutions. One approach is to measure the difference in the loss of each previous task induced by training on the current task. Specifically, after the agent assimilates the $t$-th task, it uses the trained policy $\pi^{(t)}$  to compute the off-line loss $\ell_t^{(t')}$ over the replay data from each task $\Taskt[t^\prime]: t^\prime\in\{1,\ldots,t\}$, and computes the baseline performance on the current task with a random policy $\ell_{\mathrm{rand}}^{(t)}$. Then, the agent weights each previous task's data by how much its loss has departed from its original value:
    \begin{align}
        \nonumber \Delta_\ell(t^\prime, t) =& \frac{\ell_{t}^{(t^\prime)} - \ell_{t^\prime}^{(t^\prime)}}{\ell_{\mathrm{rand}}^{(t^\prime)} - \ell_{t^\prime}^{(t^\prime)}}\\
        w_{t^\prime} =& \begin{cases} 
            1 -  \Delta_\ell(t^\prime, t) \quad&\text{if } 0 < \Delta_\ell(t^\prime, t) < 1\\
            0 \quad&\text{if } \Delta_\ell(t^\prime, t) \geq 1\\
            1 \quad&\text{if } \Delta_\ell(t^\prime, t) \leq 0 \enspace.
        \end{cases}
    \end{align}
    Note that this mechanism can only behave entirely compositionally if a single component of the environment is changing at a time. Then, if the agent knows which components are candidates for shifting, it can detect \textit{how} those modules are changing. Otherwise, if multiple modules could be changing, the loss weighting scheme would not be able to differentiate between modules within a task. In consequence, it would treat all modules as varying jointly and behave as a standard monolithic nonstationary approach. Going back to the robot example, the agent would not be able to distinguish changes in lighting conditions from motor degradation by just examining the loss, and consequently would down-weight any previous task whose light \textit{or} motor components has changed. In contrast, if only the light component varies over time, the robot could automatically discover which previous tasks' lighting conditions match the current task's via loss weighting. Moreover, if a task with an outdated light component shares a \textit{different} component (e.g., the robot component) with the current task, then data from this task could still be used. In the experiments, the agent only computed weights for modules that were changing and that were shared with the current task.
    \item {\em Representational distance weighting}\hspace{2em} To circumvent this limitation, one alternative is to instead measure the distance in representational space. This method can trivially be applied separately to each module of previous tasks, effectively leveraging the compositionality of the environment changes. Similar to the previous approach, after the agent assimilates the $t$-th task, the agent uses the policy $\pi^{(t)}$  to compute the mean of the representation of each module $i$ on each task $\mu_{t}^{(t^\prime,i)}$, and computes the baseline representational mean on the current task using a random policy $\mu_{\mathrm{rand}}^{(t,i)}$. The agent then weights the loss for each previous task by how much the representation has shifted from its original value:
    \begin{align}
        \nonumber \Delta_{\mu_i}(t, t^\prime) =&  \frac{\|\mu_{t}^{(t^\prime,i)} - \mu_{t^\prime}^{(t^\prime,i)}\|_2^2 }{\|\mu_{\mathrm{rand}}^{(t^\prime,i)} - \mu_{t^\prime}^{(t^\prime,i)}\|_2^2}\\
        w_{t^\prime,i} =&\begin{cases} 
            1 - \Delta_{\mu_i}(t, t^\prime) \quad&\text{if } 0 <  \Delta_{\mu_i}(t, t^\prime)  < 1\\
            0 \quad&\text{if }  \Delta_{\mu_i}(t, t^\prime)  \geq 1\\
            1 \quad&\text{if }  \Delta_{\mu_i}(t, t^\prime)  \leq 0 \enspace.
        \end{cases}
    \end{align}
    To make comparisons fair, the evaluation also informed the representational-distance approach of which modules were candidates for changing, and the agent obtained the overall weight for each task by multiplying the weights for all modules $i$ that were shared with the current task $w_{t^\prime}=\prod_{i\in \pi_{t}\cap\pi_{t^\prime}} w_{t^\prime, i}$ (slightly abusing the notation of $\pi$). 
\end{itemize}

\section{Experimental Evaluation}
\label{sec:ExperimentalEvaluationNonstat}

The purpose of this experimental evaluation was to demonstrate that compositionality indeed permits capturing changes to individual aspects of the environment, improving the overall performance of lifelong learners in nonstationary settings. In particular, the goal was to study the benefits of selectively retraining on past tasks based on the estimated shift of their components shared with the current task. 
For this, the evaluation tested the approaches presented in Section~\ref{sec:LifelongNonstationaryRLAlgo} on a variety of settings with different nonstationarity assumptions. The evaluation in this section repeated each experiment over six trials, varying the random seed that controls the parameter initialization and the ordering over tasks. 

\subsection{Nonstationary Evaluation Settings}

The evaluation considered three different conditions on the nonstationarity of the environment. This section describes all such conditions and the rationale behind them.

\paragraph{Cyclical changes} A single element of the environment changes cyclically throughout the agent's lifetime. This is akin to periodic changes like lighting conditions from day to night or weather patterns over seasons. The main implication of these types of changes is that at some point in time, in order to perform well on the current task distribution, the agent must disregard knowledge about tasks that were learned under different conditions. However, this knowledge is still useful for future tasks that the agent might encounter once the distribution reverts to an earlier state. Consequently, it is useful for the agent to keep all of the data it has encountered in memory (even data that appears irrelevant now) in case it becomes relevant in the future. The specific experiment conducted under this setting applied a horizontal perturbation to the agent's perceived target object locations, and the magnitude and direction of this perturbation cycled from $-3$ to $+3$ and back in steps of $1$. For simplicity, the environment applied this change equally to all colors of target objects.

\paragraph{Linear degradation} A single element of the environment changes over time, but the change is progressive instead of cyclical. This is evocative of a robot's motor power degrading over time. In this setting, the agent will never be required to perform a task under older states of the distribution, and therefore in principle it could safely discard data from tasks that have become obsolete. However, even if the agent keeps the data stored in memory, it would still be capable of adapting to the up-to-date distribution as long as it does not select any of the data from the outdated environment. The environment implemented this type of change as a decrease in the agent's likelihood of successfully executing a specific action (\texttt{turn\_left}, in this case), and applied it equally to all choices of agent dynamics.

\paragraph{Multiple elements} While single-element changes permit analyzing the effects of nonstationarity and compositionality on lifelong learning agents, many real-world settings exhibit temporal changes in multiple dimensions simultaneously. In particular, \NonstatRL{} permits addressing settings in which each element changes in different temporal patterns. As one example of such situations, the evaluation conducted an experiment where the environment applied both the cyclical and linear examples above jointly.

\subsection{Baselines} 

The experiments compared \NonstatRL{} against two baselines:
\begin{itemize}
    \item \textit{STL}, which does not share any knowledge across tasks.
    \item \textit{\compRL{}}, which assumes that nothing about the environment changes over time and modules can be perfectly reused across all tasks. 
\end{itemize}
Compositional agents received the ground-truth task decomposition, in order to isolate the problem of nonstationarity from that of discovering the compositional structure. In addition, the environment informed nonstationary agents of which elements of the environment were candidates for changing (e.g., target modules in the case of cyclical changes), and the oracle knew exactly which previous tasks matched the current task's distribution. 

\subsection{Results}

\begin{figure}[t!]
    \centering
    \begin{subfigure}[b]{\textwidth}
        \centering
        \includegraphics[trim={0.1cm 0.1cm 0.1cm 0.15cm}, clip,height=0.395cm]{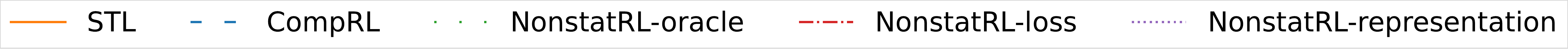}
    \end{subfigure}\\
    \begin{subfigure}[b]{0.35\textwidth}
        \includegraphics[trim={0.8cm 0.3cm 1.8cm 1.2cm}, clip, height=4cm]{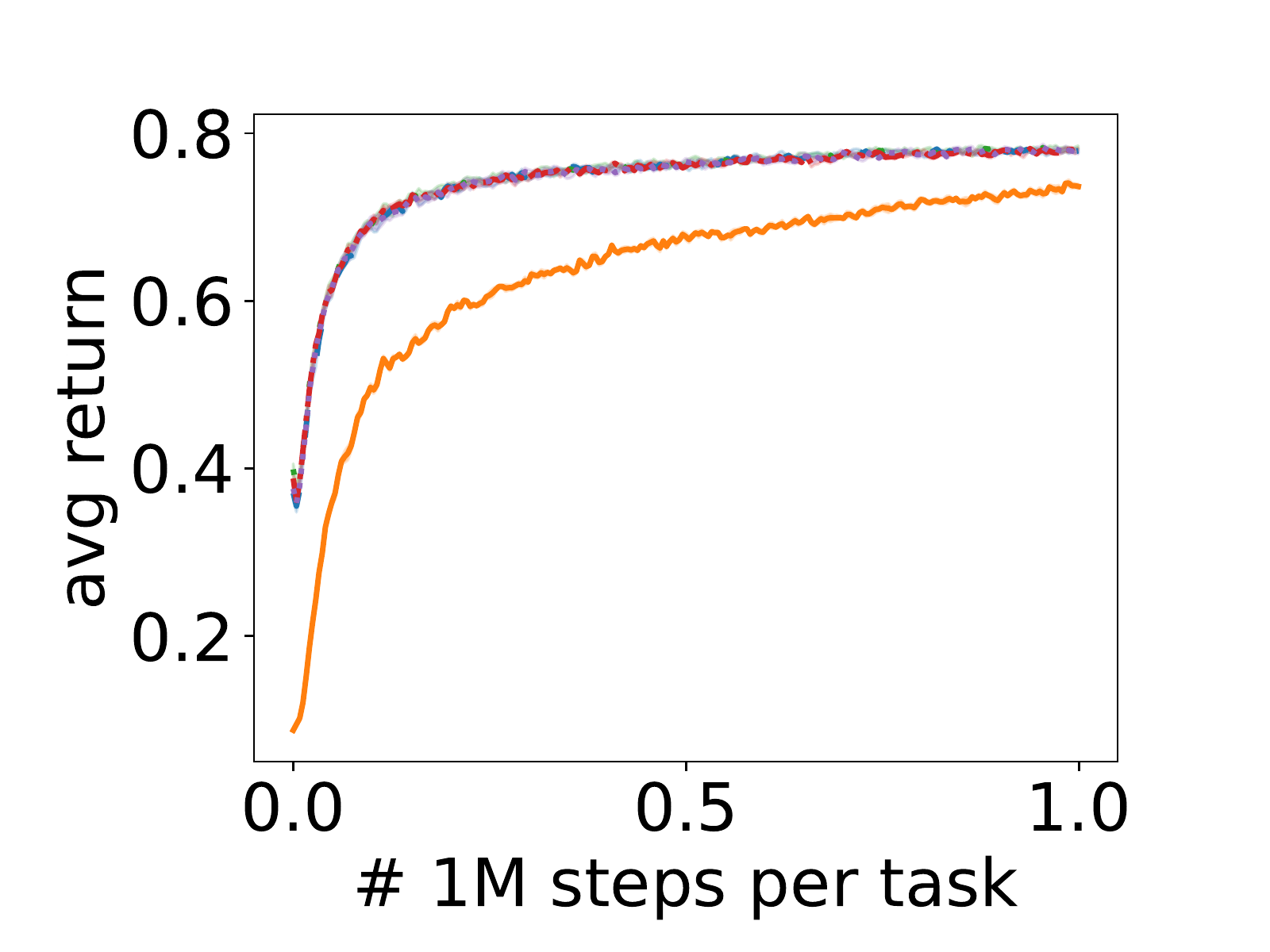}
        \caption{Cyclical}
    \end{subfigure}%
    \begin{subfigure}[b]{0.335\textwidth}
        \includegraphics[trim={1.6cm 0.3cm 1.8cm 1.2cm}, clip,height=4cm]{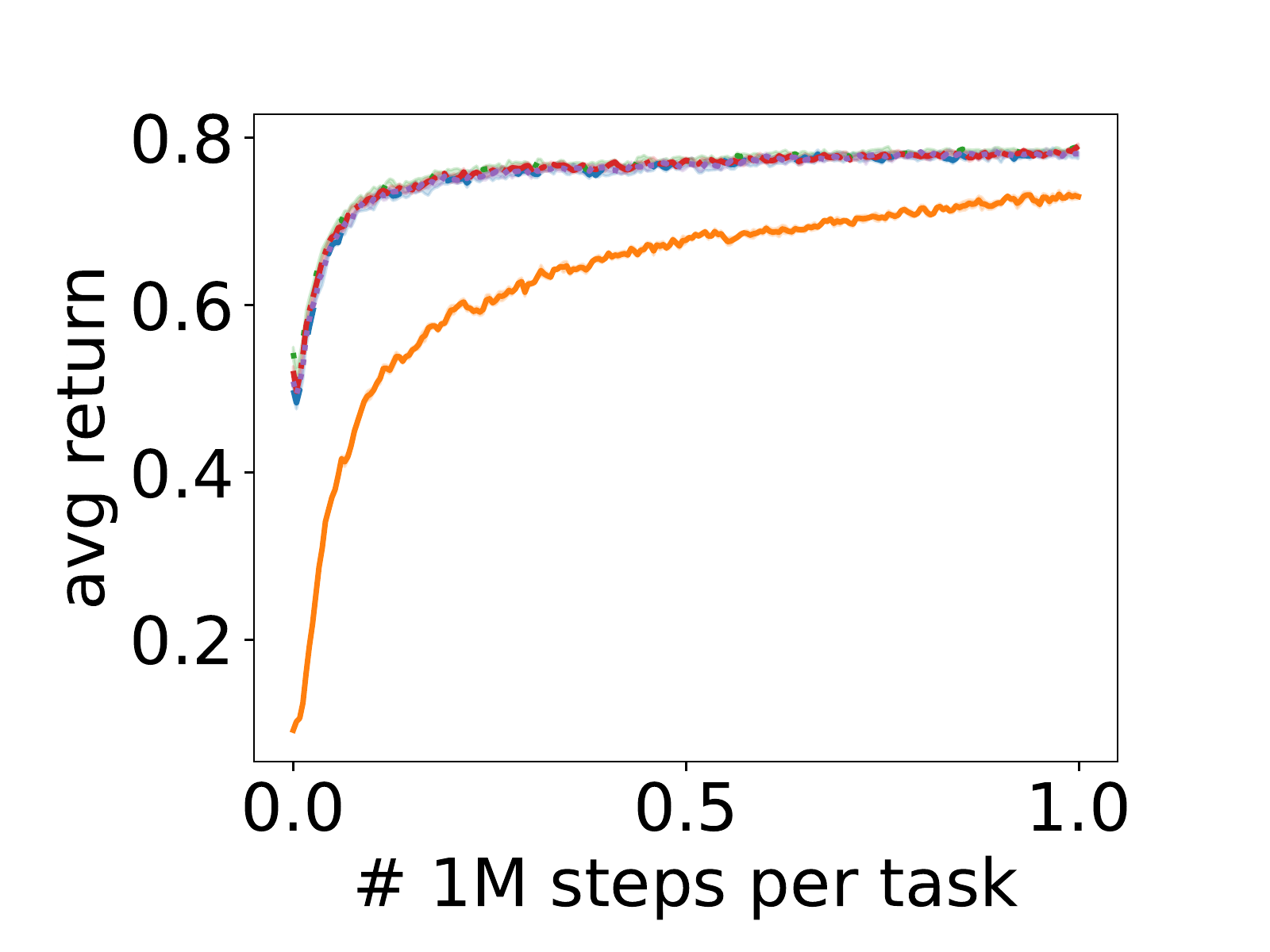}
        \caption{Linear}
    \end{subfigure}%
    \begin{subfigure}[b]{0.335\textwidth}
        \includegraphics[trim={1.6cm 0.3cm 1.8cm 1.2cm}, clip,height=4cm]{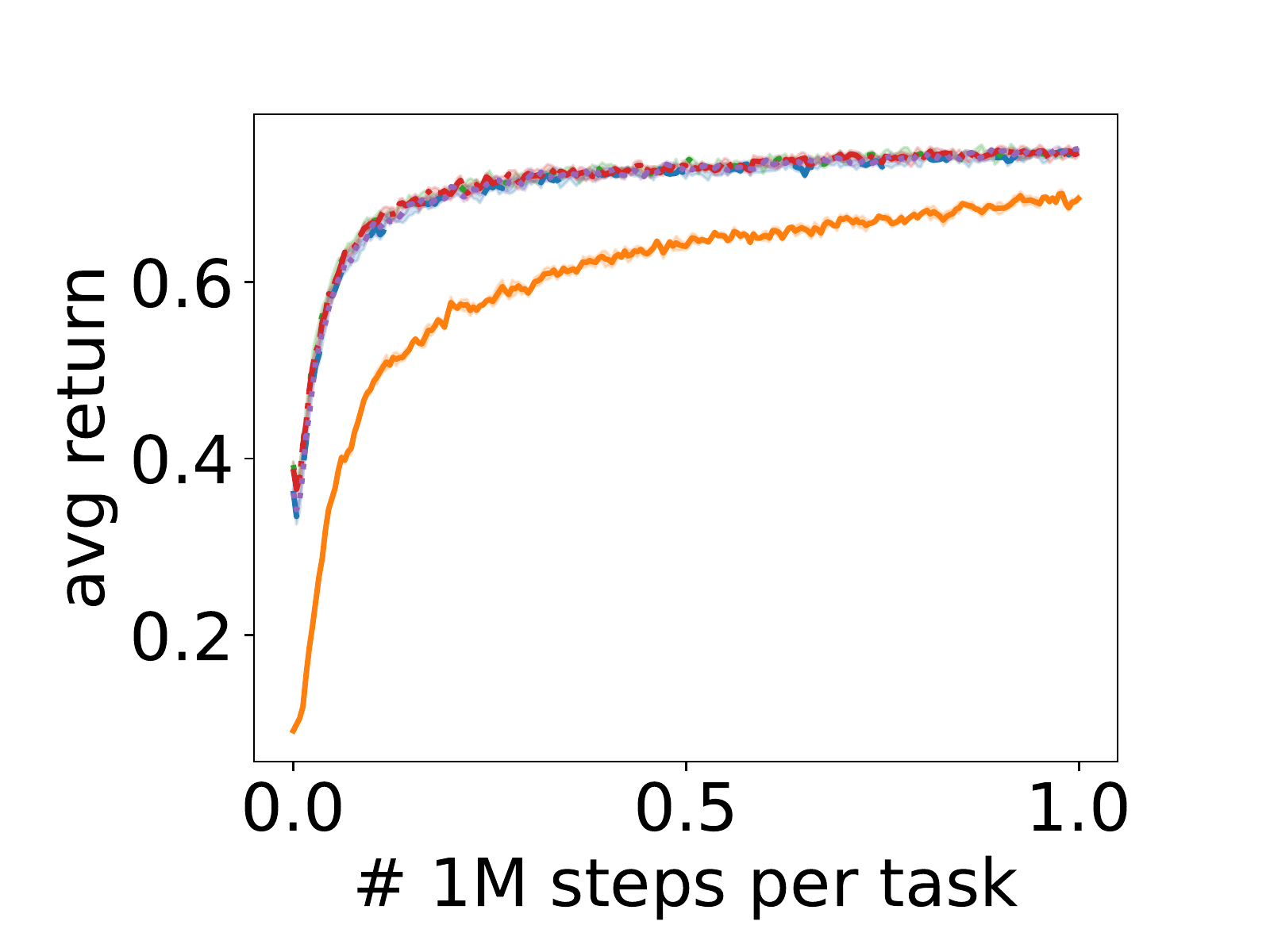}
        \caption{Multiple}
    \end{subfigure}%
    \caption[Learning curves of compositional nonstationary RL in 2-D discrete tasks.]{Average returns of STL, \compRL{}, and variants of \NonstatRL{} on $\numTasks=64$ nonstationary compositional 2-D discrete tasks. All agents are able to transfer knowledge effectively to accelerate the training with respect to STL, despite the shared components varying over time. Shaded regions represent standard errors across six seeds.}
    \label{fig:nonstatLearningCurves}
\end{figure}

To assess the usefulness of the evaluation in this chapter, the first relevant question is whether lifelong agents in this setting can still leverage accumulated knowledge, even if part of that knowledge may have become outdated. The results of the exploration substage of assimilation, summarized in Figure~\ref{fig:nonstatLearningCurves}, demonstrate that even in this nonstationary setting, compositional algorithms are able to substantially accelerate the learning with respect to STL. Interestingly, variants of \NonstatRL{} exhibit no noticeable differences in their training performance, and even the stationary algorithm, \compRL{}, performs equally well.

However, recall that, during the exploration stage, there is no requirement for the agent to maintain performance on earlier tasks, because it trains on a copy of the shared parameters. Therefore, all that Figure~\ref{fig:nonstatLearningCurves} shows is that accumulated knowledge is useful for learning the current task, but says nothing about whether this knowledge can be incorporated into the set of shared modules. To answer this latter question, Table~\ref{tab:nonstatFinal} shows the overall performance across all \textit{currently valid} tasks after the agent accommodates knowledge about the current task, averaged over the lifetime of the agent (i.e., over the sequence of tasks). This is a version of the standard final performance considered in previous chapters, but adapted to the nonstationary setting by considering only valid tasks and repeating the evaluation at various points during the training. The results show that algorithms that explicitly account for nonstationarity all perform better than \compRL{}. While the differences were small, they were consistent across nonstationary settings and across trials. 

\begin{table}[t!]
    \centering
    \caption[Performance on current task distribution in 2-D discrete tasks.]{Average performance on the currently valid distribution of 2-D discrete tasks. \NonstatRL{} tracks which tasks are relevant and optimizes performance only on those tasks, achieving higher performance. Standard errors across six seeds reported after the $\pm$. }
    \label{tab:nonstatFinal}
    \begin{tabular}{l|ccc}
    {} &                  Cyclical &                    Linear &                  Multiple \\
    \hline
    STL                      &  $0.735${\tiny$\pm0.007$} &  $0.730${\tiny$\pm0.005$} &  $0.697${\tiny$\pm0.002$} \\
    CompRL                   &  $0.803${\tiny$\pm0.007$} &  $0.813${\tiny$\pm0.005$} &  $0.760${\tiny$\pm0.005$} \\
    NonstatRL-oracle         &  $0.815${\tiny$\pm0.005$} &  $0.822${\tiny$\pm0.005$} &  $0.784${\tiny$\pm0.006$} \\
    NonstatRL-loss           &  $0.811${\tiny$\pm0.008$} &  $0.816${\tiny$\pm0.006$} &  $0.775${\tiny$\pm0.012$} \\
    NonstatRL-representation &  $0.815${\tiny$\pm0.005$} &  $0.820${\tiny$\pm0.006$} &  $0.771${\tiny$\pm0.007$} \\
    \end{tabular}
\end{table}

\section{Summary}
This chapter presented \NonstatRL{}, an extended version of the compositional RL method introduced in Chapter~\ref{cha:RL}, that equips RL agents with the ability to deal with nonstationarity in the environment. In particular, \NonstatRL{} assumes that individual components of the environment vary independently of each other, and tracks these variations to leverage only the most relevant data during off-line RL experience replay. The primary objective of the empirical evaluation in this chapter was to demonstrate that, if agents are able to target their learning updates to the individual components of the environment that are changing, they would be able to attain improved performance on the tasks that are possible under the current data distribution. Consequently, this chapter introduced a simple suite of approaches for determining the validity of previous tasks and evaluated them empirically, showing that modularity indeed improves nonstationary lifelong RL performance. 

This assumption of modular changes to the environment is more general than the standard setting, which assumes that all elements of the environment vary in similar patterns---i.e., the latter case can be recovered as a special case of the former. Moreover, none of the approaches discussed above make any assumptions about \textit{how} the components of the task might be changing over time (e.g., smoothness assumptions), making them applicable to a range of situations. However, in cases where such assumptions can be made, it would be possible to combine \NonstatRL{} with existing methods that make specific assumptions about the nature of the distribution shift.

This dissertation created the \NonstatRL{} extension of \compRL{} primarily as a proof of concept. As such, it served the purpose of demonstrating that modular nonstationarity permits effectively capturing changes in the environment. Future work should develop these ideas further, constructing realistic evaluation benchmarks for the nonstationary lifelong RL setting and crafting algorithms that leverage the notion of modular nonstationarity. 

\biblio

\chapter{\benchmark{}: A Compositional Reinforcement Learning Benchmark}
\label{cha:Benchmark}

\section{Introduction}
Benchmarks have become key drivers of progress in AI and ML research over the last couple of decades, as they facilitate the development and evaluation of new ideas and foster reproducible comparisons against existing methods. As a contribution to such advancements, this chapter presents \benchmark{}, a simulated robotic manipulation benchmark for the evaluation of compositional multitask RL approaches.

\benchmark{} follows the motivation of the robotic manipulation domain of Chapter~\ref{cha:RL}, and substantially extends it by adding a new dimension of task variation: task objectives. Compared to the previous evaluation domain, this new compositional axis translates into two key benefits: 1)~it leads to combinatorially more tasks, for a total of $\numTasks=256$, and 2)~it makes each task considerably more complex, by requiring the agent to go far beyond lifting an object and toward longer-term behaviors. In addition, \benchmark{} deliberately balances the trade-off between difficulty and attainability: the complexity and high diversity of the tasks ensures that existing compositional and general MTL approaches struggle to discover and exploit the compositional structure of the tasks, yet these methods do make progress and exhibit hints of compositional behaviors, as demonstrated empirically. 

The chapter begins with a detailed discussion of the design considerations behind \benchmark{}, diving into the details of each task component, the observation and action spaces, and the reward functions. In particular, the key insight that enables the creation of such a large benchmark is its compositional construction, which immediately grants \benchmark{} combinatorially many tasks. The discussion then describes a set of benchmarking guidelines and suggested evaluation settings for future work to leverage, promoting reproducibility. The second major contribution of this chapter is an empirical evaluation of three representative existing RL approaches: STL, end-to-end or monolithic MTL, and compositional MTL. As mentioned above, the results of this evaluation demonstrate that these methods exhibit certain compositional properties, but fall short from solving \benchmark{}. The exposition then looks toward the future and presents a set of current limitations and possible extensions of the benchmark. Overall, this chapter opens a variety of questions for future investigation.

\begin{figure}[t!]
\captionsetup[subfigure]{aboveskip=1pt}
    \begin{subfigure}[b]{0.25\textwidth}
        \centering
            \captionsetup{width=0.95\linewidth,justification=centering,singlelinecheck=false}
            \includegraphics[width=0.9\linewidth, trim={15cm 0cm 15cm 0cm}, clip]{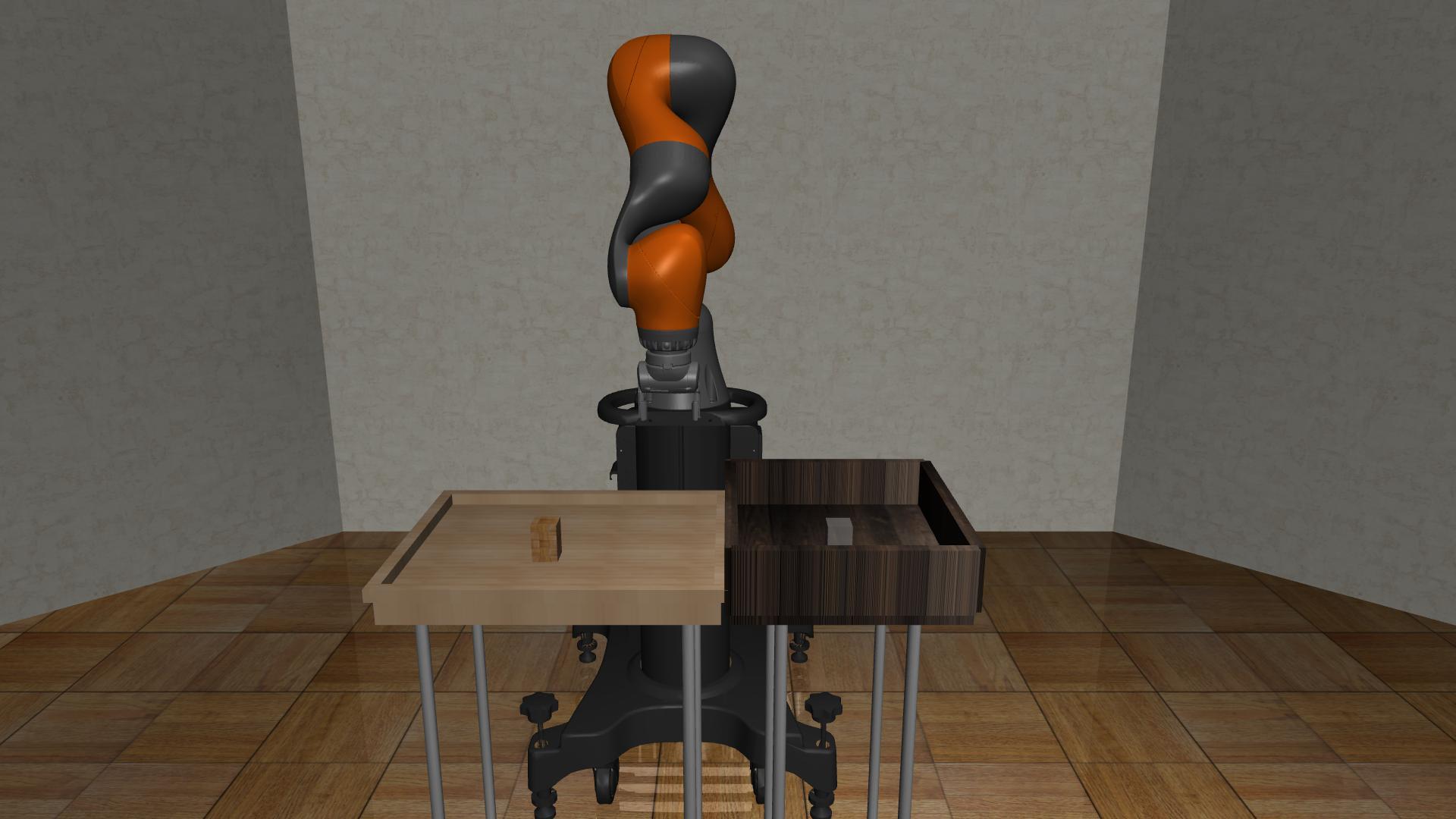}
            \caption*{\IIWA{}\\\boxObject{}\\ \noObstacle{}\\\pickPlace{}}
            \label{fig:arena1}
    \end{subfigure}%
    \begin{subfigure}[b]{0.25\textwidth}
        \centering
            \captionsetup{width=0.95\linewidth,justification=centering,singlelinecheck=false}
            \includegraphics[width=0.9\linewidth, trim={15cm 0cm 15cm 0cm}, clip]{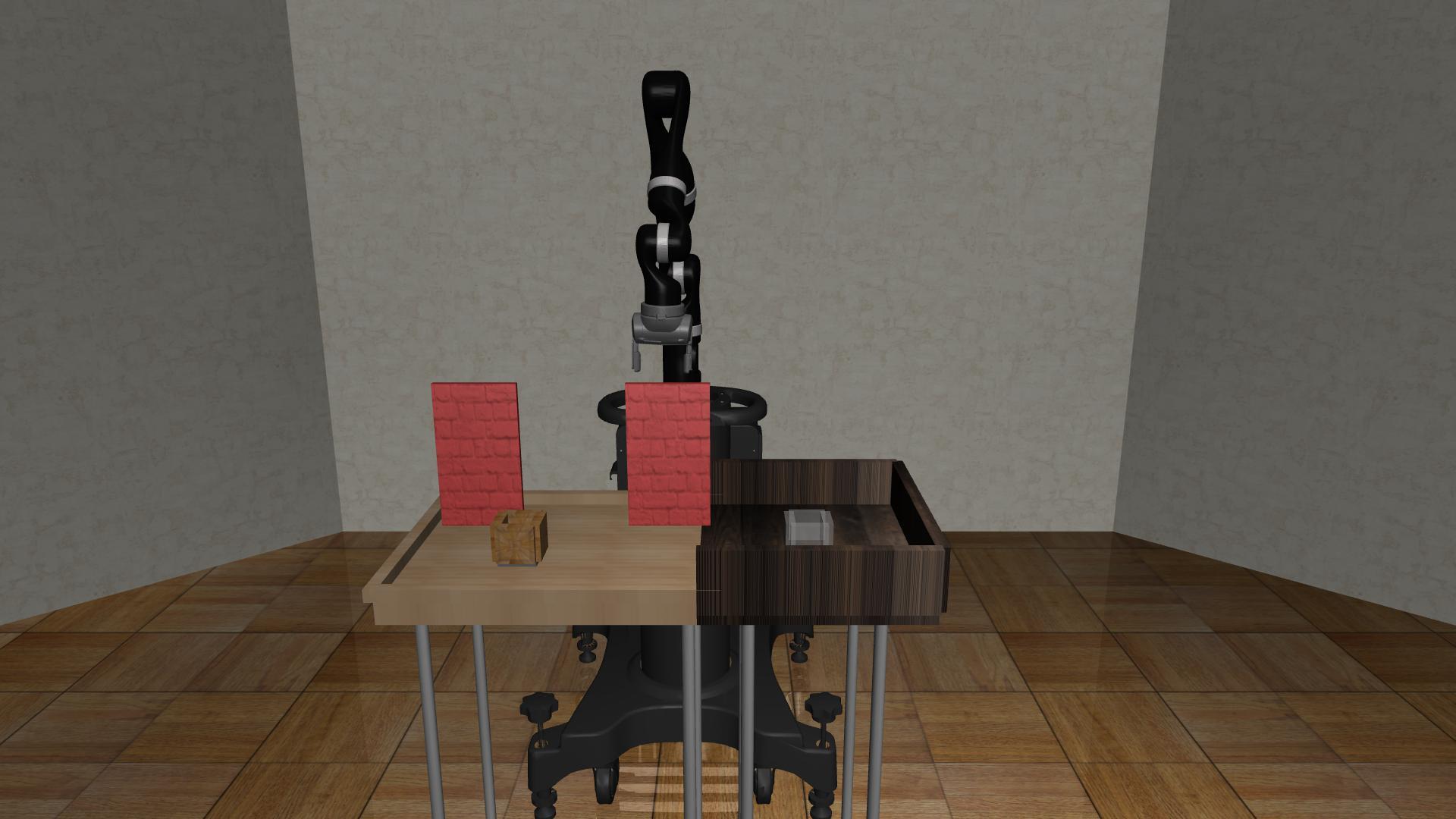}
            \caption*{\Jaco{}\\ \hollowBox{}\\ \objectDoor{}\\ \push{}}
    \end{subfigure}%
    \begin{subfigure}[b]{0.25\textwidth}
        \centering
            \captionsetup{width=0.95\linewidth,justification=centering,singlelinecheck=false}
            \includegraphics[width=0.9\linewidth, trim={15cm 0cm 15cm 0cm}, clip]{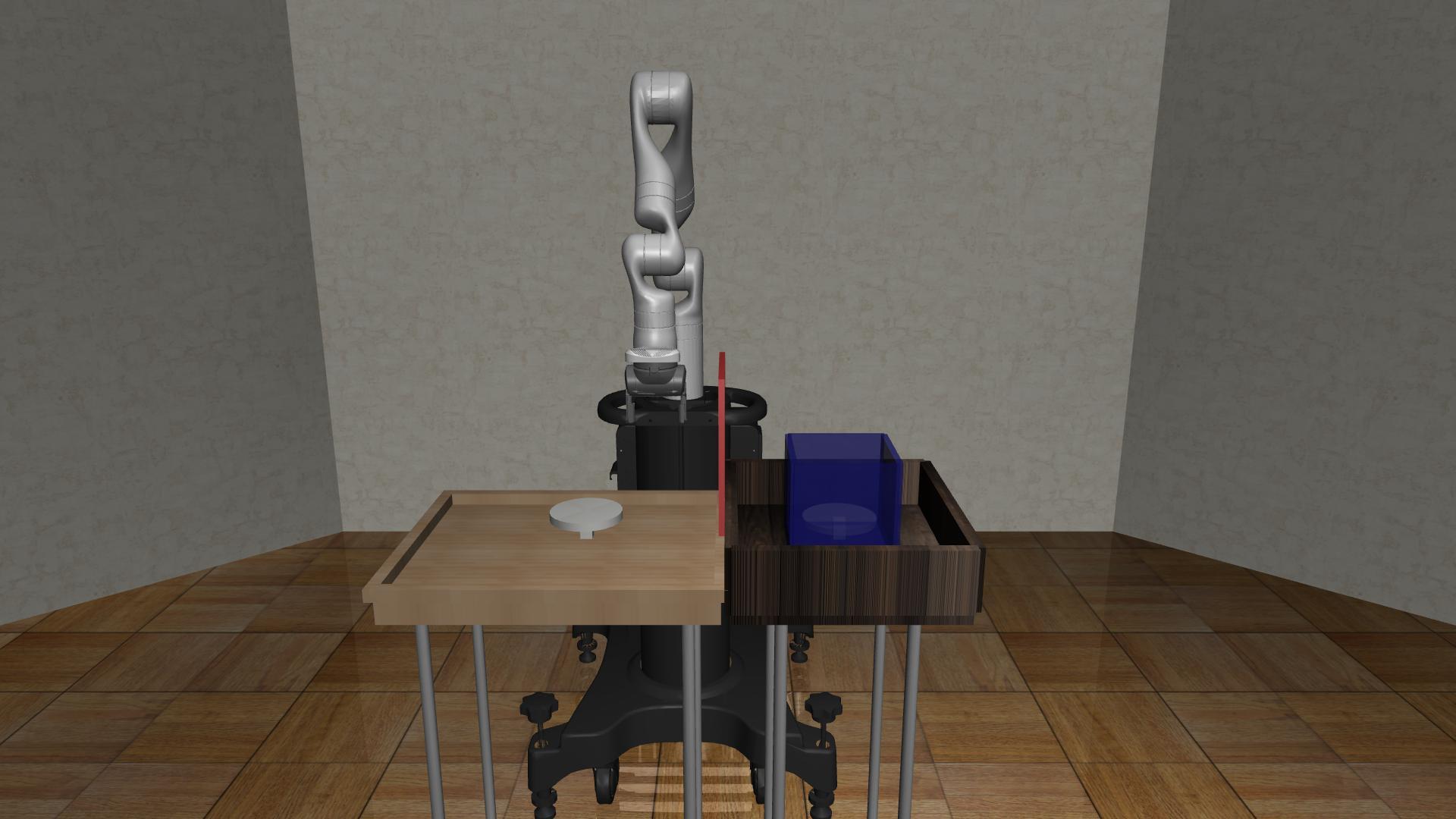}
            \caption*{\Kinova{}\\ \plate{}\\ \goalWall{}\\ \trashcan{}}
    \end{subfigure}%
    \begin{subfigure}[b]{0.25\textwidth}
        \centering
            \captionsetup{width=0.95\linewidth,justification=centering,singlelinecheck=false}
            \includegraphics[width=0.9\linewidth, trim={15cm 0cm 15cm 0cm}, clip]{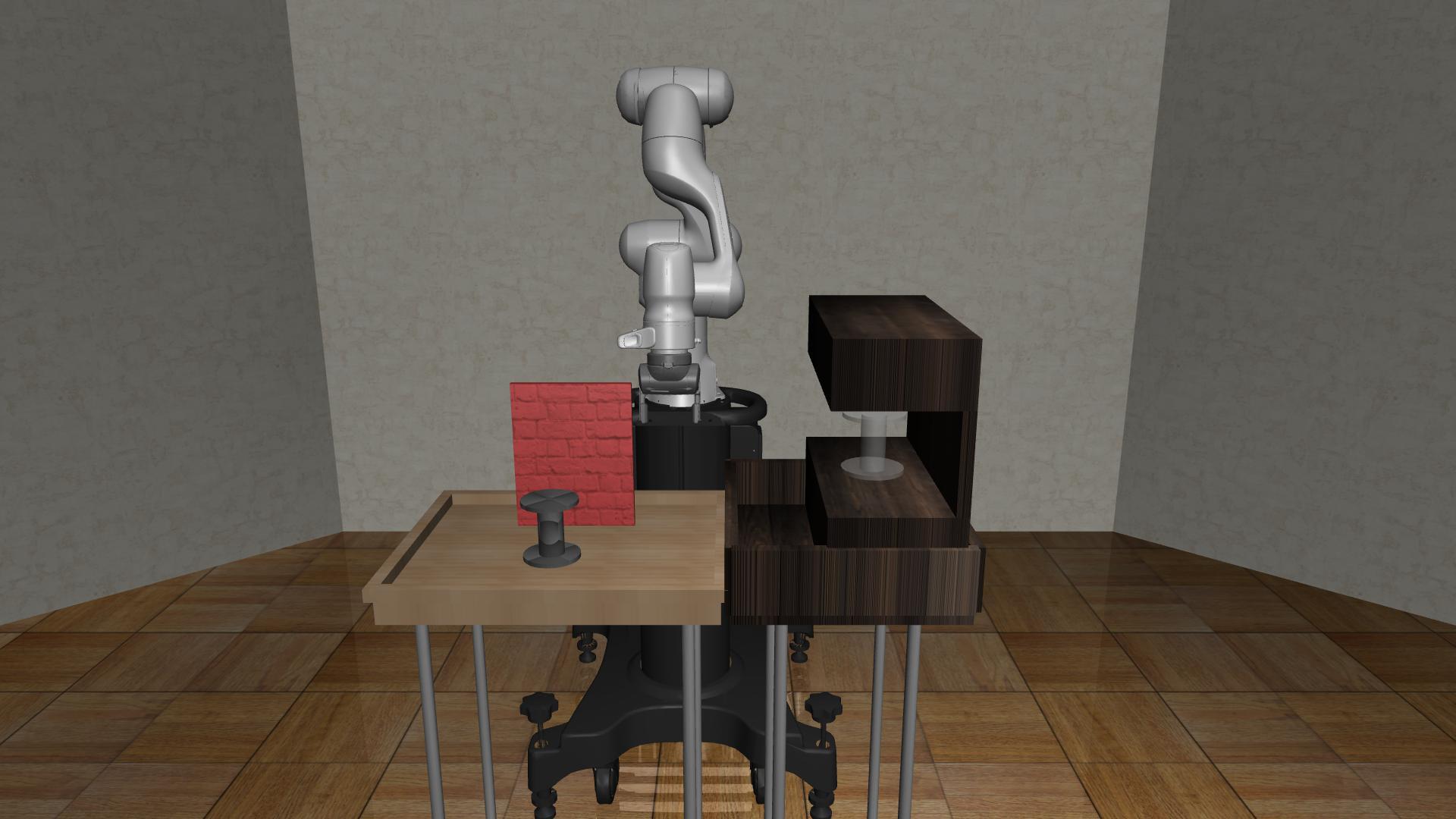}
            \caption*{\Panda{}\\ \dumbbell{}\\ \objectWall{}\\ \shelf{}}
    \end{subfigure}
    \caption[Initial conditions for \benchmark{} tasks.]{Initial conditions of four \benchmark{} tasks, containing all elements of each compositional axis. Robots: \IIWA{}, \Jaco{}, \Kinova{}, and \Panda{}. Objects: \boxObject{}, \hollowBox{}, \plate{}, and \dumbbell{}. Obstacles: \noObstacle{}, \objectDoor{}, \goalWall{}, \objectWall{}. Objectives: \pickPlace{}, \push{}, \trashcan{}, \shelf{}.}\label{fig:arenas}
\end{figure}

\section{The \benchmark{} Benchmark for Compositional Reinforcement Learning} 
\label{sec:benchmark}

\benchmark{} is a benchmark of simulated robotic manipulation tasks explicitly designed to study the ability of RL algorithms to learn functional decompositions of the solutions to the tasks, yet more broadly applicable to multitask and lifelong RL. The key idea of \benchmark{} is to build the tasks themselves compositionally, so that 1)~it contains combinatorially many (distinct) tasks, and 2)~tasks are explicitly compositionally related. 
Figure~\ref{fig:arenas} illustrates a set of sampled tasks, and Appendix~\ref{app:VisualizationOfAllTasks} shows all tasks. 

\subsection{Task Design}

\benchmark{}, built upon \robosuite{}~\citep{robosuite2020}, centers on four compositional axes: robots, objects, obstacles, and objectives. There are four elements of each type, so that combining them yields a total of $\numTasks=256$ tasks. Within each axis, the design of the elements is such that a policy that succeeds at one task is very unlikely to succeed at another task---and the optimal policy for one task is even less likely to be optimal for another task. 

All tasks take place in a two-bin environment, with bins of equal sizes across all tasks. Objects start each episode in the left bin, and the target location is in the right bin. This standardization encourages the agents to find the commonalities between the tasks. In addition, the benchmark uses crafted rewards to facilitate learning each individual task. 

\subsubsection{Task Components}
\paragraph{Robots} 
The first axis of \benchmark{} uses robot arms with different kinematic configurations, so a policy that works on one robot arm cannot be directly applied to another arm. To ensure compatibility with existing multitask RL methods, which require the dimensionality of the observation and action spaces to be compatible across tasks, \benchmark{} uses only 7-DoF manipulators. In particular, the four robot manipulators that perform the tasks are: KUKA's \IIWA{}, Kinova's \Jaco{}, Franka's \Panda{}, and Kinova's \Kinova{}. These arms vary in sizes, kinematic configurations, and position and torque limits, leading to semantic discrepancies between their observations and actions that require the agent to specialize its control policy for each arm. All arms use the Rethink Robotics two-finger gripper to manipulate objects.

\paragraph{Objects} The benchmark next considers four objects of distinct shapes that require different grasping strategies. 
The \boxObject{} is a cuboid that can be picked up from the top. The \hollowBox{} mirrors the shape of an open package, with a size sufficiently large that the gripper cannot grasp it by both sides like the \boxObject{}, and must instead grip one of its edges. The \dumbbell{} is placed upright, and its weights are larger than the gripper, and so the manipulator can only grasp it horizontally by the bar. The \texttt{plate}'s diameter is also greater than the gripper size, and therefore can only be grasped horizontally by the edge.

\paragraph{Obstacles}
The third axis of variation in \benchmark{} is a set of four obstacles that require distinct behaviors. 
The \objectWall{} is a brick wall placed between the robot and the object, while the \objectDoor{} is a similarly placed doorway between two brick walls. These two obstacles require avoiding opposite regions of the space while reaching for the object. The \goalWall{} is also a brick wall, but is placed between the left and right bins, blocking the direct path to the goal after grasping the object. Additionally, the benchmark includes tasks with \noObstacle{}.

\paragraph{Task objectives} The final compositional axis is a set of different task objectives: \pickPlace{} an object into the right bin, \push{} the object from the left to the right bin, drop the object into a \trashcan{}, and place the object on a \shelf{}. The overall trajectory required to attain each of these objectives is behaviorally distinct.

Thanks to combinatorial explosion, there are $256$ possible combinations of these components, leading to a set of $\numTasks=256$ highly varied tasks. Each of the tasks requires a fundamentally unique policy, but the construction of the tasks reveals exactly how they relate to one another, enabling researchers to extract insights about the kind of compositionality that deep multitask RL methods exhibit.

\subsubsection{Observation and Action Spaces}
\label{sec:observationSpace}

The observation space factors into the following elements, tied to the task components described in the previous section: 
\begin{itemize}
\item \textit{Robot observation}\hspace{2em}The proprioceptive portion of the observation includes the robot's joint positions, joint velocities, end effector pose, finger positions, and finger velocities.

\item \textit{Object observation}\hspace{2em}The agent observes both the absolute position and orientation of the object in world coordinates, as well as its position and orientation with respect to the robot's end effector. Note that this observation does not give away any information that distinguishes objects from one another (e.g., their geometric properties).

\item \textit{Obstacle observation}\hspace{2em}The agent also observes the absolute and relative positions and orientations of the obstacles. Similarly, this does not give away what is the free space of the environment (e.g., \objectWall{} and \objectDoor{} are always placed in the same location, but they block off opposite parts of the space).

\item \textit{Goal observation}\hspace{2em}The observation also contains the absolute and relative position and orientation of the goal, which is fixed at the center of the target region (e.g., the right bin or the shelf), as well as the relative position of the goal with respect to the object. However, to simplify the learning of \pickPlace{}, \trashcan{}, and \shelf{} tasks, reaching any arbitrary location in the target region solves each task. 
In contrast, \push{} tasks do require reaching the location specified in the observation.

\item \textit{Task observation}\hspace{2em}The agent may also be given access to a multi-hot indicator that identifies each of the components of the task (i.e., the robot, object, obstacle, and objective). This can be used as a task descriptor for MTL training.
\end{itemize}

The action space for each task is eight-dimensional. The first seven dimensions provide target positions for each robot joint. 
Under the hood, a proportional-derivative (PD) controller executes the motor commands that follow the joint positions provided by the agent---this is directly handled by \robosuite{} and is completely transparent to the agent. The eighth dimension is a binary action that indicates whether the gripper should be open or closed. 

\subsubsection{Reward Functions}
\label{sec:RewardFunctions}

While \benchmark{} supports sparse rewards for successful completion, this leads to an extremely hard exploration problem. Consequently, to isolate the problem of compositional MTL, \benchmark{} provides a crafted reward that encourages exploration in stages, such that each stage leads the agent to a state that is closer to task completion.

During the \textbf{reach} stage, the environment rewards the agent for reducing the distance to the object. This stage terminates once the agent \textbf{grasps} the object, which gives a binary reward. These two initial stages are common to all objectives. In all tasks except for \push{}, the environment next rewards the agent for \textbf{lifting} the object up to a given height. In the case of \shelf{} tasks, the reward then encourages the agent to \textbf{align} the gripper with the horizontal plane, facing the shelf. The next stage rewards the agent for \textbf{approaching} the goal location based on the horizontal distance. In \pickPlace{} tasks, the reward then encourages the agent to \textbf{lower} the object down to the bin. In \trashcan{} tasks, the environment instead rewards the agent for \textbf{dropping} the object while above the trash can with a binary reward. 

The final stage is a binary \textbf{success} reward. The \pickPlace{} tasks succeed if the object is in the bin and the robot is near the object; this latter constraint differentiates \pickPlace{} and \trashcan{} tasks. Solving the \push{} tasks involves placing the object near the goal location. The agent succeeds on \trashcan{} tasks if the object is inside the trashcan and the gripper is \textit{not}. The success criterion for \shelf{} tasks is that the object is on the shelf. 

The maximum possible reward is $\Rewards=1$ and the agent only attains it upon successfully executing the task. Table~\ref{tab:stagedRewards} summarizes the stages of each task objective, and the following items include precise formulas for the task objective rewards.

\begin{table}[t]
    \centering
    \caption[Reward stages per task objective in \benchmark{}.]{Reward stages per task objective.}
    \label{tab:stagedRewards}
    \begin{tabular}{l|l@{}l@{}l@{}l@{}l@{}l@{}l}
        Task & \multicolumn{7}{c}{Stages}\\
        \hline
        \pickPlace{} & reach $\rightarrow\,\,$& grasp $\rightarrow\,\,$& lift $\rightarrow\,\,$& & approach $\rightarrow\,\,$& lower $\rightarrow\,\,$ & success  \\
        \push{} & reach $\rightarrow\,\,$& grasp $\rightarrow\,\,$& & & approach $\rightarrow\,\,$ & & success \\
        \trashcan{} & reach $\rightarrow\,\,$& grasp $\rightarrow\,\,$& lift $\rightarrow\,\,$& & approach $\rightarrow\,\,$& drop $\rightarrow\,\,$ & success   \\
        \shelf{} & reach $\rightarrow\,\,$& grasp $\rightarrow\,\,$& lift $\rightarrow\,\,$& align $\rightarrow\,\,$& approach $\rightarrow\,\,$ & & success \\
    \end{tabular}
\end{table}

\begingroup
\allowdisplaybreaks
\begin{itemize}[leftmargin=!,labelindent=5pt,itemindent=30pt]
\item\pickPlace{} tasks:
\begin{flalign}
    \nonumber \Rewards_{\mathrm{reach}} =& 0.2 ( 1 - \tanh(10 \mathrm{target\_dist}))&&&\\
    \nonumber \Rewards_{\mathrm{grasp}} =& \begin{cases}
        1 \quad &\text{if grasping}\\
        0 \quad &\text{otherwise}
    \end{cases}
    &&\\
    \nonumber \Rewards_{\mathrm{lift}} =& \begin{cases}
        0.3 + 0.2 (1 - \tanh(5 \mathrm{z\_dist\_target\_height})) \quad &\text{if }\Rewards_{\mathrm{grasp}} > 0\\
        0 \quad &\text{otherwise}
    \end{cases}&&\\
    \nonumber \Rewards_{\mathrm{approach}} =& \begin{cases}
        \Rewards_{\mathrm{lift}} + 0.2 (1 - \tanh(2 \mathrm{goal\_xy\_dist})) \quad &\text{if } \Rewards_{\mathrm{lift}} > 0.45 \text{ \& object not above bin}\\
        0.5 + 0.2 (1 - \tanh(2 \mathrm{goal\_xy\_dist})) \quad &\text{if } \Rewards_{\mathrm{lift}} > 0.45 \text{ \& object above bin}\\
        0 \quad &\text{otherwise}
    \end{cases}&&\\
    \nonumber \Rewards_{\mathrm{lower}} =& \begin{cases}
        0.7 + 0.2(1-\tanh(5\mathrm{z\_dist\_bin})) \quad &\text{if object above bin \& } \Rewards_{\mathrm{grasp}} > 0\\
        0 \quad &\text{otherwise}
    \end{cases}&&\\ 
    \nonumber \Rewards_{\mathrm{success}} =& \begin{cases}
        1 \quad &\text{if object in bin \& } \Rewards_{\mathrm{reach}} > 0.07\\
        0 \quad &\text{otherwise}
    \end{cases}&&\\
    \Rewards =& \max_{\mathrm{stage}}\Rewards_{\mathrm{stage}}&&
\end{flalign}
    
\item\push{} tasks:
\begin{flalign}
    \nonumber\Rewards_{\mathrm{reach}} =& 0.2 ( 1 - \tanh(10 \mathrm{target\_dist}))&&\\
    \nonumber\Rewards_{\mathrm{grasp}} =& \begin{cases}
        1 \quad &\text{if grasping}\\
        0 \quad &\text{otherwise}
    \end{cases}&&
    \\
    \nonumber\Rewards_{\mathrm{approach}} =& \begin{cases}
        0.3 + 0.4 (1 - \tanh(5 \mathrm{goal\_xy\_dist})) \quad &\text{if }\Rewards_{\mathrm{grasp}} > 0\\
        0 \quad &\text{otherwise}
    \end{cases}&&\\
    \nonumber\Rewards_{\mathrm{success}} =& \begin{cases}
        1 \quad &\text{if } \text{goal\_xy\_dist} \leq 0.03\\
        0 \quad &\text{otherwise}
    \end{cases}&&\\
    \Rewards =& \max_{\mathrm{stage}}\Rewards_{\mathrm{stage}}&& 
\end{flalign}

\item\trashcan{} tasks:
\begin{flalign}
    \nonumber\Rewards_{\mathrm{reach}} =& 0.2 ( 1 - \tanh(10 \mathrm{target\_dist}))&&\\
    \nonumber\Rewards_{\mathrm{grasp}} =& \begin{cases}
        1 \quad &\text{if grasping \& object not in can}\\
        0 \quad &\text{otherwise}
    \end{cases}&&
    \\
    \nonumber\Rewards_{\mathrm{lift}} =& \begin{cases}
        0.3 + 0.2 (1 - \tanh(5 \mathrm{z\_dist\_target\_height})) \quad &\text{if }\Rewards_{\mathrm{grasp}} > 0\\
        0 \quad &\text{otherwise}
    \end{cases}&&\\
    \nonumber\Rewards_{\mathrm{approach}} =& \begin{cases}
        \Rewards_{\mathrm{lift}} + 0.2 (1 - \tanh(2 \mathrm{goal\_xy\_dist})) \quad &\text{if } \Rewards_{\mathrm{lift}} > 0.45 \text{ \& object not above can}\\
        0.5 + 0.2 (1 - \tanh(2 \mathrm{goal\_xy\_dist})) \quad &\text{if } \Rewards_{\mathrm{lift}} > 0.45 \text{ \& object above can}\\
        0 \quad &\text{otherwise}
    \end{cases}&&\\
    \nonumber\Rewards_{\mathrm{drop}} =& \begin{cases}
        0.95 \quad &\text{if object above can \& } \Rewards_{\mathrm{grasp}} = 0\\
        0 \quad &\text{otherwise}
    \end{cases}&&\\ 
    \nonumber\Rewards_{\mathrm{success}} =& \begin{cases}
        1 \quad &\text{if object trash can \& gripper not in can}\\
        0 \quad &\text{otherwise}
    \end{cases}&&\\
    \Rewards =& \max_{\mathrm{stage}}\Rewards_{\mathrm{stage}}&&
\end{flalign}

\item \shelf{} tasks:
\begin{flalign}
    \nonumber\Rewards_{\mathrm{reach}} =& 0.2 ( 1 - \tanh(10 \mathrm{target\_dist}))&&\\
    \nonumber\Rewards_{\mathrm{grasp}} =& \begin{cases}
        1 \quad &\text{if grasping}\\
        0 \quad &\text{otherwise}
    \end{cases}&&
    \\
    \nonumber\Rewards_{\mathrm{lift}} =& \begin{cases}
        0.3 + 0.2 (1 - \tanh(5 \mathrm{z\_dist\_target\_height})) \quad &\text{if }\Rewards_{\mathrm{grasp}} > 0\\
        0 \quad &\text{otherwise}
    \end{cases}&&\\
    \nonumber\Rewards_{\mathrm{align}} =& \begin{cases}
        0.5 + 0.3 (1 - \tanh(\mathrm{y\_axis\_orientation})) \quad &\text{if object in front of shelf}\\
        0 \quad &\text{otherwise}
    \end{cases}&&\\
    \nonumber\Rewards_{\mathrm{approach}} =& \begin{cases}
        0.8 + 0.1 (1-\tanh(5\mathrm{y\_dist\_shelf})) \quad &\text{if object in front of shelf \& } \Rewards_{\mathrm{align}} > 0.6\\
        0 \quad &\text{otherwise}
    \end{cases}&&\\ 
    \nonumber\Rewards_{\mathrm{success}} =& \begin{cases}
        1 \quad &\text{if object in shelf}\\
        0 \quad &\text{otherwise}
    \end{cases}&&\\
    \Rewards =& \max_{\mathrm{stage}}\Rewards_{\mathrm{stage}}&&
\end{flalign}
\end{itemize}
\endgroup

\subsubsection{Episode Initialization and Termination}

Upon initialization of each new episode, the environment places the graspable object in a random location of the left bin. Tasks that contain an obstacle restrict the object's initial location to the regions of the space that would explicitly require the robot to circumvent the obstacle. The goal location is at the center of the target region, which is itself at some fixed location in the right bin, and the robot arm begins at a fixed position with the gripper facing downward. Figure~\ref{fig:arenas} displays sampled initial conditions.

Each episode terminates after a horizon of $H=500$ time steps. 
In addition, \push{} tasks terminate if the robot lifts the object more than a set threshold above the table, in order to avoid success if the robot executes a pick-and-place strategy.

\subsection{Evaluation Settings}
\label{sec:EvaluationSettings}

\benchmark{} evaluates agents for their training speed and final performance on a set of \textit{training} tasks, akin to training sets in supervised STL settings. While this is a measure of training performance, it is the standard evaluation setting of the majority of works in RL. After training, the benchmark evaluates agents on a \textit{test} set of unseen tasks. Both of these evaluations explore the ability of agents to discover compositional properties of the tasks.

\subsubsection{Metrics}

Two metrics measure the performance of agents over the training and test tasks. For an agent evaluated over $\numTasks$ tasks, with $M$ evaluation trajectories for each task, each trajectory of length $H$, \benchmark{} computes the average metrics as follows:

\paragraph{Return} The first metric is the standard cumulative returns:
\begin{align}
    \bar{R} = \frac{1}{\numTasks M}\sum_{t=0}^{\numTasks}\sum_{j=0}^{M}\sum_{i=0}^{H} \Rewards^{(t)}(\state_i,\action_i)\enspace.
\end{align}
This is the usual RL evaluation criterion, and directly relates to the optimization objective.

\paragraph{Success} The second metric is the success rate, based on each task's definition of success:
\begin{align}
    \bar{S} = \frac{1}{\numTasks M} \sum_{t=0}^{\numTasks}\sum_{j=0}^{M} \max_{i\in[0,H]}  \mathds{1} \Big[\Rewards^{(t)}(\state_i, \action_i) = 1\Big] \enspace,
\end{align}
where $\mathds{1}$ is the indicator function. Note that a trajectory is successful if at \textit{any} time, the agent is in a success state.

\subsubsection{Evaluation on Training Tasks}

Evaluations first assess the agent's performance on the tasks that it trains on. An agent that is capable of extracting the compositional properties of the tasks should be able to achieve transfer across the tasks. Ideally, this transfer should translate to both faster convergence in terms of the number of samples required to learn, as well as higher final performance after convergence. In particular, evaluations in this setting should compare agents against equivalent STL agents that use the same training mechanism but do so individually on every task, without any notion of shared knowledge.

\subsubsection{Evaluation on Test Tasks}

The key property that \benchmark{} assesses is the ability of approaches to combine trained components in novel combinations to handle new tasks. As per Chapter~\ref{cha:RL}, this can take the following two forms:

\paragraph{Zero-shot generalization with task descriptors}
If the observation provides the agent with the multi-hot indicators described in Section~\ref{sec:observationSpace}, then the agent could (in principle) solve new, unseen tasks without any training on them. This would be possible only if the agent learns the compositional structure of the tasks and is able to combine its existing components into a solution to the new task. Intuitively, after learning 1)~the \pickPlace{} task with the \boxObject{} object avoiding the \objectDoor{} obstacle using the \IIWA{} arm, and 2)~the \push{} task with the \plate{} object avoiding the \objectWall{} obstacle using the \Panda{} arm, if the agent knows how each component relates to the overall task, it could for example swap the \IIWA{} and \Panda{} arms and solve the opposite tasks without additional training. 

\paragraph{Fast adaptation without task descriptors} Alternatively, the benchmark might not inform the agent of which components make up the current task, and require it to discover this information through experience. The goal of the agent should then be to discover this information as rapidly as possible in order to solve the new task with little experience. 

\subsubsection{Access to State Decomposition}
\label{sec:StateDecomposition}
The compositional architecture of \citet{devin2017learning} and those introduced in Chapter~\ref{cha:RL} require knowledge about which components of the observation affect which parts of the architecture. While this information is readily available in \benchmark{}, fair comparisons require noting whether the agent is given this decomposition of the state. Note that both zero-shot and fast-adaptation settings could be targeted with or without the state decomposition.

\subsubsection{Sample of Training Tasks}
\label{sec:sampleTrainingTasks}

Understanding the compositional capabilities of RL algorithms requires a careful study of the sample of combinations (i.e., tasks) that is provided to the agent for training. \benchmark{} proposes the following evaluation settings:

\paragraph{Uniform sampling} The simplest setting samples the training tasks uniformly at random, and requires the agent to generalize to all possible combinations of the seen components. The agent therefore must learn to combine its knowledge in different ways after having seen each component in various combinations.

\paragraph{Restricted sampling} This much-harder setting restricts the training to a single task for one of the components and many tasks for other components---e.g., in \benchmark{}\textbackslash{}\IIWA{}, the agent sees only one \IIWA{} task and must generalize to all other \IIWA{} tasks. This is akin to Experiment $3$ in the work of \citet{lake2018generalization}, which demonstrated that this is an onerous problem even in the supervised setting. While a complete evaluation would require various choices of restricted arms, objects, obstacles, and objectives, initial evaluations could focus on these four proposed settings: \benchmark{}\textbackslash{}\IIWA{}, \benchmark{}\textbackslash{}\hollowBox{}, \benchmark{}\textbackslash{}\objectWall{}, and \benchmark{}\textbackslash{}\pickPlace{}.

\paragraph{Smaller-scale benchmarks} While large benchmarks like \benchmark{} are appealing for studying multitask RL at scale, developing ideas in such large task sets is often (unfortunately) prohibitively time consuming. Given the compositional nature of \benchmark{}, it is trivial to extract smaller-scale benchmarks that maintain the properties of the original, full-scale benchmark. For example, \benchmark{}$\cap$\IIWA{} considers only the $\numTasks=64$ \IIWA{} tasks. One interesting property of such reduced benchmarks is that they permit studying the difficulty of generalization across certain axes (e.g., if an agent can transfer knowledge across objects but not across robots, then it would perform much better on \benchmark{}$\cap$\IIWA{} than on the full \benchmark{}). Future evaluations should evaluate agents on the following smaller-scale benchmarks: \benchmark{}$\cap$\IIWA{}, \benchmark{}$\cap$\hollowBox{}, \benchmark{}$\cap$\noObstacle{}, and \benchmark{}$\cap$\pickPlace{}---the easiest elements along each compositional axis.

\section{Benchmarking Existing Reinforcement Learning Methods on \benchmark{}} 
\label{sec:experiments}

The empirical evaluation in this section had two primary objectives. First, to demonstrate that \benchmark{} is a useful evaluation benchmark in terms of: 1)~existing algorithms making progress toward solving the problems, 2)~the tasks exhibiting compositional properties, and 3)~existing approaches leaving substantial room for improvement in performance. 
Second, to provide benchmarking results of existing algorithms for future work to leverage. 

At the time of this dissertation, a paper introducing \benchmark{} was under double-blind peer review. Consequently, the anonimized code is available at \url{https://github.com/collassubmission91/CompoSuite-Code}, which will eventually redirect to the public version.

\subsection{Experimental Setting}

The underlying RL algorithm used for all  evaluations was the modified version of PPO described in Chapter~\ref{cha:RL}. In particular, recall that the policy used a $\tanh$ activation in the output layer of the network and a constant (fixed) variance. These two modifications were critical for enabling agents to explore meaningful solutions to the tasks in \benchmark{}. 
The experiments evaluated three types of agents built upon this base algorithm:
\begin{itemize}
    \item \textit{STL} agents that trained on each task individually, without any knowledge-sharing across tasks. Lack of sharing precludes these agents from generalizing to unseen tasks, and consequently the evaluation only assessed their performance on the training tasks. Additionally, the environment withheld the task descriptor from the observation, as it would appear as a constant to each STL agent.
    \item \textit{MTL} agents that trained a {\em shared model} for all tasks, using the task descriptor in the observation to help differentiate between tasks and learn to specialize the policy for each task. Given the need for the MTL agents to encode multiple policies in a single model, these agents used a larger capacity than an individual STL agent.
    \item \textit{Compositional} agents that constructed a \textit{different} model for each task from a set of \textit{shared} components. The architecture was a variant of the modular network of Chapter~\ref{cha:RL} that establishes each policy from a set of modules, with one module for each robot, object, obstacle, and objective. The relevant state component from Section~\ref{sec:observationSpace} constituted the input to each module, and the task descriptor selected the correct modules. For fairness, the \textit{overall} number of parameters across modules was equivalent to that of MTL agents.
 \end{itemize}

The experiments evaluated each agent in three settings: the full \benchmark{}, the smaller-scale \benchmark{}$\cap$\IIWA{}, and the restricted \benchmark{}\textbackslash{}\pickPlace{}. Each of these settings gave a subset of the tasks to the agents for training, and evaluated agents for their speed and final performance over the training tasks. After training, the benchmark additionally evaluated the MTL and compositional agents for their ability to solve unseen tasks without any additional training by leveraging the task descriptors. 

\subsection{Hyperparameters}

Table~\ref{tab:hyperparametersBenchmark} reports the hyperparameters that each agent used for training. Hyperparameter tuning obtained these values via grid search on a set of tasks using STL, and maintained those for the MTL and compositional agents.

\begin{table}[t]
    \centering
        \caption[Summary of hyperparameters used for evaluations on \benchmark{}.]{Summary of optimized hyperparameters used by the baselines.}
        \label{tab:hyperparametersBenchmark}
        \begin{tabular}{l|c|c|c}
        Hyperparameter & STL & MTL & Compositional Learner \\
         \hline
        $\gamma$ & $0.99$ & $0.99$ & $0.99$ \\
        \# layers & $2$ & $2$ & - \\
        \# hidden units & $64$ & $256$ & - \\
        \# steps per task per update & $16{,}000$ & $16{,}000$ & $16{,}000$ \\
        \# total step per task & $10{,}000{,}000$ & $10{,}000{,}000$ & $10{,}000{,}000$ \\
        PPO clip value & $0.2$ & $0.2$ & $0.2$ \\
        $\pi$ learning rate & $1e^{-4}$ & $1e^{-4}$ & $1e^{-4}$ \\
        $V$ learning rate & $1e^{-4}$ & $1e^{-4}$ & $1e^{-4}$ \\
        \# $\pi$ update iterations & $128$ & $128$ & $128$ \\
        \# $V$ update iterations & $128$ & $128$ & $128$ \\
        target KL & $0.02$ & $0.02$ & $0.02$ \\
    \end{tabular}
    \label{tab:hyperparams}
\end{table}

\subsection{Compositional Network Architecture}

\begin{figure}[b!]
\centering
    \includegraphics[width=0.74\linewidth]{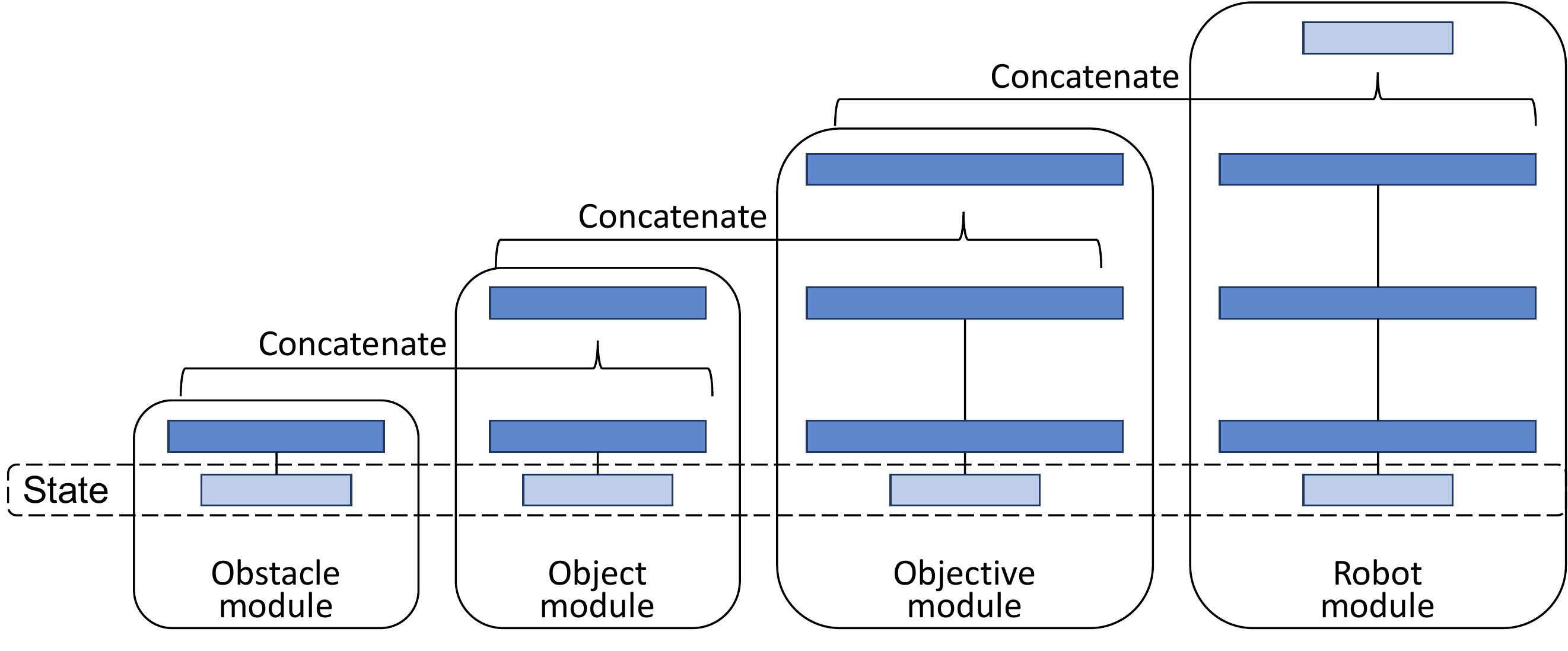}
    \caption[Modular architecture used for learning compositional policies on \benchmark{}.]{Modular architecture used for learning compositional policies.}
    \label{fig:ComposuiteArchitecture}
\end{figure}

The network architecture for the compositional agent, illustrated in Figure~\ref{fig:ComposuiteArchitecture}, follows a graph structure similar to that of Chapter~\ref{cha:RL}. The architecture consists of a total of $16$ MLP modules, each of which maps to one of the components in \benchmark{}. Specifically, there are four obstacle, object, objective, and robot modules, respectively. Each module corresponds to a level in the graph hierarchy, such that the previous level's MLP passes its output as input to the final layer of the MLP of the current level. The architecture first processes the obstacle observation. Every obstacle module consists of a single-hidden-layer MLP with $32$ hidden units---because this is the first level, there is no additional input other than the obstacle observation. The second level processes the object observation via two hidden layers of size $32$, additionally consuming as input to the second layer the obstacle observation processed by the corresponding module. The object module feeds into the second layer of the objective module, which consists of three layers of size $64$. Finally, the objective module's output acts as an input into the third layer of the robot module, which has three hidden layers of size $64$. In the value function network $V$, the robot module outputs the estimated value, while in the policy network $\pi$, it outputs the mean of the Gaussian policy.

\begin{figure}[t!]
\centering
    \begin{subfigure}[b]{\textwidth}
            \centering
                \includegraphics[trim={0.1cm 0.12cm 0.1cm 0.15cm}, clip,height=0.45cm]{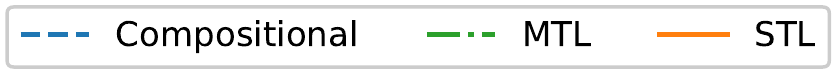}
        \end{subfigure}\\
    \begin{subfigure}[b]{0.45\textwidth}
    \centering
        \includegraphics[height=4.5cm, trim={0.2cm 0.0cm 1.5cm 0.8cm}, clip]{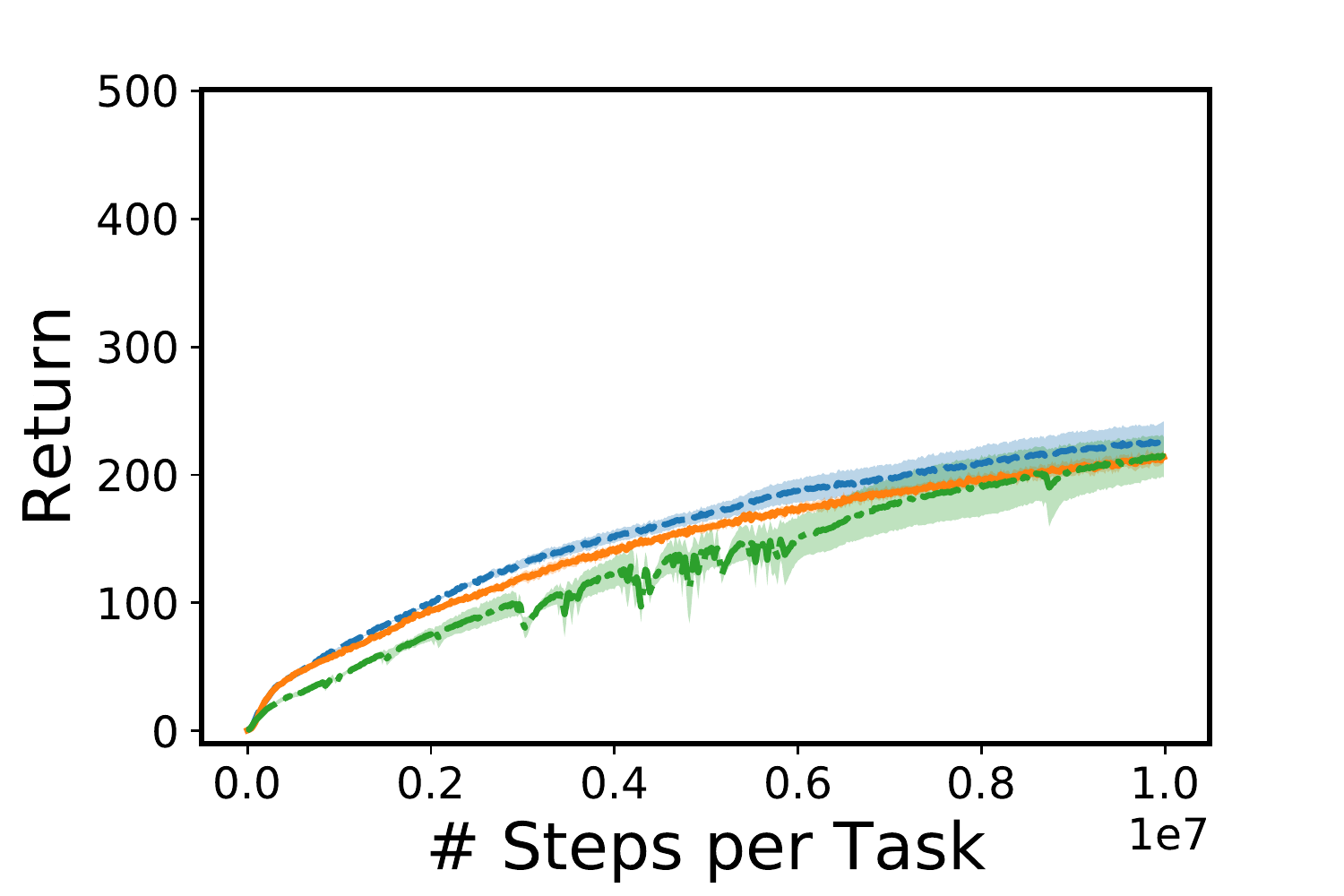}
    \end{subfigure}%
    \begin{subfigure}[b]{0.45\textwidth}
    \centering
        \includegraphics[height=4.5cm, trim={.2cm 0.0cm 1.5cm 0.8cm}, clip]{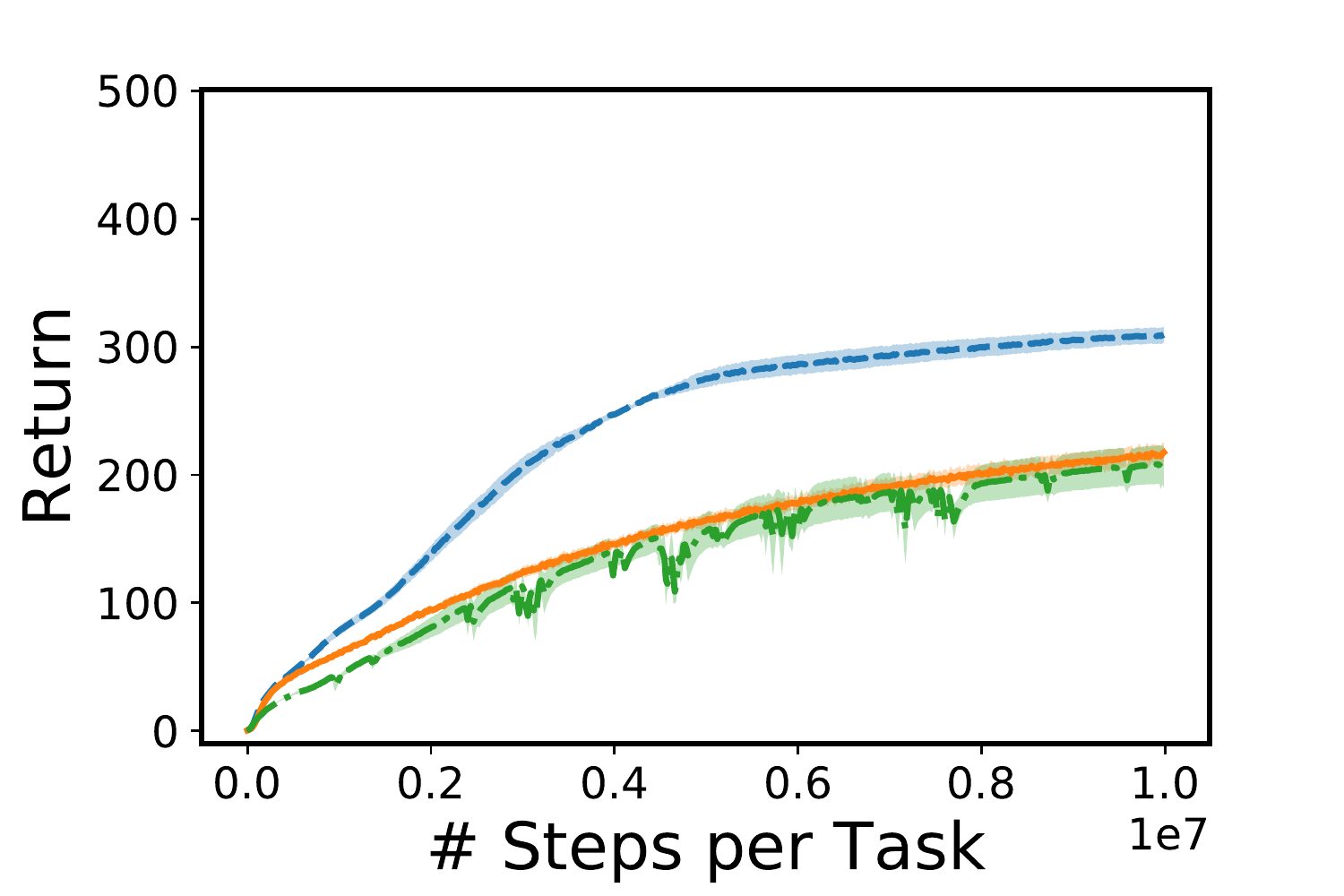}
    \end{subfigure}\\
    \vspace{1em}
    \begin{subfigure}[b]{0.45\textwidth}
    \centering
        \includegraphics[height=4.5cm, trim={0.2cm 0.0cm 1.5cm 0.8cm}, clip]{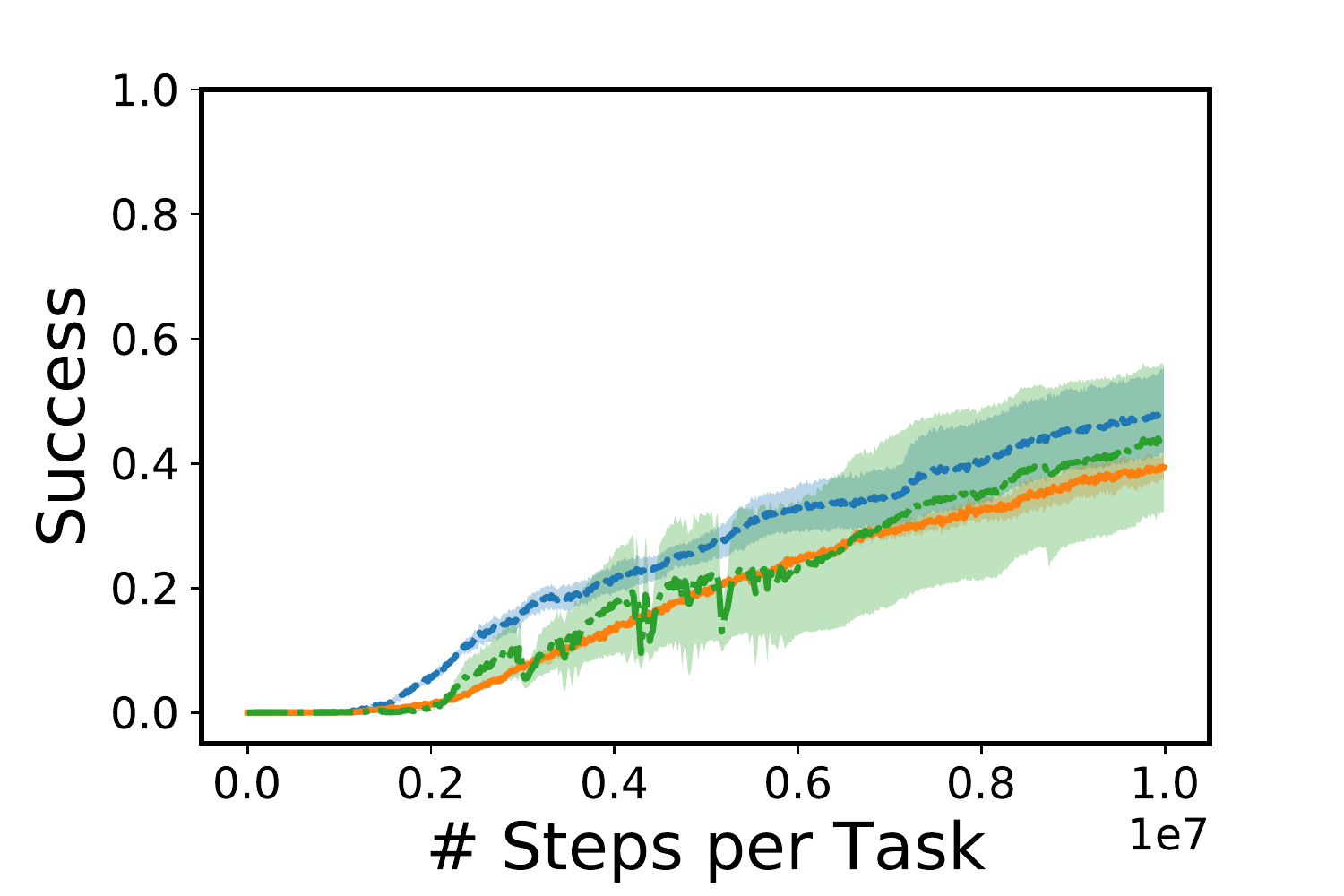}
        \caption{$\numTasks=56$ tasks}
        \label{fig:fullBenchmarkCurvesSmall}
    \end{subfigure}%
    \begin{subfigure}[b]{0.45\textwidth}
    \centering
        \includegraphics[height=4.5cm, trim={.2cm 0.0cm 1.5cm 0.8cm}, clip]{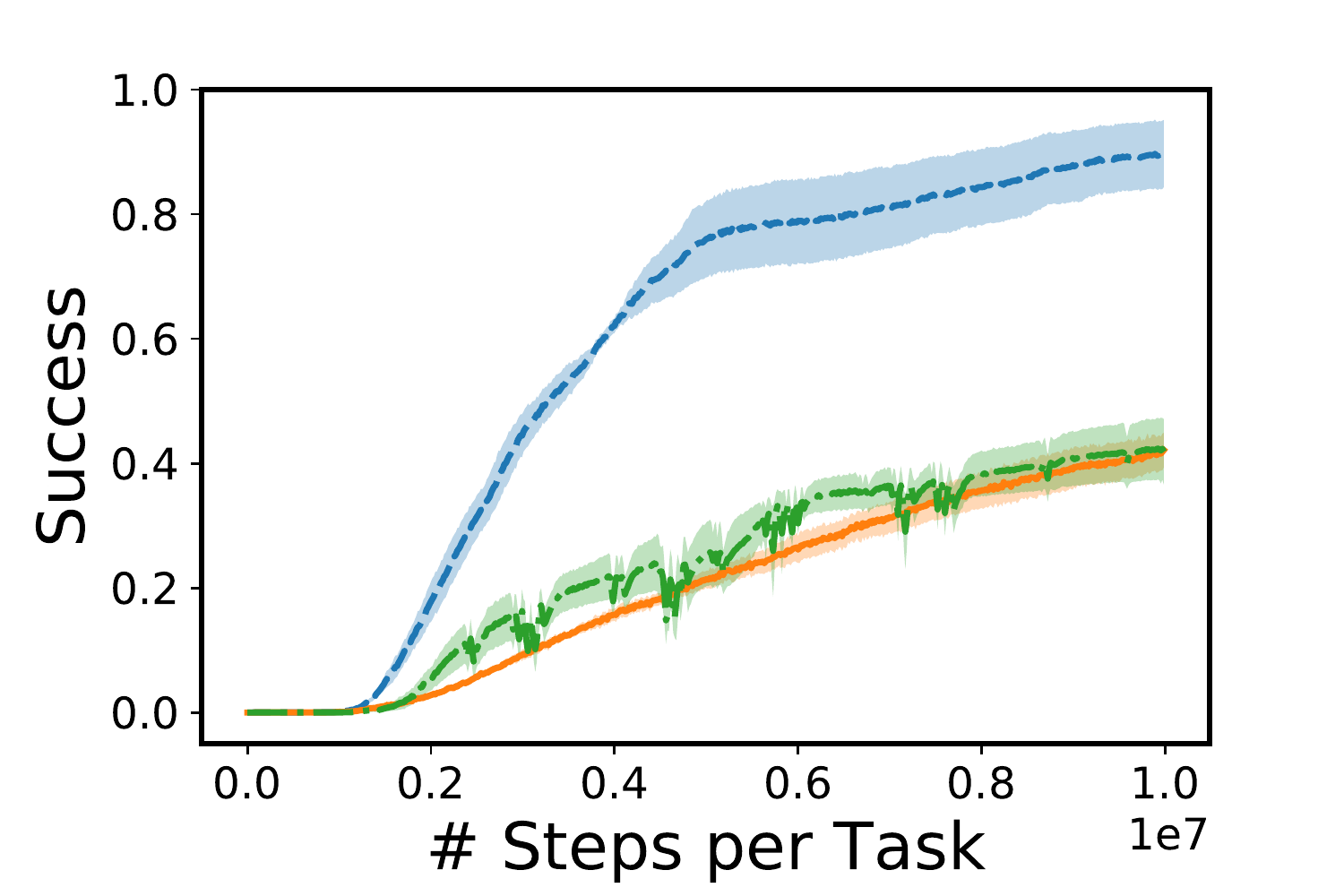}
        \caption{$\numTasks=224$ tasks}
        \label{fig:fullBenchmarkCurvesLarge}
    \end{subfigure}%
    \caption[Training task learning curves on the full \benchmark{} benchmark.]{Evaluation on training tasks for the full \benchmark{} benchmark. The MTL agent was not capable of accelerating the learning substantially with respect to the STL agents according to either metric. The compositional agent, when trained on a larger set of tasks, performed noticeably better, demonstrating that leveraging the compositional structure of \benchmark{} leads to improved training performance. Y-axes span the attainable ranges, and shaded regions represent standard errors across three seeds.}
    \label{fig:fullBenchmarkCurves}
\end{figure}

\subsection{Evaluation of Baselines on the Full \benchmark{} Benchmark}
\label{sec:MTexp}

The first experiment evaluated the agents on the main \benchmark{} benchmark, uniformly sampling tasks for training; Figure~\ref{fig:fullBenchmarkCurves} presents the corresponding learning curves. After training for $10$ million time steps on each task, the STL agents had a success rate of around $40\%$. When the training set was a small portion of the whole set of tasks, the MTL and compositional agents only slightly improved upon the STL agents. However, when training on a larger set of tasks, the compositional agent learned much faster, achieving more than twice as much success. In contrast, MTL results did not improve with the larger training set. This suggests that the MTL agent does not appropriately share knowledge across tasks, and instead separately allocates capacity in the network to different tasks. As it sees more tasks, it progressively exhausts its capacity. Instead, the compositional agent shares components appropriately, and additional training tasks improve the agent's ability to leverage these commonalities. This demonstrates that \benchmark{} tasks are indeed compositionally related, and that exploiting these relations leads to improved performance.

\begin{table}[b!]
    \caption[Zero-shot generalization on the full \benchmark{} benchmark.]{Zero-shot generalization on the full \benchmark{} benchmark. Both agents achieved generalization when trained on many tasks, but fell short from solving the benchmark with fewer training tasks. Standard errors across three seeds reported after the $\pm$.}
    \label{tab:zeroshotFull}
    \centering
        \begin{tabular}{l|cc|cc}
         & \multicolumn{2}{c|}{$T=56$ tasks} & \multicolumn{2}{c}{$T=224$ tasks} \\
         & MTL & Compositional & MTL & Compositional\\
         \hline
        Return & $115.79${\tiny$ \pm 18.01$} & $64.26${\tiny$\pm 7.51$} & $199.98${\tiny$\pm15.69$} & $301.92${\tiny$\pm6.91$} \\
        Success & $0.18${\tiny$\pm 0.11$} & $0.08${\tiny$\pm 0.02$} &  $0.41${\tiny$\pm0.05$} & $0.88${\tiny$\pm0.05$}\\
        \end{tabular}
\end{table}

After training, the experiment evaluated the agents on the \benchmark{} tasks that they did not train on. Intuitively, an agent that correctly decomposes the tasks should achieve high performance on these test tasks by adequately recombining its learned components. Results in Table~\ref{tab:zeroshotFull} show that the learners struggled to generalize to unseen tasks when trained on $\numTasks=56$ tasks, but performed remarkably well (comparatively to their training performance) when trained on $\numTasks=224$ tasks. With the smaller training set, even though the \textit{training} performance was similar for both approaches, the compositional agent achieved substantially worse \textit{zero-shot} performance. This demonstrates that, while the compositional approach can indeed capture the compositional properties of the tasks, this requires observing a large portion of the tasks.
One important question is whether the MTL agent automatically learned compositional knowledge that allowed it to solve unseen tasks. The alternative explanation would be that the agent instead found similar tasks in the training set and used the policy for those for generalization. To test this, a simple experiment found the most similar training task to each test task and used its policy to predict zero-shot performance. Concretely, for every task $\Taskt$ on which the agent achieved some zero-shot success, its closest policy $\pi^{(t')}$ ($t'\neq t$) was the one that performed best on task $\Taskt$; this would have been the best policy to choose, and so one would expect the performance of policy $\pi^{(t')}$ to correlate to that of policy $\pi^{(t)}$. 
However, the coefficients of determination between the policies' success rates were very low: $\mathbf{R^2=0.19}$ and $\mathbf{R^2=0.03}$ for the MTL and compositional agents, respectively. This shows that the generalization was unlikely to come from using trained policies for different tasks, but rather from correctly leveraging the compositional properties of the tasks.

\subsection[Evaluation of Baselines on the Smaller-Scale \benchmark{} $\cap$ IIWA Benchmark]{Evaluation of Baselines on the Smaller-Scale \benchmark{}\emph{$\cap$\IIWA{}} Benchmark}

The next experiment evaluated the three baseline agents on the reduced \benchmark{}$\cap$\IIWA{} benchmark, in order to 1)~propose a computationally cheaper setting to facilitate progress and 2)~shed light on the relative difficulty of generalizing across different \benchmark{} axes. Figure~\ref{fig:smallscaleCurves} contains the learning curves on the training tasks. The relative performance of the compositional agent with respect to the STL agents was similar to that obtained after training on \emph{over $\mathit{200}$ tasks} for the full \benchmark{}, demonstrating that the agent is capable of discovering the compositional structure of this reduced benchmark with far fewer training tasks. On the other hand, the MTL agent performed noticeably better in this simplified setting. Two (compatible) hypotheses potentially explain this improvement: 1)~there are fewer tasks, so the model capacity is not a limiting factor, and 2)~it is easier to transfer knowledge across tasks that use a single robot manipulator, as used by the large majority of multitask RL works; Section~\ref{sec:additionalExperiments} gives evidence toward this latter hypothesis.

\begin{figure}[t!]
\centering
    \begin{subfigure}[b]{\textwidth}
            \centering
                \includegraphics[trim={0.1cm 0.12cm 0.1cm 0.15cm}, clip,height=0.45cm]{chapter6/Figures/legend.pdf}
        \end{subfigure}\\
    \begin{subfigure}[b]{0.45\textwidth}
    \centering
        \includegraphics[height=4.5cm, trim={0.2cm 0.0cm 1.5cm 0.8cm}, clip]{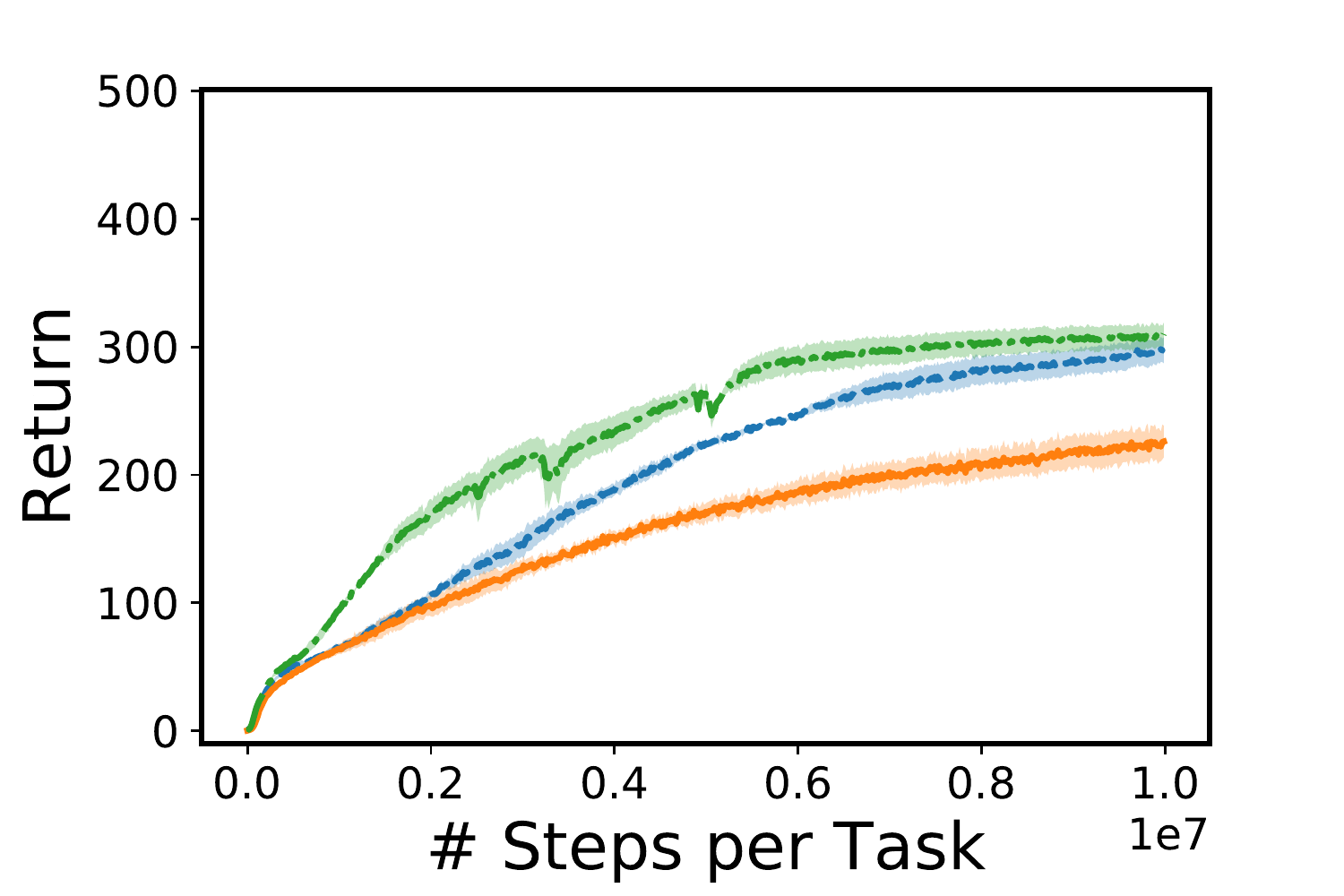}
    \end{subfigure}
    \begin{subfigure}[b]{0.45\textwidth}
    \centering
        \includegraphics[height=4.5cm, trim={.2cm 0.0cm 1.5cm 0.8cm}, clip]{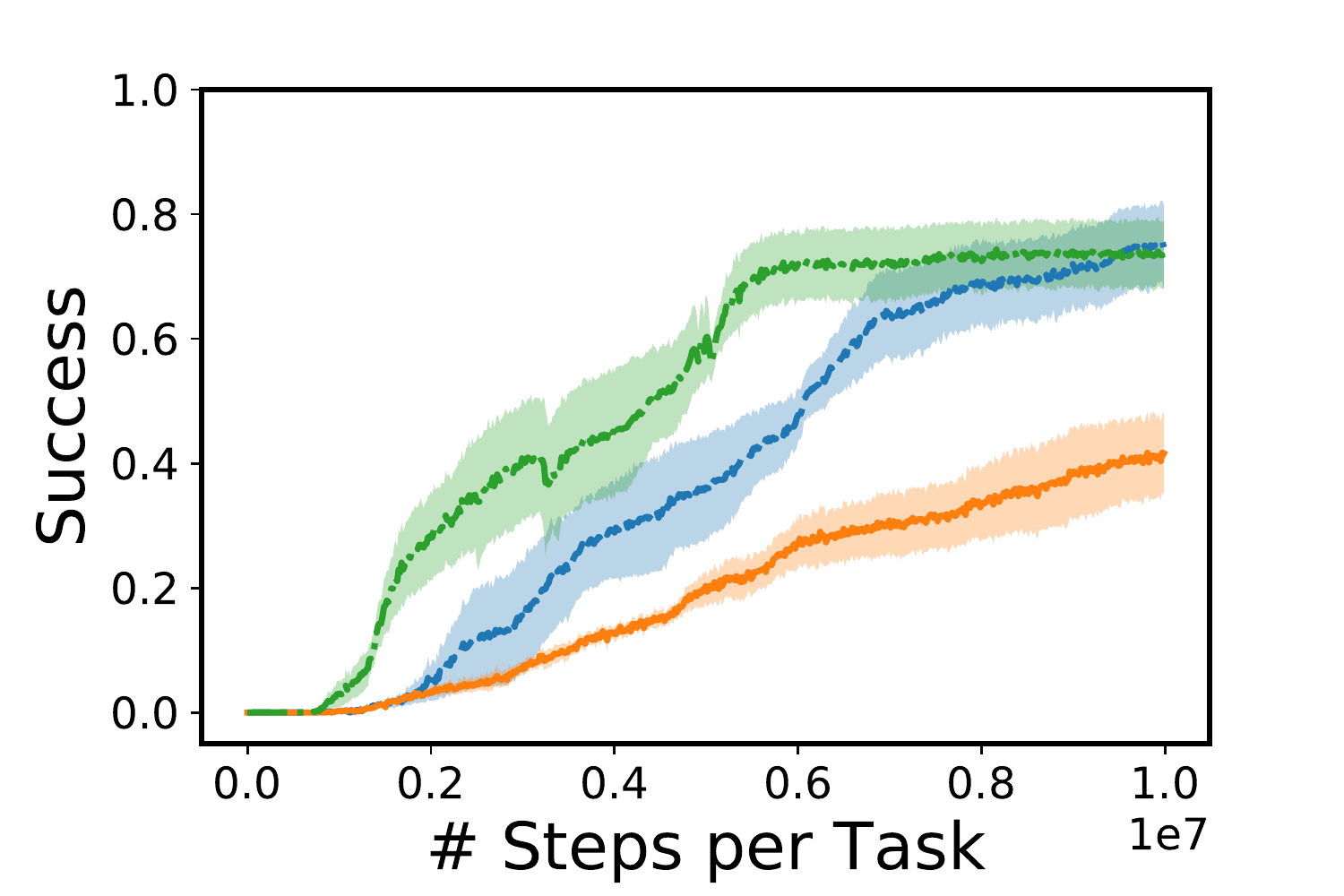}
    \end{subfigure}%
    \caption[Training task learning curves on the smaller-scale  \benchmark{} $\cap$ IIWA benchmark.]{Evaluation on $\numTasks=32$ training tasks for the smaller-scale  \benchmark{}$\cap$\IIWA{} benchmark. The MTL and compositional agents outperformed the STL agents under both metrics. This shows that sharing knowledge across tasks with a single robot arm is easier than across different robot arms when not explicitly leveraging compositionality. Y-axes span the attainable ranges, and shaded regions represent standard errors across three seeds.}
    \label{fig:smallscaleCurves}
\end{figure}

\begin{table}[t!]
    \caption[Zero-shot generalization on the smaller-scale \benchmark{} $\cap$ IIWA  benchmark.]{Zero-shot generalization on the smaller-scale \benchmark{}$\cap$\IIWA{}  benchmark. Similar to the full benchmark, the compositional agent struggled to generalize with few tasks, yet the MTL agent did generalize. Standard errors across three seeds reported after the $\pm$.}    \label{tab:zeroshotSmallscale}
    \centering
        \begin{tabular}{l|cc}
         & MTL & Compositional\\
         \hline
        Return & $232.79${\tiny$\pm 17.48$} & $79.85${\tiny$\pm 19.24$} \\
        Success &  $0.49${\tiny$\pm 0.06$} & $0.12${\tiny$\pm 0.06$}\\
        \end{tabular}
\end{table}

Table~\ref{tab:zeroshotSmallscale} presents the zero-shot results on this reduced benchmark. The MTL agent achieved notably high performance, but the compositional agent was incapable of generalizing, likely due to the small number of training tasks. Recall that the compositional agent only trains each module on the subset of tasks that shares that module. Consequently, each parameter was trained on a small number of tasks, which was insufficient for zero-shot generalization.

\subsection[Evaluation of Baselines on the Restricted \benchmark{} \textbackslash{} Pick-and-Place Benchmark]{Evaluation of Baselines on the Restricted \benchmark{}\emph{\textbackslash{}\pickPlace{}} Benchmark}

The evaluation results presented so far consider a relatively simple compositional problem: the agent trains on multiple combinations of all components, and seeks to generalize to new combinations. These previous results already expose the shortcomings of existing approaches in the compositional setting. However, future approaches that solve these simple settings would still fall short from achieving the full spectrum of compositional capabilities. In particular, agents should ideally learn components that generalize to unseen tasks even if they only see those components in one single combination. To study this setting, the next experiment evaluated the three agents on the restricted \benchmark{}\textbackslash{}\pickPlace{} benchmark. Figure~\ref{fig:holdoutCurves} shows the learning performance over the training tasks, which include exactly one \pickPlace{} task. Performance was close to that on the full benchmark, because the training distributions are similar: there are $55$ combinations of $15$ components (plus one single \pickPlace{} task), compared to $56$ combinations of $16$ components.

\begin{figure}[t!]
\centering
    \begin{subfigure}[b]{\textwidth}
            \centering
                \includegraphics[trim={0.1cm 0.12cm 0.1cm 0.15cm}, clip,height=0.45cm]{chapter6/Figures/legend.pdf}
        \end{subfigure}\\
    \begin{subfigure}[b]{0.45\textwidth}
    \centering
        \includegraphics[height=4.5cm, trim={0.2cm 0.0cm 1.5cm 0.8cm}, clip]{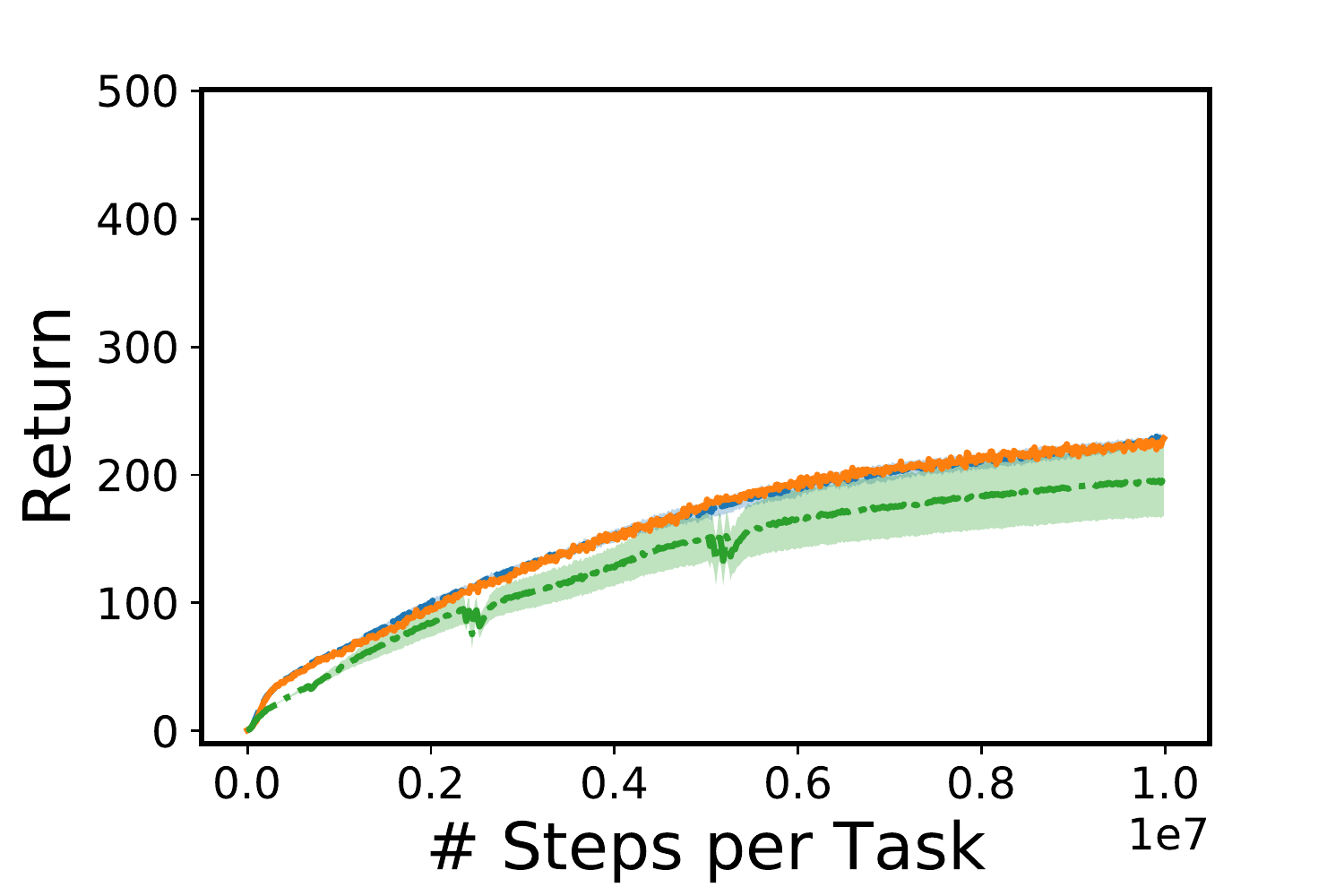}
    \end{subfigure}
    \begin{subfigure}[b]{0.45\textwidth}
    \centering
        \includegraphics[height=4.5cm, trim={.2cm 0.0cm 1.5cm 0.8cm}, clip]{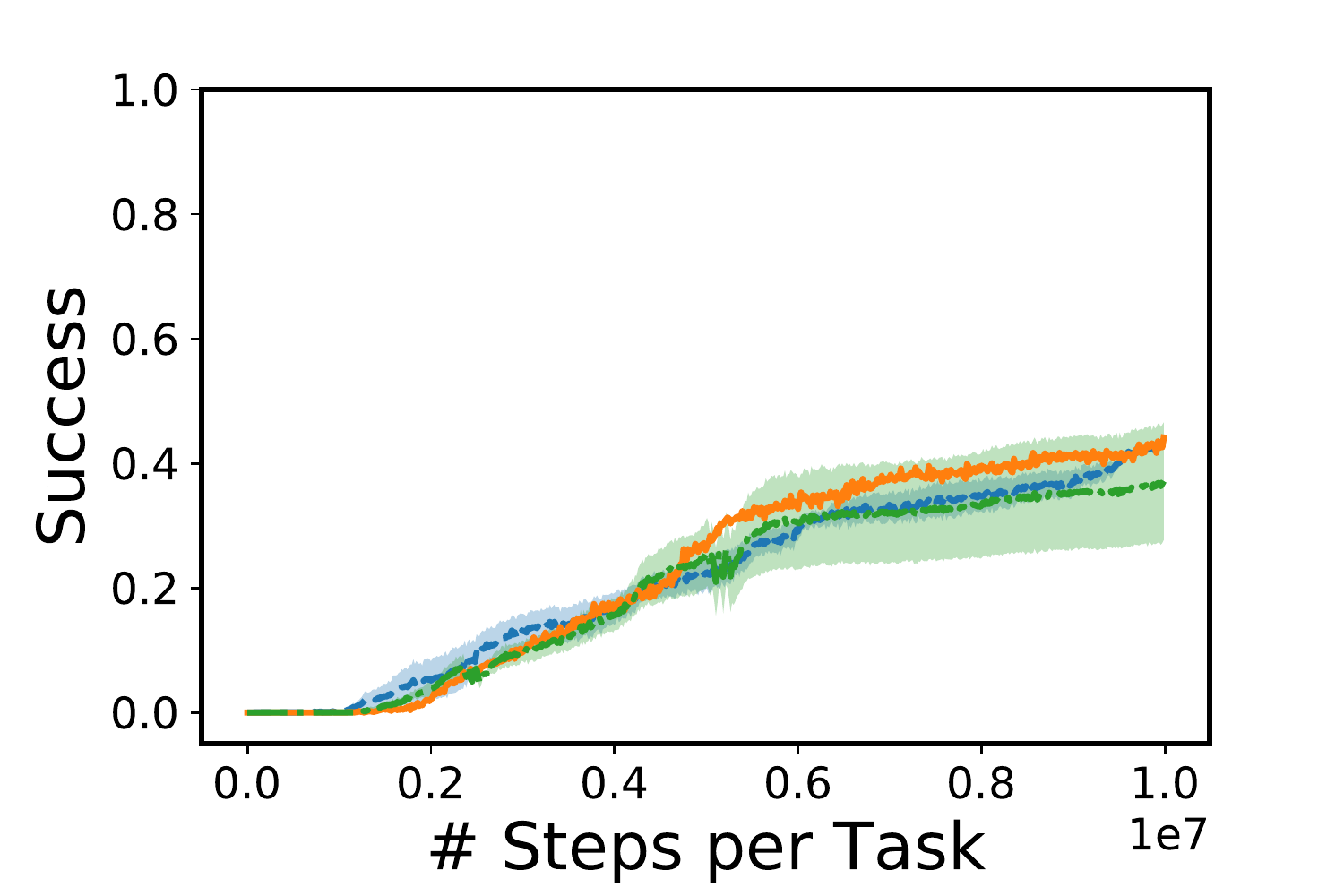}
    \end{subfigure}%
    \caption[Training task learning curves on the restricted \benchmark{} \textbackslash{} pick-and-place benchmark.]{Evaluation on $\numTasks=32$ training tasks for the restricted \benchmark{}\textbackslash{}\pickPlace{} benchmark. The training set included a single \pickPlace{} task, and all other tasks were from the remaining objectives. Results are nearly identical to those on the full \benchmark{}, as expected---major differences were expected in \textit{zero-shot} performance. Y-axes span the attainable ranges, and shaded regions represent standard errors across three seeds.}
    \label{fig:holdoutCurves}
\end{figure}

\begin{table}[b!]
    \caption[Zero-shot generalization on the restricted \benchmark{} \textbackslash{} pick-and-place  benchmark.]{Zero-shot generalization on the restricted \benchmark{}\textbackslash{}\pickPlace{} benchmark, only over \pickPlace{} tasks. Both agents failed at generalizing substantially to the restricted objective. Standard errors across three seeds reported after the $\pm$.}    
    \label{tab:zeroshotHoldout}
    \centering
        \begin{tabular}{l|cc}
         & MTL & Compositional\\
         \hline
        Return  & $74.61${\tiny$\pm9.97$} & $16.63${\tiny$\pm6.71$}\\
        Success & $0.09${\tiny$\pm0.04$} & $0.01${\tiny$\pm0.01$} \\
        \end{tabular}
\end{table}

Table~\ref{tab:zeroshotHoldout} shows that both agents failed to generalize to the unseen \pickPlace{} tasks in this restricted setting. The small amount of zero-shot generalization achieved by the MTL agent was almost entirely on tasks with the same robot arm that was used in the single \pickPlace{} training task, as shown in Table~\ref{tab:deaggreagtedZeroshotHoldout}.  In contrast, the compositional agent was completely incapable of generalizing to unseen \pickPlace{} tasks.

\begin{table}[t!]
    \centering
    \caption[Zero-shot success for the MTL agent on \benchmark{} \textbackslash{} pick-and-place for shared and not-shared elements.]{Zero-shot success for the MTL agent on \benchmark{}\textbackslash{}\pickPlace{}, separated by tasks that share (or not) each element with the trained \pickPlace{} task (e.g., if the training \pickPlace{} task used the \IIWA{} arm, \IIWA{} tasks go on the left and non-\IIWA{} tasks go on the right). Most generalization was on tasks that shared the training robot. Standard errors across three seeds reported after the $\pm$.}
    \label{tab:deaggreagtedZeroshotHoldout}
    \begin{tabular}{@{}l|c|c@{}}
        Element & Trained & Untrained \\
         \hline
        robot    & $0.30${\tiny$\pm0.10$} & $0.03${\tiny$\pm0.02$} \\
        object   & $0.14${\tiny$\pm0.04$} & $0.08${\tiny$\pm0.04$} \\
        obstacle & $0.14${\tiny$\pm0.04$}  & $0.08${\tiny$\pm0.03$} \\
    \end{tabular}
\end{table}

\subsection{Empirical Analysis of \benchmark{} Properties}
\label{sec:additionalExperiments}

The following evaluations shifted focus to verifying two important properties of \benchmark{}: that the large majority of tasks are learnable by current RL mechanisms, and that the tasks are not only compositional, but also highly varied.

\paragraph{Learnability of tasks}The combination of elements into the combinatorially many tasks in \benchmark{} opens up the question of whether some of these configurations might lead to potentially unsolvable tasks for current RL algorithms. For example, there might be configurations that restrict the physical space in such a way that the robot arm cannot fulfill a task objective.  If \benchmark{} tasks were unsolvable by current RL methods, that would conflate the difficulty of compositional reasoning with the difficulty of solving RL tasks. 
To validate that this is not the case, this experiment gave each task a score corresponding to the performance  of the best agent across all those trained so far (taking the maximum across experiments \textit{and} random seeds). For any task with a score of $0$, experiments so far provide no evidence that the task is learnable, because no agent solved it to any extent. Figure~\ref{fig:maximum_success} shows the result of this computation. Only \textit{a single task} received a score of $0$ (place the \plate{} on the \shelf{} with the \Panda{} arm avoiding the \goalWall{}), indicating that it {\em may} be unlearnable. This corroborates that almost all tasks are solvable with existing RL methods.

\begin{figure}[t!]
\centering
    \hfill
    \begin{subfigure}[b]{1.\textwidth}
        \centering
            \includegraphics[trim={0.1cm 0.1cm 0.1cm 0.15cm}, clip,height=0.45cm]{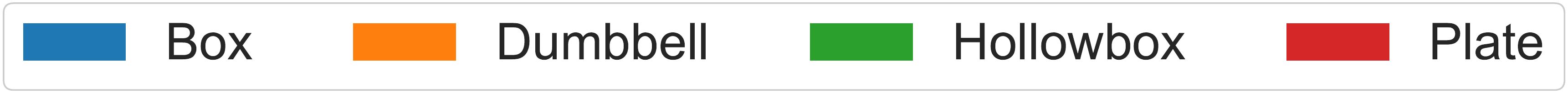}
    \end{subfigure}\hfill
    \hspace{0.16\textwidth}
    \vspace{-1em}
    \begin{subfigure}[b]{1.\textwidth}
        \centering
            \includegraphics[width=\linewidth, trim={0.3cm 0.0cm 1.5cm 0.8cm}]{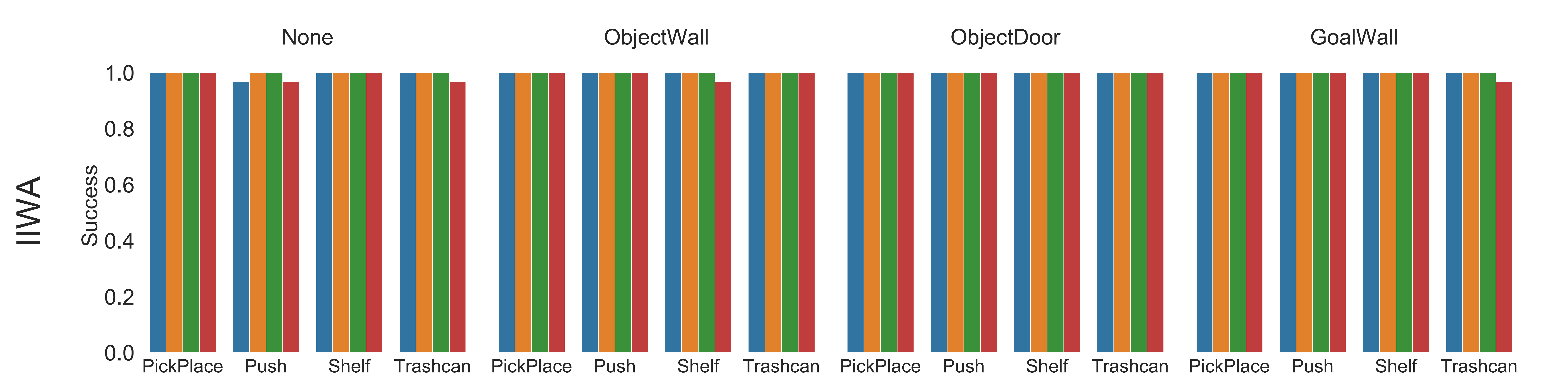}
    \end{subfigure}\\
    \hfill
    \begin{subfigure}[b]{1.\textwidth}
        \centering
            \includegraphics[width=\linewidth, trim={0.3cm 0.0cm 1.5cm 0.8cm}]{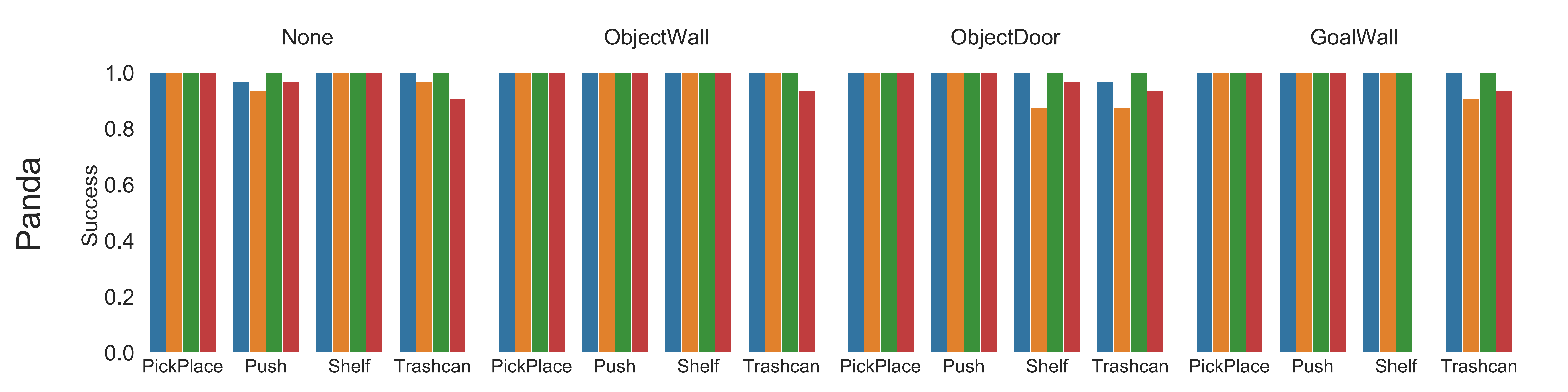}
    \end{subfigure}\\
    \hfill
    \begin{subfigure}[b]{1.\textwidth}
        \centering
            \includegraphics[width=\linewidth, trim={0.3cm 0.0cm 1.5cm 0.8cm}]{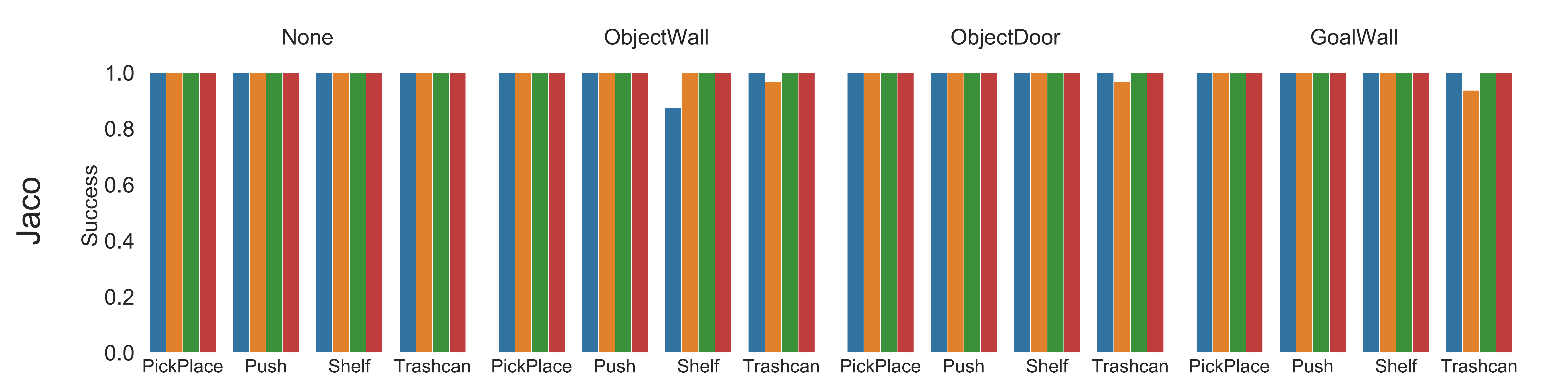}
    \end{subfigure}\\
    \hfill
    \begin{subfigure}[b]{1.\textwidth}
        \centering
            \includegraphics[width=\linewidth, trim={0.3cm 0.0cm 1.5cm 0.8cm}]{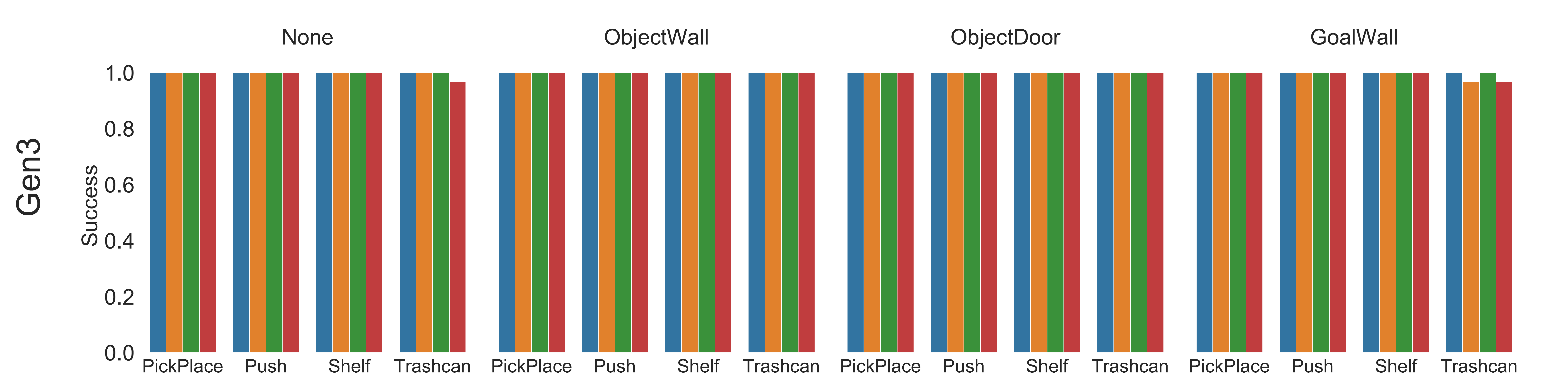}
    \end{subfigure}\\
    \vspace{-0.5em}
    \caption[Maximum success rate attained for each \benchmark{} task across all trained agents.]{Maximum success rate attained for each task across all trained agents. All tasks except one were solved at least once.}
    \label{fig:maximum_success}
\end{figure}

\begin{figure}[t!]
\centering
    \begin{subfigure}[b]{\textwidth}
        \centering
            \includegraphics[trim={0.1cm 0.1cm 0.1cm 0.1cm}, clip,height=0.35cm]{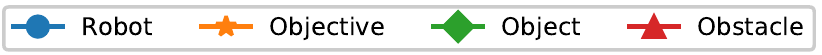}
    \end{subfigure}\\
    \begin{subfigure}[b]{0.26\textwidth}
        \centering
            \includegraphics[height=3.2cm, trim={0.0cm, 0.15cm, 0.35cm, 0.6cm}, clip]{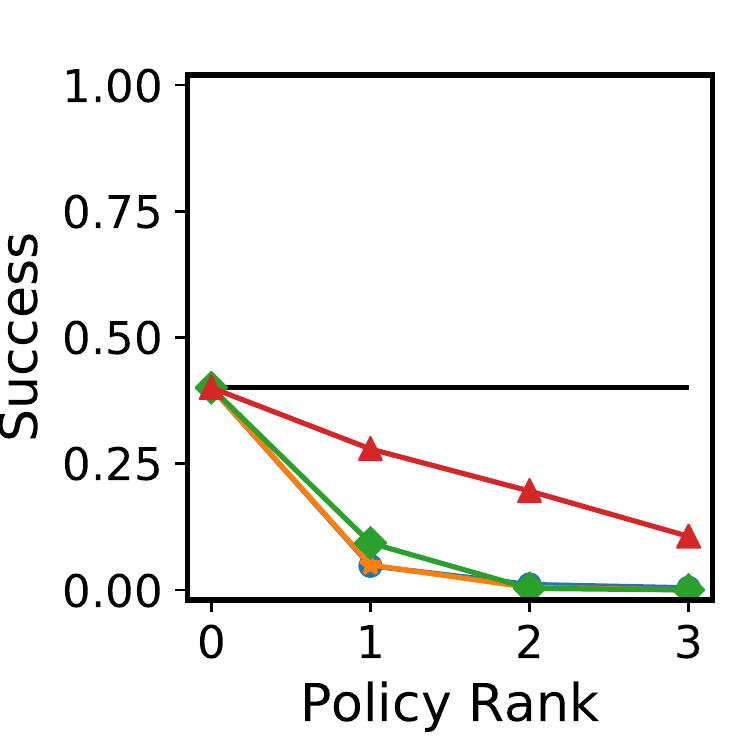}
        \caption*{MTL train}
    \end{subfigure}%
    \begin{subfigure}[b]{0.243\textwidth}
        \centering
            \includegraphics[height=3.2cm, trim={0.65cm, 0.15cm, 0.35cm, 0.6cm}, clip]{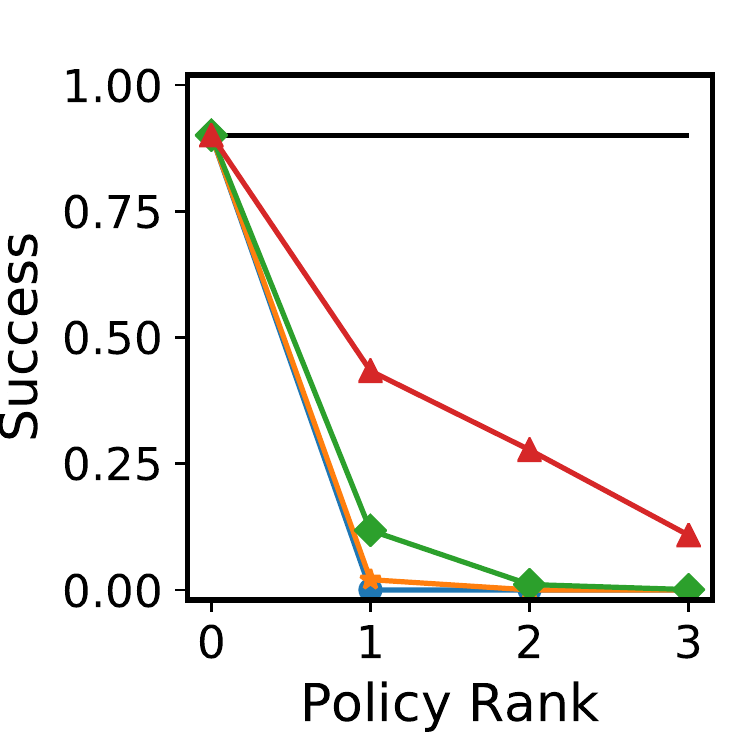}
        \caption*{Compositional train}
    \end{subfigure}%
    \begin{subfigure}[b]{0.248\textwidth}
        \centering
            \includegraphics[height=3.2cm, trim={0.5cm, 0.15cm, 0.35cm, 0.6cm}, clip]{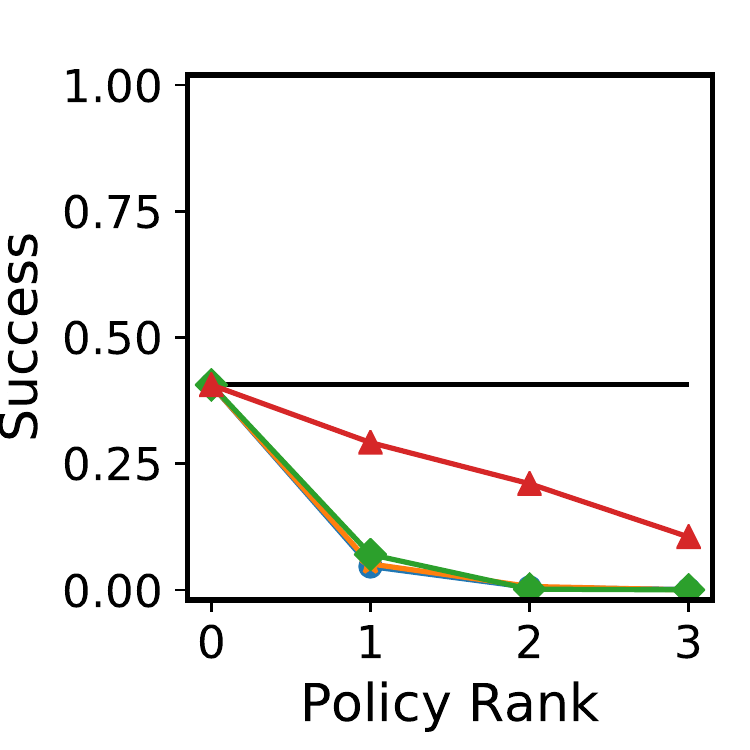}
        \caption*{MTL zero-shot}
    \end{subfigure}%
    \begin{subfigure}[b]{0.248\textwidth}
        \centering
            \includegraphics[height=3.2cm, trim={0.5cm, 0.15cm, 0.35cm, 0.6cm}, clip]{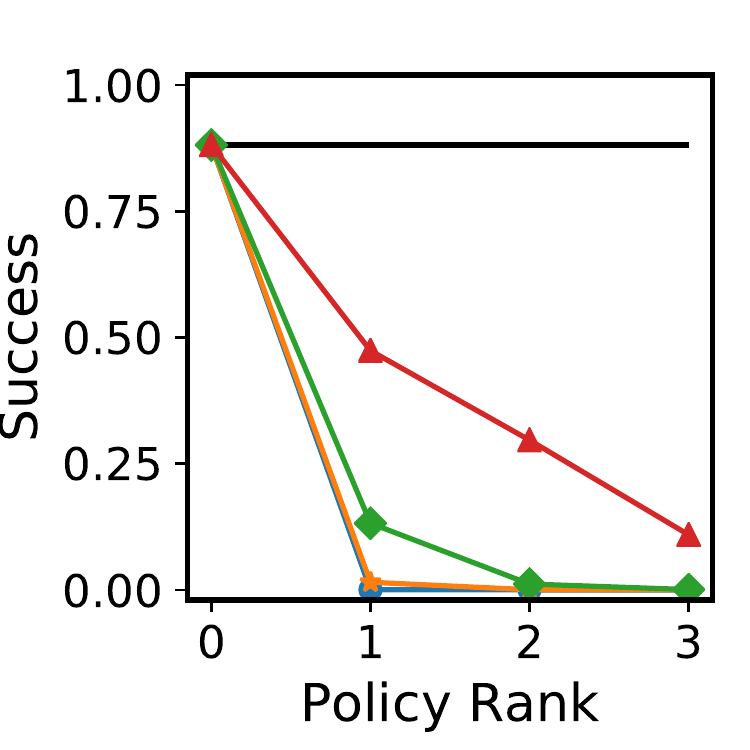}
        \caption*{Compositional zero-shot}
    \end{subfigure}%
    \caption[Task similarity of \benchmark{} tasks.]{Performance when providing incorrect task descriptors, with a single changed component: position $0$ corresponds to the correct descriptor, and positions $j>0$ correspond to the $j$-th ranked descriptors of each type. Using the wrong descriptors leads to poor performance, confirming the diversity of \benchmark{} tasks.}
    \label{fig:task_sim_mtl}
\end{figure}

\paragraph{Diversity of tasks} Another valid concern is that it might be possible for the agent to solve multiple tasks with a single policy if the tasks are very similar, implying that compositional reasoning is not necessary for generalization. To verify that this is not the case in \benchmark{}, after training the MTL and compositional agents over $\numTasks=224$ tasks, this evaluation assessed their performance if they observed the \textit{incorrect} task descriptor. In particular, for a given task $\Taskt$, the agent acted on the environment of task $\Taskt$ using the descriptors for all tasks $\Taskt[t^\prime]$ that varied in a single component from $\Taskt$ (i.e., the tasks most similar to $\Taskt$). Figure~\ref{fig:task_sim_mtl} shows the result of sorting the performances with these incorrect descriptors separately for each axis to find the rank of each incorrect component (e.g., the rank-$2$ robot for task $\Taskt$ is the robot that achieved the second-best performance when used as the descriptor for task $\Taskt$), and averaging the sorted performances. The results demonstrate that using the incorrect robot, objective, or object descriptor consistently leads to substantially degraded performance, particularly for the compositional agent; this means that the agent is specialized to each of the components. Using the incorrect obstacle descriptor has a much smaller impact, particularly for the MTL agent. This shows that the MTL agent learns policies that are somewhat agnostic to the obstacles, which is possible for example by avoiding all possible obstacle regions.
Additionally, these results show that varying the robot arm causes the most drastic drop in performance, demonstrating that solving tasks with varied robots is a challenging problem, yet existing benchmarks consider tasks with a single robot arm.

\section{Scope, Limitations, and Extensions}

The design of \benchmark{} emphasized studying the compositional properties of multitask RL algorithms. As such, while it can be used to investigate multiple other problems, it is not intended to cover the spectrum of open questions in multitask RL. This section discusses limitations and potential extensions to the use of \benchmark{}.

\paragraph{Reliance on PPO} To provide a fair comparison across single-task and multitask learners, all agents used PPO~(\citealp{schulman2017proximal}) as the base RL algorithm, built off of the \texttt{Spinning Up} implementation~\citep{SpinningUp2018}. While future research can use any base learning method, only evaluations that use the same PPO implementation could fairly compare against the benchmarking results presented here.

\paragraph{Input space} The input space used in the evaluation is a 94-dimensional, symbolic description of the environment grounded in the system dynamics. However, there is also broad interest from the robot learning community in RL with richer observations (e.g., visual inputs). While such an evaluation falls outside the scope of this dissertation, the benchmark implementation allows users to request a multicamera visual observation instead of the low-dimensional observation.

\paragraph{Task descriptors} Part of the observation space is a multi-hot indicator that describes the components that make up the current task. While this permits assessing the interesting property of zero-shot compositional generalization, there are other questions that might benefit from withholding this information from the agent. As one example, the agent might only receive a task index that indicates \textit{which} task it must solve, but not \textit{how} it relates to other tasks. Alternatively, the observation could give the agent no indication of the current task at all and require it to extract this information from data. Note that the symbolic observation does not contain sufficient information to unequivocally identify the task without task descriptors, and so the agent would need to extract this information from \textit{trajectories} of interaction instead. On the other hand, the images in the visual observations do contain sufficient information to differentiate the tasks.

\paragraph{Other forms of composition} While \benchmark{} was designed around functional composition as described in Chapter~\ref{cha:RL}, the benchmark can also be used for other forms of composition. In particular, the standardization of the environments and the use of stage-wise rewards makes this a useful domain for evaluating skill discovery and sequencing. For example, the agent could learn skills for reaching a location, grasping an object, and lifting, all of which are useful for multiple \benchmark{} tasks. Note that standard representations of skills would only work for one individual arm.

\paragraph{Lifelong learning} Another very natural extension of \benchmark{} is to use it in the lifelong learning setting as defined in Chapter~\ref{cha:RL}. The agent would face \benchmark{} tasks one after the next, and need to perform well on all previously seen tasks. The goal of the agent would be to learn each new task as quickly as possible by leveraging accumulated knowledge, and to retain performance on the earlier tasks upon training on new tasks. Given the sequential nature of lifelong learning, it might be prohibitively expensive to train the agent over the full variant of \benchmark{}, but the smaller-scale variants described in Section~\ref{sec:sampleTrainingTasks} would be feasible; the experiments of Chapter~\ref{cha:RL} already used sequences of tasks of similar lengths. 

\paragraph{Sim2real transfer} Learning a multitude of tasks in simulation is a common strategy used to transfer policies from simulation to the real world (sim2real). Since \benchmark{} uses simulated versions of four robot arms that are commercially available, it could additionally be leveraged to study this promising direction.

\section{Summary}
This chapter introduced \benchmark{}, a large-scale robotic manipulation benchmark for studying the novel problem of functionally compositional RL introduced in this dissertation. \benchmark{} builds upon the simpler benchmarks presented in Chapter~\ref{cha:RL}, leveraging the power of combinatorics to create hundreds of highly diverse tasks and opening the door for the study of multitask and lifelong RL at scale. In particular, \benchmark{} is designed to study the ability of approaches to discover the decomposition of complex problems into simpler subproblems whose solutions can be combined to solve the overall task. Once appropriate components have been found, they could be combined in novel ways to solve \textit{new} RL problems that the agent has never trained on. 

The empirical evaluation assessed the performance of two MTL approaches: a monolithic agent that uses a single network to solve all tasks, and a modular agent that composes different policies for each task from a set of shared components. The results on a range of conditions under \benchmark{} demonstrate that these existing methods  show promising compositional properties, but these results also expose that existing methods are far from solving the problem of compositional RL. Progress in that direction will enable RL approaches to automatically detect commonalities across diverse problems, leverage these commonalities to facilitate learning, and eventually handle far more complex tasks than is possible today.

\biblio

\chapter{Conclusion}
\label{cha:Conclusion}

This dissertation presented a thorough investigation of the novel problem of lifelong learning of functionally compositional knowledge. The central question that this dissertation sought to answer was \textit{how can ML agents learn compositional knowledge structures in a lifelong learning setting}. The key technical contribution toward answering this question was a general-purpose framework, a sort of algorithm sketch, that permits agents to discover these compositional structures in a variety of settings. The answer to the fundamental question, derived from this framework, is that the design of algorithms for lifelong compositional learning should consider parts of the problem in separate stages. 

The two primary stages, named \textit{assimilation} and \textit{accommodation} due to their connection to Piagetian theory, alternate between discovering the best combination of components for the current task and learning the best instantiation of those components for all seen tasks. Concretely, in the assimilation stage, the agent should discover the best way to combine its existing components into a solution to the current task. As examples, the algorithms in this dissertation use backpropagation, PG training, and discrete optimization as the mechanisms for tackling this problem. In the accommodation stage, the agent should incorporate any new knowledge it discovered during training on the new task to improve its accumulated knowledge. This dissertation experimented with various choices, including regularization, experience replay, and a closed-form approximate optimization. In particular, this dissertation developed an experience replay method via off-line RL that replicates the benefits of replay data from the supervised setting.

In order for the second stage to succeed, it is critical that the agent finds a reasonably good solution in the first stage. Otherwise, the combination of existing components would be of low quality and any knowledge accumulated during assimilation would be insufficient for proper accommodation. This creates the need for a preliminary \textit{initialization} stage, where the agent sets up its initial set of components, encouraging them to be reusable across multiple tasks. 

An extensive empirical evaluation, with various algorithms, benchmarks, and learning paradigms, consistently yielded that this separation of the learning process into stages enables substantially improved lifelong learning performance, supporting the thesis of this dissertation (Section~\ref{sec:Thesis}). This improvement manifested itself in the form of forward transfer, avoidance of forgetting, backward transfer, and limited growth. In addition, the evaluation led to the conclusion that compositional lifelong learners can learn highly diverse tasks, which noncompositional learners systematically fail to achieve. 

\section{Summary of Technical Contributions}

Along the way to answer these questions, this dissertation developed a number of technical contributions, summarized as follows:
\begin{enumerate}
    \item A formalization of the novel problem of lifelong compositional learning in terms of a compositional problem graph.
    \item A set of nine lifelong learning algorithms for discovering compositional structures in the supervised setting. Each algorithm combines one structural model configuration from linear model combinations, soft layer ordering, and soft gating, with one mechanism for avoiding forgetting, between na\"ive fine-tuning, EWC, and experience replay.
    \item Component dropout, an approach that enables a lifelong learner to automatically detect when it must incorporate a new module into its model, by approximately training a version of the model with the new module and a version without the new module.
    \item A formalization of the lifelong compositional RL problem in terms of functional composition, leveraging the graph formalism from the supervised setting.
    \item \lpgftw{}, an efficient lifelong PG learning algorithm with theoretical guarantees, and with strong empirical performance on a sequence of $\numTasks=48$ highly diverse robotic manipulation tasks.
    \item \compRL{}, an algorithm that uses explicitly modular architectures to learn decomposed solutions to compositional RL tasks. The method performed well on two long sequences of diverse tasks in discrete- and continuous-action settings, achieving zero-shot, forward, and backward transfer.
    \item Off-line RL replay, a general-purpose mechanism for avoiding forgetting in multistage lifelong RL approaches, that demonstrably reduces forgetting in compositional and noncompositional settings.
    \item \NonstatRL{}, a simple extension to \compRL{} that handles nonstationary lifelong sequences of tasks, where the variations in the environment occur independently along multiple dimensions.
    \item \benchmark{}, a large-scale simulated robotic manipulation benchmark for the study of compositional, multitask, and lifelong RL. 
\end{enumerate}

Despite the high performance of the proposed algorithms, this dissertation did not focus on the creation of highly optimized methods toward achieving state-of-the-art performance. Instead, the purpose of these methods was to expose concepts such as multistage training and compositionality, which will hopefully become accessible to the general ML community. In particular, the hope was to demonstrate that these intuitively appealing ideas of modularity and compositionality, with simple training mechanisms, enable lifelong learning agents to solve task sequences that were previously out of reach.

\section{Future Directions}

This dissertation posed the novel problem of lifelong learning of functionally compositional knowledge, both in the supervised and the RL settings. As a nascent field, lifelong compositional learning has a range of open questions, which future investigations should tackle. This (speculative) section elucidates a subset of such questions that are most likely to substantially impact the broader field of AI.

\paragraph{Real-world applications} While the evaluations in this dissertation were inspired by realistic robotic applications, it remains an open question how the proposed approaches would fare upon deployment on physical robots. More broadly, this is a challenge faced by the larger research field. Benchmark data sets and RL environments enable fast development and fair performance comparisons, both of which are useful for accelerating progress. However, \textit{applied} research should progress. In particular, lifelong learning has so far been disconnected from real-world deployments, partly because of the artificial nature of task-based lifelong learning. Future work tackling realistic applications, ideally involving embodied agents, that complement fundamental lifelong learning developments would have massive impact.

\paragraph{Task-free lifelong learning} As hinted at above, one significant step in the direction of deployed lifelong learning would be to move away from the task-based lifelong learning formulation. Some recent works have placed efforts in this direction, but this still remains a severely underdeveloped area. In particular, it is critical that future instantiations of the problem do not assume that individual inputs contain sufficient information to determine the agent's current objective. This assumption, which most task-agnostic works to date make, is unfortunately unrealistic. Instead, real-world (embodied) lifelong learning would require the agent to explore and study the environment over a stream of temporally correlated inputs to discover its current objective.

\paragraph{Flexible compositionality} This dissertation evaluated approaches in two settings: noncompositional and compositional. Noncompositional evaluations served to demonstrate the flexibility of the methods, while compositional evaluations served to study the compositional properties of the algorithms. The real world is neither of these two extremes: it has a multitude of compositional properties, but many tasks require highly specialized knowledge. Devising techniques (and corresponding evaluation domains) that explicitly reason about when compositional or specialized knowledge is required would constitute another significant step toward deployed lifelong learning.  

\paragraph{Other forms of composition} The approaches in this dissertation all tackle the problem of functional compositionality. Yet, as discussed extensively in Chapter~\ref{cha:RelatedWork}, there are numerous other forms of composition. Specifically, the RL community has developed a variety of temporal, representational, logical, and morphological views of the notion of compositionality. Each of these formulations is promising toward developing agents that accumulate knowledge and compose it in combinatorially many ways to solve a wide range of diverse tasks. Intuitively, the high-level idea of separating the lifelong learning process into initialization, assimilation, and accommodation stages would permit learning these various forms of composition. Developing concrete instantiations of this intuition could potentially be highly impactful. 

\paragraph{Moving beyond deep learning} This final comment, part recommendation for future work and part reflection, encourages future work to look beyond deep learning in the development of lifelong and compositional learners. This dissertation leveraged neural net modules as the main form of compositional structures. The reason for this choice was primarily practical: neural networks and backpropagation are today the most powerful tools available in ML, and they permitted abstracting away the many nuances of statistical learning and optimization, and focus instead on the intuition of knowledge compositionality. Partially-non-deep-learning portions of this dissertation considered linear classifiers in Chapter~\ref{cha:Supervised}, as well as linear policies, closed-form approximate optimization, and discrete search in Chapter~\ref{cha:RL}. Historical evidence suggests that the tools of the future will be different from (the current version of) deep learning, and consequently this dissertation encourages future research to not focus exclusively on deep learning, and develop approaches to lifelong compositional learning that look outside of deep learning as well. 

\biblio

\end{mainf}

\appendix

\newenvironment{appendixf}{}{}
\titleformat{\chapter}[hang]{\large\center}{APPENDIX \thechapter}{0 pt}{ : } 
\titlespacing*{\chapter}{0pt}{-33 pt}{6 pt} 

\begin{appendixf}
\addtocontents{toc}{\protect\setcounter{tocdepth}{-1}} 
\cleardoublepage
\addtocontents{toc}{\protect\setcounter{tocdepth}{1}} 
\addtocontents{toc}{\protect\renewcommand{\protect\cftchappresnum}{APPENDIX }}
\begin{landscape}
\chapter{Categorized Related Works on Lifelong Learning and Compositional Learning}
\label{app:RelatedWork}

\footnotesize
\begin{longtable}{llllllp{9cm}}
\caption[Related work categorization.]{A categorization of existing works into six axes. The vast majority of work on lifelong learning has not learned explicitly compositional structures, while most efforts on compositional learning have operated in the MTL or STL settings.}\\
\toprule
Comp.~type & Lifelong? & Mechanism & Struc.~given & Struc.~type & Domain &                                                                                                                                                                                                                                                                                                                                                                                                                                                                                                                                                                                                                                                                                                                                                                                                                                                                                                                                                                                                                                                                                                                                                                                                                                                                                                                                                                                                 References \\
\midrule
\midrule
\endfirsthead
\toprule
Comp.~type & Lifelong? & Mechanism & Struc.~given & Struc.~type & Domain &                                                                                                                                                                                                                                                                                                                                                                                                                                                                                                                                                                                                                                                                                                                                                                                                                                                                                                                                                                                                                                                                                                                                                                                                                                                                                                                                                                                                 References \\
\midrule
\midrule
\endhead
           None &       Lifelong &        Supervised &                 No &          None &      Vision &  \citet{kirkpatrick2017overcoming}, \citet{ritter2018online}, \citet{zenke2017continual}, \citet{chaudhry2018riemannian}, \citet{serra2018overcoming}, \citet{yoon2020scalable}, \citet{jung2020continual}, \citet{cha2021cpr}, \citet{chaudhry2020continual}, \citet{saha2021gradient}, \citet{deng2021flattening}, \citet{duncker2020organizing}, \citet{nguyen2018variational}, \citet{ahn2019uncertainty}, \citet{loo2021generalized}, \citet{zhang2021variational}, \citet{kumar2021bayesian}, \citet{li2017learning}, \citet{benjamin2018measuring}, \citet{titsias2020functional}, \citet{pan2020continual}, \citet{wang2021afec}, \citet{chaudhry2019tiny}, \citet{lopez2017gradient}, \citet{chaudhry2018efficient}, \citet{riemer2018learning}, \citet{gupta2020lookahead}, \citet{mirzadeh2021linear}, \citet{raghavan2021formalizing}, \citet{guo2020improved}, \citet{pham2021contextual}, \citet{knoblauch2020optimal}, \citet{derakhshani2021kernel}, \citet{oswald2020continual}, \citet{henning2021posterior}, \citet{shin2017continual}, \citet{singh2020calibrating}, \citet{yoon2018lifelong}, \citet{hung2019compacting}, \citet{adel2020continual}, \citet{lee2021sharing}, \citet{ke2020continual}, \citet{hurtado2021optimizing}, \citet{yoon2021federated}, \citet{mirzadeh2020understanding}, \citet{ramasesh2021anatomy}, \citet{lee2021continual}, \citet{ehret2021continual}, \citet{schwarz2018progress} \\
\bottomrule
\end{longtable}

\end{landscape}

\chapter{Full Results on Lifelong Compositional Supervised Learning}
\label{app:FullResultsSupervised}

For completeness, this appendix includes expanded results from Figures~\ref{fig:softOrderingBars} and \ref{fig:softOrderingCurves}, corresponding to soft layer ordering. Figure~\ref{fig:softOrderingBarsNotAveraged} is a more detailed version of Figure~\ref{fig:softOrderingBars}, and shows the test accuracy immediately after each task was trained and after all tasks had been trained, separately for each data set. Compositional algorithms conforming to the proposed framework achieve a better trade-off than others in flexibility and stability, leading to good adaptability to each task with little forgetting of previous tasks. Similarly, Figure~\ref{fig:softOrderingCurvesNotAveraged} shows learning curves like those in Figure~\ref{fig:softOrderingCurves}, for each data set. Baselines that train components and structures jointly all exhibit a decay in the performance of earlier tasks as the learning of future tasks progresses, whereas methods conforming to the proposed framework do not. Results for soft gating nets display a similar behavior.

\begin{figure}[H]
\centering
    \begin{subfigure}[b]{0.35\textwidth}
        \includegraphics[height=2.5cm, trim={0cm 0.5cm 0cm 1.8cm}, clip]{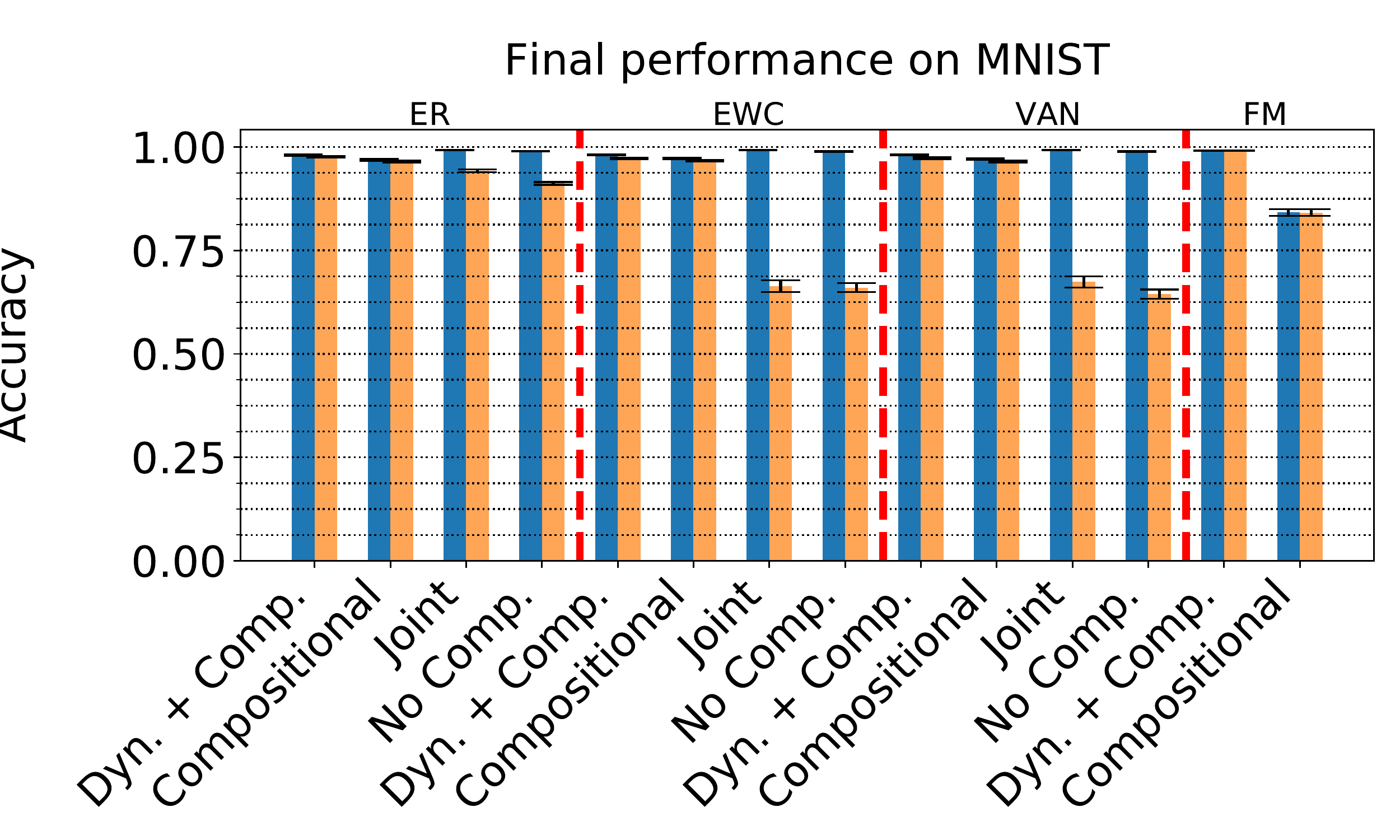}
        \caption{\MNIST}
    \end{subfigure}%
    \begin{subfigure}[b]{0.32\textwidth}
        \includegraphics[height=2.5cm, trim={1.4cm 0.5cm 0cm 1.8cm}, clip]{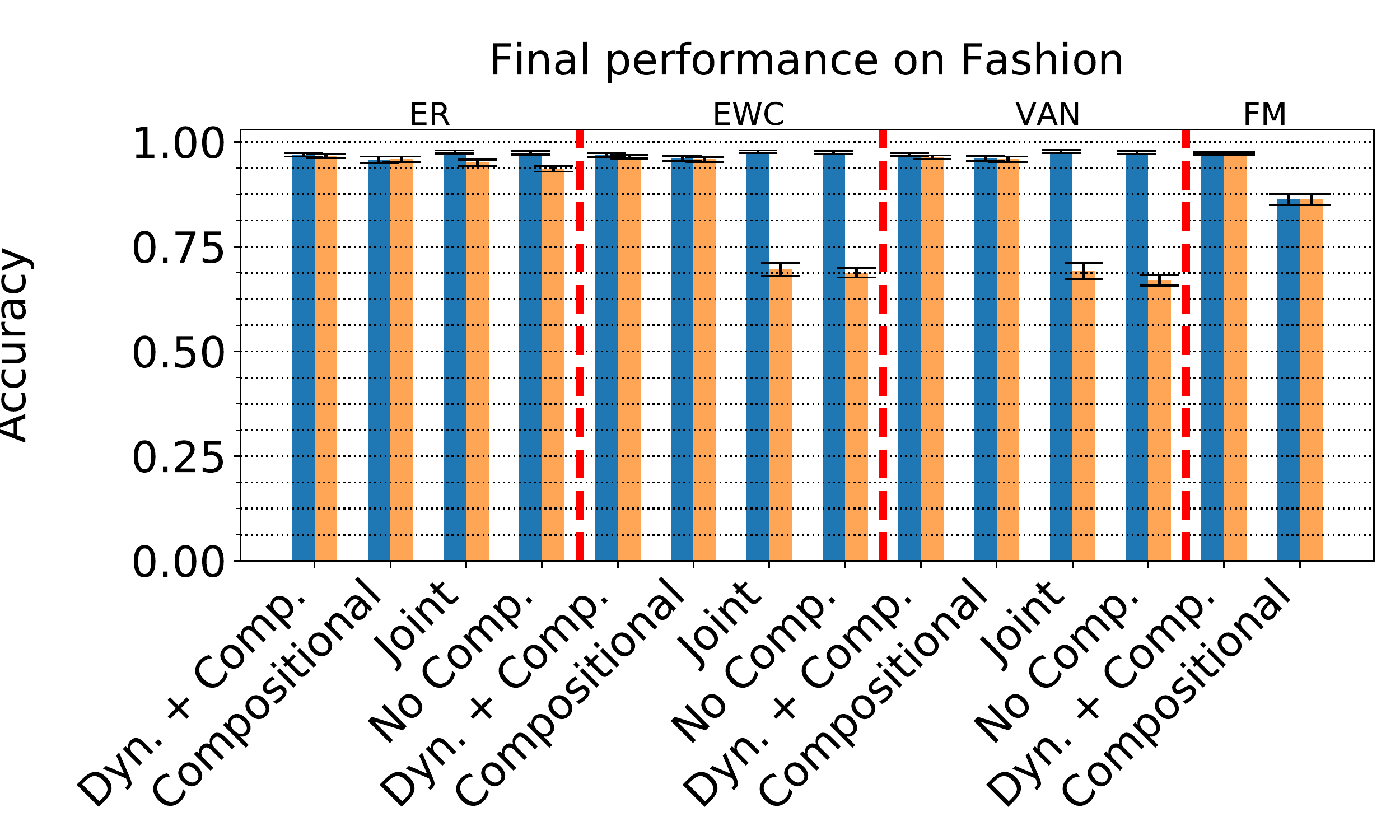}
         \caption{\Fashion}
    \end{subfigure}%
    \begin{subfigure}[b]{0.32\textwidth}
        \includegraphics[height=2.5cm, trim={1.4cm 0.5cm 0cm 1.8cm}, clip]{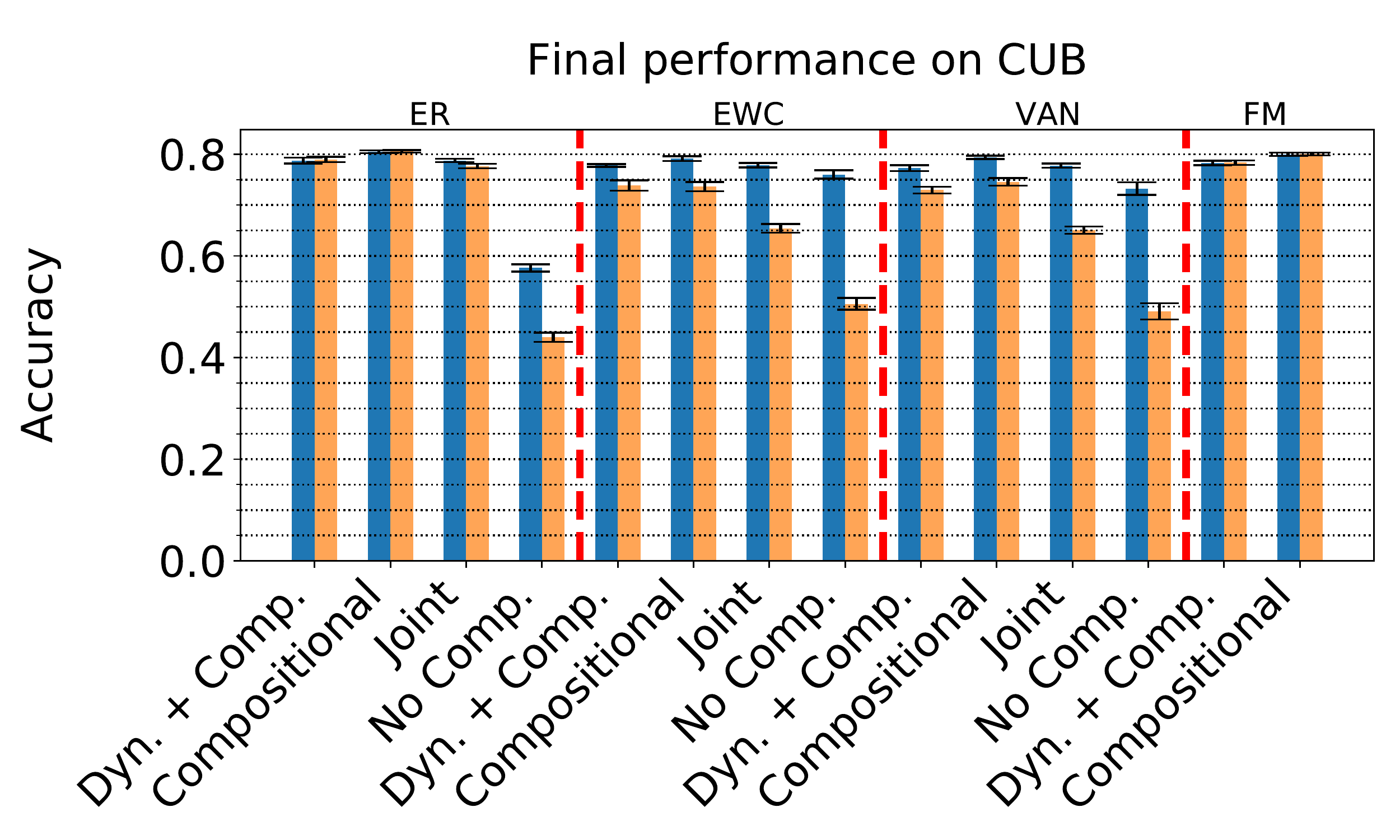}
        \caption{\CUB}
    \end{subfigure}\\
    \vspace{1em}
    \begin{subfigure}[b]{0.35\textwidth}
        \includegraphics[height=2.5cm, trim={0cm 0.5cm 0cm 1.8cm}, clip]{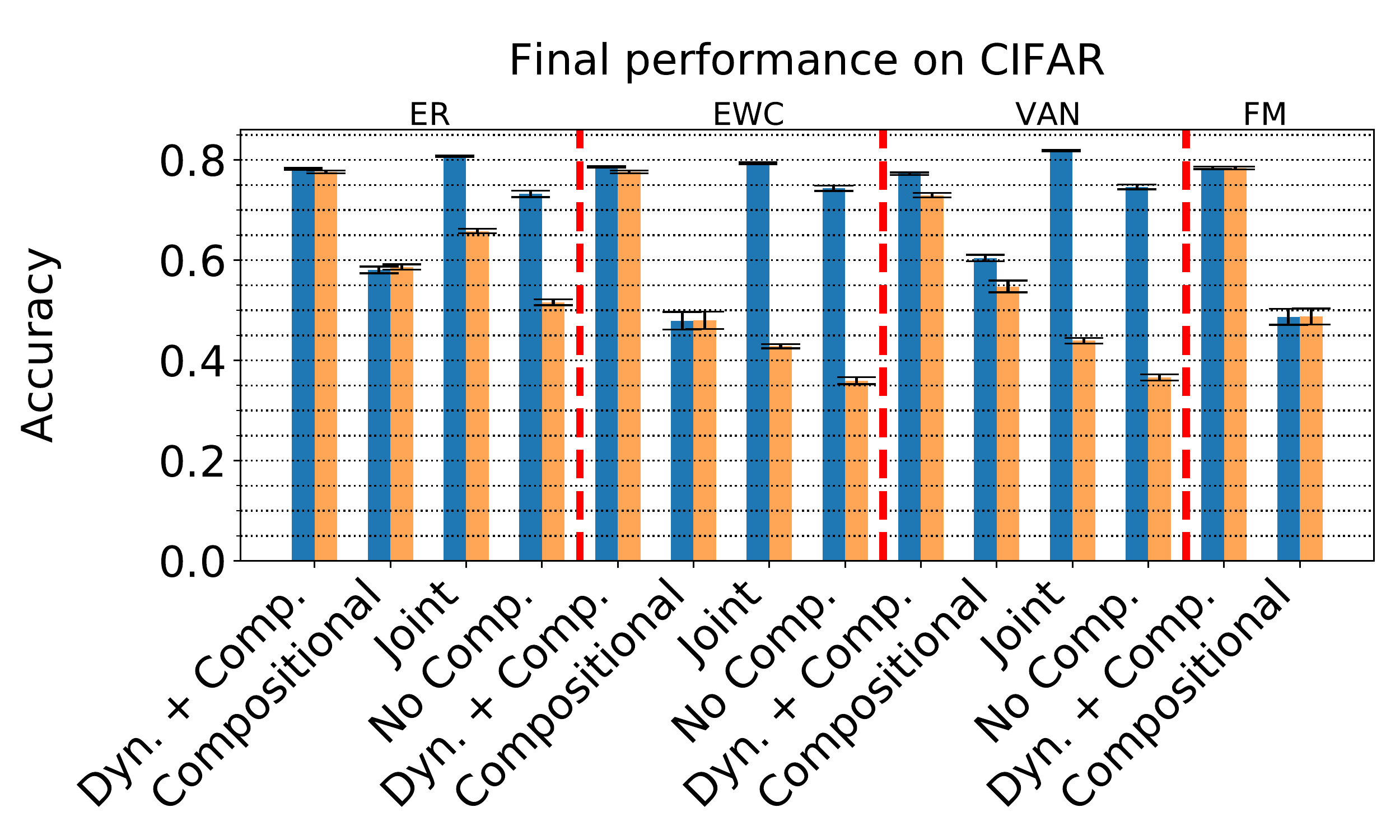}
        \caption{\CIFAR}
    \end{subfigure}%
    \begin{subfigure}[b]{0.32\textwidth}
        \includegraphics[height=2.5cm, trim={1.4cm 0.5cm 0cm 1.8cm}, clip]{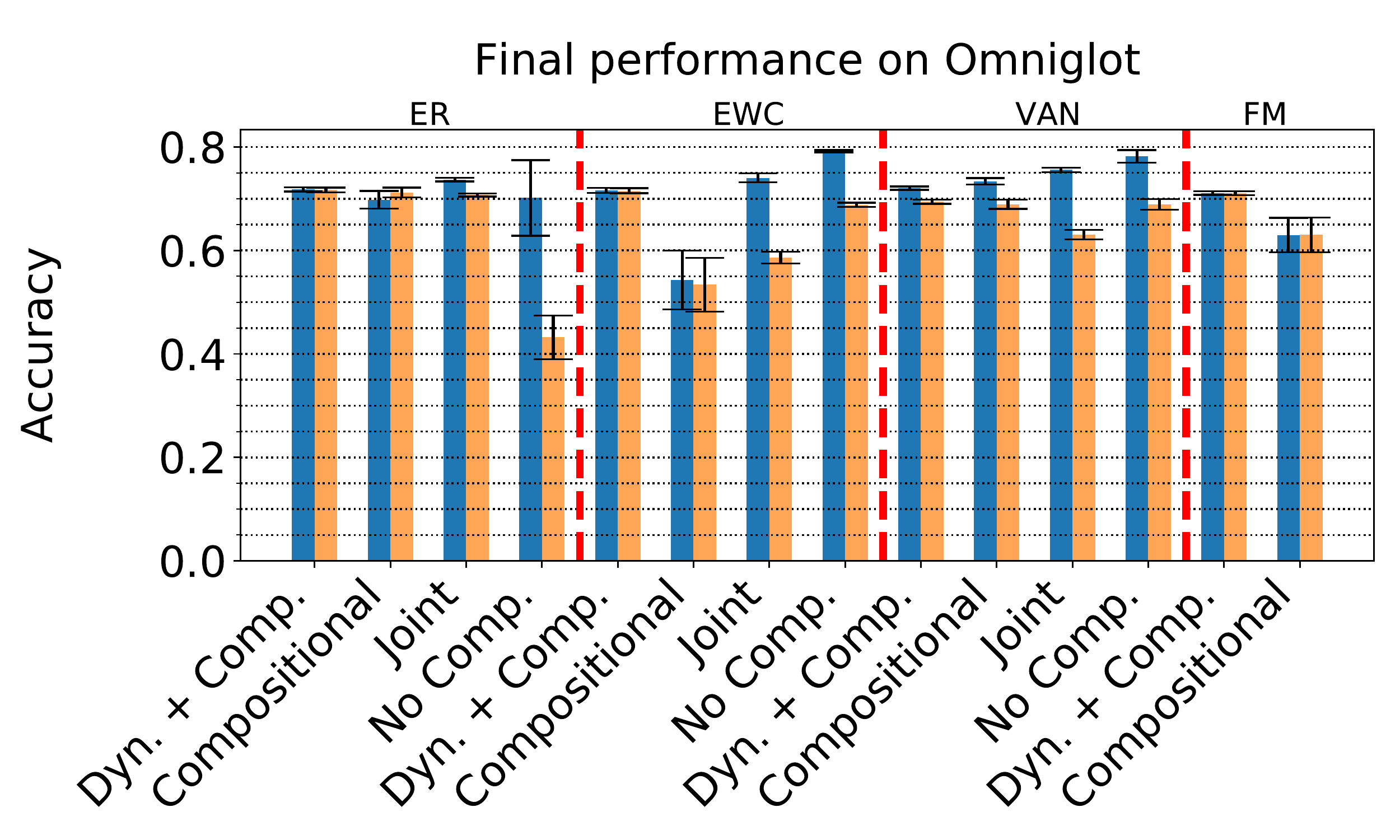}
        \caption{\Omniglot}
    \end{subfigure}%
    \begin{subfigure}[b]{0.32\textwidth}
        \centering
        \raisebox{4.2em}{\includegraphics[width=0.5\linewidth, trim={0.1cm, 0.1cm, 0.1cm, 0.1cm}, clip]{chapter3/Figures/barcharts/bar_legend.pdf}}
    \end{subfigure}%
    \caption[Full performance of supervised lifelong composition after each task and after all tasks: soft ordering.]{Soft layer ordering accuracy. Compositional agents outperformed baselines in most data sets for every adaptation method. Dyn.~+ comp. agents further improved performance, leading to these methods being strongest. Error bars denote standard errors across ten seeds.}%
    \label{fig:softOrderingBarsNotAveraged}
\end{figure}

\begin{figure}[t!]
\centering
    \begin{subfigure}[b]{0.08\textwidth}
    \caption*{}
    \end{subfigure}%
    \begin{subfigure}[b]{0.223\textwidth}
        \caption*{\ \ ER Dyn.~+ Comp.}
    \end{subfigure}%
    \begin{subfigure}[b]{0.213\textwidth}
        \caption*{\ ER Compositional}
    \end{subfigure}%
    \begin{subfigure}[b]{0.213\textwidth}
        \caption*{\ \ ER Joint}
    \end{subfigure}%
    \begin{subfigure}[b]{0.213\textwidth}
        \caption*{ER No Components}
    \end{subfigure}\\
    \vspace{0.5em}
    \begin{subfigure}[b]{0.1\textwidth}
        \raisebox{2.3em}{\small\MNIST{}}
    \end{subfigure}%
    \begin{subfigure}[b]{0.228\textwidth}
        \includegraphics[height=1.85cm, trim={0.4cm 1.8cm 3.cm 1.9cm}, clip]{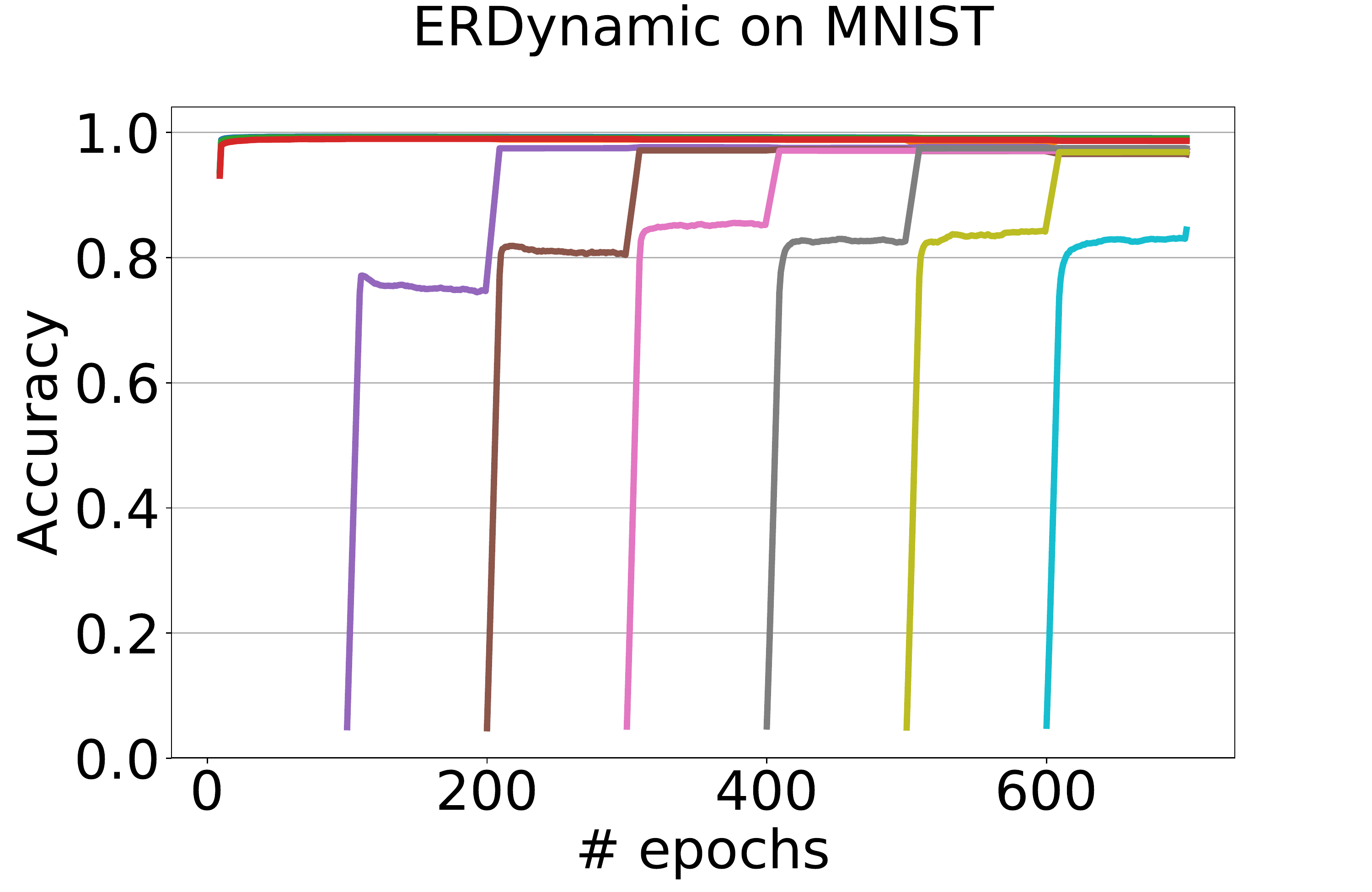}
    \end{subfigure}%
    \begin{subfigure}[b]{0.218\textwidth}
        \includegraphics[height=1.85cm, trim={1.7cm 1.8cm 3.cm 1.9cm}, clip]{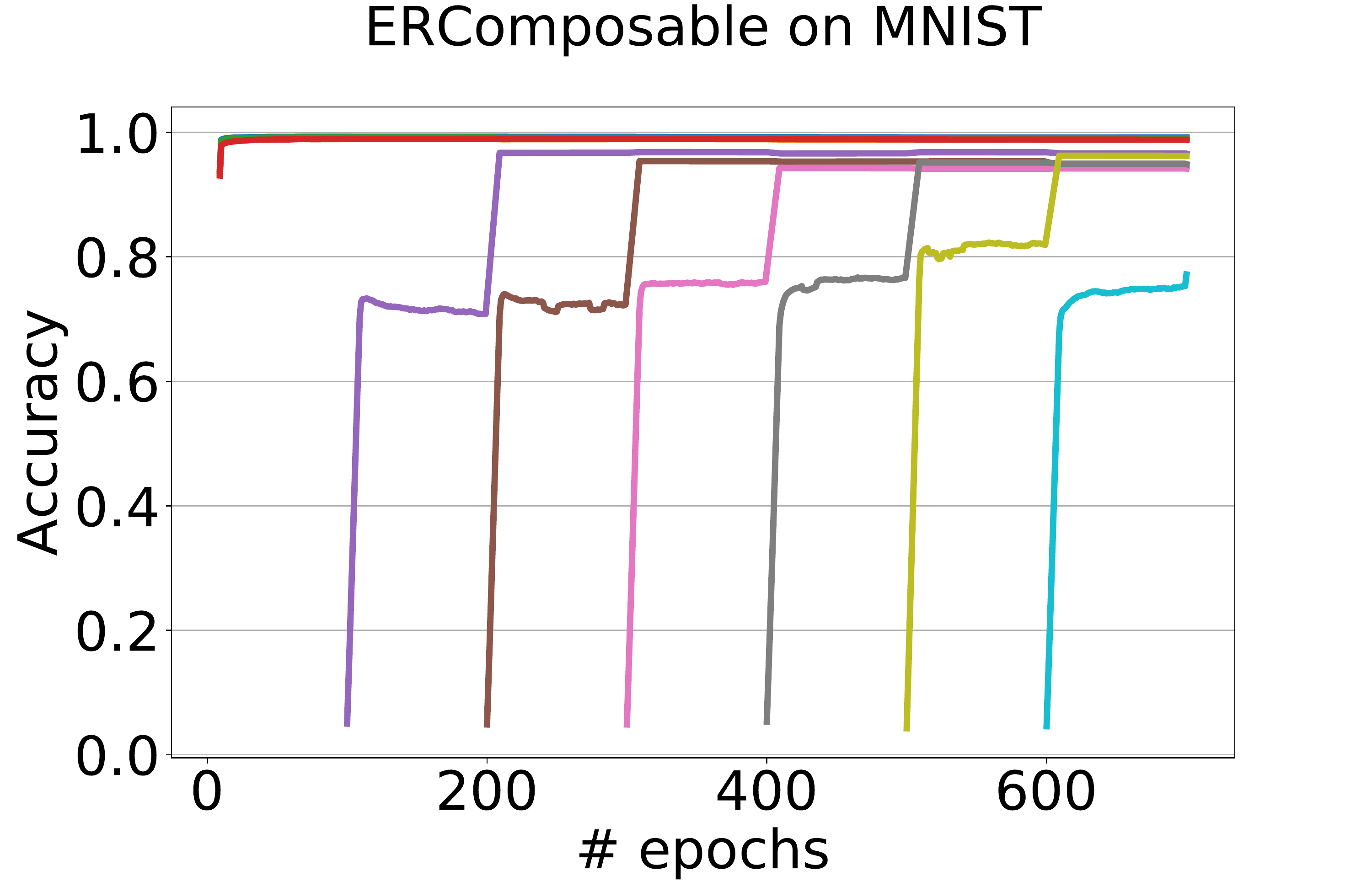}
    \end{subfigure}%
    \begin{subfigure}[b]{0.218\textwidth}
        \includegraphics[height=1.85cm, trim={1.7cm 1.8cm 3.cm 1.9cm}, clip]{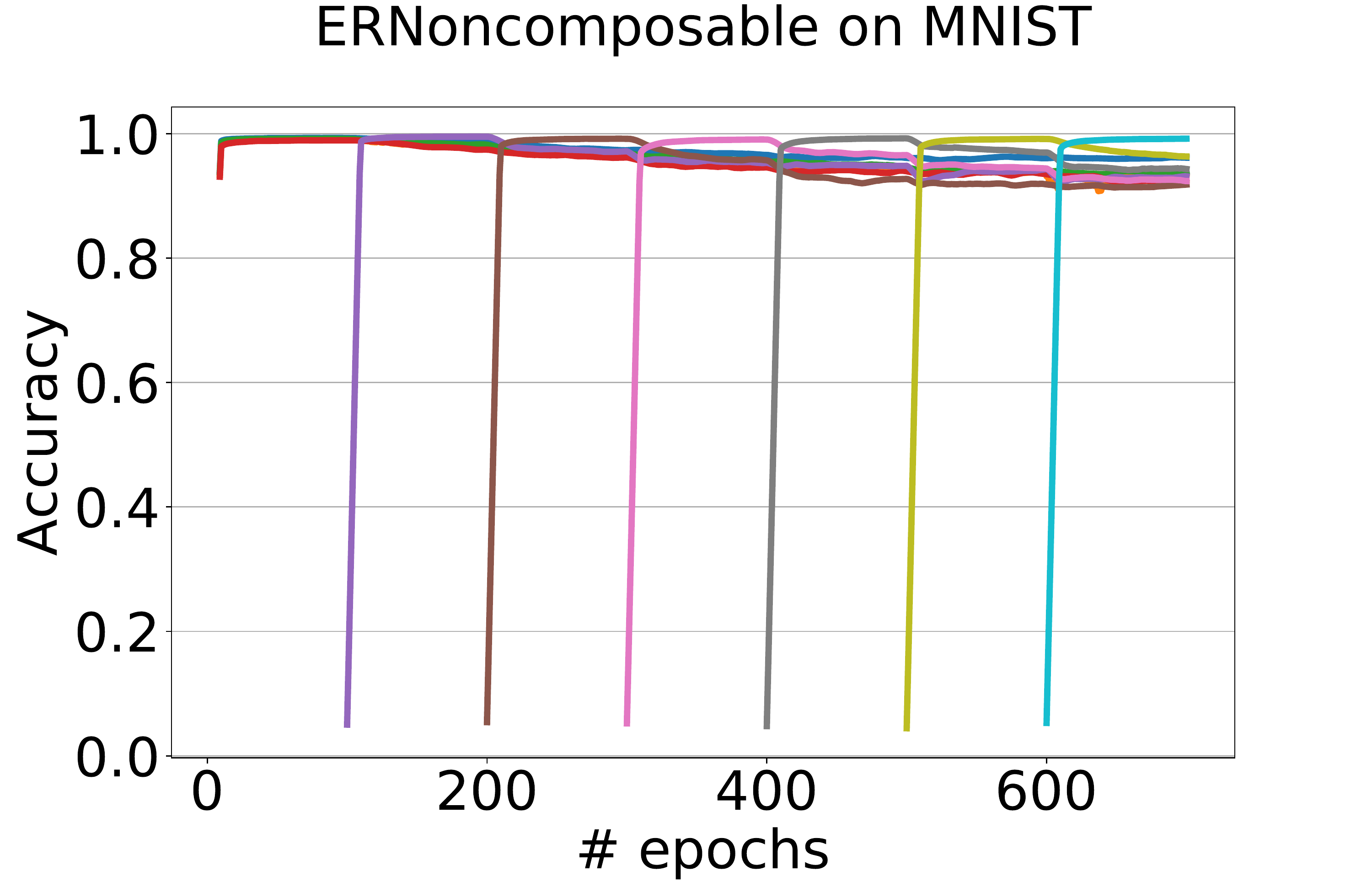}
    \end{subfigure}%
    \begin{subfigure}[b]{0.218\textwidth}
        \includegraphics[height=1.85cm, trim={1.7cm 1.8cm 3.cm 1.9cm}, clip]{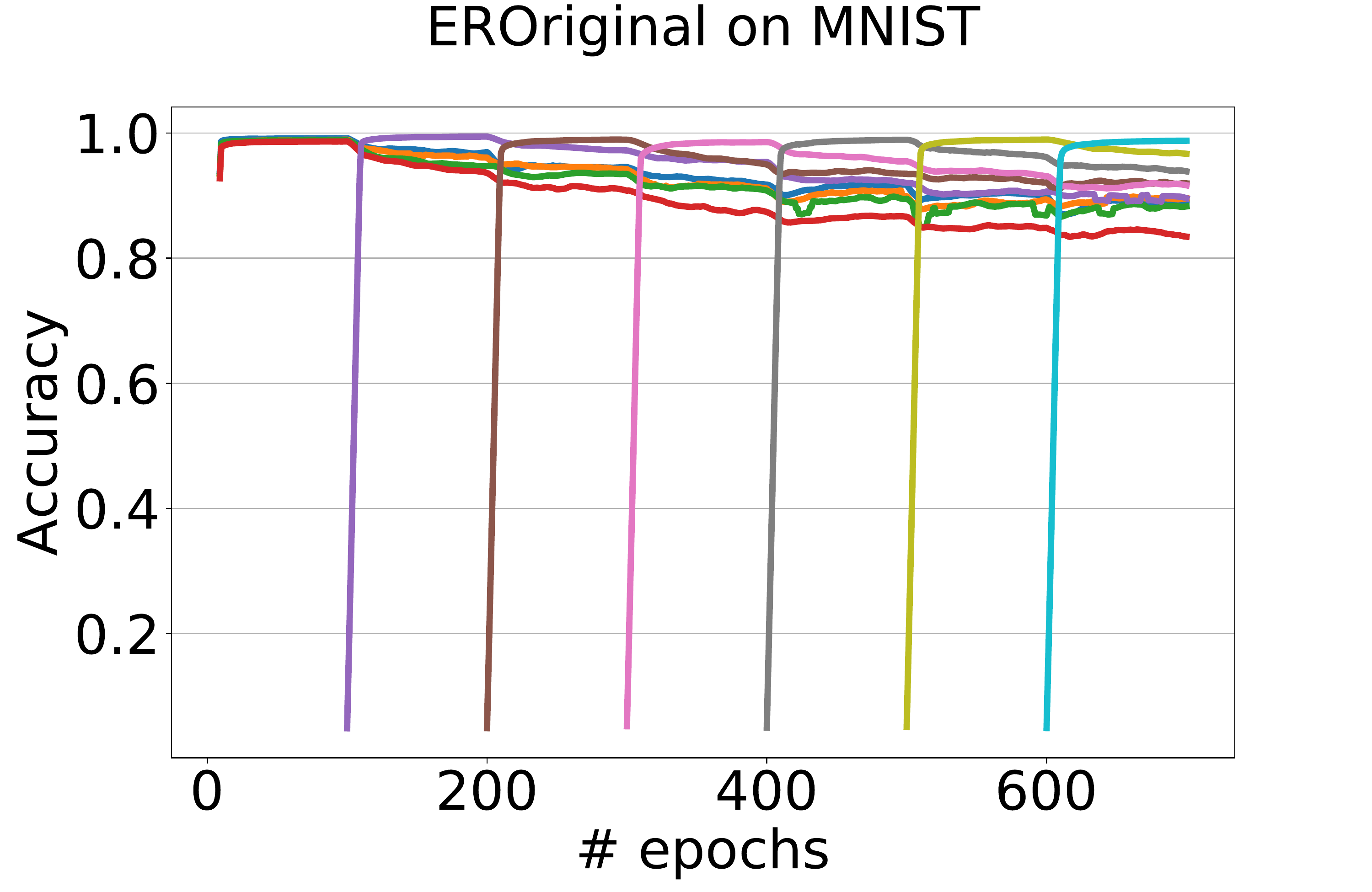}
    \end{subfigure}\\
    \vspace{0.5em}
    \begin{subfigure}[b]{0.1\textwidth}
        \raisebox{2.3em}{\small\Fashion{}}
    \end{subfigure}%
    \begin{subfigure}[b]{0.228\textwidth}
        \includegraphics[height=1.85cm, trim={0.4cm 1.8cm 3.cm 1.9cm}, clip]{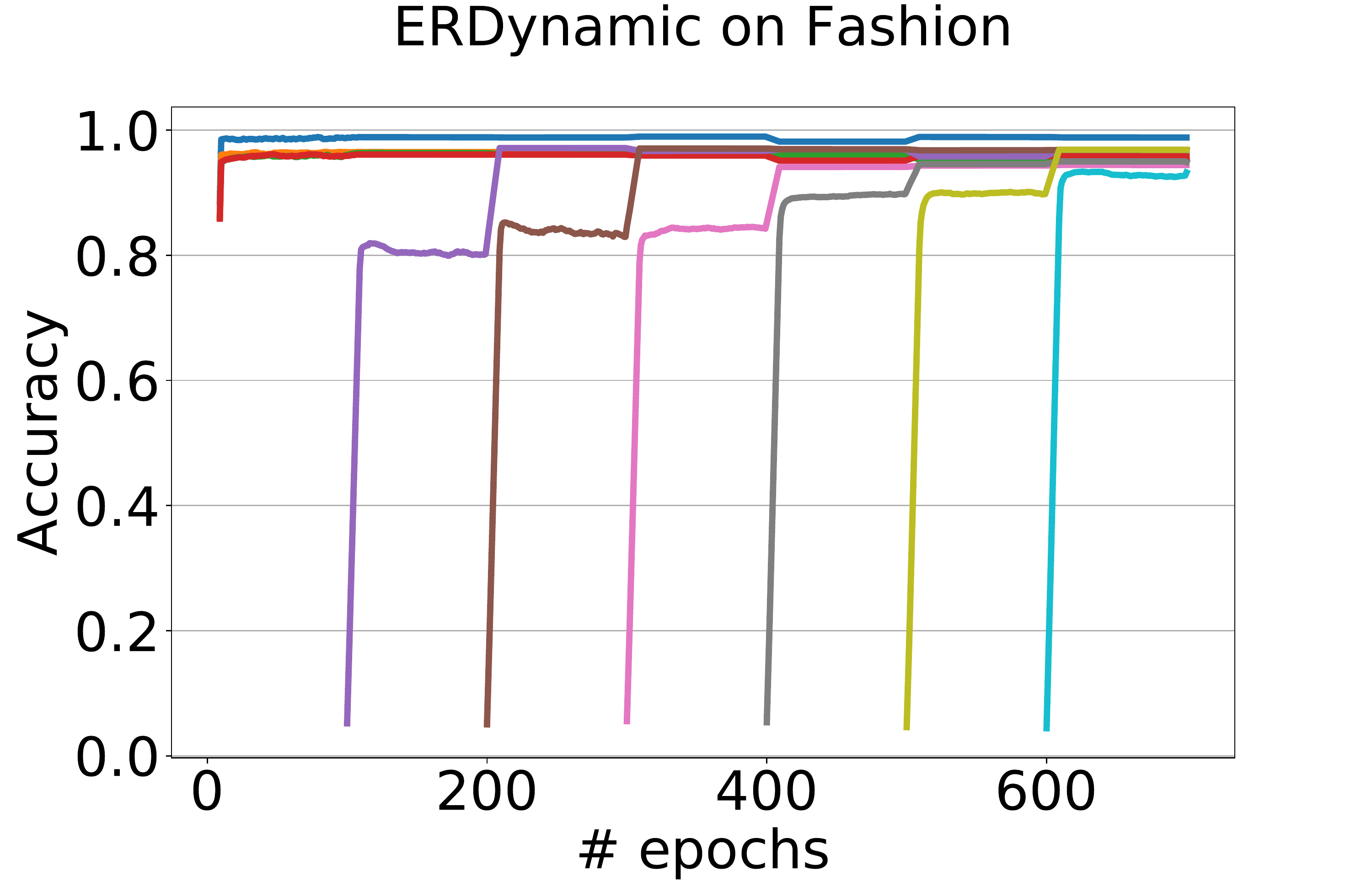}
    \end{subfigure}%
    \begin{subfigure}[b]{0.218\textwidth}
        \includegraphics[height=1.85cm, trim={1.7cm 1.8cm 3.cm 1.9cm}, clip]{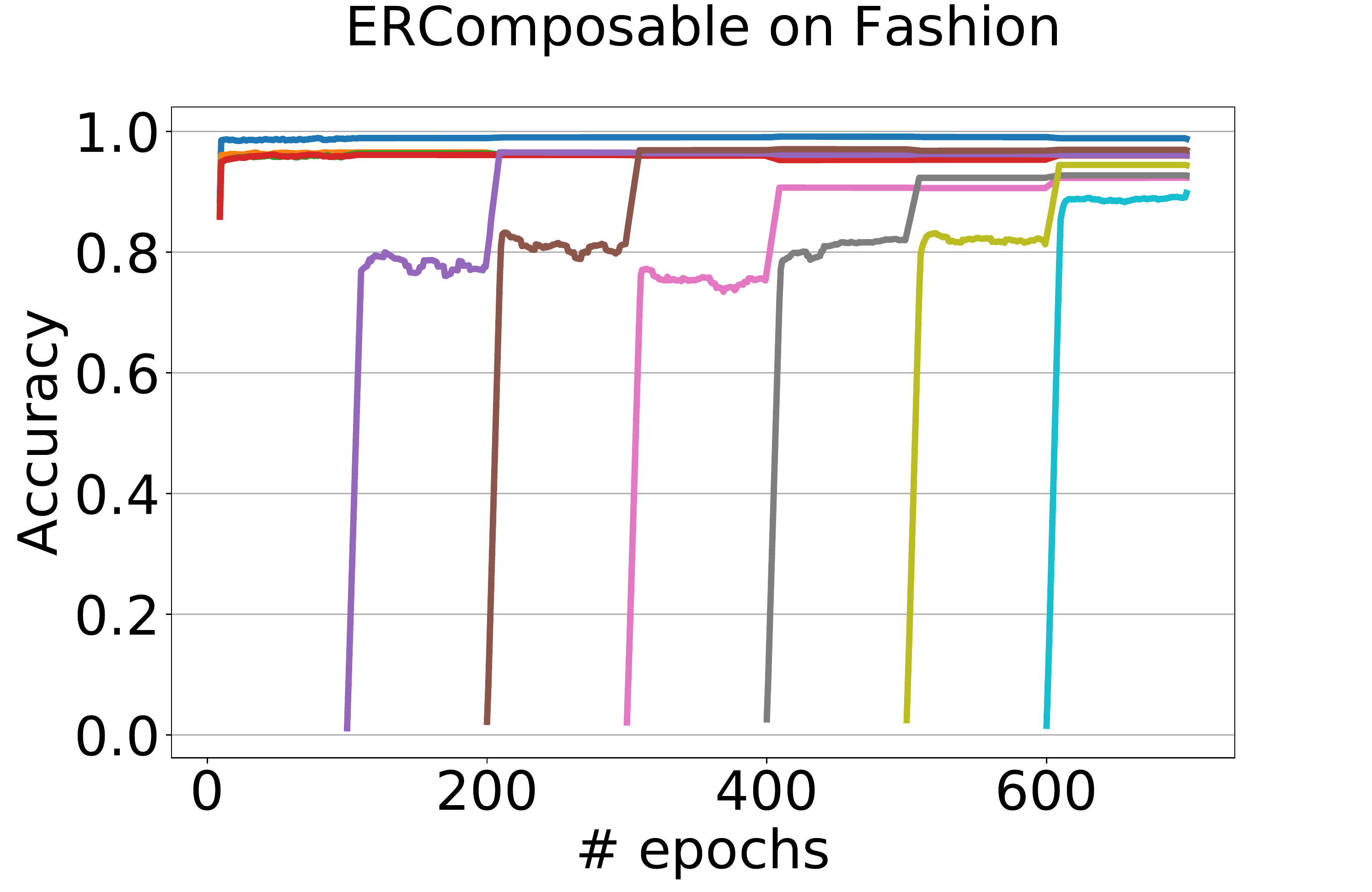}
    \end{subfigure}%
    \begin{subfigure}[b]{0.218\textwidth}
        \includegraphics[height=1.85cm, trim={1.7cm 1.8cm 3.cm 1.9cm}, clip]{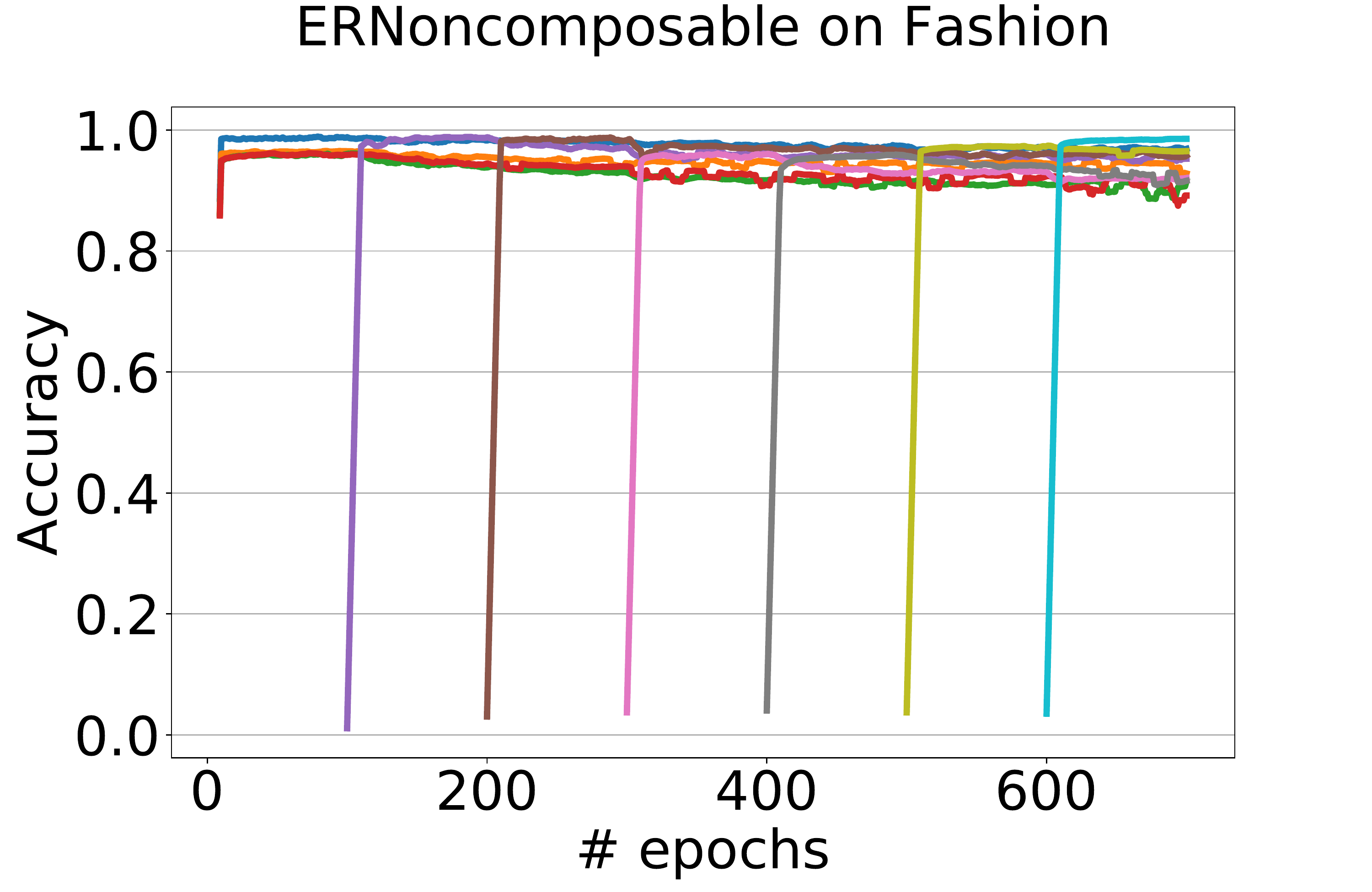}
    \end{subfigure}%
    \begin{subfigure}[b]{0.218\textwidth}
        \includegraphics[height=1.85cm, trim={1.7cm 1.8cm 3.cm 1.9cm}, clip]{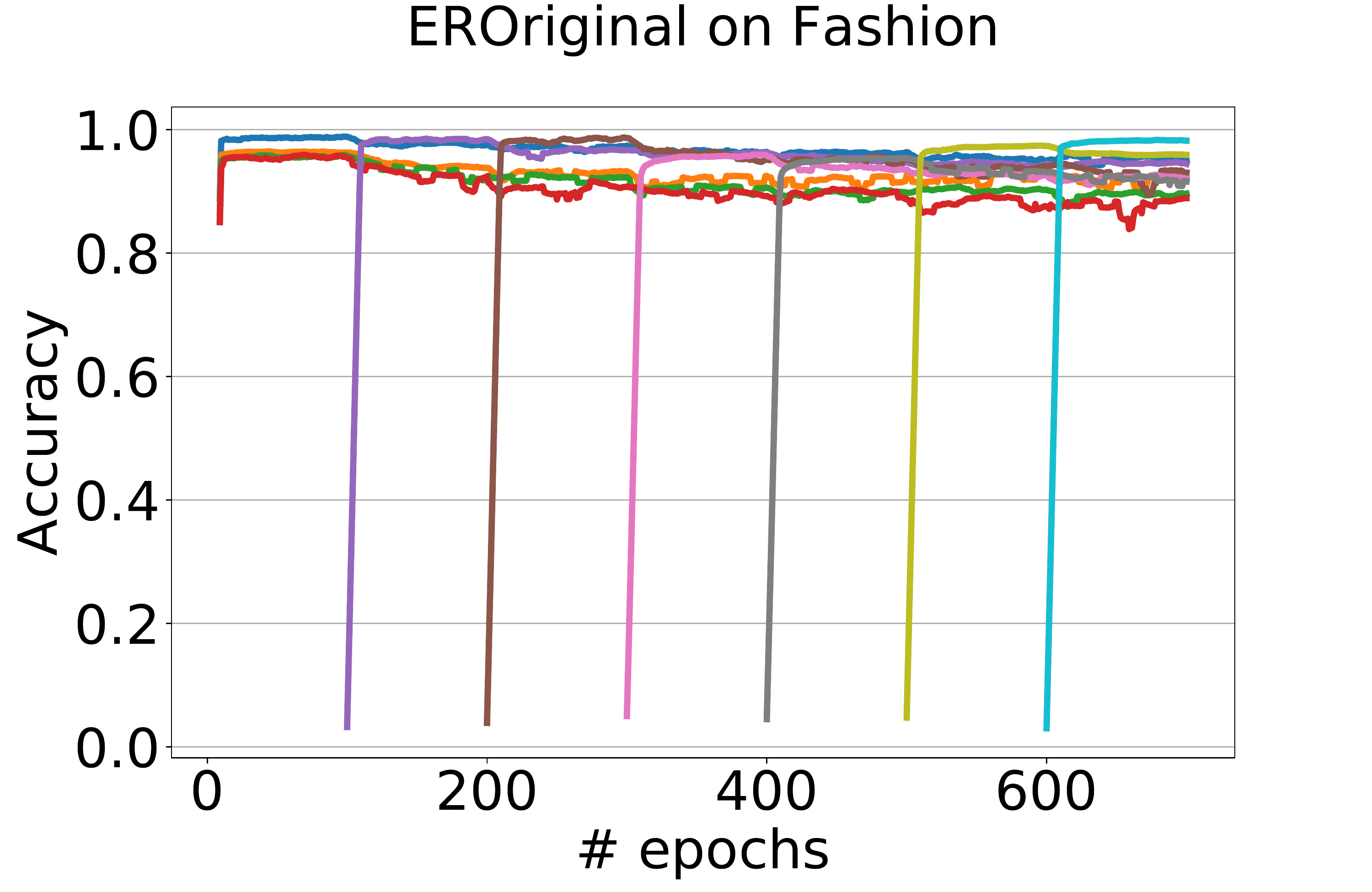}
    \end{subfigure}\\
    \vspace{0.5em}
    \begin{subfigure}[b]{0.1\textwidth}
        \raisebox{2.3em}{\small\CUB{}}
    \end{subfigure}%
    \begin{subfigure}[b]{0.228\textwidth}
        \includegraphics[height=1.85cm, trim={0.4cm 1.8cm 3.cm 1.9cm}, clip]{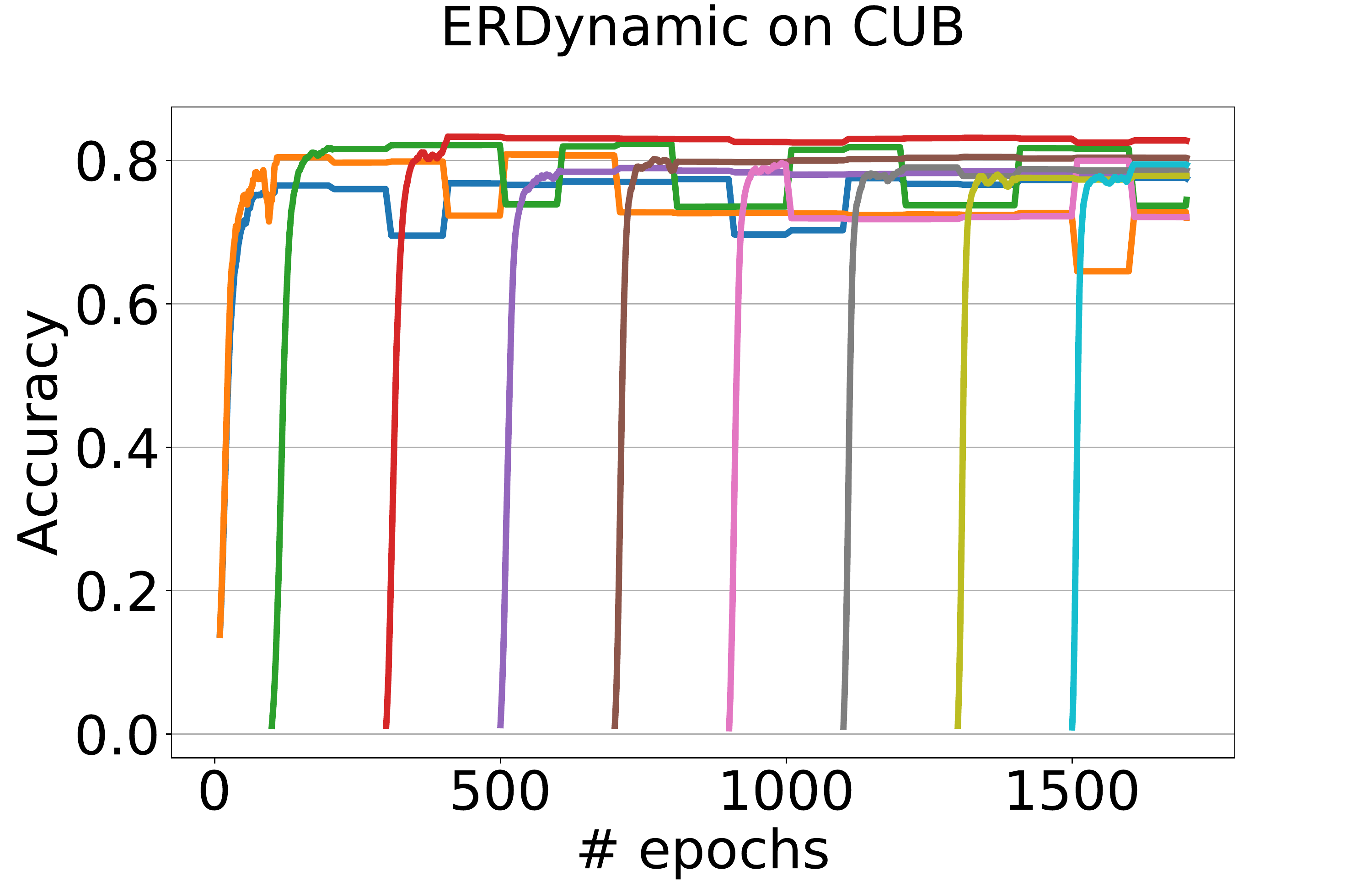}
    \end{subfigure}%
    \begin{subfigure}[b]{0.218\textwidth}
        \includegraphics[height=1.85cm, trim={1.7cm 1.8cm 3.cm 1.9cm}, clip]{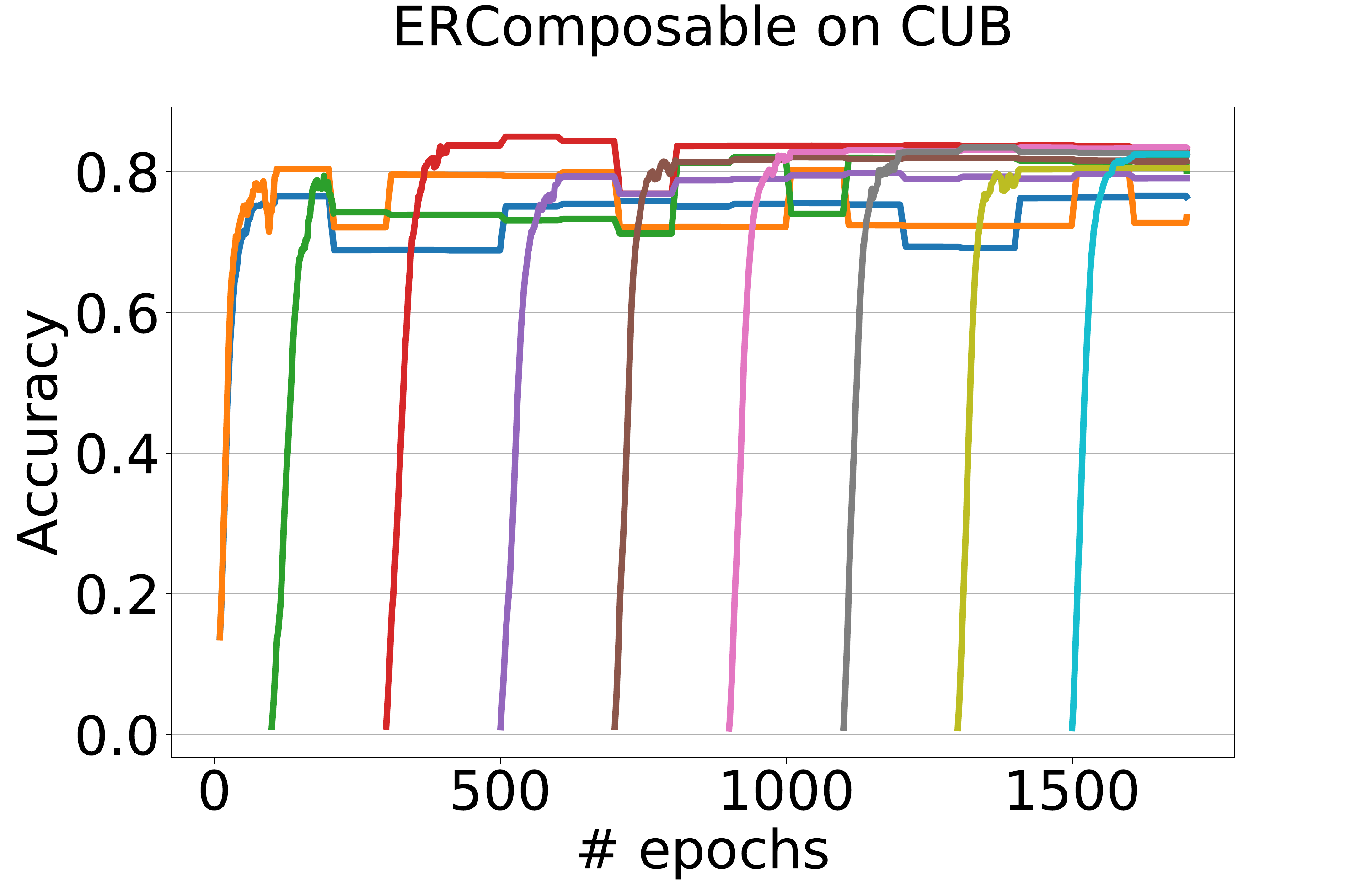}
    \end{subfigure}%
    \begin{subfigure}[b]{0.218\textwidth}
        \includegraphics[height=1.85cm, trim={1.7cm 1.8cm 3.cm 1.9cm}, clip]{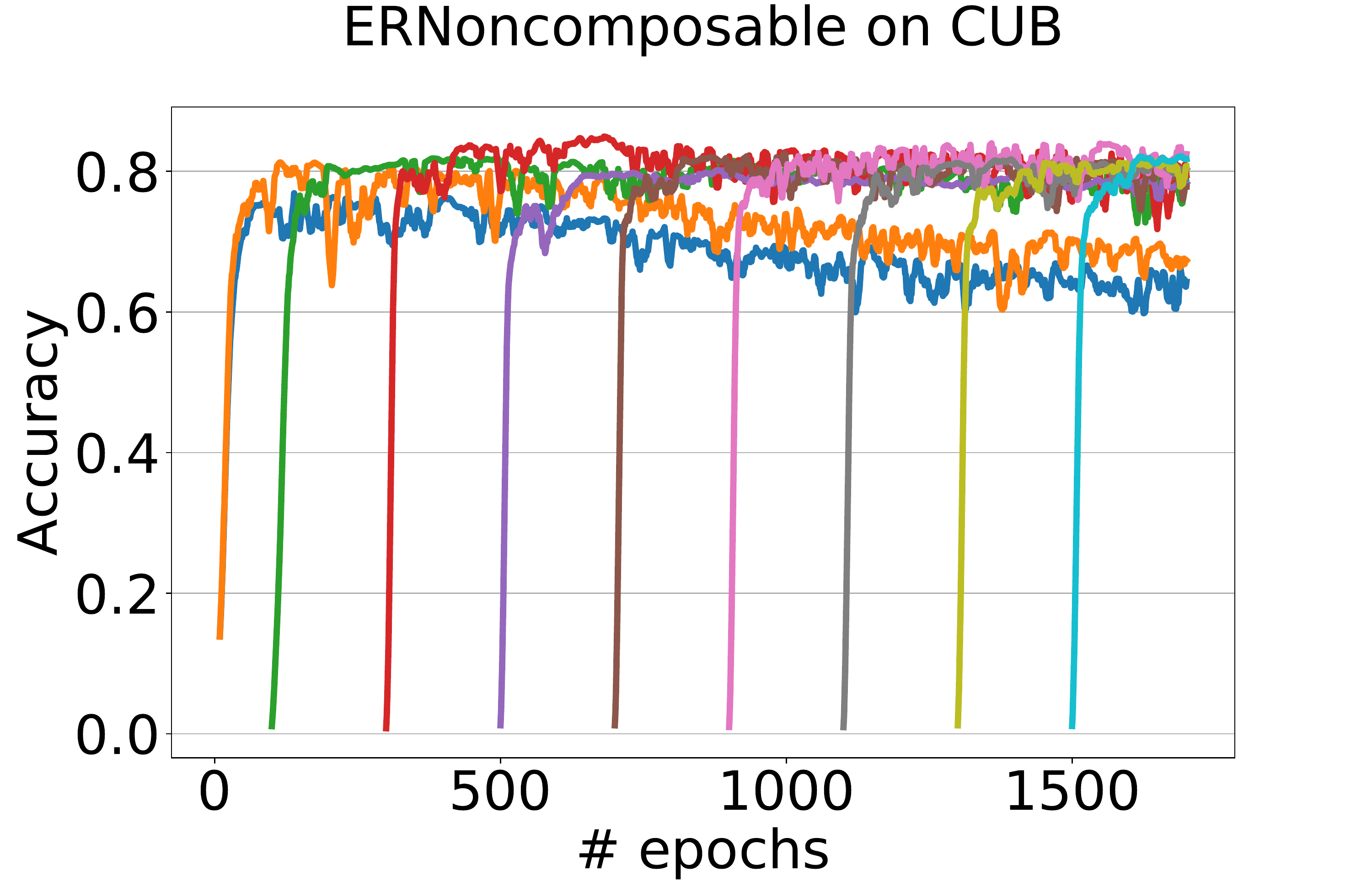}
    \end{subfigure}%
    \begin{subfigure}[b]{0.218\textwidth}
        \includegraphics[height=1.85cm, trim={1.7cm 1.8cm 3.cm 1.9cm}, clip]{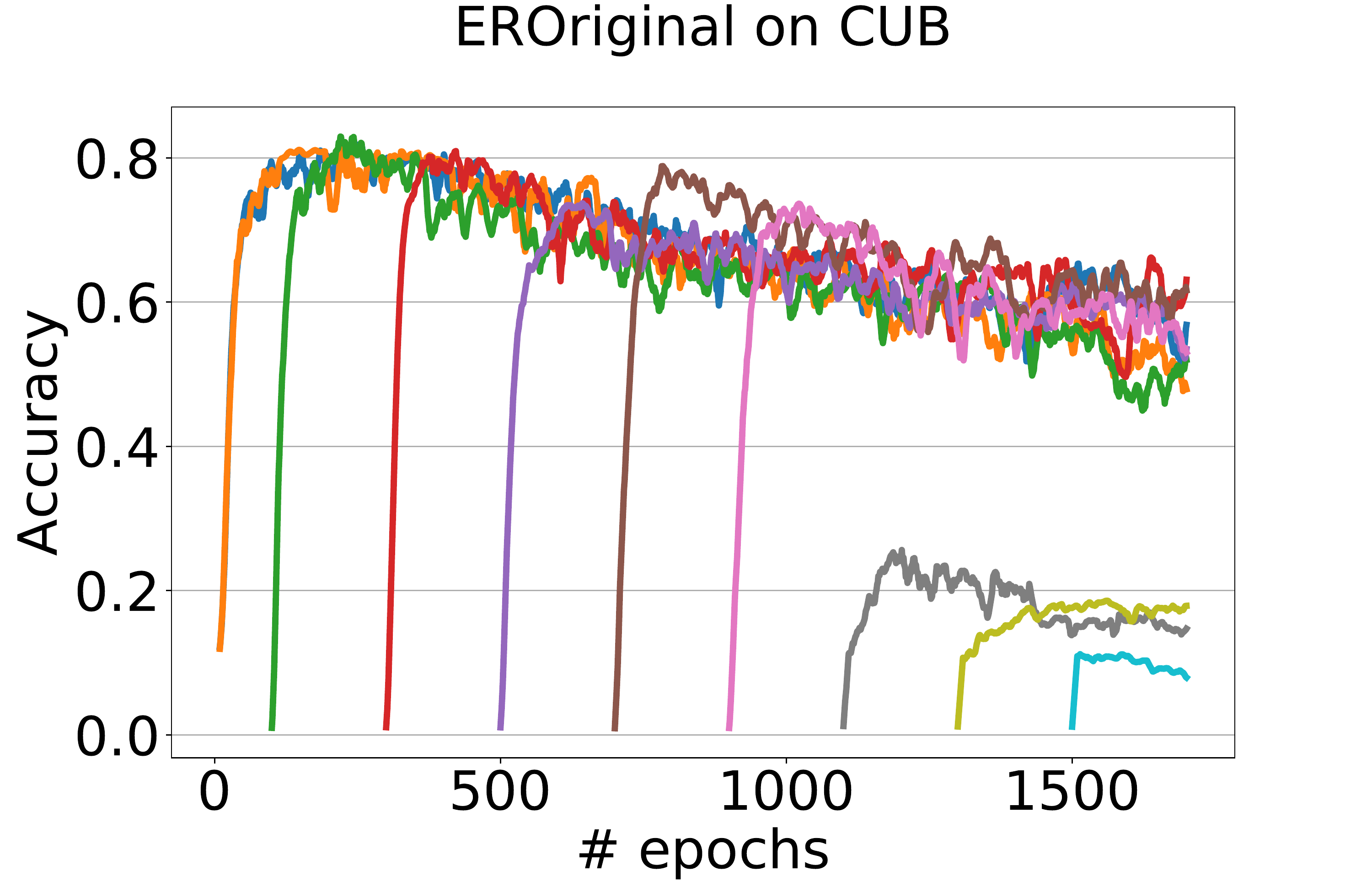}
    \end{subfigure}\\
    \vspace{0.5em}
    \begin{subfigure}[b]{0.1\textwidth}
        \raisebox{2.3em}{\small\CIFAR{}}
    \end{subfigure}%
    \begin{subfigure}[b]{0.228\textwidth}
        \includegraphics[height=1.85cm, trim={0.4cm 1.8cm 3.cm 1.9cm}, clip]{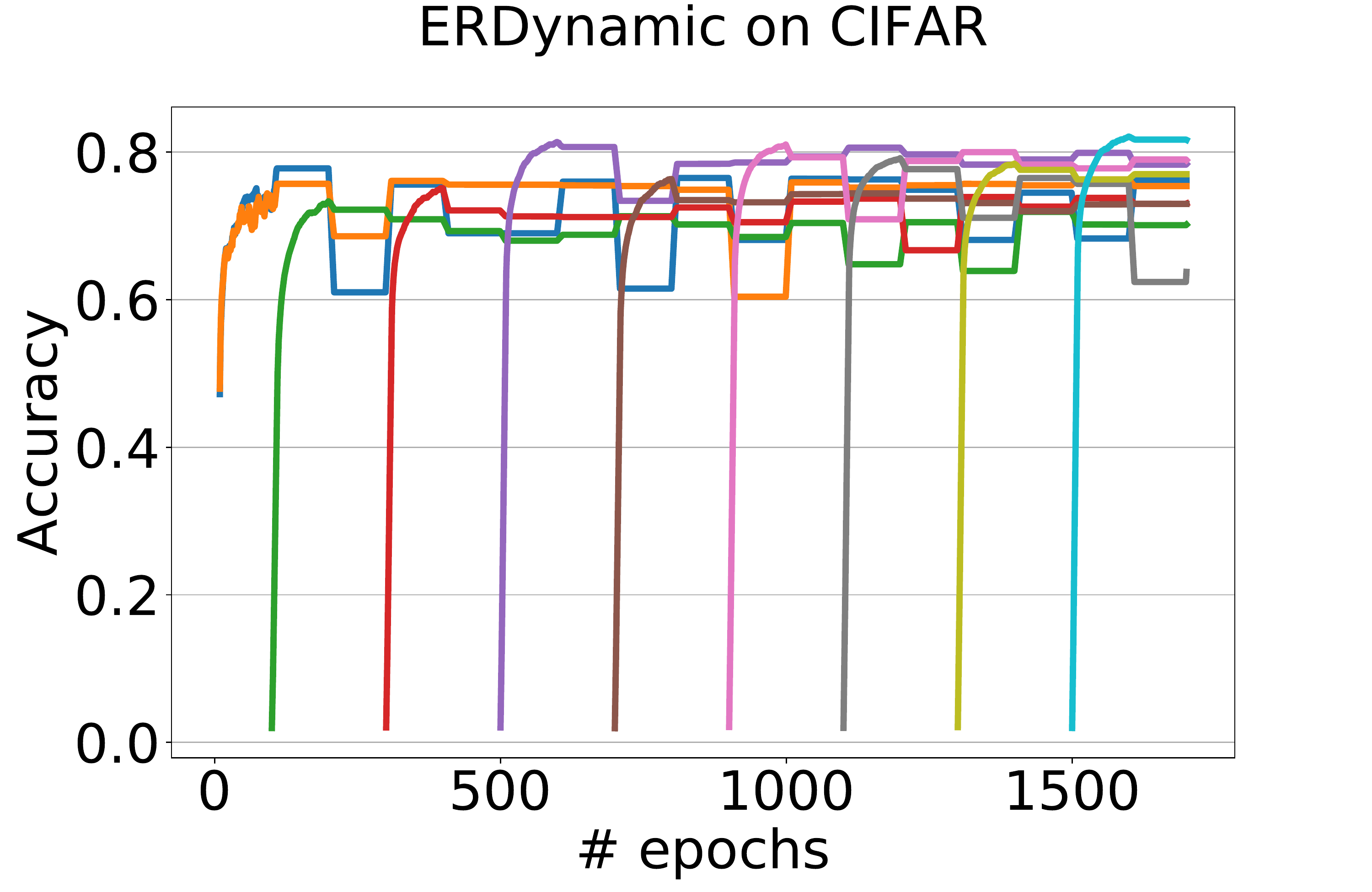}
    \end{subfigure}%
    \begin{subfigure}[b]{0.218\textwidth}
        \includegraphics[height=1.85cm, trim={1.7cm 1.8cm 3.cm 1.9cm}, clip]{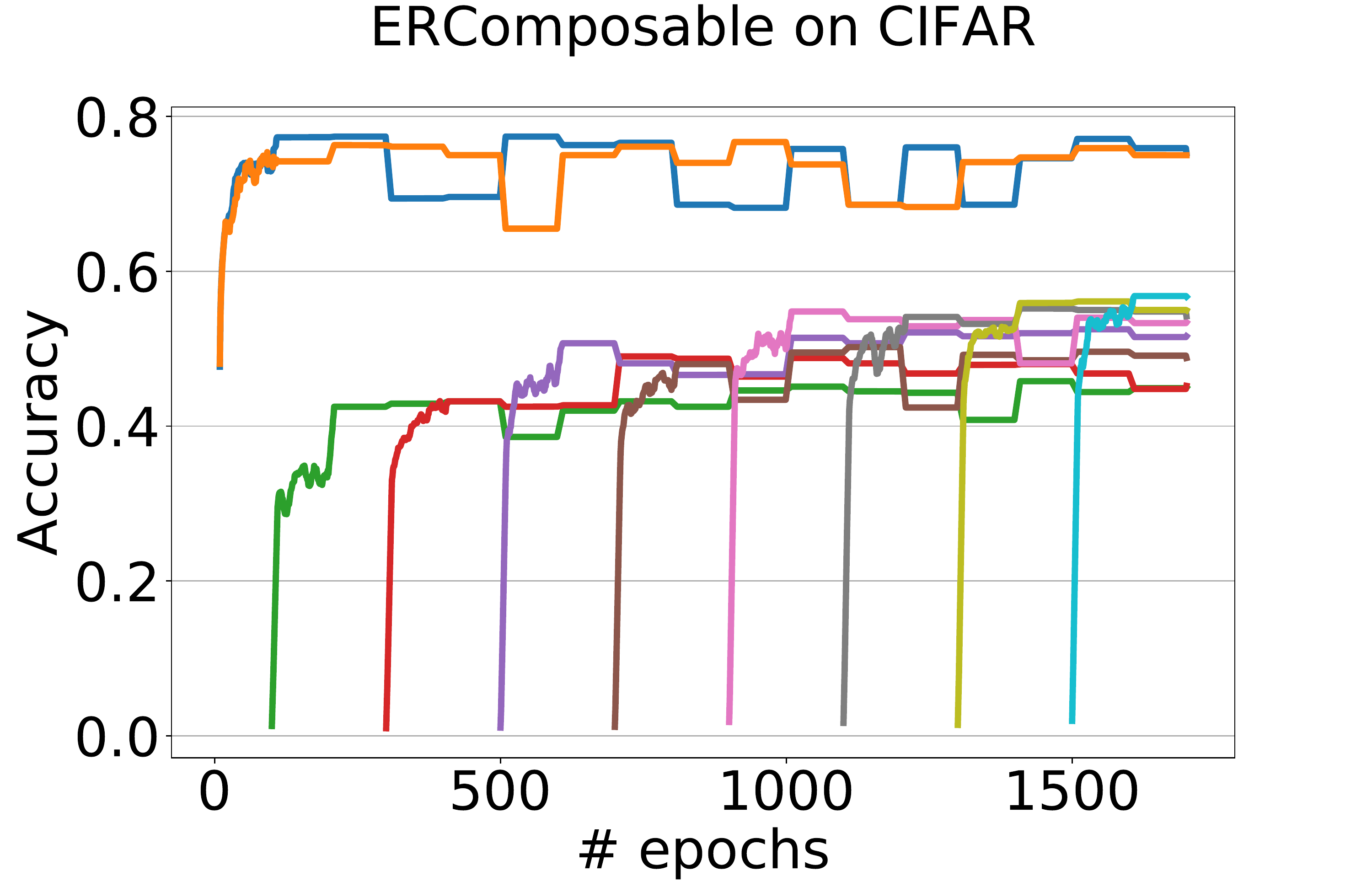}
    \end{subfigure}%
    \begin{subfigure}[b]{0.218\textwidth}
        \includegraphics[height=1.85cm, trim={1.7cm 1.8cm 3.cm 1.9cm}, clip]{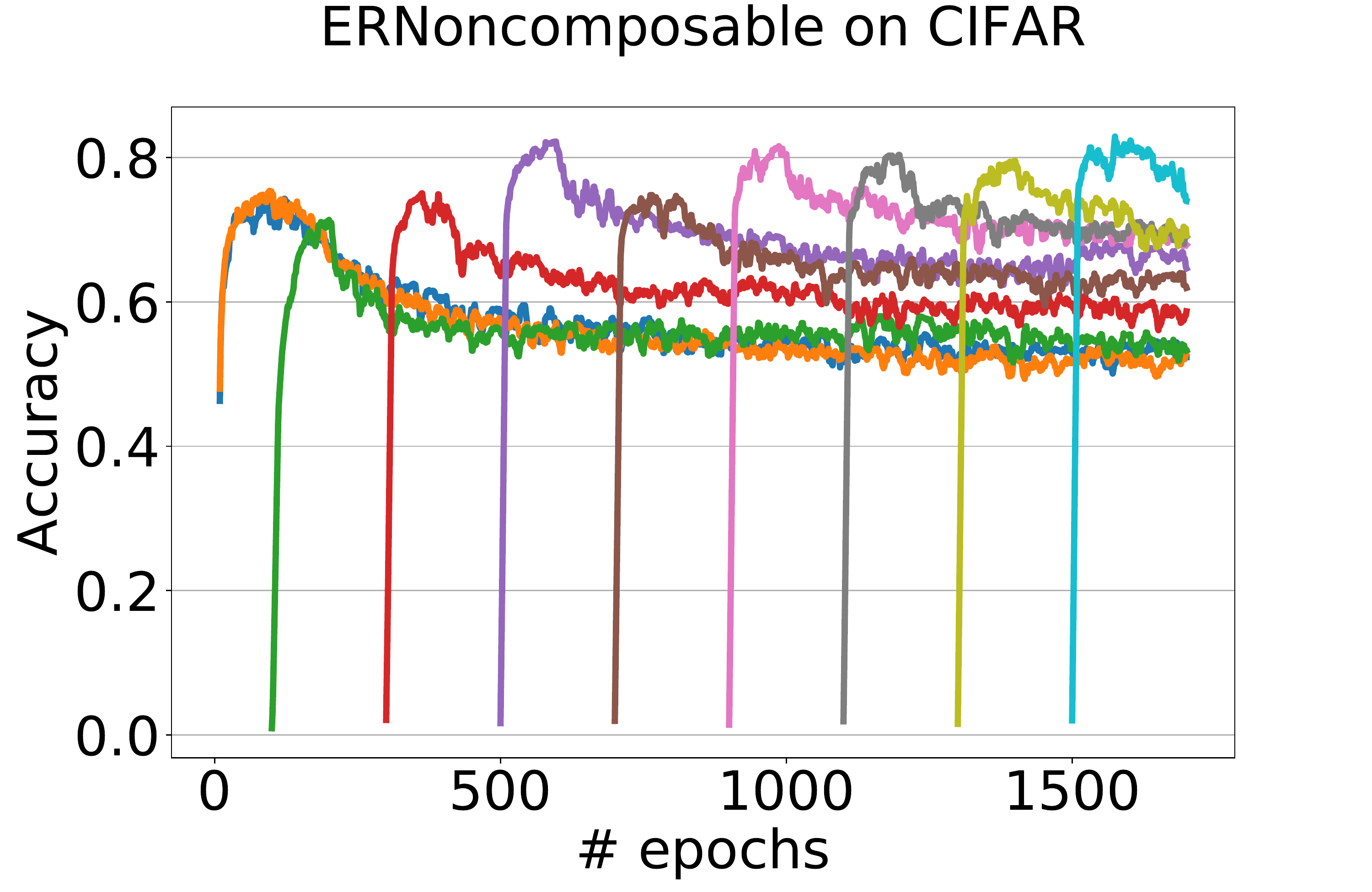}
    \end{subfigure}%
    \begin{subfigure}[b]{0.218\textwidth}
        \includegraphics[height=1.85cm, trim={1.7cm 1.8cm 3.cm 1.9cm}, clip]{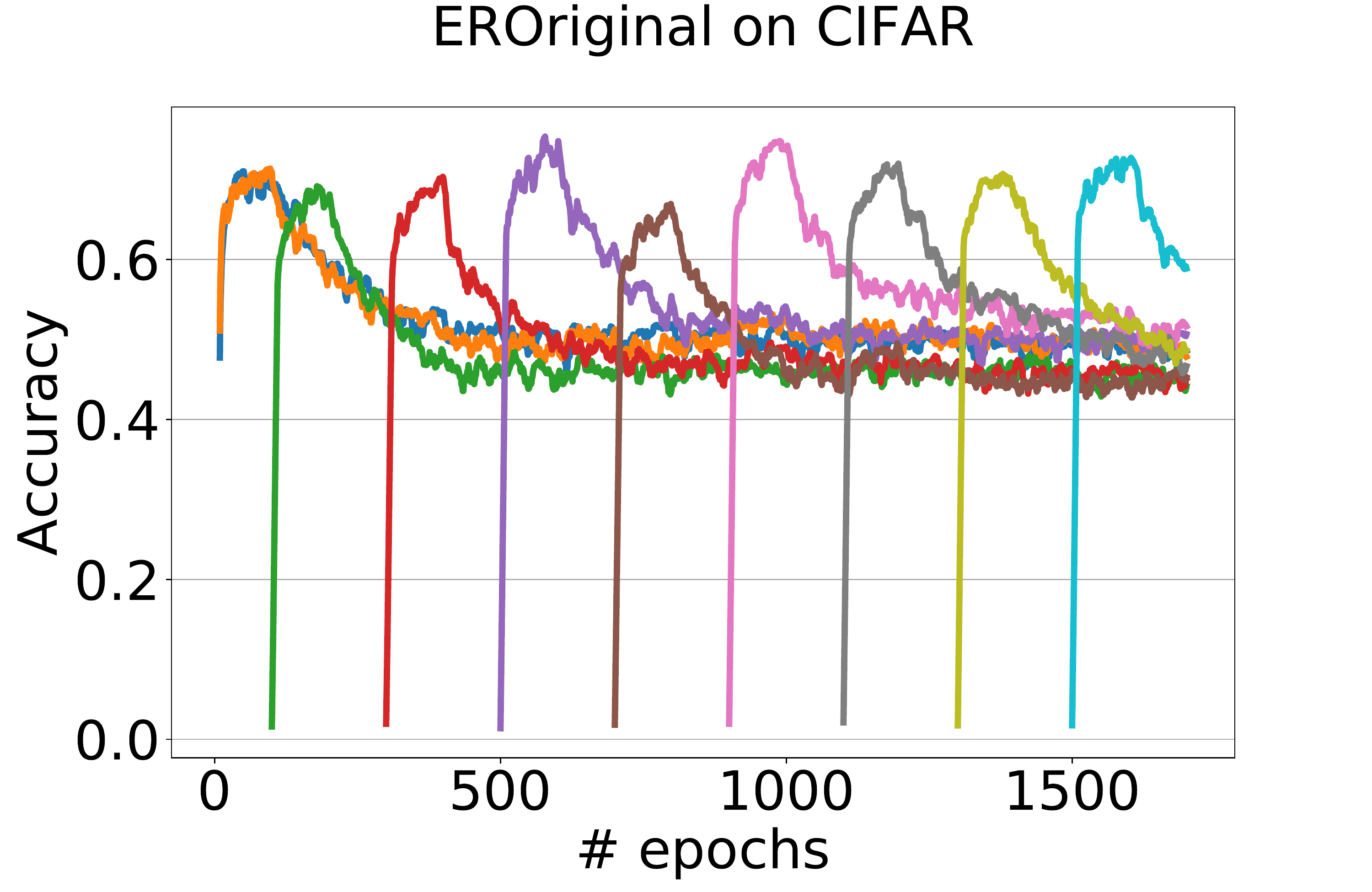}
    \end{subfigure}\\
    \vspace{0.5em}
    \begin{subfigure}[b]{0.1\textwidth}
        \raisebox{2.8em}{\small\Omniglot{}}
    \end{subfigure}%
    \begin{subfigure}[b]{0.228\textwidth}
        \includegraphics[height=2.02cm, trim={0.4cm 0.3cm 3.cm 1.9cm}, clip]{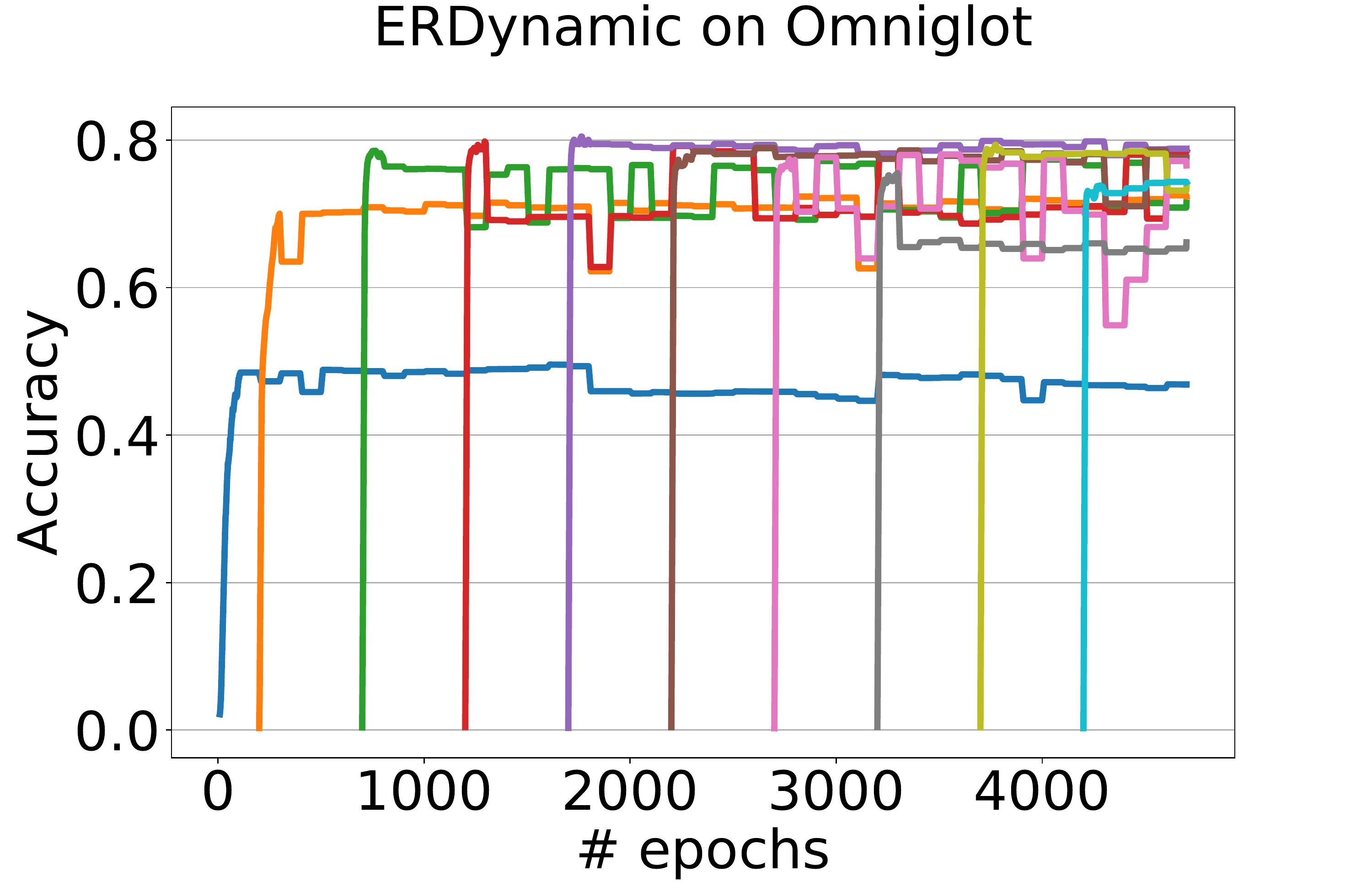}
    \end{subfigure}%
    \begin{subfigure}[b]{0.218\textwidth}
        \includegraphics[height=2.02cm, trim={1.7cm 0.3cm 3.cm 1.9cm}, clip]{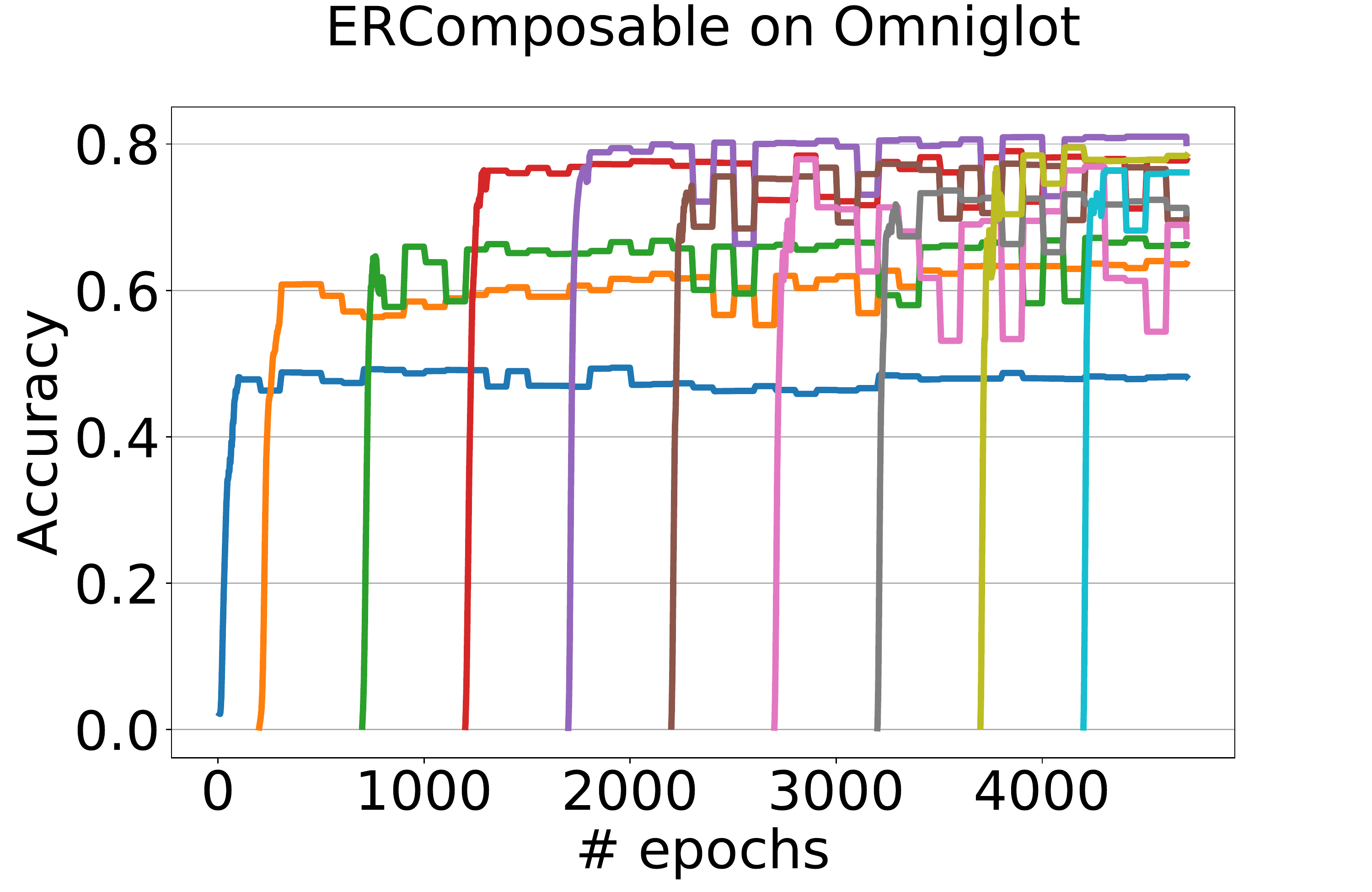}
    \end{subfigure}%
    \begin{subfigure}[b]{0.218\textwidth}
        \includegraphics[height=2.02cm, trim={1.7cm 0.3cm 3.cm 1.9cm}, clip]{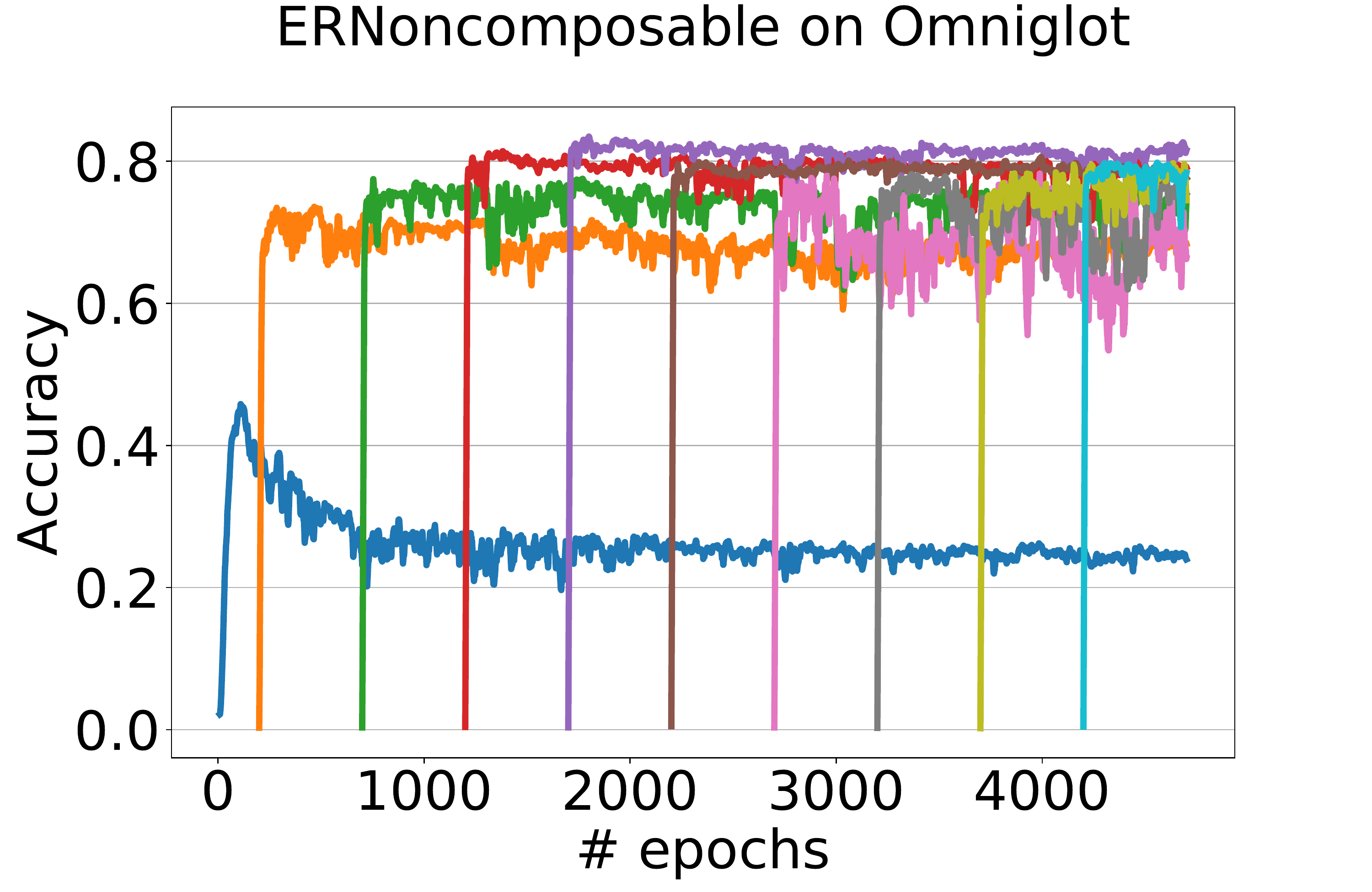}
    \end{subfigure}%
    \begin{subfigure}[b]{0.218\textwidth}
        \includegraphics[height=2.02cm, trim={1.7cm 0.3cm 3.cm 1.9cm}, clip]{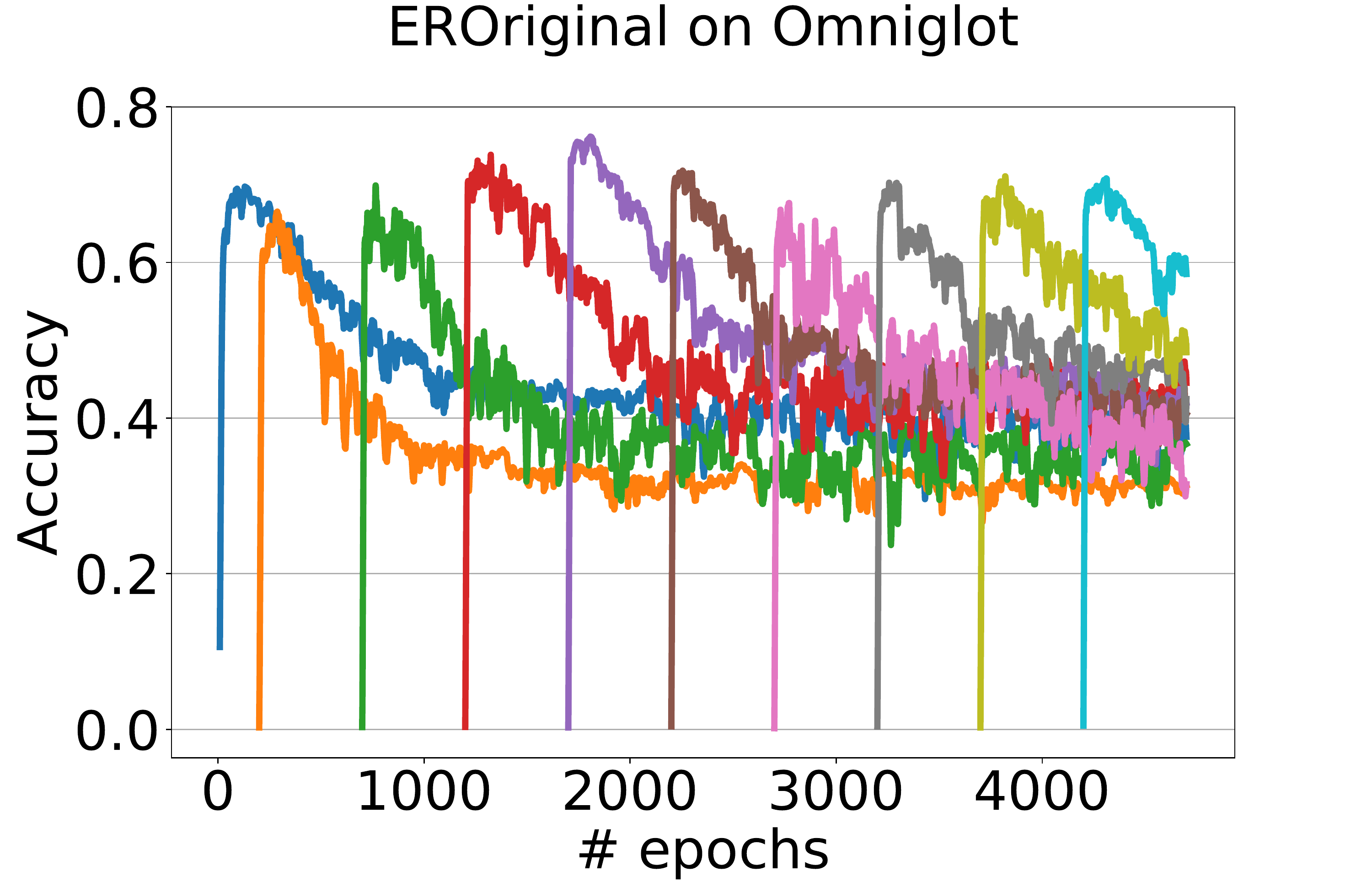}
    \end{subfigure}
    \caption[Full learning curves for supervised lifelong composition: soft ordering.]{Smoothed learning curves with soft layer ordering using \ER{}. Compositional methods did not exhibit decaying performance of early tasks, while joint and no-components baselines did.}%
    \label{fig:softOrderingCurvesNotAveraged}
\end{figure}

\chapter{Additional Results Using EWC for Reinforcement Learning on OpenAI Gym MuJoCo Domains}
\label{app:EWCAdditionalResultsMuJoCo}

The evaluation of Chapter~\ref{cha:RL} on OpenAI MuJoCo domains considered six different variants of EWC, by varying two different choices. The first choice was whether to share the variance of the Gaussian policies across the different tasks. Sharing the variance enables the algorithm to start from a more deterministic policy, thereby achieving higher initial performance, at the cost of reducing task-specific exploration. The second choice was the exact form of the regularization penalty. In the original EWC formulation, the regularization term applied to the PG objective was $-\frac{\lambda}{2}\sum_{\that=1}^{t-1}\Big\|\btheta-\alphathat\Big\|_{\Hthat}^2$. \citet{huszar2018note} noted that this does not correspond to the correct Bayesian formulation, and proposed to instead use \mbox{$-\frac{\lambda}{2} \Big\|\btheta-\balpha^{(t-1)}\Big\|_{\bH^{(t-1)}}^2$}, where $\balpha^{(t-1)}$ and $\bH^{(t-1)}$ capture all the information from tasks $1$ through $t-1$ in the Bayesian setting. Beyond these two choices of regularization, the experiments considered an additional choice where $\lambda$ is scaled by $\frac{1}{t-1}$ in order for the penalty not to increase linearly with the number of tasks. Hyperparameter tuning independently obtained the optimal values for each version, as described in Chapter~\ref{cha:RL}.

\begin{table}[p]
    \captionof{table}[Performance of different versions of EWC on MuJoCo domains used to evaluate lifelong compositional RL with \lpgftw{}.]{Results with different versions of EWC. The first digit differentiates variants with shared $\sigma$ (1) and task-specific $\sigma$ (2), and the second digit differentiates between Husz{\'a}r regularization (1), EWC regularization scaled by $\frac{1}{t-1}$ (2), and the original EWC regularization (3). EWC regularization with $\sigma$ shared across tasks (boldfaced) was the most consistent, so this was chosen for the experiments in Section~\ref{sec:ExperimentalResultsSimple}. NaN's indicate that the learned policies became unstable, leading to failures in the simulator.}
    \label{tab:EWCAdditionalResults}
    \centering
    \begin{tabular}{@{}l|l|c|c|c@{}}
                Domain  & Algorithm     & Start   & Tune    &Final\\
\hline\hline
\multirow{6}{*}{HC\_G}
  & EWC 1, 1   & $-1245$  & $-2917$  & $-3.97e4$\\
  & EWC 1, 2   & $1796$  & $2409$   & $2603$\\
  & \textbf{EWC 1, 3}   & $\bf{1666}$   & $\bf{2225}$   & \bf{2254}\\
  & EWC 2, 1   & $-778$   & $1384$   & $1797$\\
  & EWC 2, 2   & $-762$   & $1565$   & $2238$\\
  & EWC 2, 3   & $-6.58e5$  & $-7e5$  & $-1.05e7$\\
\hline
\multirow{6}{*}{HC\_BP}
  & EWC 1, 1   & $1029$   & $1748$   & $1522$\\
  & EWC 1, 2   & $1132$   & $1769$   & $1588$\\
  & \textbf{EWC 1, 3}   & $\bf{1077}$   & $\bf{1716}$   & $\bf{1571}$\\
  & EWC 2, 1   & $-892$   & $1308$   & $1521$\\
  & EWC 2, 2   & $-1.79e5$  & $-2.16e5$  & $-1.53e6$\\
  & EWC 2, 3   & $-1.1e6$   & $-1.01e6$   & $-5.23e6$\\
\hline
\multirow{6}{*}{Ho\_G}
  & EWC 1, 1   & $1301$   & $2252$   & $1522$\\
  & EWC 1, 2   & $1339$   & $2322$   & $1836$\\
  & \textbf{EWC 1, 3}   & $\bf{1434}$   & $\bf{2488}$   & $\bf{1732}$\\
  & EWC 2, 1   & $872$  & $2616$   & $2089$\\
  & EWC 2, 2   & $930$  & $2582$   & $1900$\\
  & EWC 2, 3   & $939$  & $2520$   & $2029$\\
\hline
\multirow{6}{*}{Ho\_BP}
  & EWC 1, 1   & $613$  & $1508$   & $793$\\
  & EWC 1, 2   & $385$  & $920$  & $43$\\
  & \textbf{EWC 1, 3}   & $\bf{424}$  & $\bf{936}$  & $\bf{31}$\\
  & EWC 2, 1   & $615$  & $2142$   & $1011$\\
  & EWC 2, 2   & $620$  & $2119$   & $1120$\\
  & EWC 2, 3   & $613$  & $2138$   & $928$\\
\hline
\multirow{6}{*}{W\_G}
  & EWC 1, 1   & $1293$   & $2052$   & $303$\\
  & EWC 1, 2   & $-2132$  & $-2181$  &NaN\\
  & \textbf{EWC 1, 3}   & $\bf{2192}$   & $\bf{3901}$   & $\bf{2325}$\\
  & EWC 2, 1   & $-2269$  & $-1915$  & $-2490$\\
  & EWC 2, 2   & $-3.12e4$   & $-3.23e4$   & $-1.15e5$\\
  & EWC 2, 3   & $-8.98e4$   & $-9.65e4$   & $-1.59e5$\\
\hline
\multirow{6}{*}{W\_BP}
  & EWC 1, 1   & $1237$   & $3055$   & $1382$\\
  & EWC 1, 2   & $1148$   & $2800$   & $1306$\\
  & \textbf{EWC 1, 3}   & $\bf{744}$  & $\bf{2000}$   & $-\bf{128}$\\
  & EWC 2, 1   &NaN   &NaN   &NaN\\
  & EWC 2, 2   & $1027$   & $3687$   & $1416$\\
  & EWC 2, 3   &NaN   &NaN   &NaN
    \end{tabular}
\end{table}

Table~\ref{tab:EWCAdditionalResults} summarizes the results obtained for each variant of EWC. The only version that consistently learned each task's policy (tune) was the original EWC regularization with the variance shared across tasks. This was also the only variant for which the final performance was never unreasonably low. Therefore, the experiments in Chapter~\ref{cha:RL} used this version.

\chapter{Proofs of Theoretical Guarantees of \lpgftw{}}
\label{app:proofsOfTheoreticalGuarantees}

This appendix presents complete proofs for the three results on the convergence of \lpgftw{} described in Chapter~\ref{cha:RL}. First, recall the definitions of the actual objective the agent wants to optimize:
\begin{align*}
    g_t(\bL) =& \frac{1}{t}\sum_{\that=1}^{t}\max_{\sthat} \bigg[ \Big\|\alphathat -\bL\sthat\Big\|_{\Hthat}^2 + {\gthat}^\top\Big(\bL\sthat-\alphathat\Big) -  \mu\Big\|\sthat\Big\|_1 \bigg] - \lambda\|\bL\|_{\mathsf{F}}^2\enspace,
\end{align*}
the surrogate objective used for optimizing $\bL$:
\begin{align*}
    \hat{g}_{t}(\bL) =& -\hspace{-0.3em}\lambda\|\bL\|_{\mathsf{F}}^2 + \frac{1}{t} \sum_{\that=1}^{t} \hat{\ell}\Big(\bL, \sthat, \alphathat, \Hthat, \gthat\Big)\enspace,
\end{align*}
and the expected objective:
\begin{align*}
    g(\bL) =& \mathbb{E}_{\Ht, \gt, \alphat} \bigg[\max_{\bs}\hat{\ell}\Big(\bL, \bs, \alphat, \Ht, \gt\Big) \bigg]\enspace,
\end{align*}
with $
    \hat{\ell}(\bL, \bs, \balpha, \bH, \bg) = -\mu\|\bs\|_1 + \|\balpha-\bL\bs\|_{\bH}^2 + \bg^\top(\bL\bs-\balpha)$. 
The convergence results of \lpgftw{} are summarized as: 1)~the knowledge base $\Lt$ becomes increasingly stable, 2)~$\hat{g}_{t}$, $g_{t}$, and $g$ converge to the same value, and 3)~$\Lt$ converges to a stationary point of $g$. These results, given below as Propositions~\ref{prop:app_Stability}, \ref{prop:app_Convergence}, and \ref{prop:app_StationaryPoint}, are based on the following assumptions:
\renewcommand{\theenumi}{\Alph{enumi}}
\begin{enumerate}
\itemsep0em
    \item \label{asmp:app_iid}The tuples $\Big(\Ht,\gt\Big)$ are drawn \iid~from a distribution with compact support.
    \item \label{asmp:app_mixing}The sequence $\Big\{\alphat\Big\}_{t=1}^{\infty}$ is stationary and $\phi$-mixing.
    \item \label{asmp:app_magnitude} The magnitude of $\Objt(\bm{0})$ is bounded by $B$.
    \item \label{asmp:app_uniqueness}For all $\bL$, $\Ht$, $\gt$, and $\alphat$, the largest eigenvalue (smallest in magnitude) of $\bL_{\gamma}^\top\Ht\bL_{\gamma}$ is at most $-2\kappa$, with $\kappa>0$, where $\gamma$ is the set of non-zero indices of \mbox{$\st=\argmax_{\bs}\hat{\ell}\Big(\bL, \bs,\Ht, \gt, \alphat\Big)$}. The non-zero elements of the unique maximizing $\st$ are: \mbox{$\st_{\gamma} = \big(\bL_{\gamma}^\top\Ht\bL_{\gamma}\big)^{-1}\bigg(\bL^\top\Big(\Ht\alphat - \gt\Big) - \mu\sign\Big(\st_{\gamma}\Big)\bigg)$}.
\end{enumerate}
\renewcommand{\theenumi}{\arabic{enumi}}
Note that the $\alphat$'s are not independently obtained, so the proof cannot assume they are \iid~like \citet{ruvolo2013ella}. Therefore, this proof uses a weaker assumption on the sequence of $\alphat$'s found by \lpgftw{}, which enables using the Donsker theorem~\citep{billingsley1968convergence} and the Glivenko-Cantelli theorem~\citep{baklanov2006strong}.

\begin{claim}
    \label{claim:magnitudes}
    $\exists\ c_1,c_2,c_3 \in \Reals$ such that no element of $\Lt$, $\st$, and $\alphat$ has magnitude greater than $c_1$, $c_2$, and $c_3$, respectively,  $\forall t \in \{1,\ldots,\infty\}$.
\end{claim}
\begin{proof}
    This proof corroborates the claim by strong induction. In the base case, $\bL_{1}$ is given by $\argmax_{\bepsilon} \Obj^{(1)}(\bepsilon) - \lambda\|\bepsilon\|_{2}^{2}$. If $\bepsilon=\bm{0}$, the objective becomes $\Obj^{(1)}(\bm{0})$, which is bounded by Assumption~\ref{asmp:app_magnitude}. This implies that if $\bepsilon$ grows too large, $-\lambda\|\bepsilon\|_2$ would be too negative, and then it would not be a maximizer. $\bs^{(1)}=1$ per Algorithm~\ref{alg:LifelongPolicyInitialization}, and so $\balpha^{(1)}=\bL_{1}$, which is bounded. 
    
    Then, for $t\leq k$, $\st$ and $\epsilont$ are given by \mbox{$\argmax_{\bs,\bepsilon} \Objt(\LtMinusOne\bs + \bepsilon) - \mu\|\bs\|_1 - \lambda\|\bepsilon\|_2^2$}. If $\bs=\bm{0}$ and $\bepsilon=\bm{0}$, this becomes $\Objt(\bm{0})$, which is again bounded, and therefore neither $\bepsilon$ nor $\bs$ may grow too large. The bound on $\alphat$ follows by induction, since $\alphat=\LtMinusOne\st$. Moreover, since only the $t-$th column of $\bL$ is modified by setting it to $\bepsilon$, $\Lt$ is also bounded. For $t>k$, the same argument applies to $\st$ and therefore to $\alphat$. $\Lt$ is then given by $\argmax_{\bL} -\lambda\|\bL\|_{\mathsf{F}}^2 + \frac{1}{t} \sum_{\that}^{t} \|\bL\sthat - \alphathat\|_{\Hthat}+ {\gthat}^\top(\bL\sthat - \alphathat)$. If $\Lt=\bm{0}$, the objective for task $\Task^{(\that)}$ becomes ${\alphathat}^\top \Hthat \alphathat + {\gthat}^\top \alphathat$. By Assumption~\ref{asmp:app_iid} and strong induction, this is bounded for all $\that\leq t$, so if any element of $\bL$ is too large, $\bL$ would not be a maximizer because of the regularization term. 
\end{proof}

\begin{proposition}
\label{prop:app_Stability}
    $\bL_{t} - \bL_{t-1} = O(\frac{1}{t})\enspace$.
\end{proposition}
\begin{proof}
    The first step is to show that $\hat{g}_{t}\!-\!\hat{g}_{t-1}$ is Lipschitz with constant $O\left(\frac{1}{t}\right)$. For this, note that $\hat{\ell}$ is Lipschitz in $\bL$ with a constant independent of $t$, since it is a quadratic function over a compact region with bounded coefficients. Next, compute the difference:
\clearpage
\vspace*{-6em}
\begin{align*}
    \hat{g}_{t}(\bL) - \hat{g}_{t-1}(\bL) =& \frac{1}{t} \hat{\ell}\left(\bL,\st,\alphat, \Ht, \gt\right) + \frac{1}{t}\sum_{\that=1}^{t-1} \hat{\ell}\left(\bL, \sthat, \alphathat, \Hthat, \gthat\right) \\ &- \frac{1}{t-1}\sum_{\that=1}^{t-1}\hat{\ell}\left(\bL, \sthat, \alphathat, \Hthat, \gthat\right) \\
        =& \frac{1}{t}\hat{\ell}\left(\bL,\st,\alphat, \Ht, \gt\right) + \frac{1}{t(t-1)}\sum_{\that=1}^{t-1} \hat{\ell}\left(\bL, \sthat, \alphathat, \Hthat, \gthat\right)\enspace.
\end{align*}%
Therefore, $\hat{g}_{t}\!-\!\hat{g}_{t\!-\!1}$ has a Lipschitz constant $O\left(\frac{1}{t}\right)$, since it is the difference of two terms divided by $t$: $\hat{\ell}$ and an average over $t\!-\!1$ terms, whose Lipschitz constant is bounded by the largest Lipschitz constant of the terms.

Let $\xi_t$ be the Lipschitz constant of $\hat{g}_{t}-\hat{g}_{t-1}$. This gives:
\begin{align*}
    \hat{g}_{t-1}(\bL_{t-1}) -\hat{g}_{t-1}(\bL_{t}) =&\  \hat{g}_{t-1}(\LtMinusOne) - \hat{g}_{t}(\LtMinusOne) + \hat{g}_{t}(\LtMinusOne) -  \hat{g}_{t}(\Lt) + \hat{g}_{t}(\Lt) - \hat{g}_{t-1}(\Lt)\\
    \leq&\  \hat{g}_{t-1}(\LtMinusOne) - \hat{g}_{t}(\LtMinusOne) + \hat{g}_{t}(\Lt) - \hat{g}_{t-1}(\Lt)\\
    =& -(\hat{g}_{t}-\hat{g}_{t-1})(\LtMinusOne) + (\hat{g}_{t}-\hat{g}_{t-1})(\Lt)
    \leq \xi_{t}\|\Lt-\LtMinusOne\|_\mathsf{F}\enspace.
\end{align*}
Moreover, since $\LtMinusOne$ maximizes $\hat{g}_{t-1}$ and the $\ell_2$ regularization term ensures that the maximum eigenvalue of the Hessian of $\hat{g}_{t-1}$ is upper-bounded by $-2\lambda$, then: 
\[
\hat{g}_{t-1}(\LtMinusOne)-\hat{g}_{t-1}(\Lt)\geq\lambda\|\Lt-\LtMinusOne\|_\mathsf{F}^2\enspace.
\]
Combining these two inequalities gives:
\begin{equation*}
    \|\Lt-\LtMinusOne\|_{\mathsf{F}} \leq \frac{\xi_{t}}{\lambda}=O\left(\frac{1}{t}\right)\qedhere
\end{equation*}
\end{proof}
The critical step for adapting the proof from \citet{ruvolo2013ella} to \lpgftw{} is to introduce the following lemma, which shows the equality of the maximizers of $\ell$ and $\hat{\ell}$.

\begin{lemma}
\label{lma:app_EqualityOfMaximizers} $
    \hat{\ell}\Big(\bL_{t}, \bs^{(t+1)}, \balpha^{(t+1)}, \bH^{(t+1)}, \bg^{(t+1)}\Big) = \max_{\bs} \hat{\ell}\Big(\bL_{t}, \bs, \balpha^{(t+1)}, \bH^{(t+1)}, \bg^{(t+1)}\Big)\enspace$.
\end{lemma}

\begin{proof}
    Showing this requires the following to hold:
\begin{align*}
    \bs^{(t+1)} =\argmax_{\bs} \ell(\bL_{t}, \bs) = \argmax_{\bs} \hat{\ell}\Big(\bL_{t}, \bs, \balpha^{(t+1)}, \bH^{(t+1)}, \bg^{(t+1)}\Big)\enspace.
\end{align*}

First, compute the gradient of $\ell$, given by:
\begin{align*}
    \nabla_{\bs}\ell(\Lt, \bs)
         =& -\mu\sign(\bs) + \Lt^\top \nabla_{\btheta}\Obj^{(t+1)}(\btheta)\bigg\lvert_{\btheta=\bL_{t}\bs}\enspace.
\end{align*}
Since $\bs^{(t+1)}$ is the maximizer of $\ell$, then:
\begin{align}
    \label{equ:gradEllH}
    \nabla_{\bs}\ell(\bL_{t}, \bs)\bigg\lvert_{\bs=\bs^{(t+1)}} \hspace{-3em}= -\mu\sign\Big(\bs^{(t+1)}\Big) + \bL_{t}^\top\bg^{(t+1)} 
        = \bm{0}\enspace.
\end{align}
Now compute the gradient of $\hat{\ell}$ and evaluate it at $\bs^{(t+1)}$:
\begin{align*}
    \nabla_{\bs}\hat{\ell}\Big(\!\Lt, \bs, \balpha^{(t+1)}\!\!, \bH^{(t+1)}\!\!, \bg^{(t+1)}\!\Big) = -\!\mu\sign\Big(\!\bs^{(t+1)}\!\Big) + \bL_{t}\bg^{(t+1)} - 2\bL_{t}^\top\bH^{(t+1)}\Big(\!\balpha^{(t+1)} \!-\! \bL_{t}\bs\!\Big)
\end{align*}
\begin{align*}
    \nabla_{\bs}\hat{\ell}\Big(\bL_{t}, \bs, \balpha^{(t+1)}, \bH^{(t+1)}, \bg^{(t+1)}\Big) \bigg\lvert_{\bs=\bs^{(t+1)}} = -\mu\sign\Big(\bs^{(t+1)}\Big) + \bL_{t}^\top\bg^{(t+1)}= \bm{0}\enspace,
\end{align*}
since it matches Equation~\ref{equ:gradEllH}. By Assumption~\ref{asmp:app_uniqueness}, $\hat{\ell}$ has a unique maximizer $\bs^{(t+1)}$.
\end{proof}

Before stating the next lemma, define:
\begin{align*}
    \bs^* =& \beta\Big(\bL, \alphat, \Ht, \gt\Big) = \argmax_{\bs} \hat{\ell}\Big(\bL, \bs, \alphat, \Ht, \gt\Big)\enspace.
\end{align*}

\begin{lemma}\ 
\label{lma:app_Lipschitz}
\renewcommand{\theenumi}{\Alph{enumi}}
\begin{enumerate}
\itemsep0em
    \item \label{part:app_firstLemmaLipschitz} $\max_{\bs}\hat{\ell}\Big(\bL, \bs, \alphat, \Ht, \gt\Big)$ is continuously differentiable in $\bL$ with $$\nabla_{\bL}\max_{\bs}\hat{\ell}(\bL, \bs, \alphat, \Ht, \gt) =\Big[-2\Ht\bs^* + \gt\Big]{\bs^*}^\top\enspace.$$
    \item \label{part:app_secondLemmaLipschitz} $g$ is continuously differentiable and $$\nabla_{\bL} g(\bL) = -2\lambda\bI + \mathbb{E}\Big[\nabla_{\bL} \max_{\bs}\hat{\ell}\Big(\bL, \bs, \alphat, \Ht, \gt\Big) \Big]\enspace.$$
    \item \label{part:app_thirdLemmaLipschitz} $\nabla_{\bL} g(\bL)$ is Lipschitz in the space of latent components $\bL$ that obey Claim~\ref{claim:magnitudes}.
\end{enumerate}
\renewcommand{\theenumi}{\arabic{enumi}}
\end{lemma}
\begin{proof}
    To prove Part~\ref{part:app_firstLemmaLipschitz}, apply a corollary to Theorem 4.1 in~\citep{bonnans1998optimization}. 
    This corollary states that if $\hat{\ell}$ is continuously differentiable in $\bL$ (which it clearly is) and has a unique maximizer $\st$ for any $\alphat$, $\Ht$, and $\gt$ (which is guaranteed by Assumption~\ref{asmp:app_uniqueness}), then \mbox{$\nabla_{\bL}\min_{\bs}\hat{\ell}\Big(\bL, \bs, \alphat\!, \Ht\!, \gt\Big)$} exists and is equal to \mbox{$\nabla_{\bL}\hat{\ell}\Big(\bL, \bs^*, \alphat\!, \Ht\!, \gt\Big)$}, given by \mbox{${\Big[\!\!-2\Ht\bs^* + \gt\Big]{\bs^*}^\top}$}. Part~\ref{part:app_secondLemmaLipschitz} follows since, by Assumption~\ref{asmp:app_iid} and Claim~\ref{claim:magnitudes}, the tuple $\Big(\Ht, \gt, \alphat\Big)$ is drawn from a distribution with compact support. 
    
    To prove Part~\ref{part:app_thirdLemmaLipschitz}, first show that $\beta$ is Lipschitz in $\bL$ with constant independent of $\alphat$, $\Ht$, and $\gt$. Part~\ref{part:app_thirdLemmaLipschitz} follows due to the form of the gradient of $g$ with respect to $\bL$. The function $\beta$ is continuous in its arguments since $\hat{\ell}$ is continuous and by Assumption~\ref{asmp:app_uniqueness} has a unique maximizer. Next, define \mbox{$\rho\Big(\bL, \Ht, \gt, \alphat, j\Big) = \bl_{j}^\top\Big[2\Ht\Big(\bL\bs^{*}-\alphat\Big)+\gt\Big]$}, where  $\bl_{j}$ is the $j$-th column of $\bL$. The argument of \citet{fuchs2005recovery} yields the following conditions:
\begin{align}
    \nonumber\Big|\rho\Big(\bL, \Ht, \gt, \alphat, j\Big)\Big| =& \mu \Longleftrightarrow \bs_{j}^{*}\neq0\\
    \Big|\rho\Big(\bL, \Ht, \gt, \alphat, j\Big)\Big| <& \mu \Longleftrightarrow \bs_{j}^{*}=0\enspace. 
    \label{equ:rhoLeqMu}
\end{align}
Let $\gamma$ be the set of indices $j$ such that $\Big|\rho\Big(\bL, \Ht, \gt, \alphat, j\Big)\Big| = \mu$. Since $\rho$ is continuous in $\bL$, $\Ht$, $\gt$, and $\alphat$, there must exist an open neighborhood $V$ around $\Big(\bL, \Ht, \gt, \alphat\Big)$ such that for all $\Big(\bL^\prime, {\Ht}^\prime, {\gt}^\prime, {\alphat}^\prime\Big)\in V$ and $j\notin\gamma$, $\Big|\rho\Big(\bL^\prime, {\Ht}^\prime, {\gt}^\prime, {\alphat}^\prime, j\Big)\Big| < \mu$. By Equation~\ref{equ:rhoLeqMu}, it follows that $\beta\Big(\bL^\prime, {\Ht}^\prime, {\gt}^\prime, {\alphat}^\prime\Big)_{j}=0, \forall j\notin\gamma$.

Next, define a new objective:
\begin{align*}
    \bar{\ell}(\bL_{\gamma},\bs_{\gamma},\balpha,\bH,\bg) =& \|\balpha-\bL_{\gamma}\bs_{\gamma}\|_{\bH}^2 + \bg^\top(\bL_{\gamma}\bs_{\gamma}-\balpha)- \mu\|\bs_{\gamma}\|_1\enspace.
\end{align*}
By Assumption~\ref{asmp:app_uniqueness}, $\bar{\ell}$ is strictly concave with a Hessian bounded by $-2\kappa$, implying that:
\begin{align}
    \nonumber
    \bar{\ell}\Big(\!\bL_{\gamma}, \beta\Big(\!\bL, {\alphat}, {\Ht},&\, {\gt}\!\Big)_{\gamma}, \alphat, \Ht, \gt\!\Big)\!-\!\bar{\ell}\Big(\!\bL_{\gamma}, \beta\Big(\!\bL^\prime, {\alphat}^\prime, {\Ht}^\prime, {\gt}^\prime\!\Big)_{\gamma}, \alphat, \Ht, \gt\!\Big)\\
        \geq& \kappa\Big\|\beta\Big(\bL^\prime, {\alphat}^\prime, {\Ht}^\prime, {\gt}^\prime\Big)_{\gamma}
        \label{equ:ellBarDistance}
            - \beta\Big(\bL, {\alphat}, {\Ht}, {\gt}\Big)_{\gamma} \Big\|_2^2\enspace.
\end{align}

On the other hand, by Assumption~\ref{asmp:app_iid} and Claim~\ref{claim:magnitudes}, $\bar{\ell}$ is Lipschitz in its second argument with constant ${e_1\|\bL_{\gamma} - \bL_{\gamma}^\prime\|_\mathsf{F}} + {e_2 \|\balpha-\balpha^\prime\|_2} + {e_3\|\bH-\bH^\prime\|_\mathsf{F}}+{e_4\|\bg-\bg^\prime\|_2}$, where $e_{1\text{--}4}$ are all constants independent of any of the arguments. Combining this with Equation~\ref{equ:ellBarDistance} gives:
\begin{align*}
    \Big\|\beta\Big(\bL^\prime, {\alphat}^\prime,&\, {\Ht}^\prime, {\gt}^\prime\Big) - \beta\Big(\bL, {\alphat}, {\Ht}, {\gt}\Big) \Big\|=\\
        &\Big\|\beta\Big(\bL^\prime, {\alphat}^\prime, {\Ht}^\prime, {\gt}^\prime\Big)_{\gamma}- \beta\Big(\bL, {\alphat}, {\Ht}, {\gt}\Big)_{\gamma} \Big\|\\
        \leq& \frac{e_1\|\bL_{\gamma} - \bL_{\gamma}^\prime\|_\mathsf{F} + e_2 \Big\|\alphat-{\alphat}^\prime\Big\|_2}{\kappa}+ \frac{e_3\Big\|\Ht-{\Ht}^\prime\Big\|_\mathsf{F} + e_4\Big\|\gt-{\gt}^\prime\Big\|_2}{\kappa}\enspace.
\end{align*}
Therefore, $\beta$ is locally Lipschitz. Since the domain of $\beta$ is compact by Assumption~\ref{asmp:app_iid} and Claim~\ref{claim:magnitudes}, this implies that $\beta$ is uniformly Lipschitz, and therefore $\nabla g$ is Lipschitz as well.
\end{proof}

\begin{proposition}
\label{prop:app_Convergence}~
    \begin{enumerate}
        \item \label{part:app_firstPropositionConvergence} $\hat{g}_{t}(\bL_{t})$ converges a.s.
        \item \label{part:app_secondPropositionConvergence}$g_{t}(\bL_{t}) -\hat{g}_{t}(\bL_{t})$ converges a.s.~to 0
        \item \label{part:app_thirdPropositionConvergence} {$g_{t}(\bL_{t}) - \hat{g}(\bL_{t})$ converges a.s.~to 0}
        \item \label{part:app_fourthPropositionConvergence} $g(\bL_{t})$ converges a.s.
    \end{enumerate}
\end{proposition}

\begin{proof}
Begin by defining the stochastic process $u_t = \hat{g}_{t}(\bL)$.
The proof shows that this process is a quasi-martingale and by a theorem by \citet{fisk1965quasi}, it converges almost surely. 
\begin{align}
    \nonumber
    u_{t+1} - u_{t} =& \hat{g}_{t+1}(\bL_{t+1}) - \hat{g}_{t}(\bL_{t}) = \hat{g}_{t+1}(\bL_{t+1}) - \hat{g}_{t+1}(\bL_{t})+ \hat{g}_{t+1}(\bL_{t}) - \hat{g}_{t}(\bL_{t})\\
        \nonumber
        =& (\hat{g}_{t+1}(\bL_{t+1}) - \hat{g}_{t+1}(\bL_{t}))+ \frac{g_{t}(\bL_{t}) - \hat{g}_{t}(\bL_{t})}{t+1}\\
            &+ \frac{\max_{\bs}\hat{\ell}\Big(\bL_{t}, \bs, \balpha^{(t+1)}, \bH^{(t+1)}, \bg^{(t+1)}\Big)}{t+1}- \frac{ g_{t}(\bL_{t})}{t+1}
        \enspace, \label{equ:uTSequence}
\end{align}
which made use of the fact that:
\begin{align*}
    \hat{g}_{t+1}(\bL_{t}) =& \frac{\hat{\ell}\Big(\bL_{t}, \bs^{(t+1)}, \balpha^{(t+1)}, \bH^{(t+1)}, \bg^{(t+1)}\Big) }{t+1} + \frac{t}{t+1}\hat{g}_{t}(\bL_{t})\\
        =& \frac{\max_{\bs}\hat{\ell}\Big(\bL_{t}, \bs, \balpha^{(t+1)}, \bH^{(t+1)}, \bg^{(t+1)}\Big)}{t+1}  + \frac{t}{t+1}\hat{g}_{t}(\bL_{t})\enspace,
\end{align*}
where the second equality holds by Lemma~\ref{lma:app_EqualityOfMaximizers}. 

The next step is to show that the sum of positive and negative variations in Equation~\ref{equ:uTSequence} are bounded. By an argument similar to a lemma by \citet{bottou1998online}, the sum of positive variations of $u_t$ is bounded, since $\hat{g}$ is upper-bounded by Assumption~\ref{asmp:app_magnitude}. Therefore, it suffices to show that the sum of negative variations is bounded. The first term on the first line of Equation~\ref{equ:uTSequence} is guaranteed to be positive since $\bL_{t+1}$ maximizes $\hat{g}_{t+1}$. Additionally, since $g_{t}$ is always at least as large as $\hat{g}_{t}$, the second term on the first line is also guaranteed to be positive. Therefore, this step can focus on the second line.
\begin{align}
    \nonumber
    \mathbb{E}[u_{t+1}-u_{t} \mid \mathcal{I}_{t}] \geq& \frac{\mathbb{E}\Big[\max_{\bs}\hat{\ell}\Big(\bL_{t}, \bs, \balpha^{(t+1)}, \bH^{(t+1)}, \bg^{(t+1)}\Big)\mid\mathcal{I}_t\Big]}{t+1} - \frac{g_t(\Lt)}{t+1}\\
    \nonumber
    =& \frac{g(\Lt)-g_{t}(\Lt)}{t+1} 
    \geq -\frac{\|g - g_{t}\|_\infty}{t+1}\enspace,
\end{align}
where $\mathcal{I}_{t}$ represents all the $\alphathat$'s, $\Hthat$'s, and $\gthat$'s up to time $t$. Hence, showing that $\sum_{t=1}^{\infty}\frac{\|g-g_t\|_\infty}{t+1}<\infty$ will prove that $u_t$ is a quasi-martingale that converges almost surely.

To prove this, apply the following corollary of the Donsker theorem~\Citep{van2000asymptotic}:
\hfill\begin{adjustwidth}{1cm}{1cm}
\singlespacing
    Let $\mathcal{F} = \{f_{\btheta} : \mathcal{X} \mapsto \Reals, \btheta\in \bm{\Theta}\}$ be a set of measurable functions indexed by a bounded subset $\bm{\Theta}$ of $\Reals^{d}$. Suppose that there exists a constant $K$ such that:
    \begin{equation*}
        |f_{\btheta_1}(x)- f_{\btheta_2}(x)| \leq K \|\btheta_1-\btheta_2\|_2
    \end{equation*}
    for every $\btheta_1,\btheta_2\in\bm{\Theta}$ and $x\in \mathcal{X}$. Then, $\mathcal{F}$ is P-Donsker and for any $f\in\mathcal{F}$, we define:
    \begin{align*}
        \mathbb{P}_n f =& \frac{1}{n} \sum_{i=1}^n f(X_i)\\
        \mathbb{P}f =& \mathbb{E}_{X} [f(X)]\\
        \mathbb{G}_n f =& \sqrt{n} (\mathbb{P}_n f -\mathbb{P}f)\enspace.
    \end{align*}
    If $\mathbb{P}f^2\leq\delta^2$ and $\|f\|_\infty<B$ and the random elements are Borel measurable, then:
    \begin{equation*}
        \mathbb{E}[\sup_{f\in\mathcal{F}} |\mathbb{G}_n f|] = O(1)\enspace.
    \end{equation*}
\end{adjustwidth}
In order to apply this corollary to this analysis, consider a set of functions $\mathcal{F}$ indexed by $\bL$, given by $f_{\bL}\Big(\Ht,\gt,\alphat\Big)=\max_{\bs} \hat{\ell}\Big(\bL,\bs,\alphat,\Ht,\gt\Big)$, whose domain is all possible tuples $\Big(\Ht, \gt, \alphat\Big)$. The expected value of $f^2$ is bounded for all $f\in\mathcal{F}$ since  $\hat{\ell}$ is bounded by Claim~\ref{claim:magnitudes}. Second, $\|f\|_\infty$ is bounded given Claim~\ref{claim:magnitudes} and Assumption~\ref{asmp:app_iid}. Finally, by Assumptions~\ref{asmp:app_iid} and \ref{asmp:app_mixing}, the corollary applies to the tuples $\Big(\Ht, \gt, \alphat\Big)$ \citep{billingsley1968convergence}. Therefore:
\begin{align*}
    \mathbb{E} \Bigg[\sqrt{t} \Bigg\|\Bigg(\frac{1}{t} \sum_{\that=1}^{t} \max_{\bs} \hat{\ell}\Big(\bL,\bs,\alphathat,\Hthat,\gthat\Big)\Bigg) -&\, \mathbb{E}\Big[ \max_{\bs}\hat{\ell}\Big(\bL,\bs,\alphathat,\Hthat,\gthat\Big)\Big]\Bigg\|_\infty\Bigg] = O(1)\\
    \Longrightarrow& \mathbb{E}[\|g_{t}(\bL) - g(\bL) \|_\infty] = O\left(\frac{1}{\sqrt{t}}\right)\enspace.
\end{align*}
Therefore, $\exists\ c_3 \in \Reals$ such that $\mathbb{E}[\|g_{t}-g\|_\infty]<\frac{c_3}{\sqrt{t}}$:
\begin{align*}
    \sum_{t=1}^\infty \mathbb{E}\Big[\mathbb{E}[u_{t+1}-u_{t}\mid\mathcal{I}_t]^{-}\Big] \geq& \sum_{t=1}^\infty -\frac{\mathbb{E}[\|g_t - g\|_\infty]}{t+1}
        > \sum_{t=1}^\infty -\frac{c_3}{t^{\frac{3}{2}}} = - O(1)\enspace,
\end{align*}
where $i^{-}\!=\!\min(i, 0)$. This shows that the sum of negative variations of $u_t$ is bounded, so $u_{t}$ is a quasi-martingale and thus converges almost surely~\citep{fisk1965quasi}. This proves Part~\ref{part:app_firstPropositionConvergence} of Proposition~\ref{prop:app_Convergence}.

Next, show that $u_t$ being a quasi-martingale implies the almost sure convergence of the fourth line of Equation~\ref{equ:uTSequence}. To see this, note that since $u_t$ is a quasi-martingale and the sum of its positive variations is bounded, and since the term on the fourth line of Equation~\ref{equ:uTSequence}, $\frac{g_{t}(\Lt)-\hat{g}_{t}(\Lt)}{t+1}$, is  positive, the sum of that term from 1 to infinity must be bounded:
\begin{equation}
    \label{equ:PropositionConvergenceSecondPart}
    \sum_{t=1}^\infty \frac{g_{t}(\Lt)-\hat{g}_{t}(\Lt)}{t+1} < \infty\enspace.
\end{equation}
To complete the proof of Part~\ref{part:app_secondPropositionConvergence} of Proposition~\ref{prop:app_Convergence}, consider the following lemma: Let $a_n, b_n$ be two real sequences such that for all $n$, \mbox{$a_n\geq0$}, \mbox{$b_n\geq0$}, \mbox{$\sum_{j=1}^\infty a_j=\infty$}, \mbox{$\sum_{j=1}^\infty a_j b_j<\infty$}, \mbox{$\exists\ K > 0$} such that \mbox{$|b_{n+1} - b_n| < Ka_n$}. Then, \mbox{$\lim_{n\to\infty} b_n=0$}. Define the sequences \mbox{$a_t=\frac{1}{t+1}$} and \mbox{$b_t=g_t(\Lt) - \hat{g}_t(\Lt)$}; clearly these are both positive sequences and \mbox{$\sum_{t=1}^{\infty}a_t=\infty$}. By Equation~\ref{equ:PropositionConvergenceSecondPart}, \mbox{$\sum_{t=1}^\infty a_n b_n < \infty$}. Finally, since $g_t$ and $\hat{g}_t$ are bounded and Lipschitz with constant independent of $t$ and $\bL_{t+1}-\Lt=O\left(\frac{1}{t}\right)$, all of the assumptions are verified, which implies that $g_{t} - \hat{g}_{t}$ converges a.s.~to 0.

By Part~\ref{part:app_secondPropositionConvergence} and the Glivenko-Cantelli theorem, $\lim_{t\to\infty}\|g - g_{t}\|_\infty=0$, which implies that $g$ must converge almost surely. By transitivity, $\lim_{t\to\infty} g(\Lt) - \hat{g}_{t}(\Lt)=0$, showing Parts~\ref{part:app_thirdPropositionConvergence} and \ref{part:app_fourthPropositionConvergence}.
\end{proof}

\begin{proposition}
\label{prop:app_StationaryPoint}
    The distance between $\Lt$ and the set of all stationary points of $g$ converges a.s.~to 0.
\end{proposition}
\begin{proof}
First, $\nabla_{\bL}\hat{g}_{t}$ is Lipschitz with a constant independent of $t$, since the gradient of $\hat{g}_{t}$ is linear, $\st$, $\Ht$, $\gt$, and $\alphat$ are bounded, and the summation in $\hat{g}_{t}$ is normalized by $t$. Next, define an arbitrary non-zero matrix $\bm{U}$ of the same dimensionality as $\bL$. The fact that $g_{t}$ upper-bounds $\hat{g}_{t}$ implies that:
\clearpage
\vspace*{-6em}
\begin{align*}
    g_{t}(\Lt+\bm{U}) \geq& \hat{g}_{t}(\Lt+\bm{U}) \Longrightarrow
    \lim_{t\to\infty} g(\Lt+\bm{U}) \geq \lim_{t\to\infty} \hat{g}_{t}(\Lt+\bm{U})\enspace,
\end{align*}
which used the fact that $\lim_{t\to\infty}g_{t}=\lim_{t\to\infty}g$. Let $h_{t}>0$ be a sequence of positive real numbers that converges to $0$. Taking the first-order Taylor expansion on both sides of the inequality and using the fact that $\nabla g$ and $\nabla \hat{g}$ are both Lipschitz with constant independent of $t$ gives:
\begin{align*}
    \lim&_{t\to\infty} g_{t}(\Lt) \!+\! \tr\!\big(h_t\bm{U}^\top\nabla g_{t}(\Lt)\big) \!+\! O(h_t\bm{U}) \geq \lim_{t\to\infty} \hat{g}_{t}(\Lt) \!+\! \tr\!\big(h_t\bm{U}^\top\nabla \hat{g}_{t}(\Lt)\big) \!+\! O(h_t\bm{U})\enspace\!.
\end{align*}
Since $\lim_{t\to\infty} g(\Lt) - \hat{g}(\Lt)=0$ a.s.~and $\lim_{t\to\infty}h_t=0$:
\begin{align*}
    \lim_{t\to\infty} \Bigg(\frac{1}{\|\bm{U}\|_\mathsf{F}}\bm{U}^\top&\nabla g(\Lt)\Bigg) \geq \lim_{t\to\infty} \Bigg(\frac{1}{\|\bm{U}\|_\mathsf{F}}\bm{U}^\top\nabla \hat{g}(\Lt)\Bigg) \enspace.
\end{align*}
Since this inequality has to hold for every $\bm{U}$, then $\lim_{t\to\infty}\nabla g(\Lt) = \lim_{t\to\infty} \nabla\hat{g}_{t}(\Lt)$. Since $\Lt$ minimizes $\hat{g}_{t}$, then $\nabla\hat{g}_{t}(\Lt)=\bm{0}$. This implies that $\nabla g(\Lt)=\bm{0}$, which is a sufficient first-order condition for $\Lt$ to be a stationary point of $g$.
\end{proof}

\chapter{Visualization of All \benchmark{} Tasks}
\label{app:VisualizationOfAllTasks}

\benchmark{} consists of a total of $256$ possible combinations of elements, each representing a separate task. Figures~\ref{fig:iiwa_viz}--~\ref{fig:kinova_viz} show each of the different robot arms in action, solving the diversity of tasks in \benchmark{}. 

\begin{figure}[H]
\centering
    \hspace{0.16\textwidth}
    \centering
        \includegraphics[width=0.95\linewidth, trim={0.3cm 0.0cm 1.5cm 0.8cm}]{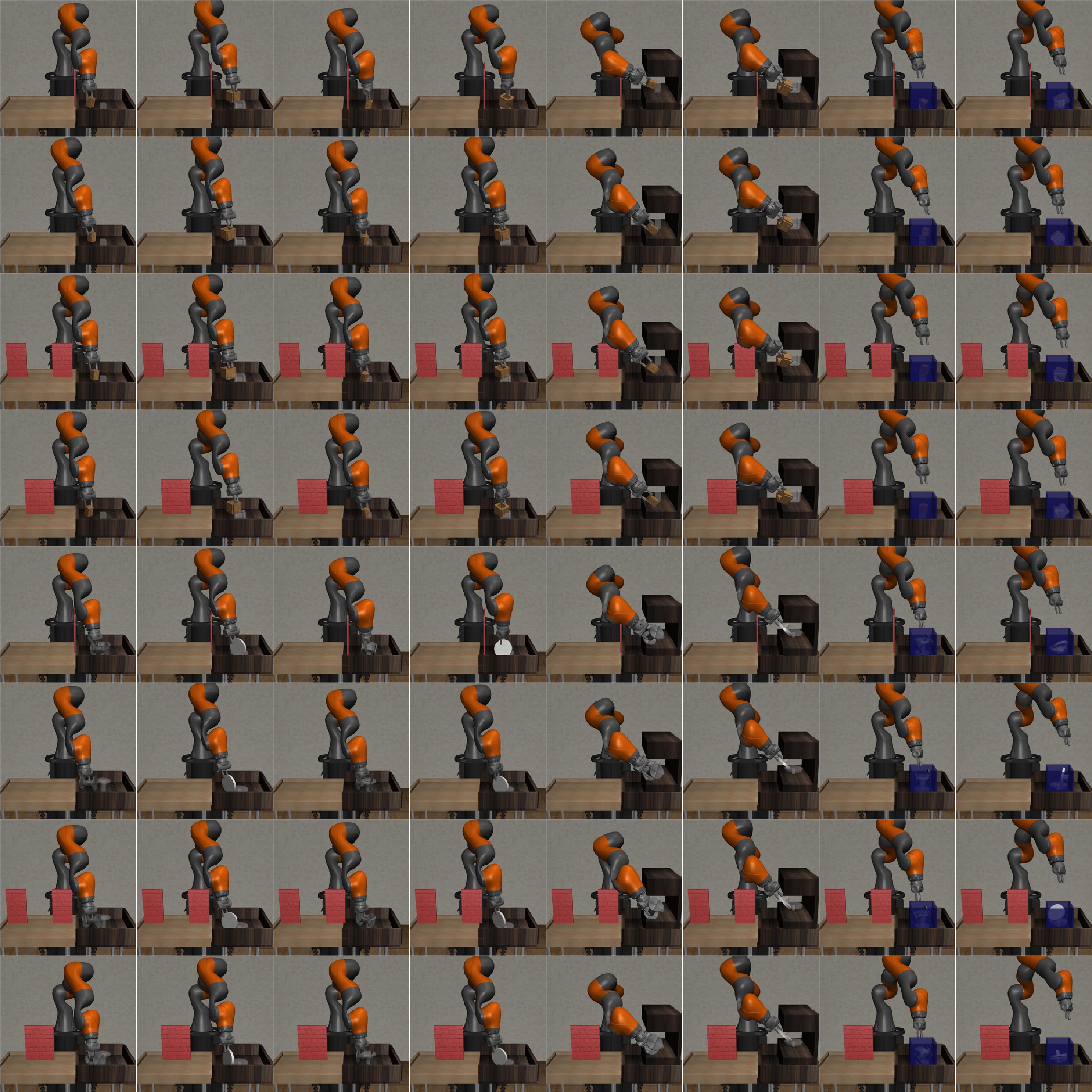}
    \caption[Visualization of the $64$ IIWA tasks.]{Visualization of the $64$ \IIWA{} tasks.}
    \label{fig:iiwa_viz}
\end{figure}

\begin{figure}[t!]
\centering
    \hspace{0.16\textwidth}
    \centering
        \includegraphics[width=0.95\linewidth, trim={0.3cm 0.0cm 1.5cm 0.8cm}]{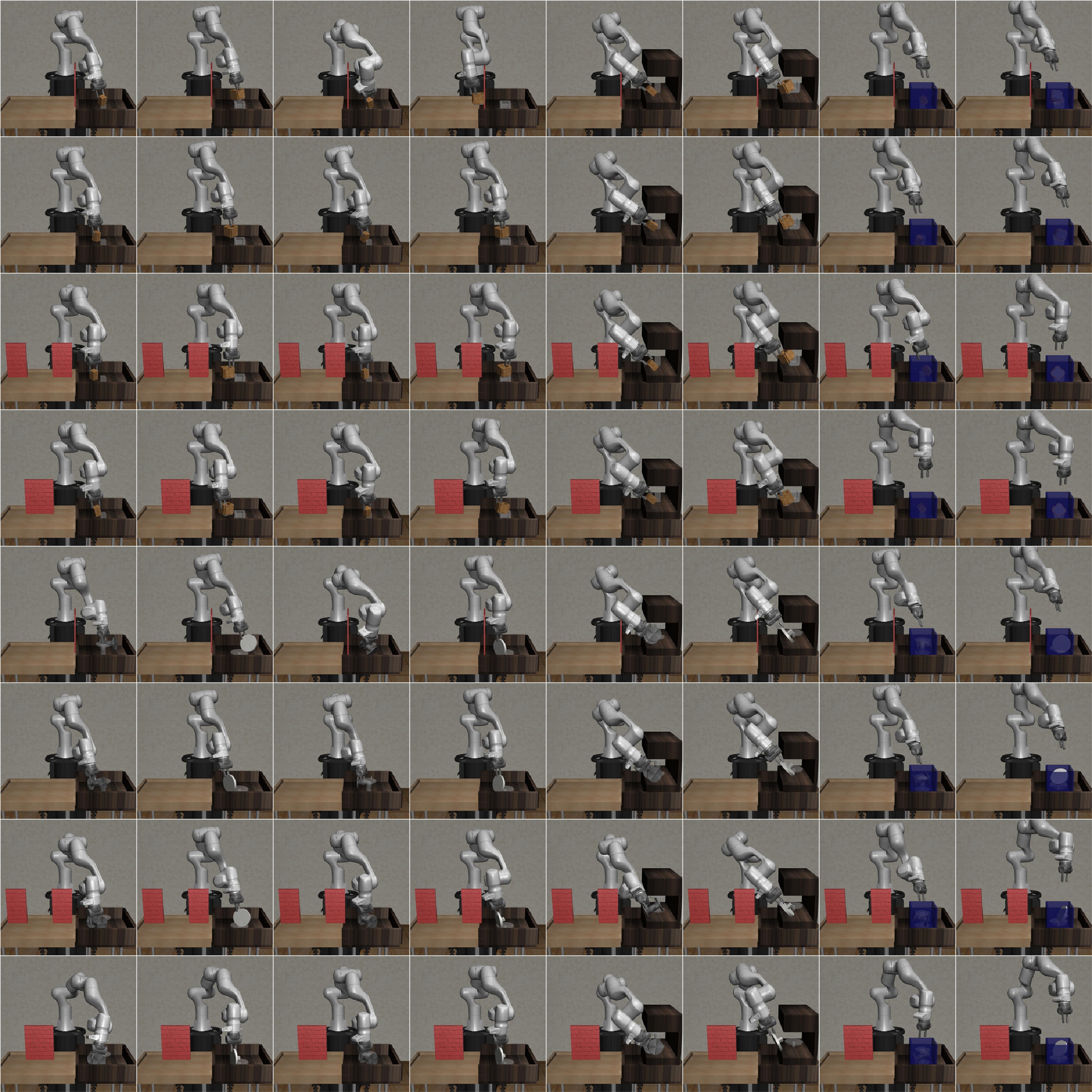}
    \caption[Visualization of the $64$ Panda tasks.]{Visualization of the $64$ \Panda{} tasks.}
    \label{fig:panda_viz}
\end{figure}

\begin{figure}[t!]
\centering
    \hspace{0.16\textwidth}
    \centering
        \includegraphics[width=0.95\linewidth, trim={0.3cm 0.0cm 1.5cm 0.8cm}]{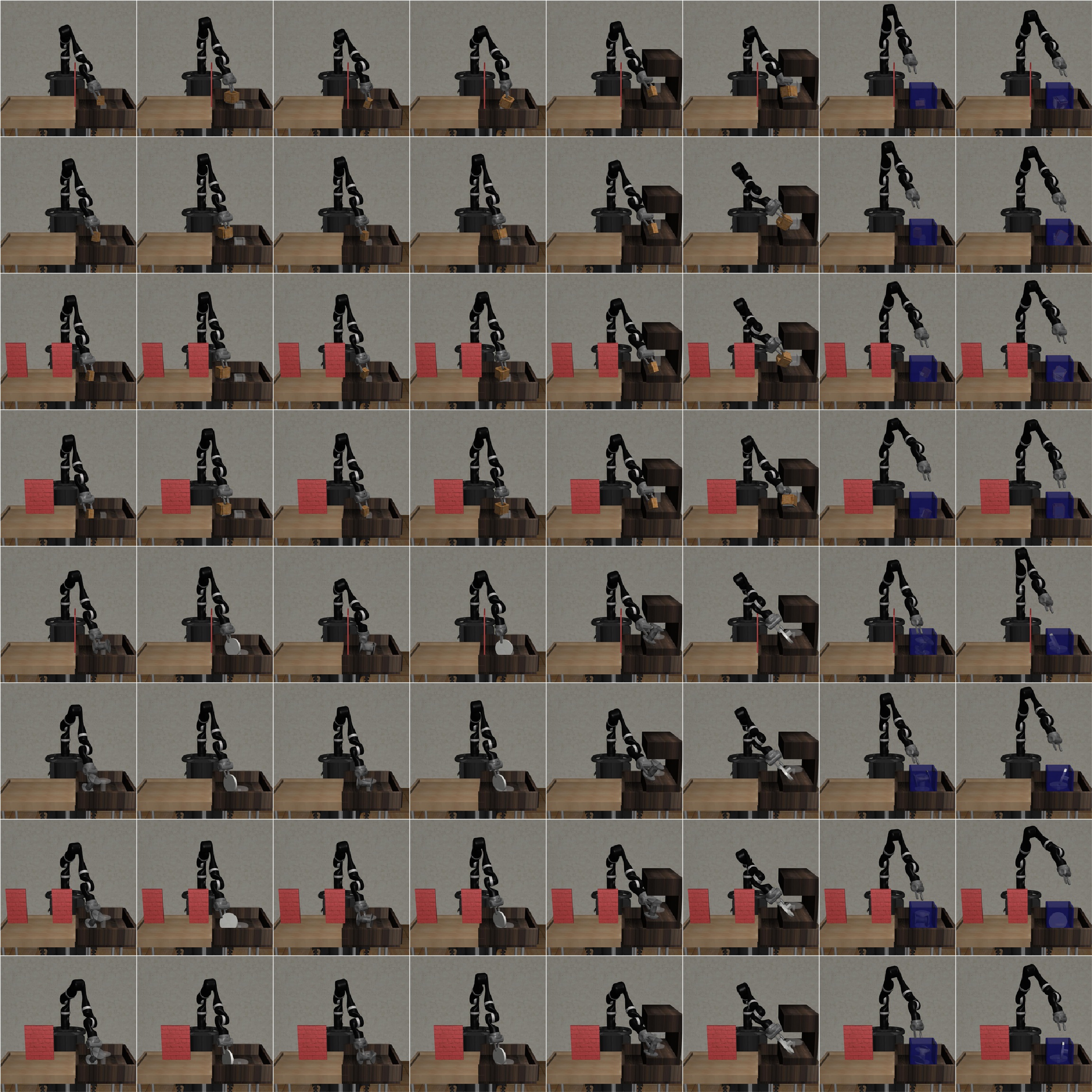}
    \caption[Visualization of the $64$ Jaco tasks.]{Visualization of the $64$ \Jaco{} tasks.}
    \label{fig:jaco_viz}
\end{figure}

\begin{figure}[t!]
\centering
    \hspace{0.16\textwidth}
    \centering
        \includegraphics[width=0.95\linewidth, trim={0.3cm 0.0cm 1.5cm 0.8cm}]{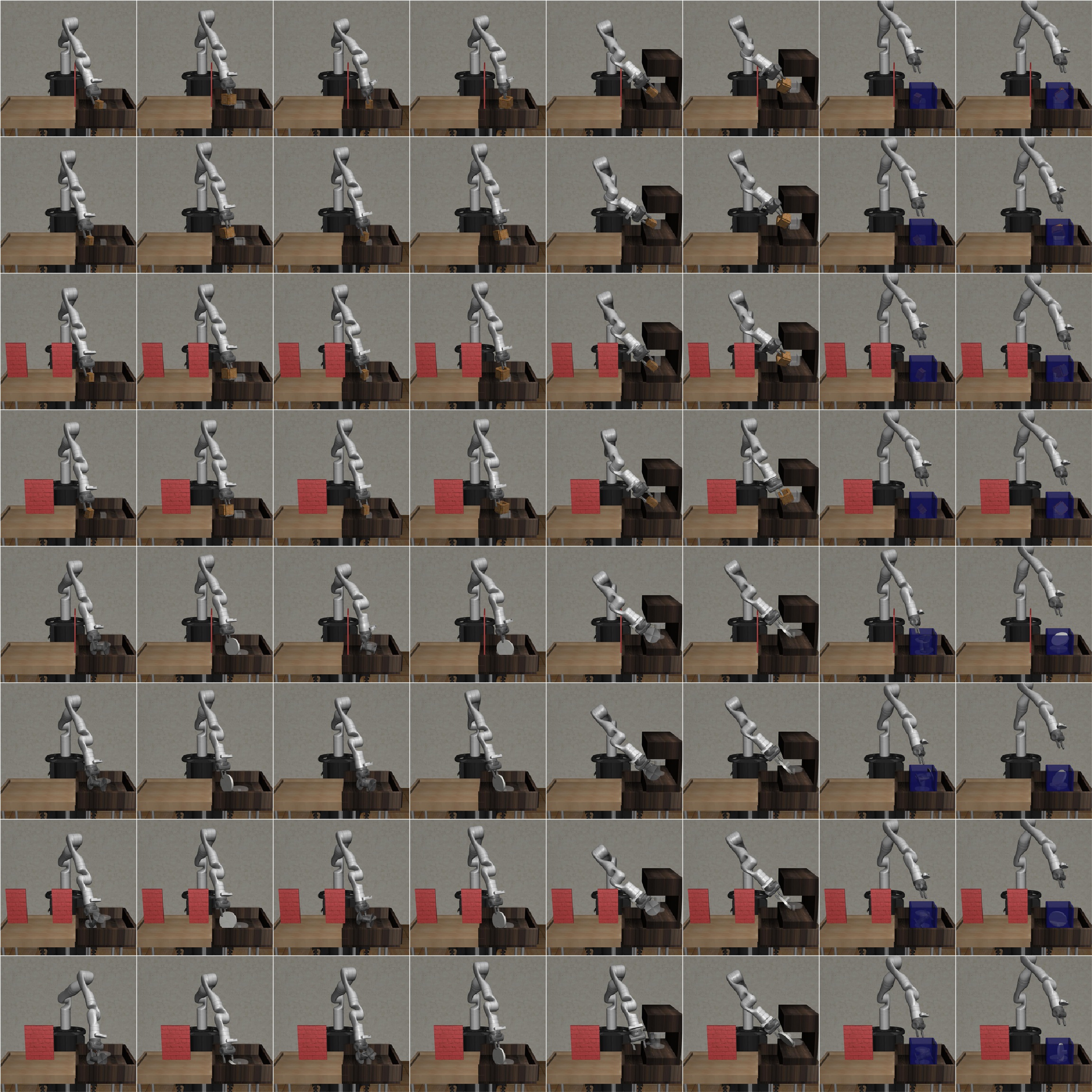}
    \caption[Visualization of the $64$ Kinova tasks.]{Visualization of the $64$ \Kinova{} tasks.}
    \label{fig:kinova_viz}
\end{figure}
\end{appendixf}

\singlespacing
\newenvironment{bibliof}{}{}
\titleformat{\chapter}[hang]{\large\center}{\thechapter}{0 pt}{}
\titlespacing*{\chapter}{0pt}{-25 pt}{6 pt} 
\begin{bibliof}

\renewcommand\bibname{BIBLIOGRAPHY}
\addcontentsline{toc}{chapter}{BIBLIOGRAPHY}
\setcitestyle{numbers} 
\bibliographystyle{apalike}
\bibliography{thesis}

\begin{thebibliography}{}

\bibitem[Abel et~al., 2018]{abel2018state}
Abel, D., Arumugam, D., Lehnert, L., and Littman, M. (2018).
\newblock State abstractions for lifelong reinforcement learning.
\newblock In {\em Proceedings of the 35th International Conference on Machine
  Learning (ICML-18)}, pages 10--19.

\bibitem[Achiam, 2018]{SpinningUp2018}
Achiam, J. (2018).
\newblock Spinning up in deep reinforcement learning.
\newblock \url{https://github.com/openai/spinningup}.

\bibitem[Achille et~al., 2018]{achille2018life}
Achille, A., Eccles, T., Matthey, L., Burgess, C., Watters, N., Lerchner, A.,
  and Higgins, I. (2018).
\newblock Life-long disentangled representation learning with cross-domain
  latent homologies.
\newblock In {\em Advances in Neural Information Processing Systems 31
  (NeurIPS-18)}, pages 9873--9883.

\bibitem[Adel et~al., 2020]{adel2020continual}
Adel, T., Zhao, H., and Turner, R.~E. (2020).
\newblock Continual learning with adaptive weights ({CLAW}).
\newblock In {\em 8th International Conference on Learning Representations
  (ICLR-20)}.

\bibitem[Agarwala et~al., 2021]{agarwala2021one}
Agarwala, A., Das, A., Juba, B., Panigrahy, R., Sharan, V., Wang, X., and
  Zhang, Q. (2021).
\newblock One network fits all? {M}odular versus monolithic task formulations
  in neural networks.
\newblock In {\em 9th International Conference on Learning Representations
  (ICLR-21)}.

\bibitem[Ahmed et~al., 2021]{ahmed2021causalworld}
Ahmed, O., Tr{\"a}uble, F., Goyal, A., Neitz, A., Wuthrich, M., Bengio, Y.,
  Sch{\"o}lkopf, B., and Bauer, S. (2021).
\newblock Causal{W}orld: A robotic manipulation benchmark for causal structure
  and transfer learning.
\newblock In {\em 9th International Conference on Learning Representations
  (ICLR-21)}.

\bibitem[Ahn et~al., 2019]{ahn2019uncertainty}
Ahn, H., Cha, S., Lee, D., and Moon, T. (2019).
\newblock Uncertainty-based continual learning with adaptive regularization.
\newblock In {\em Advances in Neural Information Processing Systems 32
  (NeurIPS-19)}.

\bibitem[Aksan et~al., 2020]{aksan2020cose}
Aksan, E., Deselaers, T., Tagliasacchi, A., and Hilliges, O. (2020).
\newblock {CoSE}: Compositional stroke embeddings.
\newblock In {\em Advances in Neural Information Processing Systems 33
  (NeurIPS-20)}, pages 10041--10052.

\bibitem[Akula et~al., 2021]{akula2021robust}
Akula, A., Jampani, V., Changpinyo, S., and Zhu, S.-C. (2021).
\newblock Robust visual reasoning via language guided neural module networks.
\newblock In {\em Advances in Neural Information Processing Systems 34
  (NeurIPS-21)}, pages 11041--11053.

\bibitem[Aky{\"u}rek et~al., 2021]{akyurek2021learning}
Aky{\"u}rek, E., Aky{\"u}rek, A.~F., and Andreas, J. (2021).
\newblock Learning to recombine and resample data for compositional
  generalization.
\newblock In {\em 9th International Conference on Learning Representations
  (ICLR-21)}.

\bibitem[Al-Shedivat et~al., 2018]{al-shedivat2018continuous}
Al-Shedivat, M., Bansal, T., Burda, Y., Sutskever, I., Mordatch, I., and
  Abbeel, P. (2018).
\newblock Continuous adaptation via meta-learning in nonstationary and
  competitive environments.
\newblock In {\em 6th International Conference on Learning Representations
  (ICLR-18)}.

\bibitem[Alet et~al., 2018]{alet2018modular}
Alet, F., Lozano-Perez, T., and Kaelbling, L.~P. (2018).
\newblock Modular meta-learning.
\newblock In {\em Proceedings of the 2nd Conference on Robot Learning
  (CoRL-18)}, pages 856--868.

\bibitem[Alet et~al., 2019]{alet2019neural}
Alet, F., Weng, E., Lozano-P{\'e}rez, T., and Kaelbling, L.~P. (2019).
\newblock Neural relational inference with fast modular meta-learning.
\newblock In {\em Advances in Neural Information Processing Systems 32
  (NeurIPS-19)}.

\bibitem[Aljundi et~al., 2019a]{aljundi2019online}
Aljundi, R., Belilovsky, E., Tuytelaars, T., Charlin, L., Caccia, M., Lin, M.,
  and Page-Caccia, L. (2019a).
\newblock Online continual learning with maximal interfered retrieval.
\newblock In {\em Advances in Neural Information Processing Systems 32
  (NeurIPS-19)}.

\bibitem[Aljundi et~al., 2017]{aljundi2017expert}
Aljundi, R., Chakravarty, P., and Tuytelaars, T. (2017).
\newblock Expert gate: Lifelong learning with a network of experts.
\newblock In {\em Proceedings of the 2017 IEEE Conference on Computer Vision
  and Pattern Recognition (CVPR-17)}, pages 3366--3375.

\bibitem[Aljundi et~al., 2019b]{aljundi2019gradient}
Aljundi, R., Lin, M., Goujaud, B., and Bengio, Y. (2019b).
\newblock Gradient based sample selection for online continual learning.
\newblock In {\em Advances in Neural Information Processing Systems 32
  (NeurIPS-19)}.

\bibitem[Andreas, 2019]{andreas2019measuring}
Andreas, J. (2019).
\newblock Measuring compositionality in representation learning.
\newblock In {\em 7th International Conference on Learning Representations
  (ICLR-19)}.

\bibitem[Andreas et~al., 2016]{andreas2016neural}
Andreas, J., Rohrbach, M., Darrell, T., and Klein, D. (2016).
\newblock Neural module networks.
\newblock In {\em Proceedings of the 2016 IEEE Conference on Computer Vision
  and Pattern Recognition (CVPR-16)}, pages 39--48.

\bibitem[Arad~Hudson and Zitnick, 2021]{arad2021compositional}
Arad~Hudson, D. and Zitnick, L. (2021).
\newblock Compositional transformers for scene generation.
\newblock In {\em Advances in Neural Information Processing Systems 34
  (NeurIPS-21)}, pages 9506--9520.

\bibitem[Atzmon et~al., 2020]{atzmon2020causal}
Atzmon, Y., Kreuk, F., Shalit, U., and Chechik, G. (2020).
\newblock A causal view of compositional zero-shot recognition.
\newblock In {\em Advances in Neural Information Processing Systems 33
  (NeurIPS-20)}, pages 1462--1473.

\bibitem[Ayub and Wagner, 2021]{ayub2021eec}
Ayub, A. and Wagner, A. (2021).
\newblock {EEC}: Learning to encode and regenerate images for continual
  learning.
\newblock In {\em 9th International Conference on Learning Representations
  (ICLR-21)}.

\bibitem[Bacon et~al., 2017]{bacon2017option}
Bacon, P.-L., Harb, J., and Precup, D. (2017).
\newblock The option-critic architecture.
\newblock In {\em Proceedings of the Thirty-First AAAI Conference on Artificial
  Intelligence (AAAI-17)}, pages 1726--1734.

\bibitem[Bahdanau et~al., 2018]{bahdanau2018systematic}
Bahdanau, D., Murty, S., Noukhovitch, M., Nguyen, T.~H., de~Vries, H., and
  Courville, A. (2018).
\newblock Systematic generalization: What is required and can it be learned?
\newblock In {\em 6th International Conference on Learning Representations
  (ICLR-18)}.

\bibitem[Baklanov, 2006]{baklanov2006strong}
Baklanov, E.~A. (2006).
\newblock The strong law of large numbers for {L}-statistics with dependent
  data.
\newblock {\em Siberian Mathematical Journal}, 47(6):975--979.

\bibitem[Banayeeanzade et~al., 2021]{banayeeanzade2021generative}
Banayeeanzade, M., Mirzaiezadeh, R., Hasani, H., and Soleymani, M. (2021).
\newblock Generative vs.~discriminative: Rethinking the meta-continual
  learning.
\newblock In {\em Advances in Neural Information Processing Systems 34
  (NeurIPS-21)}, pages 21592--21604.

\bibitem[Barreto et~al., 2018]{barreto2018transfer}
Barreto, A., Borsa, D., Quan, J., Schaul, T., Silver, D., Hessel, M.,
  Mankowitz, D., Zidek, A., and Munos, R. (2018).
\newblock Transfer in deep reinforcement learning using successor features and
  generalised policy improvement.
\newblock In {\em Proceedings of the 35th International Conference on Machine
  Learning (ICML-18)}, pages 501--510.

\bibitem[Beaulieu et~al., 2020]{beaulieu2020learning}
Beaulieu, S., Frati, L., Miconi, T., Lehman, J., Stanley, K.~O., Clune, J., and
  Cheney, N. (2020).
\newblock Learning to continually learn.
\newblock In {\em Proceedings of the 24th European Conference on Artificial
  Intelligence (ECAI-20)}, pages 992--1001.

\bibitem[Bellemare et~al., 2013]{bellemare2013arcade}
Bellemare, M.~G., Naddaf, Y., Veness, J., and Bowling, M. (2013).
\newblock The arcade learning environment: An evaluation platform for general
  agents.
\newblock {\em Journal of Artificial Intelligence Research (JAIR)},
  47:253--279.

\bibitem[Benjamin et~al., 2019]{benjamin2018measuring}
Benjamin, A., Rolnick, D., and Kording, K. (2019).
\newblock Measuring and regularizing networks in function space.
\newblock In {\em 7th International Conference on Learning Representations
  (ICLR-19)}.

\bibitem[Billingsley, 1968]{billingsley1968convergence}
Billingsley, P. (1968).
\newblock {\em Convergence of probability measures}.
\newblock John Wiley \& Sons.

\bibitem[Bonnans and Shapiro, 1998]{bonnans1998optimization}
Bonnans, J.~F. and Shapiro, A. (1998).
\newblock Optimization problems with perturbations: A guided tour.
\newblock {\em SIAM Review}, 40(2):228--264.

\bibitem[Borsos et~al., 2020]{borsos2020coresets}
Borsos, Z., Mutny, M., and Krause, A. (2020).
\newblock Coresets via bilevel optimization for continual learning and
  streaming.
\newblock In {\em Advances in Neural Information Processing Systems 33
  (NeurIPS-20)}, pages 14879--14890.

\bibitem[Bo{\v{s}}njak et~al., 2017]{bovsnjak2017programming}
Bo{\v{s}}njak, M., Rockt{\"a}schel, T., Naradowsky, J., and Riedel, S. (2017).
\newblock Programming with a differentiable {F}orth interpreter.
\newblock In {\em Proceedings of the 34th International Conference on Machine
  Learning (ICML-17)}, pages 547--556.

\bibitem[Bottou, 2009]{bottou1998online}
Bottou, L. (2009).
\newblock On-line learning and stochastic approximations.
\newblock In Saad, D., editor, {\em On-line learning in neural networks},
  chapter~2, pages 9--42. Cambridge University Press.

\bibitem[{Bou Ammar} et~al., 2015]{bouammar2015autonomous}
{Bou Ammar}, H., Eaton, E., Luna, J.~M., and Ruvolo, P. (2015).
\newblock Autonomous cross-domain knowledge transfer in lifelong policy
  gradient reinforcement learning.
\newblock In {\em Proceedings of the Twenty-Fourth International Joint
  Conference on Artificial Intelligence (IJCAI-15)}, pages 3345--3351.

\bibitem[{Bou Ammar} et~al., 2014]{bouammar2014online}
{Bou Ammar}, H., Eaton, E., Ruvolo, P., and Taylor, M. (2014).
\newblock Online multi-task learning for policy gradient methods.
\newblock In {\em Proceedings of the 31st International Conference on Machine
  Learning (ICML-14)}, pages 1206--1214.

\bibitem[Brockman et~al., 2016]{brockman2016openai}
Brockman, G., Cheung, V., Pettersson, L., Schneider, J., Schulman, J., Tang,
  J., and Zaremba, W. (2016).
\newblock Open{AI} {G}ym.
\newblock {\em arXiv preprint arXiv:1606.01540}.

\bibitem[Brunskill and Li, 2014]{brunskill2014pac}
Brunskill, E. and Li, L. (2014).
\newblock {PAC}-inspired option discovery in lifelong reinforcement learning.
\newblock In {\em Proceedings of the 31st International Conference on Machine
  Learning (ICML-14)}, pages 316--324.

\bibitem[Bunel et~al., 2018]{bunel2018leveraging}
Bunel, R., Hausknecht, M., Devlin, J., Singh, R., and Kohli, P. (2018).
\newblock Leveraging grammar and reinforcement learning for neural program
  synthesis.
\newblock In {\em 6th International Conference on Learning Representations
  (ICLR-18)}.

\bibitem[Buzzega et~al., 2020]{buzzega2020dark}
Buzzega, P., Boschini, M., Porrello, A., Abati, D., and Calderara, S. (2020).
\newblock Dark experience for general continual learning: A strong, simple
  baseline.
\newblock In {\em Advances in Neural Information Processing Systems 33
  (NeurIPS-20)}, pages 15920--15930.

\bibitem[Bylard et~al., 2021]{bylard2021composable}
Bylard, A., Bonalli, R., and Pavone, M. (2021).
\newblock Composable geometric motion policies using multi-task pullback bundle
  dynamical systems.
\newblock In {\em 2021 IEEE International Conference on Robotics and Automation
  (ICRA-21)}, pages 7464--7470.

\bibitem[Caccia et~al., 2020a]{caccia2020onlinelearned}
Caccia, L., Belilovsky, E., Caccia, M., and Pineau, J. (2020a).
\newblock Online learned continual compression with adaptive quantization
  modules.
\newblock In {\em Proceedings of the 37th International Conference on Machine
  Learning (ICML-20)}, pages 1240--1250.

\bibitem[Caccia et~al., 2020b]{caccia2020onlinefast}
Caccia, M., Rodriguez, P., Ostapenko, O., Normandin, F., Lin, M., Page-Caccia,
  L., Laradji, I.~H., Rish, I., Lacoste, A., V{\'a}zquez, D., and Charlin, L.
  (2020b).
\newblock Online fast adaptation and knowledge accumulation ({OSAKA}): A new
  approach to continual learning.
\newblock In {\em Advances in Neural Information Processing Systems 33
  (NeurIPS-20)}, pages 16532--16545.

\bibitem[Cai et~al., 2017]{cai2017making}
Cai, J., Shin, R., and Song, D. (2017).
\newblock Making neural programming architectures generalize via recursion.
\newblock In {\em 5th International Conference on Learning Representations
  (ICLR-17)}.

\bibitem[Cha et~al., 2021]{cha2021cpr}
Cha, S., Hsu, H., Hwang, T., Calmon, F., and Moon, T. (2021).
\newblock {CPR}: Classifier-projection regularization for continual learning.
\newblock In {\em 9th International Conference on Learning Representations
  (ICLR-21)}.

\bibitem[Chang et~al., 2019]{chang2018automatically}
Chang, M., Gupta, A., Levine, S., and Griffiths, T.~L. (2019).
\newblock Automatically composing representation transformations as a means for
  generalization.
\newblock In {\em 7th International Conference on Learning Representations
  (ICLR-19)}.

\bibitem[Chang et~al., 2021]{chang2021modularity}
Chang, M., Kaushik, S., Levine, S., and Griffiths, T. (2021).
\newblock Modularity in reinforcement learning via algorithmic independence in
  credit assignment.
\newblock In {\em Proceedings of the 38th International Conference on Machine
  Learning (ICML-21)}, pages 1452--1462.

\bibitem[Chaudhry et~al., 2018]{chaudhry2018riemannian}
Chaudhry, A., Dokania, P.~K., Ajanthan, T., and Torr, P.~H. (2018).
\newblock Riemannian walk for incremental learning: Understanding forgetting
  and intransigence.
\newblock In {\em Proceedings of the European Conference on Computer Vision
  (ECCV-18)}, pages 532--547.

\bibitem[Chaudhry et~al., 2020]{chaudhry2020continual}
Chaudhry, A., Khan, N., Dokania, P., and Torr, P. (2020).
\newblock Continual learning in low-rank orthogonal subspaces.
\newblock In {\em Advances in Neural Information Processing Systems 33
  (NeurIPS-20)}, pages 9900--9911.

\bibitem[Chaudhry et~al., 2019a]{chaudhry2018efficient}
Chaudhry, A., Ranzato, M., Rohrbach, M., and Elhoseiny, M. (2019a).
\newblock Efficient lifelong learning with {A-GEM}.
\newblock In {\em 7th International Conference on Learning Representations
  (ICLR-19)}.

\bibitem[Chaudhry et~al., 2019b]{chaudhry2019tiny}
Chaudhry, A., Rohrbach, M., Elhoseiny, M., Ajanthan, T., Dokania, P.~K., Torr,
  P.~H., and Ranzato, M. (2019b).
\newblock On tiny episodic memories in continual learning.
\newblock {\em arXiv preprint arXiv:1902.10486}.

\bibitem[Chen et~al., 2020a]{chen2020mitigating}
Chen, H.-J., Cheng, A.-C., Juan, D.-C., Wei, W., and Sun, M. (2020a).
\newblock Mitigating forgetting in online continual learning via instance-aware
  parameterization.
\newblock In {\em Advances in Neural Information Processing Systems 33
  (NeurIPS-20)}, pages 17466--17477.

\bibitem[Chen et~al., 2021]{chen2021long}
Chen, T., Zhang, Z., Liu, S., Chang, S., and Wang, Z. (2021).
\newblock Long live the lottery: The existence of winning tickets in lifelong
  learning.
\newblock In {\em 9th International Conference on Learning Representations
  (ICLR-21)}.

\bibitem[Chen et~al., 2020b]{chen2020compositional}
Chen, X., Liang, C., Yu, A.~W., Song, D., and Zhou, D. (2020b).
\newblock Compositional generalization via neural-symbolic stack machines.
\newblock In {\em Advances in Neural Information Processing Systems 33
  (NeurIPS-20)}, pages 1690--1701.

\bibitem[Chen et~al., 2020c]{chen2020modular}
Chen, Y., Friesen, A.~L., Behbahani, F., Doucet, A., Budden, D., Hoffman, M.,
  and de~Freitas, N. (2020c).
\newblock Modular meta-learning with shrinkage.
\newblock In {\em Advances in Neural Information Processing Systems 33
  (NeurIPS-20)}, pages 2858--2869.

\bibitem[Chen and Liu, 2018]{chen2018lifelong}
Chen, Z. and Liu, B. (2018).
\newblock Lifelong machine learning.
\newblock In {\em Synthesis Lectures on Artificial Intelligence and Machine
  Learning \#38}. Morgan \& Claypool Publishers.

\bibitem[Cheng et~al., 2021]{cheng2021rmpflow}
Cheng, C.-A., Mukadam, M., Issac, J., Birchfield, S., Fox, D., Boots, B., and
  Ratliff, N. (2021).
\newblock {RMP\textit{flow}}: A geometric framework for generation of multitask
  motion policies.
\newblock {\em IEEE Transactions on Automation Science and Engineering},
  18(3):968--987.

\bibitem[Chevalier-Boisvert et~al., 2019]{chevalier-boisvert2018babyai}
Chevalier-Boisvert, M., Bahdanau, D., Lahlou, S., Willems, L., Saharia, C.,
  Nguyen, T.~H., and Bengio, Y. (2019).
\newblock Baby{AI}: First steps towards grounded language learning with a human
  in the loop.
\newblock In {\em 7th International Conference on Learning Representations
  (ICLR-19)}.

\bibitem[Chevalier-Boisvert et~al., 2018]{gym_minigrid}
Chevalier-Boisvert, M., Willems, L., and Pal, S. (2018).
\newblock Minimalistic gridworld environment for {OpenAI Gym}.
\newblock \url{https://github.com/maximecb/gym-minigrid}.

\bibitem[Chrysakis and Moens, 2020]{chrysakis2020online}
Chrysakis, A. and Moens, M.-F. (2020).
\newblock Online continual learning from imbalanced data.
\newblock In {\em Proceedings of the 37th International Conference on Machine
  Learning (ICML-20)}, pages 1952--1961.

\bibitem[Clavera et~al., 2019]{clavera2018learning}
Clavera, I., Nagabandi, A., Liu, S., Fearing, R.~S., Abbeel, P., Levine, S.,
  and Finn, C. (2019).
\newblock Learning to adapt in dynamic, real-world environments through
  meta-reinforcement learning.
\newblock In {\em 7th International Conference on Learning Representations
  (ICLR-19)}.

\bibitem[Colas et~al., 2019]{colas2019curious}
Colas, C., Fournier, P., Chetouani, M., Sigaud, O., and Oudeyer, P.-Y. (2019).
\newblock {CURIOUS}: Intrinsically motivated modular multi-goal reinforcement
  learning.
\newblock In {\em Proceedings of the 36th International Conference on Machine
  Learning (ICML-19)}, pages 1331--1340.

\bibitem[Csord{\'a}s et~al., 2021]{csordas2021are}
Csord{\'a}s, R., van Steenkiste, S., and Schmidhuber, J. (2021).
\newblock Are neural nets modular? {I}nspecting functional modularity through
  differentiable weight masks.
\newblock In {\em 9th International Conference on Learning Representations
  (ICLR-21)}.

\bibitem[D'Amario et~al., 2021]{damario2021how}
D'Amario, V., Sasaki, T., and Boix, X. (2021).
\newblock How modular should neural module networks be for systematic
  generalization?
\newblock In {\em Advances in Neural Information Processing Systems 34
  (NeurIPS-21)}, pages 23374--23385.

\bibitem[Dayan and Hinton, 1993]{dayan1993feudal}
Dayan, P. and Hinton, G.~E. (1993).
\newblock Feudal reinforcement learning.
\newblock In {\em Advances in Neural Information Processing Systems 6
  (NIPS-93)}, pages 271--278.

\bibitem[de~Masson~d'Autume et~al., 2019]{demasson2019episodic}
de~Masson~d'Autume, C., Ruder, S., Kong, L., and Yogatama, D. (2019).
\newblock Episodic memory in lifelong language learning.
\newblock In {\em Advances in Neural Information Processing Systems 32
  (NeurIPS-19)}.

\bibitem[Del~Chiaro et~al., 2020]{delchiaro2020ratt}
Del~Chiaro, R., Twardowski, B., Bagdanov, A., and van~de Weijer, J. (2020).
\newblock {RATT}: Recurrent attention to transient tasks for continual image
  captioning.
\newblock In {\em Advances in Neural Information Processing Systems 33
  (NeurIPS-20)}, pages 16736--16748.

\bibitem[Deng et~al., 2021]{deng2021flattening}
Deng, D., Chen, G., Hao, J., Wang, Q., and Heng, P.-A. (2021).
\newblock Flattening sharpness for dynamic gradient projection memory benefits
  in continual learning.
\newblock In {\em Advances in Neural Information Processing Systems 34
  (NeurIPS-21)}, pages 18710--18721.

\bibitem[Deng et~al., 2009]{deng2009imagenet}
Deng, J., Dong, W., Socher, R., Li, L.-J., Li, K., and Fei-Fei, L. (2009).
\newblock Image{N}et: A large-scale hierarchical image database.
\newblock In {\em Proceedings of the 2009 IEEE Conference on Computer Vision
  and Pattern Recognition (CVPR-09)}, pages 248--255.

\bibitem[Derakhshani et~al., 2021]{derakhshani2021kernel}
Derakhshani, M.~M., Zhen, X., Shao, L., and Snoek, C. (2021).
\newblock Kernel continual learning.
\newblock In {\em Proceedings of the 38th International Conference on Machine
  Learning (ICML-21)}, pages 2621--2631.

\bibitem[Devin et~al., 2019]{devin2019compositional}
Devin, C., Geng, D., Abbeel, P., Darrell, T., and Levine, S. (2019).
\newblock Compositional plan vectors.
\newblock In {\em Advances in Neural Information Processing Systems 32
  (NeurIPS-19)}.

\bibitem[Devin et~al., 2017]{devin2017learning}
Devin, C., Gupta, A., Darrell, T., Abbeel, P., and Levine, S. (2017).
\newblock Learning modular neural network policies for multi-task and
  multi-robot transfer.
\newblock In {\em Proceedings of the 2017 IEEE International Conference on
  Robotics and Automation (ICRA-17)}, pages 2169--2176.

\bibitem[Devlin et~al., 2019]{devlin2019bert}
Devlin, J., Chang, M.-W., Lee, K., and Toutanova, K. (2019).
\newblock {BERT}: Pre-training of deep bidirectional transformers for language
  understanding.
\newblock In {\em Proceedings of the 2019 Conference of the North American
  Chapter of the Association for Computational Linguistics: Human Language
  Technologies (NAACL-HLT-19)}, pages 4171--4186.

\bibitem[Dietterich, 2000]{dietterich2000hierarchical}
Dietterich, T.~G. (2000).
\newblock Hierarchical reinforcement learning with the {MAXQ} value function
  decomposition.
\newblock {\em Journal of Artificial Intelligence Research (JAIR)},
  13:227--303.

\bibitem[Doyle, 1979]{doyle1979truth}
Doyle, J. (1979).
\newblock A truth maintenance system.
\newblock {\em Artificial Intelligence}, 12(3):231--272.

\bibitem[Du et~al., 2020]{du2020compositional}
Du, Y., Li, S., and Mordatch, I. (2020).
\newblock Compositional visual generation with energy based models.
\newblock In {\em Advances in Neural Information Processing Systems 33
  (NeurIPS-20)}, pages 6637--6647.

\bibitem[Duncker et~al., 2020]{duncker2020organizing}
Duncker, L., Driscoll, L., Shenoy, K.~V., Sahani, M., and Sussillo, D. (2020).
\newblock Organizing recurrent network dynamics by task-computation to enable
  continual learning.
\newblock In {\em Advances in Neural Information Processing Systems 33
  (NeurIPS-20)}, pages 14387--14397.

\bibitem[Egorov et~al., 2021]{egorov2021boovae}
Egorov, E., Kuzina, A., and Burnaev, E. (2021).
\newblock {BooVAE}: Boosting approach for continual learning of {VAE}.
\newblock In {\em Advances in Neural Information Processing Systems 34
  (NeurIPS-21)}, pages 17889--17901.

\bibitem[Ehret et~al., 2021]{ehret2021continual}
Ehret, B., Henning, C., Cervera, M., Meulemans, A., von Oswald, J., and Grewe,
  B.~F. (2021).
\newblock Continual learning in recurrent neural networks.
\newblock In {\em 9th International Conference on Learning Representations
  (ICLR-21)}.

\bibitem[Fernando et~al., 2017]{fernando2017pathnet}
Fernando, C., Banarse, D., Blundell, C., Zwols, Y., Ha, D., Rusu, A.~A.,
  Pritzel, A., and Wierstra, D. (2017).
\newblock Path{N}et: Evolution channels gradient descent in super neural
  networks.
\newblock {\em arXiv preprint arXiv:1701.08734}.

\bibitem[Fisk, 1965]{fisk1965quasi}
Fisk, D. (1965).
\newblock Quasi-martingales.
\newblock {\em Transactions of the American Mathematical Society},
  120(3):369--389.

\bibitem[Frankle and Carbin, 2019]{frankle2018lottery}
Frankle, J. and Carbin, M. (2019).
\newblock The lottery ticket hypothesis: Finding sparse, trainable neural
  networks.
\newblock In {\em 7th International Conference on Learning Representations
  (ICLR-19)}.

\bibitem[Fuchs, 2005]{fuchs2005recovery}
Fuchs, J.-J. (2005).
\newblock Recovery of exact sparse representations in the presence of bounded
  noise.
\newblock {\em IEEE Transactions on Information Theory}, 51(10):3601--3608.

\bibitem[Fujimoto et~al., 2019a]{fujimoto2019benchmarking}
Fujimoto, S., Conti, E., Ghavamzadeh, M., and Pineau, J. (2019a).
\newblock Benchmarking batch deep reinforcement learning algorithms.
\newblock {\em arXiv preprint arXiv:1910.01708}.

\bibitem[Fujimoto et~al., 2019b]{fujimoto2019off}
Fujimoto, S., Meger, D., and Precup, D. (2019b).
\newblock Off-policy deep reinforcement learning without exploration.
\newblock In {\em Proceedings of the 36th International Conference on Machine
  Learning (ICML-19)}, pages 2052--2062.

\bibitem[Fujimoto et~al., 2018]{fujimoto2018functionapprox}
Fujimoto, S., van Hoof, H., and Meger, D. (2018).
\newblock Addressing function approximation error in actor-critic methods.
\newblock In {\em Proceedings of the 35th International Conference on Machine
  Learning (ICML-18)}, pages 1587--1596.

\bibitem[Garcia and Thomas, 2019]{garcia2019meta}
Garcia, F. and Thomas, P.~S. (2019).
\newblock A meta-{MDP} approach to exploration for lifelong reinforcement
  learning.
\newblock In {\em Advances in Neural Information Processing Systems 32
  (NeurIPS-19)}, pages 5692--5701.

\bibitem[Gatys et~al., 2016]{gatys2016image}
Gatys, L.~A., Ecker, A.~S., and Bethge, M. (2016).
\newblock Image style transfer using convolutional neural networks.
\newblock In {\em Proceedings of the 2016 IEEE Conference on Computer Vision
  and Pattern Recognition (CVPR-16)}, pages 2414--2423.

\bibitem[Gaunt et~al., 2017]{gaunt2017differentiable}
Gaunt, A.~L., Brockschmidt, M., Kushman, N., and Tarlow, D. (2017).
\newblock Differentiable programs with neural libraries.
\newblock In {\em Proceedings of the 34th International Conference on Machine
  Learning (ICML-17)}, pages 1213--1222.

\bibitem[Ghazi et~al., 2019]{ghazi2019recursive}
Ghazi, B., Panigrahy, R., and Wang, J. (2019).
\newblock Recursive sketches for modular deep learning.
\newblock In {\em Proceedings of the 36th International Conference on Machine
  Learning (ICML-19)}, pages 2211--2220.

\bibitem[Gomez et~al., 2019]{gomez2019learning}
Gomez, A.~N., Zhang, I., Kamalakara, S.~R., Madaan, D., Swersky, K., Gal, Y.,
  and Hinton, G.~E. (2019).
\newblock Learning sparse networks using targeted dropout.
\newblock {\em arXiv preprint arXiv:1905.13678}.

\bibitem[Gordon et~al., 2020]{gordon2020permutation}
Gordon, J., Lopez-Paz, D., Baroni, M., and Bouchacourt, D. (2020).
\newblock Permutation equivariant models for compositional generalization in
  language.
\newblock In {\em 8th International Conference on Learning Representations
  (ICLR-20)}.

\bibitem[Goyal et~al., 2021]{goyal2021recurrent}
Goyal, A., Lamb, A., Hoffmann, J., Sodhani, S., Levine, S., Bengio, Y., and
  Sch{\"o}lkopf, B. (2021).
\newblock Recurrent independent mechanisms.
\newblock In {\em 9th International Conference on Learning Representations
  (ICLR-21)}.

\bibitem[Gu et~al., 2017]{gu2017deep}
Gu, S., Holly, E., Lillicrap, T., and Levine, S. (2017).
\newblock Deep reinforcement learning for robotic manipulation with
  asynchronous off-policy updates.
\newblock In {\em Proceedings of the 2017 IEEE international conference on
  robotics and automation (ICRA-17)}, pages 3389--3396.

\bibitem[Guo et~al., 2020a]{guo2020hierarchical}
Guo, Y., Lin, Z., Lou, J.-G., and Zhang, D. (2020a).
\newblock Hierarchical poset decoding for compositional generalization in
  language.
\newblock In {\em Advances in Neural Information Processing Systems 33
  (NeurIPS-20)}, pages 6913--6924.

\bibitem[Guo et~al., 2020b]{guo2020improved}
Guo, Y., Liu, M., Yang, T., and Rosing, T. (2020b).
\newblock Improved schemes for episodic memory-based lifelong learning.
\newblock In {\em Advances in Neural Information Processing Systems 33
  (NeurIPS-20)}, pages 1023--1035.

\bibitem[Gupta et~al., 2020a]{gupta2020lookahead}
Gupta, G., Yadav, K., and Paull, L. (2020a).
\newblock Look-ahead meta learning for continual learning.
\newblock In {\em Advances in Neural Information Processing Systems 33
  (NeurIPS-20)}, pages 11588--11598.

\bibitem[Gupta et~al., 2020b]{gupta2020neuralmodule}
Gupta, N., Lin, K., Roth, D., Singh, S., and Gardner, M. (2020b).
\newblock Neural module networks for reasoning over text.
\newblock In {\em 8th International Conference on Learning Representations
  (ICLR-20)}.

\bibitem[Gupta et~al., 2020c]{gupta2020neuraltopic}
Gupta, P., Chaudhary, Y., Runkler, T., and Schuetze, H. (2020c).
\newblock Neural topic modeling with continual lifelong learning.
\newblock In {\em Proceedings of the 37th International Conference on Machine
  Learning (ICML-20)}, pages 3907--3917.

\bibitem[Gur et~al., 2021]{gur2021environment}
Gur, I., Jaques, N., Miao, Y., Choi, J., Tiwari, M., Lee, H., and Faust, A.
  (2021).
\newblock Environment generation for zero-shot compositional reinforcement
  learning.
\newblock In {\em Advances in Neural Information Processing Systems 34
  (NeurIPS-21)}, pages 4157--4169.

\bibitem[Haarnoja et~al., 2018]{haarnoja2018composable}
Haarnoja, T., Pong, V., Zhou, A., Dalal, M., Abbeel, P., and Levine, S. (2018).
\newblock Composable deep reinforcement learning for robotic manipulation.
\newblock In {\em Proceedings of the 2018 IEEE International Conference on
  Robotics and Automation (ICRA-18)}, pages 6244--6251.

\bibitem[{Henderson} et~al., 2017]{henderson2017multitask}
{Henderson}, P., {Chang}, W.-D., {Shkurti}, F., {Hansen}, J., {Meger}, D., and
  {Dudek}, G. (2017).
\newblock Benchmark environments for multitask learning in continuous domains.
\newblock {\em ICML Lifelong Learning: A Reinforcement Learning Approach
  Workshop}.

\bibitem[Henning et~al., 2021]{henning2021posterior}
Henning, C., Cervera, M., D'Angelo, F., von Oswald, J., Traber, R., Ehret, B.,
  Kobayashi, S., Grewe, B.~F., and Sacramento, J. (2021).
\newblock Posterior meta-replay for continual learning.
\newblock In {\em Advances in Neural Information Processing Systems 34
  (NeurIPS-21)}, pages 14135--14149.

\bibitem[Hinton et~al., 2012]{hinton2012improving}
Hinton, G.~E., Srivastava, N., Krizhevsky, A., Sutskever, I., and
  Salakhutdinov, R.~R. (2012).
\newblock Improving neural networks by preventing co-adaptation of feature
  detectors.
\newblock {\em arXiv preprint arXiv:1207.0580}.

\bibitem[Hu et~al., 2017]{hu2017learning}
Hu, R., Andreas, J., Rohrbach, M., Darrell, T., and Saenko, K. (2017).
\newblock Learning to reason: End-to-end module networks for visual question
  answering.
\newblock In {\em Proceedings of the 2017 IEEE International Conference on
  Computer Vision (ICCV-17)}, pages 804--813.

\bibitem[Hu et~al., 2019]{hu2019overcoming}
Hu, W., Lin, Z., Liu, B., Tao, C., Tao, Z., Ma, J., Zhao, D., and Yan, R.
  (2019).
\newblock Overcoming catastrophic forgetting via model adaptation.
\newblock In {\em 7th International Conference on Learning Representations
  (ICLR-19)}.

\bibitem[Huang et~al., 2020]{huang2020one}
Huang, W., Mordatch, I., and Pathak, D. (2020).
\newblock One policy to control them all: Shared modular policies for
  agent-agnostic control.
\newblock In {\em Proceedings of the 37th International Conference on Machine
  Learning (ICML-20)}, pages 4455--4464.

\bibitem[Hung et~al., 2019]{hung2019compacting}
Hung, C.-Y., Tu, C.-H., Wu, C.-E., Chen, C.-H., Chan, Y.-M., and Chen, C.-S.
  (2019).
\newblock Compacting, picking and growing for unforgetting continual learning.
\newblock In {\em Advances in Neural Information Processing Systems 32
  (NeurIPS-19)}.

\bibitem[Hurtado et~al., 2021]{hurtado2021optimizing}
Hurtado, J., Raymond, A., and Soto, A. (2021).
\newblock Optimizing reusable knowledge for continual learning via
  metalearning.
\newblock In {\em Advances in Neural Information Processing Systems 34
  (NeurIPS-21)}, pages 14150--14162.

\bibitem[Husz{\'a}r, 2018]{huszar2018note}
Husz{\'a}r, F. (2018).
\newblock Note on the quadratic penalties in elastic weight consolidation.
\newblock {\em Proceedings of the National Academy of Sciences (PNAS)}, pages
  E2496--E2497.

\bibitem[Huynh and Elhamifar, 2020]{huynh2020compositional}
Huynh, D. and Elhamifar, E. (2020).
\newblock Compositional zero-shot learning via fine-grained dense feature
  composition.
\newblock In {\em Advances in Neural Information Processing Systems 33
  (NeurIPS-20)}, pages 19849--19860.

\bibitem[Isele and Cosgun, 2018]{isele2018selective}
Isele, D. and Cosgun, A. (2018).
\newblock Selective experience replay for lifelong learning.
\newblock In {\em Proceedings of the Thirty-Second AAAI Conference on
  Artificial Intelligence (AAAI-18)}, pages 3302--3309.

\bibitem[Isele et~al., 2016]{isele2016using}
Isele, D., Rostami, M., and Eaton, E. (2016).
\newblock Using task features for zero-shot knowledge transfer in lifelong
  learning.
\newblock In {\em Proceedings of the Twenty-Fifth International Joint
  Conference on Artificial Intelligence (IJCAI-16)}, pages 1620--1626.

\bibitem[James et~al., 2020]{james2020rlbench}
James, S., Ma, Z., Arrojo, D.~R., and Davison, A.~J. (2020).
\newblock {RLB}ench: The robot learning benchmark \& learning environment.
\newblock {\em IEEE Robotics and Automation Letters}, 5(2):3019--3026.

\bibitem[Javed and White, 2019]{javed2019metalearning}
Javed, K. and White, M. (2019).
\newblock Meta-learning representations for continual learning.
\newblock In {\em Advances in Neural Information Processing Systems 32
  (NeurIPS-19)}.

\bibitem[Jerfel et~al., 2019]{jerfel2019reconciling}
Jerfel, G., Grant, E., Griffiths, T., and Heller, K.~A. (2019).
\newblock Reconciling meta-learning and continual learning with online mixtures
  of tasks.
\newblock In {\em Advances in Neural Information Processing Systems 32
  (NeurIPS-19)}.

\bibitem[Jin et~al., 2021]{jin2021gradientbased}
Jin, X., Sadhu, A., Du, J., and Ren, X. (2021).
\newblock Gradient-based editing of memory examples for online task-free
  continual learning.
\newblock In {\em Advances in Neural Information Processing Systems 34
  (NeurIPS-21)}, pages 29193--29205.

\bibitem[Johnson et~al., 2017]{johnson2017inferring}
Johnson, J., Hariharan, B., van~der Maaten, L., Hoffman, J., Fei-Fei, L.,
  Lawrence~Zitnick, C., and Girshick, R. (2017).
\newblock Inferring and executing programs for visual reasoning.
\newblock In {\em Proceedings of the 2017 IEEE International Conference on
  Computer Vision (ICCV-17)}, pages 2989--2998.

\bibitem[Joseph and Balasubramanian, 2020]{joseph2020meta}
Joseph, K.~J. and Balasubramanian, V.~N. (2020).
\newblock Meta-consolidation for continual learning.
\newblock In {\em Advances in Neural Information Processing Systems 33
  (NeurIPS-20)}, pages 14374--14386.

\bibitem[Jothimurugan et~al., 2019]{jothimurugan2019composable}
Jothimurugan, K., Alur, R., and Bastani, O. (2019).
\newblock A composable specification language for reinforcement learning tasks.
\newblock In {\em Advances in Neural Information Processing Systems 32
  (NeurIPS-19)}.

\bibitem[Jothimurugan et~al., 2021]{jothimurugan2021compositional}
Jothimurugan, K., Bansal, S., Bastani, O., and Alur, R. (2021).
\newblock Compositional reinforcement learning from logical specifications.
\newblock In {\em Advances in Neural Information Processing Systems 34
  (NeurIPS-21)}, pages 10026--10039.

\bibitem[Jung et~al., 2020]{jung2020continual}
Jung, S., Ahn, H., Cha, S., and Moon, T. (2020).
\newblock Continual learning with node-importance based adaptive group sparse
  regularization.
\newblock In {\em Advances in Neural Information Processing Systems 33
  (NeurIPS-20)}, pages 3647--3658.

\bibitem[Kakade, 2002]{kakade2002natural}
Kakade, S.~M. (2002).
\newblock A natural policy gradient.
\newblock In {\em Advances in Neural Information Processing Systems 15
  (NIPS-02)}, pages 1531--1538.

\bibitem[Kao et~al., 2021]{kao2021natural}
Kao, T.-C., Jensen, K., van~de Ven, G.~M., Bernacchia, A., and Hennequin, G.
  (2021).
\newblock Natural continual learning: Success is a journey, not (just) a
  destination.
\newblock In {\em Advances in Neural Information Processing Systems 34
  (NeurIPS-21)}, pages 28067--28079.

\bibitem[Kaplanis et~al., 2019]{kaplanis2019policy}
Kaplanis, C., Shanahan, M., and Clopath, C. (2019).
\newblock Policy consolidation for continual reinforcement learning.
\newblock In {\em Proceedings of the 36th International Conference on Machine
  Learning (ICML-19)}, pages 3242--3251.

\bibitem[Kapoor et~al., 2021]{kapoor2021variational}
Kapoor, S., Karaletsos, T., and Bui, T.~D. (2021).
\newblock Variational auto-regressive {G}aussian processes for continual
  learning.
\newblock In {\em Proceedings of the 38th International Conference on Machine
  Learning (ICML-21)}, pages 5290--5300.

\bibitem[Ke et~al., 2020]{ke2020continual}
Ke, Z., Liu, B., and Huang, X. (2020).
\newblock Continual learning of a mixed sequence of similar and dissimilar
  tasks.
\newblock In {\em Advances in Neural Information Processing Systems 33
  (NeurIPS-20)}, pages 18493--18504.

\bibitem[Ke et~al., 2021]{ke2021achieving}
Ke, Z., Liu, B., Ma, N., Xu, H., and Shu, L. (2021).
\newblock Achieving forgetting prevention and knowledge transfer in continual
  learning.
\newblock In {\em Advances in Neural Information Processing Systems 34
  (NeurIPS-21)}, pages 22443--22456.

\bibitem[Keysers et~al., 2020]{keysers2020measuring}
Keysers, D., Sch{\"a}rli, N., Scales, N., Buisman, H., Furrer, D., Kashubin,
  S., Momchev, N., Sinopalnikov, D., Stafiniak, L., Tihon, T., Tsarkov, D.,
  Wang, X., van Zee, M., and Bousquet, O. (2020).
\newblock Measuring compositional generalization: A comprehensive method on
  realistic data.
\newblock In {\em 8th International Conference on Learning Representations
  (ICLR-20)}.

\bibitem[Kim et~al., 2019]{kim2019visual}
Kim, S.~W., Tapaswi, M., and Fidler, S. (2019).
\newblock Visual reasoning by progressive module networks.
\newblock In {\em 7th International Conference on Learning Representations
  (ICLR-19)}.

\bibitem[Kirkpatrick et~al., 2017]{kirkpatrick2017overcoming}
Kirkpatrick, J., Pascanu, R., Rabinowitz, N., Veness, J., Desjardins, G., Rusu,
  A.~A., Milan, K., Quan, J., Ramalho, T., Grabska-Barwinska, A., Hassabis, D.,
  Clopath, C., Kumaran, D., and Hadsell, R. (2017).
\newblock Overcoming catastrophic forgetting in neural networks.
\newblock {\em Proceedings of the National Academy of Sciences (PNAS)},
  114(13):3521--3526.

\bibitem[Kirsch et~al., 2018]{kirsch2018modular}
Kirsch, L., Kunze, J., and Barber, D. (2018).
\newblock Modular networks: Learning to decompose neural computation.
\newblock In {\em Advances in Neural Information Processing Systems 31
  (NeurIPS-18)}, pages 2408--2418.

\bibitem[Knoblauch et~al., 2020]{knoblauch2020optimal}
Knoblauch, J., Husain, H., and Diethe, T. (2020).
\newblock Optimal continual learning has perfect memory and is {NP}-hard.
\newblock In {\em Proceedings of the 37th International Conference on Machine
  Learning (ICML-20)}, pages 5327--5337.

\bibitem[Konda and Borkar, 1999]{konda1999actor}
Konda, V.~R. and Borkar, V.~S. (1999).
\newblock Actor-critic--type learning algorithms for {M}arkov decision
  processes.
\newblock {\em SIAM Journal on control and Optimization}, 38(1):94--123.

\bibitem[Konidaris and Barto, 2009]{konidaris2009skilldiscovery}
Konidaris, G. and Barto, A. (2009).
\newblock Skill discovery in continuous reinforcement learning domains using
  skill chaining.
\newblock In {\em Advances in Neural Information Processing Systems 22
  (NIPS-09)}.

\bibitem[Krizhevsky and Hinton, 2009]{krizhevsky2009learning}
Krizhevsky, A. and Hinton, G. (2009).
\newblock Learning multiple layers of features from tiny images.
\newblock Technical report, University of Toronto.

\bibitem[Kumar et~al., 2021]{kumar2021bayesian}
Kumar, A., Chatterjee, S., and Rai, P. (2021).
\newblock {B}ayesian structural adaptation for continual learning.
\newblock In {\em Proceedings of the 38th International Conference on Machine
  Learning (ICML-21)}, pages 5850--5860.

\bibitem[Kumar et~al., 2019]{kumar2019stabilizing}
Kumar, A., Fu, J., Soh, M., Tucker, G., and Levine, S. (2019).
\newblock Stabilizing off-policy {Q}-learning via bootstrapping error
  reduction.
\newblock In {\em Advances in Neural Information Processing Systems 32
  (NeurIPS-19)}.

\bibitem[Kumar et~al., 2020]{kumar2020understanding}
Kumar, A., Ma, T., and Liang, P. (2020).
\newblock Understanding self-training for gradual domain adaptation.
\newblock In {\em Proceedings of the 37th International Conference on Machine
  Learning (ICML-20)}, pages 5468--5479.

\bibitem[Kurle et~al., 2020]{kurle2020Continual}
Kurle, R., Cseke, B., Klushyn, A., van~der Smagt, P., and G{\"u}nnemann, S.
  (2020).
\newblock Continual learning with {B}ayesian neural networks for non-stationary
  data.
\newblock In {\em 8th International Conference on Learning Representations
  (ICLR-20)}.

\bibitem[Lake and Baroni, 2018]{lake2018generalization}
Lake, B. and Baroni, M. (2018).
\newblock Generalization without systematicity: On the compositional skills of
  sequence-to-sequence recurrent networks.
\newblock In {\em Proceedings of the 35th International Conference on Machine
  Learning (ICML-18)}, pages 2873--2882.

\bibitem[Lake, 2019]{lake2019compositional}
Lake, B.~M. (2019).
\newblock Compositional generalization through meta sequence-to-sequence
  learning.
\newblock In {\em Advances in Neural Information Processing Systems 32
  (NeurIPS-19)}.

\bibitem[Lake et~al., 2015]{lake2015human}
Lake, B.~M., Salakhutdinov, R., and Tenenbaum, J.~B. (2015).
\newblock Human-level concept learning through probabilistic program induction.
\newblock {\em Science}, 350(6266):1332--1338.

\bibitem[Lao et~al., 2020]{lao2020continuous}
Lao, Q., Jiang, X., Havaei, M., and Bengio, Y. (2020).
\newblock Continuous domain adaptation with variational domain-agnostic feature
  replay.
\newblock {\em arXiv preprint arXiv:2003.04382}.

\bibitem[Laroche et~al., 2019]{laroche2019safe}
Laroche, R., Trichelair, P., and Combes, R. T.~D. (2019).
\newblock Safe policy improvement with baseline bootstrapping.
\newblock In {\em Proceedings of the 36th International Conference on Machine
  Learning (ICML-19)}, pages 3652--3661.

\bibitem[Le{C}un et~al., 1998]{gradient1998lecun}
Le{C}un, Y., Bottou, L., Bengio, Y., and Haffner, P. (1998).
\newblock Gradient-based learning applied to document recognition.
\newblock {\em Proceedings of the IEEE}, 86(11):2278--2324.

\bibitem[Lee et~al., 2021a]{lee2021sharing}
Lee, S., Behpour, S., and Eaton, E. (2021a).
\newblock Sharing less is more: Lifelong learning in deep networks with
  selective layer transfer.
\newblock In {\em Proceedings of the 38th International Conference on Machine
  Learning (ICML-21)}, pages 6065--6075.

\bibitem[Lee et~al., 2021b]{lee2021continual}
Lee, S., Goldt, S., and Saxe, A. (2021b).
\newblock Continual learning in the teacher-student setup: Impact of task
  similarity.
\newblock In {\em Proceedings of the 38th International Conference on Machine
  Learning (ICML-21)}, pages 6109--6119.

\bibitem[Lee et~al., 2020]{lee2020neural}
Lee, S., Ha, J., Zhang, D., and Kim, G. (2020).
\newblock A neural {D}irichlet process mixture model for task-free continual
  learning.
\newblock In {\em 8th International Conference on Learning Representations
  (ICLR-20)}.

\bibitem[Lee et~al., 2019a]{lee2019learning}
Lee, S., Stokes, J., and Eaton, E. (2019a).
\newblock Learning shared knowledge for deep lifelong learning using
  deconvolutional networks.
\newblock In {\em Proceedings of the Twenty-Eighth International Joint
  Conference on Artificial Intelligence (IJCAI-19)}, pages 2837--2844.

\bibitem[Lee et~al., 2019b]{lee2019composing}
Lee, Y., Sun, S.-H., Somasundaram, S., Hu, E., and Lim, J.~J. (2019b).
\newblock Composing complex skills by learning transition policies with
  proximity reward induction.
\newblock In {\em 7th International Conference on Learning Representations
  (ICLR-19)}.

\bibitem[Levine et~al., 2020]{levine2020offline}
Levine, S., Kumar, A., Tucker, G., and Fu, J. (2020).
\newblock Offline reinforcement learning: Tutorial, review, and perspectives on
  open problems.
\newblock {\em arXiv preprint arXiv:2005.01643}.

\bibitem[Li et~al., 2021a]{li2021rmp2}
Li, A., Cheng, C.-A., Rana, M.~A., Xie, M., Van~Wyk, K., Ratliff, N., and
  Boots, B. (2021a).
\newblock {RMP${}^2$}: A structured composable policy class for robot learning.
\newblock {\em arXiv preprint arXiv:2103.05922}.

\bibitem[Li et~al., 2019]{li2019learn}
Li, X., Zhou, Y., Wu, T., Socher, R., and Xiong, C. (2019).
\newblock Learn to grow: A continual structure learning framework for
  overcoming catastrophic forgetting.
\newblock In {\em Proceedings of the 36th International Conference on Machine
  Learning (ICML-19)}, pages 3925--3934.

\bibitem[Li et~al., 2020a]{li2020learning}
Li, Y., He, H., Wu, J., Katabi, D., and Torralba, A. (2020a).
\newblock Learning compositional {K}oopman operators for model-based control.
\newblock In {\em 8th International Conference on Learning Representations
  (ICLR-20)}.

\bibitem[Li et~al., 2021b]{li2021solving}
Li, Y., Wu, Y., Xu, H., Wang, X., and Wu, Y. (2021b).
\newblock Solving compositional reinforcement learning problems via task
  reduction.
\newblock In {\em 9th International Conference on Learning Representations
  (ICLR-21)}.

\bibitem[Li et~al., 2020b]{li2020compositional}
Li, Y., Zhao, L., Church, K., and Elhoseiny, M. (2020b).
\newblock Compositional language continual learning.
\newblock In {\em 8th International Conference on Learning Representations
  (ICLR-20)}.

\bibitem[Li and Hoiem, 2017]{li2017learning}
Li, Z. and Hoiem, D. (2017).
\newblock Learning without forgetting.
\newblock {\em IEEE Transactions on Pattern Analysis and Machine Intelligence
  (TPAMI)}, 40(12):2935--2947.

\bibitem[Lillicrap et~al., 2016]{lillicrap2015continuous}
Lillicrap, T.~P., Hunt, J.~J., Pritzel, A., Heess, N., Erez, T., Tassa, Y.,
  Silver, D., and Wierstra, D. (2016).
\newblock Continuous control with deep reinforcement learning.
\newblock In {\em 4th International Conference on Learning Representations
  (ICLR-16)}.

\bibitem[Lin et~al., 2020]{lin2020rd2}
Lin, Z., Yang, D., Zhao, L., Qin, T., Yang, G., and Liu, T.-Y. (2020).
\newblock {RD{$^2$}}: Reward decomposition with representation disentanglement.
\newblock In {\em Advances in Neural Information Processing Systems 33
  (NeurIPS-20)}, pages 11298--11308.

\bibitem[Lin et~al., 2019]{lin2019distributional}
Lin, Z., Zhao, L., Yang, D., Qin, T., Liu, T.-Y., and Yang, G. (2019).
\newblock Distributional reward decomposition for reinforcement learning.
\newblock In {\em Advances in Neural Information Processing Systems 32
  (NeurIPS-19)}.

\bibitem[Liu et~al., 2020]{liu2020compositional}
Liu, Q., An, S., Lou, J.-G., Chen, B., Lin, Z., Gao, Y., Zhou, B., Zheng, N.,
  and Zhang, D. (2020).
\newblock Compositional generalization by learning analytical expressions.
\newblock In {\em Advances in Neural Information Processing Systems 33
  (NeurIPS-20)}, pages 11416--11427.

\bibitem[Loo et~al., 2021]{loo2021generalized}
Loo, N., Swaroop, S., and Turner, R.~E. (2021).
\newblock Generalized variational continual learning.
\newblock In {\em 9th International Conference on Learning Representations
  (ICLR-21)}.

\bibitem[Lopez-Paz and Ranzato, 2017]{lopez2017gradient}
Lopez-Paz, D. and Ranzato, M. (2017).
\newblock Gradient episodic memory for continual learning.
\newblock In {\em Advances in Neural Information Processing Systems 30
  (NIPS-17)}, pages 6467--6476.

\bibitem[Lu et~al., 2021]{lu2021resetfree}
Lu, K., Grover, A., Abbeel, P., and Mordatch, I. (2021).
\newblock Reset-free lifelong learning with skill-space planning.
\newblock In {\em 9th International Conference on Learning Representations
  (ICLR-21)}.

\bibitem[McCloskey and Cohen, 1989]{mccloskey1989catastrophic}
McCloskey, M. and Cohen, N.~J. (1989).
\newblock Catastrophic interference in connectionist networks: The sequential
  learning problem.
\newblock In {\em Psychology of Learning and Motivation}, volume~24, pages
  109--165. Elsevier.

\bibitem[Meyerson and Miikkulainen, 2018]{meyerson2018beyond}
Meyerson, E. and Miikkulainen, R. (2018).
\newblock Beyond shared hierarchies: Deep multitask learning through soft layer
  ordering.
\newblock In {\em 6th International Conference on Learning Representations
  (ICLR-18)}.

\bibitem[Meyerson and Miikkulainen, 2019]{meyerson2019modular}
Meyerson, E. and Miikkulainen, R. (2019).
\newblock Modular universal reparameterization: Deep multi-task learning across
  diverse domains.
\newblock In {\em Advances in Neural Information Processing Systems 32
  (NeurIPS-19)}.

\bibitem[Mirzadeh et~al., 2021]{mirzadeh2021linear}
Mirzadeh, S.~I., Farajtabar, M., Gorur, D., Pascanu, R., and Ghasemzadeh, H.
  (2021).
\newblock Linear mode connectivity in multitask and continual learning.
\newblock In {\em 9th International Conference on Learning Representations
  (ICLR-21)}.

\bibitem[Mirzadeh et~al., 2020]{mirzadeh2020understanding}
Mirzadeh, S.~I., Farajtabar, M., Pascanu, R., and Ghasemzadeh, H. (2020).
\newblock Understanding the role of training regimes in continual learning.
\newblock In {\em Advances in Neural Information Processing Systems 33
  (NeurIPS-20)}, pages 7308--7320.

\bibitem[Mittal et~al., 2020]{mittal2020learning}
Mittal, S., Lamb, A., Goyal, A., Voleti, V., Shanahan, M., Lajoie, G., Mozer,
  M., and Bengio, Y. (2020).
\newblock Learning to combine top-down and bottom-up signals in recurrent
  neural networks with attention over modules.
\newblock In {\em Proceedings of the 37th International Conference on Machine
  Learning (ICML-20)}, pages 6972--6986.

\bibitem[Mnih et~al., 2016]{mnih2016asynchronous}
Mnih, V., Badia, A.~P., Mirza, M., Graves, A., Lillicrap, T., Harley, T.,
  Silver, D., and Kavukcuoglu, K. (2016).
\newblock Asynchronous methods for deep reinforcement learning.
\newblock In {\em Proceedings of The 33rd International Conference on Machine
  Learning (ICML-16)}, pages 1928--1937.

\bibitem[Mnih et~al., 2015]{mnih2015human}
Mnih, V., Kavukcuoglu, K., Silver, D., Rusu, A.~A., Veness, J., Bellemare,
  M.~G., Graves, A., Riedmiller, M., Fidjeland, A.~K., Ostrovski, G., et~al.
  (2015).
\newblock Human-level control through deep reinforcement learning.
\newblock {\em Nature}, 518(7540):529--533.

\bibitem[Mu et~al., 2020]{mu2020refactoring}
Mu, T., Gu, J., Jia, Z., Tang, H., and Su, H. (2020).
\newblock Refactoring policy for compositional generalizability using
  self-supervised object proposals.
\newblock In {\em Advances in Neural Information Processing Systems 33
  (NeurIPS-20)}, pages 8883--8894.

\bibitem[Nagabandi et~al., 2019]{nagabandi2018deep}
Nagabandi, A., Finn, C., and Levine, S. (2019).
\newblock Deep online learning via meta-learning: Continual adaptation for
  model-based {RL}.
\newblock In {\em 7th International Conference on Learning Representations
  (ICLR-19)}.

\bibitem[Nangue~Tasse et~al., 2020]{nangue2020boolean}
Nangue~Tasse, G., James, S., and Rosman, B. (2020).
\newblock A {B}oolean task algebra for reinforcement learning.
\newblock In {\em Advances in Neural Information Processing Systems 33
  (NeurIPS-20)}, pages 9497--9507.

\bibitem[Nekoei et~al., 2021]{nekoei2021continuous}
Nekoei, H., Badrinaaraayanan, A., Courville, A., and Chandar, S. (2021).
\newblock Continuous coordination as a realistic scenario for lifelong
  learning.
\newblock In {\em Proceedings of the 38th International Conference on Machine
  Learning (ICML-21)}, pages 8016--8024.

\bibitem[Nguyen et~al., 2018]{nguyen2018variational}
Nguyen, C.~V., Li, Y., Bui, T.~D., and Turner, R.~E. (2018).
\newblock Variational continual learning.
\newblock In {\em 6th International Conference on Learning Representations
  (ICLR-18)}.

\bibitem[Nye et~al., 2020]{nye2020learning}
Nye, M., Solar-Lezama, A., Tenenbaum, J., and Lake, B.~M. (2020).
\newblock Learning compositional rules via neural program synthesis.
\newblock In {\em Advances in Neural Information Processing Systems 33
  (NeurIPS-20)}, pages 10832--10842.

\bibitem[Open{AI} et~al., 2020]{openai2020learning}
Open{AI}, Andrychowicz, M., Baker, B., Chociej, M., J{\'o}zefowicz, R., McGrew,
  B., Pachocki, J., Petron, A., Plappert, M., Powell, G., Ray, A., Schneider,
  J., Sidor, S., Tobin, J., Welinder, P., Weng, L., and Zaremba, W. (2020).
\newblock Learning dexterous in-hand manipulation.
\newblock {\em International Journal of Robotics Research}, 39(1):3--20.

\bibitem[Ostapenko et~al., 2021]{ostapenko2021continual}
Ostapenko, O., Rodriguez, P., Caccia, M., and Charlin, L. (2021).
\newblock Continual learning via local module composition.
\newblock In {\em Advances in Neural Information Processing Systems 34
  (NeurIPS-21)}, pages 30298--30312.

\bibitem[Pahuja et~al., 2019]{pahuja2019structure}
Pahuja, V., Fu, J., Chandar, S., and Pal, C. (2019).
\newblock Structure learning for neural module networks.
\newblock In {\em Proceedings of the Beyond Vision and LANguage: inTEgrating
  Real-world kNowledge (LANTERN)}, pages 1--10. Association for Computational
  Linguistics.

\bibitem[Pan et~al., 2020]{pan2020continual}
Pan, P., Swaroop, S., Immer, A., Eschenhagen, R., Turner, R., and Khan, M.~E.
  (2020).
\newblock Continual deep learning by functional regularisation of memorable
  past.
\newblock In {\em Advances in Neural Information Processing Systems 33
  (NeurIPS-20)}, pages 4453--4464.

\bibitem[Pathak et~al., 2019]{pathak2019learning}
Pathak, D., Lu, C., Darrell, T., Isola, P., and Efros, A.~A. (2019).
\newblock Learning to control self-assembling morphologies: A study of
  generalization via modularity.
\newblock In {\em Advances in Neural Information Processing Systems 32
  (NeurIPS-19)}.

\bibitem[Peng et~al., 2019]{peng2019mcp}
Peng, X.~B., Chang, M., Zhang, G., Abbeel, P., and Levine, S. (2019).
\newblock {MCP}: Learning composable hierarchical control with multiplicative
  compositional policies.
\newblock In {\em Advances in Neural Information Processing Systems 32
  (NeurIPS-19)}.

\bibitem[Pentina and Lampert, 2015]{pentina2015lifelong}
Pentina, A. and Lampert, C.~H. (2015).
\newblock Lifelong learning with non-i.i.d.~tasks.
\newblock In {\em Advances in Neural Information Processing Systems 28
  (NIPS-15)}.

\bibitem[Pham et~al., 2021a]{pham2021dualnet}
Pham, Q., Liu, C., and Hoi, S. (2021a).
\newblock {DualNet}: Continual learning, fast and slow.
\newblock In {\em Advances in Neural Information Processing Systems 34
  (NeurIPS-21)}, pages 16131--16144.

\bibitem[Pham et~al., 2021b]{pham2021contextual}
Pham, Q., Liu, C., Sahoo, D., and Hoi, S. (2021b).
\newblock Contextual transformation networks for online continual learning.
\newblock In {\em 9th International Conference on Learning Representations
  (ICLR-21)}.

\bibitem[Piaget, 1976]{piaget1976piaget}
Piaget, J. (1976).
\newblock Piaget's theory.
\newblock In Inhelder, B., Chipman, H.~H., and Zwingmann, C., editors, {\em
  Piaget and His School: A Reader in Developmental Psychology}, pages 11--23.
  Springer, Berlin, Heidelberg.

\bibitem[Pierrot et~al., 2019]{pierrot2019learning}
Pierrot, T., Ligner, G., Reed, S.~E., Sigaud, O., Perrin, N., Laterre, A., Kas,
  D., Beguir, K., and de~Freitas, N. (2019).
\newblock Learning compositional neural programs with recursive tree search and
  planning.
\newblock In {\em Advances in Neural Information Processing Systems 32
  (NeurIPS-19)}.

\bibitem[Qin et~al., 2021]{qin2021bns}
Qin, Q., Hu, W., Peng, H., Zhao, D., and Liu, B. (2021).
\newblock {BNS}: Building network structures dynamically for continual
  learning.
\newblock In {\em Advances in Neural Information Processing Systems 34
  (NeurIPS-21)}, pages 20608--20620.

\bibitem[Raghavan and Balaprakash, 2021]{raghavan2021formalizing}
Raghavan, K. and Balaprakash, P. (2021).
\newblock Formalizing the generalization-forgetting trade-off in continual
  learning.
\newblock In {\em Advances in Neural Information Processing Systems 34
  (NeurIPS-21)}.

\bibitem[Rahaman et~al., 2021]{rahaman2021spatially}
Rahaman, N., Goyal, A., Gondal, M.~W., Wuthrich, M., Bauer, S., Sharma, Y.,
  Bengio, Y., and Sch{\"o}lkopf, B. (2021).
\newblock Spatially structured recurrent modules.
\newblock In {\em 9th International Conference on Learning Representations
  (ICLR-21)}.

\bibitem[Rajasegaran et~al., 2019]{rajasegaran2019random}
Rajasegaran, J., Hayat, M., Khan, S., Khan, F.~S., and Shao, L. (2019).
\newblock Random path selection for incremental learning.
\newblock In {\em Advances in Neural Information Processing Systems 32
  (NeurIPS-19)}, pages 12669--12679.

\bibitem[Rajeswaran et~al., 2017]{rajeswaran2017towards}
Rajeswaran, A., Lowrey, K., Todorov, E.~V., and Kakade, S.~M. (2017).
\newblock Towards generalization and simplicity in continuous control.
\newblock In {\em Advances in Neural Information Processing Systems 30
  (NIPS-17)}, pages 6550--6561.

\bibitem[Ramasesh et~al., 2021]{ramasesh2021anatomy}
Ramasesh, V.~V., Dyer, E., and Raghu, M. (2021).
\newblock Anatomy of catastrophic forgetting: Hidden representations and task
  semantics.
\newblock In {\em 9th International Conference on Learning Representations
  (ICLR-21)}.

\bibitem[Rao et~al., 2019]{rao2019continual}
Rao, D., Visin, F., Rusu, A., Pascanu, R., Teh, Y.~W., and Hadsell, R. (2019).
\newblock Continual unsupervised representation learning.
\newblock In {\em Advances in Neural Information Processing Systems 32
  (NeurIPS-19)}, pages 7647--7657.

\bibitem[Reed and de~Freitas, 2016]{reed2016neural}
Reed, S. and de~Freitas, N. (2016).
\newblock Neural programmers-interpreters.
\newblock In {\em 4th International Conference on Learning Representations
  (ICLR-16)}.

\bibitem[Ren et~al., 2021]{ren2021wandering}
Ren, M., Iuzzolino, M.~L., Mozer, M.~C., and Zemel, R. (2021).
\newblock Wandering within a world: Online contextualized few-shot learning.
\newblock In {\em 9th International Conference on Learning Representations
  (ICLR-21)}.

\bibitem[Ren et~al., 2020]{ren2020compositional}
Ren, Y., Guo, S., Labeau, M., Cohen, S.~B., and Kirby, S. (2020).
\newblock Compositional languages emerge in a neural iterated learning model.
\newblock In {\em 8th International Conference on Learning Representations
  (ICLR-20)}.

\bibitem[Riemer et~al., 2019]{riemer2018learning}
Riemer, M., Cases, I., Ajemian, R., Liu, M., Rish, I., Tu, Y., , and Tesauro,
  G. (2019).
\newblock Learning to learn without forgetting by maximizing transfer and
  minimizing interference.
\newblock In {\em 7th International Conference on Learning Representations
  (ICLR-19)}.

\bibitem[Ritter et~al., 2018]{ritter2018online}
Ritter, H., Botev, A., and Barber, D. (2018).
\newblock Online structured {L}aplace approximations for overcoming
  catastrophic forgetting.
\newblock In {\em Advances in Neural Information Processing Systems 31
  (NeurIPS-18)}, pages 3738--3748.

\bibitem[Rolnick et~al., 2019]{rolnick2019experience}
Rolnick, D., Ahuja, A., Schwarz, J., Lillicrap, T., and Wayne, G. (2019).
\newblock Experience replay for continual learning.
\newblock In {\em Advances in Neural Information Processing Systems 32
  (NeurIPS-19)}.

\bibitem[Rosenbaum et~al., 2019]{rosenbaum2019routing}
Rosenbaum, C., Cases, I., Riemer, M., and Klinger, T. (2019).
\newblock Routing networks and the challenges of modular and compositional
  computation.
\newblock {\em arXiv preprint arXiv:1904.12774}.

\bibitem[Rosenbaum et~al., 2018]{rosenbaum2018routing}
Rosenbaum, C., Klinger, T., and Riemer, M. (2018).
\newblock Routing networks: Adaptive selection of non-linear functions for
  multi-task learning.
\newblock In {\em 6th International Conference on Learning Representations
  (ICLR-18)}.

\bibitem[Rostami, 2021]{rostami2021lifelong}
Rostami, M. (2021).
\newblock Lifelong domain adaptation via consolidated internal distribution.
\newblock In {\em Advances in Neural Information Processing Systems 34
  (NeurIPS-21)}, pages 11172--11183.

\bibitem[Rostami et~al., 2020]{rostami2020using}
Rostami, M., Isele, D., and Eaton, E. (2020).
\newblock Using task descriptions in lifelong machine learning for improved
  performance and zero-shot transfer.
\newblock {\em Journal of Artificial Intelligence Research (JAIR)},
  67:673--704.

\bibitem[Ruis et~al., 2021]{ruis2021independent}
Ruis, F., Burghouts, G., and Bucur, D. (2021).
\newblock Independent prototype propagation for zero-shot compositionality.
\newblock In {\em Advances in Neural Information Processing Systems 34
  (NeurIPS-21)}, pages 10641--10653.

\bibitem[Rusu et~al., 2016]{rusu2016progressive}
Rusu, A.~A., Rabinowitz, N.~C., Desjardins, G., Soyer, H., Kirkpatrick, J.,
  Kavukcuoglu, K., Pascanu, R., and Hadsell, R. (2016).
\newblock Progressive neural networks.
\newblock {\em arXiv preprint arXiv:1606.04671}.

\bibitem[Ruvolo and Eaton, 2013]{ruvolo2013ella}
Ruvolo, P. and Eaton, E. (2013).
\newblock {ELLA}: An efficient lifelong learning algorithm.
\newblock In {\em Proceedings of the 30th International Conference on Machine
  Learning (ICML-13)}, pages 507--515.

\bibitem[Sacerdoti, 1974]{sacerdoti1974planning}
Sacerdoti, E.~D. (1974).
\newblock Planning in a hierarchy of abstraction spaces.
\newblock {\em Artificial Intelligence}, 5(2):115--135.

\bibitem[Saha et~al., 2021]{saha2021gradient}
Saha, G., Garg, I., and Roy, K. (2021).
\newblock Gradient projection memory for continual learning.
\newblock In {\em 9th International Conference on Learning Representations
  (ICLR-21)}.

\bibitem[Saqur and Narasimhan, 2020]{saqur2020multimodal}
Saqur, R. and Narasimhan, K. (2020).
\newblock Multimodal graph networks for compositional generalization in visual
  question answering.
\newblock In {\em Advances in Neural Information Processing Systems 33
  (NeurIPS-20)}, pages 3070--3081.

\bibitem[Schulman et~al., 2015]{schulman2015trust}
Schulman, J., Levine, S., Abbeel, P., Jordan, M., and Moritz, P. (2015).
\newblock Trust region policy optimization.
\newblock In {\em Proceedings of the 32nd International Conference on Machine
  Learning (ICML-15)}, pages 1889--1897.

\bibitem[Schulman et~al., 2017]{schulman2017proximal}
Schulman, J., Wolski, F., Dhariwal, P., Radford, A., and Klimov, O. (2017).
\newblock Proximal policy optimization algorithms.
\newblock {\em arXiv preprint arXiv:1707.06347}.

\bibitem[Schwarz et~al., 2018]{schwarz2018progress}
Schwarz, J., Czarnecki, W., Luketina, J., Grabska-Barwinska, A., Teh, Y.~W.,
  Pascanu, R., and Hadsell, R. (2018).
\newblock Progress {\&} compress: A scalable framework for continual learning.
\newblock In {\em Proceedings of the 35th International Conference on Machine
  Learning (ICML-18)}, pages 4528--4537.

\bibitem[Serr{\`a} et~al., 2018]{serra2018overcoming}
Serr{\`a}, J., Sur{\'i}s, D., Miron, M., and Karatzoglou, A. (2018).
\newblock Overcoming catastrophic forgetting with hard attention to the task.
\newblock In {\em Proceedings of the 35th International Conference on Machine
  Learning (ICML-18)}, pages 4548--4557.

\bibitem[Shazeer et~al., 2017]{shazeer2017outrageously}
Shazeer, N., Mirhoseini, A., Maziarz, K., Davis, A., Le, Q., Hinton, G., and
  Dean, J. (2017).
\newblock Outrageously large neural networks: The sparsely-gated
  mixture-of-experts layer.
\newblock In {\em 5th International Conference on Learning Representations
  (ICLR-17)}.

\bibitem[Shin et~al., 2017]{shin2017continual}
Shin, H., Lee, J.~K., Kim, J., and Kim, J. (2017).
\newblock Continual learning with deep generative replay.
\newblock In {\em Advances in Neural Information Processing Systems 30
  (NIPS-17)}, pages 2990--2999.

\bibitem[Silver et~al., 2016]{silver2016mastering}
Silver, D., Huang, A., Maddison, C.~J., Guez, A., Sifre, L., van Den~Driessche,
  G., Schrittwieser, J., Antonoglou, I., Panneershelvam, V., Lanctot, M.,
  et~al. (2016).
\newblock Mastering the game of {Go} with deep neural networks and tree search.
\newblock {\em Nature}, 529(7587):484--489.

\bibitem[Silver et~al., 2017]{silver2017mastering}
Silver, D., Schrittwieser, J., Simonyan, K., Antonoglou, I., Huang, A., Guez,
  A., Hubert, T., Baker, L., Lai, M., Bolton, A., et~al. (2017).
\newblock Mastering the game of {Go} without human knowledge.
\newblock {\em Nature}, 550(7676):354--359.

\bibitem[Singh et~al., 2020]{singh2020calibrating}
Singh, P., Verma, V.~K., Mazumder, P., Carin, L., and Rai, P. (2020).
\newblock Calibrating {CNN}s for lifelong learning.
\newblock In {\em Advances in Neural Information Processing Systems 33
  (NeurIPS-20)}, pages 15579--15590.

\bibitem[Sinha et~al., 2020]{sinha2020evaluating}
Sinha, K., Sodhani, S., Pineau, J., and Hamilton, W.~L. (2020).
\newblock Evaluating logical generalization in graph neural networks.
\newblock {\em arXiv preprint arXiv:2003.06560}.

\bibitem[Skorokhodov and Elhoseiny, 2021]{skorokhodov2021class}
Skorokhodov, I. and Elhoseiny, M. (2021).
\newblock Class normalization for (continual)? generalized zero-shot learning.
\newblock In {\em 9th International Conference on Learning Representations
  (ICLR-21)}.

\bibitem[Sprechmann et~al., 2018]{sprechmann2018memorybased}
Sprechmann, P., Jayakumar, S., Rae, J., Pritzel, A., Badia, A.~P., Uria, B.,
  Vinyals, O., Hassabis, D., Pascanu, R., and Blundell, C. (2018).
\newblock Memory-based parameter adaptation.
\newblock In {\em 6th International Conference on Learning Representations
  (ICLR-18)}.

\bibitem[Sun et~al., 2020]{sun2020lamol}
Sun, F.-K., Ho, C.-H., and Lee, H.-Y. (2020).
\newblock {LAMOL}: {LA}nguage {MO}deling for lifelong language learning.
\newblock In {\em 8th International Conference on Learning Representations
  (ICLR-20)}.

\bibitem[Sutton and Barto, 2018]{sutton2018reinforcement}
Sutton, R.~S. and Barto, A.~G. (2018).
\newblock {\em Reinforcement learning: An introduction}.
\newblock MIT press.

\bibitem[Sutton et~al., 1999a]{sutton2000policy}
Sutton, R.~S., McAllester, D., Singh, S., and Mansour, Y. (1999a).
\newblock Policy gradient methods for reinforcement learning with function
  approximation.
\newblock In {\em Advances in Neural Information Processing Systems 12
  (NIPS-00)}.

\bibitem[Sutton et~al., 2011]{sutton2011horde}
Sutton, R.~S., Modayil, J., Delp, M., Degris, T., Pilarski, P.~M., White, A.,
  and Precup, D. (2011).
\newblock Horde: A scalable real-time architecture for learning knowledge from
  unsupervised sensorimotor interaction.
\newblock In {\em Tenth International Conference on Autonomous Agents and
  Multiagent Systems (AAMAS-11)}, pages 761--768.

\bibitem[Sutton et~al., 1999b]{sutton1999between}
Sutton, R.~S., Precup, D., and Singh, S. (1999b).
\newblock Between {MDP}s and semi-{MDP}s: A framework for temporal abstraction
  in reinforcement learning.
\newblock {\em Artificial Intelligence}, 112(1--2):181--211.

\bibitem[Sylvain et~al., 2020]{sylvain2020locality}
Sylvain, T., Petrini, L., and Hjelm, D. (2020).
\newblock Locality and compositionality in zero-shot learning.
\newblock In {\em 8th International Conference on Learning Representations
  (ICLR-20)}.

\bibitem[Tang and Matteson, 2021]{tang2021graphbased}
Tang, B. and Matteson, D.~S. (2021).
\newblock Graph-based continual learning.
\newblock In {\em 9th International Conference on Learning Representations
  (ICLR-21)}.

\bibitem[Tessler et~al., 2017]{tessler2017deep}
Tessler, C., Givony, S., Zahavy, T., Mankowitz, D., and Mannor, S. (2017).
\newblock A deep hierarchical approach to lifelong learning in {M}inecraft.
\newblock In {\em Proceedings of the Thirty-First AAAI Conference on Artificial
  Intelligence (AAAI-17)}.

\bibitem[Thrun, 1998]{thrun1998lifelong}
Thrun, S. (1998).
\newblock Lifelong learning algorithms.
\newblock In {\em Learning to learn}, pages 181--209. Springer.

\bibitem[Titsias et~al., 2020]{titsias2020functional}
Titsias, M.~K., Schwarz, J., de~G.~Matthews, A.~G., Pascanu, R., and Teh, Y.~W.
  (2020).
\newblock Functional regularisation for continual learning with {G}aussian
  processes.
\newblock In {\em 8th International Conference on Learning Representations
  (ICLR-20)}.

\bibitem[Todorov, 2009]{todorov2009compositionality}
Todorov, E. (2009).
\newblock Compositionality of optimal control laws.
\newblock {\em Advances in Neural Information Processing Systems 22 (NIPS-09)},
  pages 1856--1864.

\bibitem[Todorov et~al., 2012]{todorov2012mujoco}
Todorov, E., Erez, T., and Tassa, Y. (2012).
\newblock {MuJoCo}: A physics engine for model-based control.
\newblock In {\em Proceedings of the 2012 IEEE/RSJ International Conference on
  Intelligent Robots and Systems (IROS-12)}, pages 5026--5033.

\bibitem[Tunyasuvunakool et~al., 2020]{tunyasuvunakool2020dmcontrol}
Tunyasuvunakool, S., Muldal, A., Doron, Y., Liu, S., Bohez, S., Merel, J.,
  Erez, T., Lillicrap, T., Heess, N., and Tassa, Y. (2020).
\newblock dm\_control: Software and tasks for continuous control.
\newblock {\em Software Impacts}, 6.

\bibitem[Valkov et~al., 2018]{valkov2018houdini}
Valkov, L., Chaudhari, D., Srivastava, A., Sutton, C., and Chaudhuri, S.
  (2018).
\newblock Houdini: Lifelong learning as program synthesis.
\newblock In {\em Advances in Neural Information Processing Systems 31
  (NeurIPS-18)}, pages 8687--8698.

\bibitem[van~de Ven et~al., 2020]{van2020brain}
van~de Ven, G.~M., Siegelmann, H.~T., and Tolias, A.~S. (2020).
\newblock Brain-inspired replay for continual learning with artificial neural
  networks.
\newblock {\em Nature Communications}, 11(1).

\bibitem[van~de Ven and Tolias, 2019]{van2019three}
van~de Ven, G.~M. and Tolias, A.~S. (2019).
\newblock Three scenarios for continual learning.
\newblock {\em arXiv preprint arXiv:1904.07734}.

\bibitem[van~der Vaart, 2000]{van2000asymptotic}
van~der Vaart, A. (2000).
\newblock {\em Asymptotic statistics}, volume~3.
\newblock Cambridge University Press.

\bibitem[van Niekerk et~al., 2019]{van2019composing}
van Niekerk, B., James, S., Earle, A., and Rosman, B. (2019).
\newblock Composing value functions in reinforcement learning.
\newblock In {\em Proceedings of the 36th International Conference on Machine
  Learning (ICML-19)}, pages 6401--6409.

\bibitem[van Seijen et~al., 2017]{vanseijen2017hybrid}
van Seijen, H., Fatemi, M., Romoff, J., Laroche, R., Barnes, T., and Tsang, J.
  (2017).
\newblock Hybrid reward architecture for reinforcement learning.
\newblock In {\em Advances in Neural Information Processing Systems 30
  (NIPS-17)}.

\bibitem[Varshney et~al., 2021]{varshney2021camgan}
Varshney, S., Verma, V.~K., Srijith, P.~K., Carin, L., and Rai, P. (2021).
\newblock {CAM-GAN}: Continual adaptation modules for generative adversarial
  networks.
\newblock In {\em Advances in Neural Information Processing Systems 34
  (NeurIPS-21)}, pages 15175--15187.

\bibitem[Veniat et~al., 2021]{veniat2021efficient}
Veniat, T., Denoyer, L., and Ranzato, M. (2021).
\newblock Efficient continual learning with modular networks and task-driven
  priors.
\newblock In {\em 9th International Conference on Learning Representations
  (ICLR-21)}.

\bibitem[Vezhnevets et~al., 2017]{vezhnevets2017feudal}
Vezhnevets, A.~S., Osindero, S., Schaul, T., Heess, N., Jaderberg, M., Silver,
  D., and Kavukcuoglu, K. (2017).
\newblock Fe{U}dal networks for hierarchical reinforcement learning.
\newblock In {\em Proceedings of the 34th International Conference on Machine
  Learning (ICML-17)}, pages 3540--3549.

\bibitem[Vinyals et~al., 2017]{vinyals2017starcraft}
Vinyals, O., Ewalds, T., Bartunov, S., Georgiev, P., Vezhnevets, A.~S., Yeo,
  M., Makhzani, A., K{\"{u}}ttler, H., Agapiou, J.~P., Schrittwieser, J., Quan,
  J., Gaffney, S., Petersen, S., Simonyan, K., Schaul, T., van Hasselt, H.,
  Silver, D., Lillicrap, T.~P., Calderone, K., Keet, P., Brunasso, A.,
  Lawrence, D., Ekermo, A., Repp, J., and Tsing, R. (2017).
\newblock Star{C}raft {II}: A new challenge for reinforcement learning.
\newblock {\em arXiv preprint arXiv:1708.04782}.

\bibitem[von Oswald et~al., 2020]{oswald2020continual}
von Oswald, J., Henning, C., Sacramento, J., and Grewe, B.~F. (2020).
\newblock Continual learning with hypernetworks.
\newblock In {\em 8th International Conference on Learning Representations
  (ICLR-20)}.

\bibitem[von Oswald et~al., 2021]{oswald2021learning}
von Oswald, J., Zhao, D., Kobayashi, S., Schug, S., Caccia, M., Zucchet, N.,
  and Sacramento, J. (2021).
\newblock Learning where to learn: Gradient sparsity in meta and continual
  learning.
\newblock In {\em Advances in Neural Information Processing Systems 34
  (NeurIPS-21)}, pages 5250--5263.

\bibitem[Wang et~al., 2021]{wang2021afec}
Wang, L., Zhang, M., Jia, Z., Li, Q., Bao, C., Ma, K., Zhu, J., and Zhong, Y.
  (2021).
\newblock {AFEC}: Active forgetting of negative transfer in continual learning.
\newblock In {\em Advances in Neural Information Processing Systems 34
  (NeurIPS-21)}, pages 22379--22391.

\bibitem[Watkins, 1989]{watkins1989learning}
Watkins, C. J. C.~H. (1989).
\newblock {\em Learning from delayed rewards}.
\newblock Ph.D.~thesis, King's College, Cambridge, United Kingdom.

\bibitem[Welinder et~al., 2010]{welinder2010caltech}
Welinder, P., Branson, S., Mita, T., Wah, C., Schroff, F., Belongie, S., and
  Perona, P. (2010).
\newblock Caltech-{UCSD} birds 200.
\newblock Technical report, California Institute of Technology.

\bibitem[Williams, 1992]{williams1992simple}
Williams, R.~J. (1992).
\newblock Simple statistical gradient-following algorithms for connectionist
  reinforcement learning.
\newblock {\em Machine Learning}, 8(3):229--256.

\bibitem[Wo{\l}czyk et~al., 2021]{wolczyk2021continual}
Wo{\l}czyk, M., Zaj{\k{a}}c, M., Pascanu, R., Kuci{\'n}ski, {\L}., and
  Mi{\l}o{\'s}, P. (2021).
\newblock Continual {W}orld: A robotic benchmark for continual reinforcement
  learning.
\newblock In {\em Advances in Neural Information Processing Systems 34
  (NeurIPS-21)}, pages 28496--28510.

\bibitem[Wu et~al., 2021]{wu2021improving}
Wu, M., Goodman, N., and Ermon, S. (2021).
\newblock Improving compositionality of neural networks by decoding
  representations to inputs.
\newblock In {\em Advances in Neural Information Processing Systems 34
  (NeurIPS-21)}, pages 26689--26700.

\bibitem[Xiao et~al., 2017]{xiao2017fashion}
Xiao, H., Rasul, K., and Vollgraf, R. (2017).
\newblock Fashion-{MNIST}: A novel image dataset for benchmarking machine
  learning algorithms.
\newblock {\em arXiv preprint arXiv:1708.07747}.

\bibitem[Xie et~al., 2021]{xie2020deep}
Xie, A., Harrison, J., and Finn, C. (2021).
\newblock Deep reinforcement learning amidst continual structured
  non-stationarity.
\newblock In {\em Proceedings of the 38th International Conference on Machine
  Learning (ICML-21)}, pages 11393--11403.

\bibitem[Xu et~al., 2018]{xu2018neural}
Xu, D., Nair, S., Zhu, Y., Gao, J., Garg, A., Fei-Fei, L., and Savarese, S.
  (2018).
\newblock Neural task programming: Learning to generalize across hierarchical
  tasks.
\newblock In {\em Proceedings of the 2018 IEEE International Conference on
  Robotics and Automation (ICRA-18)}, pages 3795--3802.

\bibitem[Xu et~al., 2020]{xu2020continual}
Xu, K., Verma, S., Finn, C., and Levine, S. (2020).
\newblock Continual learning of control primitives : Skill discovery via
  reset-games.
\newblock In {\em Advances in Neural Information Processing Systems 33
  (NeurIPS-20)}, pages 4999--5010.

\bibitem[Yang et~al., 2020]{yang2020multi}
Yang, R., Xu, H., WU, Y., and Wang, X. (2020).
\newblock Multi-task reinforcement learning with soft modularization.
\newblock In {\em Advances in Neural Information Processing Systems 33
  (NeurIPS-20)}, pages 4767--4777.

\bibitem[Yin et~al., 2021]{yin2021mitigating}
Yin, H., Yang, P., and Li, P. (2021).
\newblock Mitigating forgetting in online continual learning with neuron
  calibration.
\newblock In {\em Advances in Neural Information Processing Systems 34
  (NeurIPS-21)}, pages 10260--10272.

\bibitem[Yoon et~al., 2021]{yoon2021federated}
Yoon, J., Jeong, W., Lee, G., Yang, E., and Hwang, S.~J. (2021).
\newblock Federated continual learning with weighted inter-client transfer.
\newblock In {\em Proceedings of the 38th International Conference on Machine
  Learning (ICML-21)}, pages 12073--12086.

\bibitem[Yoon et~al., 2020]{yoon2020scalable}
Yoon, J., Kim, S., Yang, E., and Hwang, S.~J. (2020).
\newblock Scalable and order-robust continual learning with additive parameter
  decomposition.
\newblock In {\em 8th International Conference on Learning Representations
  (ICLR-20)}.

\bibitem[Yoon et~al., 2018]{yoon2018lifelong}
Yoon, J., Lee, J., Yang, E., and Hwang, S.~J. (2018).
\newblock Lifelong learning with dynamically expandable networks.
\newblock In {\em 6th International Conference on Learning Representations
  (ICLR-18)}.

\bibitem[Yu et~al., 2019]{yu2019meta}
Yu, T., Quillen, D., He, Z., Julian, R., Hausman, K., Finn, C., and Levine, S.
  (2019).
\newblock Meta-{W}orld: A benchmark and evaluation for multi-task and meta
  reinforcement learning.
\newblock In {\em Proceedings of the 3rd Conference on Robot Learning
  (CoRL-19)}.

\bibitem[Zaremba et~al., 2016]{zaremba2016learning}
Zaremba, W., Mikolov, T., Joulin, A., and Fergus, R. (2016).
\newblock Learning simple algorithms from examples.
\newblock In {\em Proceedings of the 33rd International Conference on Machine
  Learning (ICML-16)}, pages 421--429.

\bibitem[Zenke et~al., 2017]{zenke2017continual}
Zenke, F., Poole, B., and Ganguli, S. (2017).
\newblock Continual learning through synaptic intelligence.
\newblock In {\em Proceedings of the 34th International Conference on Machine
  Learning (ICML-17)}, pages 3987--3995.

\bibitem[Zeno et~al., 2018]{zeno2018task}
Zeno, C., Golan, I., Hoffer, E., and Soudry, D. (2018).
\newblock Task agnostic continual learning using online variational {B}ayes.
\newblock {\em arXiv preprint arXiv:1803.10123}.

\bibitem[Zeno et~al., 2021]{zeno2021task}
Zeno, C., Golan, I., Hoffer, E., and Soudry, D. (2021).
\newblock Task-agnostic continual learning using online variational {B}ayes
  with fixed-point updates.
\newblock {\em Neural Computation}, 33(11):3139--3177.

\bibitem[Zhang et~al., 2021]{zhang2021variational}
Zhang, Q., Fang, J., Meng, Z., Liang, S., and Yilmaz, E. (2021).
\newblock Variational continual {B}ayesian meta-learning.
\newblock In {\em Advances in Neural Information Processing Systems 34
  (NeurIPS-21)}, pages 24556--24568.

\bibitem[Zhao et~al., 2017]{zhao2017tensor}
Zhao, C., Hospedales, T.~M., Stulp, F., and Sigaud, O. (2017).
\newblock Tensor based knowledge transfer across skill categories for robot
  control.
\newblock In {\em Proceedings of the Twenty-Sixth International Joint
  Conference on Artificial Intelligence (IJCAI-17)}, pages 3462--3468.

\bibitem[Zhu et~al., 2020]{robosuite2020}
Zhu, Y., Wong, J., Mandlekar, A., and Mart{\'i}n-Mart{\'i}n, R. (2020).
\newblock robosuite: A modular simulation framework and benchmark for robot
  learning.
\newblock In {\em arXiv preprint arXiv:2009.12293}.

\end{thebibliography}

\end{bibliof}

\end{document}